\documentclass[letterpaper]{article} 
\usepackage{aaai23}  
\usepackage{times}  
\usepackage{helvet}  
\usepackage{courier}  
\usepackage[hyphens]{url}  
\usepackage{graphicx} 
\usepackage{natbib}  
\usepackage{caption} 
\usepackage{algorithm}
\usepackage{algorithmic}

\usepackage{newfloat}
\usepackage{listings}
\DeclareCaptionStyle{ruled}{labelfont=normalfont,labelsep=colon,strut=off} 
\floatstyle{ruled}
\newfloat{listing}{tb}{lst}{}
\floatname{listing}{Listing}
\usepackage{subfigure}

\newcommand{\tabref}[1]{Tab.~\ref{#1}}

\newcommand{\figref}[1]{Fig.~\ref{#1}}

\newcommand{\eg}{\textit{e.g.}}

\usepackage{amsmath}
\usepackage{amssymb}
\usepackage{multirow}

\title{Deep Digging into the Generalization of Self-Supervised Monocular Depth Estimation}
\author{
  Jinwoo Bae\textsuperscript{\rm 1}, Sungho Moon\textsuperscript{\rm 1}, and Sunghoon Im\textsuperscript{\rm 1}\\
}

\affiliations {
    \textsuperscript{\rm 1} Department of Electrical Engineering and Computer Science, DGIST, Daegu, Korea\\
    \{sjg02122, byeol3325, sunghoonim\}@dgist.ac.kr
}

\begin{document}

\maketitle

\begin{abstract}
Self-supervised monocular depth estimation has been widely studied recently. Most of the work has focused on improving performance on benchmark datasets, such as KITTI, but has offered a few experiments on generalization performance.
In this paper, we investigate the backbone networks (\eg~CNNs, Transformers, and CNN-Transformer hybrid models) toward the generalization of monocular depth estimation.
We first evaluate state-of-the-art models on diverse public datasets, which have never been seen during the network training.
Next, we investigate the effects of texture-biased and shape-biased representations using the various texture-shifted datasets that we generated.
We observe that Transformers exhibit a strong shape bias and CNNs do a strong texture-bias.
We also find that shape-biased models show better generalization performance for monocular depth estimation compared to texture-biased models. 
Based on these observations, we newly design a CNN-Transformer hybrid network with a multi-level adaptive feature fusion module, called MonoFormer.
The design intuition behind MonoFormer is to increase shape bias by employing Transformers while compensating for the weak locality bias of Transformers by adaptively fusing multi-level representations.
Extensive experiments show that the proposed method achieves state-of-the-art performance with various public datasets. Our method also shows the best generalization ability among the competitive methods.
\end{abstract}

\section{Introduction}

How do humans efficiently extract and recognize essential information from complex scenes? The biological vision system treats the object's shape as the single most crucial vision cue, compared with other cues like texture or color \cite{landau1988importance}. This enables humans, even small children, to easily recognize an object from a line drawing or a silhouette image.
It is widely known that convolutional neural networks (CNNs) are designed with inspiration from the biological neural networks in living organisms~\cite{hubel1959receptive,fukushima1988neocognitron,kriegeskorte2015deep}. 
CNNs extract the simple patterns (\textit{e.g.} edges) and then build complex patterns by successively composing early neural responses.
However, in contrast to human visual representation, recent researches \cite{geirhos2018imagenet,morrison2021exploring,tuli2021convolutional} have revealed that CNNs are strongly biased towards recognizing textures rather than shapes.
CNN-based models rationally classify labels even in images with disrupted shape structures \cite{gatys2017texture,brendel2019approximating}. On the other hand, CNN models fail to predict labels correctly in a texture-removed image whose shape is well-preserved \cite{ballester2016performance}.

Then, how does this observation affect the monocular depth estimation task? Over the past decade, monocular depth estimation has made significant progress using CNNs \cite{xiong2021self,yin2018geonet,zhou2021r,godard2019digging,guizilini20203d,casser2019depth}. 
These works show the remarkable performance on the KITTI datasets \cite{Geiger2013IJRR} even with the model trained in a self-supervised manner.
However, the experiments have been conducted on only a few driving scenes, mostly KITTI datasets, so the generality of these methods has not been closely studied.
In this paper, we study the generalization performance of the state-of-the-art methods and investigate how texture-biased representation from CNNs affects monocular depth estimation.
We evaluate state-of-the-art models trained on KITTI using six public depth datasets (SUN3D, RGBD, MVS, Scenes11, ETH3D, and Oxford Robotcar).
We also conduct experiments on three different texture-shifted datasets including texture-smoothed (Watercolor), textureless (Pencil-sketch), and texture-transferred (Style-transfer) images.
Through these extensive experiments, we determine that texture-biased models are vulnerable to generality in monocular depth estimation.

Recently, Transformers \cite{dosovitskiy2020image} have received a surge of attention for their outstanding performance in the field of computer vision \cite{carion2020end}, despite the lack of a spatial locality bias. 
Moreover, several works \cite{zhang2021delving,morrison2021exploring,park2022vision} show that Transformers have a strong shape bias, unlike CNNs. 
We also investigate the Transformers, similar to the experiments conducted for CNNs, and observe that shape bias is key to generalize depth estimation.
Thus, we propose a CNN-Transformer hybrid network, called MonoFormer, which are highly complementary to each other. 
The design intuition behind MonoFormer is to take the strong shape bias of Transformers and the spatial locality bias of low-level Transformers features projected from CNN features. 
To do so, we design a layer-wise Attention Connection Module (ACM) and a Feature Fusion Decoder (FFD). The ACM measures the importance of shape bias representation and the local details, and then the FFD adaptively fuses them for depth prediction.
The detailed ablation studies show that the shape-biased features are mostly extracted from high-level Transformers and the local details are captured at low layers. 

To verify the generality, we evaluate our KITTI-trained model on the six out-of-distribution datasets.
These experiments show MonoFormer achieves performance improvement of up to more than 30$\%$ over other CNN-based state-of-the-art models \cite{godard2019digging,zhou2021r,guizilini20203d}, 7$\%$ over a Transformer-based model \cite{dosovitskiy2020image}, and 15$\%$ over a conventional hybrid model \cite{yang2021transformer}. Our model shows strong robustness and generality regardless of the testing distributions.
By investigating the network structures, we observe that the CNNs mostly learn texture-based representation while Transformers nearly learn shape-based representation. 
We also reveal that the shape-biased models achieve superior generalization ability compared with texture-biased models on out-of-distribution training datasets. 
Our contributions can be summarized as follows:
\begin{itemize}
    \item We investigate the representation learned by CNNs, Transformers, and hybrid models for monocular depth estimation
    using various public datasets and stylized datasets. 
    \item We propose a CNN-Transformer hybrid network with multi-level feature aggregation, which complements the shape bias and spatial locality bias toward the generalization of monocular depth estimation.
    \item Extensive experiments demonstrate the effectiveness of the proposed method, and our method achieves state-of-the-art performance on KITTI datasets, diverse out-of-distribution datasets, and texture-shifted datasets.
\end{itemize}

\section{Related Work}

\subsection{Self-Supervised Monocular Depth Estimation}
Self-supervised depth estimation methods \cite{zhou2017unsupervised,godard2019digging,guizilini20203d,lyu2020hr,klingner2020self,xiong2021self} simultaneously train depth and motion network by imposing photometric consistency loss between target and source images warped by the predicted depth and motion.
Monodepth2 \cite{godard2019digging} presents a minimum reprojection loss to handle occlusions, a full-resolution multi-scale sampling method to reduce visual artifacts, and an auto-masking loss to ignore outlier pixels. 
PackNet-SfM \cite{guizilini20203d} introduces packing and unpacking blocks that leveraged 3D convolutions to learn the dense appearance and geometric information in real-time. 
HR-Depth \cite{lyu2020hr} analyzes the reason for the inaccurate depth prediction in large gradient regions and designed a skip connection to extract representative features in high resolution.

\subsection{Vision Transformers}
Recently, Transformers \cite{vaswani2017attention} start to show promises for solving computer vision tasks such as image classification \cite{dosovitskiy2020image,touvron2021training}, object detection \cite{carion2020end}, and dense prediction. \cite{zheng2021rethinking,ranftl2021vision,yang2021transformer,guizilini2022multi}. 
ViT \cite{dosovitskiy2020image} employs Transformers architecture on fixed-size image patches for image classification for the first time. DeiT \cite{touvron2021training} utilizes Knowledge distinction on ViT architecture, showing good performance only with the ImageNet dataset.
Some works \cite{ranftl2021vision,yang2021transformer} have employed Transformers for monocular depth estimation in a supervised manner. 
TransDepth \cite{yang2021transformer} utilizes multi-scale information to capture local level details.  These works \cite{zheng2021rethinking,yang2021transformer} only focus on improving performance on benchmark datasets. Previous works lack studies on whether models behave as intended in another domain dataset.

\section{Method}

\begin{figure*}[t] 
\begin{center}
\includegraphics[width=0.85\textwidth]{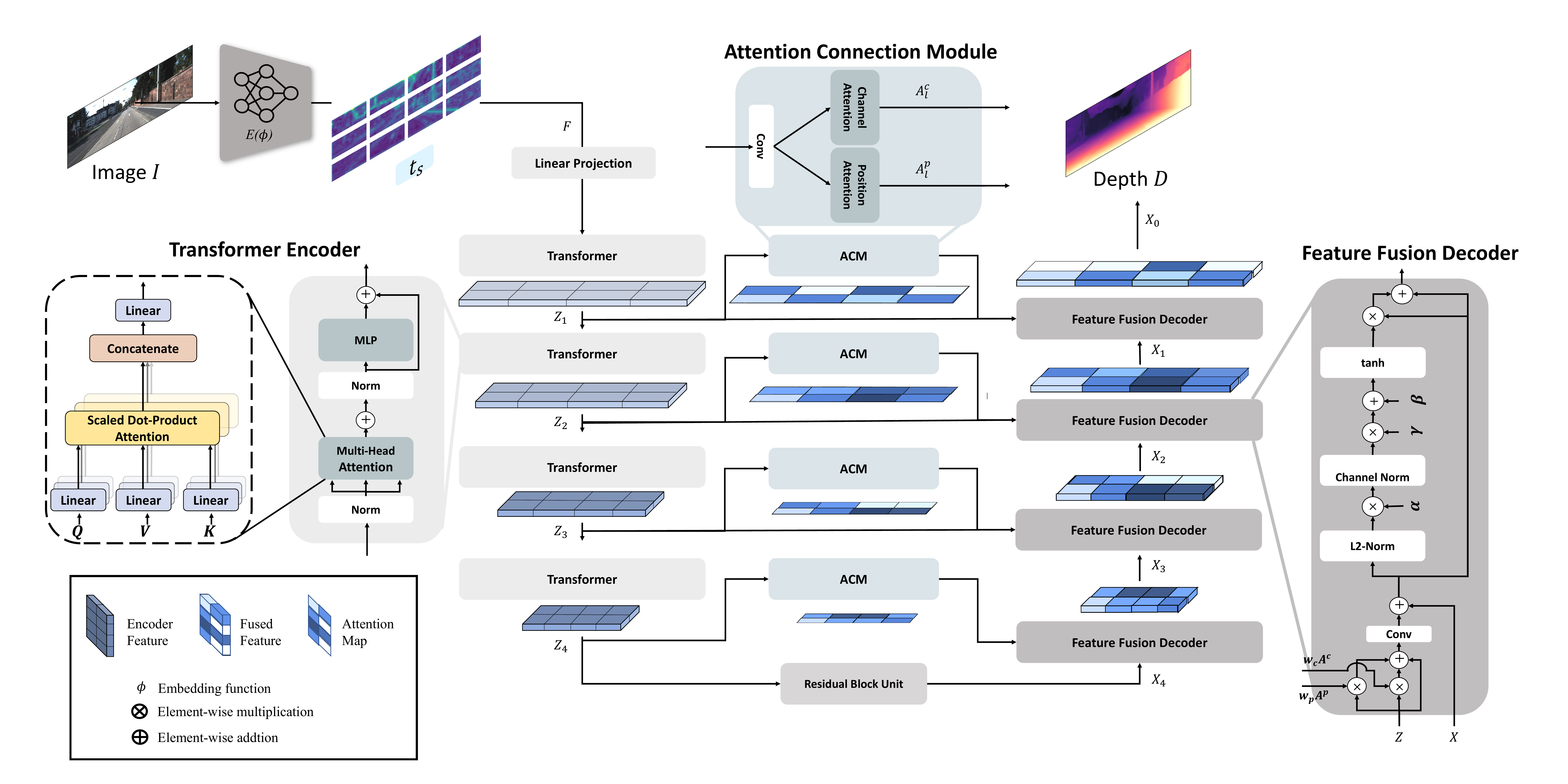}
\end{center}
\caption{\textbf{Overall Architecture.} We design an encoder-decoder structure with a multi-level feature fusion module. The encoder is composed of a CNN and Transformers. The ACM learns the channel and position attentions. The FFD adaptively fuses the encoder features using the attention maps.}
\label{figure_network_overview}
\end{figure*}

\subsection{CNN-Transformer Encoder}
\label{sec:encoder}
The encoder consists of a CNN and Transformers. We use ResNet50 \cite{he2016deep} as the CNN backbone ($E(\theta)$ in \figref{figure_network_overview}), and $L$ number of Transformers. In this work, we set the $L$ as 4. 
An input image $I$ passes through the CNN encoder to extract a feature map $F \in \mathbb{R}^{C \times H \times W}$, then the map is divided into $N$ $(= \frac{H}{16} \times \frac{W}{16})$ number of patches $p_{n} \in \mathbb{R}^{C\times 16 \times 16}$, which is utilized as the input of the first Transformer layer.
We additionally use a special token $t_s$ following the work \cite{ranftl2021vision}.
We input the patch tokens $p_{n},~n \in\{{1,...,N}\}$ and the special token $t_s$ with a learnable linear projection layer $E$ as follows:
\begin{equation}
\small
    Z_0 = [t_s;~p_1E;~p_2E;~...~;~p_{N}E],
\end{equation}
where $Z_0$ is the latent embedding vector.
The Transformer encoder consists of a Multi-head Self-Attention (MSA) layer, a Multi-Layer Perceptron (MLP) layer, and Layer Norm (LN) layers. The MLP is built with GELU non-linearity \cite{hendrycks2016gaussian}. The LN is applied before every block and residual connections apply after every block.
Self-Attention (SA) at each layer $l \in \{1,...,L\}$ is processed with the learnable parameters $W^m_Q , W^m_K , W^m_V \in \mathbb{R}^{C \times d}$ of \{query, key, value\} weight matrices, given the embedding vector $Z_l \in \mathbb{R}^{N \times C}$ as follows:
\begin{equation}
\small
\begin{gathered}
        \text{SA}^m_{l-1} = \text{softmax}(\frac{Q^m_{l-1} (K^m_{l-1})^\text{T}}{\sqrt{d}} )V^m_{l-1},~m \in \{1,...,M\},\\
    Q^m_{l-1} = Z_{l-1}W^m_Q,~K^m_{l-1} = Z_{l-1}W^m_K,~V^m_{l-1} = Z_{l-1}W^m_V,
\end{gathered}
\end{equation}
where $M$ and $d$ are the number of SA blocks and the dimension of the self-attention block, which is the same as the dimension of the weight matrices, respectively. 
The Multi-head Self-Attention (MSA) consists of the $M$ number of SA blocks with the learnable parameters of weight
matrices $W \in \mathbb{R}^{Md \times C }$ as follows:
\begin{equation}
\small
\begin{gathered}
\text{MSA}_{l-1} = Z_{l-1} +  \text{concat}(\text{SA}^1_{l-1};~\text{SA}^2_{l-1};~ \dots ;~\text{SA}^M_{l-1} )W,\\
Z_{l} = \text{MLP}(\text{LN}(\text{MSA}_{l-1}))+\text{MSA}_{l-1}.
\end{gathered}
\end{equation}
This Transformer layer is repeated $L$ times with unique learnable parameters. The outputs of the Transformers $\{Z_1,...,Z_L\}$ are utilized as the input of the following layers ACM and FFD.

\begin{figure*}[t!] 
    \centering
    \resizebox{\textwidth}{!}{
    \begin{tabular}{c@{\hspace{1mm}}c@{\hspace{1mm}}c@{\hspace{1mm}}c@{\hspace{1mm}}c@{\hspace{1mm}}}
     \includegraphics[]{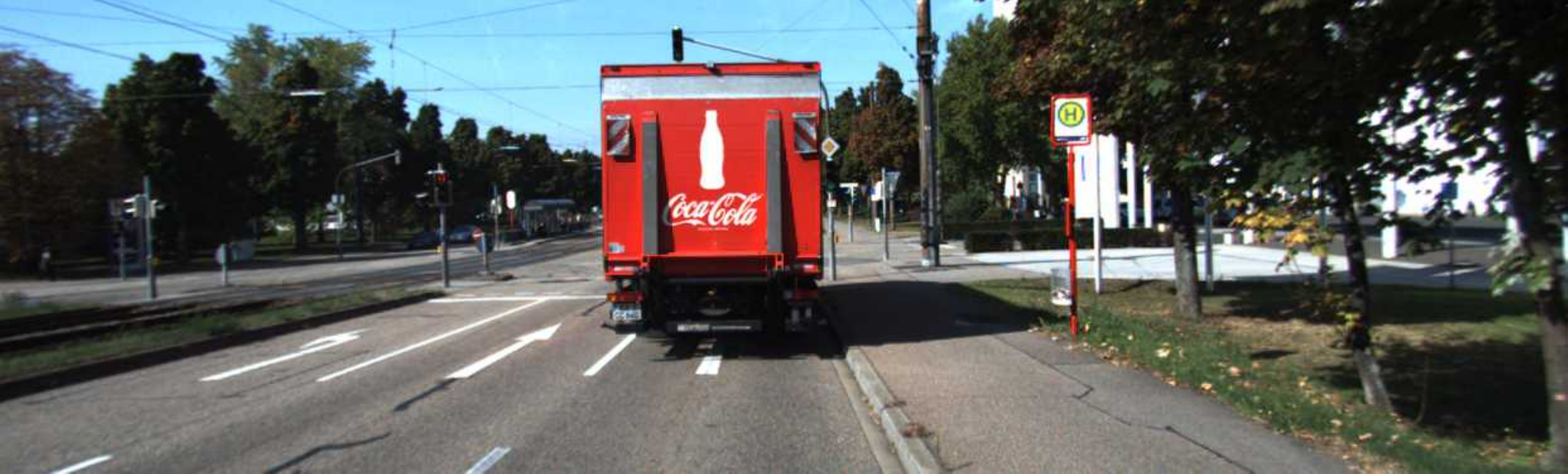}& \includegraphics[]{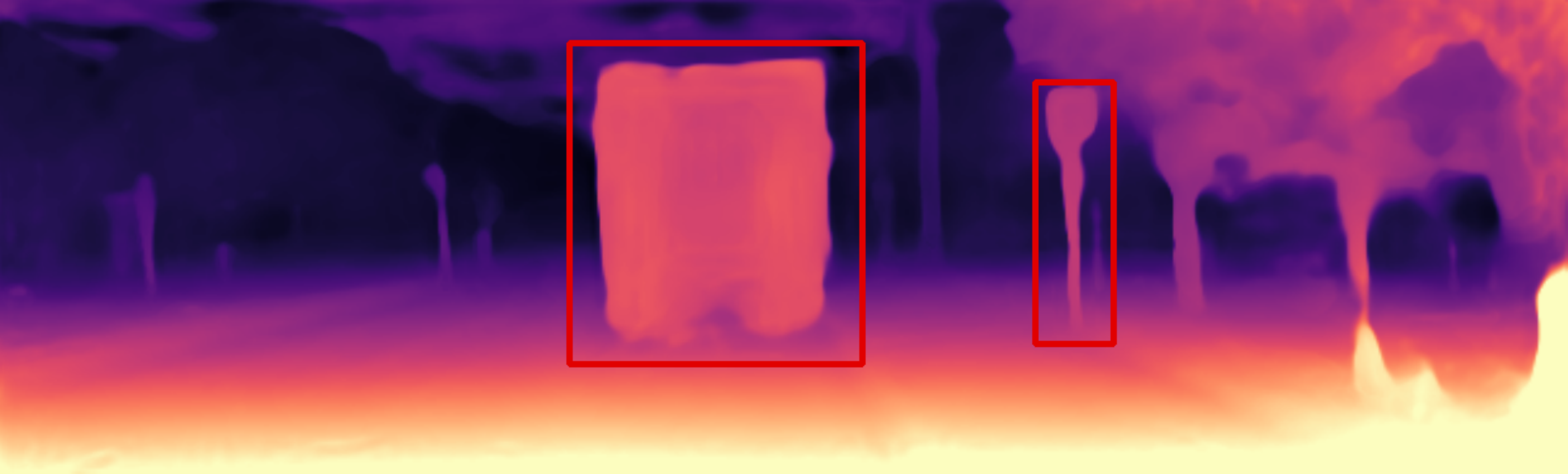}&  \includegraphics[]{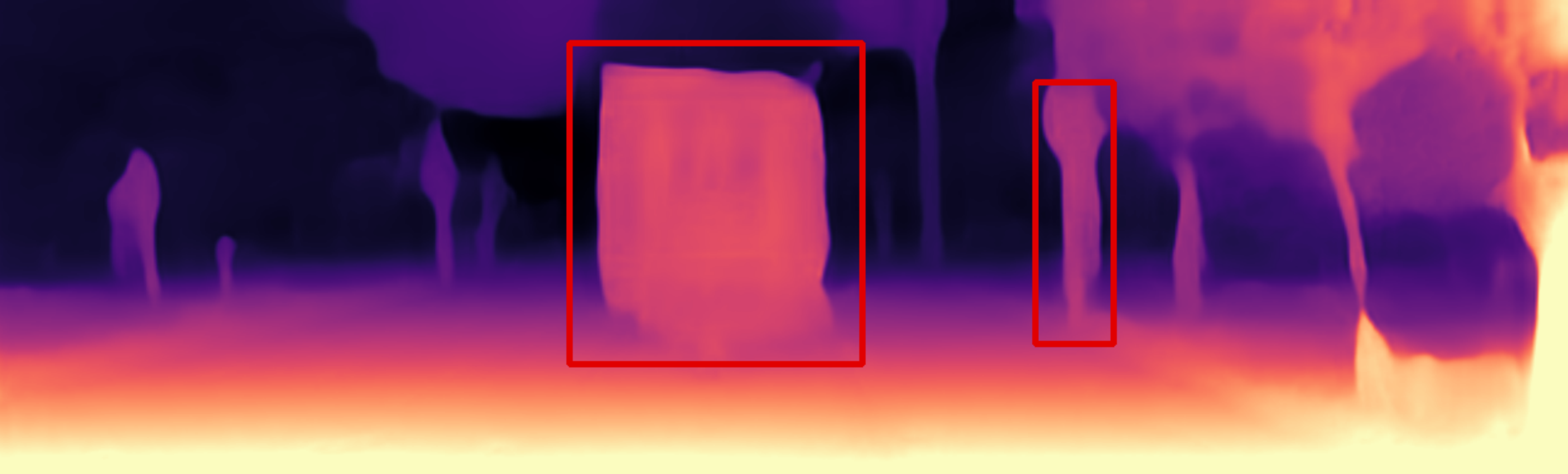}&
     \includegraphics[]{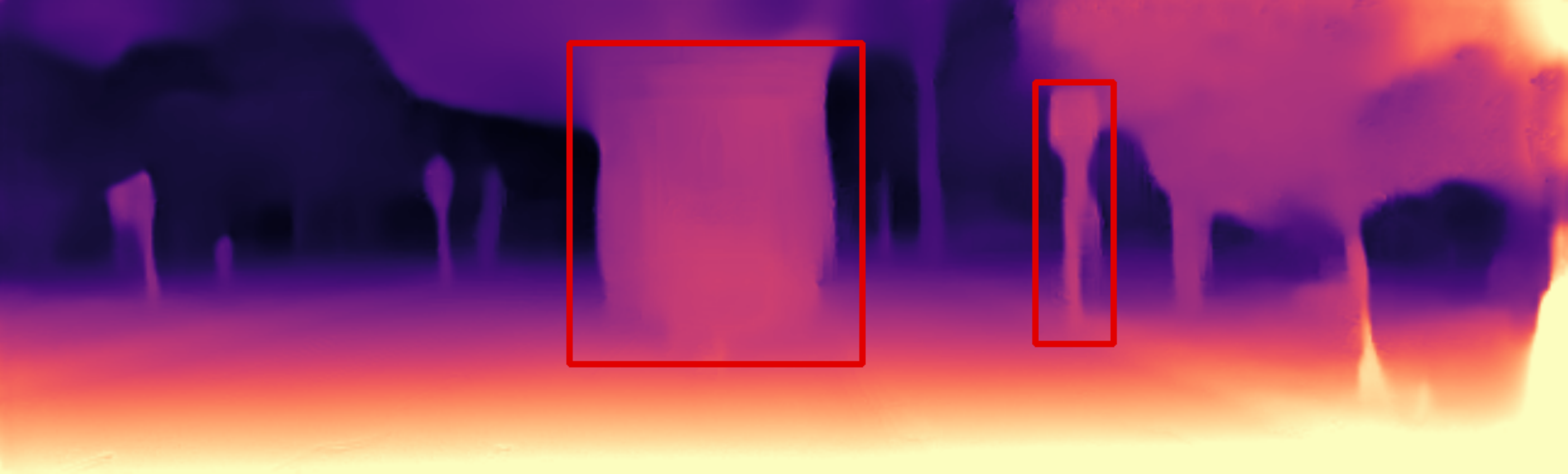}&
     \includegraphics[]{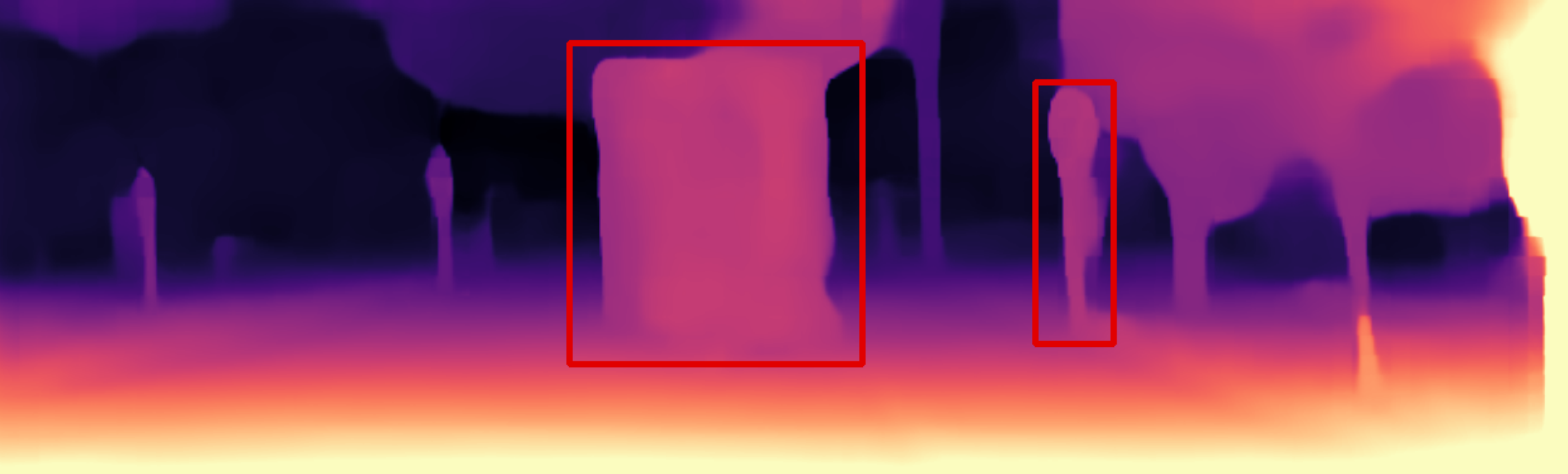}   \\
     \includegraphics[]{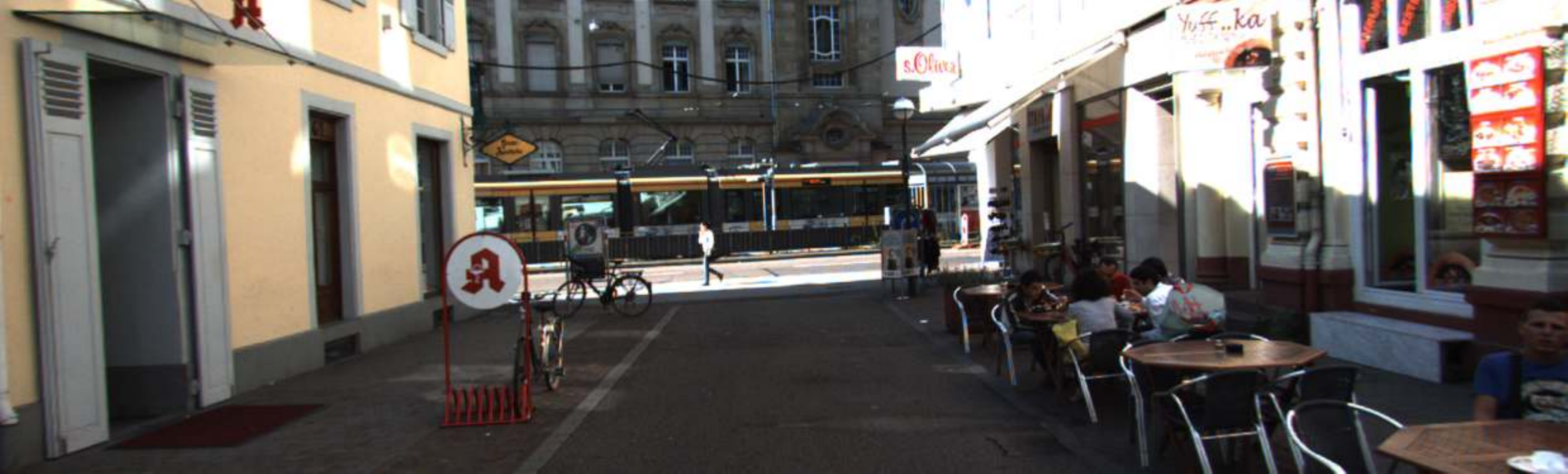}& \includegraphics[]{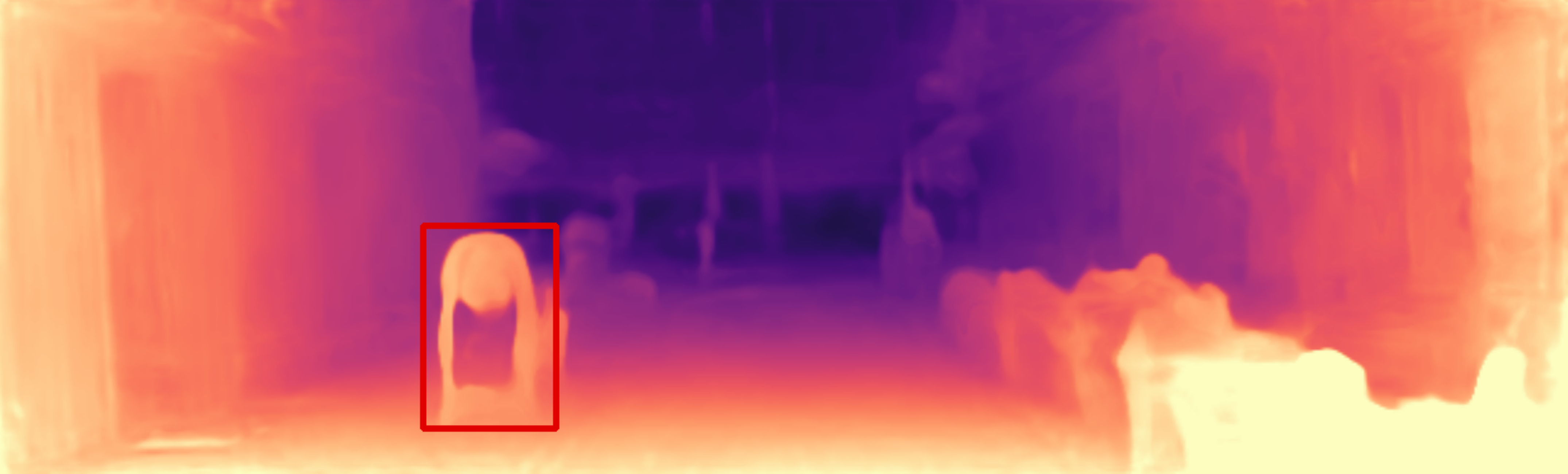}&  \includegraphics[]{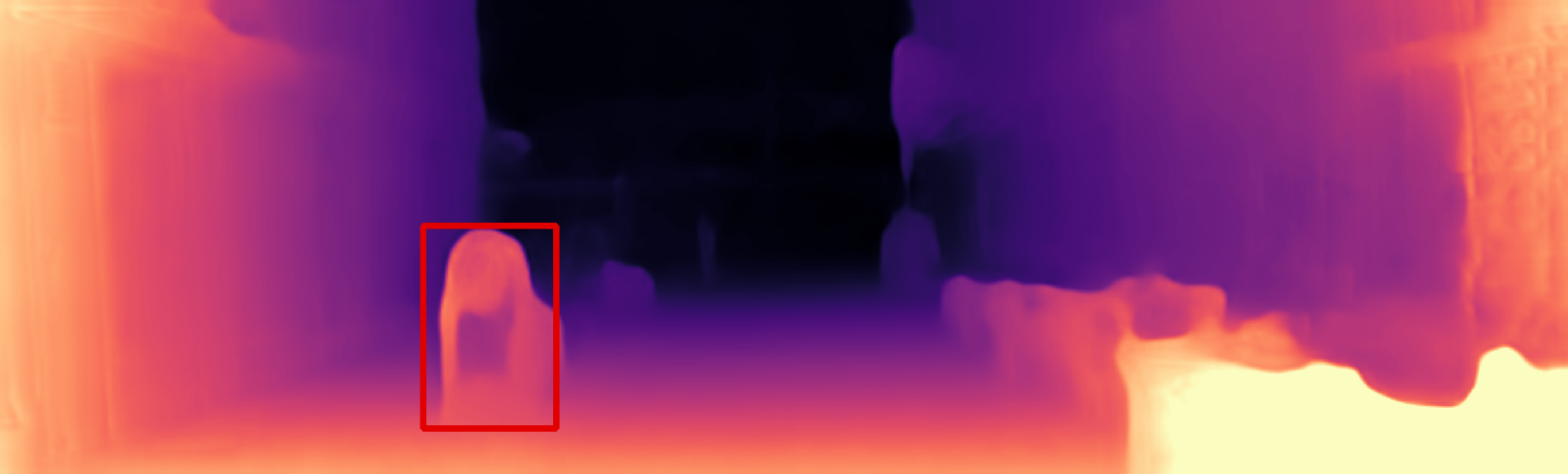}&
     \includegraphics[]{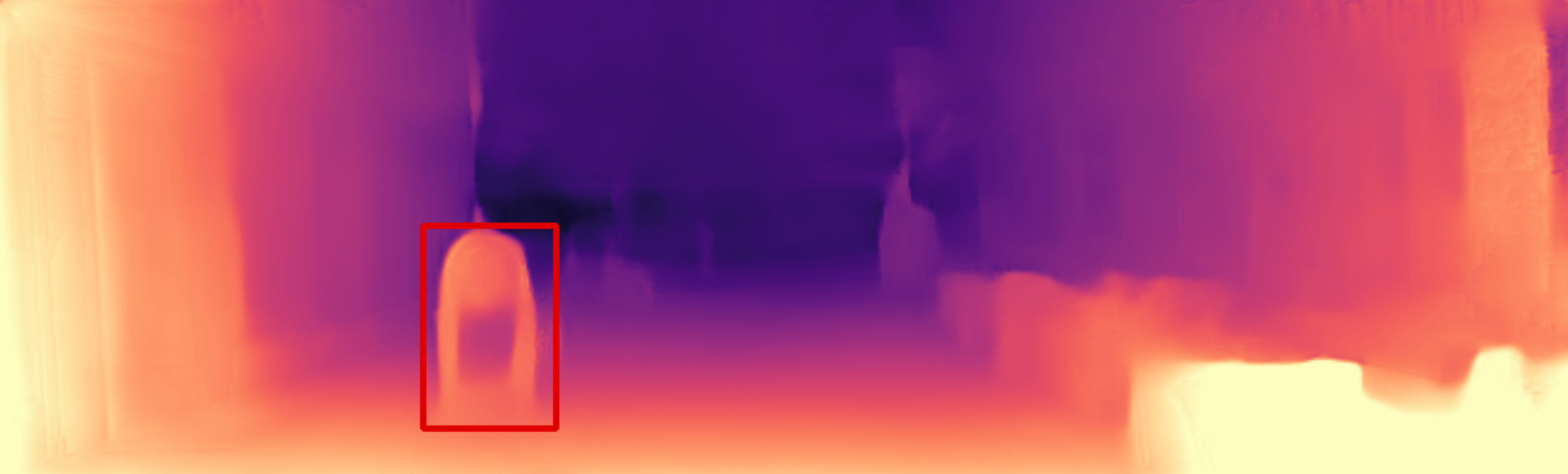}&
     \includegraphics[]{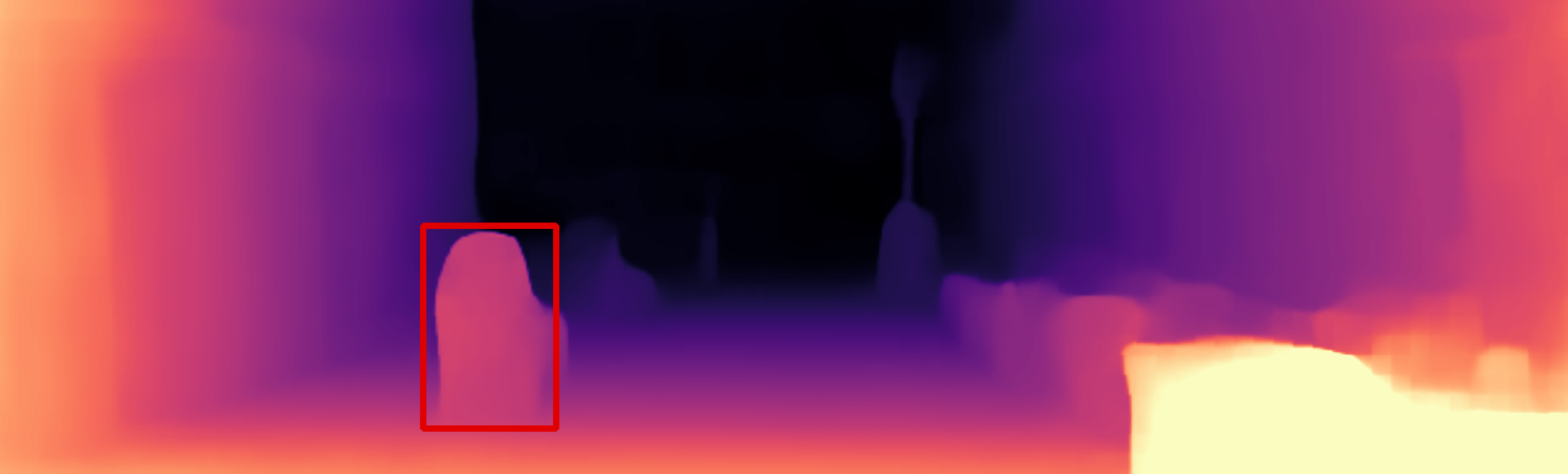}   \\
     \includegraphics[]{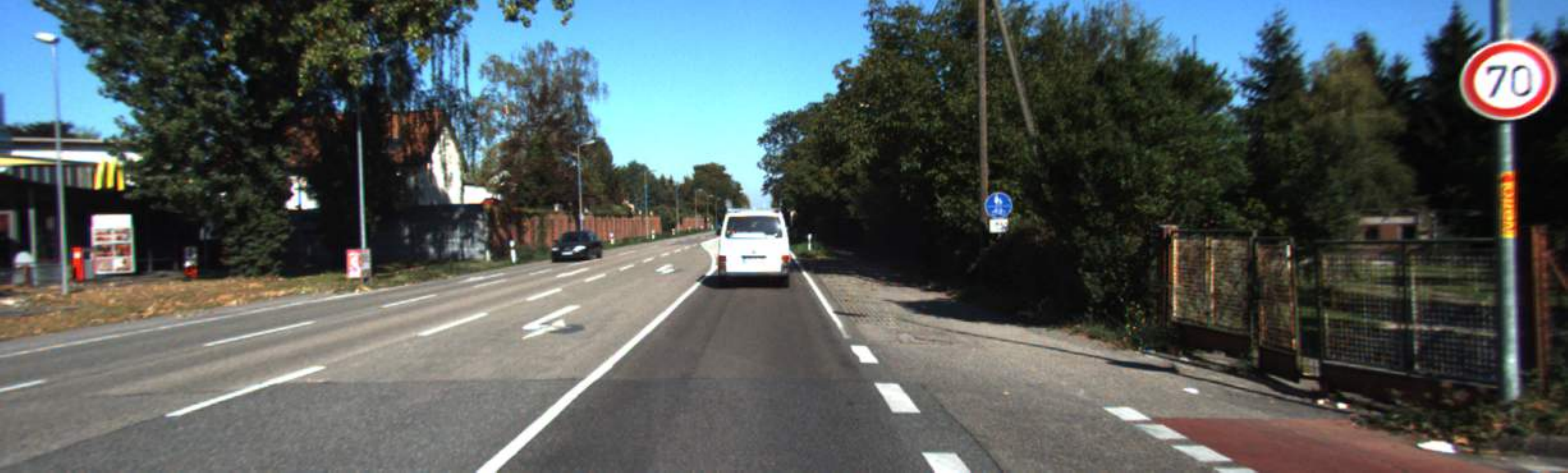}& \includegraphics[]{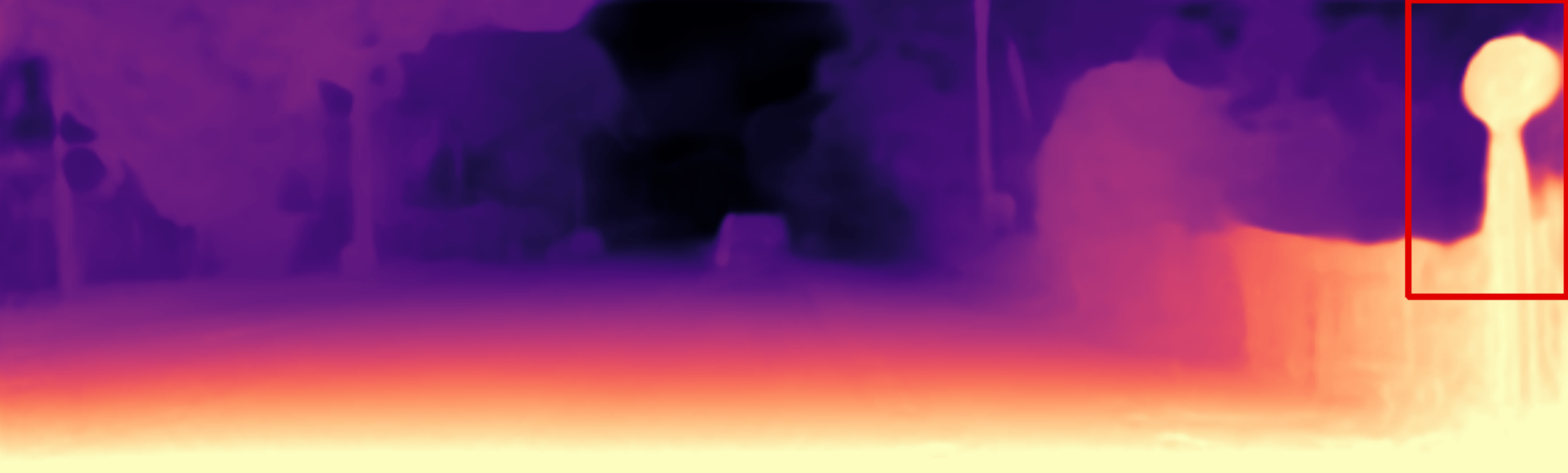}&  \includegraphics[]{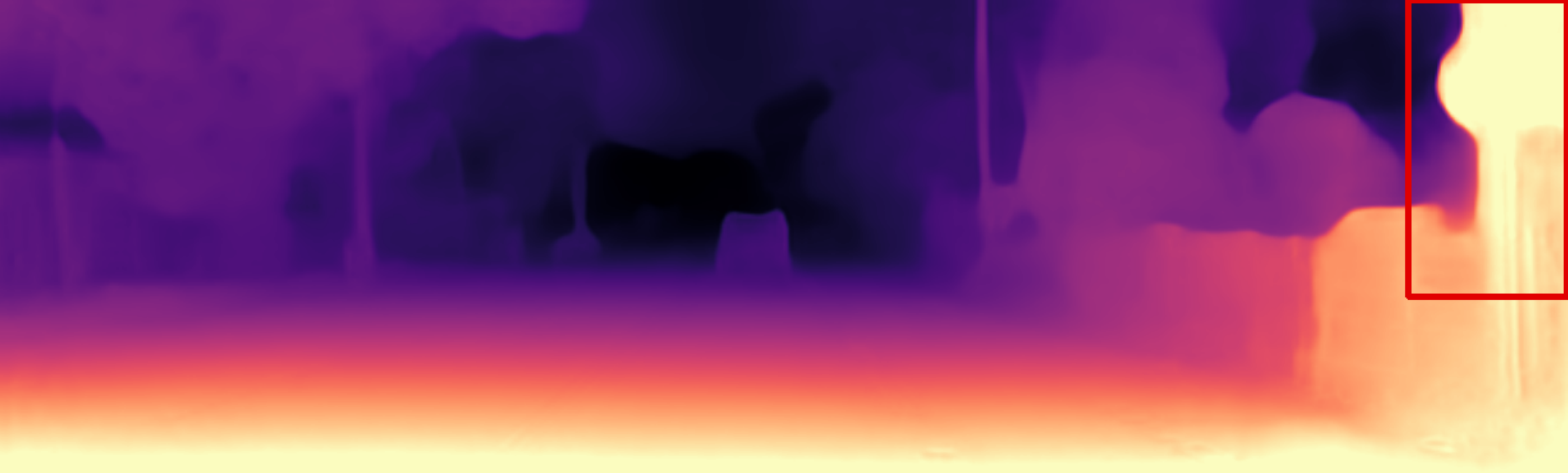}&
     \includegraphics[]{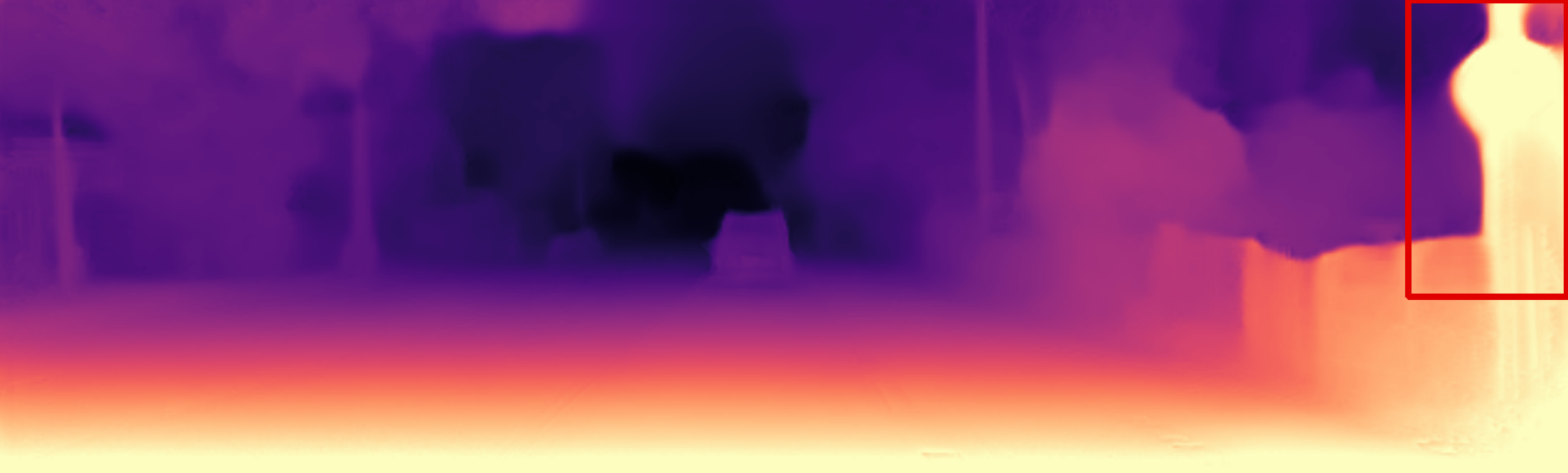}&
     \includegraphics[]{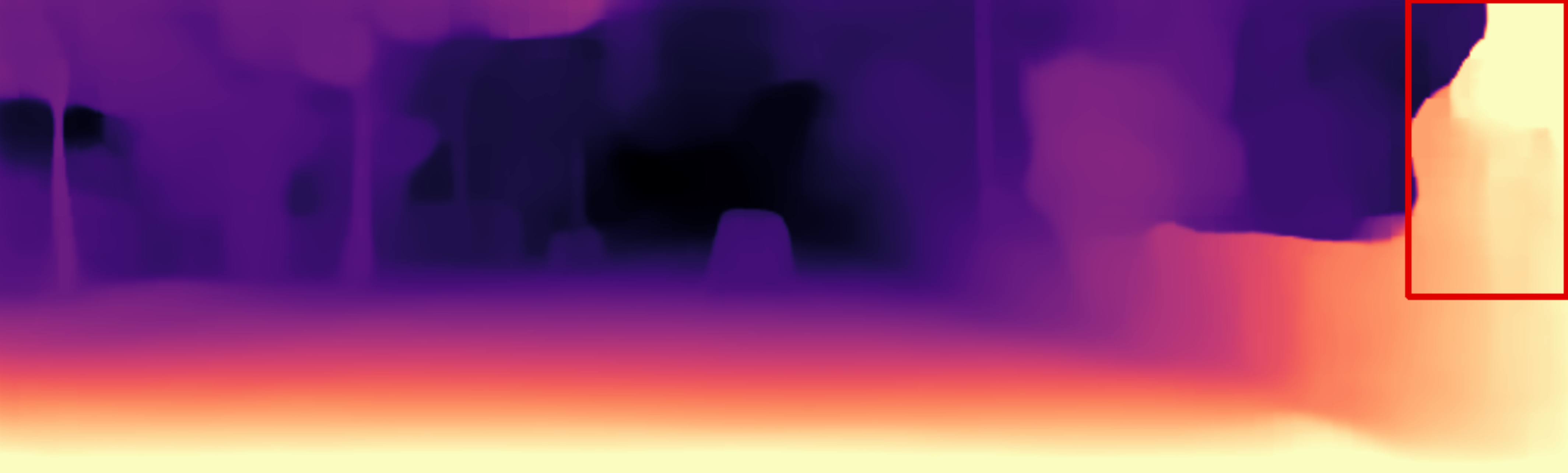}   \\
     \includegraphics[]{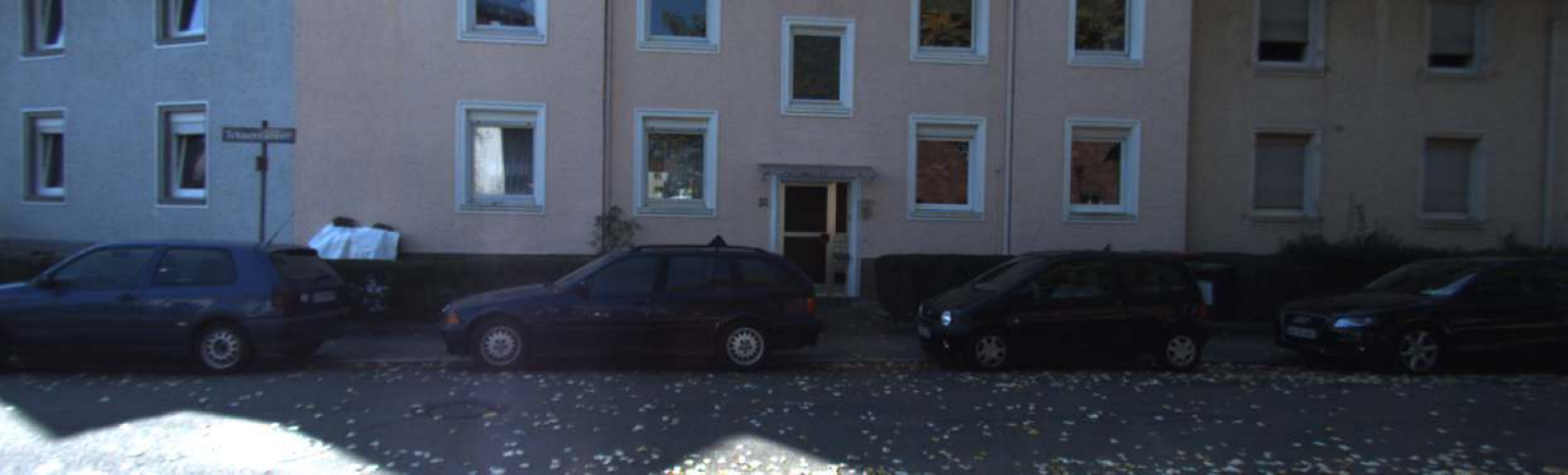}& \includegraphics[]{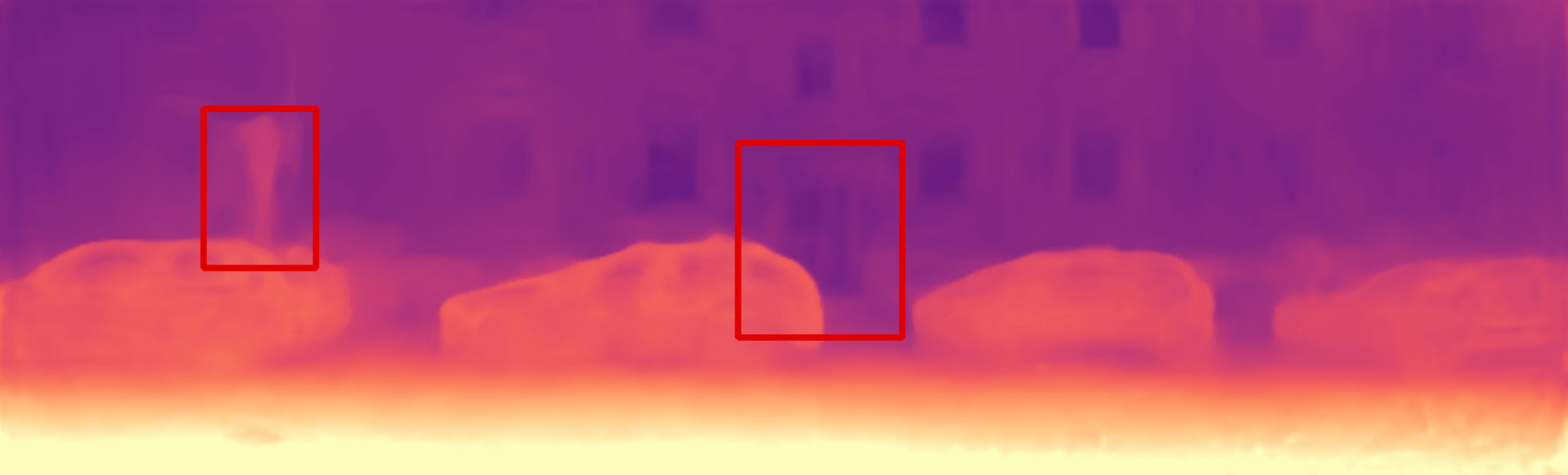}&  \includegraphics[]{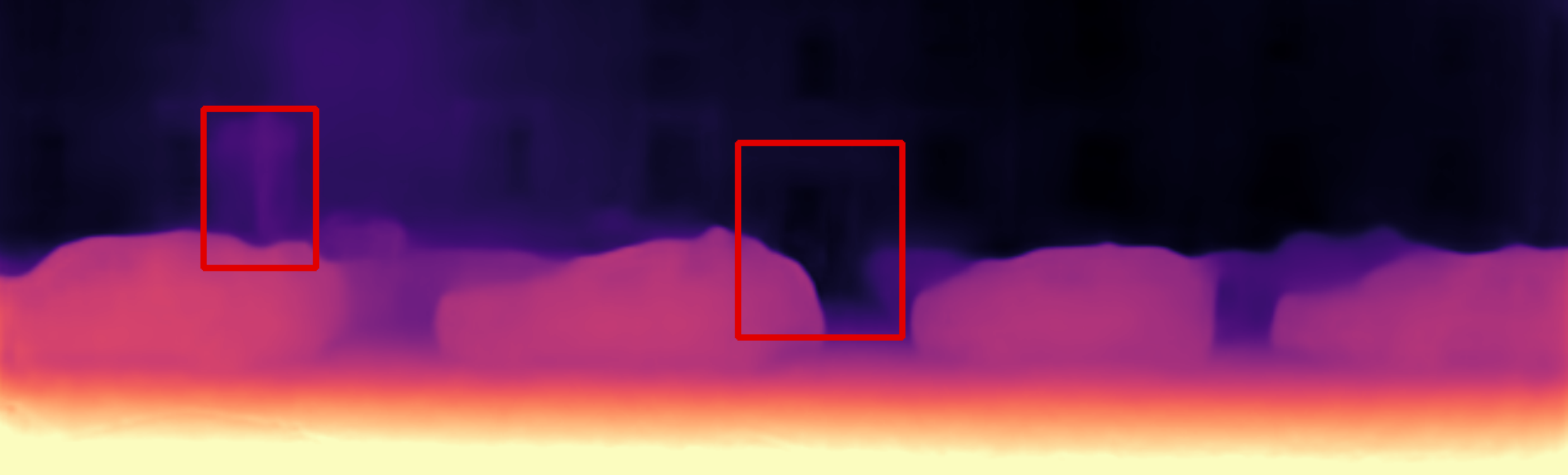}&
     \includegraphics[]{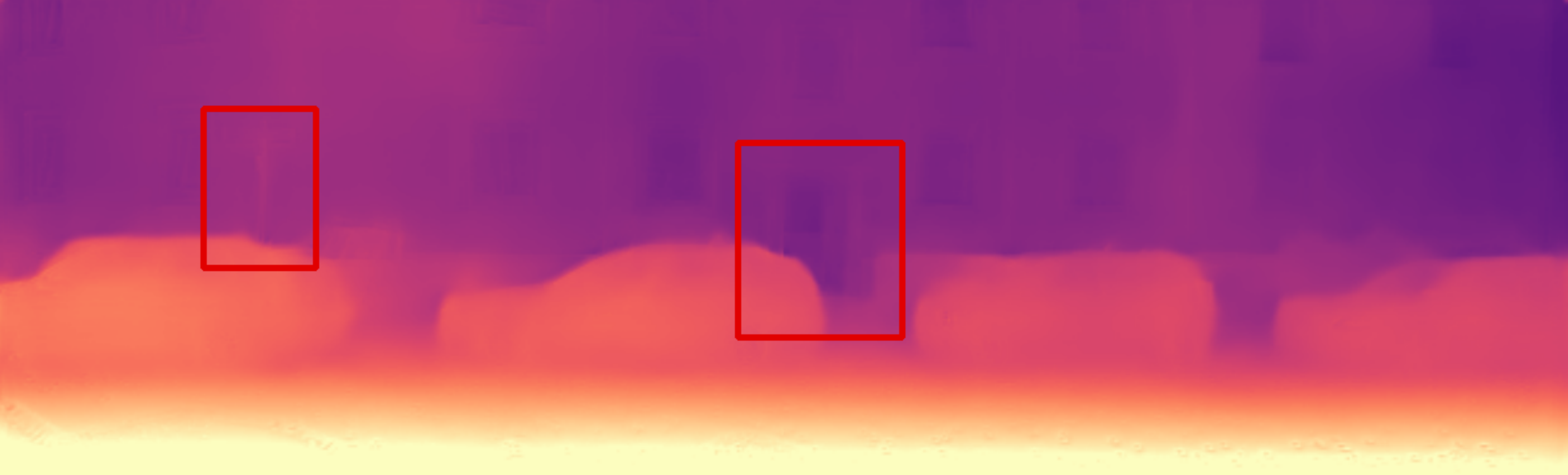}&
     \includegraphics[]{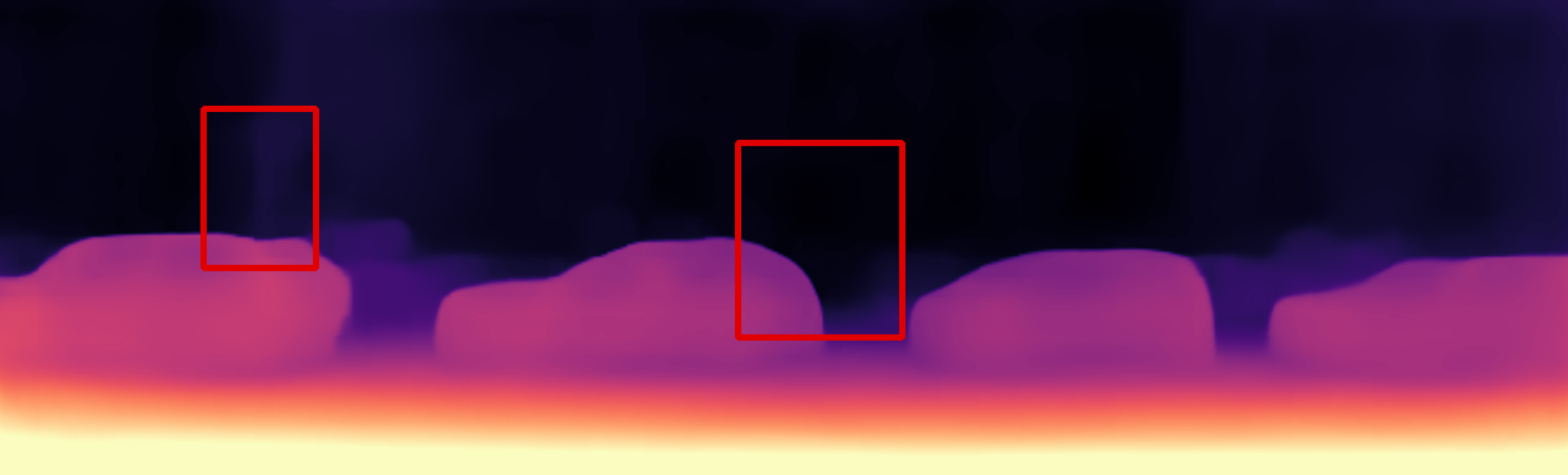}   \\
     \fontsize{120}{50} \selectfont Input images & \fontsize{120}{50} \selectfont MonoFormer (Hybrid) & \fontsize{120}{50} \selectfont Monodepth2&
     \fontsize{120}{50} \selectfont PackNet-SfM & \fontsize{120}{50} \selectfont R-MSFM6
    \end{tabular}}
    \caption{\textbf{Qualitative comparison to state-of-the-arts.} We use KITTI for training and testing.}
\label{figure_result_kitti}
\end{figure*}

\begin{figure*}[t] 
    \centering
    \resizebox{\textwidth}{!}{
    \begin{tabular}{cccccc}
    \includegraphics[]{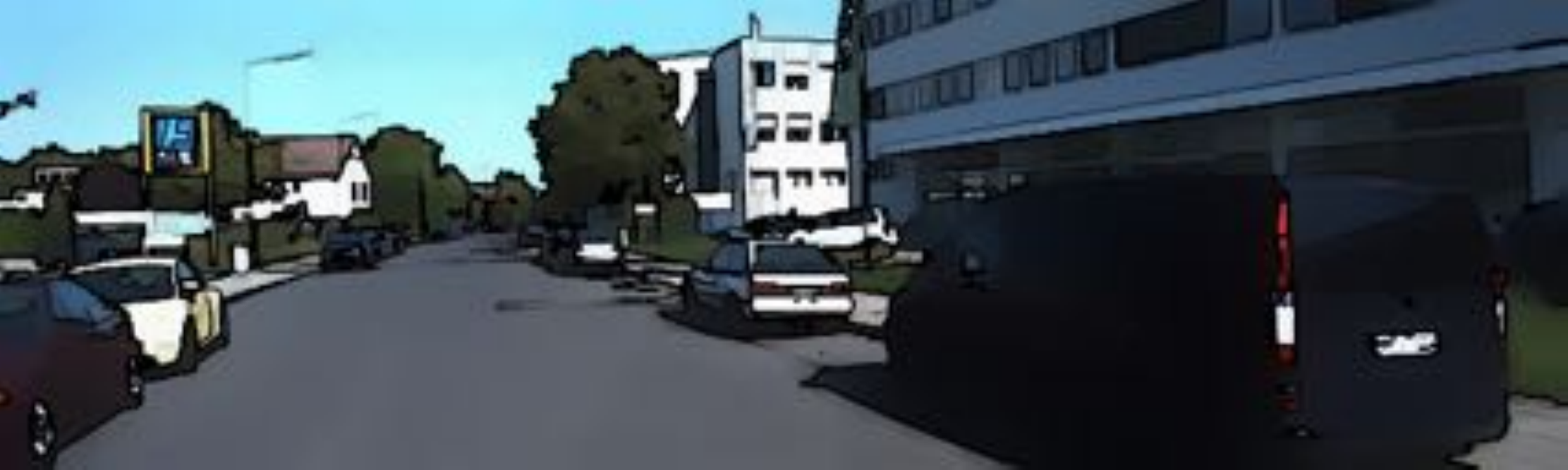}& 
    \includegraphics[]{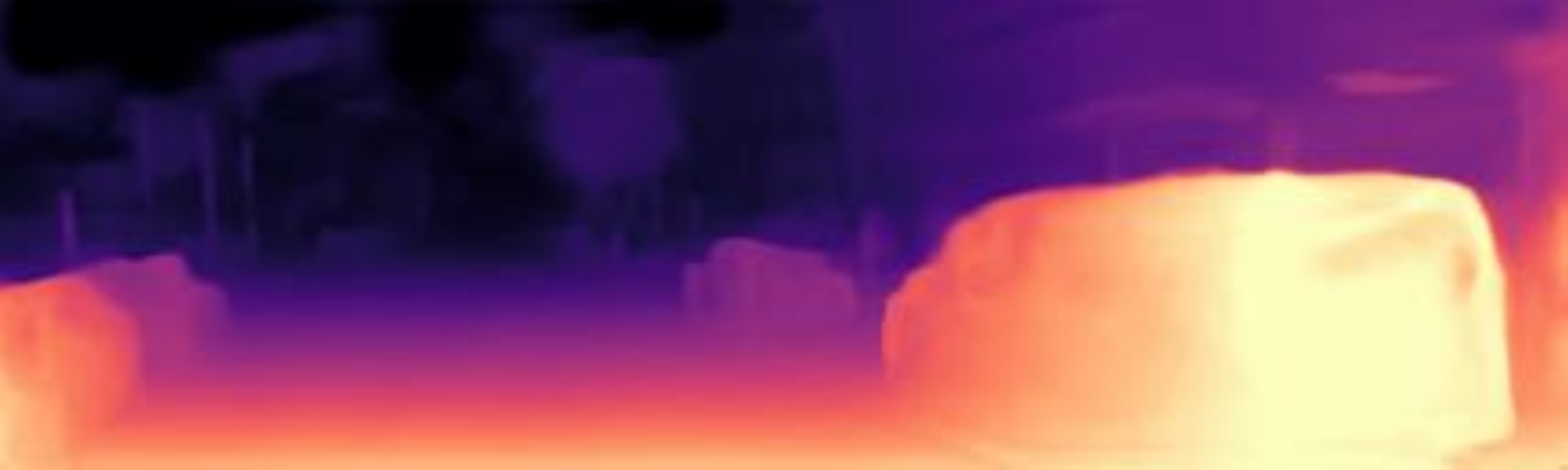}&
    \includegraphics[]{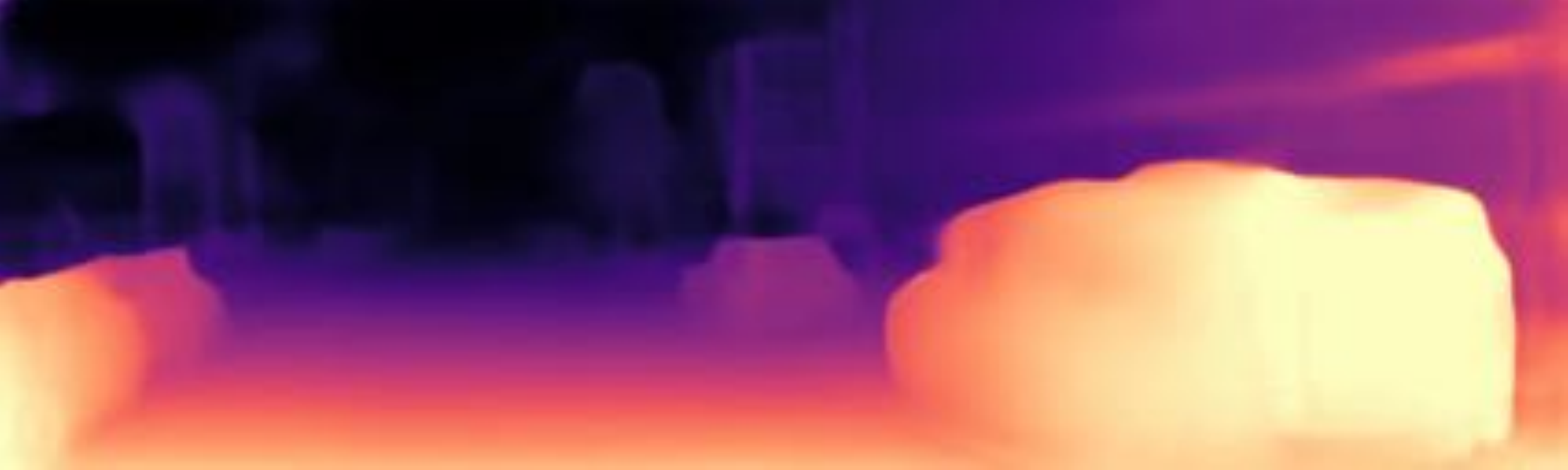}&
    \includegraphics[]{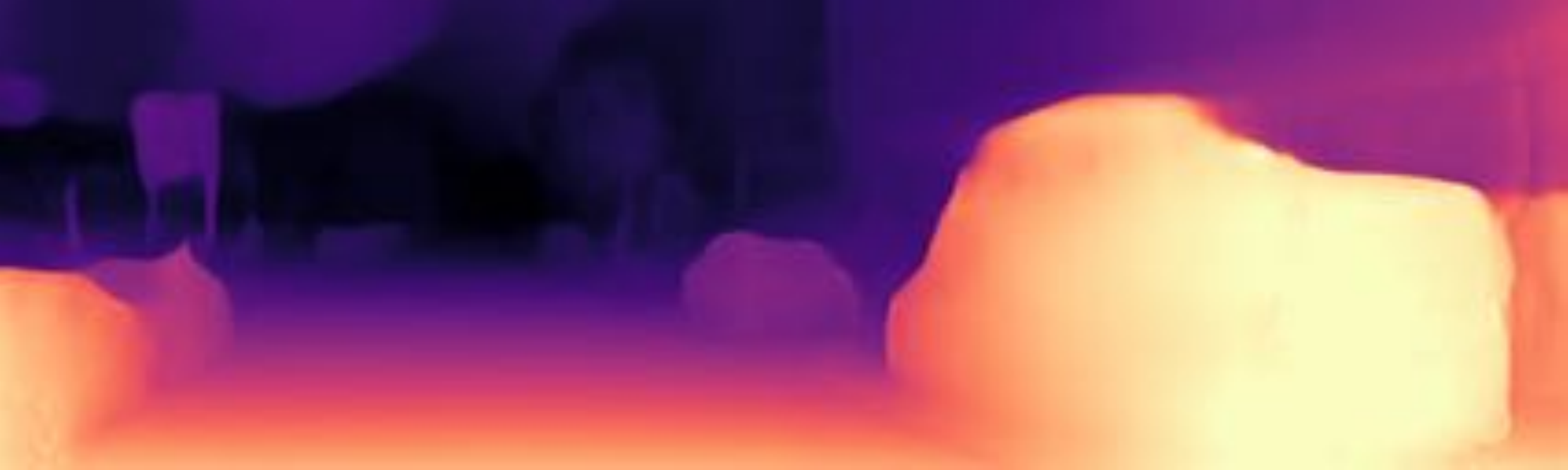}&
    \includegraphics[]{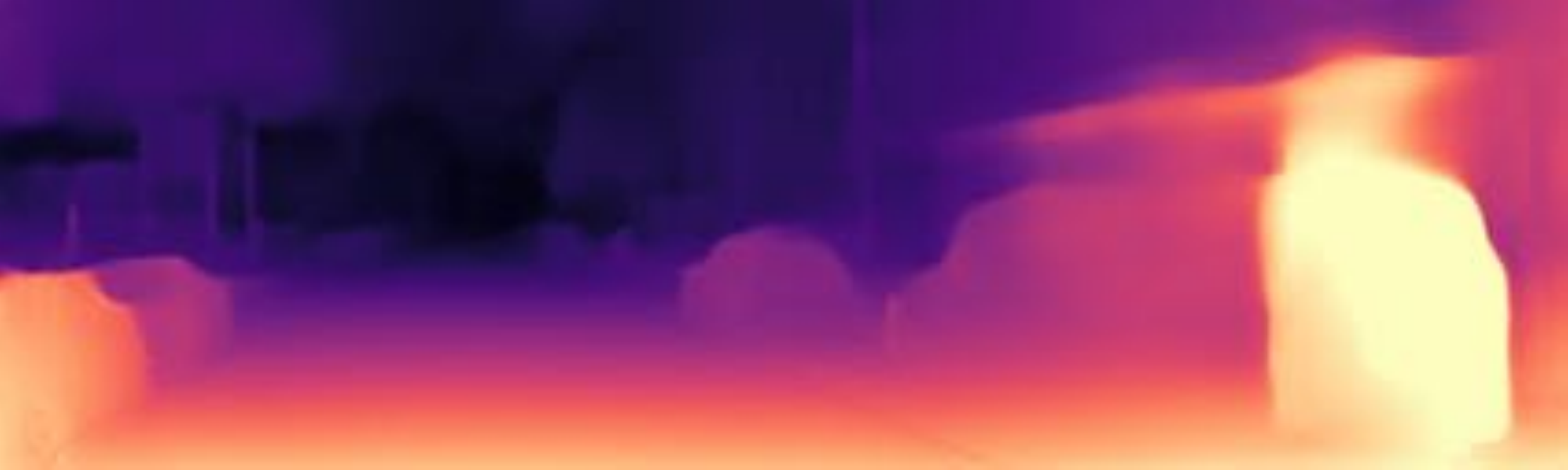}&
    \includegraphics[]{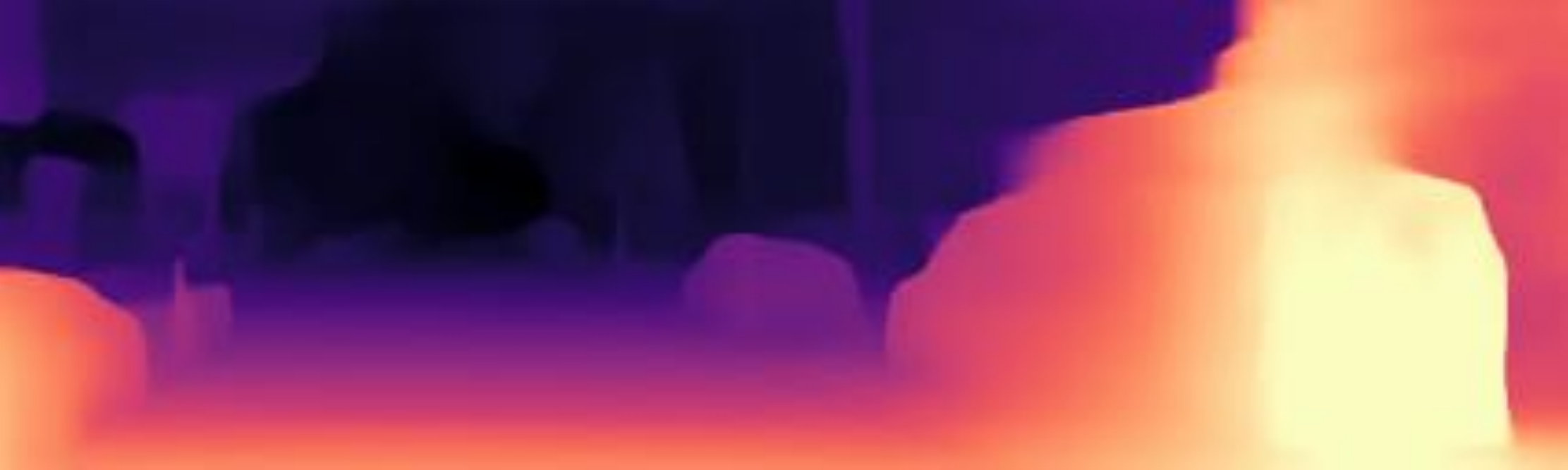} \\
    \includegraphics[]{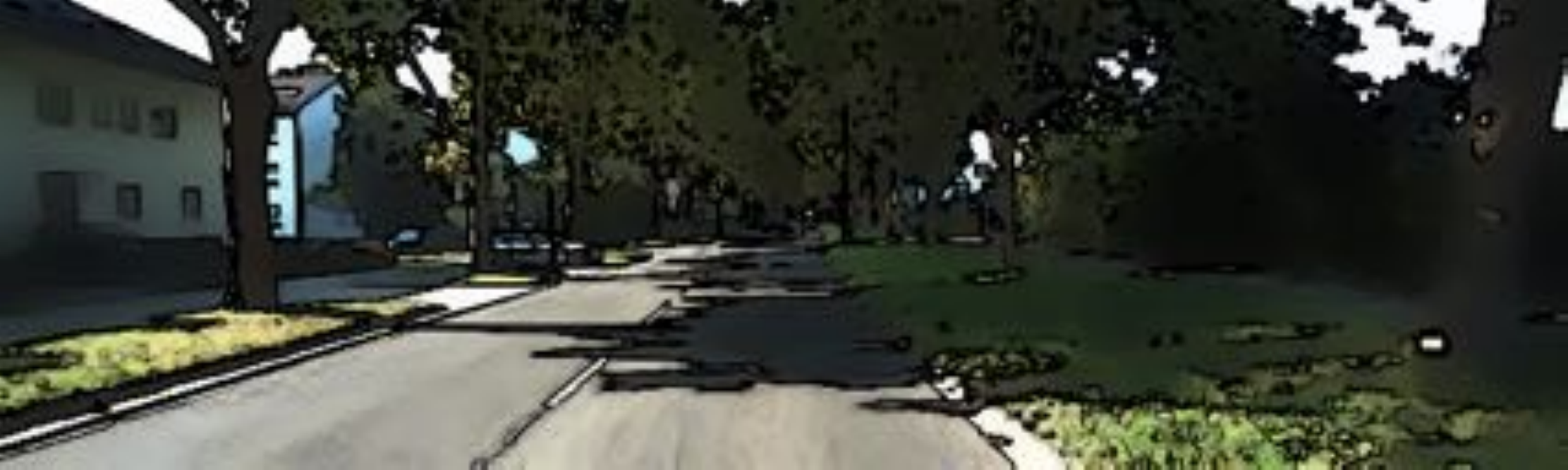}& 
    \includegraphics[]{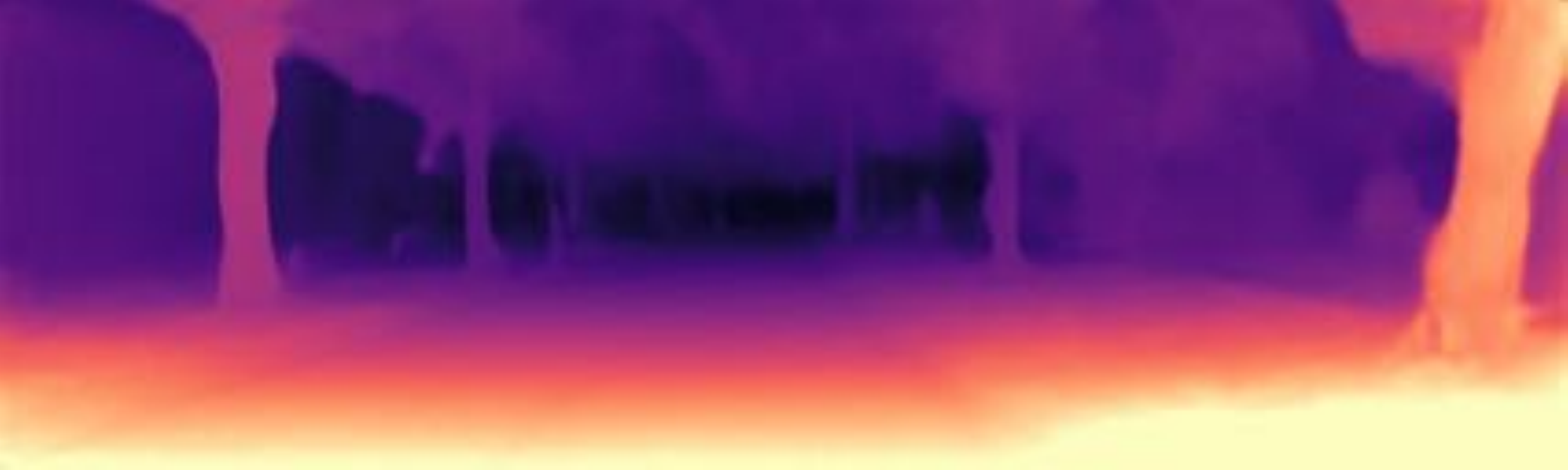}&
    \includegraphics[]{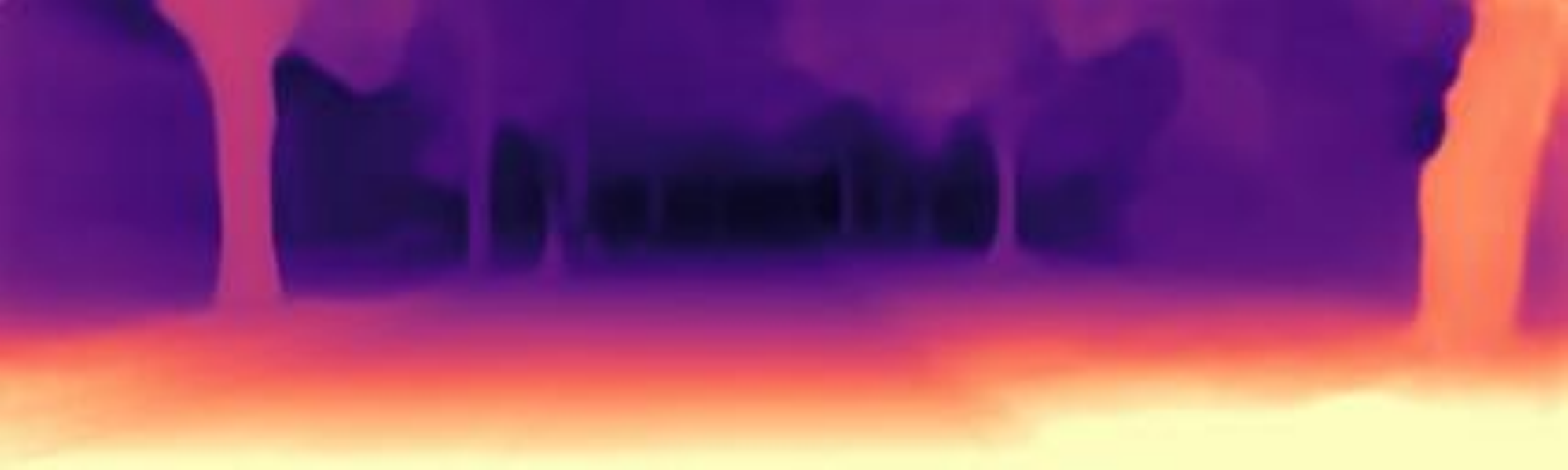}&
    \includegraphics[]{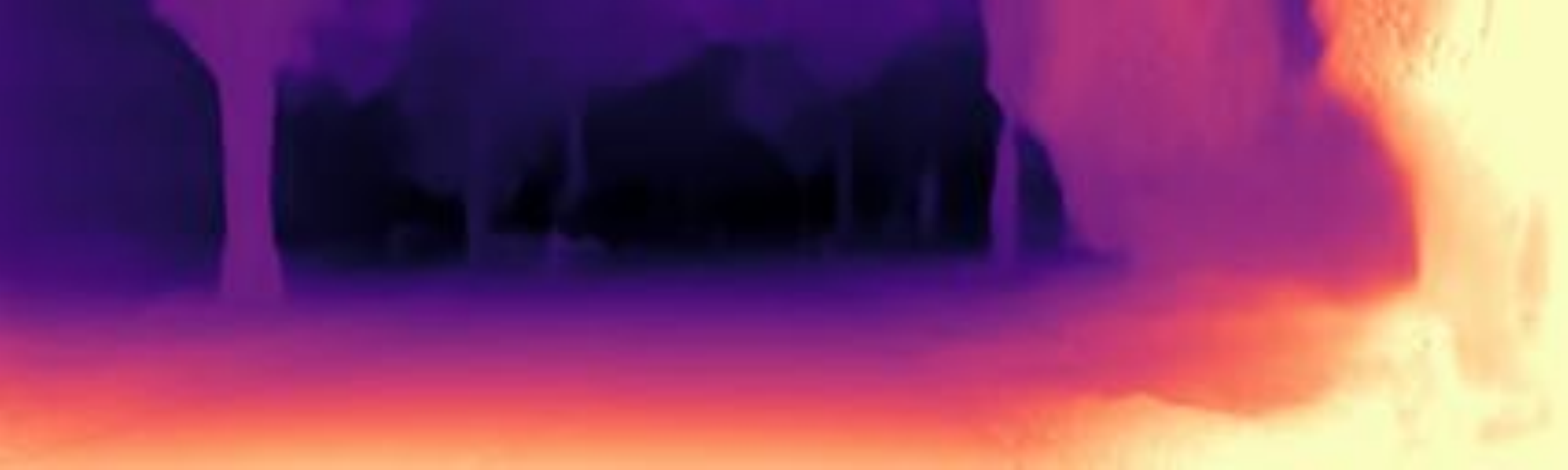}&
    \includegraphics[]{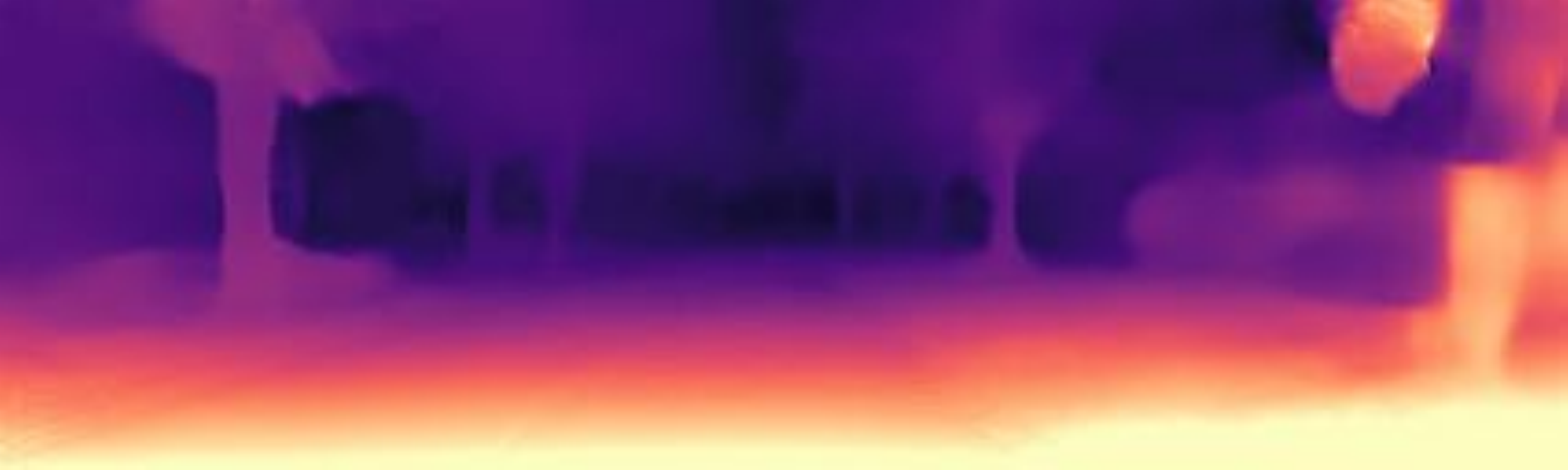}&
    \includegraphics[]{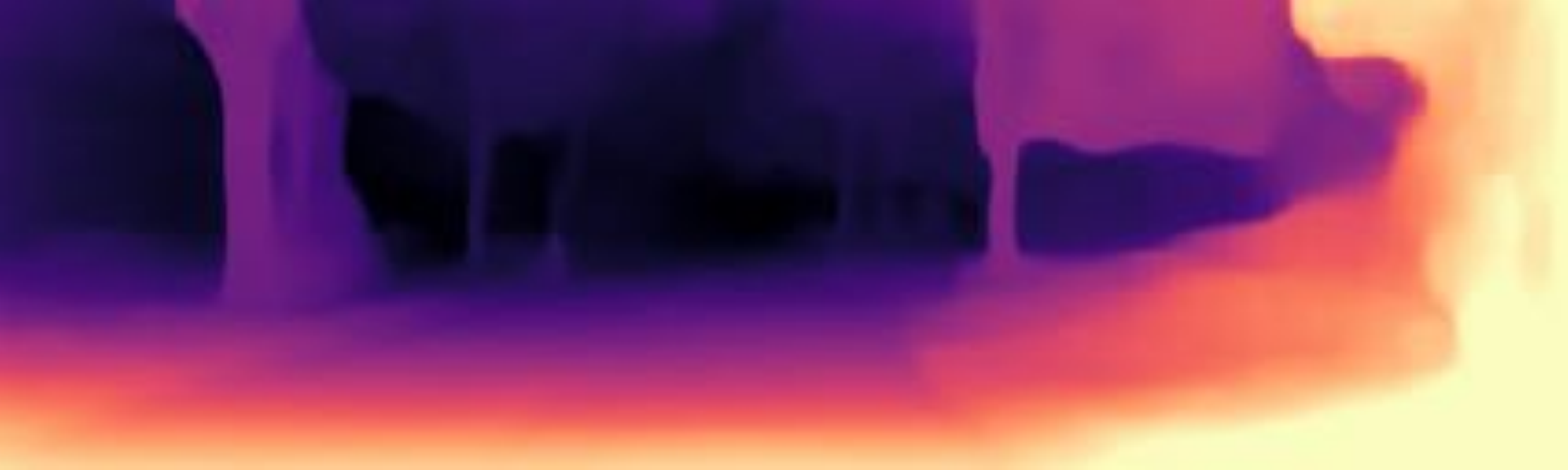} \\
    \includegraphics[]{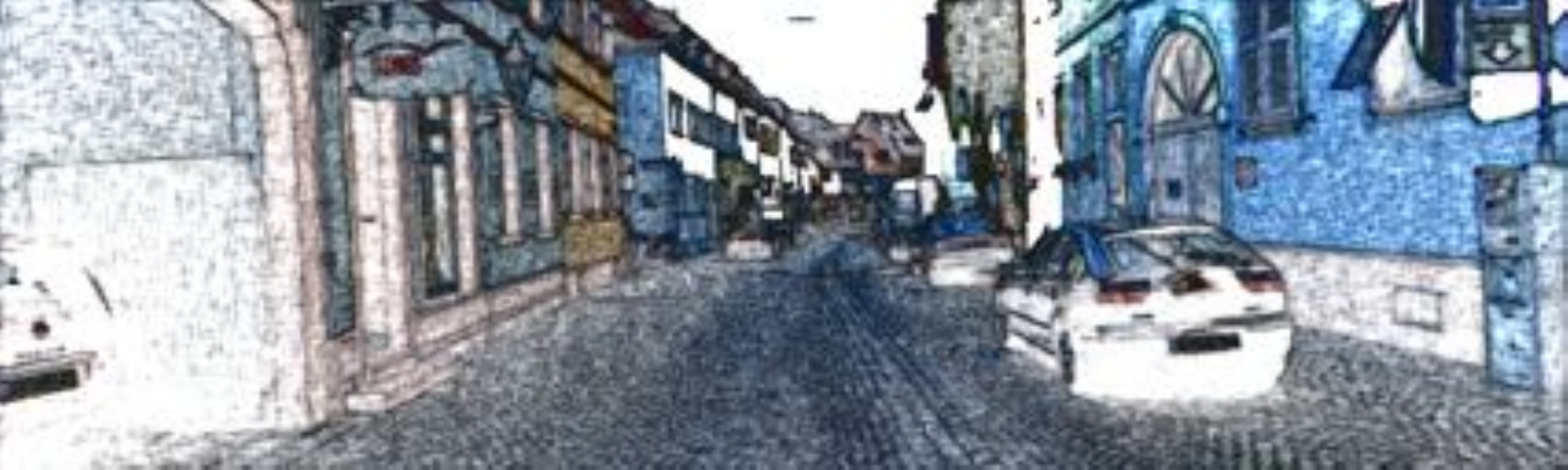}& 
    \includegraphics[]{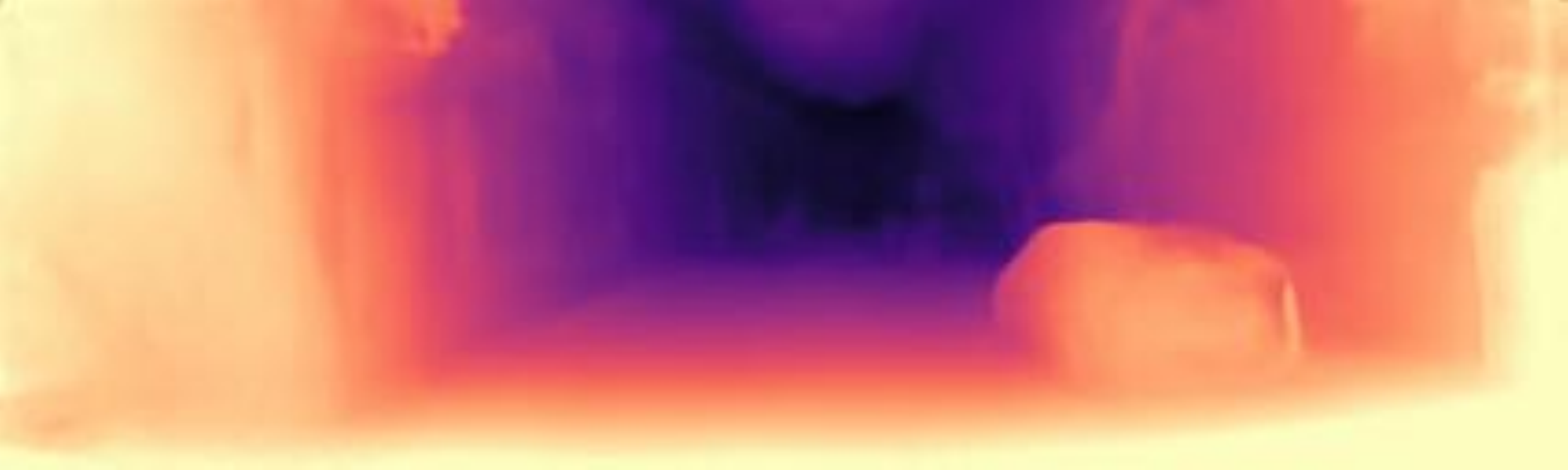}&
    \includegraphics[]{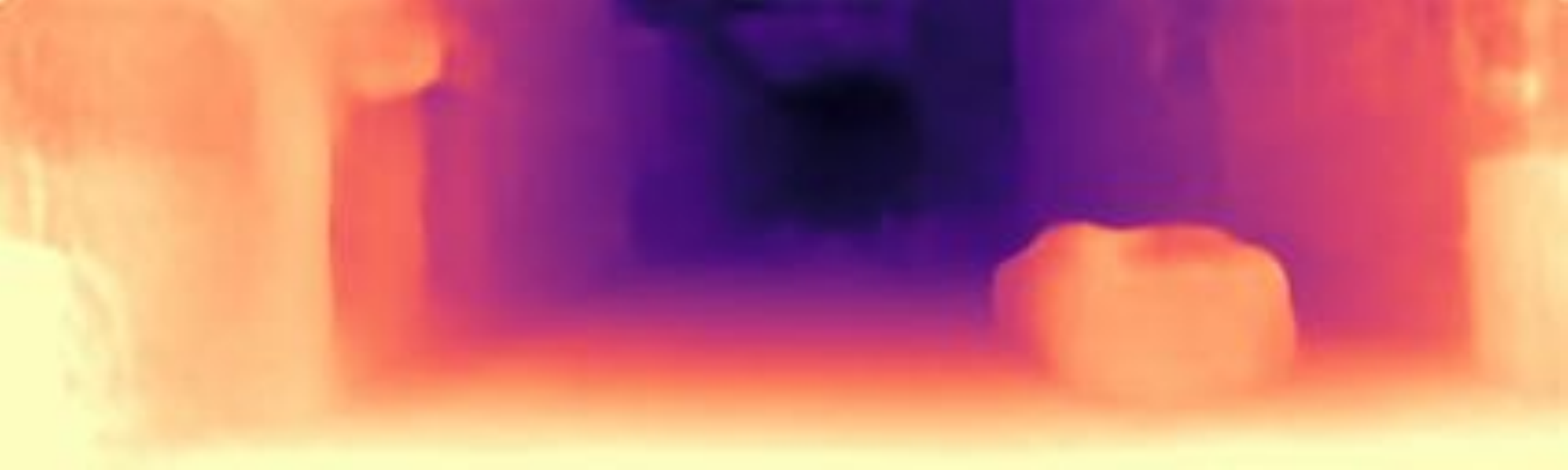}&
    \includegraphics[]{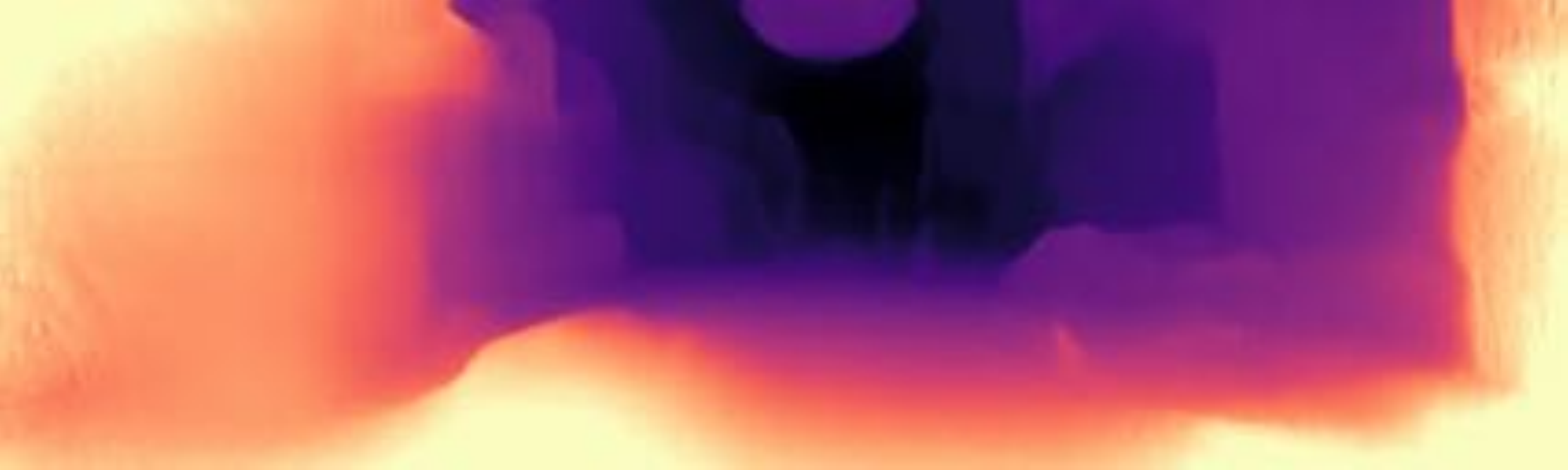}&
    \includegraphics[]{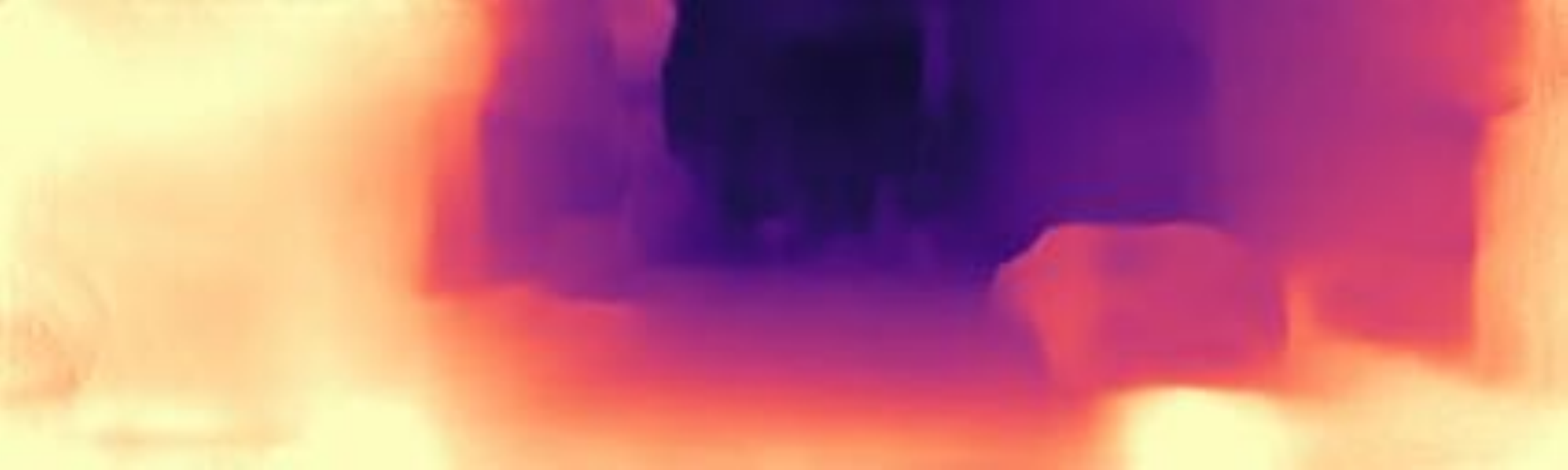}&
    \includegraphics[]{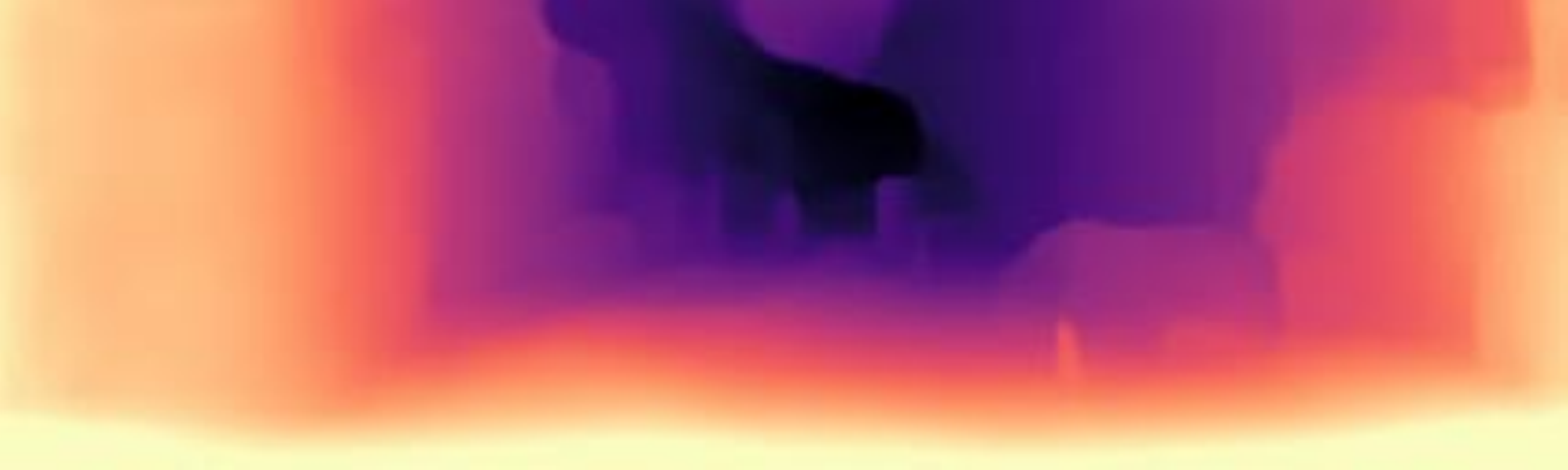}\\
    \includegraphics[]{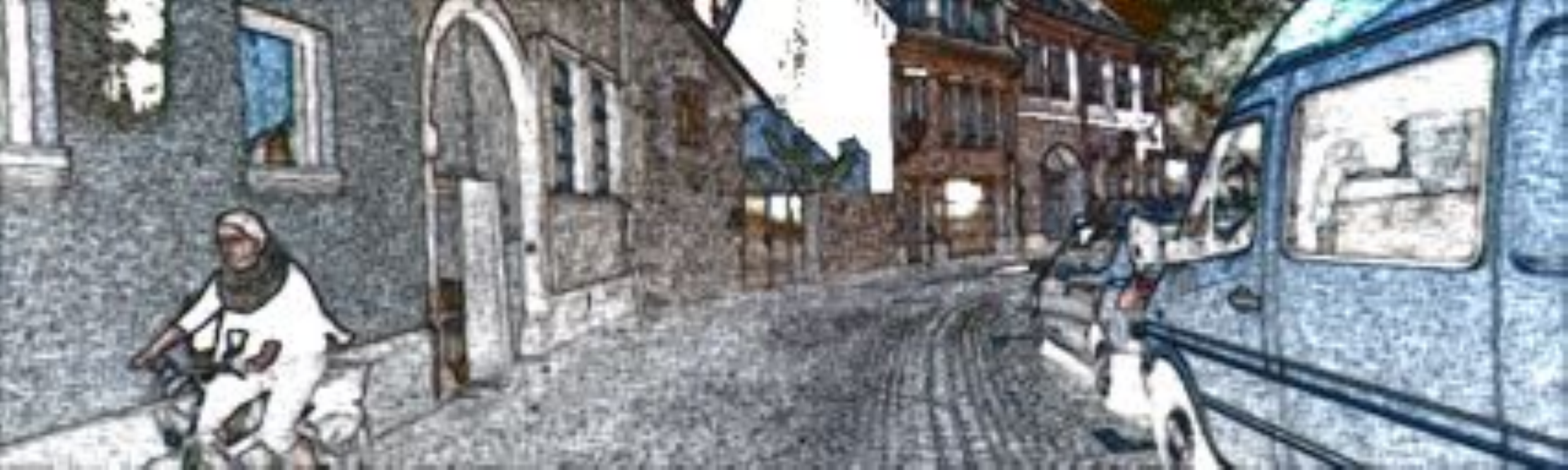}& 
    \includegraphics[]{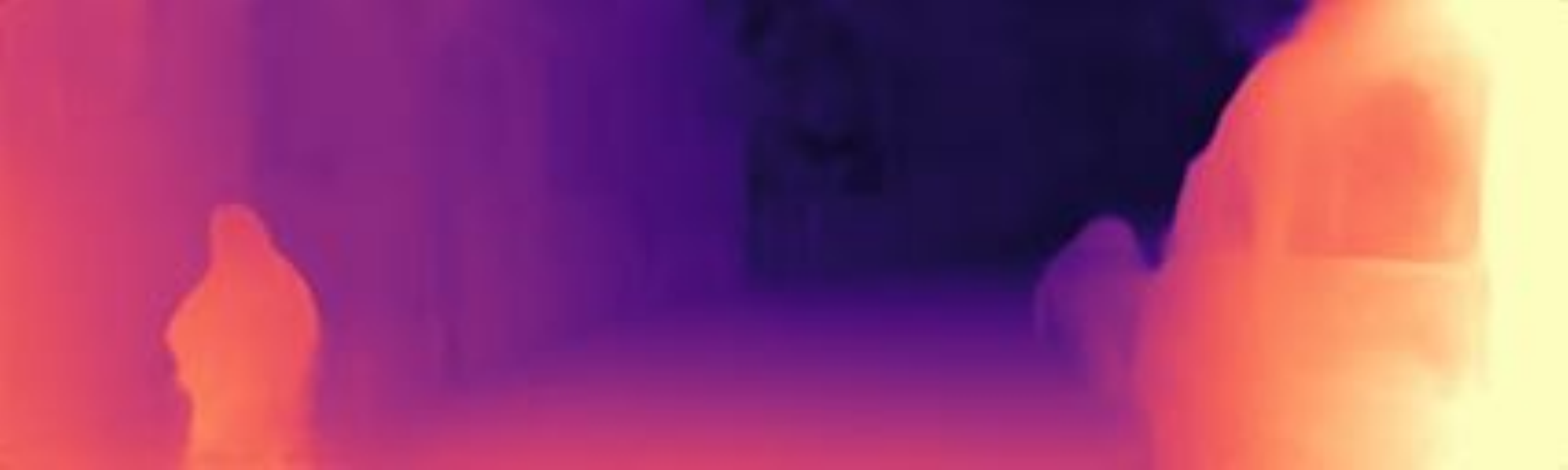}& 
    \includegraphics[]{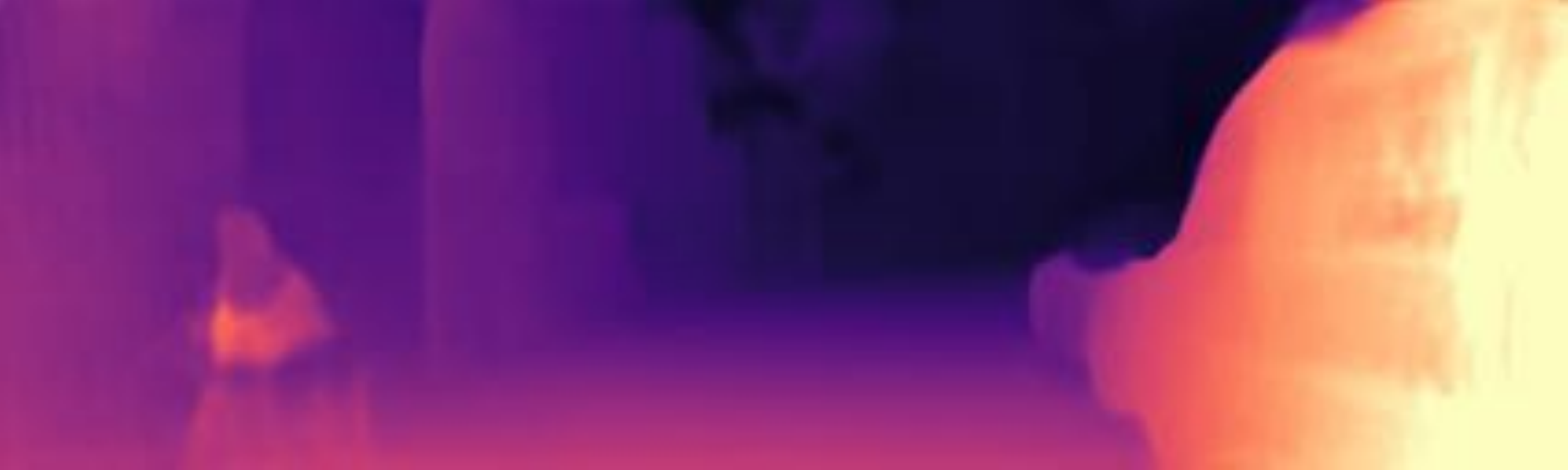}& 
    \includegraphics[]{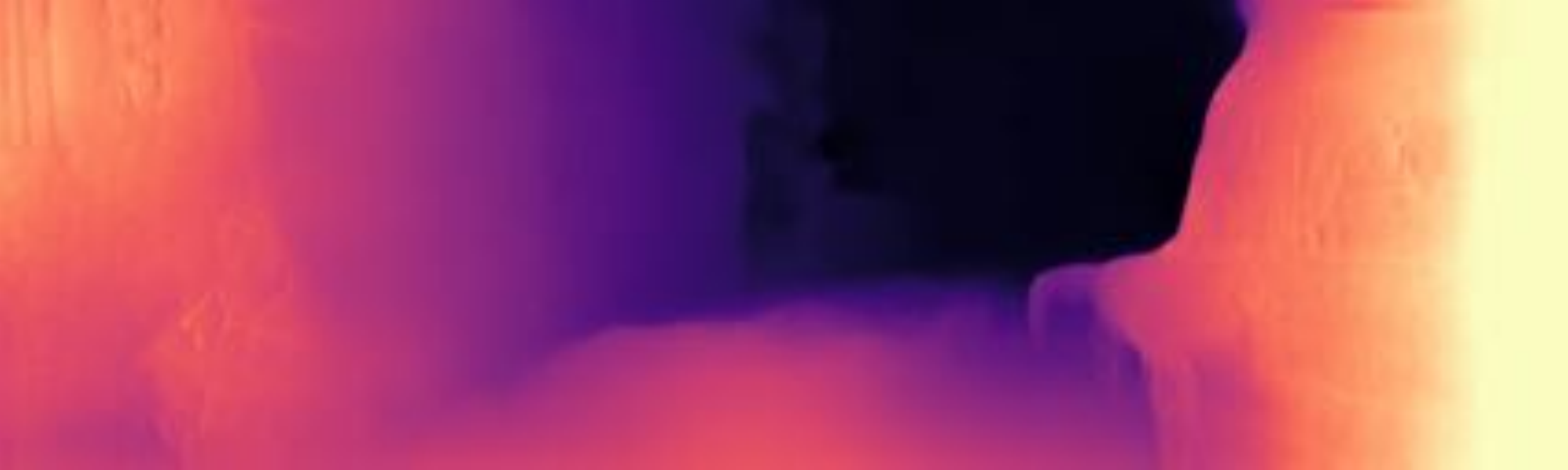}& 
    \includegraphics[]{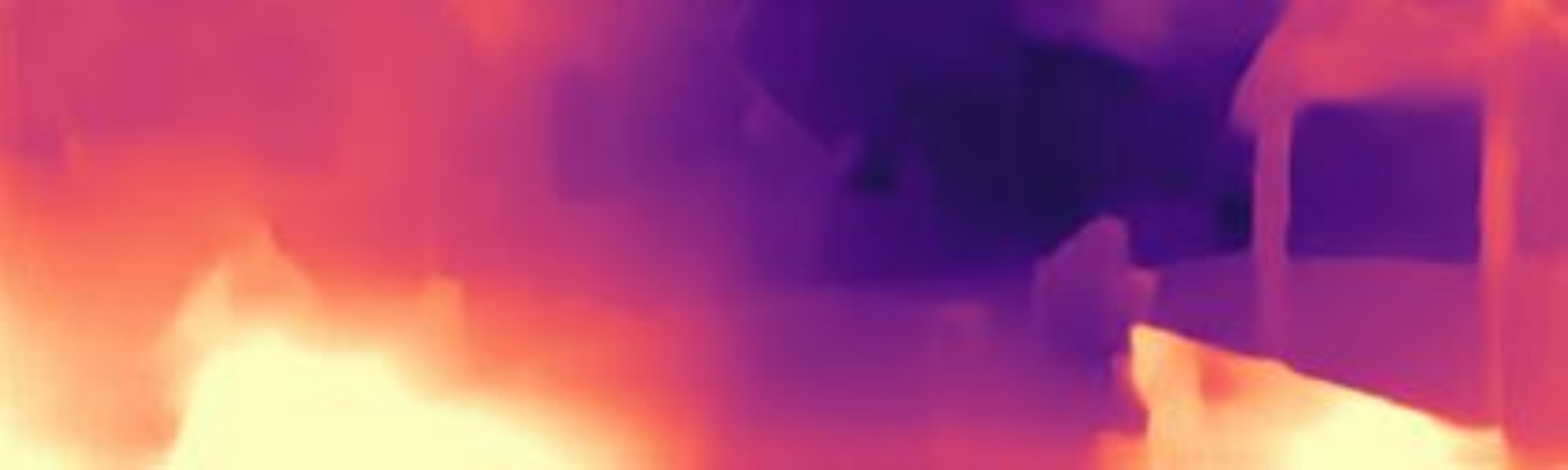}& 
    \includegraphics[]{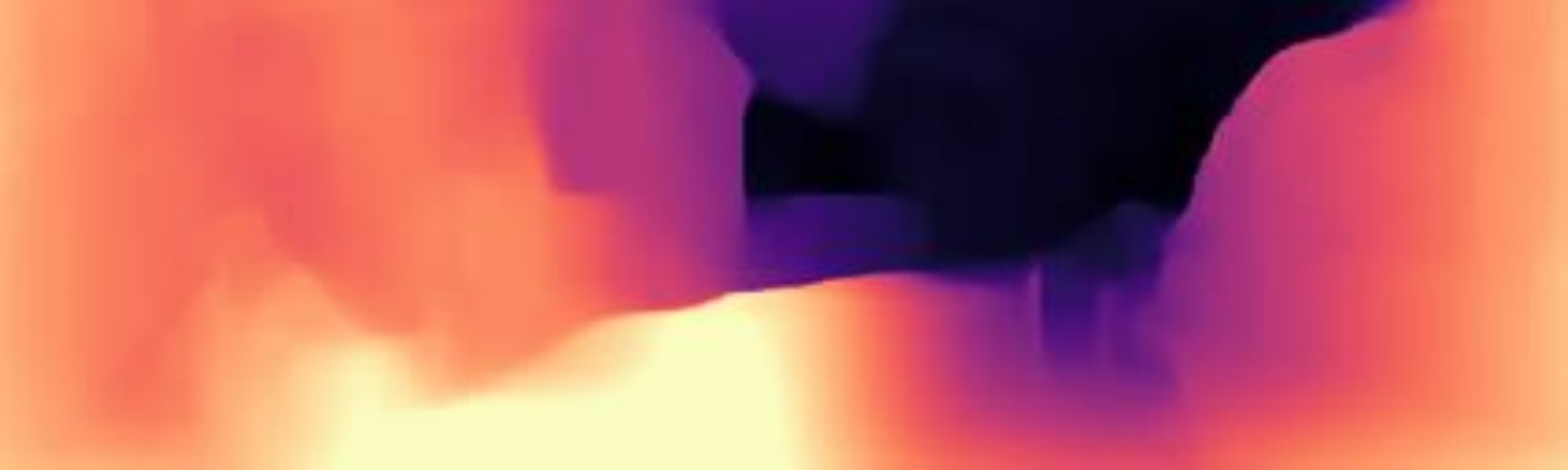}\\
    \includegraphics[]{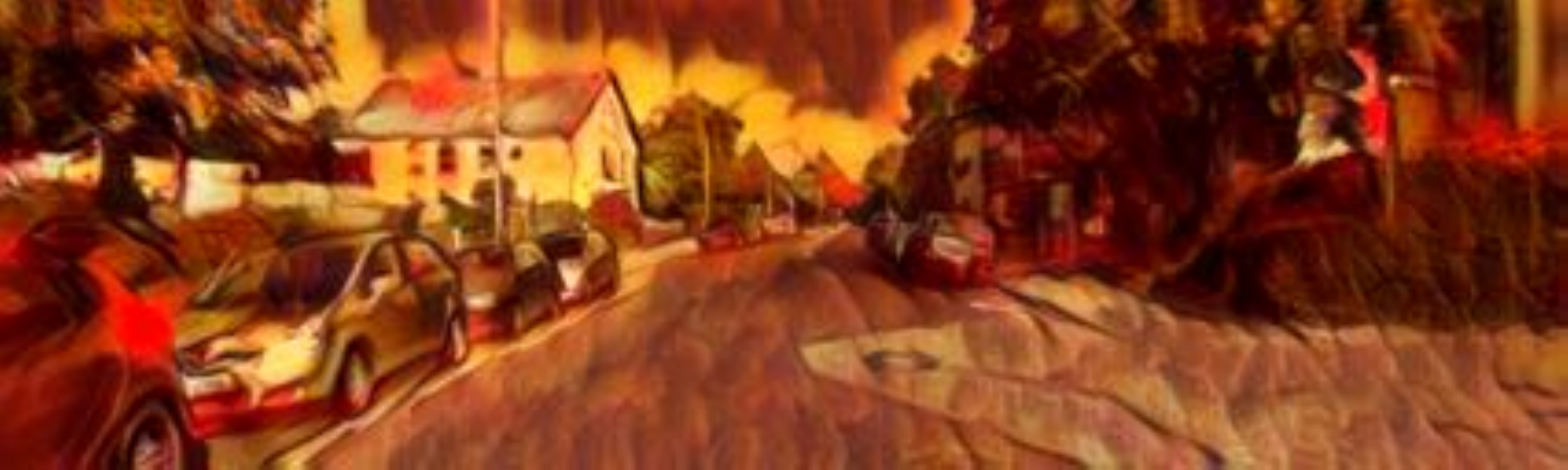}&
    \includegraphics[]{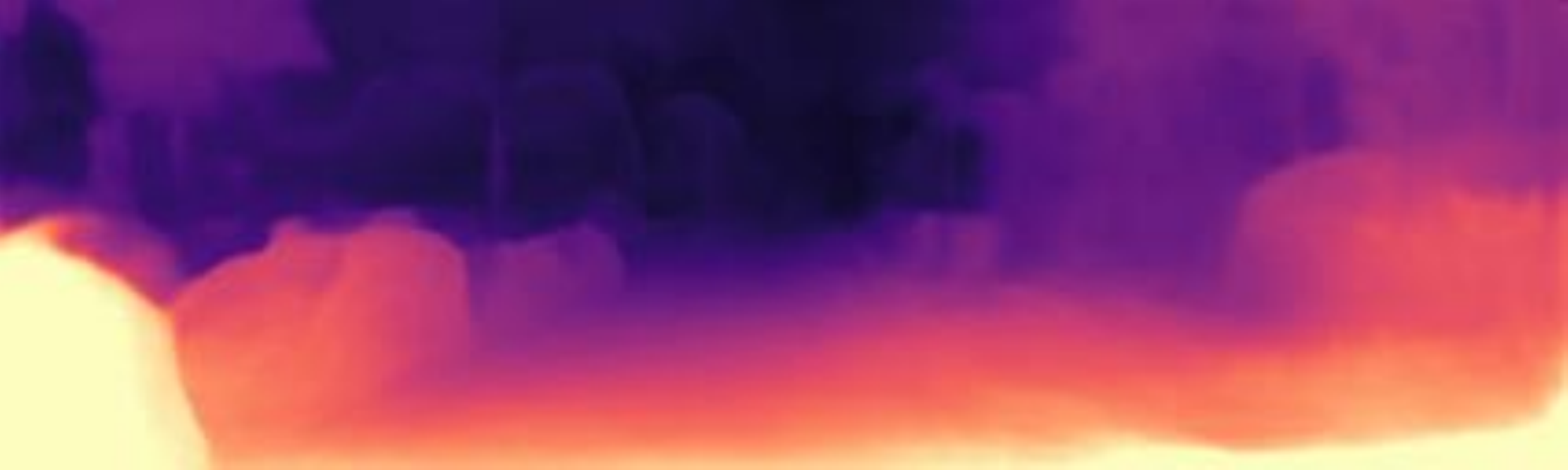}&
    \includegraphics[]{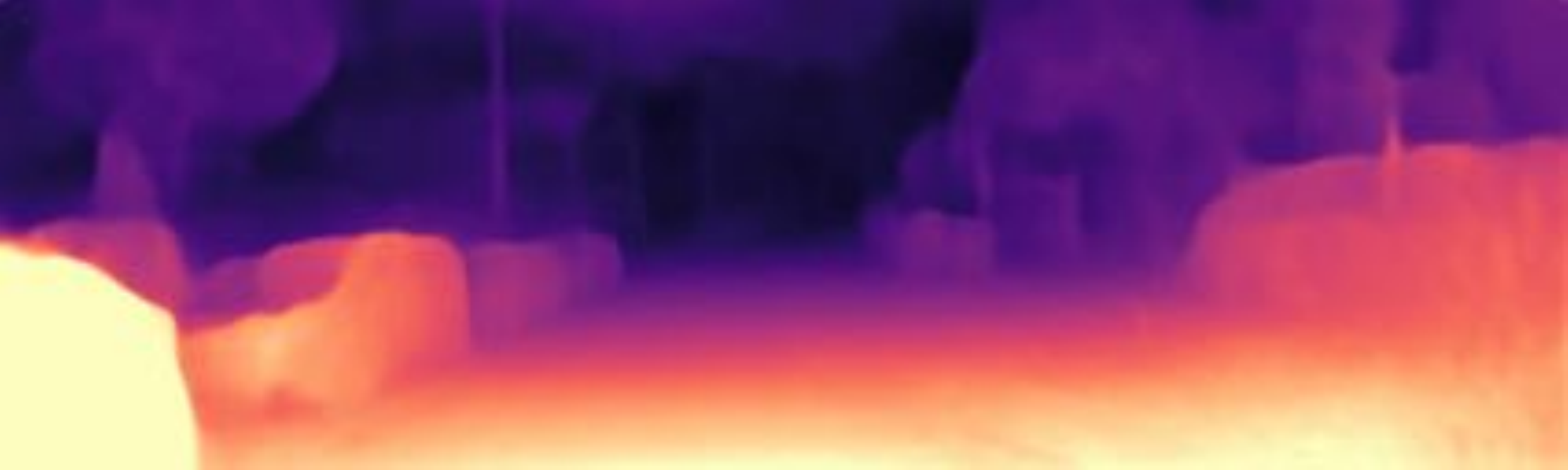}&
    \includegraphics[]{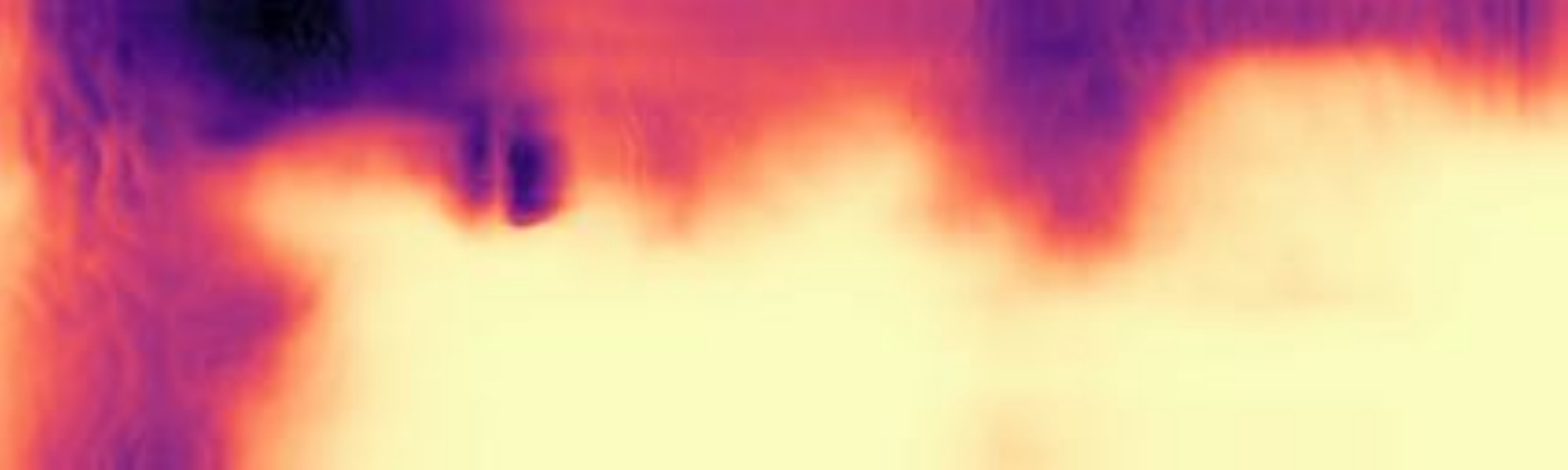}&
    \includegraphics[]{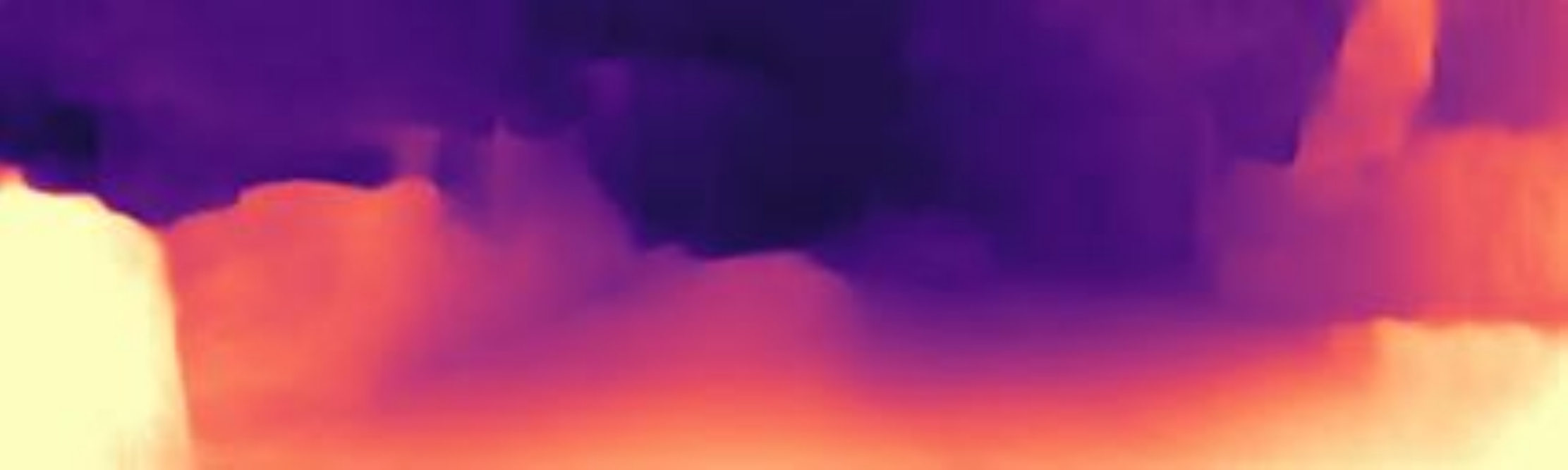}&
    \includegraphics[]{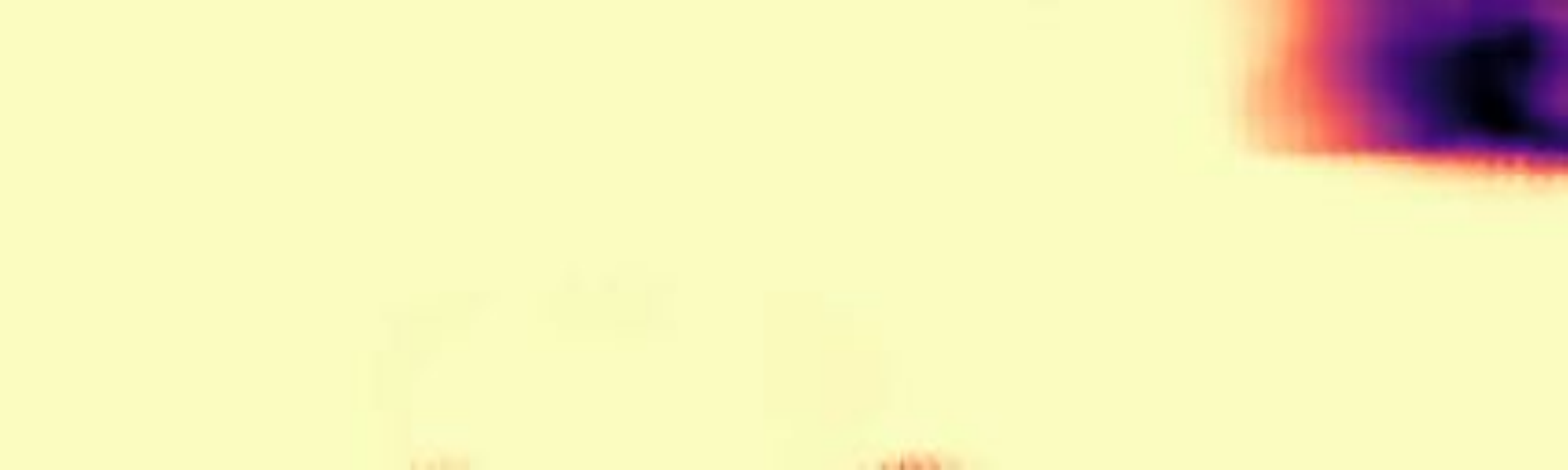}\\
    \includegraphics[]{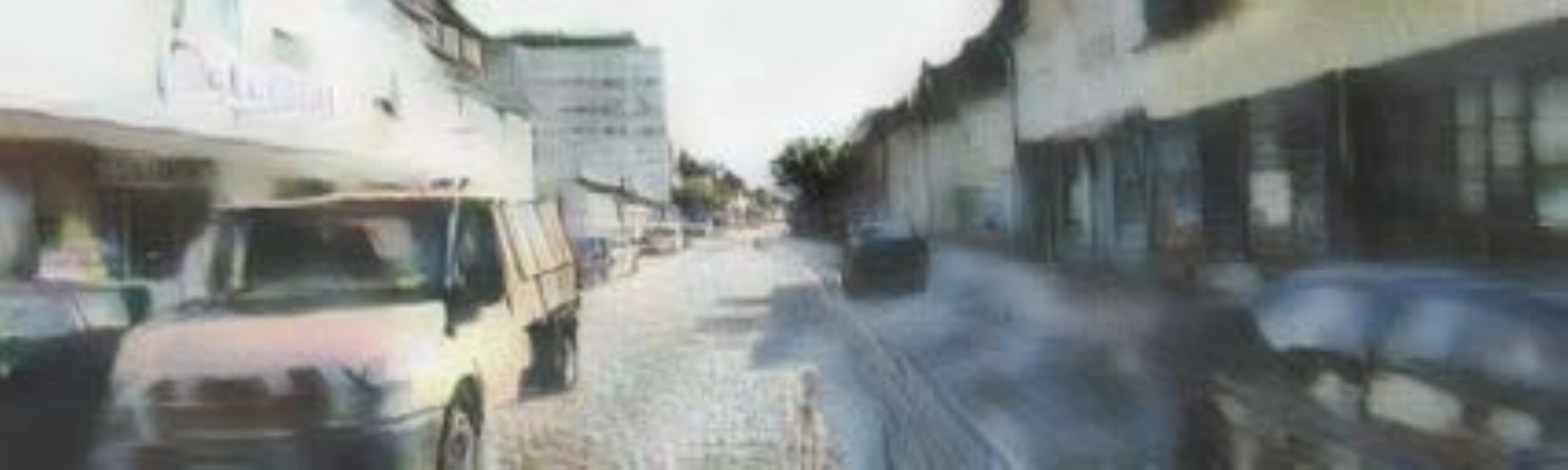}&
    \includegraphics[]{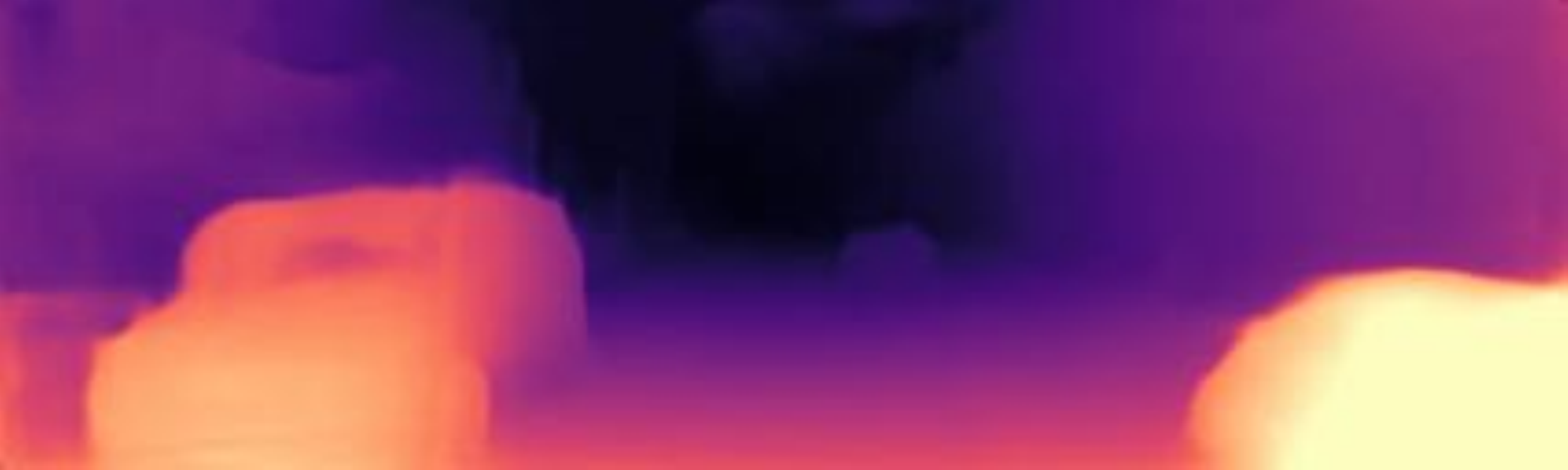}&
    \includegraphics[]{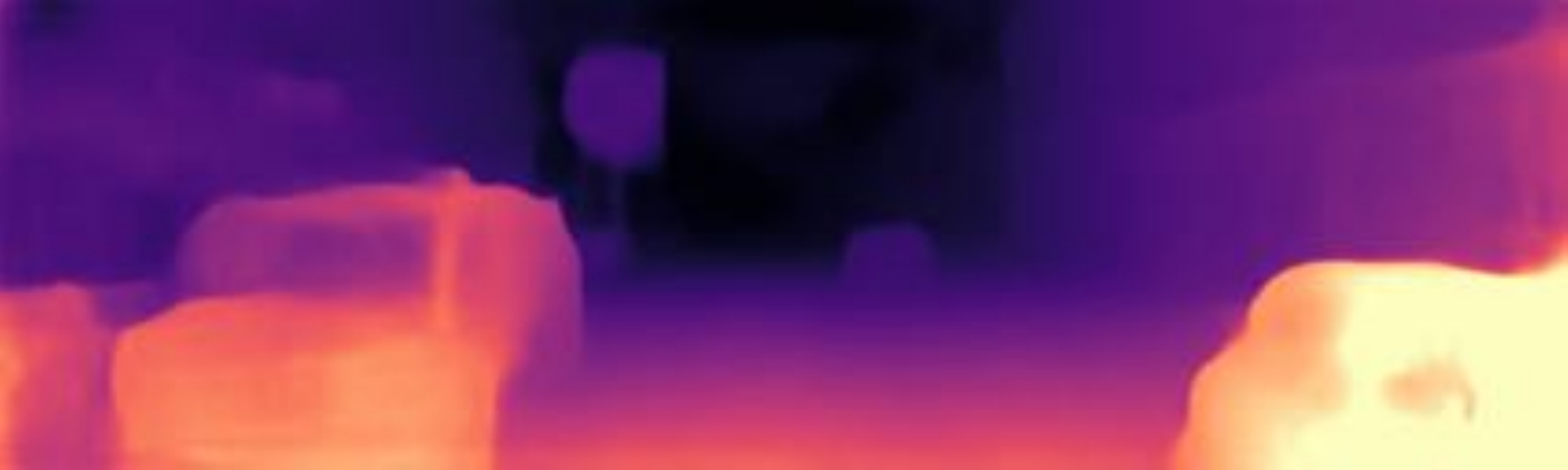}&
    \includegraphics[]{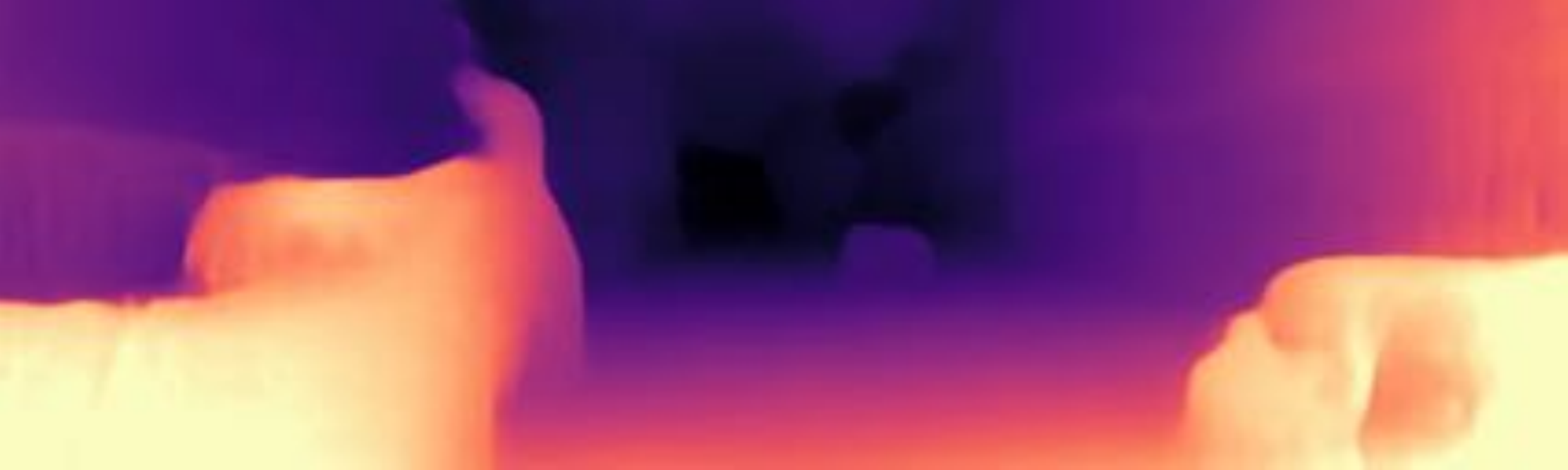}&
    \includegraphics[]{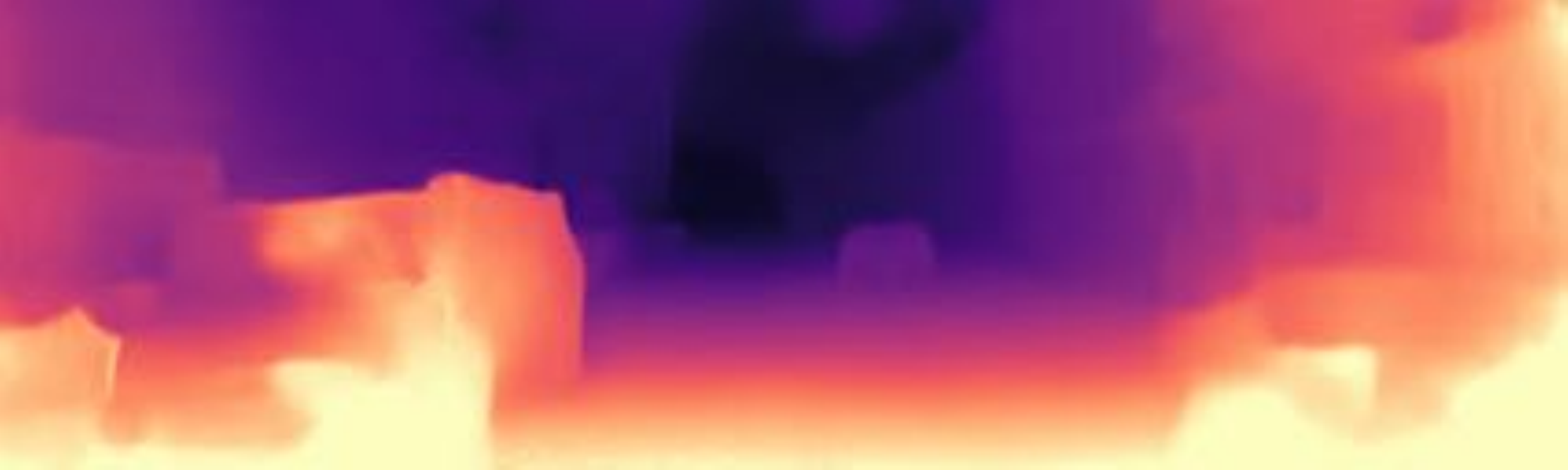}&
    \includegraphics[]{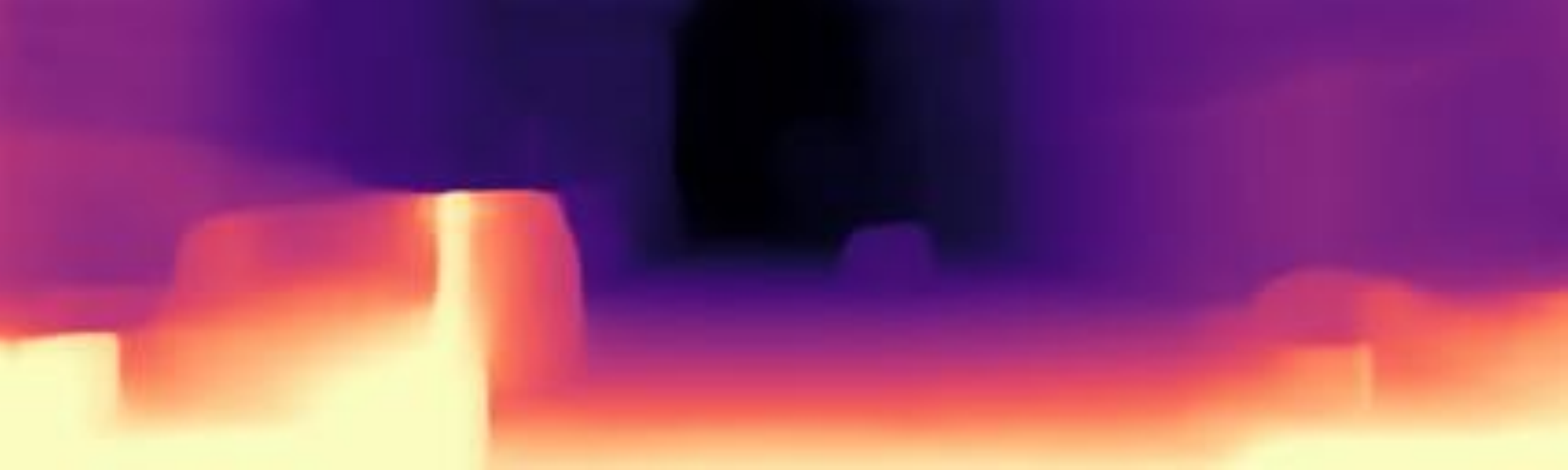}\\
     \fontsize{80}{50} \selectfont Input images & 
     \fontsize{80}{50} \selectfont Ours-Hybrid &
     \fontsize{80}{50} \selectfont Ours-ViT &
     \fontsize{80}{50} \selectfont Monodepth2 &
     \fontsize{80}{50} \selectfont PackNet-SfM & 
     \fontsize{80}{50} \selectfont R-MSFM6
    \end{tabular}}
    \caption{\textbf{Depth map results on texture-shifted datasets.} We test our hybrid/ViT models and the competitive models trained on KITTI using watercolor, pencil-sketch, and style-transfer images (Top to Bottom). Note that the Ours-Hybrid is equivalent to MonoFormer.}
\label{figure_result_texture}
\end{figure*}

\subsection{Attention Connection Module (ACM)}
\label{sec:ACM}
We design a new skip connection method, ACM, which produces the attention of global context and a semantic presentation of the feature given the features $Z_{l},~l\in\{1,...,L\}$.
The skip connection is widely utilized for the dense prediction tasks \cite{ronneberger2015u} because it helps to keep the fine detail by directly transferring the spatial information to the decoder. 
However, it has been observed that in the na\"ive skip connection method, concatenating each feature is too simple to preserve local detail, such as object boundaries \cite{zhou2018unet++}.
To tackle the problem, we introduce an ACM that produces attention weight from the spatial domain and the channel domain inspired by \cite{fu2019dual}. 
It consists of position attention, channel attention modules, and a fusion block that gathers important information from two attentions.
The position attention module produces a position attention map $A^p_{l} \in \mathbb{R}^{C \times N}$ as follows:
\begin{equation}
\small
    A^p_{l} = \text{softmax}({Q}^p_{l} ({K}^p_{l})^\text{T})V^p_{l},
\end{equation}
where ${Q}^p_l, {K}^p_l$ and ${V}^p_l$ are the query, key, and value matrices computed by passing $Z_{l}$ through a single convolutional layer.
The channel attention module directly calculate the channel attention map $ A^c_l \in \mathbb{R}^{C \times N}$ by computing the gram matrix of $Z_l$ as follows: 
\begin{equation}
\small
    A^c_l = \text{softmax}(Z_lZ_l^{\text{T}}).
\end{equation}
The position attention map $A^p_l$ and channel attention map $A^c_l$ enhance the feature representation by capturing long-range context and exploiting the inter-dependencies between each channel map, respectively. 
These two attention maps are utilized in the following section, which highlights the importance of the features.

\subsection{Feature Fusion Decoder (FFD)}
\label{sec:FFD}
The FFD gets the encoder features $Z_l$, the attention maps $A^p_l,~A^c_l$, and the output feature $X_L$ of the last Transformer layer passed through a Residual convolutional layer. 
The decoder fuses the feature $X_{L-l+1},~l \in \{1,...,L\}$ through a single Convolutional layer (Conv) and Channel Normalization (CN) with learnable parameters $\alpha, \beta$ and $\gamma$ as follows:
\begin{equation}
\small
\begin{gathered}
    X_{L-l} = \hat{X}_{L-l}[1+\tanh(\gamma(\text{CN}(\alpha||\hat{X}_{L-l}{||}_2  + \beta )],
    \\
    \hat{X}_{L-l} = \text{Conv}(w_pA^p_lZ_l + w_cA^c_lZ_l + Z_l) + X_{L-l+1},
\end{gathered}
\end{equation}
where $w_{p}$ and $w_c$ are the learnable parameters that determine the importance of the position and channel attentions \cite{zhang2019self}.
The parameter $\alpha$ works so that each channel can learn about each other individually, and $\gamma$ and $\beta$ control the activation channel-wisely following the work in \cite{yang2020gated}. 
Through this process, the FFD is able to produce a depth map from the fused features that preserve local detailed semantic representation while maintaining the global context of features.

\section{Experiments}

\subsection{Comparison on KITTI Datasets}
\label{sec_KITTI}

\begin{table}[t]
    \centering
    \resizebox{\columnwidth}{!}{
        \begin{tabular}{ccccccccc}
        \hline
        \multirow{2}{*}{Model} & \multirow{2}{*}{Backbone} & \multicolumn{4}{c}{Lower is better $\downarrow$}& \multicolumn{3}{c}{Higher is better $\uparrow$}\\ \cline{3-9} 
        && Abs Rel& Sq Rel& RMSE& RMSElog& $\delta<1.25$  & $\delta<1.25^2$& $\delta<1.25^3$ \\ \hline
        Monodepth& CNN& 0.148& 1.344& 5.972& 0.216& 0.816& 0.941& 0.976\\
        Monodepth2& CNN& 0.115& 0.903& 4.863& 0.193& 0.877& 0.959& 0.981\\
        PackNet-Sfm& CNN& 0.111& \textbf{0.785} & 4.601& 0.189& 0.878& 0.960& \textbf{0.982}\\
        SGDepth& CNN& 0.117& 0.907& 4.844& 0.196& 0.875& 0.958& 0.980\\
        R-MSFM3& CNN& 0.114& 0.815& 4.712& 0.193& 0.876& 0.959& 0.981\\
        R-MSFM6& CNN& 0.112& 0.806& 4.704& 0.191& 0.878& 0.960& 0.981\\ \hline
        Ours-ViT& Transformer& 0.118& 0.942& 4.840& 0.193& 0.873& 0.956& 0.981\\
        Ours-Hybird& CNN-Transformer& \textbf{0.104} & 0.846& \textbf{4.580} & \textbf{0.183} & \textbf{0.891} & \textbf{0.962}& \textbf{0.982}\\ \hline
        \end{tabular}}
        \caption{\textbf{Quantitative comparison to state-of-the-arts.} We evaluate models trained on KITTI (K) with an input image size of $640 \times 192$.  We only use monocular images (M) for supervision.  \textbf{Bold} is the best performance.}
        \label{table_resutl_kitti}
\end{table}

We compare our method with state-of-the-art methods, SGCDepth {\cite{xiong2021self}}, GeoNet {\cite{yin2018geonet}}, Struct2depth {\cite{casser2019depth}}, Monodepth2 {\cite{godard2019digging}}, PackNet-SfM {\cite{guizilini20203d}}, SGDepth {\cite{klingner2020self}}, R-MSFM {\cite{zhou2021r}} in \tabref{table_resutl_kitti}.
We use the KITTI Eigen split \cite{geiger2013vision,eigen2015predicting} consisting of 39,810 training, and 4,424 validation and 697 test data. 
We additionally sample data about 5\% of the total data with infinite-depth problems that mostly occur in dynamic scenes, following the work \cite{guizilini2020semantically}. 
We use typical error and accuracy metrics for depth, absolute relative (Abs Rel), square relative (Sq Rel), root-mean-square-error (RMSE), its log (RMSElog), and the ratio of inliers following the work \cite{guizilini20203d}.
The quantitative results show that the proposed method outperforms other models. 
The qualitative results in \figref{figure_result_kitti} show that our method precisely preserves object boundaries. This demonstrates that the encoder captures both global context and informative local features and transfers them to the decoder for the pixel-wise prediction.

\begin{figure*}[t]
     \newcommand\w{10cm}
    \newcommand\h{5cm}
    \begin{subfigure}
  \centering
    \resizebox{\linewidth}{!}{
    \begin{tabular}{cccccc}
    \includegraphics[]{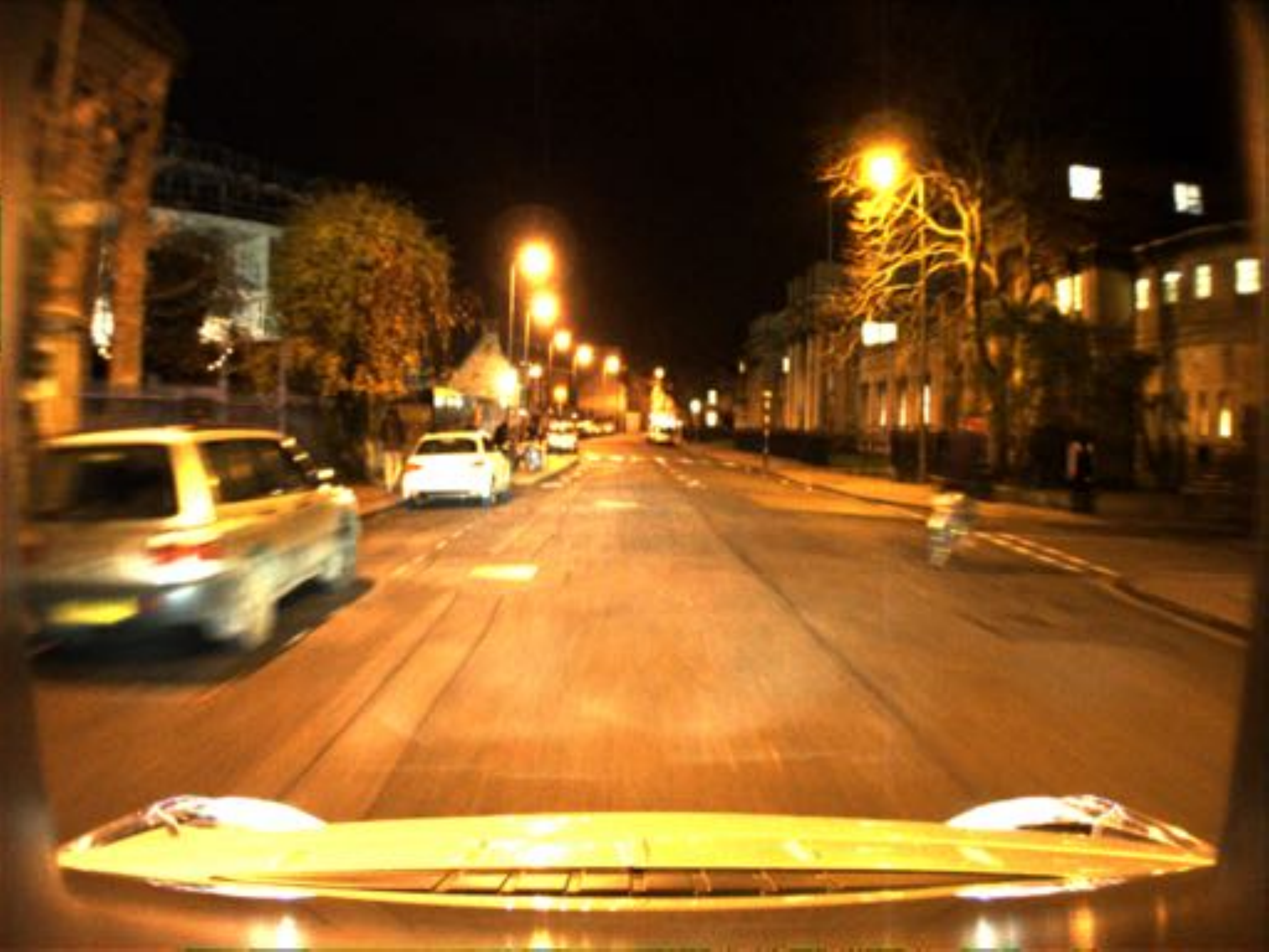}& 
    \includegraphics[]{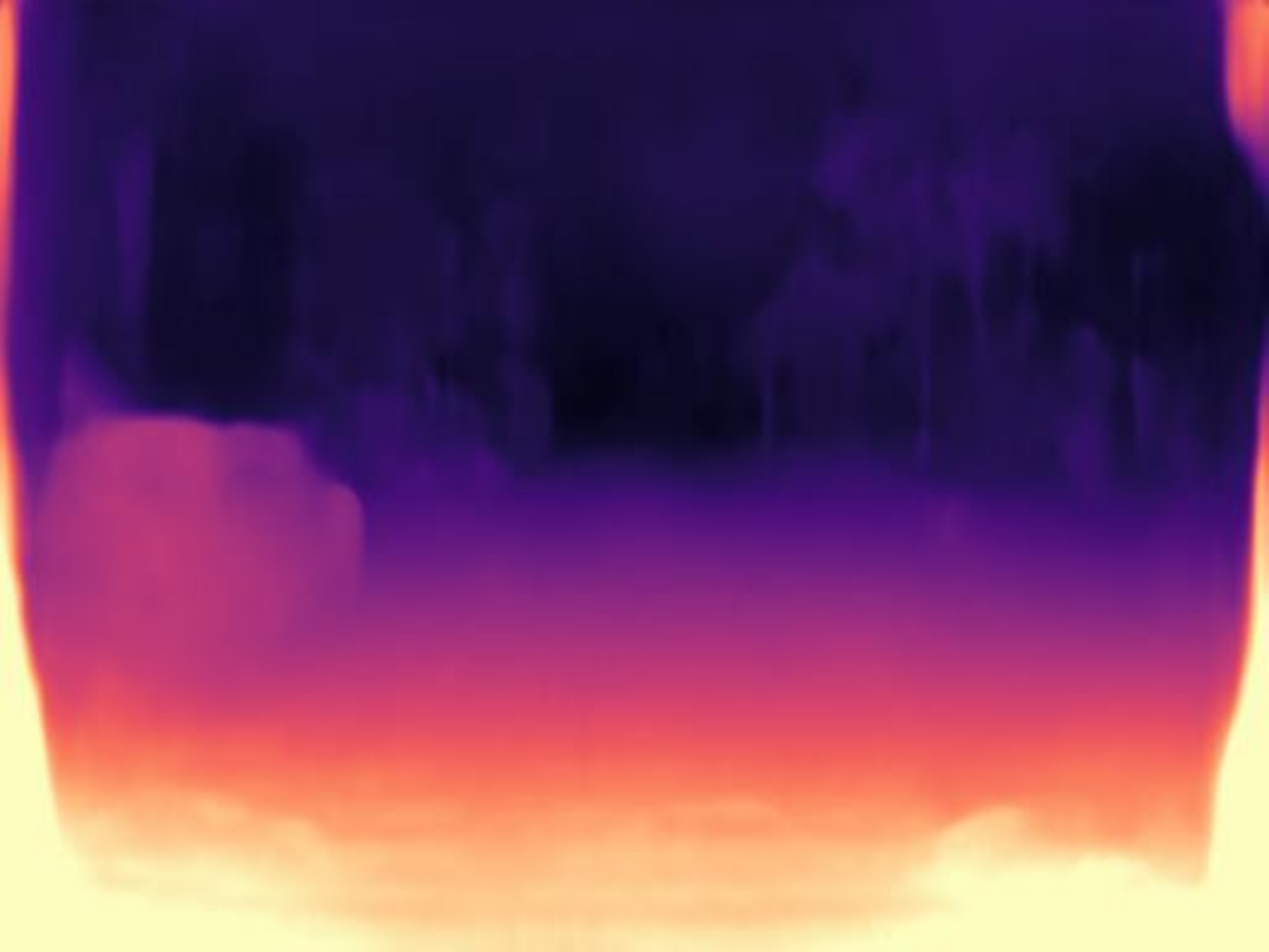}& 
    \includegraphics[]{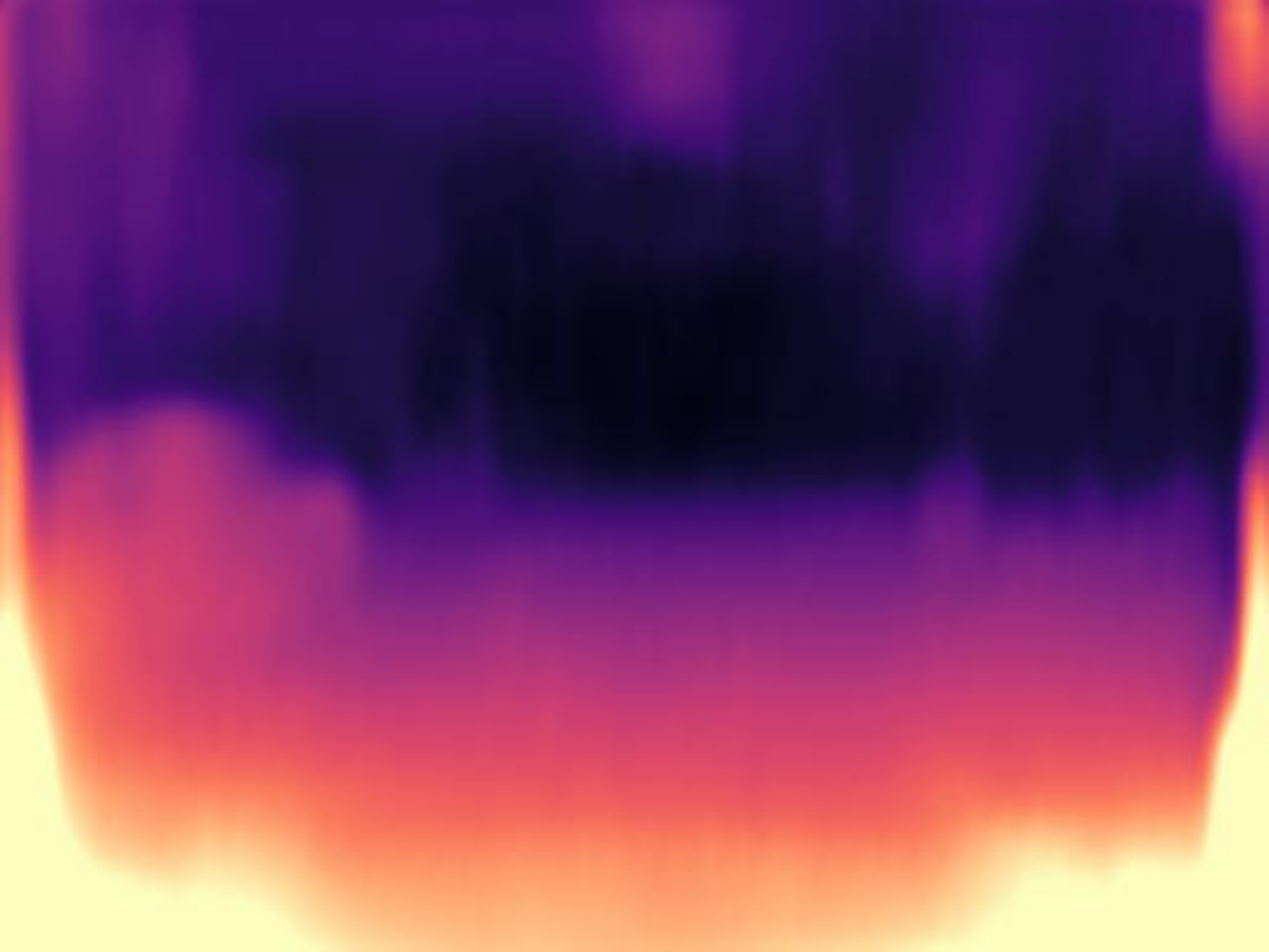}& 
    \includegraphics[]{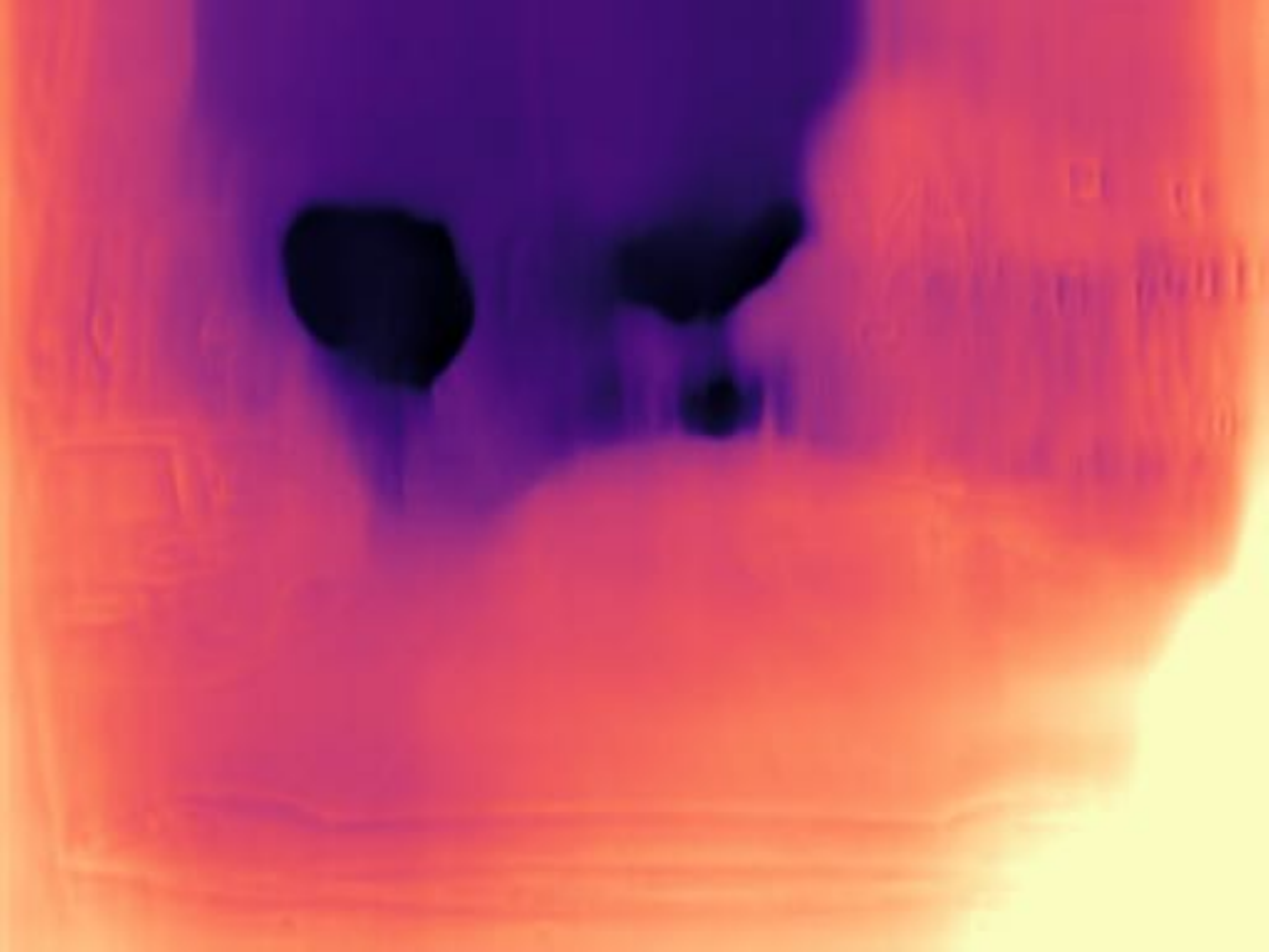}& 
    \includegraphics[]{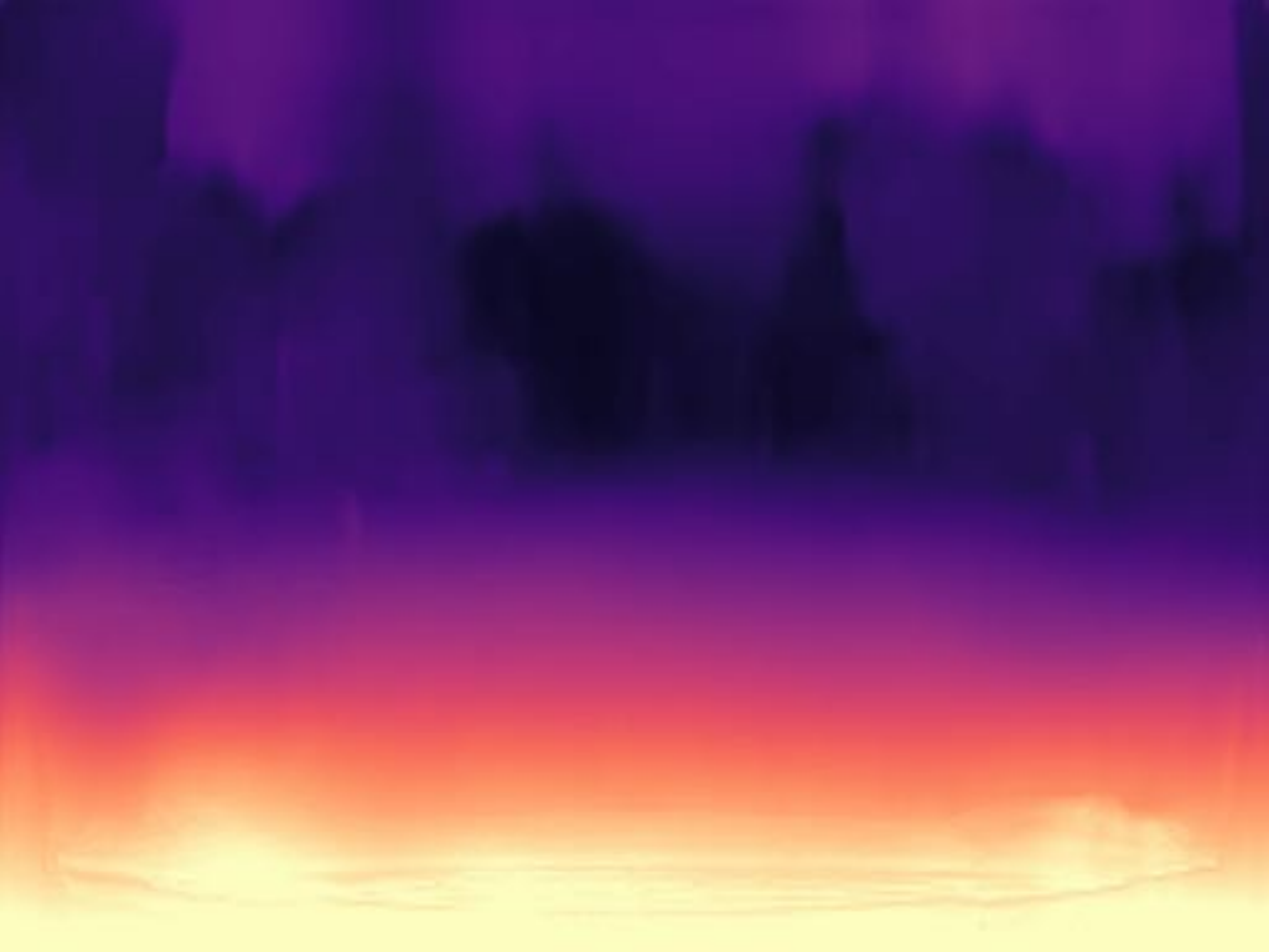}& 
    \includegraphics[]{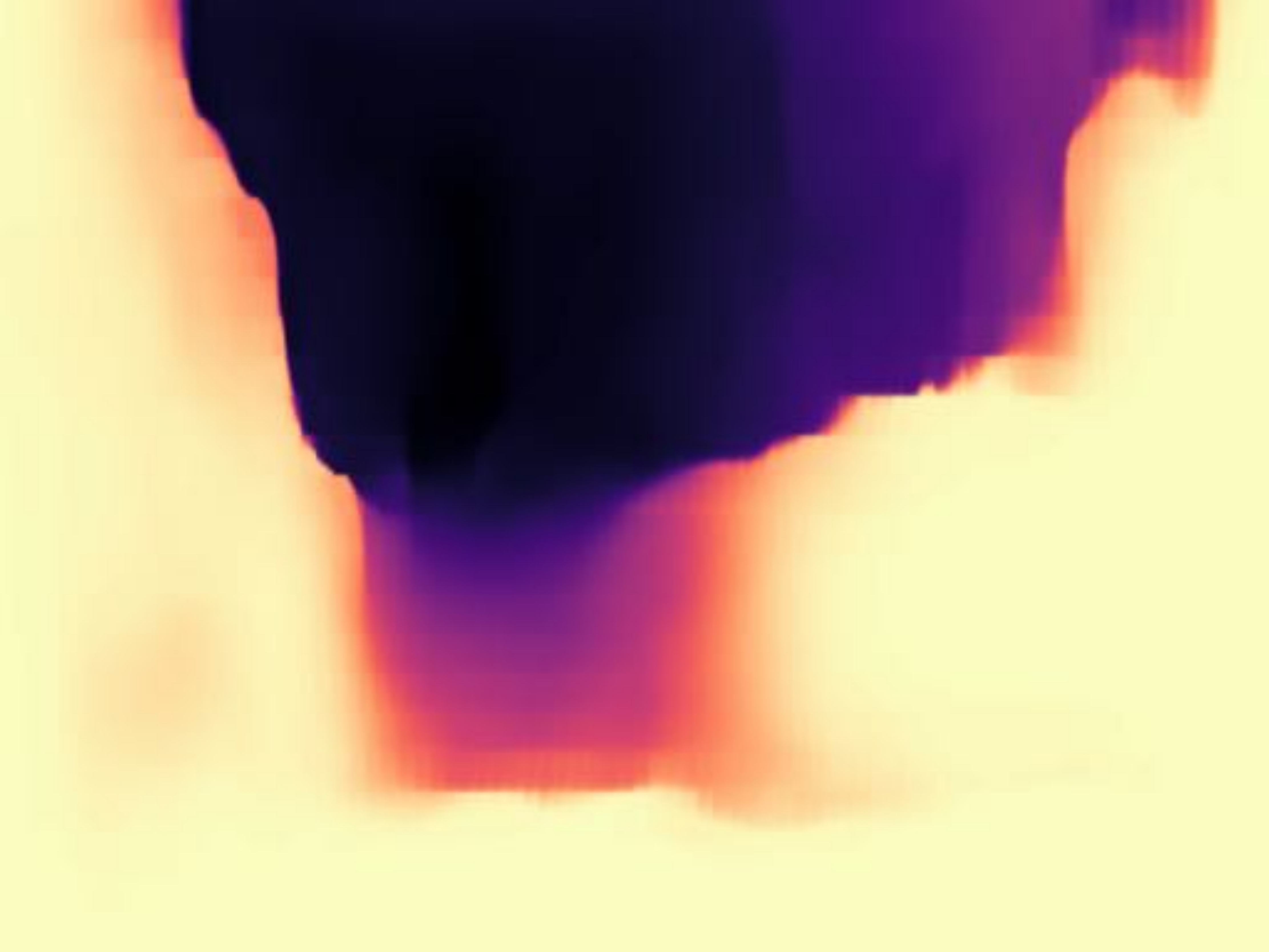} \\
     \fontsize{150}{100} \selectfont Input images & 
    \fontsize{150}{100} \selectfont Ours-Hybrid & 
    \fontsize{150}{100} \selectfont Ours-ViT &
     \fontsize{150}{100} \selectfont Monodepth2 &
     \fontsize{150}{100} \selectfont PackNet-SfM & 
     \fontsize{150}{100} \selectfont R-MSFM6
    \end{tabular}}
    \caption{\textbf{Depth Results on Oxford Robotcar Night time Dataset.}}
\label{figure_result_night}
     \end{subfigure}
     \begin{subfigure}
    \centering
    \resizebox{\linewidth}{!}{
    \begin{tabular}{ccccccc}
        \includegraphics[width=\w,height=\h]{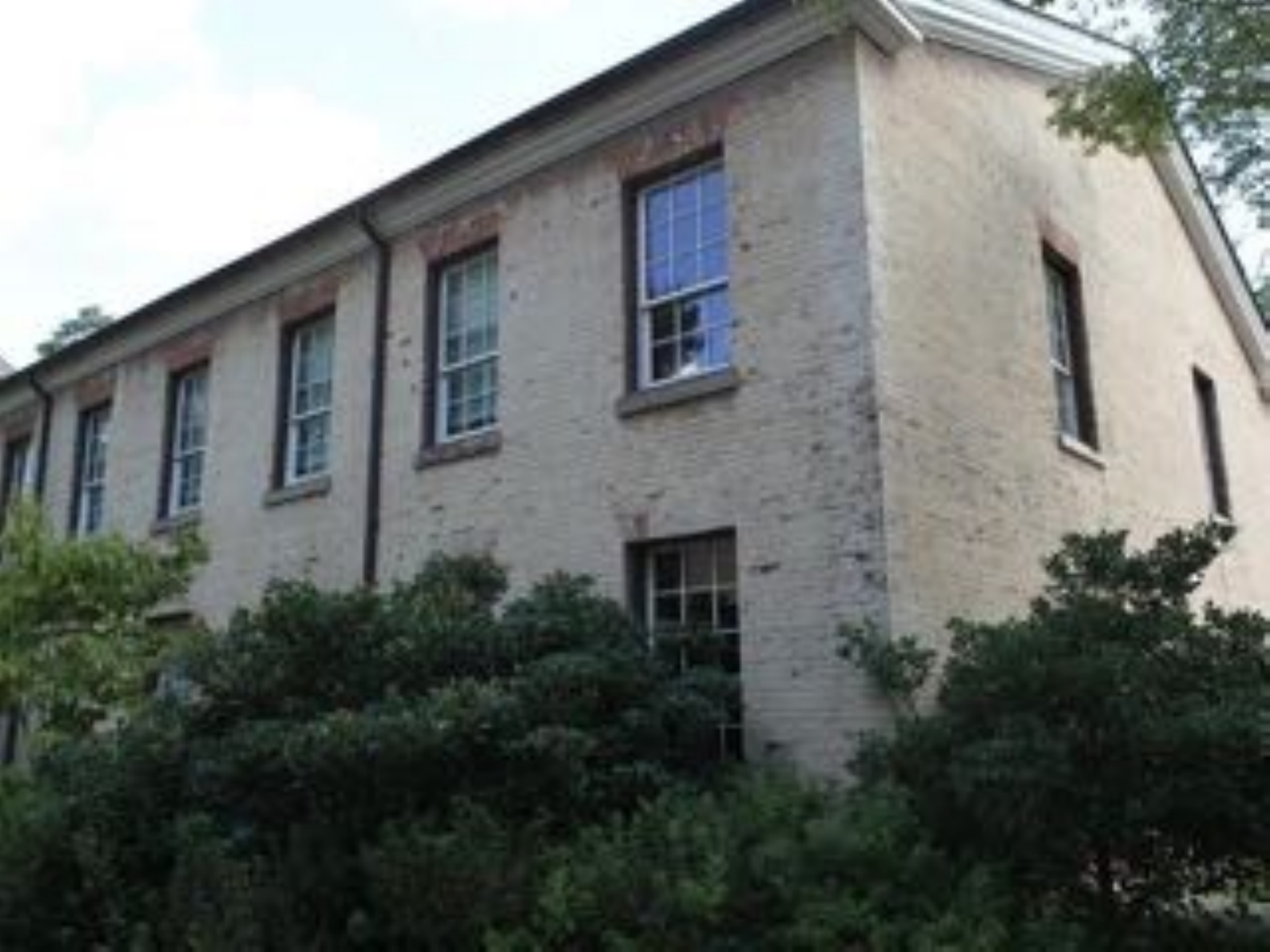}& 
        \includegraphics[width=\w,height=\h]{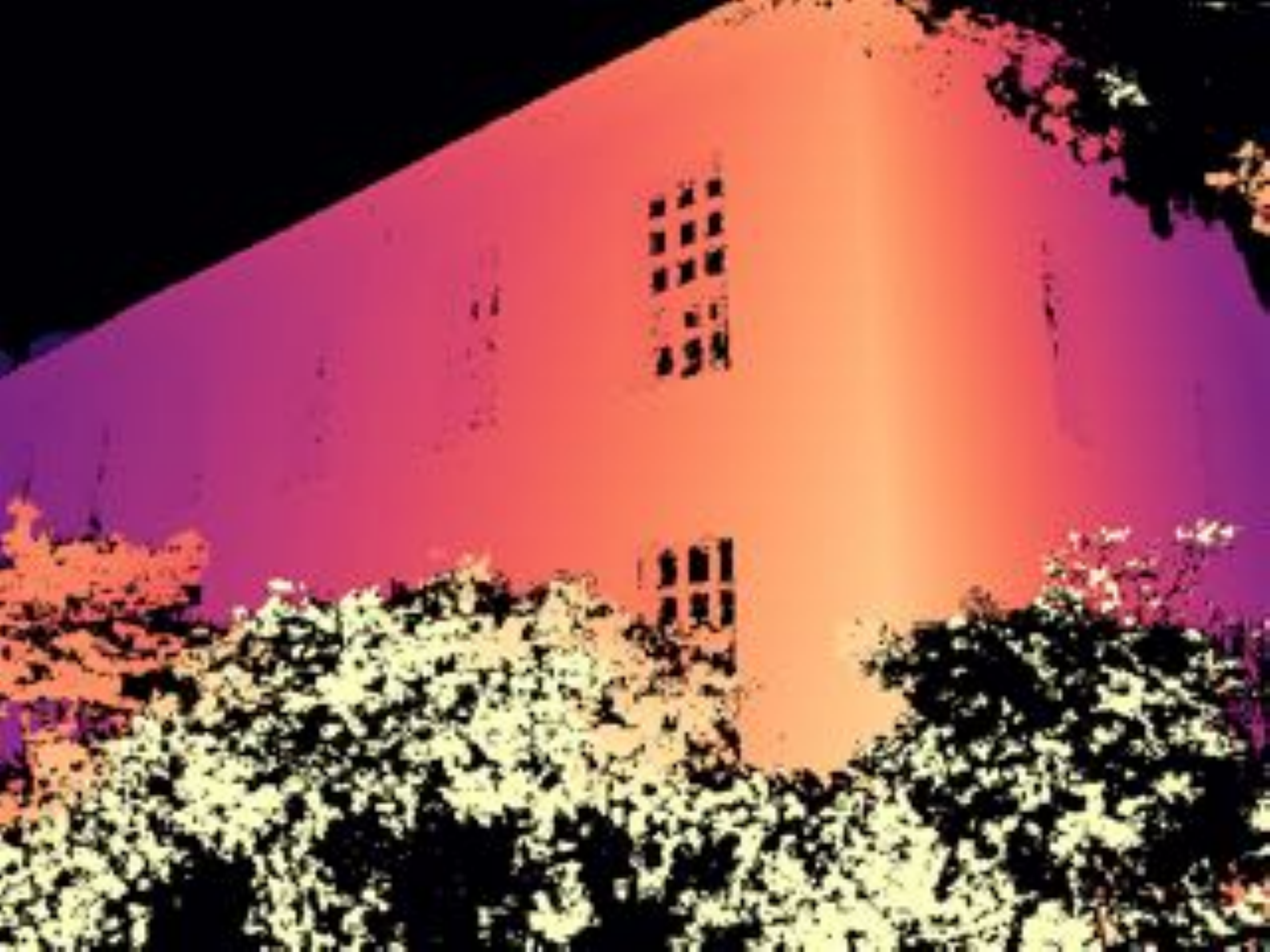}& 
        \includegraphics[width=\w,height=\h]{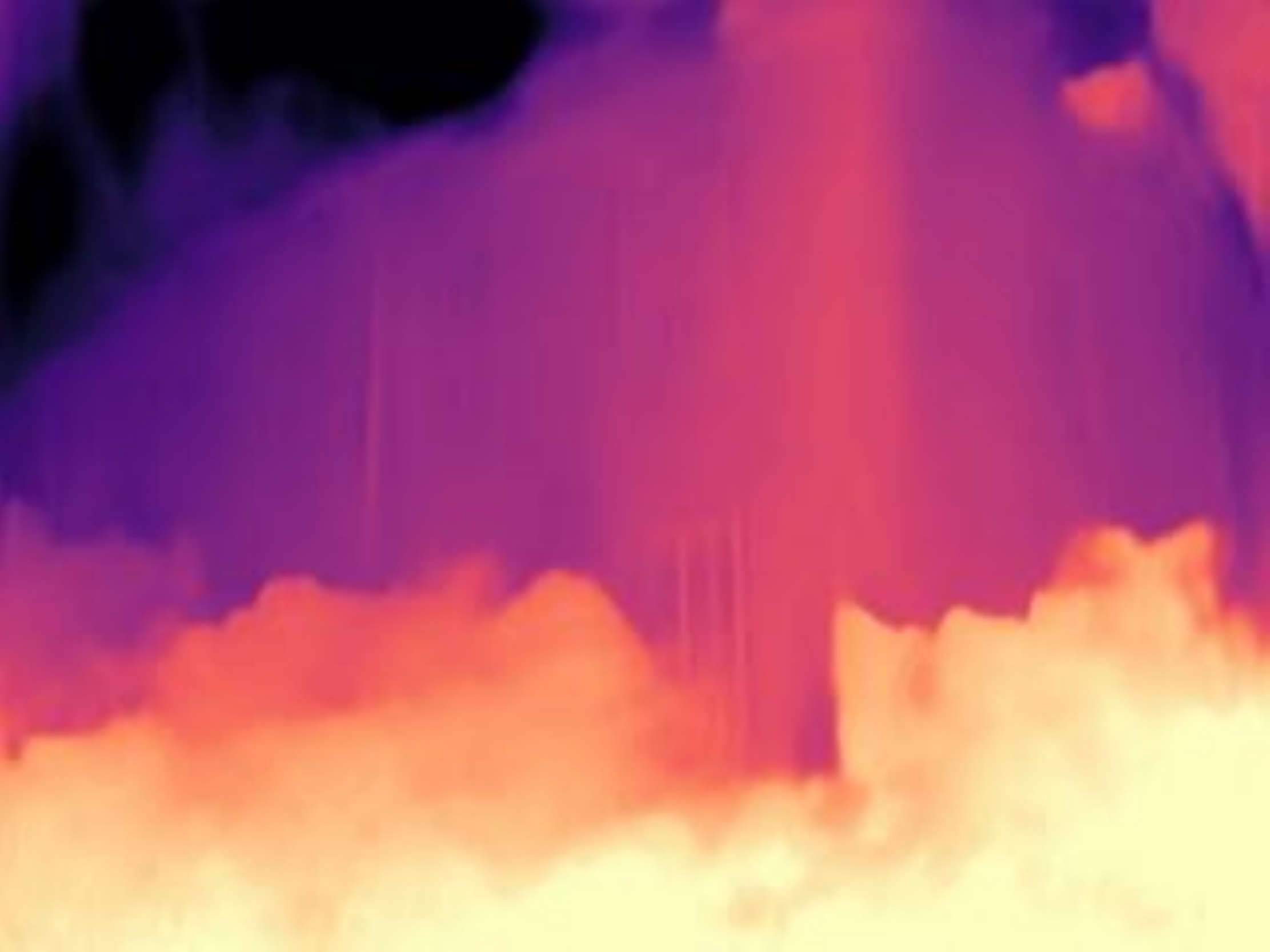}& 
        \includegraphics[width=\w,height=\h]{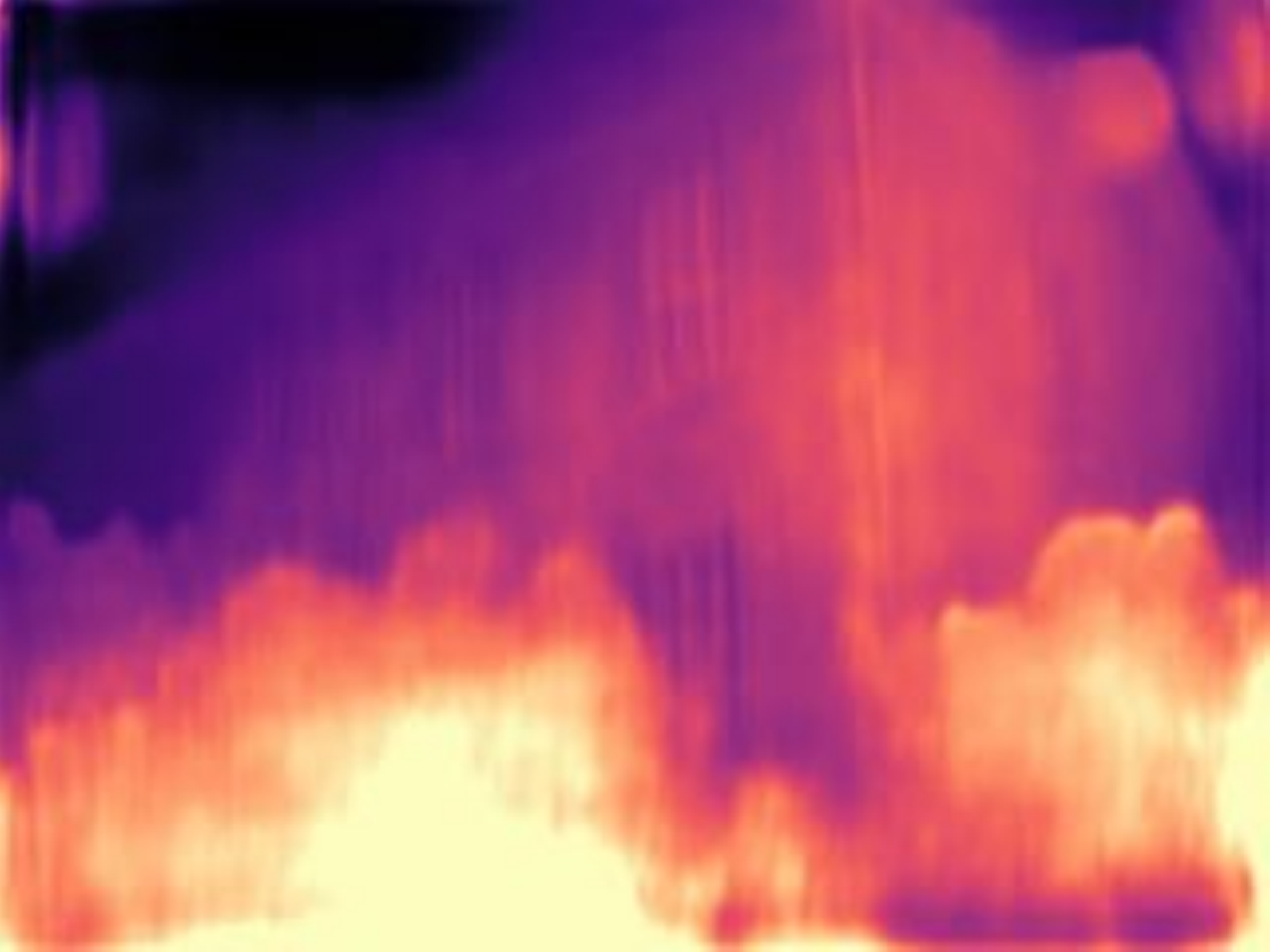}& 
        \includegraphics[width=\w,height=\h]{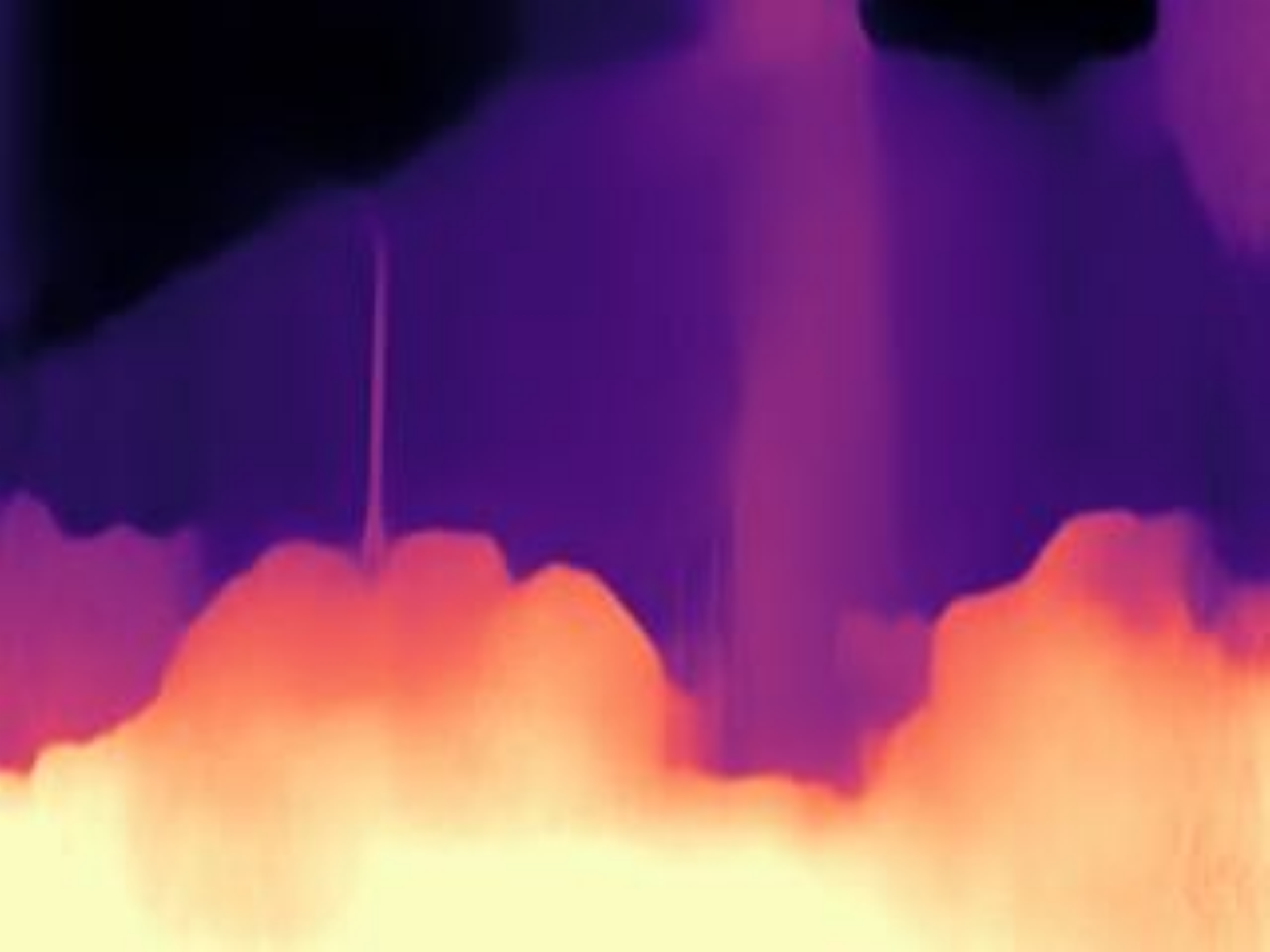}& 
        \includegraphics[width=\w,height=\h]{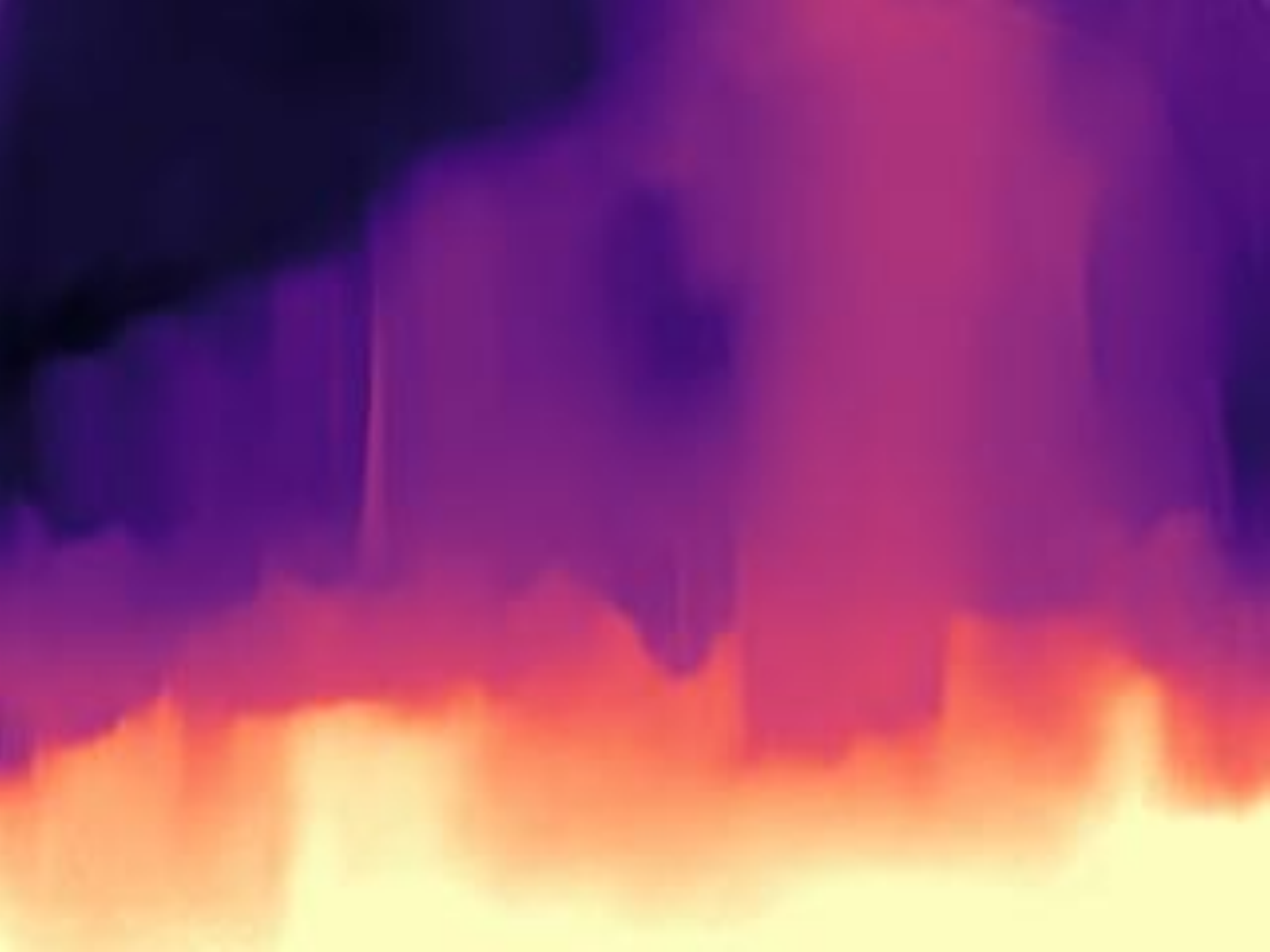}& 
        \includegraphics[width=\w,height=\h]{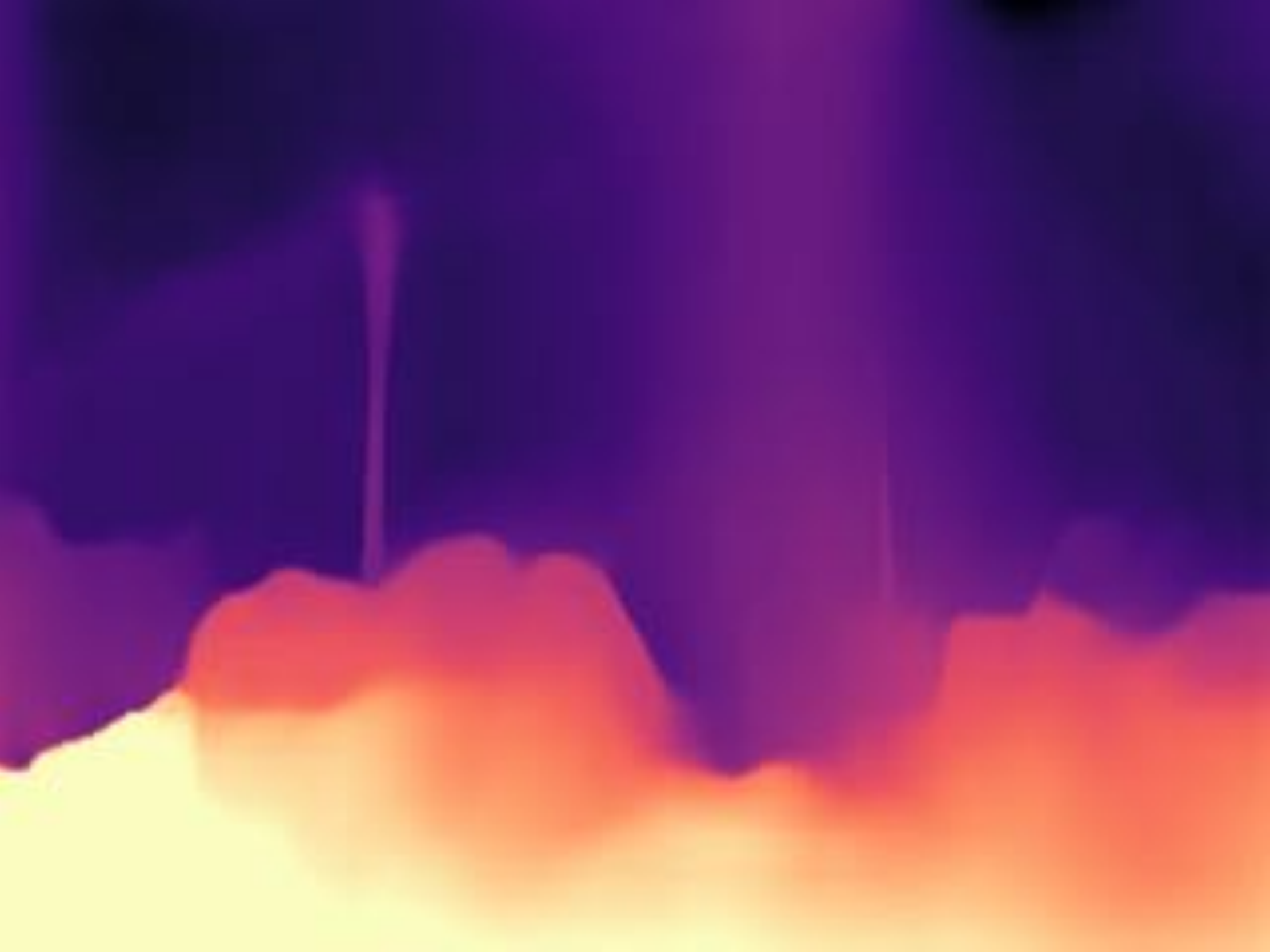}\\ 
        \includegraphics[width=\w,height=\h]{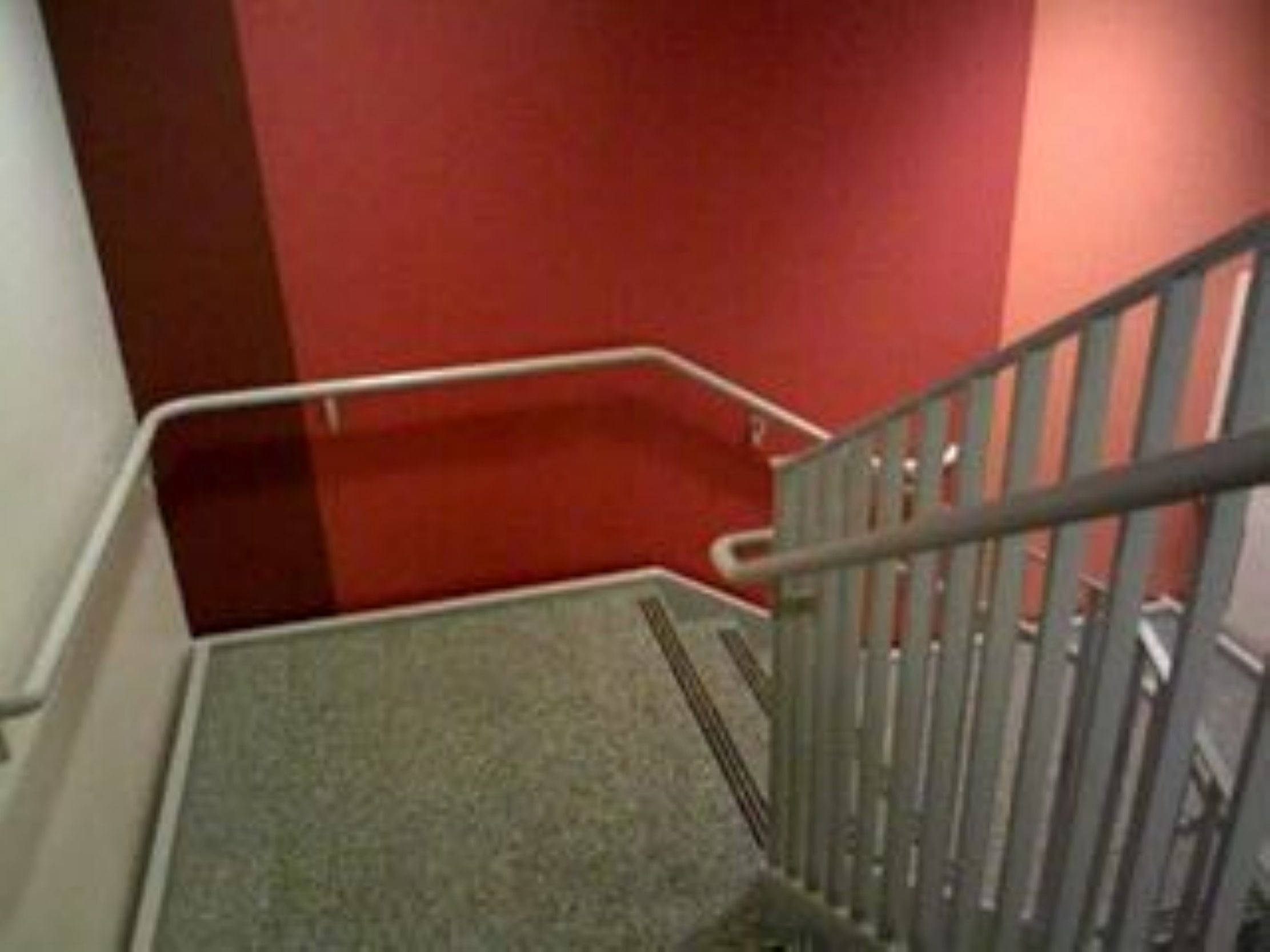}& 
        \includegraphics[width=\w,height=\h]{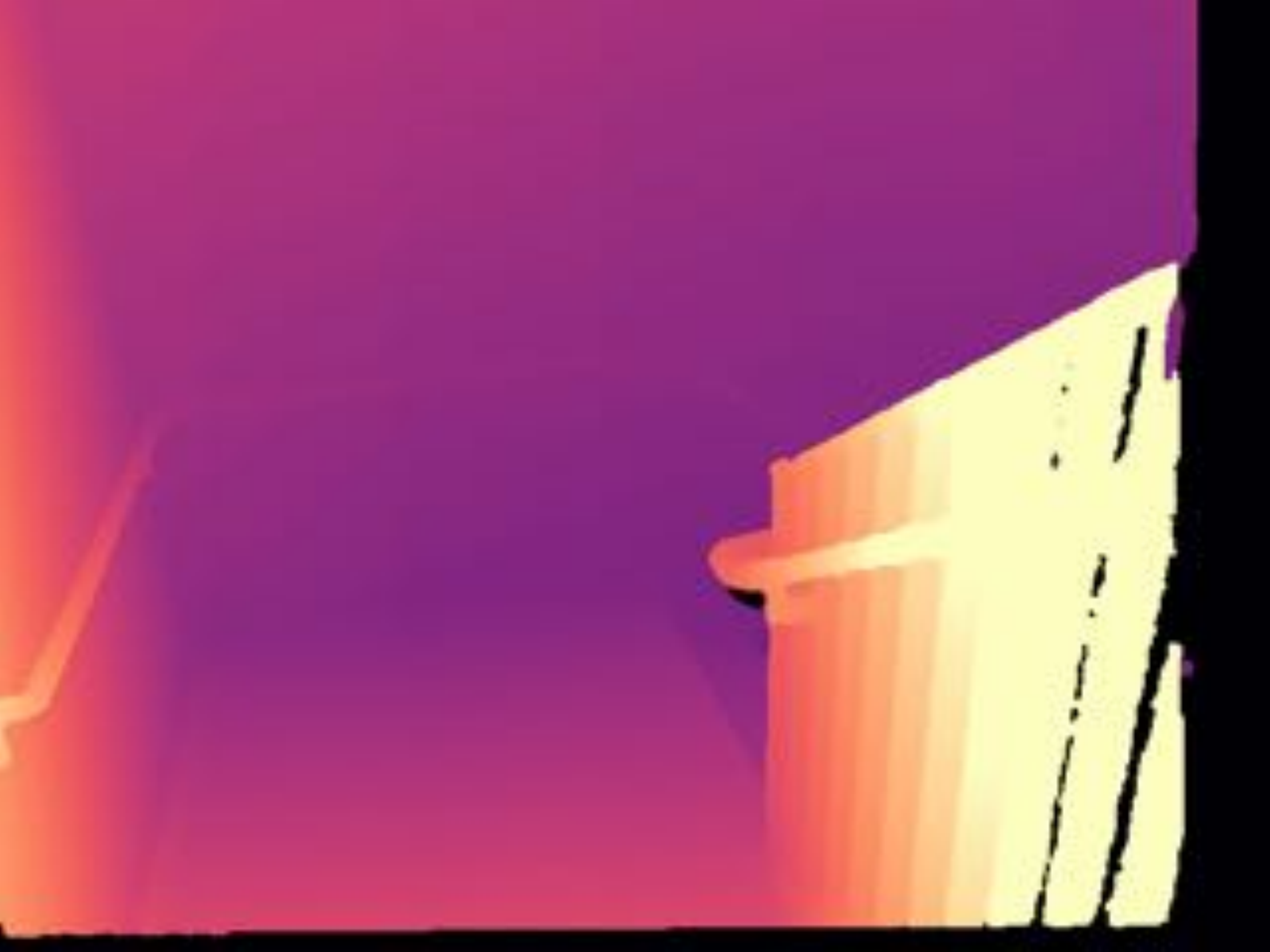}& 
        \includegraphics[width=\w,height=\h]{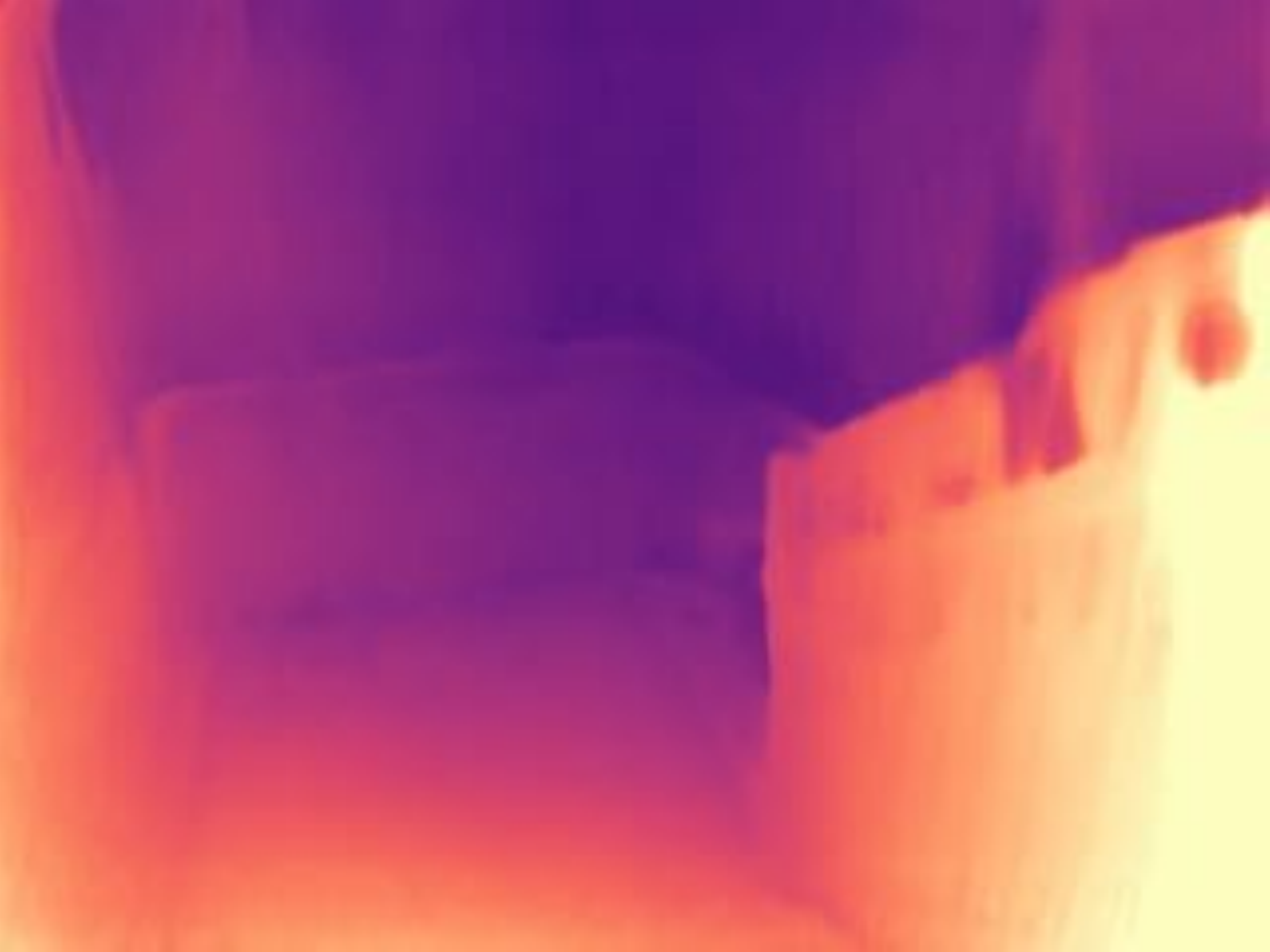}&
        \includegraphics[width=\w,height=\h]{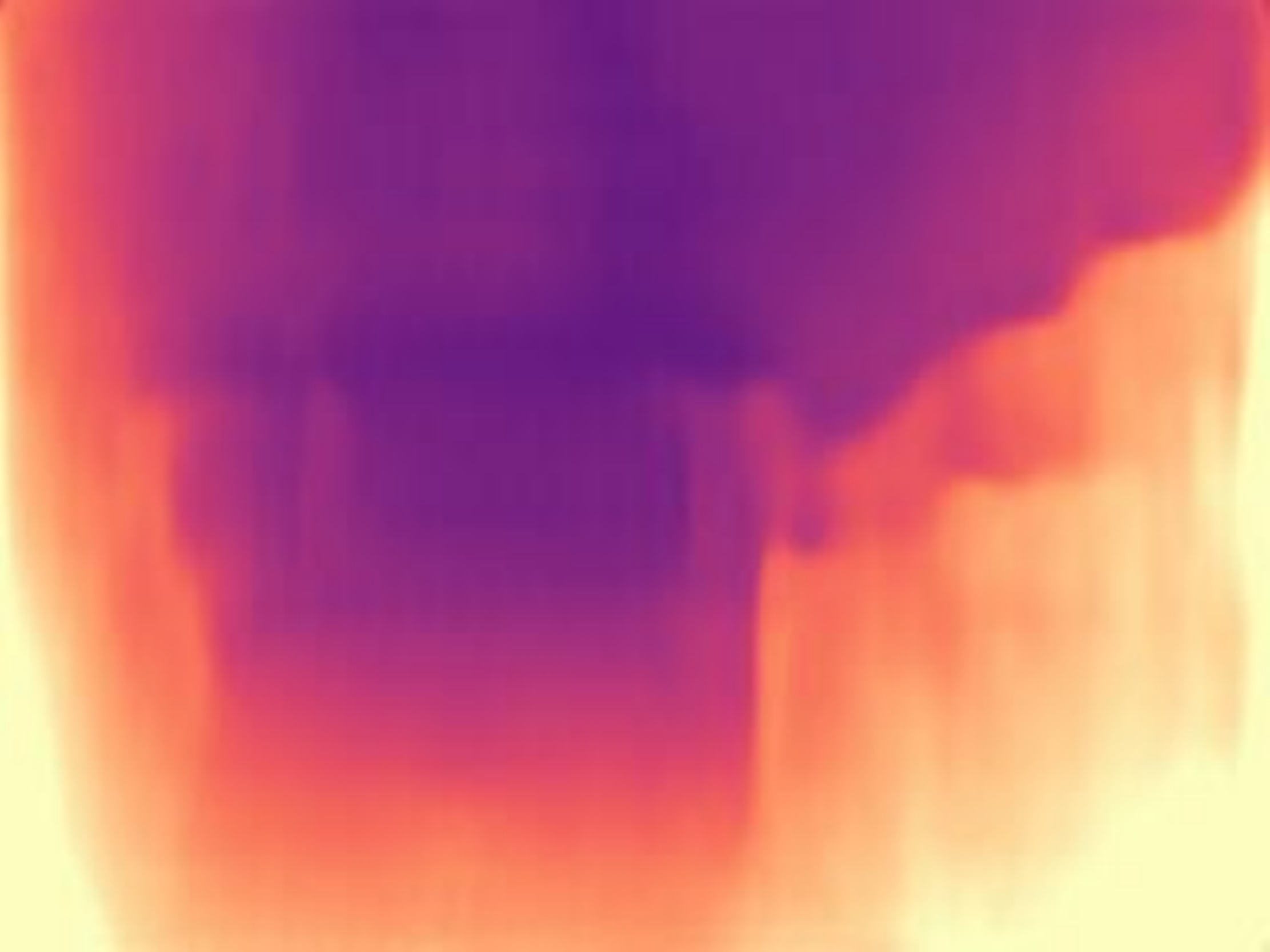}& 
        \includegraphics[width=\w,height=\h]{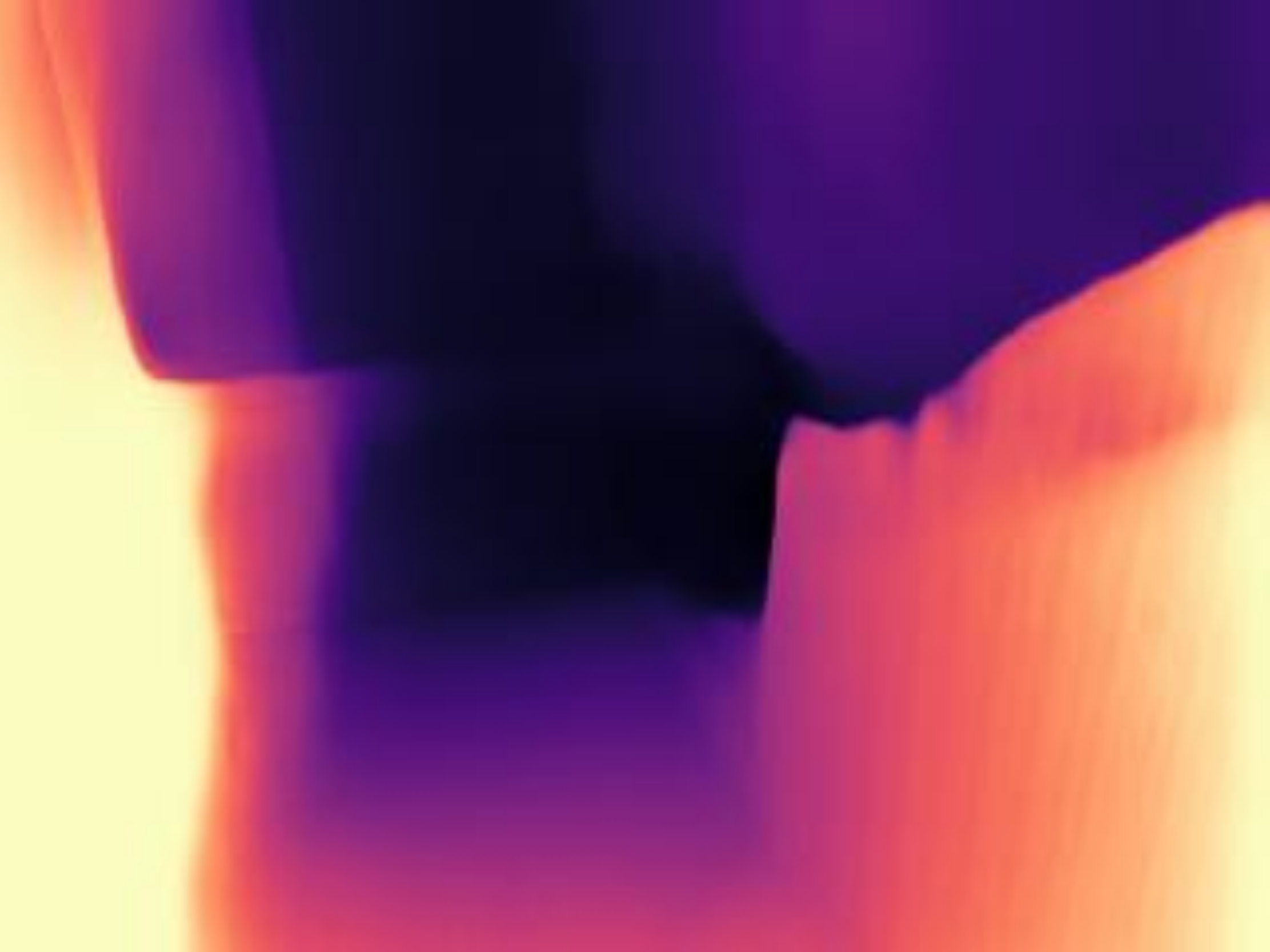}& 
        \includegraphics[width=\w,height=\h]{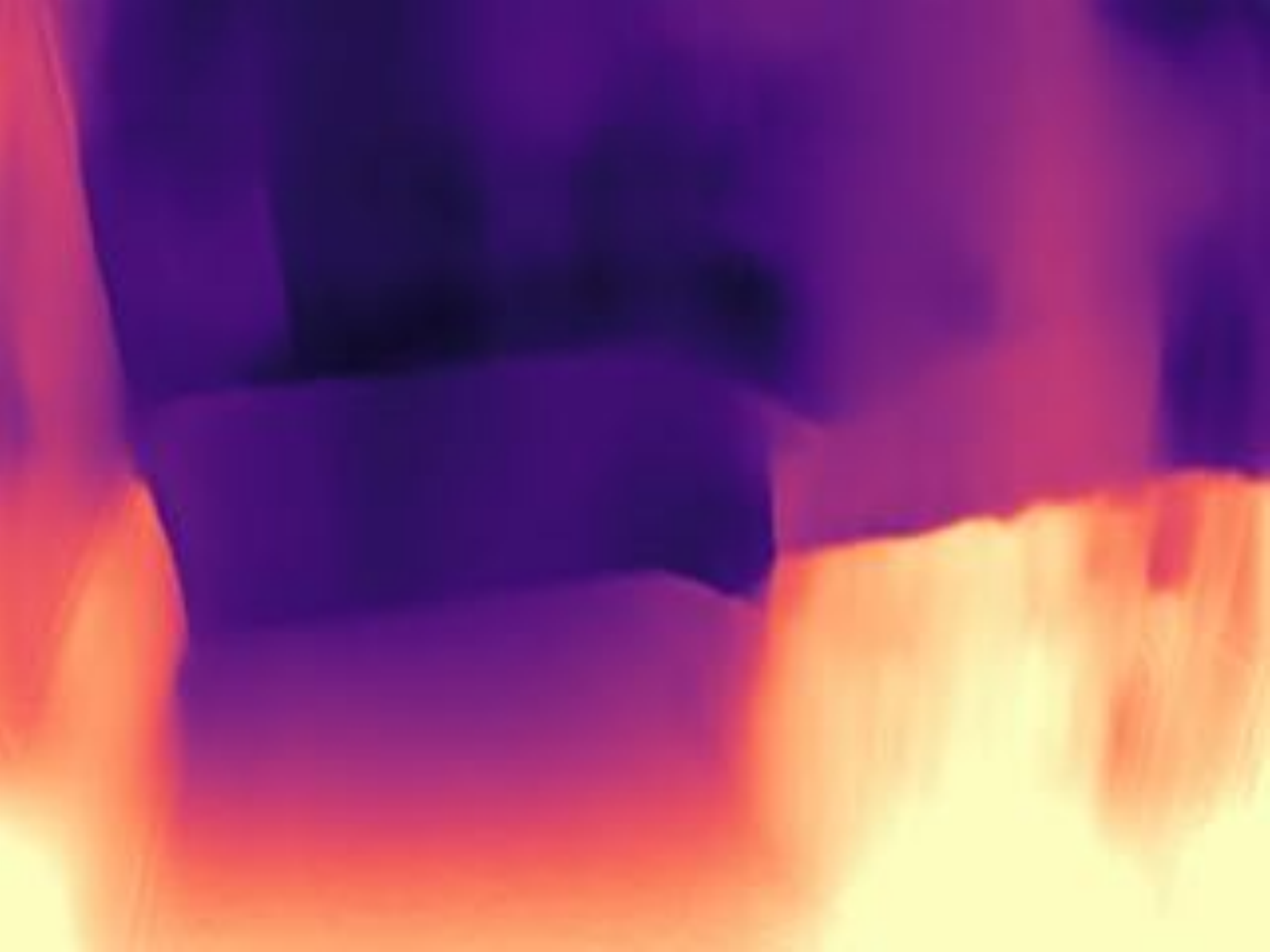}& 
        \includegraphics[width=\w,height=\h]{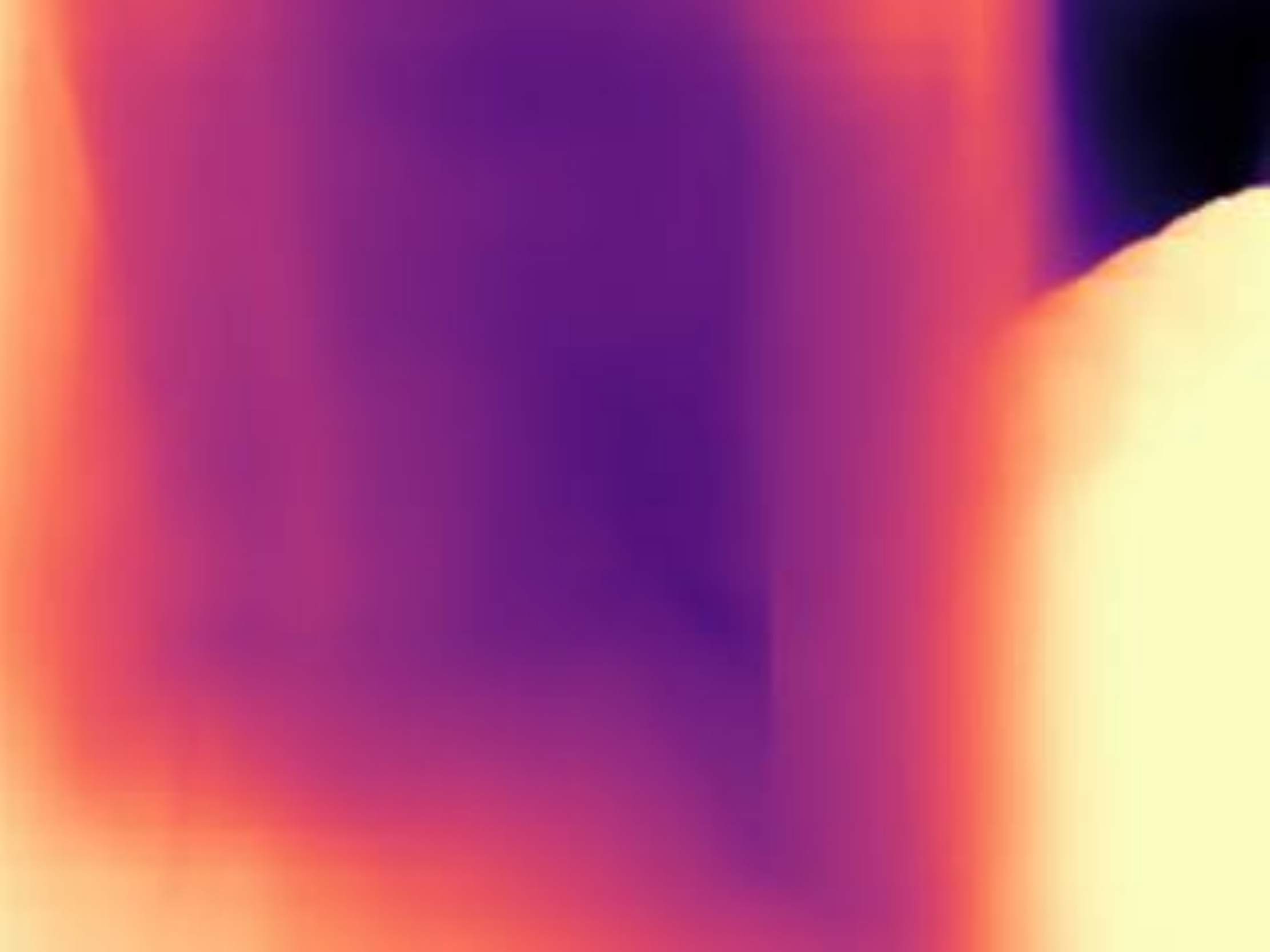}\\ 
        \includegraphics[width=\w,height=\h]{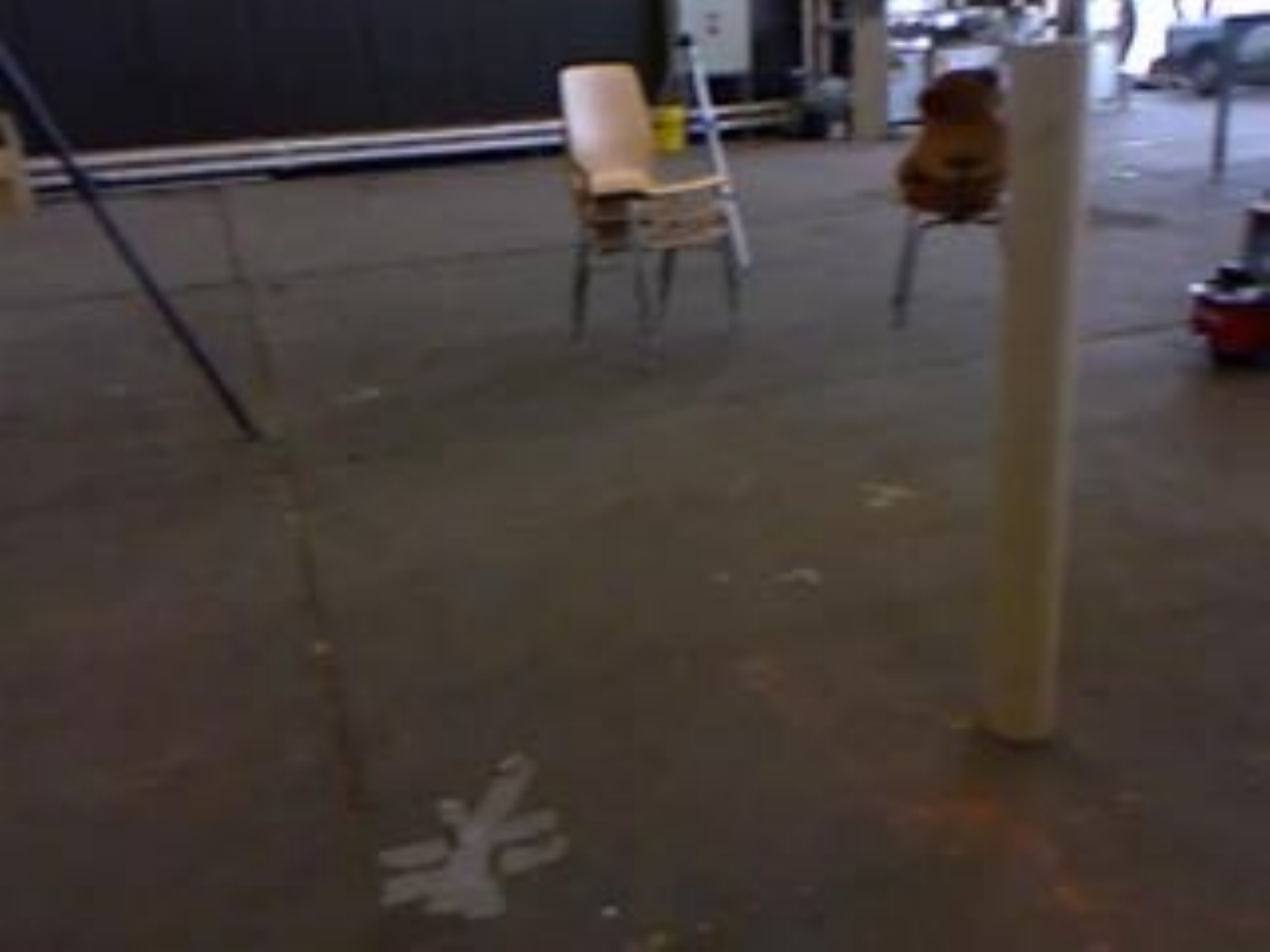}& 
        \includegraphics[width=\w,height=\h]{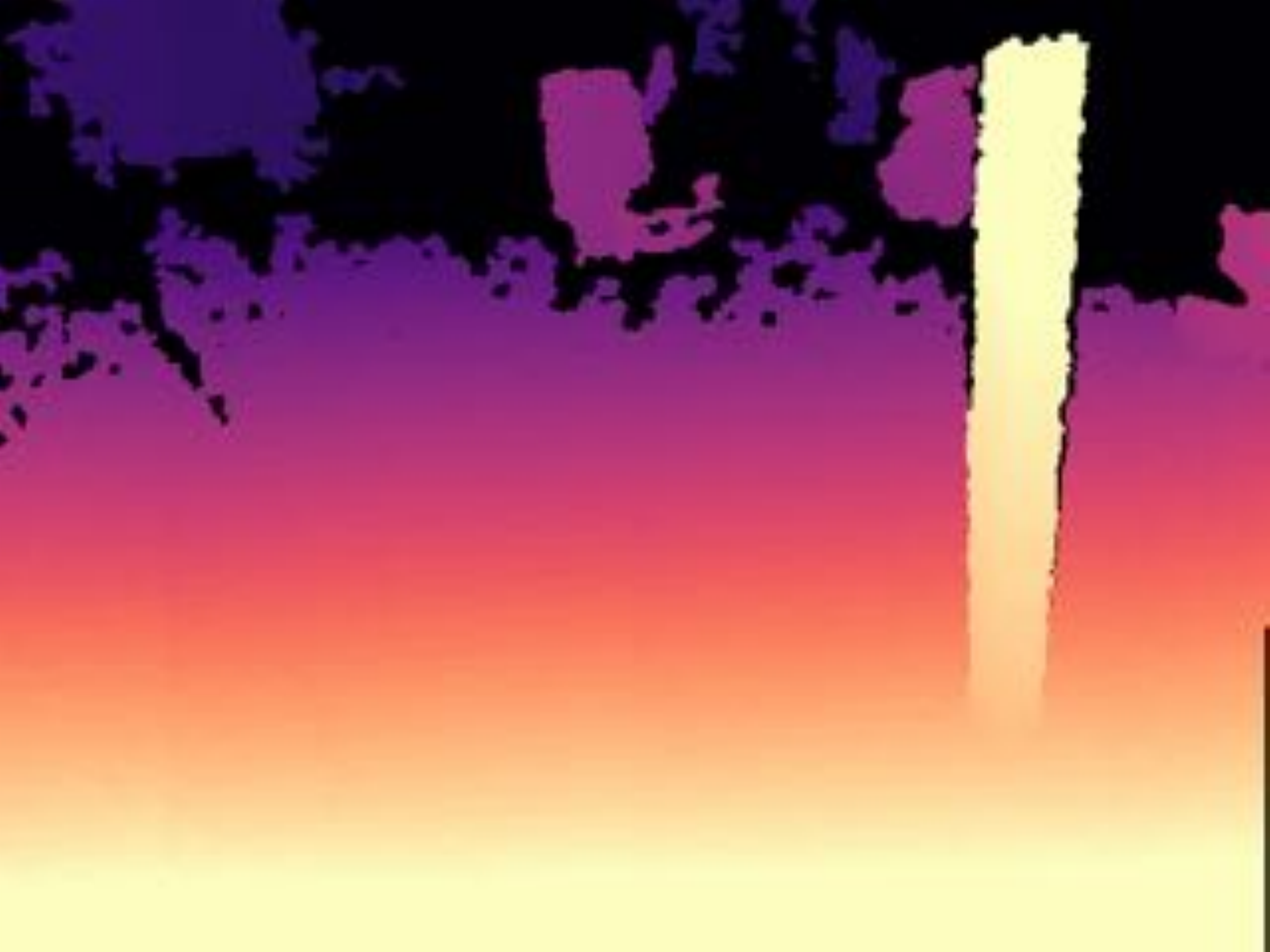}& 
        \includegraphics[width=\w,height=\h]{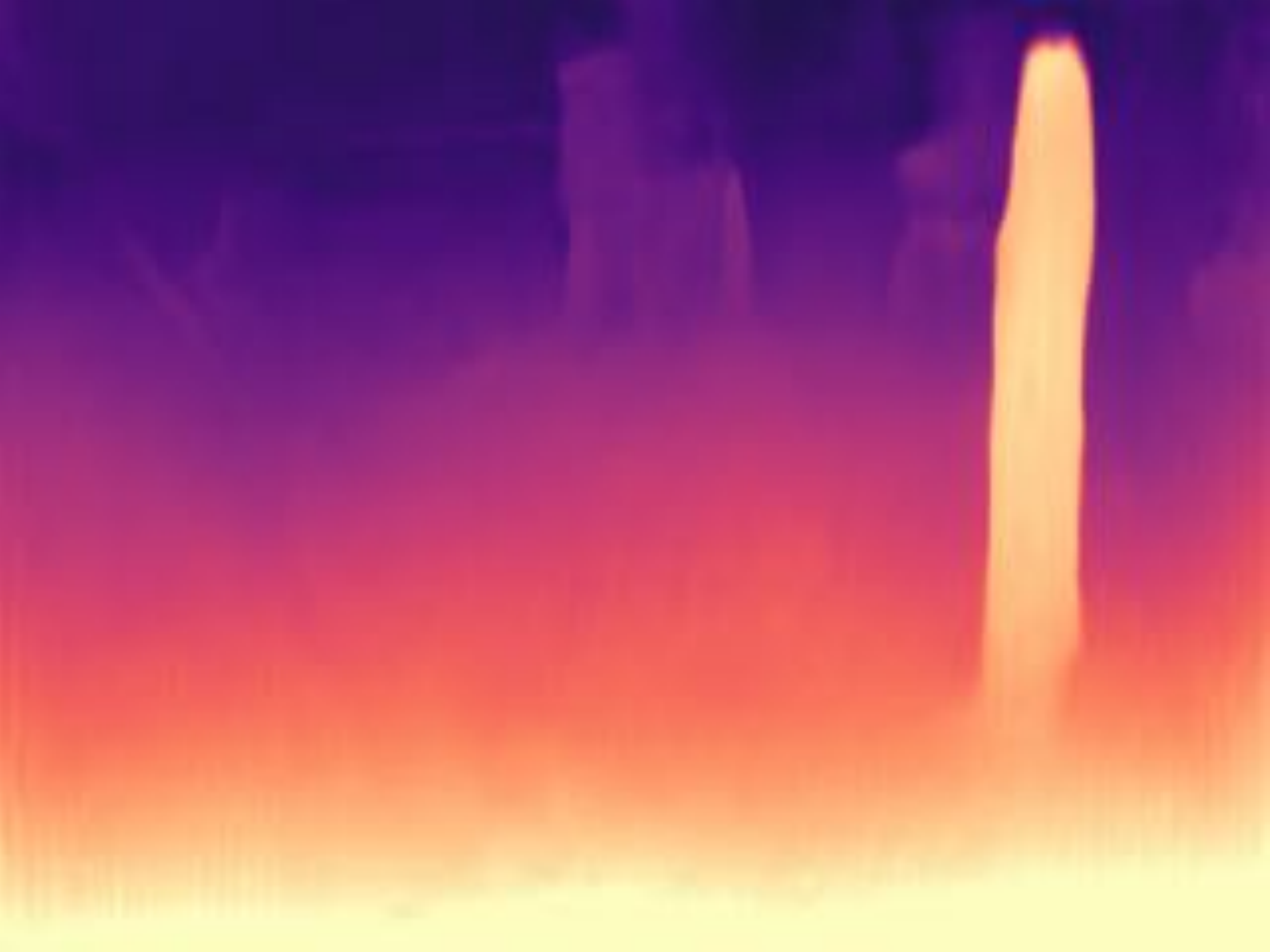}&
        \includegraphics[width=\w,height=\h]{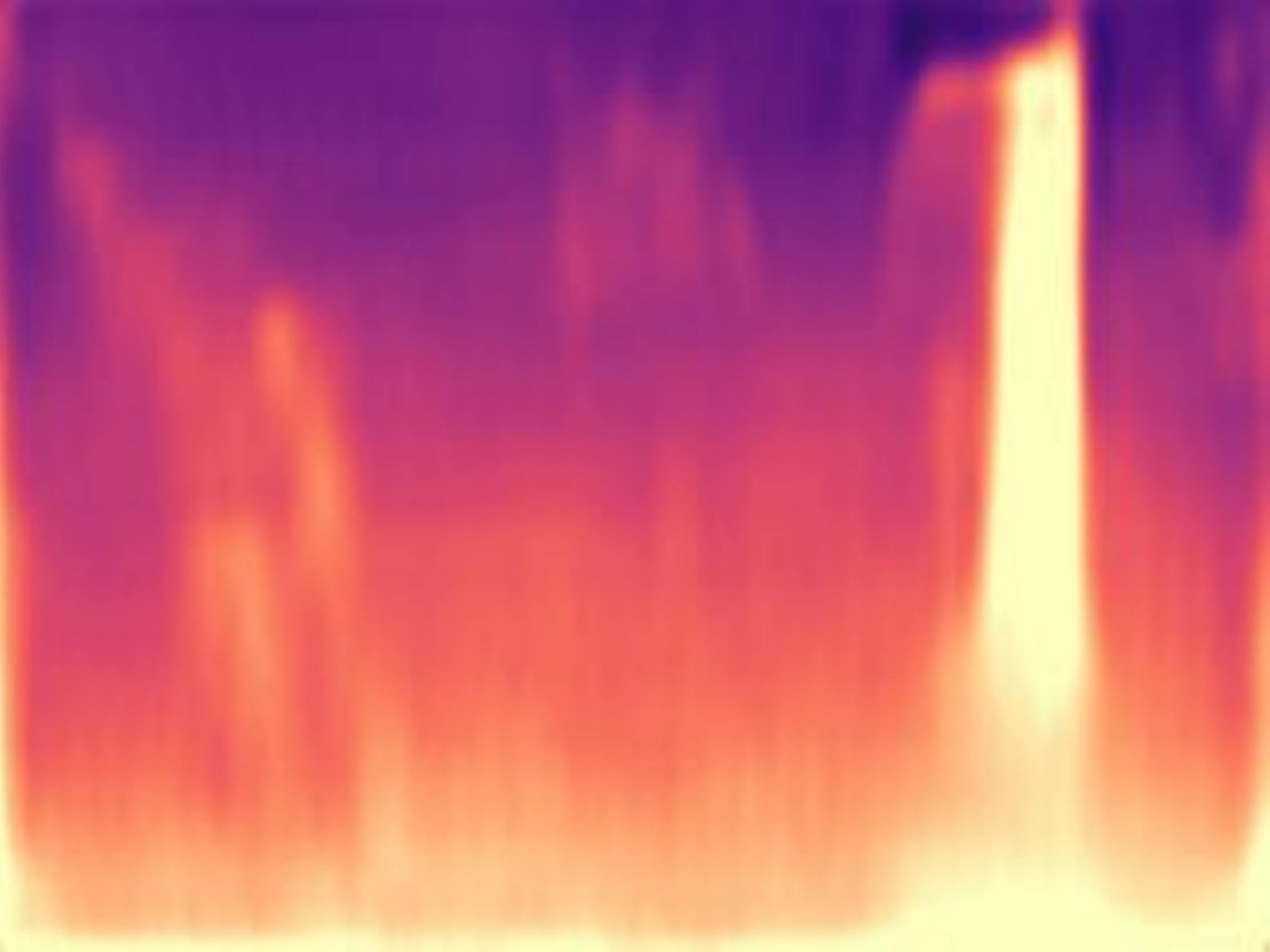}& 
        \includegraphics[width=\w,height=\h]{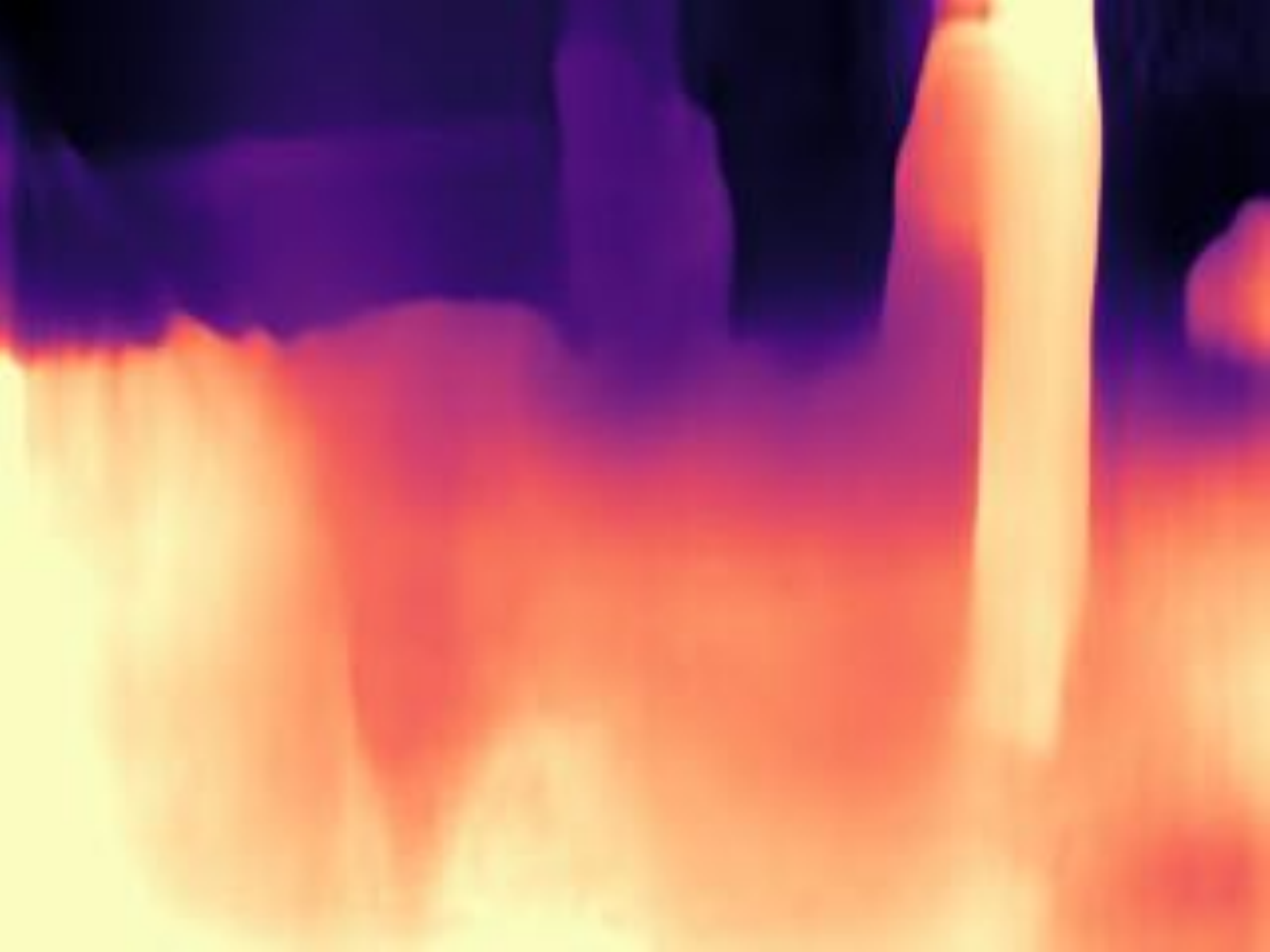}& 
        \includegraphics[width=\w,height=\h]{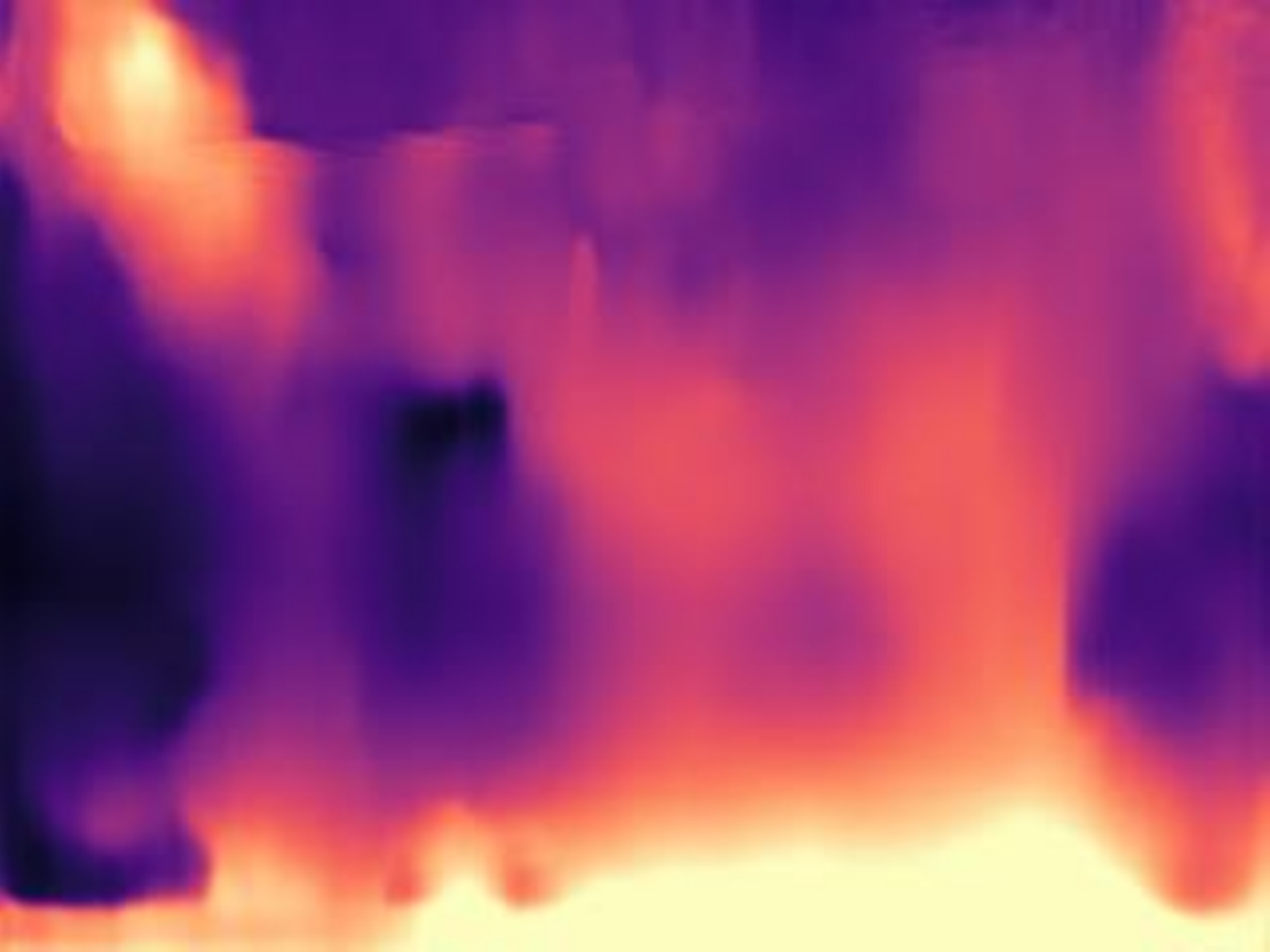}& 
        \includegraphics[width=\w,height=\h]{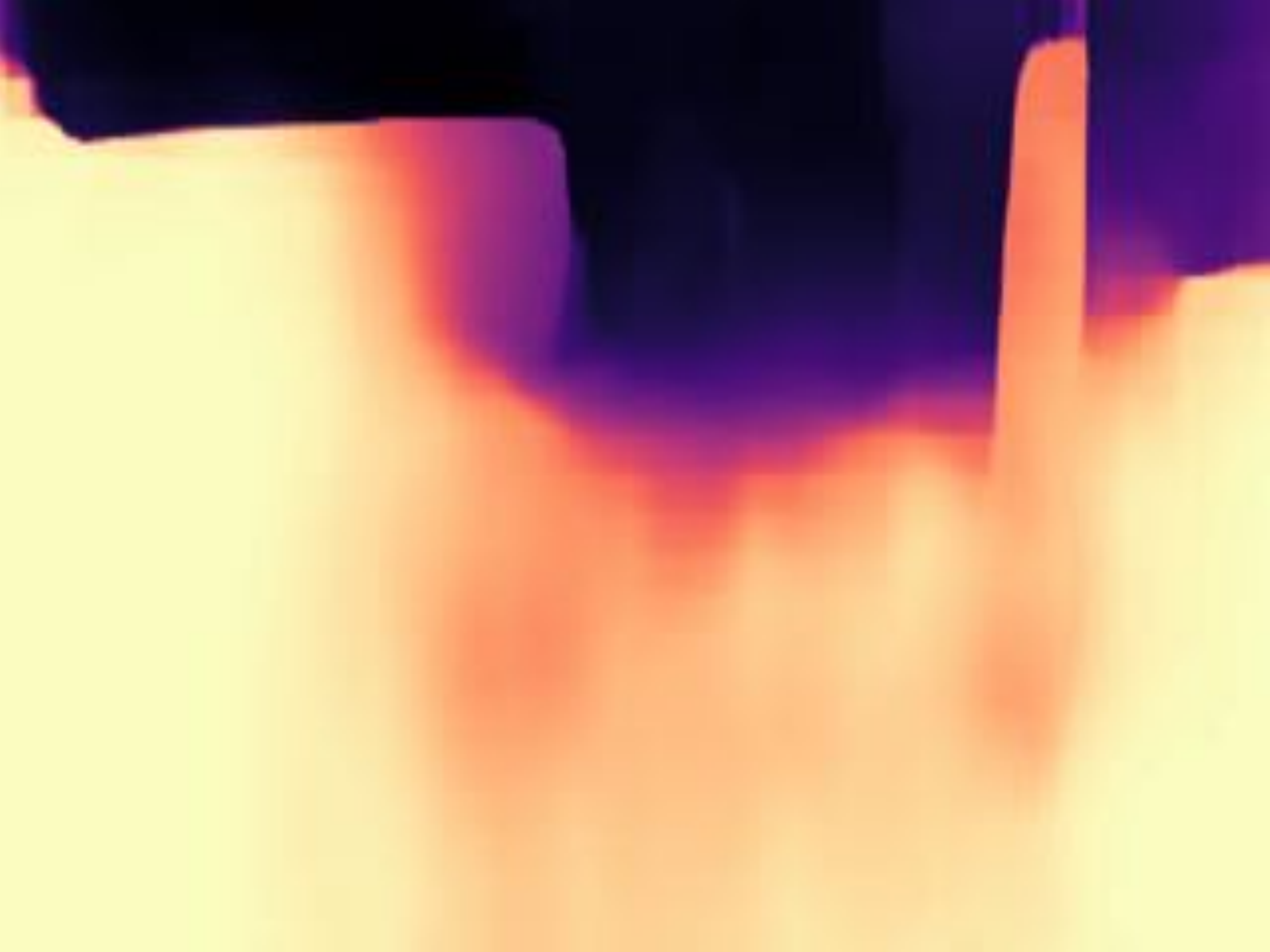}\\ 
        \includegraphics[width=\w,height=\h]{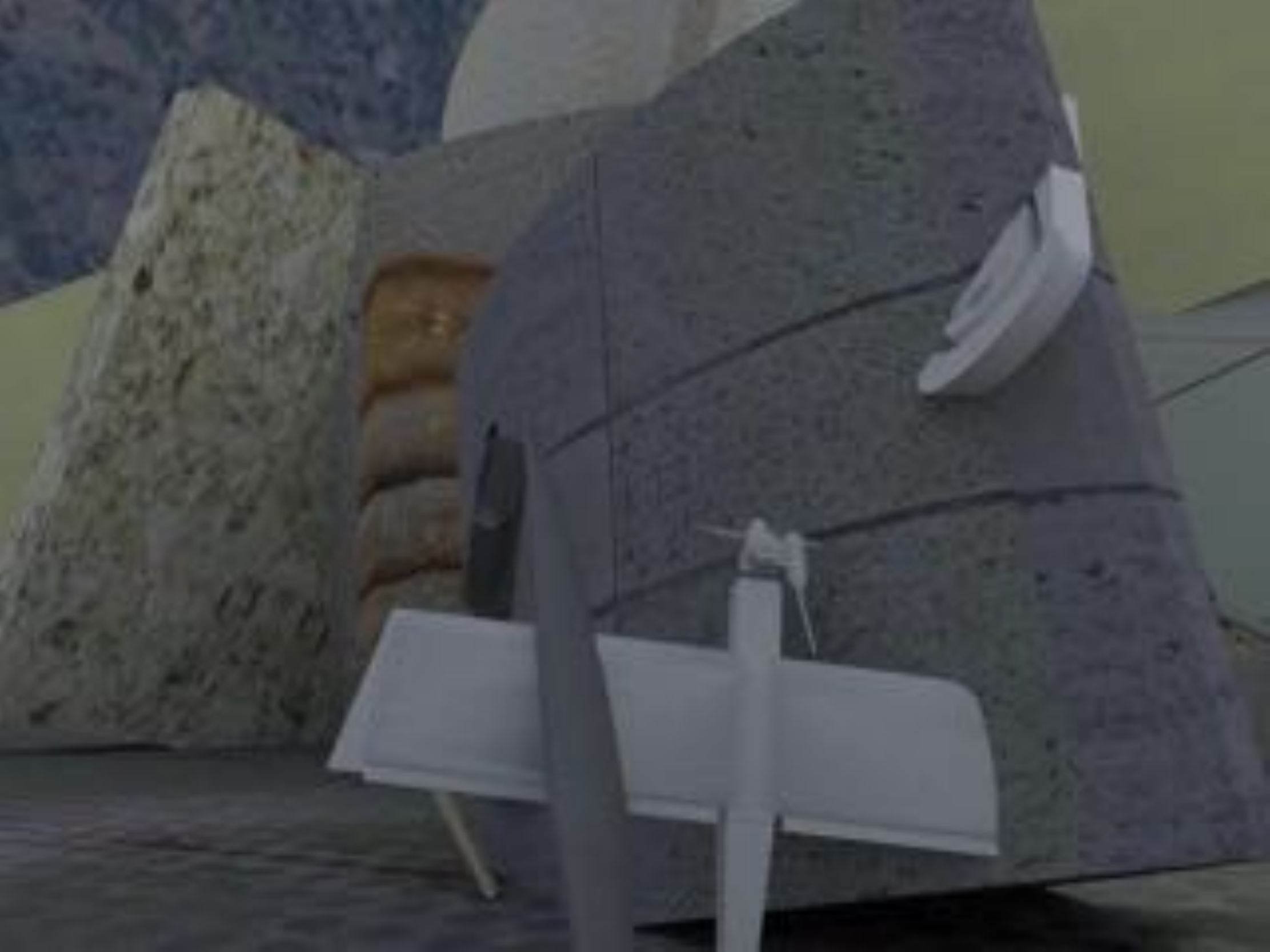}& 
        \includegraphics[width=\w,height=\h]{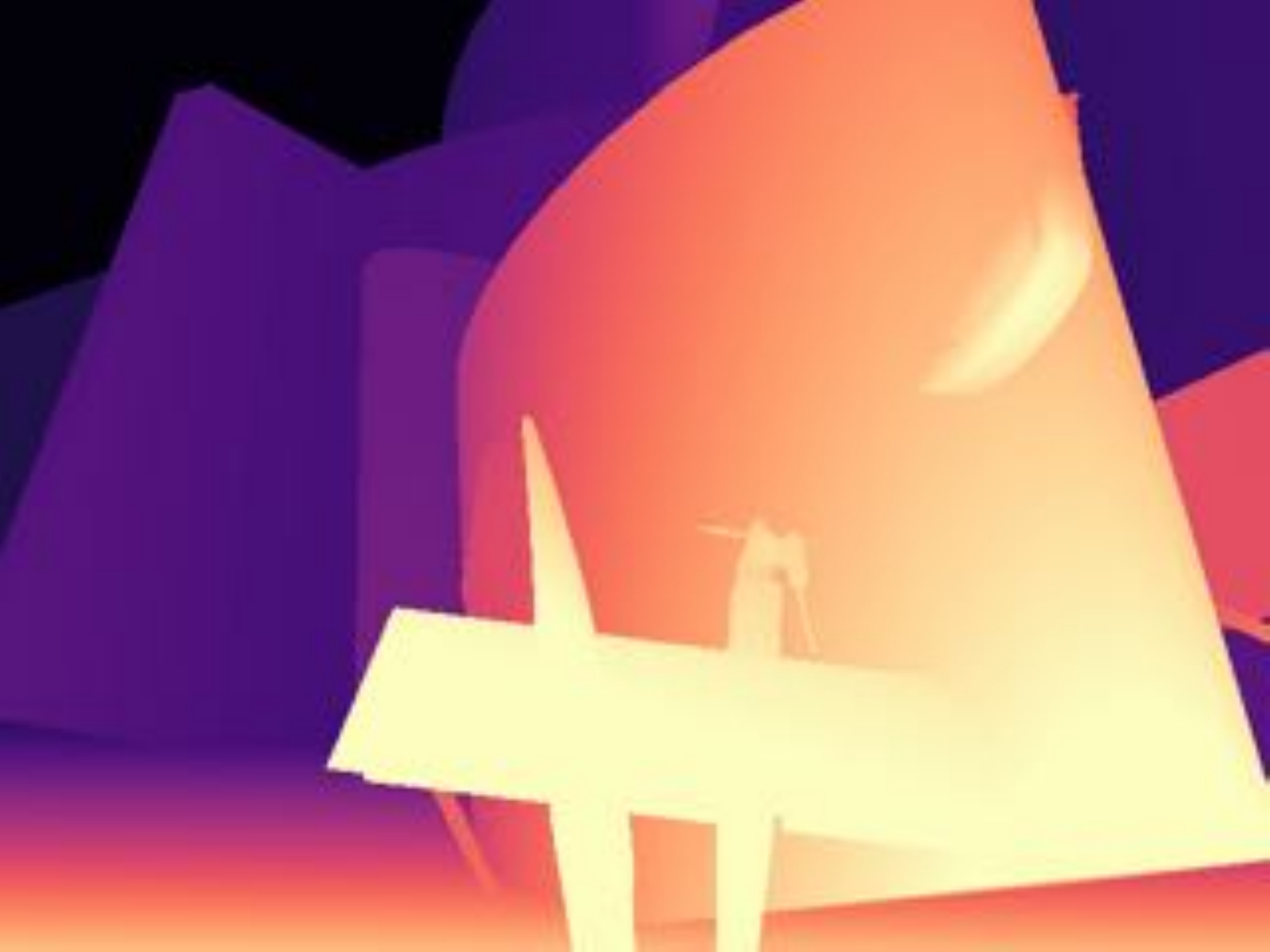}& 
        \includegraphics[width=\w,height=\h]{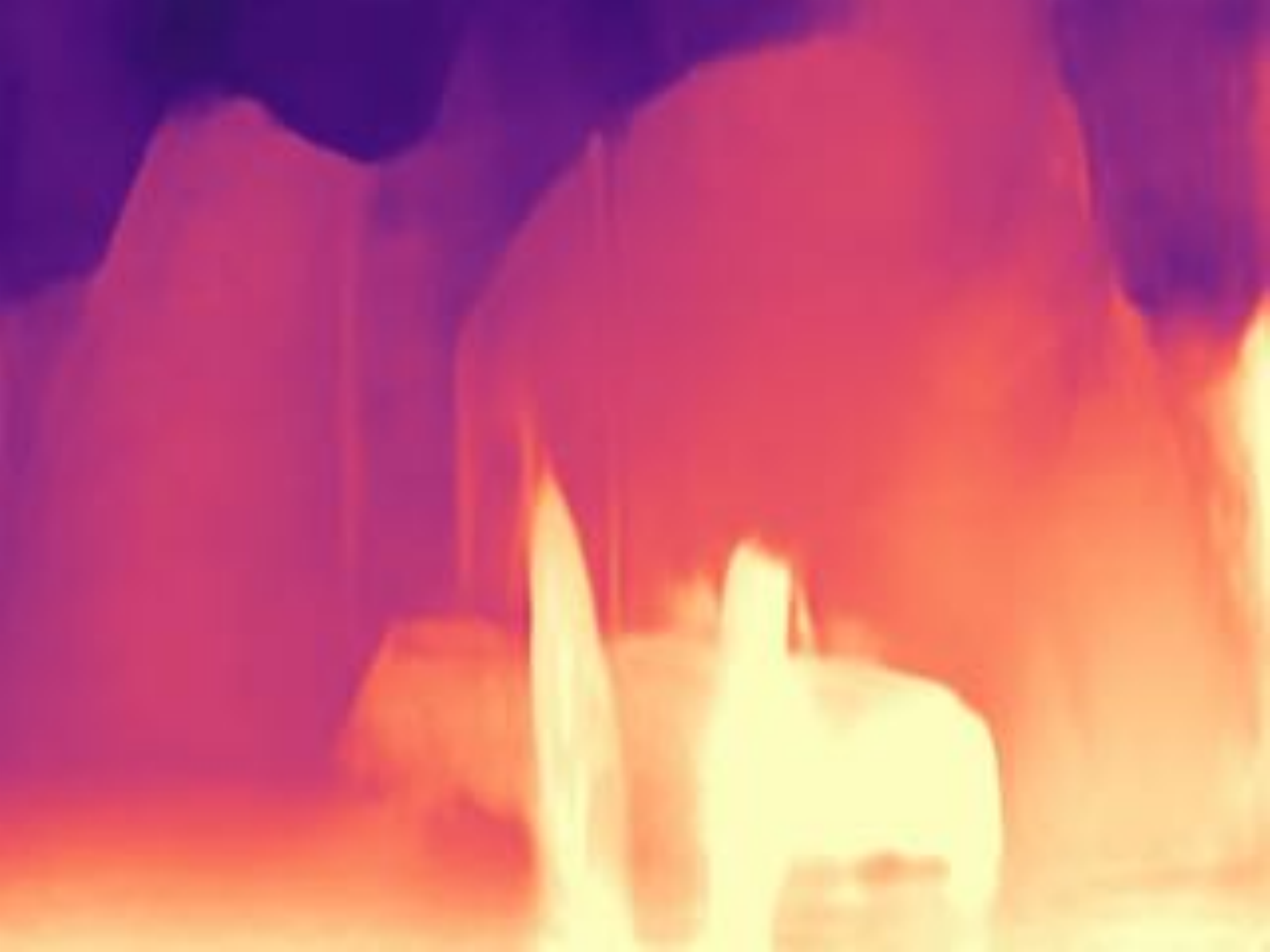}&
        \includegraphics[width=\w,height=\h]{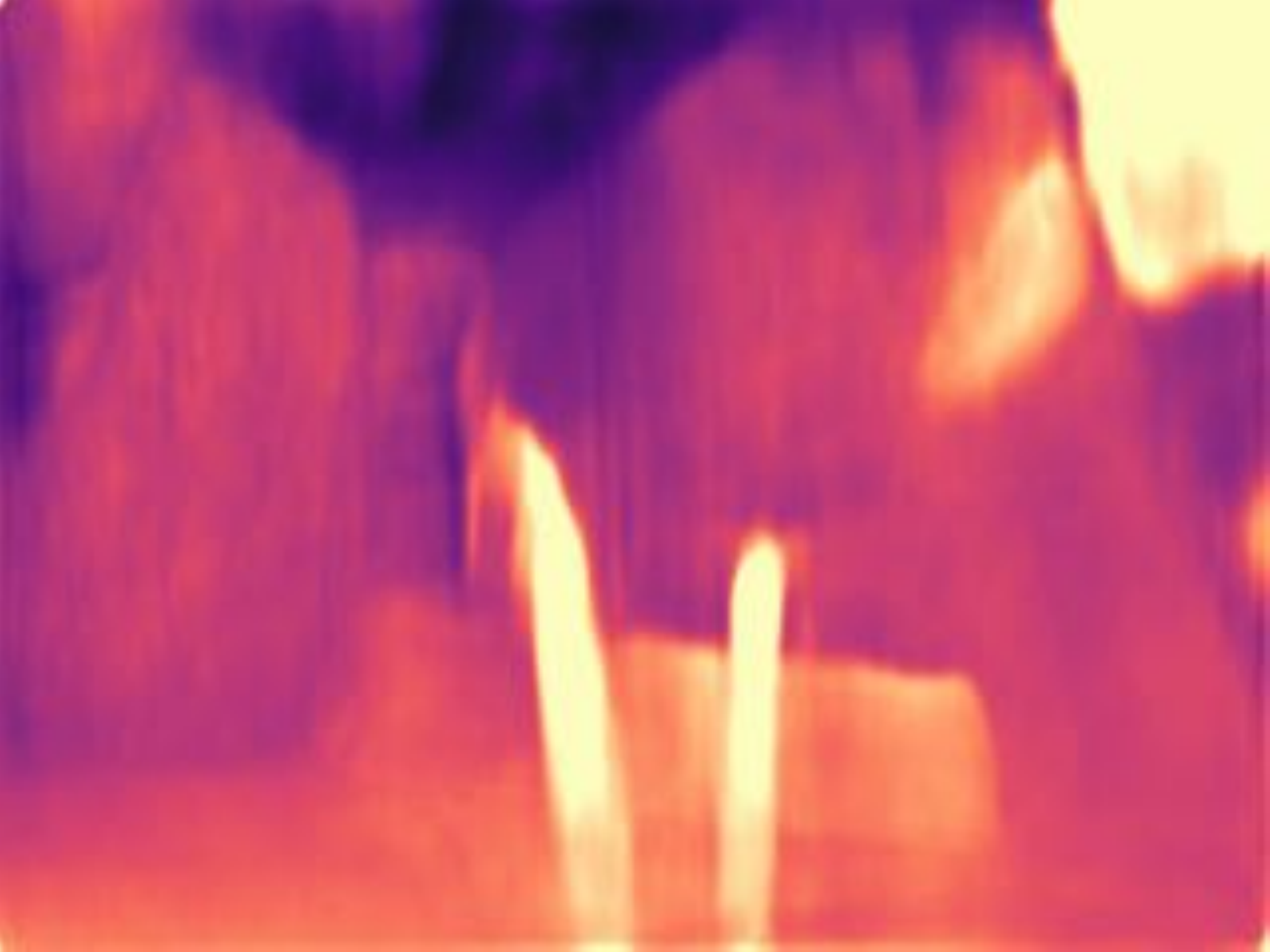}& 
        \includegraphics[width=\w,height=\h]{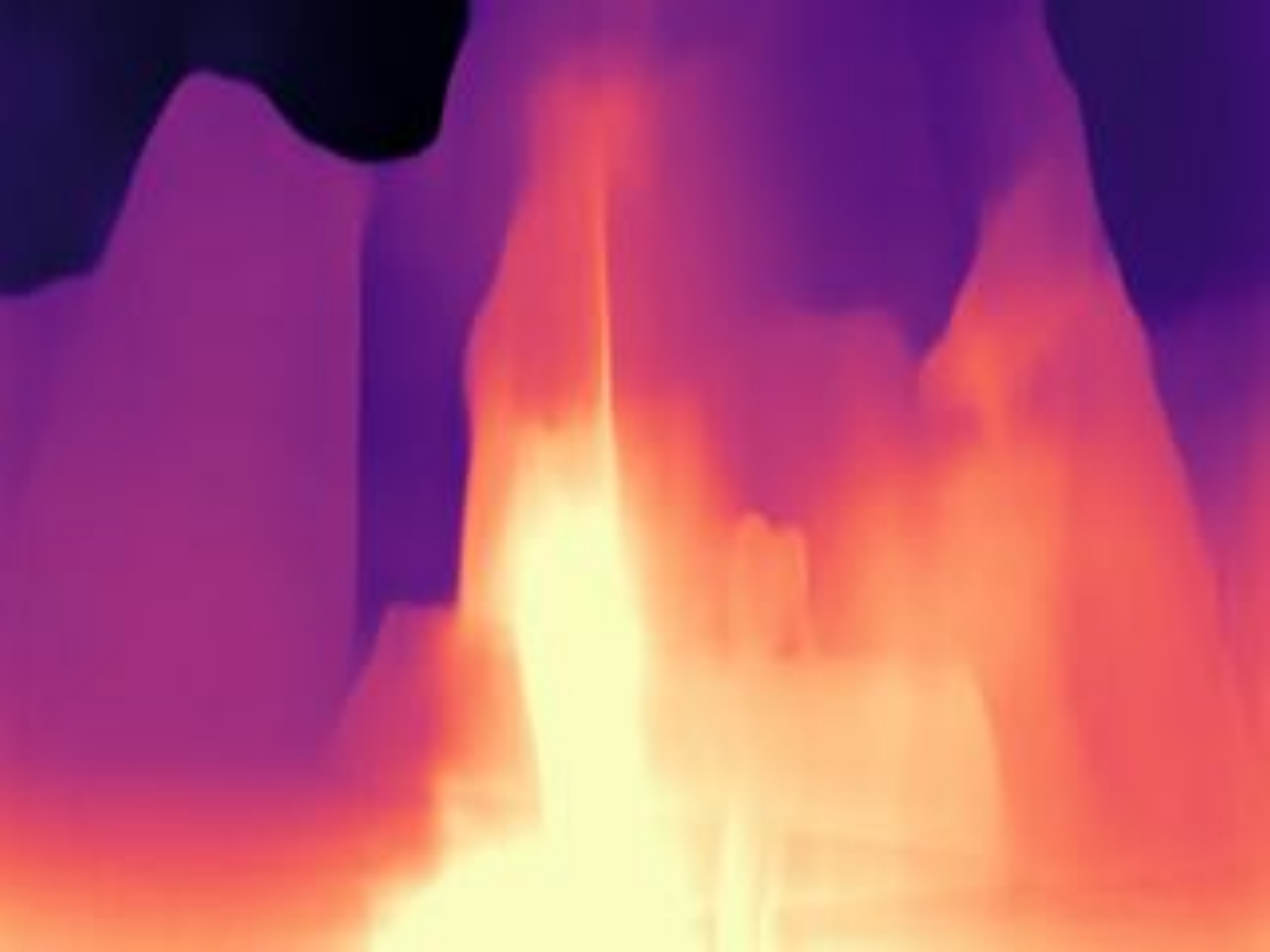}& 
        \includegraphics[width=\w,height=\h]{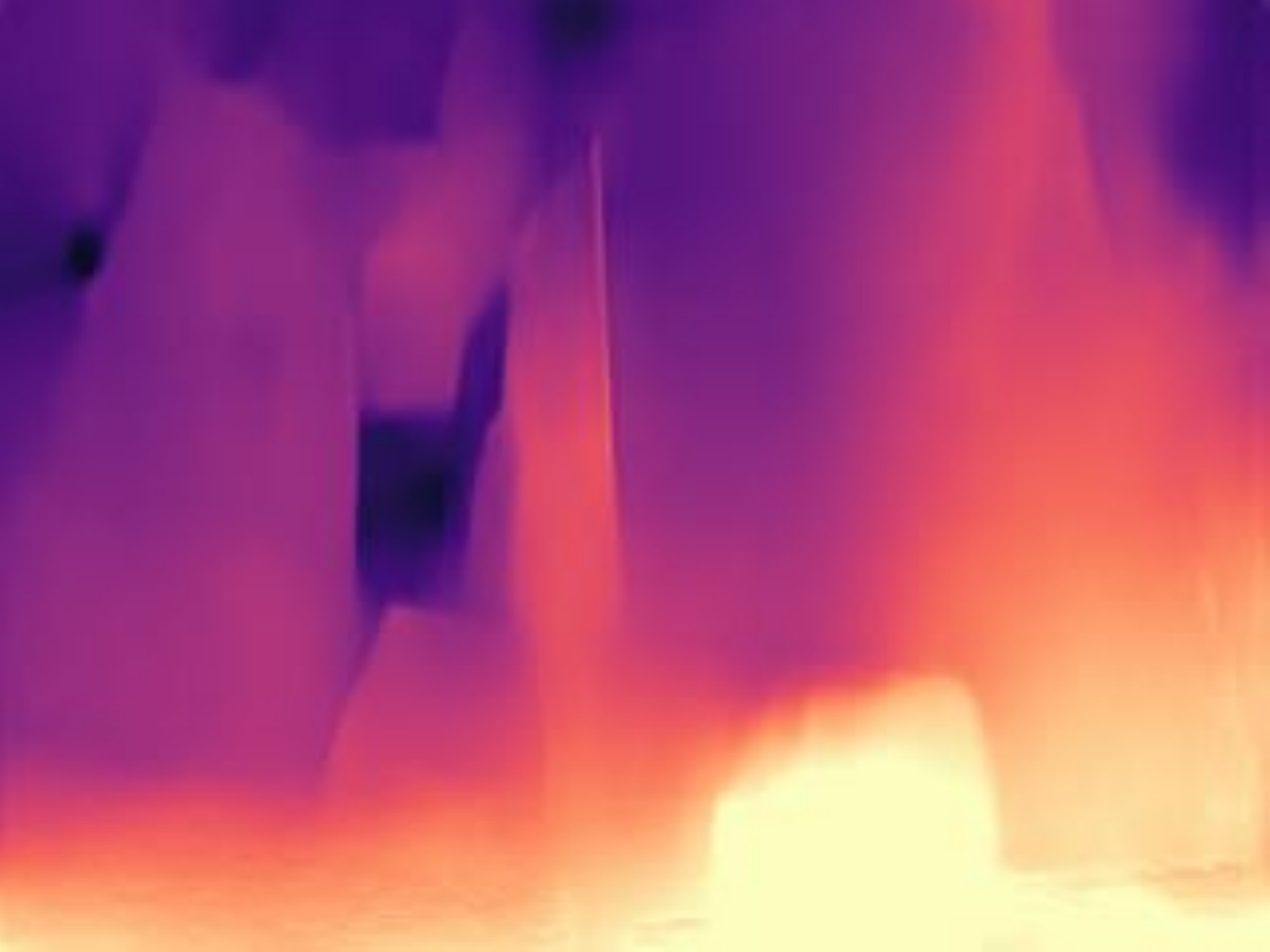}& 
        \includegraphics[width=\w,height=\h]{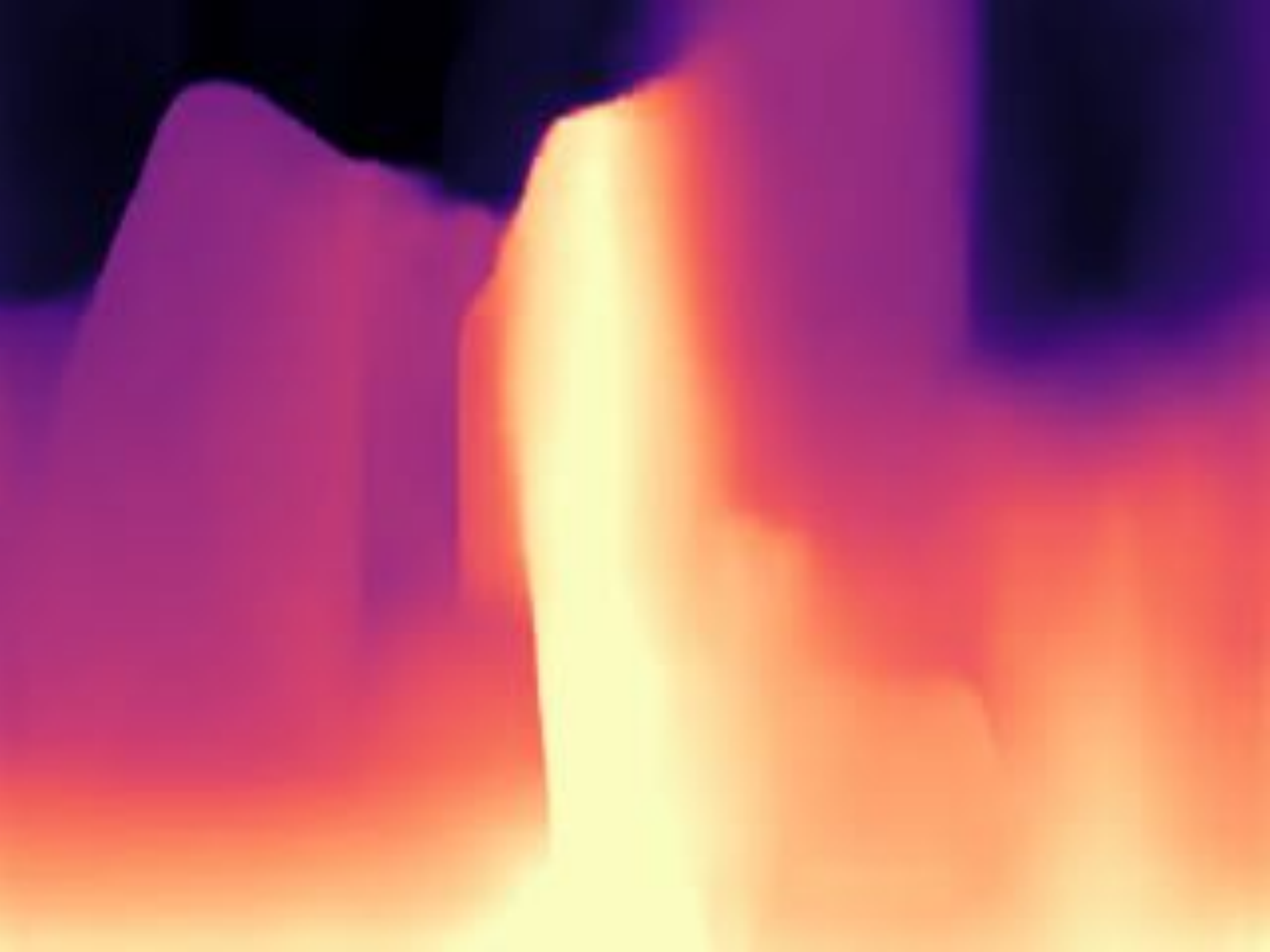}\\ 
        \includegraphics[width=\w,height=\h]{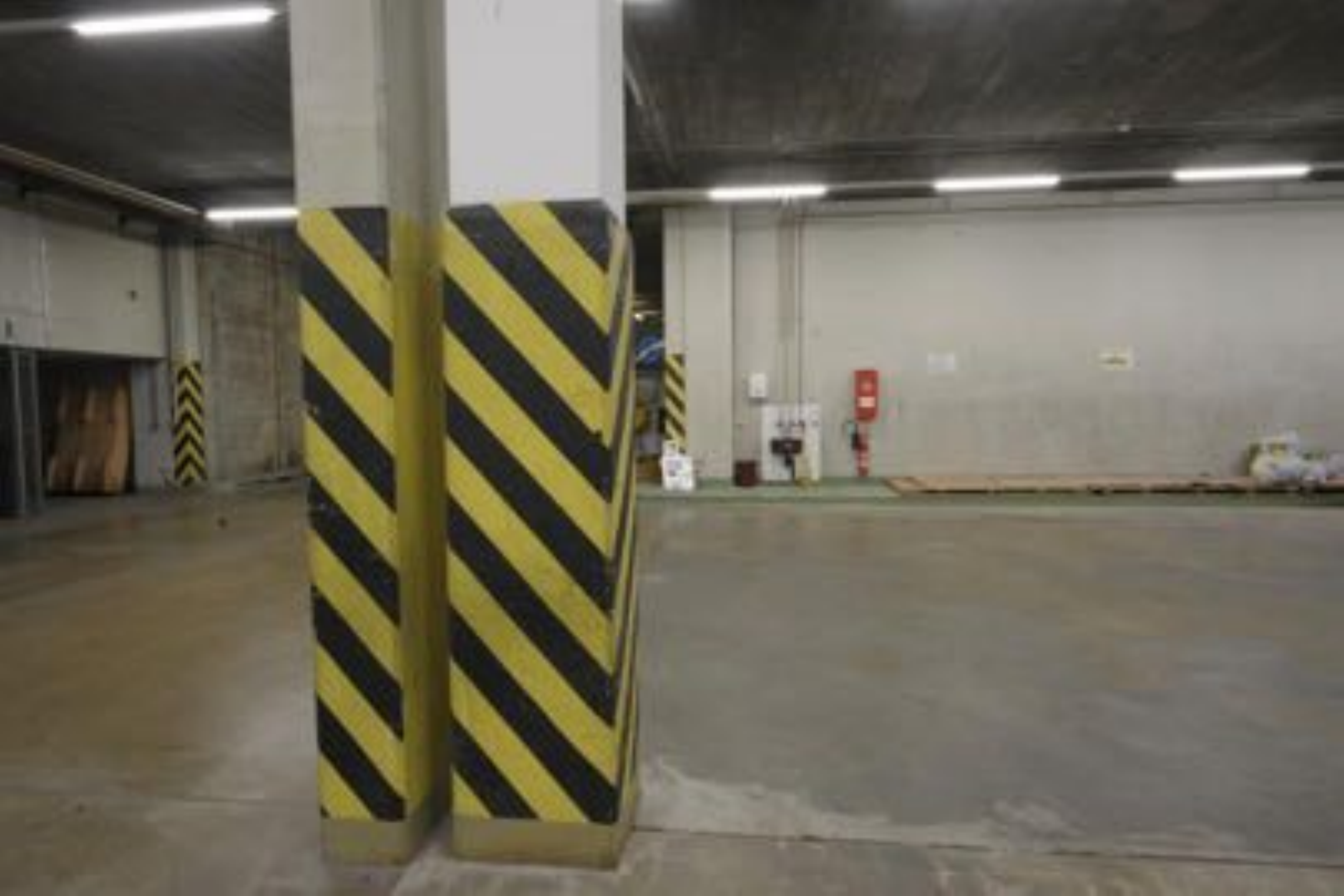}& 
        \includegraphics[width=\w,height=\h]{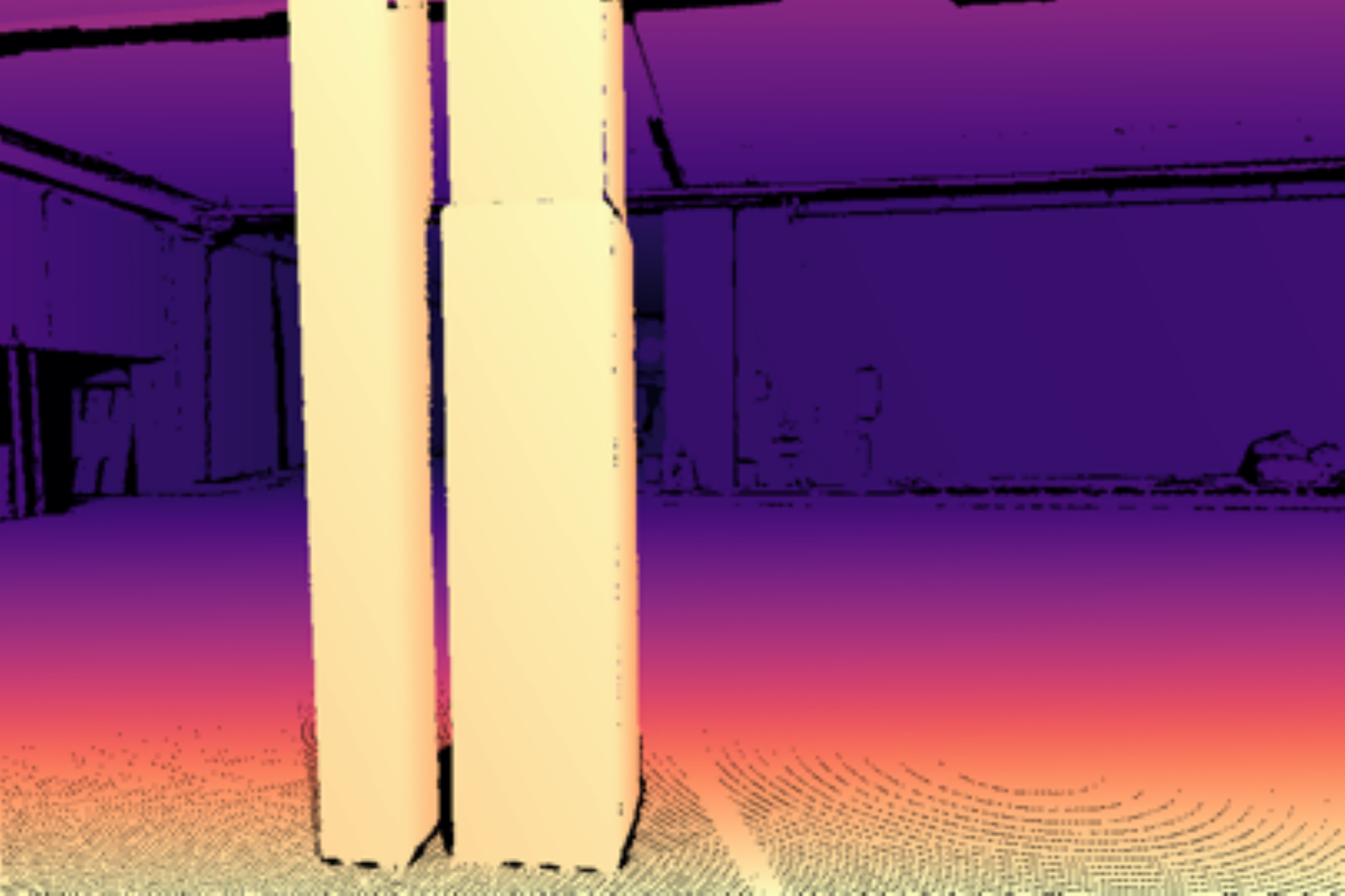}& 
        \includegraphics[width=\w,height=\h]{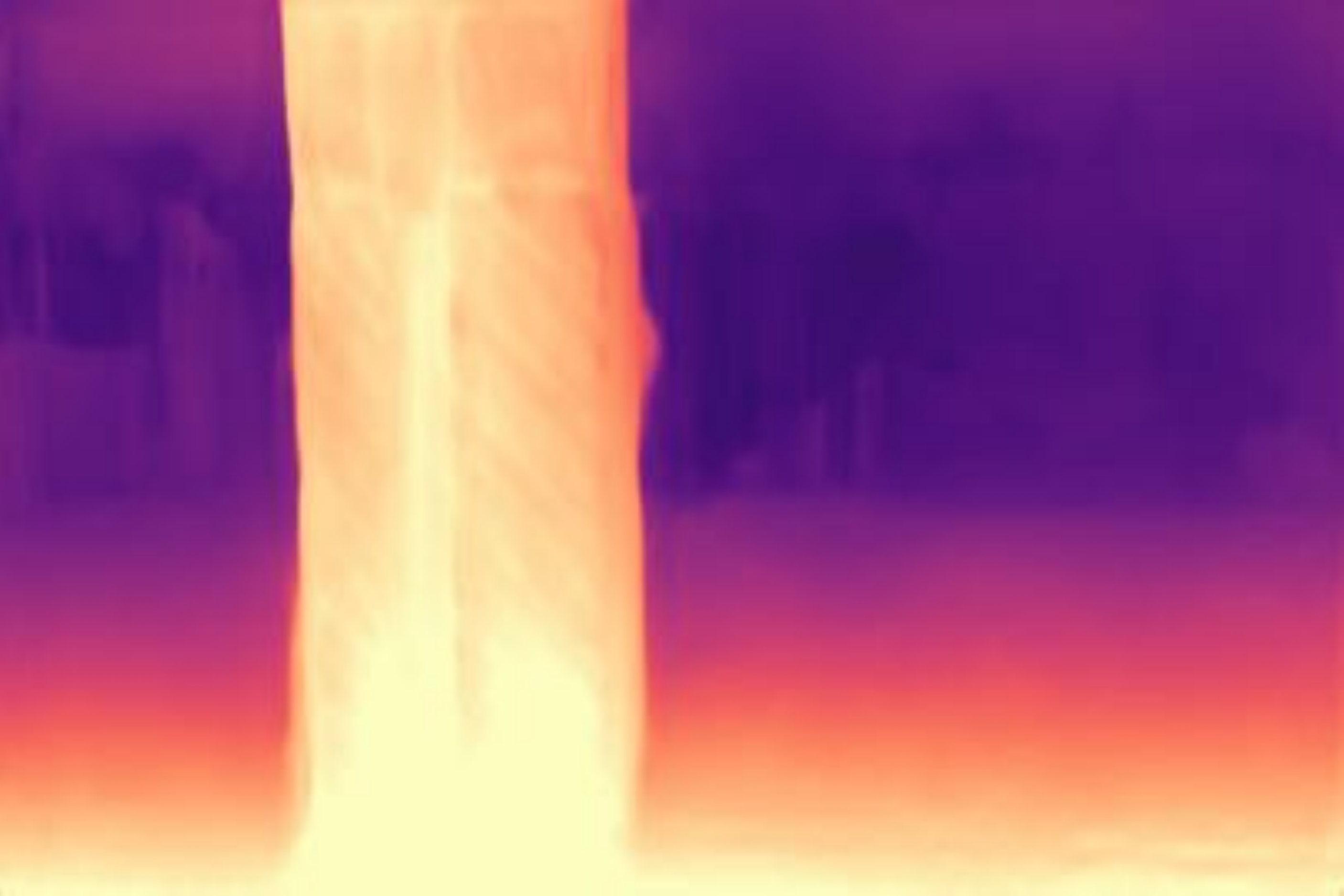}& 
        \includegraphics[width=\w,height=\h]{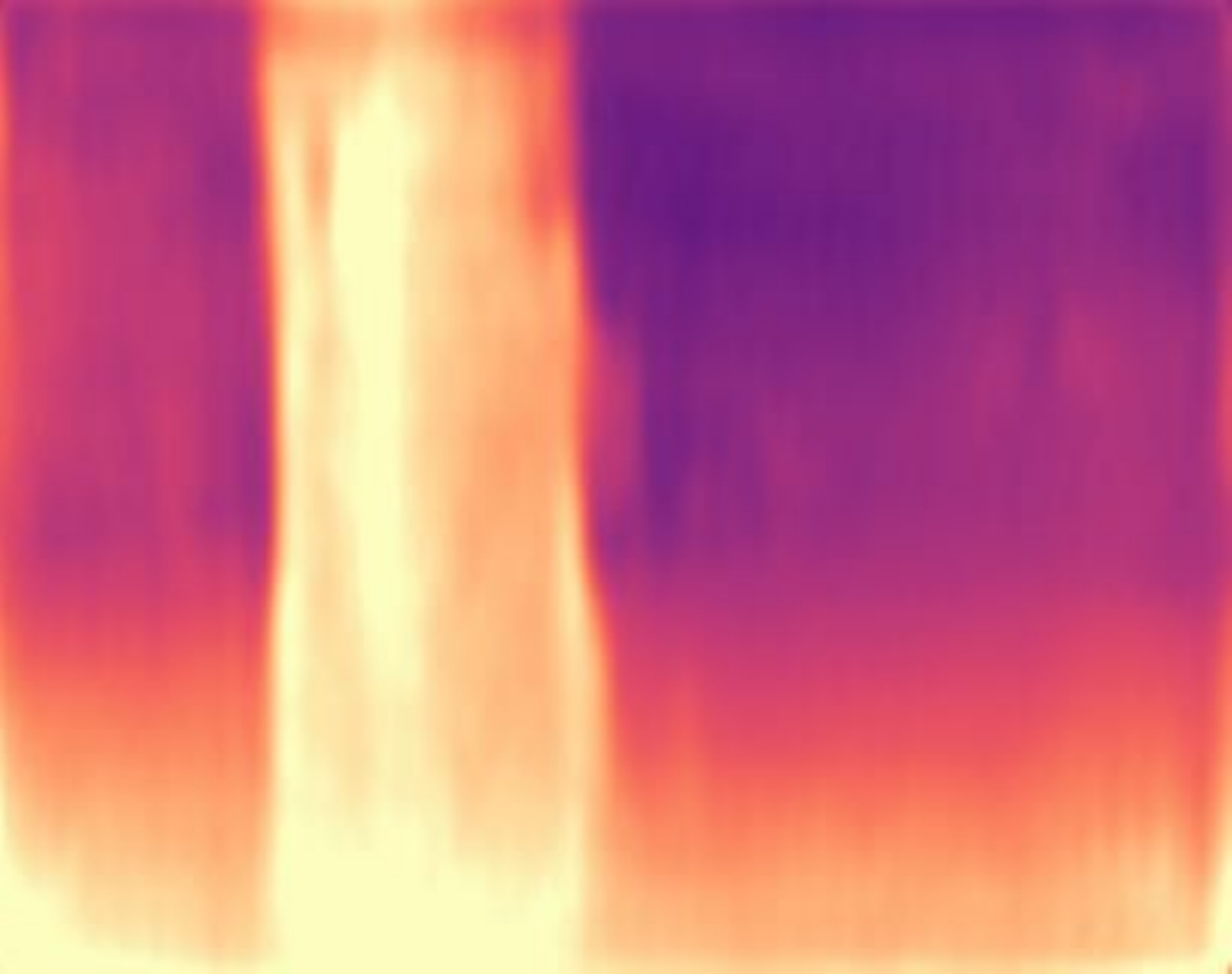}& 
        \includegraphics[width=\w,height=\h]{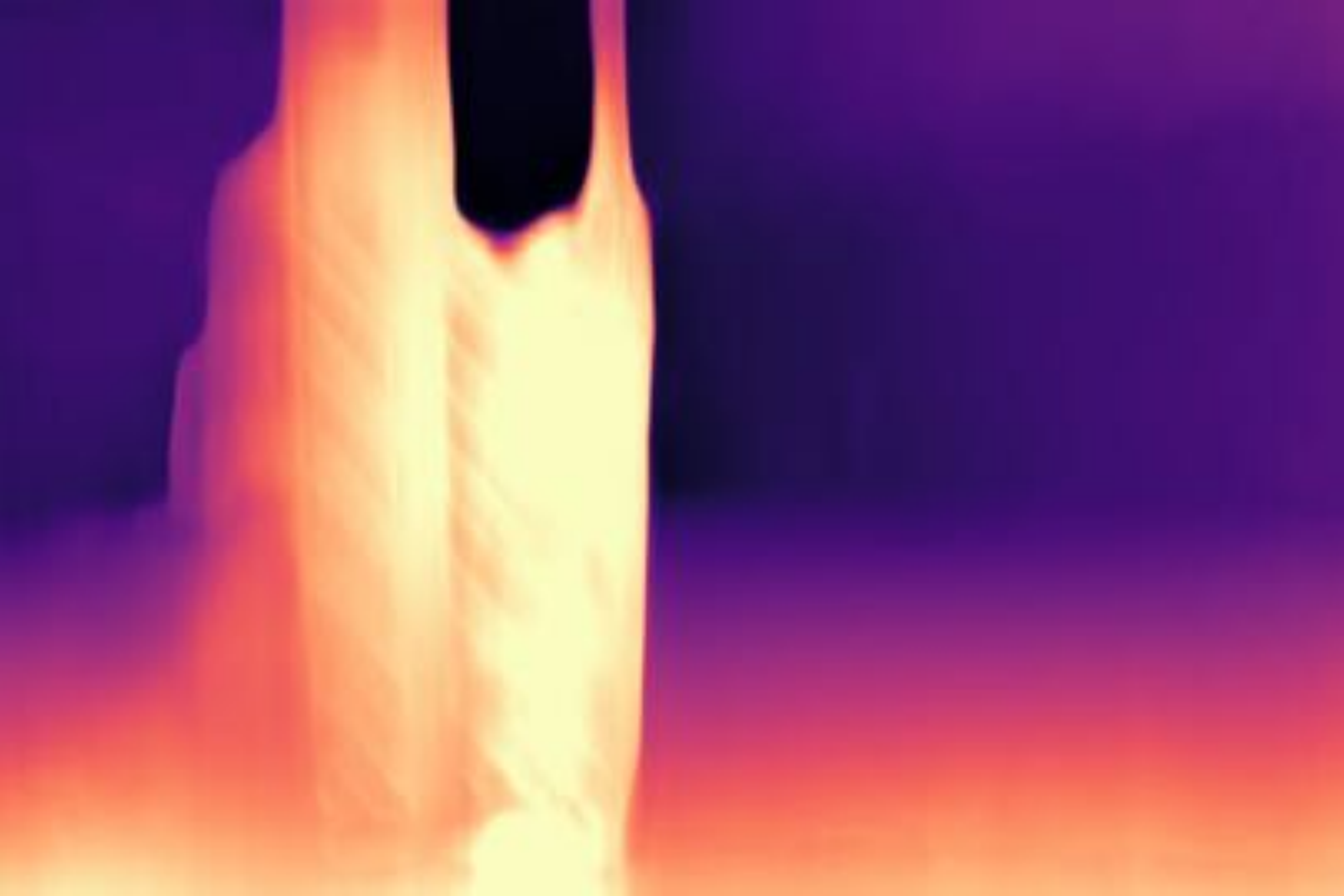}& 
        \includegraphics[width=\w,height=\h]{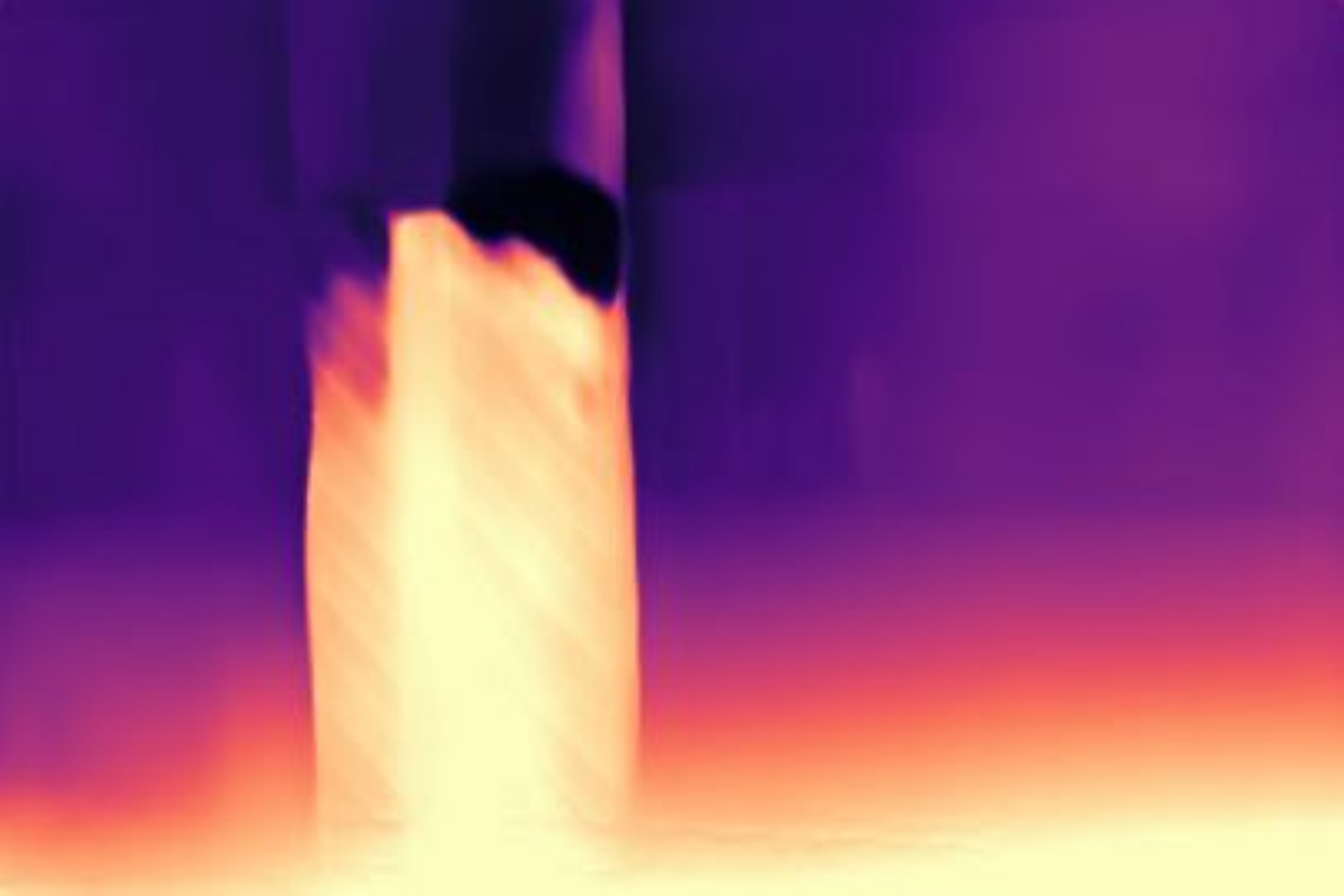}& 
        \includegraphics[width=\w,height=\h]{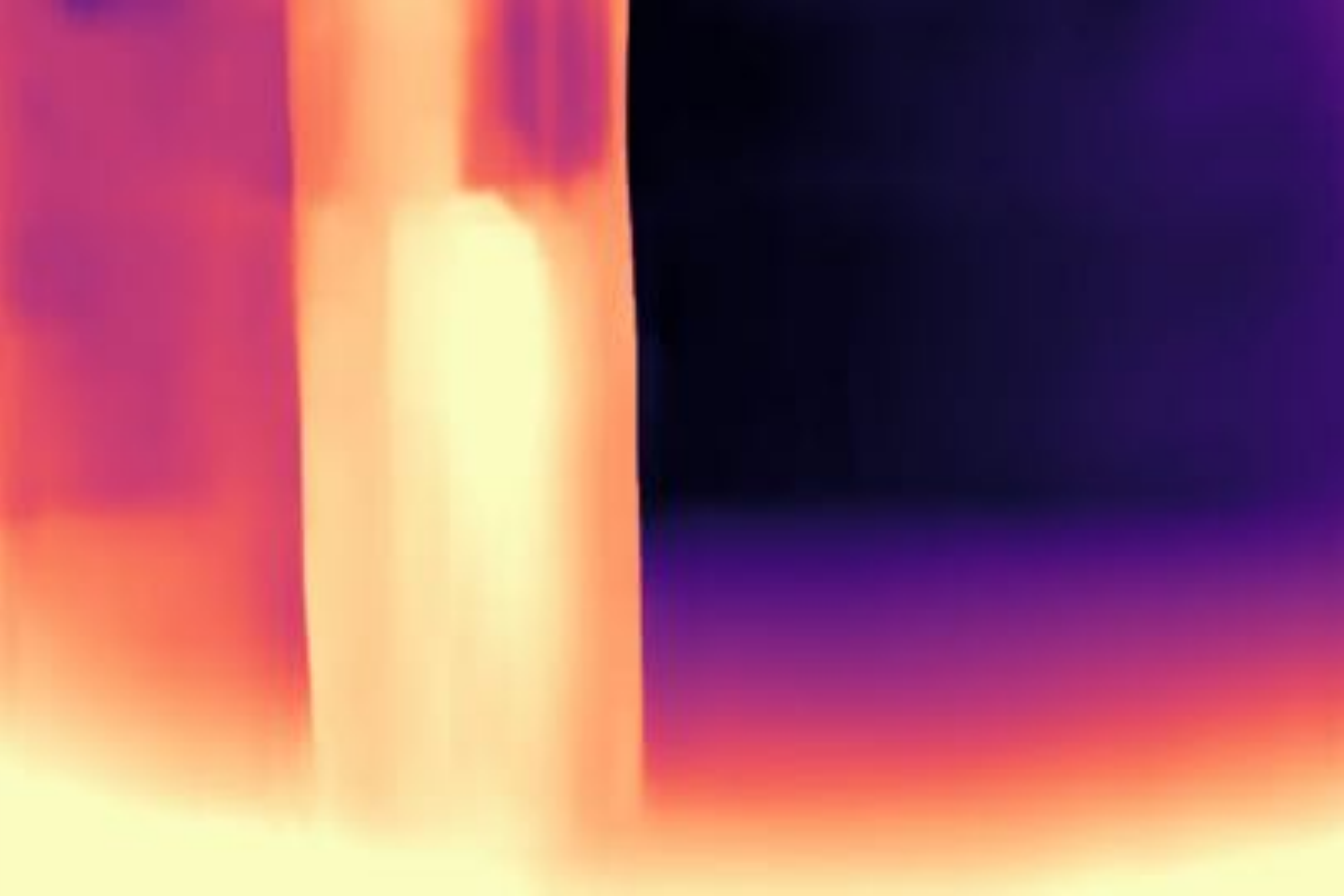}\\ 
        \includegraphics[width=\w,height=\h]{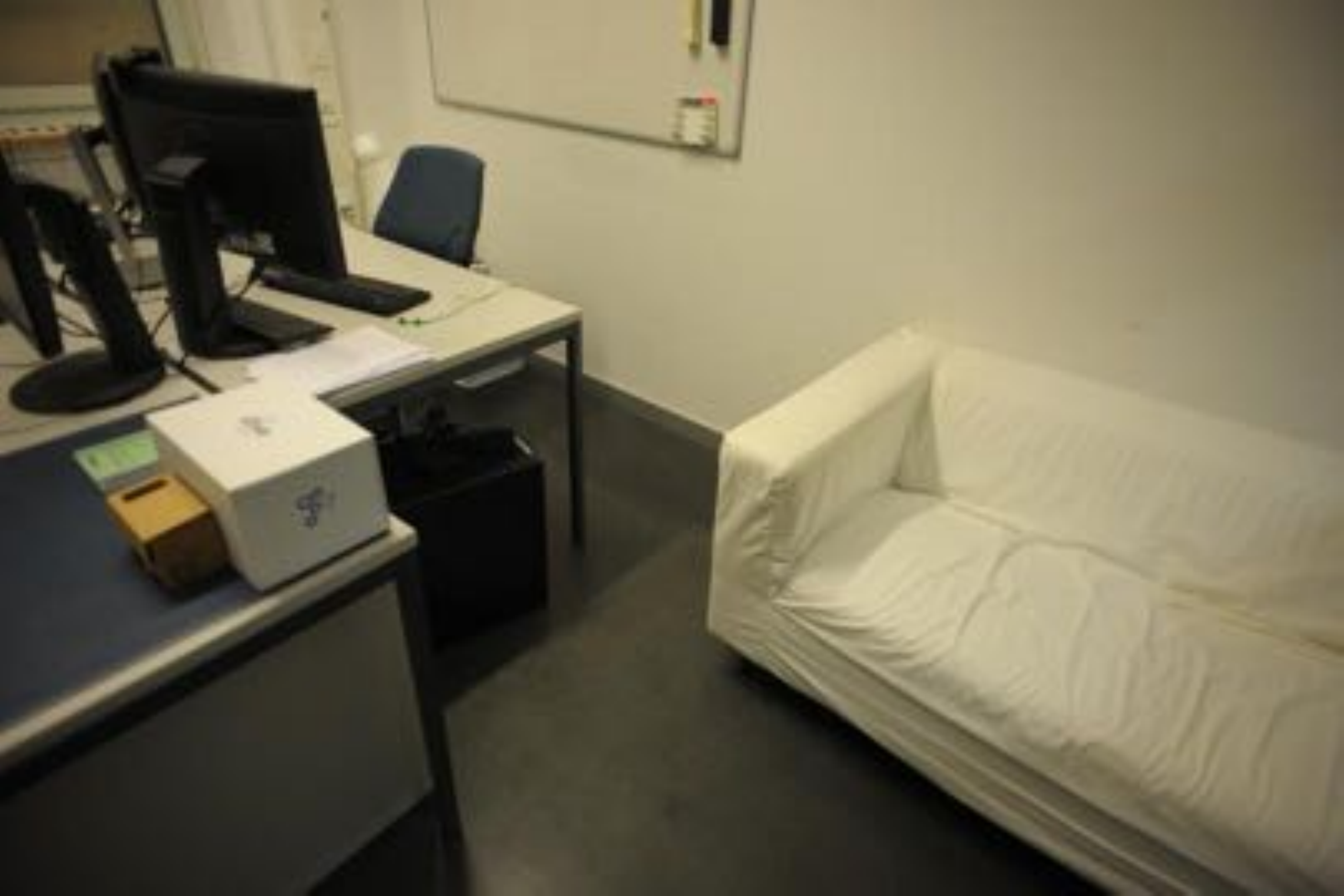}& 
        \includegraphics[width=\w,height=\h]{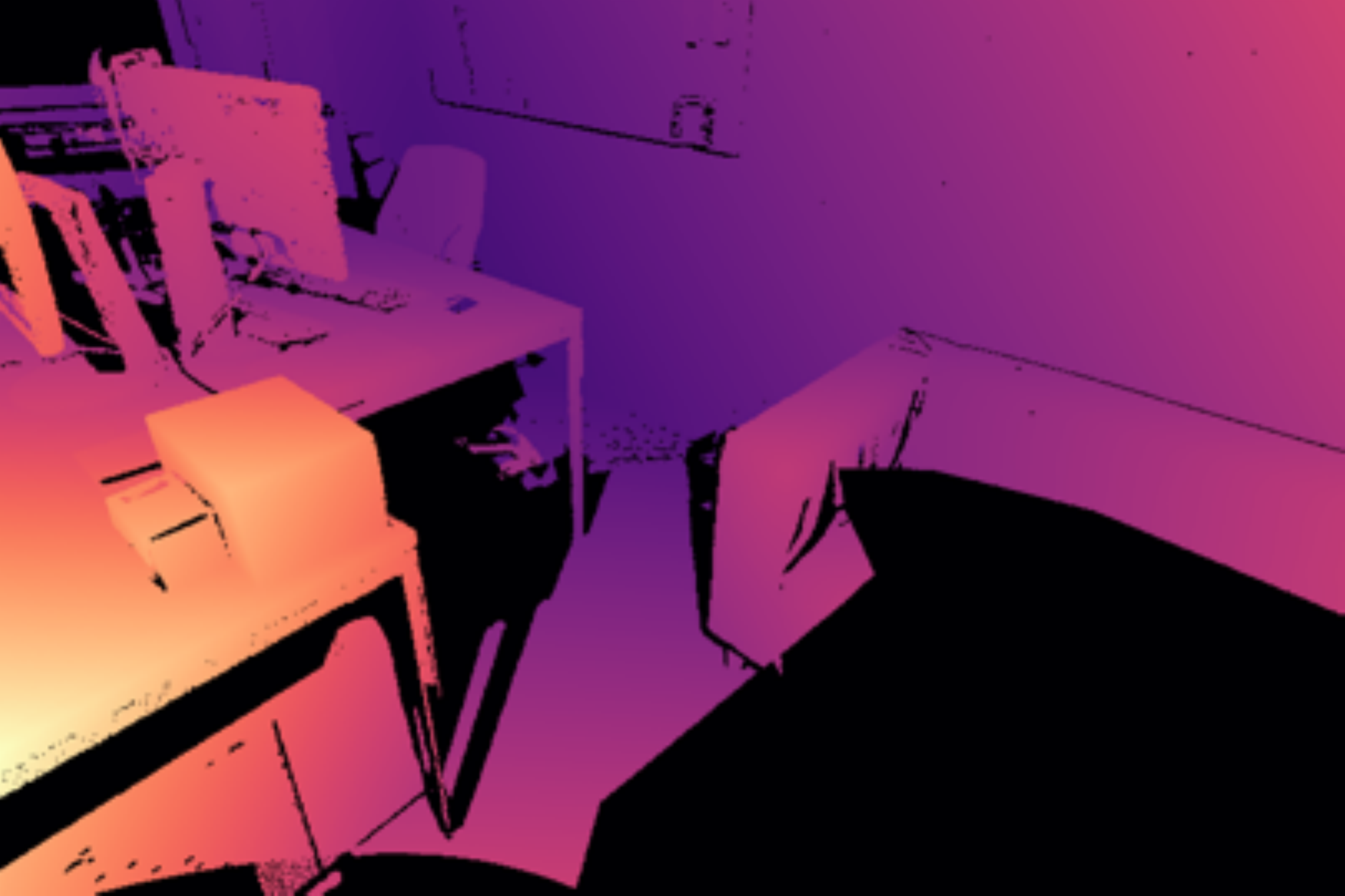}& 
        \includegraphics[width=\w,height=\h]{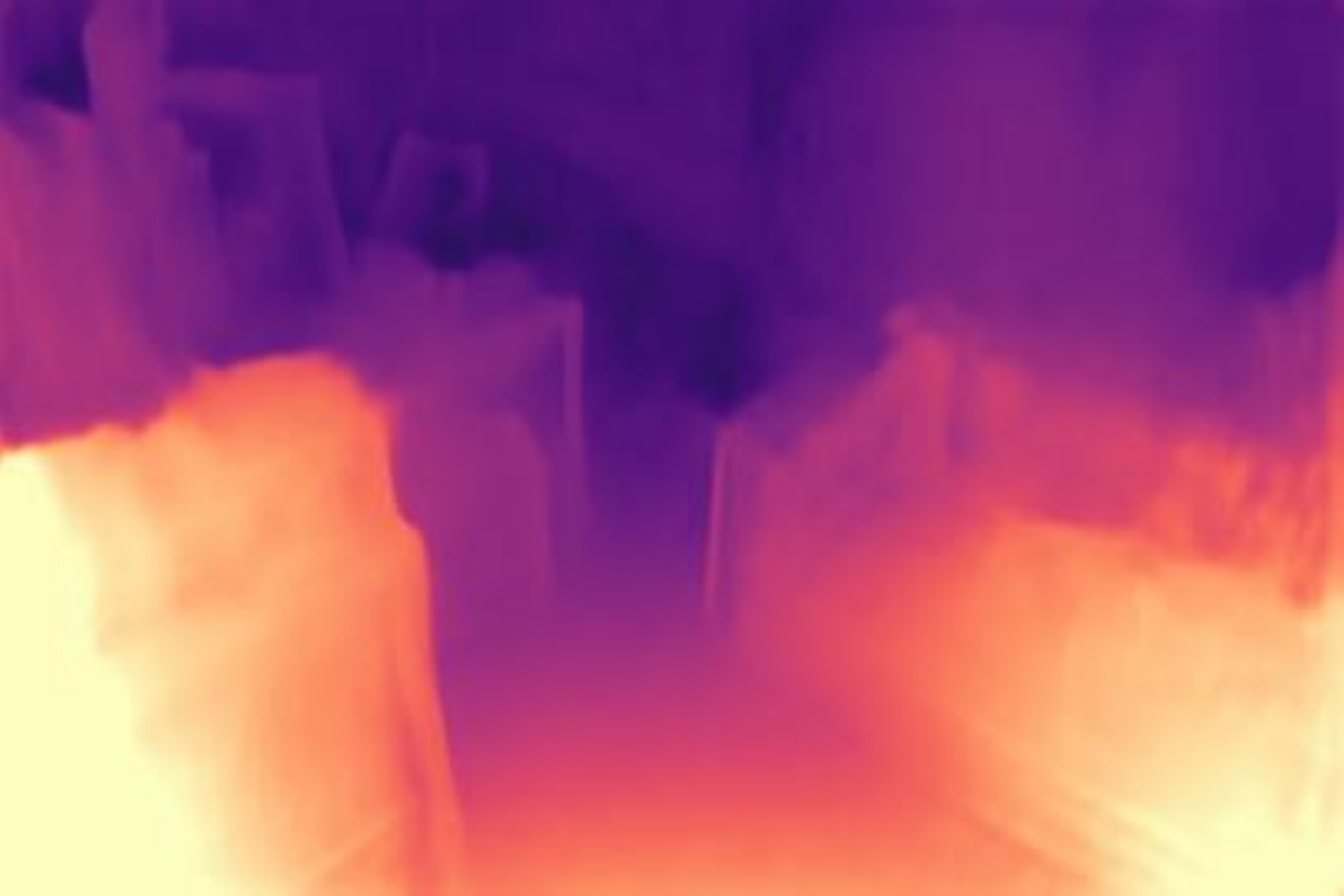}& 
        \includegraphics[width=\w,height=\h]{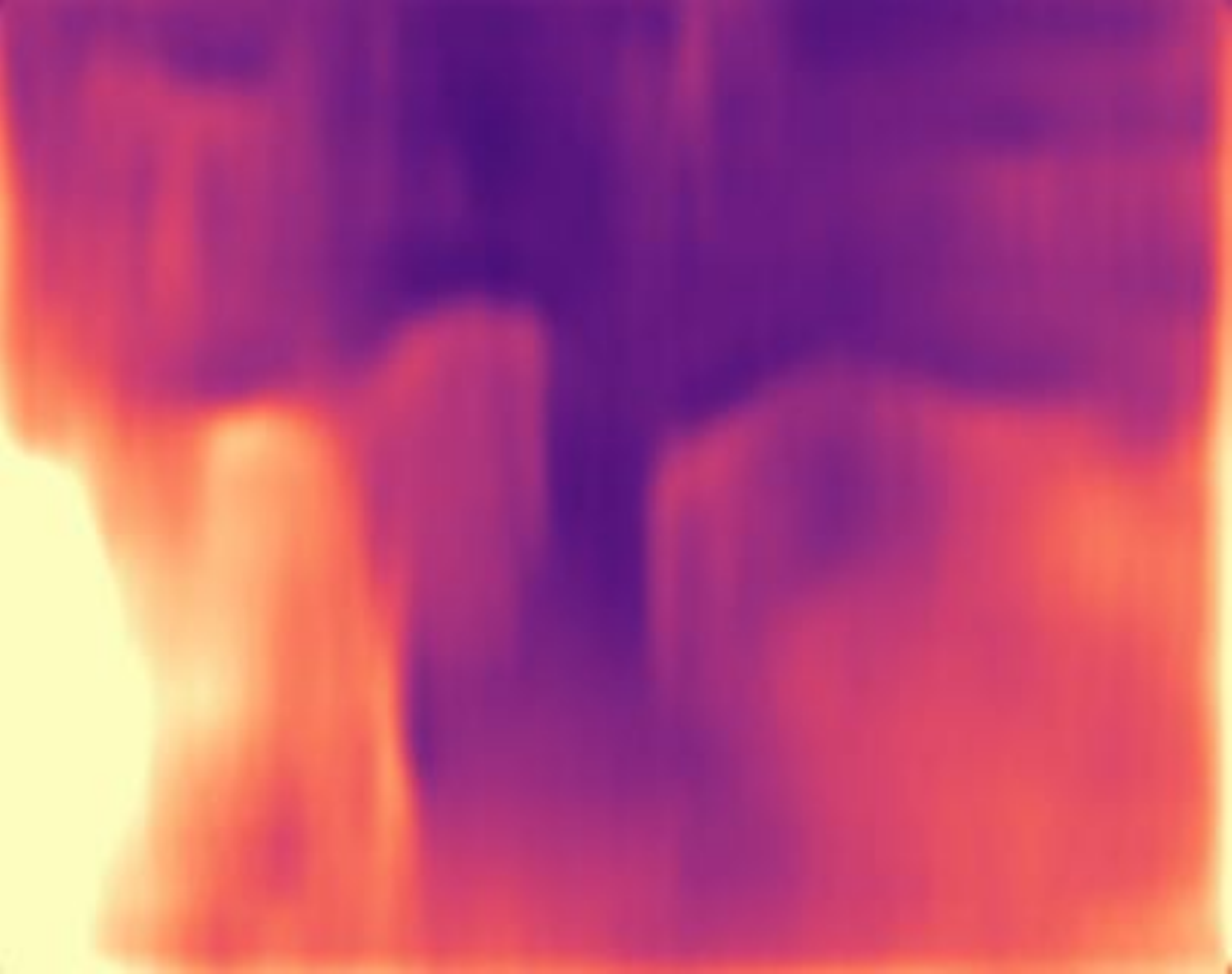}& 
        \includegraphics[width=\w,height=\h]{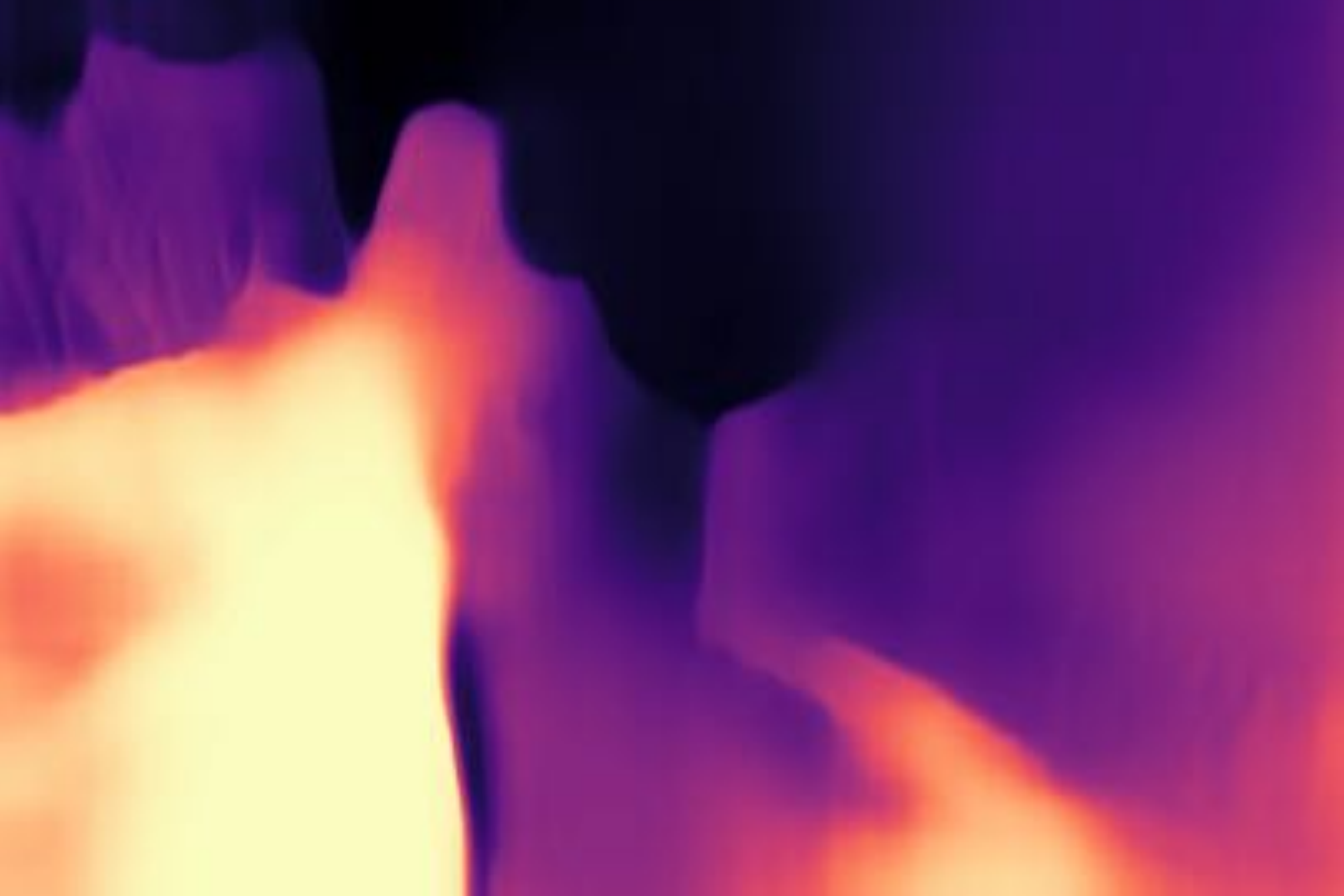}& 
        \includegraphics[width=\w,height=\h]{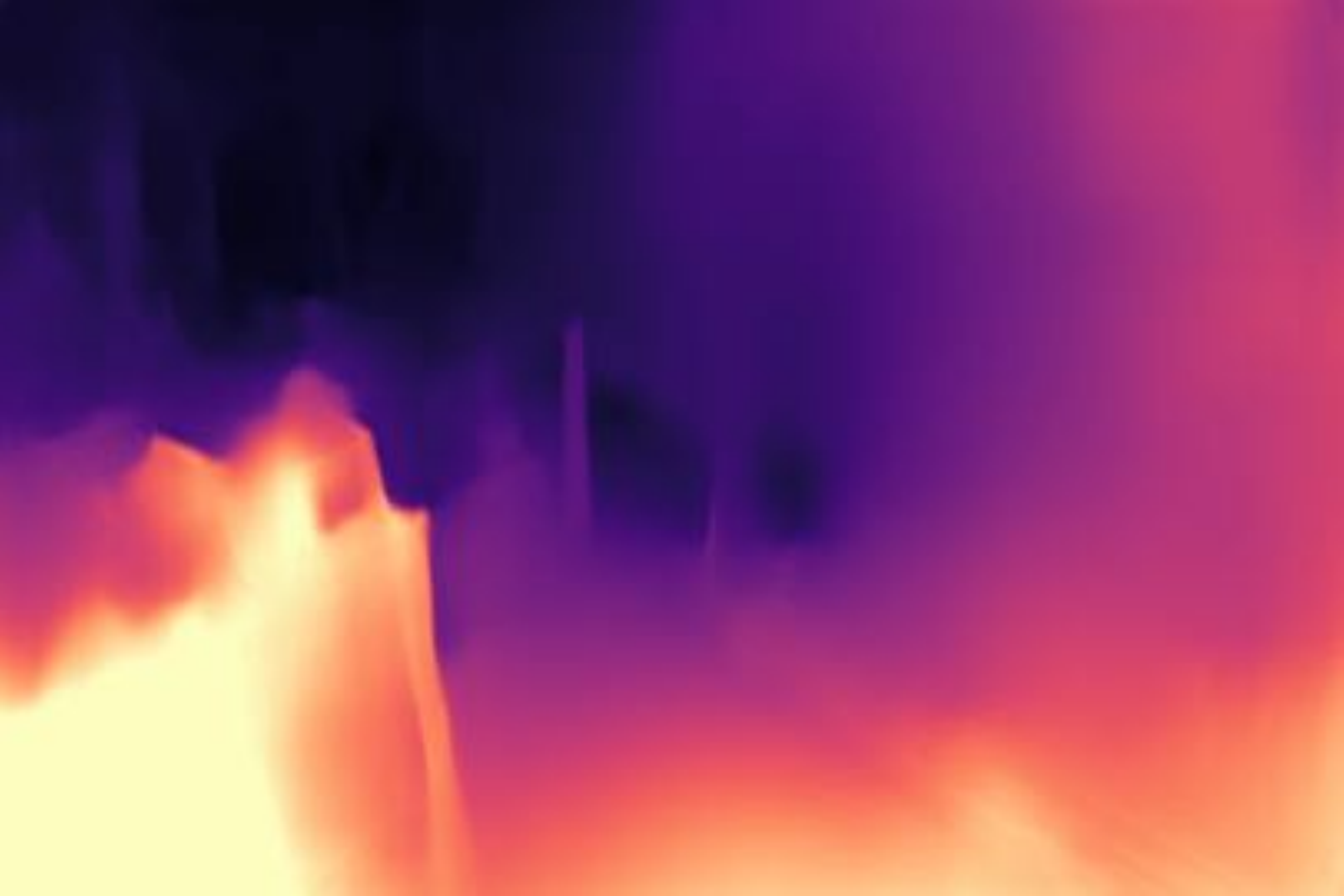}& 
        \includegraphics[width=\w,height=\h]{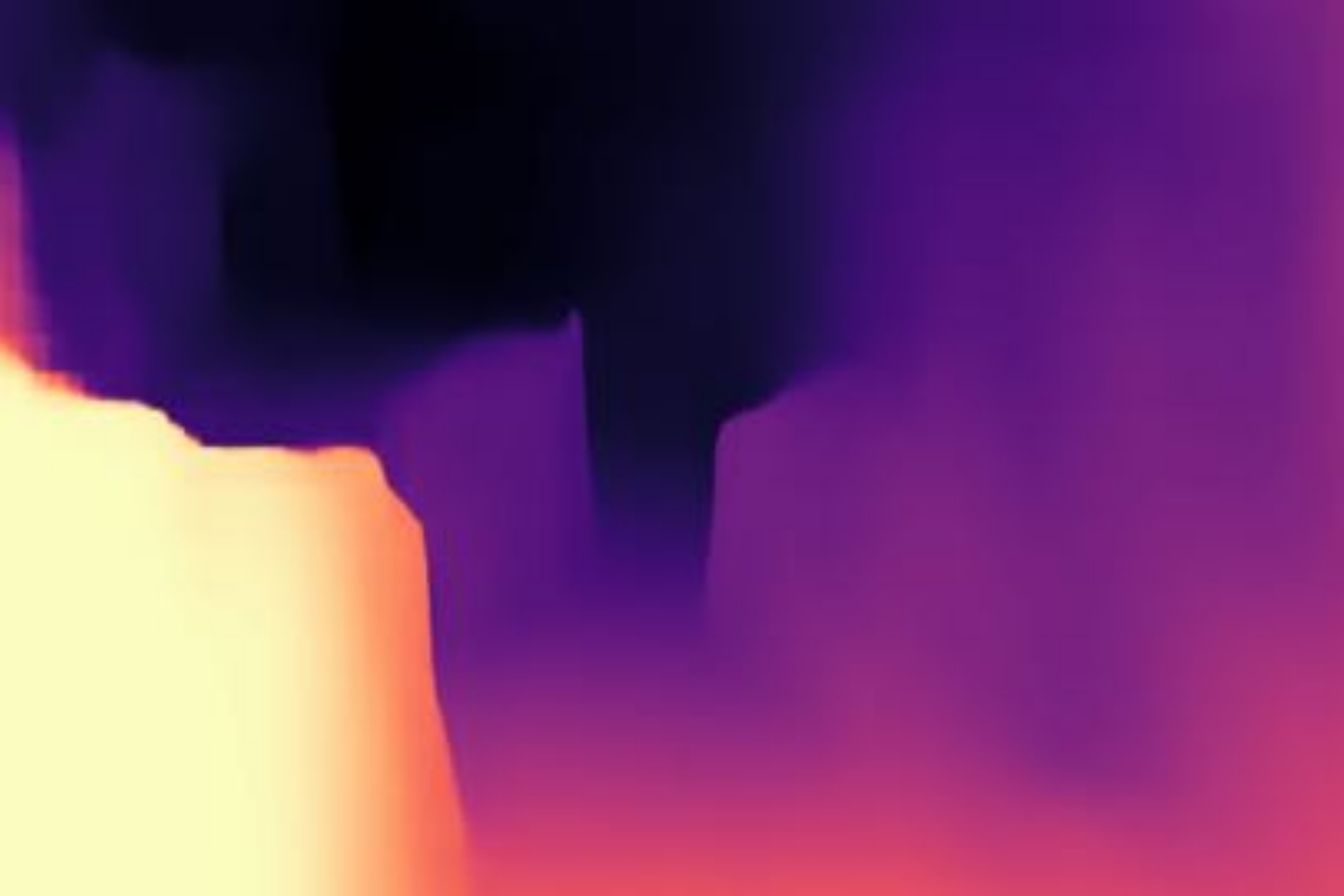}\\ 
     \fontsize{40}{40} \selectfont Input images & 
     \fontsize{40}{40} \selectfont GT depths &
     \fontsize{40}{40} \selectfont Ours-Hybrid &
     \fontsize{40}{40} \selectfont Ours-ViT &
     \fontsize{40}{40} \selectfont Monodepth2 & 
     \fontsize{40}{40} \selectfont PackNet-SfM & 
     \fontsize{40}{40} \selectfont R-MSFM6
    \end{tabular}}
    \caption{\textbf{Comparison of depth map results on various dataset.} We test our model and the competitive models trained on KITTI using MVS, SUN3D, RGBD, Scenes11 and ETH3D (Top to Bottom).}
    \label{figure_result_demons}
     \end{subfigure}
\end{figure*}

\begin{table}[t!]
\resizebox{\columnwidth}{!}{
\begin{tabular}{ccccccccc}
\hline $\text{D}^*$ & Model & Abs Rel $\downarrow$ & Sq Rel $\downarrow$ & RMSE $\downarrow$ &RMSElog $\downarrow$ & $\delta < 1.25$ $\uparrow$ & $\delta < 1.25^2 \uparrow$ & $\delta < 1.25^3 \uparrow$ \\ \hline
\multirow{5}{*}{\rotatebox[origin=c]{90}{Water-color}}  & Monodepth2       & 0.170          & 1.345          & 6.175          & 0.263          & 0.750          & 0.909          & 0.960          \\
                              & PackNet-SfM          & 0.174          & 1.364          & 6.334          & 0.264          & 0.742          & 0.906          & 0.961          \\
                              & R-MSFM6          & 0.194          & 1.613          & 7.173          & 0.302          & 0.696          & 0.876          & 0.943          \\ \cline{2-9} 
                              & Ours-ViT       & 0.152          & 1.196          & 5.668          & 0.232 & 0.799 & 0.932 & 0.973 \\
                              & Ours-Hybrid & \textbf{0.140} & \textbf{1.053} & \textbf{5.665} & \textbf{0.222}          & \textbf{0.815}          & \textbf{0.936}          & \textbf{0.975}          \\ \hline
\multirow{5}{*}{\rotatebox[origin=c]{90}{Pencil-sketch}} & Monodepth2       & 0.196          & 1.522          & 6.232          & 0.276          & 0.691          & 0.898          & 0.962          \\
                              & PackNet-SfM          & 0.204          & 1.569          & 6.568          & 0.290          & 0.670          & 0.888          & 0.957          \\
                              & R-MSFM6          & 0.217          & 1.698          & 6.719          & 0.301          & 0.647          & 0.872          & 0.951          \\ \cline{2-9} 
                              & Ours-ViT       & 0.174          & 1.311          & 5.770 & 0.248          & 0.756          & 0.920          & 0.967          \\
                              & Ours-Hybrid & \textbf{0.151} & \textbf{1.084} & \textbf{5.615}          & \textbf{0.227} & \textbf{0.786} & \textbf{0.934} & \textbf{0.976} \\ \hline
\multirow{5}{*}{\rotatebox[origin=c]{90}{Style-transfer}}  & Monodepth2       & 0.435          & 6.107          & 10.891         & 0.509          & 0.379          & 0.660          & 0.821          \\
                              & PackNet-SfM          & 0.379          & 4.462          & 9.834          & 0.470          & 0.418          & 0.708          & 0.855          \\
                              & R-MSFM6          & 0.394          & 4.667          & 10.214         & 0.490          & 0.399          & 0.680          & 0.837          \\ \cline{2-9} 
                              & Ours-ViT       & 0.378          & 4.854          & 9.869          & 0.449          & \textbf{0.447} & 0.730 & 0.869          \\
                              & Ours-Hybrid & \textbf{0.351} & \textbf{3.847} & \textbf{9.402} & \textbf{0.438} & 0.446          & \textbf{0.737} & \textbf{0.875} \\ \hline
\end{tabular}}
\caption{\textbf{Quantitative comparison on texture-shifted datasets.} $\text{D}^*$ is datasets.}
\label{table_result_texture}
\vspace{-0.4cm}
\end{table}

\subsection{Analysis of Texture-/Shape-Bias on CNN and Transformer}
\label{texture/shaep}

Generally, the texture represents a spatial color or pattern of pixel intensity in an image \cite{armi2019texture}.
To examine the influence of textures on the inference process in detail, we apply three different texture modification strategies including texture-smoothing (Watercolor), texture removal (Pencil-sketch), and texture-transfer (Style-transfer). Extensive details of image generation to facilitate replication are provided in the Appendix.
The generated images and the correspondence results are shown in \figref{figure_result_texture}. The first two images are watercolors, the middle two images and the last two images are pencil-sketch and style transferred images, respectively.
We also conduct the quantitative evaluations in \tabref{table_result_texture} using all of the KITTI test data (697 images).

In this experiment, we compare the performance of CNN-based models (Monodepth2, PackNet-SfM, R-MSFM6), a Transformer-based model (Ours-ViT), and a hybrid (Ours-Hybrid) model.
We note that Ours-Hybrid is equivalent to MonoFormer and Ours-ViT employs only the ViT  \cite{dosovitskiy2020image} encoder structure without any CNN.
Both qualitative and quantitative results of the watercolor data show that both the CNN-based and Transformer-based models produce plausible depth maps. However, the CNN-based model tends to lose more details of the object boundaries and has higher errors than the Transformer-based models.
The experiments with the pencil-sketch data and the style-transfer data show that the Transformer-based models distinguish objects (\eg~pedestrians and cars) and stuff (\eg~walls and roads) better than the CNN-based models.
Specifically, the CNN-based models produce unrecognizable depth maps on style-transfer data due to the loss of original texture information.
These experiments demonstrate our two observations. 
One is that CNNs have a strong texture bias while Transformers have a strong shape bias.
The other is that models with shape bias representation provide better generalization performance for monocular depth estimation compared to models with texture bias. 
Of particular note, MonoFormer (Ours-Hybrid) more precisely preserves object boundaries than Transformer-based model (Ours-ViT). Ours-ViT also generally produces reliable depth thanks to the shape bias of Transformers, but fails to recover details such as a pedestrians. We believe that the proposed multi-level feature fusion module captures both shape bias and the spatial locality bias.

\begin{table}[t]
\centering
\resizebox{\columnwidth}{!}{
\begin{tabular}{cccccccc}
\hline
$\text{D}^*$ & Model & Abs Rel $\downarrow$ & Sq Rel $\downarrow$ & RMSE $\downarrow$ & $\delta < 1.25 \uparrow$ & $\delta < 1.25^2 \uparrow$ & $\delta < 1.25^3 \uparrow$ \\ \hline
\multirow{4}{*}{\rotatebox[origin=c]{90}{MVS}} & Monodepth2 & 0.471 & 0.407 & 0.503 & 0.408 & 0.661 & 0.806 \\
 & PackNet-SfM & 0.449 & 0.295 & 0.429 & 0.397 & 0.670 & 0.837 \\
 & R-MSFM6 & 0.550 & 0.603 & 0.583 & 0.352 & 0.591 & 0.756 \\
 & Ours-ViT & 0.260 & 0.102 & 0.257 & 0.611 & 0.877 & 0.962 \\
 & Ours-Hybrid & \textbf{0.240} & \textbf{0.086} & \textbf{0.242} & \textbf{0.633} & \textbf{0.881} & \textbf{0.972} \\ \hline
\multirow{4}{*}{\rotatebox[origin=c]{90}{RGBD}} & Monodepth2 & 0.610 & 0.508 & 0.488 & 0.292 & 0.520 & 0.681 \\
 & PackNet-SfM & 0.593 & 0.416 & 0.460 & 0.318 & 0.562 & 0.731 \\
 & R-MSFM6 & 0.695 & 0.553 & 0.490 & 0.261 & 0.471 & 0.627 \\
 & Ours-ViT & 0.383 & 0.185 & 0.284 & 0.487 & 0.701 & 0.846 \\
 & Ours-Hybrid & \textbf{0.363} & \textbf{0.137} & \textbf{0.282} & \textbf{0.486} & \textbf{0.744} & \textbf{0.867} \\ \hline
\multirow{4}{*}{\rotatebox[origin=c]{90}{Scenes11}} & Monodepth2 & 1.647 & 0.763 & 0.356 & 0.312 & 0.529 & 0.671 \\
 & PackNet-SfM & 2.065 & 0.837 & 0.330 & 0.310 & 0.530 & 0.674 \\
 & R-MSFM6 & 1.727 & 0.726 & 0.361 & 0.280 & 0.494 & 0.636 \\
 & Ours-ViT & 1.671 & 0.657 & 0.268 & 0.355 & 0.575 & 0.713 \\
 & Ours-Hybrid & \textbf{1.511} & \textbf{0.404} & \textbf{0.255} & \textbf{0.388} & \textbf{0.615} & \textbf{0.755} \\ \hline
\multirow{4}{*}{\rotatebox[origin=c]{90}{SUN3D}} & Monodepth2 & 0.554 & 0.535 & 0.576 & 0.324 & 0.556 & 0.718 \\
 & PackNet-SfM & 0.466 & 0.336 & 0.471 & 0.350 & 0.612 & 0.792 \\
 & R-MSFM6 & 0.523 & 0.406 & 0.506 & 0.310 & 0.544 & 0.721 \\
  & Ours-ViT & 0.289 & 0.163 & 0.298 & 0.554 & 0.810 & 0.910 \\
 & Ours-Hybrid & \textbf{0.245} & \textbf{0.088} & \textbf{0.255} & \textbf{0.582} & \textbf{0.869} & \textbf{0.964} \\ \hline
\multirow{4}{*}{\rotatebox[origin=c]{90}{ETH3D}} & Monodepth2 & 1.007 & 0.780 & 0.396 & 0.318 & 0.536 & 0.687 \\
 & PackNet-SfM & 0.802 & 0.401 & 0.268 & 0.378 & 0.639 & 0.809 \\
 & R-MSFM6 & 0.943 & 0.632 & 0.366 & 0.330 & 0.541 & 0.686 \\
& Ours-ViT & 0.701 & 0.312 & 0.217 & 0.473 & 0.760 & 0.890 \\
 & Ours-Hybrid & \textbf{0.668} & \textbf{0.293} & \textbf{0.189} & \textbf{0.531} & \textbf{0.817} & \textbf{0.926} \\ \hline
\end{tabular}}
\caption{\textbf{Comparison results.} Evaluation of KITTI-trained model on diverse public datasets. $\text{D}^*$ is datasets.}
\vspace{-0.3cm}
\label{table_result_domains}
\end{table}

\subsection{Generalization Performance of CNN-Based, Transformer-Based, and Hybrid Models} 

We compare the generalization performance of all the competitive models and ours trained on the KITTI datasets \cite{geiger2013vision,eigen2015predicting}.
We test the models using public depth datasets consisting of indoor scenes (SUN3D \cite{xiao2013sun3d}, RGBD \cite{sturm2012benchmark}), synthetic scenes from graphics tools (Scenes11 \cite{ummenhofer2017demon}), outdoor building-focused scenes (MVS \cite{ummenhofer2017demon}), and night driving scenes (Oxford Robotcar \cite{maddern20171}).
We also use ETH3D \cite{schops2017multi} containing both indoor and outdoor scenes. 
The results in \figref{figure_result_night} show that the CNN-based models fail to estimate depth even though the scenes from the training and test sets share the stuff (\textit{e.g.} road and sky) and things (\textit{e.g.} cars), while the Transformer-based model keeps the details of object and scene.
We observe that the texture shifts caused by illumination changes confuse the CNN-based model to estimate accurate depth.

The test results on the other scene environments in \figref{figure_result_demons} also show aspects similar to the results in \figref{figure_result_night}.
The transformers-based models recover scene depth even in the complex scenes containing things and stuff which never been seen during training. However, the CNN-based models estimate unreliable depth maps, which keep the infinity depth mostly seen in KITTI datasets and loss the depth boundaries of objects. 
Of particular note, Ours-Hybrid produces more accurate depth maps which preserve the fine structures compared with Ours-ViT. The quantitative evaluations in \tabref{table_result_domains} show that ours outperforms all competitive methods for all datasets and all measurements. 
MonoFormer achieves performance improvement of up to more than 30$\%$ over other CNN-based state-of-the-art models and 7$\%$ over a Transformer-based model (Ours-ViT) on average in Abs Rel.
We believe that our network efficiently combines the local region information from the proposed module while keeping the shape bias representation from Transformers.

\subsection{Analysis of Feature Representation on CNN and Transformers}

Previous works~\cite{geirhos2018imagenet,esser2020disentangling,islam2021shape} propose analysis methods for the representation and the mechanisms of CNNs. They contain the method to quantify the amount of shape information and texture information in the feature representation \cite{islam2021shape}. 
Following the method, we freeze the encoder $E$ of the depth network and input the image $I$ to obtain the encoder's feature $z$ ($z = E(I)$). The mutual relationship between $z_a$ and $z_b$ is obtained using image pairs $(I_a,I_b)$ with specific semantic concepts (i.e., texture or shape features) can be used to quantify the types of features that the network has learned. We measure correlation relationships through a simple correlation coefficient $\rho$ in \eqref{coeff}.
\begin{equation}
\small
    \rho = \frac{\text{Cov}(z_a,z_b)}{\sqrt{\text{Var}(z_a)\text{Var}(z_b)}}.
\label{coeff}
\end{equation}
We estimate shape/texture dimensionality using 697 image pairs (\textit{e.g.}, KITTI eigen test images) in the texture shifted dataset and the original KITTI dataset. We calculate $\rho$ for each image pair and then sum it up. \tabref{shape-texture-est} shows that the features of the Transformer-based model involve more shape information than the CNN-based model.

\begin{table}[t]
\centering
\begin{tabular}{c|cc}
\hline
        Model & Shape & Texture \\ \hline
        Monodepht2 & 273 & 411 \\ \hline
        PackNet-Sfm & 75 & 144 \\ \hline
        R-MSFM6 & 145 & 303 \\ \hline
        MonoFormer-ViT & 697 & 228  \\ \hline
        MonoFormer-Hybrid & 334 & 275 \\ \hline
\end{tabular}
\caption{\textbf{Estimation result of shape/texture dimensionality.} All models are trained on KITTI datasets.}
\label{shape-texture-est}
\end{table}

\subsection{Ablation study}
 
\noindent \textbf{Comparison to various backbones.} We evaluate the performance of models with different backbones in \tabref{table4_ablation_backbone}. We compare ours to models whose encoder was built with either CNNs (ResNet50, ResNet101) or Transformers (ViT-B, Vit-L). We also evaluate another CNN-Transformer hybrid model, TransDepth \cite{yang2021transformer}. The results demonstrate that our model achieves the best performance among them. The numbers show that the model with Transformers (ViT) performs worse than CNN (ResNet). This is because the insufficient number of datasets are used to tackle the lack of inductive bias that Transformers typically struggle with.
The hybrid network (TransDepth) shows better performance than the pure Transformer-based network, but it still underperforms the pure CNN-based network. Meanwhile, our model outperforms CNNs, as well as the other backbone networks. We believe that this is because the proposed method effectively compensates for the lack of inductive bias in Transformers. We note that we use TransDepth whose model is provided by the author \cite{yang2021transformer} and train the model in a self-supervised manner.

\begin{table}[!t]
    \centering
    \small
    \resizebox{\columnwidth}{!}{
    \begin{tabular}{cccccc}
    \hline
    Backbone & Abs Rel $\downarrow$ & RMSE $\downarrow$  &  & $\delta < 1.25 \uparrow$ & $\delta < 1.25^2  \uparrow$ \\ \hline
    ViT-B/16 & 0.118& 4.840 &  & 0.873 & 0.956 \\
    ViT-L/16 & 0.116 & 4.832 &  & 0.875 & 0.957   \\ \hline
    ResNet50 &  0.123 & 4.690  & & 0.884  & \textbf{0.962} \\
    ResNet101 & 0.113 & \textbf{4.565}  &  & 0.875  & \textbf{0.962} \\ \hline
    TransDepth & 0.121 & 4.809&  & 0.865 & 0.957   \\
    Ours & \textbf{0.104} & 4.580 & & \textbf{0.891} & \textbf{0.962}   \\ \hline
    \end{tabular}
    }
    \caption{\textbf{Ablation study on backbone network.} We use only Transformers (ViT), CNNs (ResNet), and hybrid models (TransDepth \cite{yang2021transformer} and ours). ViT-B and ViT-L are the base and large ViT \cite{dosovitskiy2020image}, respectively. TransDepth and ours use the combination of ResNet50 and ViT-B/16. }
    \vspace{-2mm}
\label{table4_ablation_backbone}
\end{table}

\noindent \textbf{Comparison to the conventional hybrid models} 
We compare our model with existing CNN-Transformer hybrid models, TransDepth \cite{yang2021transformer}. The original TransDepth model is trained with a large number of various datasets in a supervised manner. For a fair comparison, we train the author-provided TransDepth with KITTI eigen split in a self-supervised manner. We conduct the quantitative comparison using the five out-of-distribution datasets as well as the KITTI datasets. The results in \tabref{tab:transdepth} show the performance improvement ratio from TransDepth to MonoFormer. The experiments show that the proposed method achieves performance improvement around 15$\%$ on average in Abs Rel over TransDepth. These results show that MonoFormer outperforms all the conventional hybrid models.

\begin{table}[t]
    \centering
    \small
    \resizebox{\columnwidth}{!}{
    \begin{tabular}{cccccc}
    \hline
    Datasets & Abs Rel & RMSE   &  & $\delta < 1.25$  & $\delta < 1.25^2 $  \\ \hline
    KITTI & 14.1$\%$ $\downarrow$ & 4.77$\%$ $\downarrow$ &  & 3.00$\%$ $\uparrow$ & 0.52$\%$ $\uparrow$ \\
    MVS & 19.4$\%$ $\downarrow$ & 19.2$\%$ $\downarrow$ &  & 13.1$\%$ $\uparrow$ & 5.7$\%$ $\uparrow$ \\
    RGBD &  15.6$\%$ $\downarrow$ &	16.4$\%$ $\downarrow$ &  &	13.3$\%$ $\uparrow$ &	7.7$\%$ $\uparrow$ \\
    Scenes11 & 8.7$\%$ $\downarrow$ & 11.3$\%$ $\downarrow$ &  &	10.3$\%$ $\uparrow$ &	4.9$\%$ $\uparrow$  \\ 
    SUN3D & 31.5$\%$ $\downarrow$ & 30.7$\%$ $\downarrow$ &  & 36.9$\%$ $\uparrow$ &	20.6$\%$ $\uparrow$  \\ 
    ETH3D & 6.7$\%$ $\downarrow$ & 18.3$\%$ $\downarrow$ & &	9.9$\%$ $\uparrow$ &	8.4$\%$ $\uparrow$  \\ \hline
    \end{tabular}}
    \caption{\textbf{Comparison to another hybrid model.} The error (Abs Rel, RMSE) reduction and accuracy ($\delta < 1.25$, $\delta < 1.25^2 $) improvement percentage from TransDepth~\cite{yang2021transformer} to our Monoformer.} 
    \label{tab:transdepth}
\end{table}

\begin{figure}[!t] 
    \newcommand\w{4cm}
    \centering
    \begin{tabular}{c@{\hspace{1mm}}c@{\hspace{1mm}}}
    \includegraphics[width=\w]{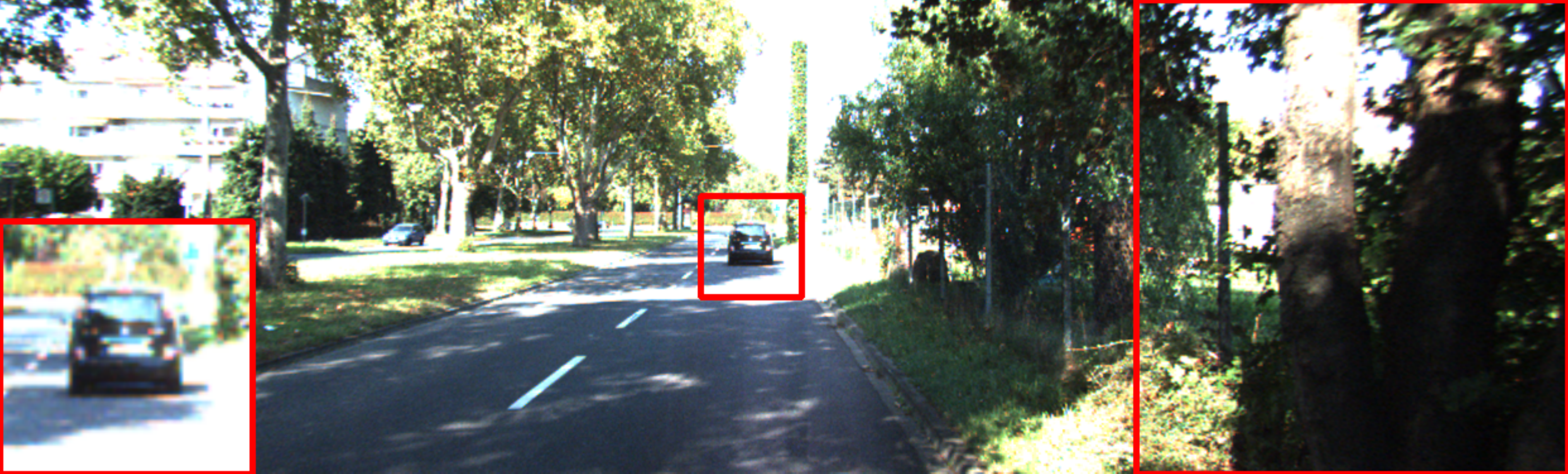} &
    \includegraphics[width=\w]{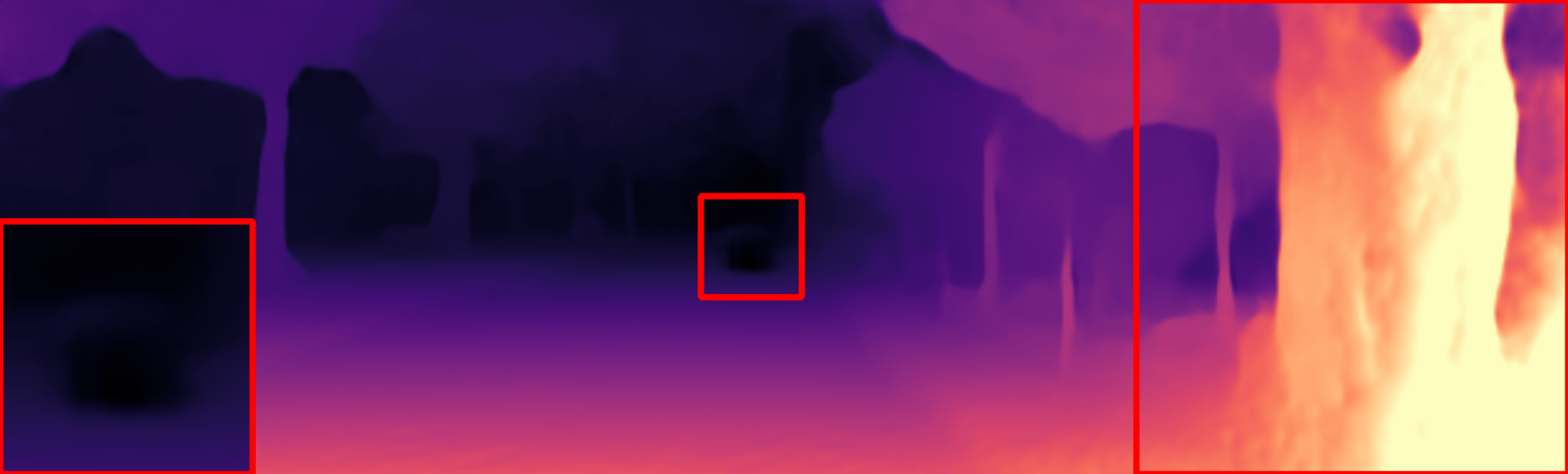}  \\
    \small{(a) Input image}  &
    \small{(b) Results with ACM}  \\
    \includegraphics[width=\w]{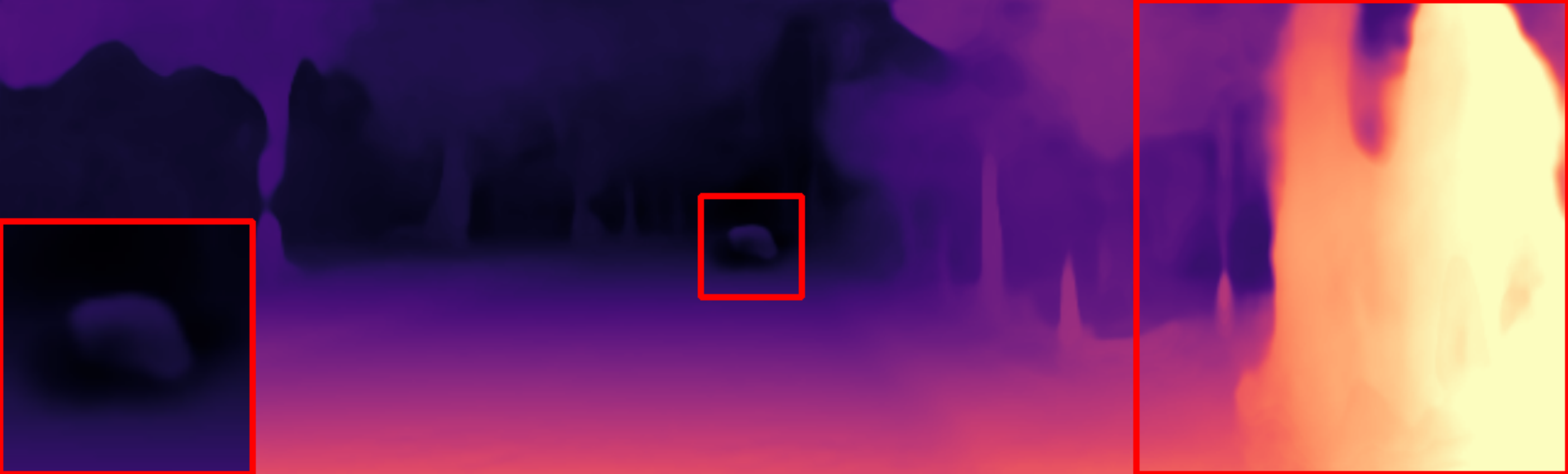}  & \includegraphics[width=\w]{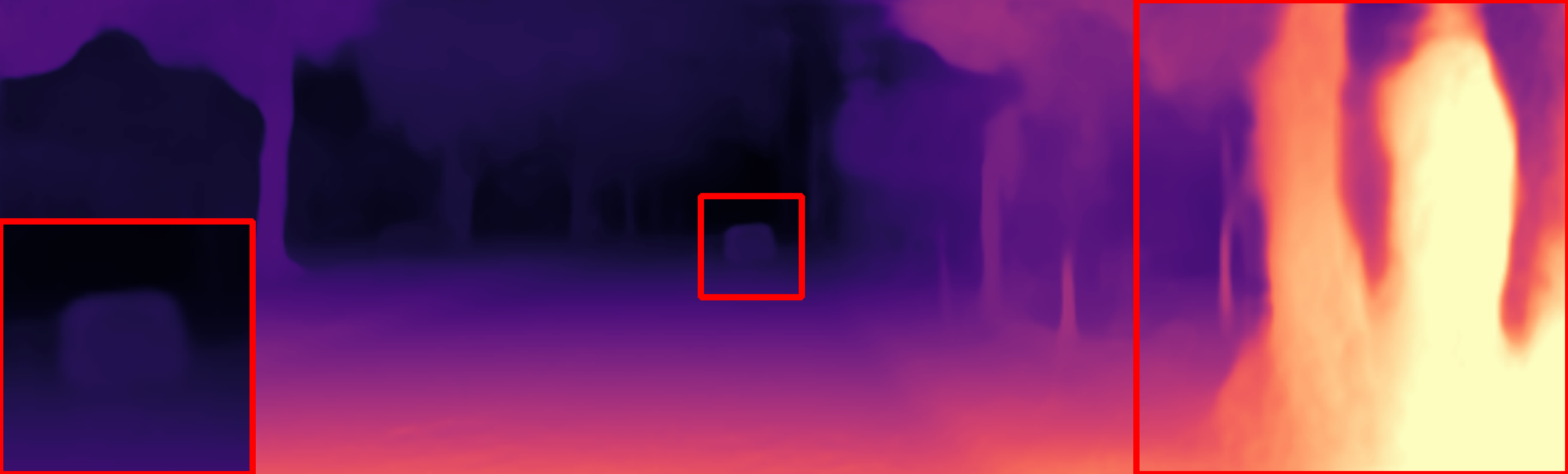}  \\
    \small{(c) Results with FFD}  &
    \small{(d) Results with ACM and FFD}
    \end{tabular}
        \captionof{figure}{\textbf{Visualization of results with/without ACM and FFD.}}
    \label{figure_ablation_modules}
\end{figure}

\noindent \textbf{Effectiveness of the proposed modules.} We conduct an ablation study to demonstrate the effectiveness of the proposed modules, ACM and FFD in \tabref{table_result_modules}.
The baseline is DPT \cite{ranftl2021vision}.
The models with only the ACM module or FFD module marginally improve the depth estimation performance, due to the absence of proper attention map fusions. On the other hand, our MonoFormer with both ACM and FFD significantly improves the performance. The results show the proposed model achieves the best performance in all measurements. The qualitative comparison in \figref{figure_ablation_modules} shows that the model with both ACM and FFD keeps clearer object boundaries, even a small car in far depth.

\begin{figure}[t!]
    \centering
    \newcommand\w{10cm}
    \resizebox{\columnwidth}{!}{
    \begin{tabular}{cc}
    \includegraphics[width=\w]{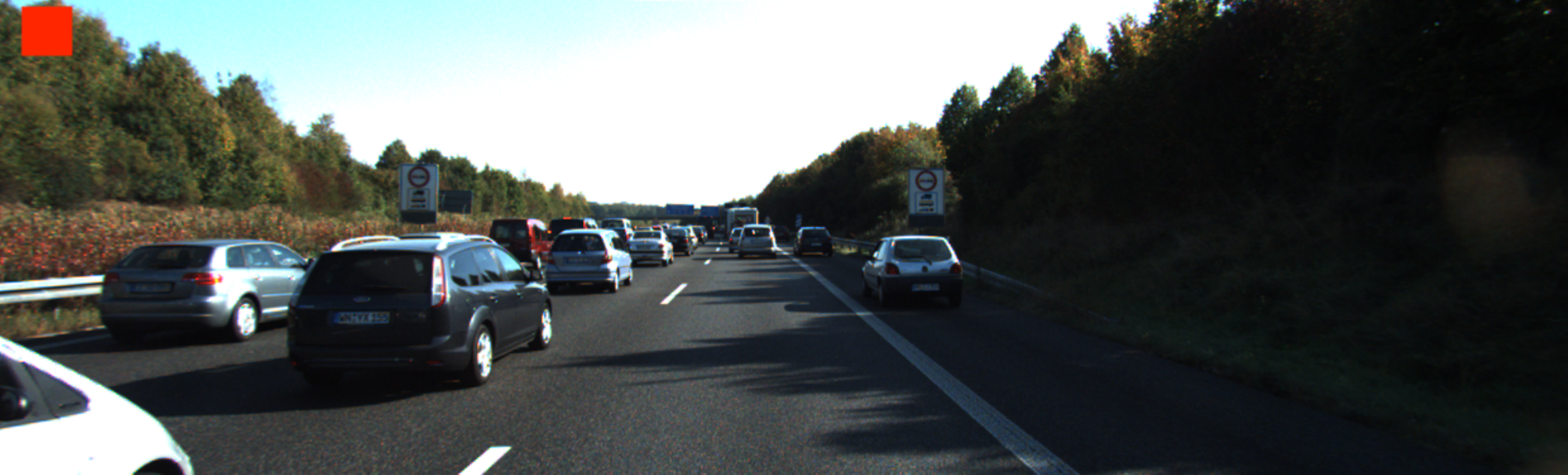}&\includegraphics[width=\w]{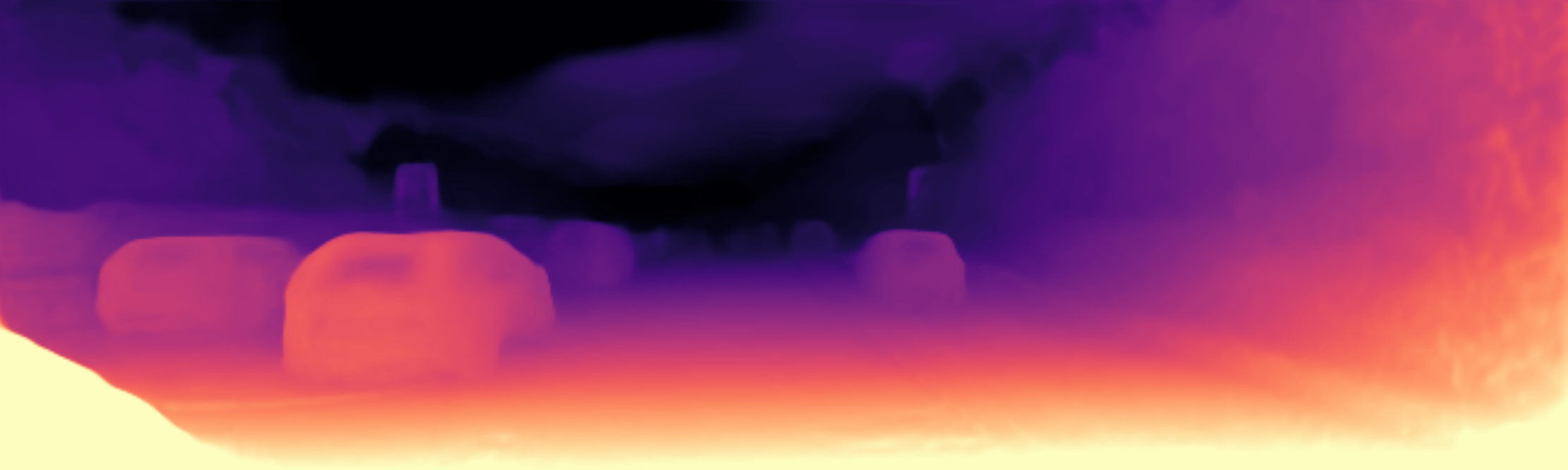}            \\
    \fontsize{20}{20} \selectfont Image $I$           &  \fontsize{20}{20} \selectfont Depth $D$         \\
    \includegraphics[width=\w]{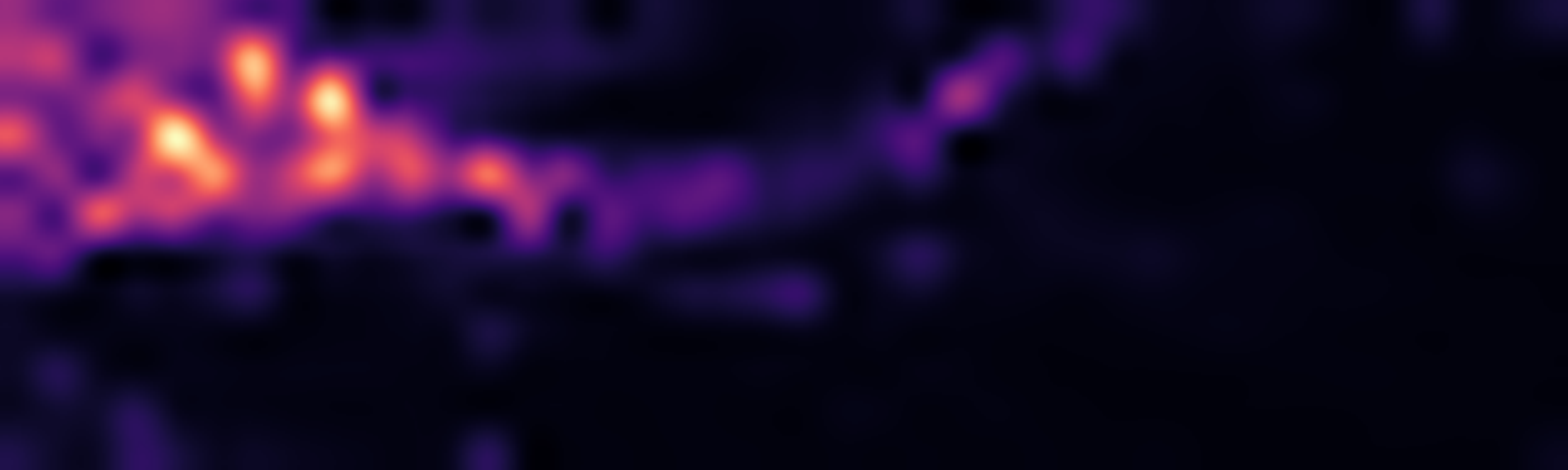}           &   \includegraphics[width=\w]{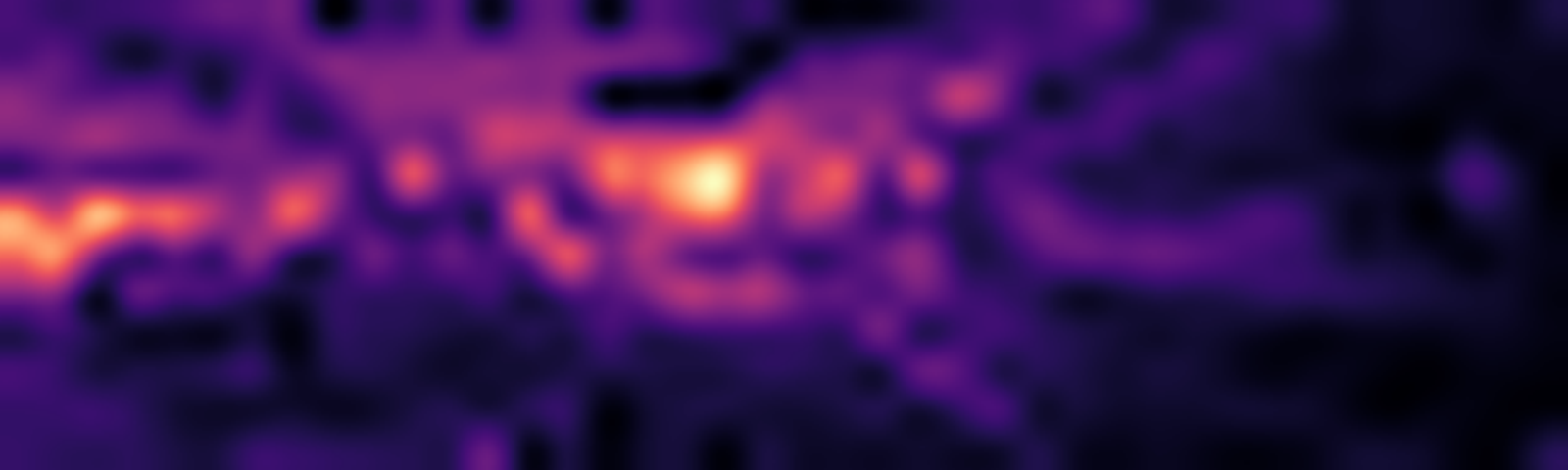}         \\
    \multicolumn{2}{c}{\fontsize{20}{20} \selectfont (a) Attention map of CNN-Transformer encoder } \\
    \includegraphics[width=\w]{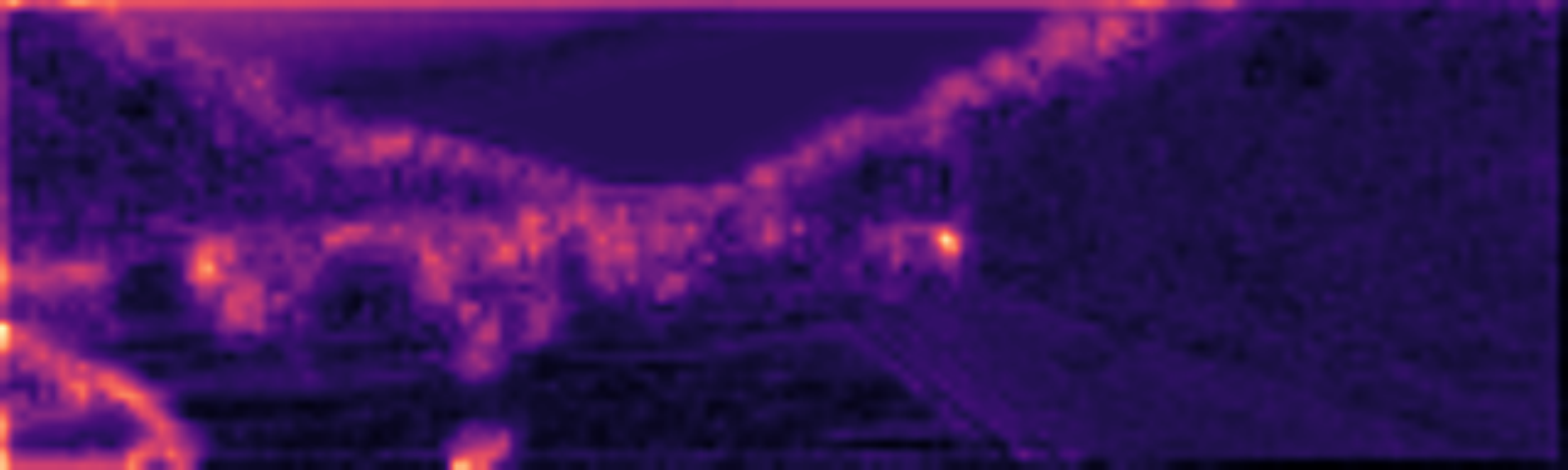}&\includegraphics[width=\w]{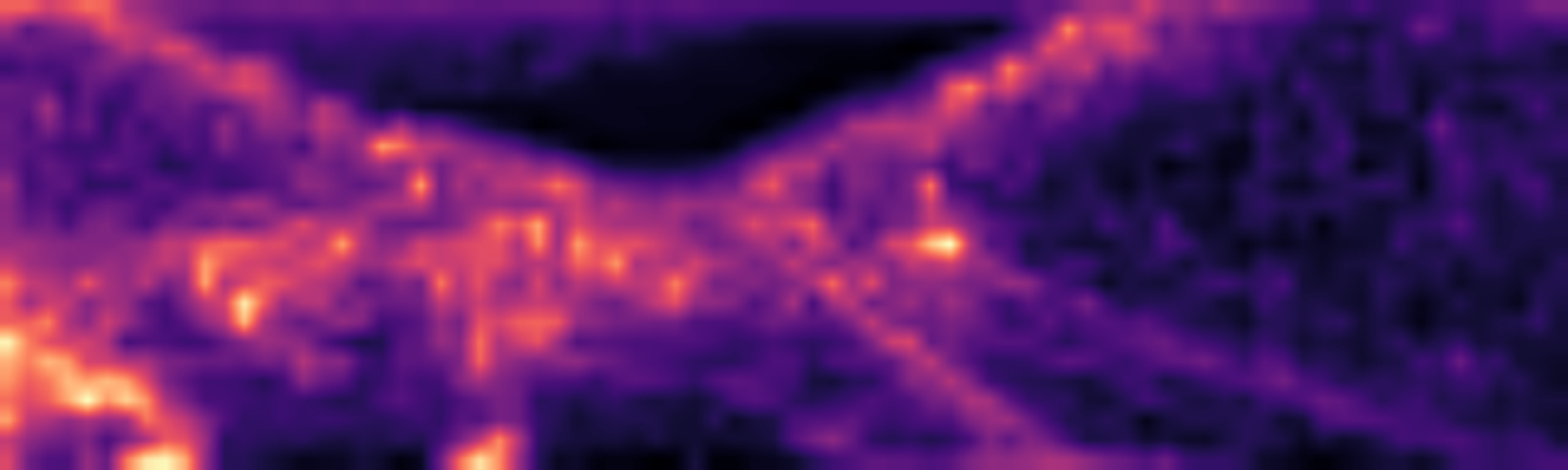}            \\
    \multicolumn{2}{c}{\fontsize{20}{20} \selectfont (b) Attention map of ACM} \\
    \includegraphics[width=\w]{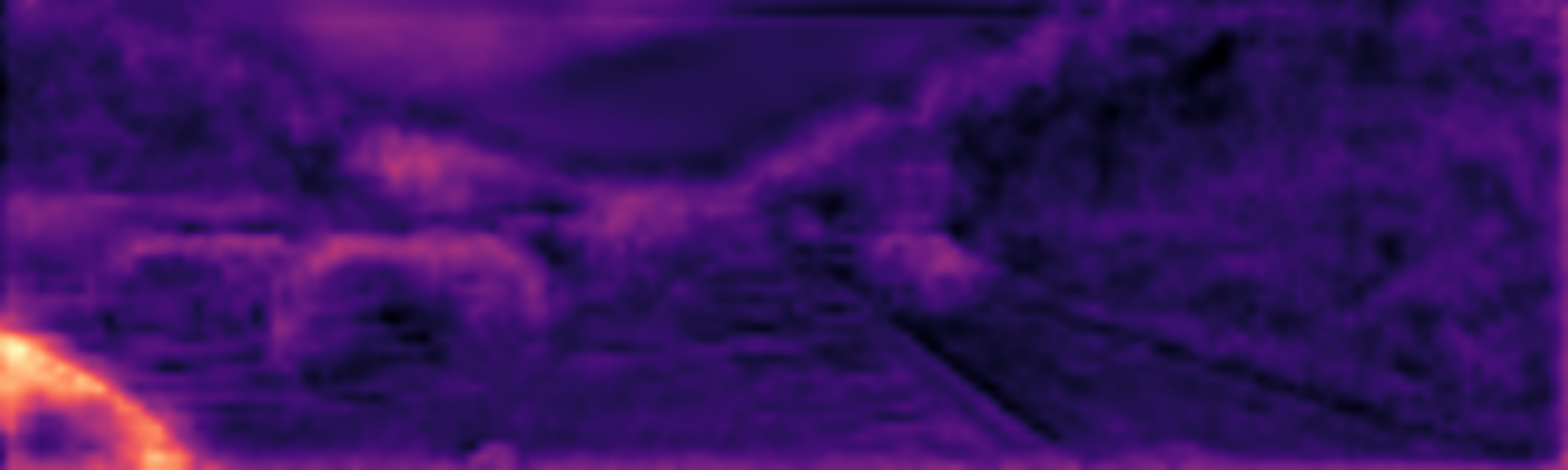}           &\includegraphics[width=\w]{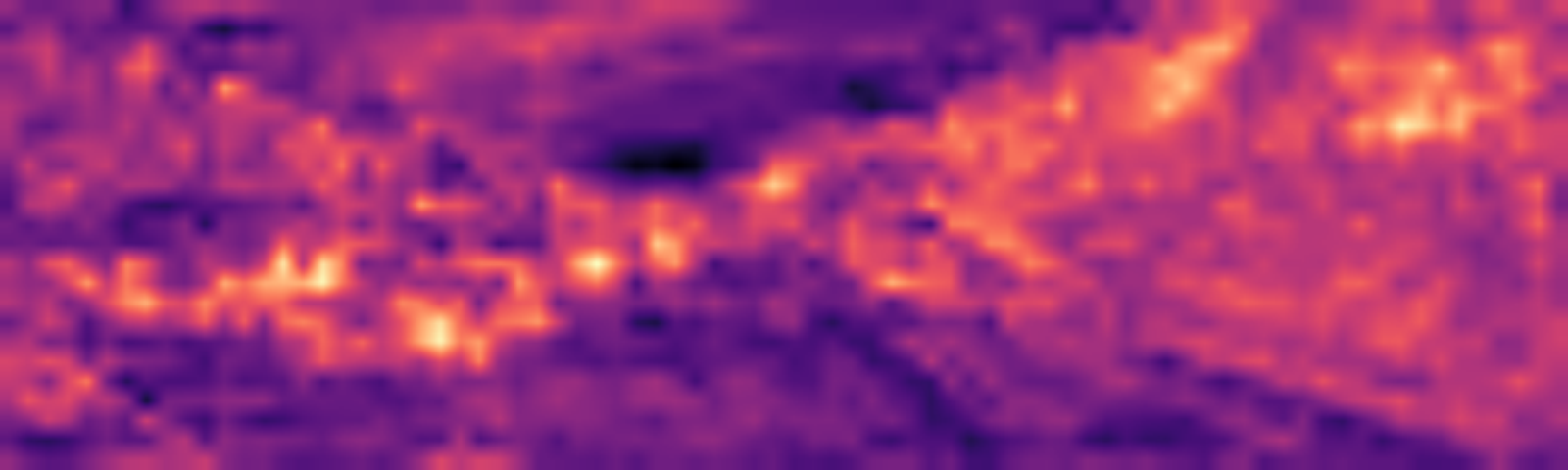}            \\
    \multicolumn{2}{c}{\fontsize{20}{20} \selectfont (c) Feature map of FFD}
    \end{tabular}
    }
        \captionof{figure}{\textbf{Visualization of attention map and feature map}. We visualize the self-attention map of the patch on the upper left corner of the image $I$. The left column from the second row is the attention map from shallow layers, whereas the right is the map from deep layers.}
    \label{figure_monoformer_attention}
    \vspace{-0.4cm}
\end{figure}

\noindent \textbf{Visualization of attention maps.}
We visualize the attention maps from the lower to higher layers of Transformers. As shown in \figref{figure_monoformer_attention}, the encoder in the shallow layer extracts local region features. The deeper the layer, the more global shape contexts are extracted. Another observation is that ACM captures more detailed attention at different depths of the encoder features. FFD enhances the encoder features by fusing them with the attention map from ACM. The fused feature captures features from coarse to fine details. These experiments show that our model is capable of accurate pixel-wise prediction as it secures adequate local details. 

\section{Conclusion}
\label{conclusion}
\begin{table}[t]
    \centering
    \resizebox{\columnwidth}{!}{
    \begin{tabular}{llllll}
    \hline
    & Abs Rel $\downarrow$ & Sq Rel $\downarrow$ & RMSE$\downarrow$ & RMSE$_{log}$ $\downarrow$ & $\delta <1.25 $ $\uparrow$ \\ \hline
    baseline     &   0.113 & 0.899  & 4.783& 0.189 & 0.882      \\
    +ACM         &   0.113 & 0.879  & 4.820& 0.189 & 0.879      \\
    +FFD         &   0.112 & 0.860  & 4.803& 0.186 & 0.879       \\
    +Both &   \textbf{0.104}  & \textbf{0.846} & \textbf{4.580} & \textbf{0.183} & \textbf{0.891}       \\ \hline
    \end{tabular}}
    \caption{\textbf{Ablation study on ACM and FFD.} Both is with ACM and FFD.}
    \label{table_result_modules}
\end{table}
In this paper, we provide three important observations for the self-supervised monocular depth estimation task: 1) \textit{ CNN-based models rely heavily on textures, while Transformer-based models rely on shapes for a monocular depth estimation task.} 2) \textit{Texture-based representations leads to poor generalization performance with texture-shift such as scene changes, illumination changes, and style changes.} 3) \textit{Shape-based representations are more helpful for a generalized monocular depth model than texture-based representations.}
Based on these observations, we propose a CNN-Transformer hybrid network, called MonoFormer, which incorporates both shape bias and spatial locality bias. The proposed model achieves the best performance among various competitive methods on diverse unseen datasets as well as KITTI datasets, by a high margin. The extensive experiments demonstrate that our MonoFormer has superior generalization ability.
We believe that the performance improvement comes from the design of strong shape-biased models, and this observation can be a useful insight to better understanding of monocular depth estimation. 

\appendix

\section{Appendix}

\subsection{Additional Qualitative Results}
\begin{figure*}[!t]
     \newcommand\w{10cm}
    \newcommand\h{5cm}
    \begin{subfigure}
       \centering
    \resizebox{\linewidth}{!}{
    \begin{tabular}{c@{\hspace{1mm}}c@{\hspace{1mm}}c@{\hspace{1mm}}c@{\hspace{1mm}}c@{\hspace{1mm}}}
     \includegraphics[]{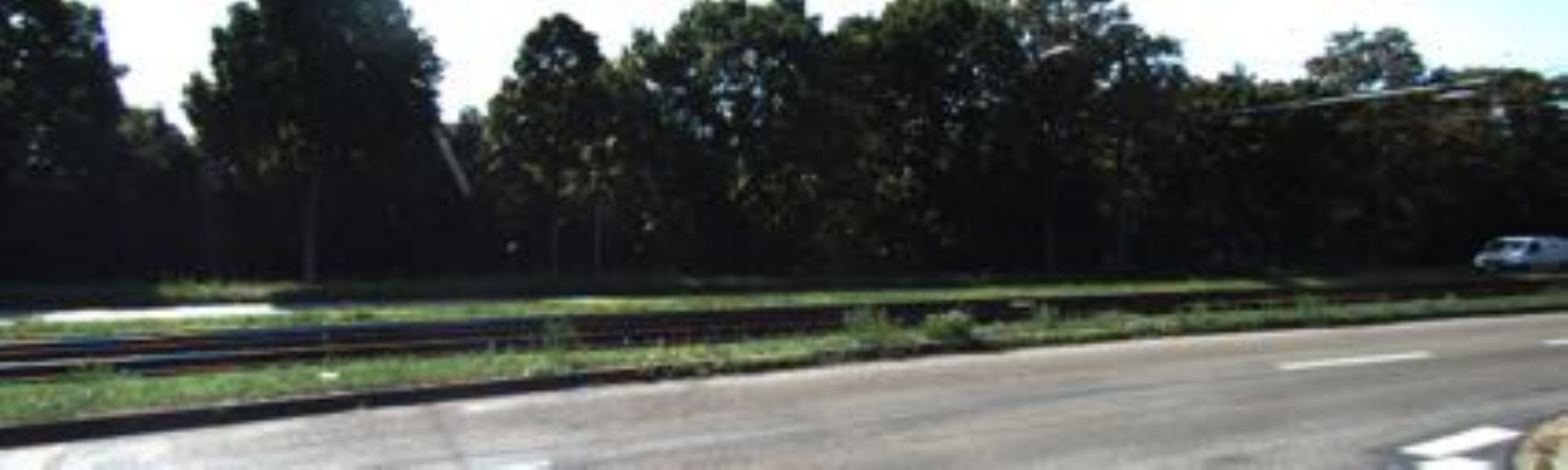}&
     \includegraphics[]{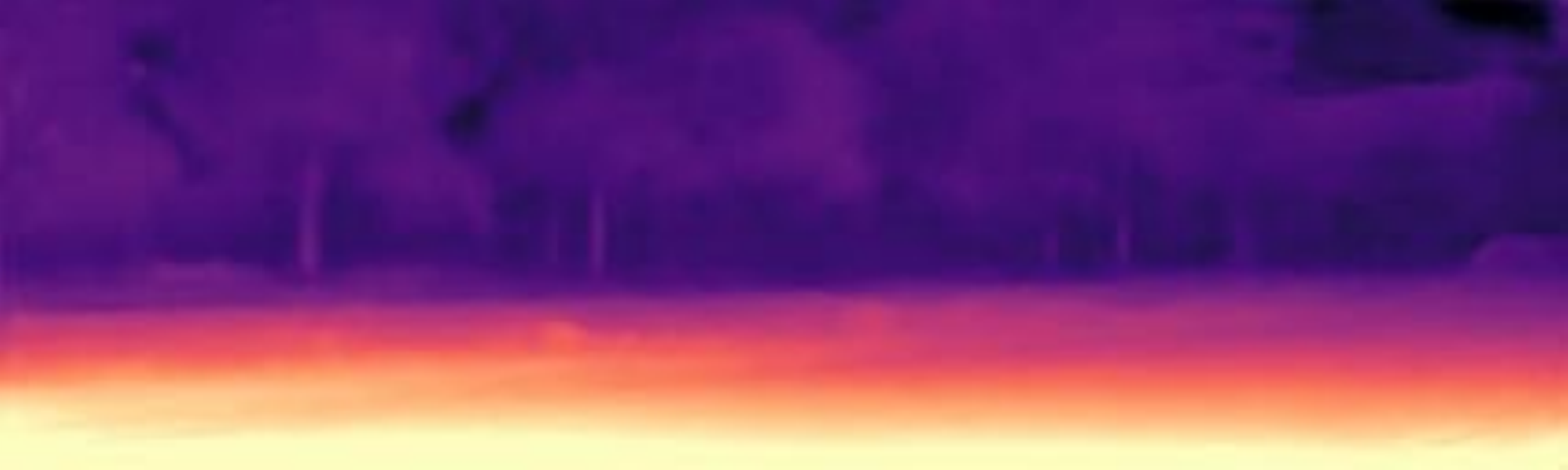}&
     \includegraphics[]{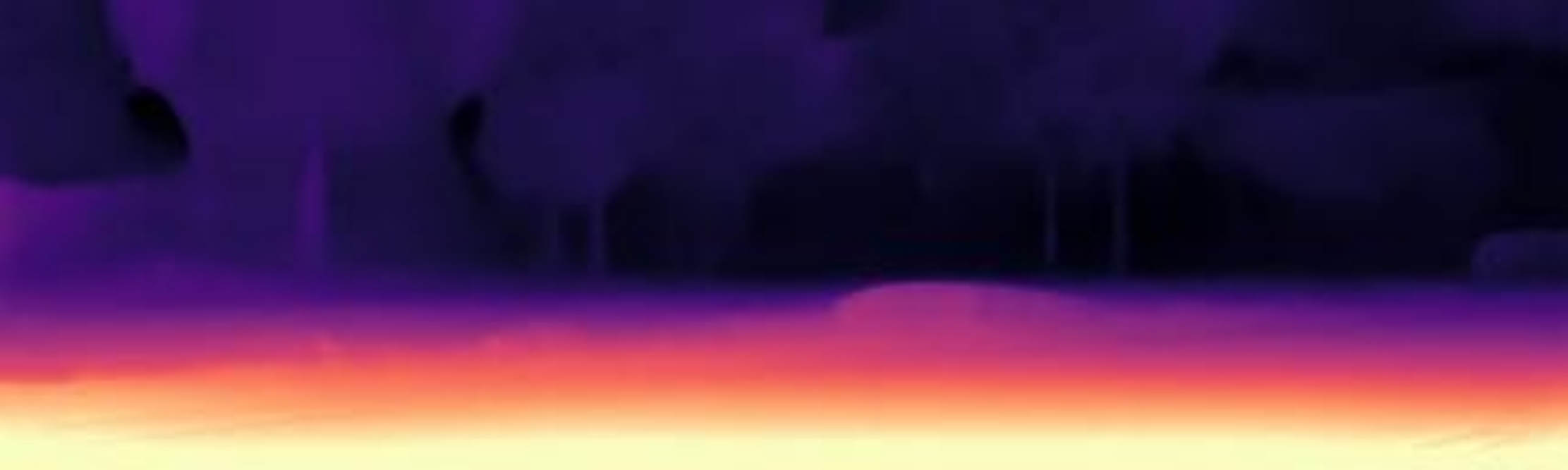}&
     \includegraphics[]{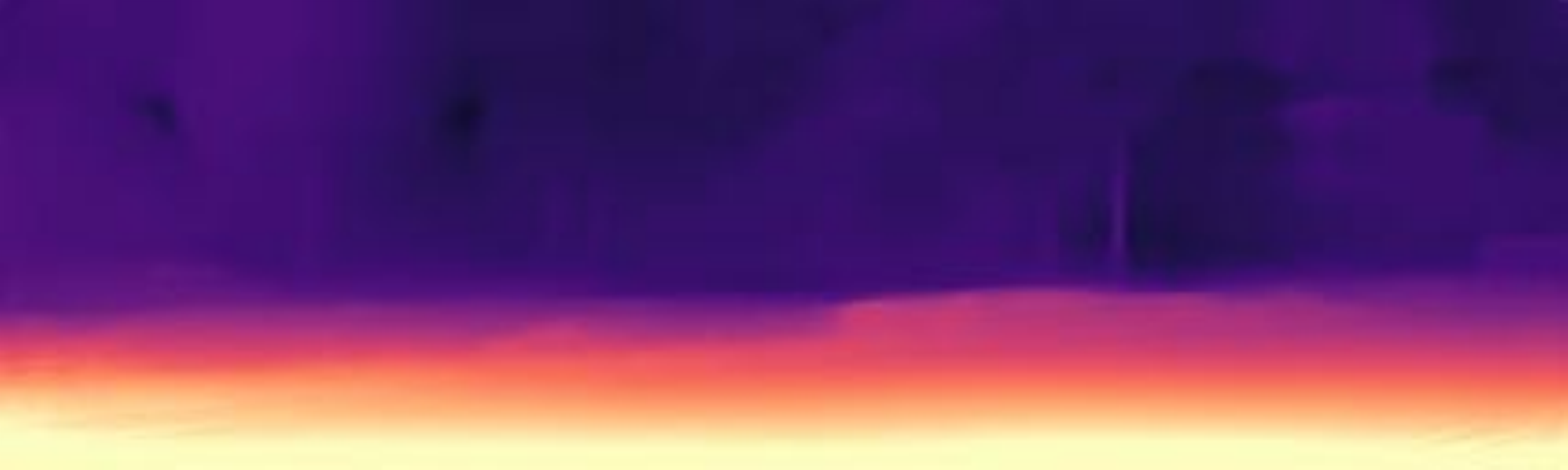}&
     \includegraphics[]{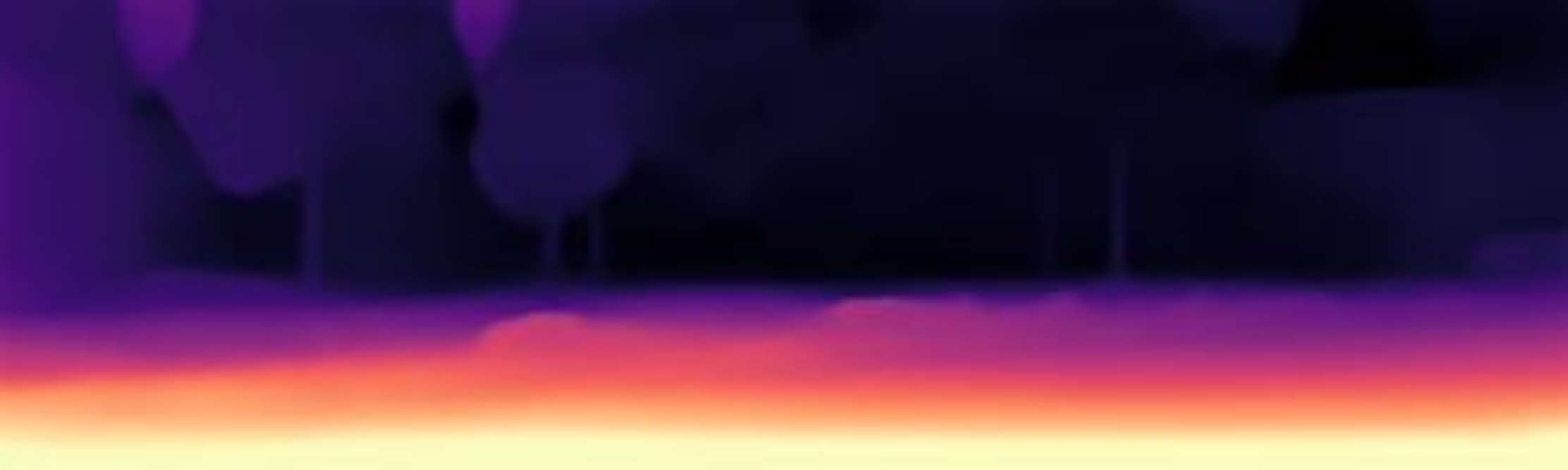}\\
     
    \includegraphics[]{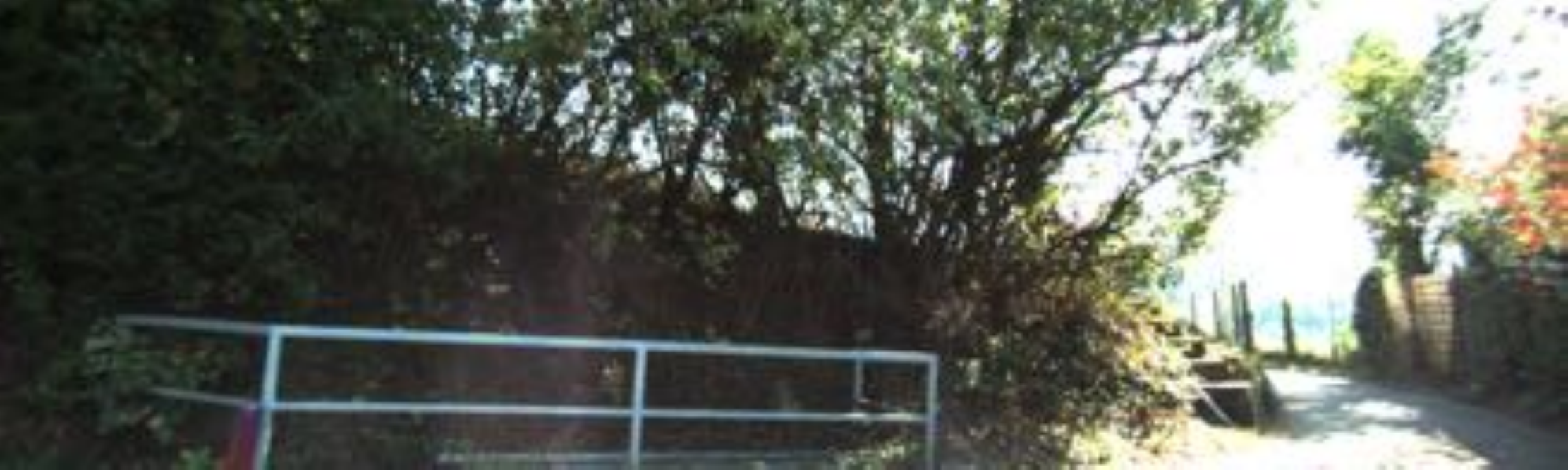}&
     \includegraphics[]{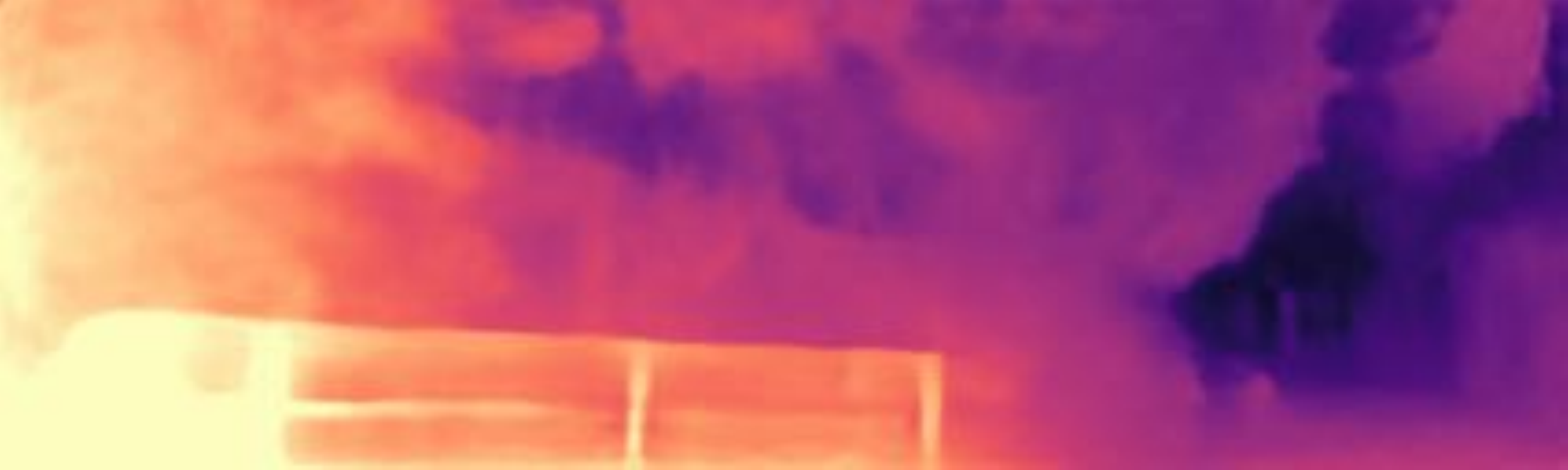}&
     \includegraphics[]{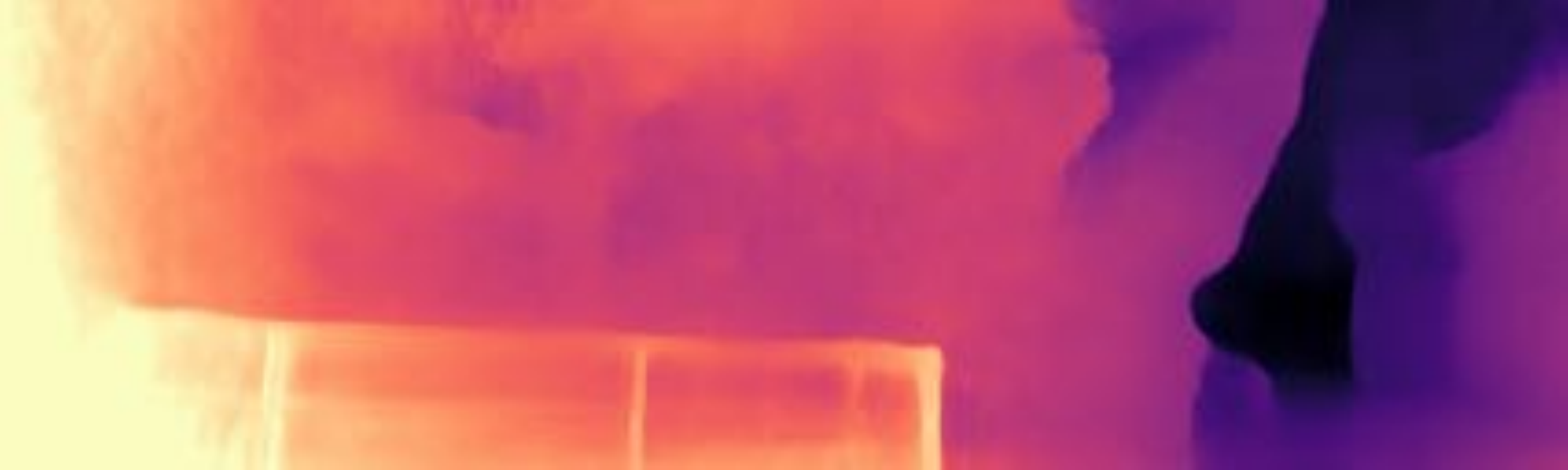}&
     \includegraphics[]{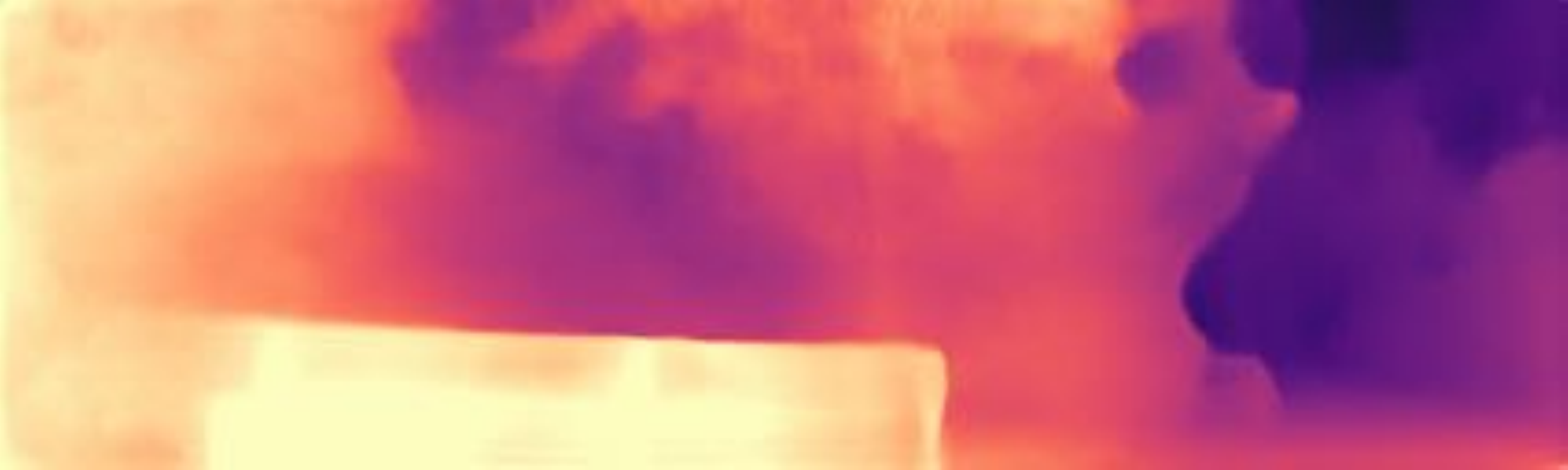}&
     \includegraphics[]{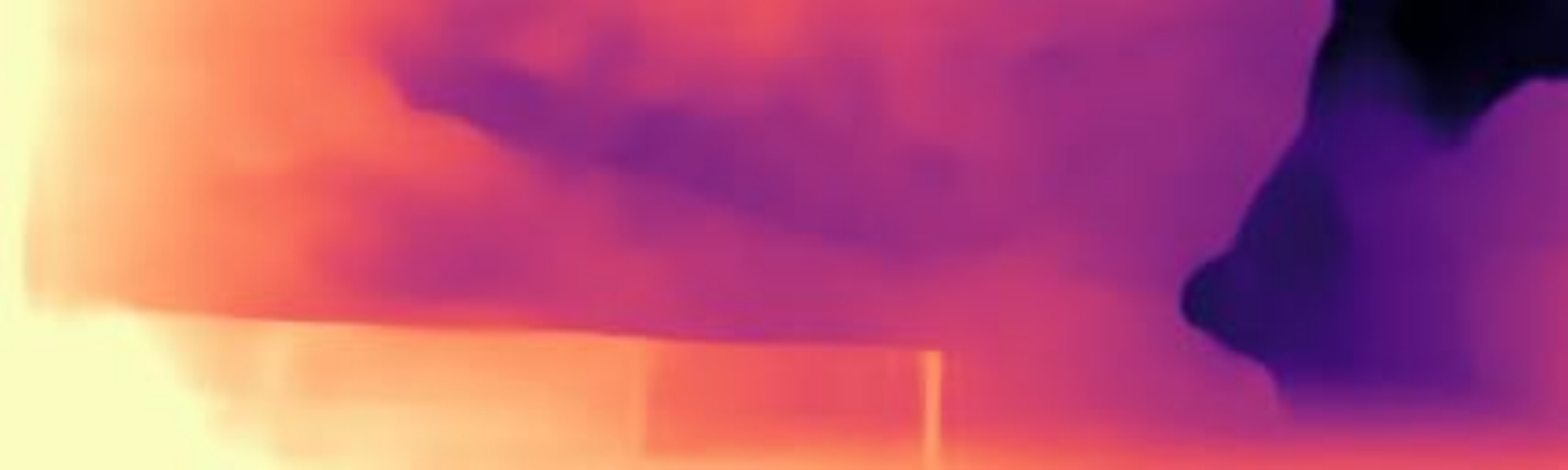}\\
     
          \includegraphics[]{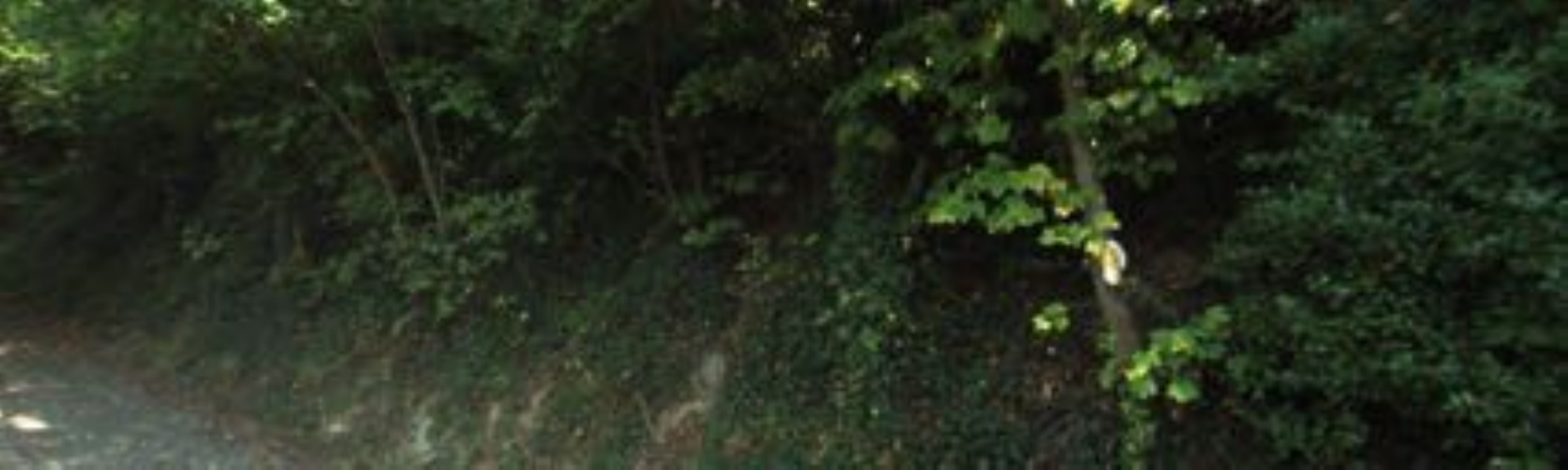}&
     \includegraphics[]{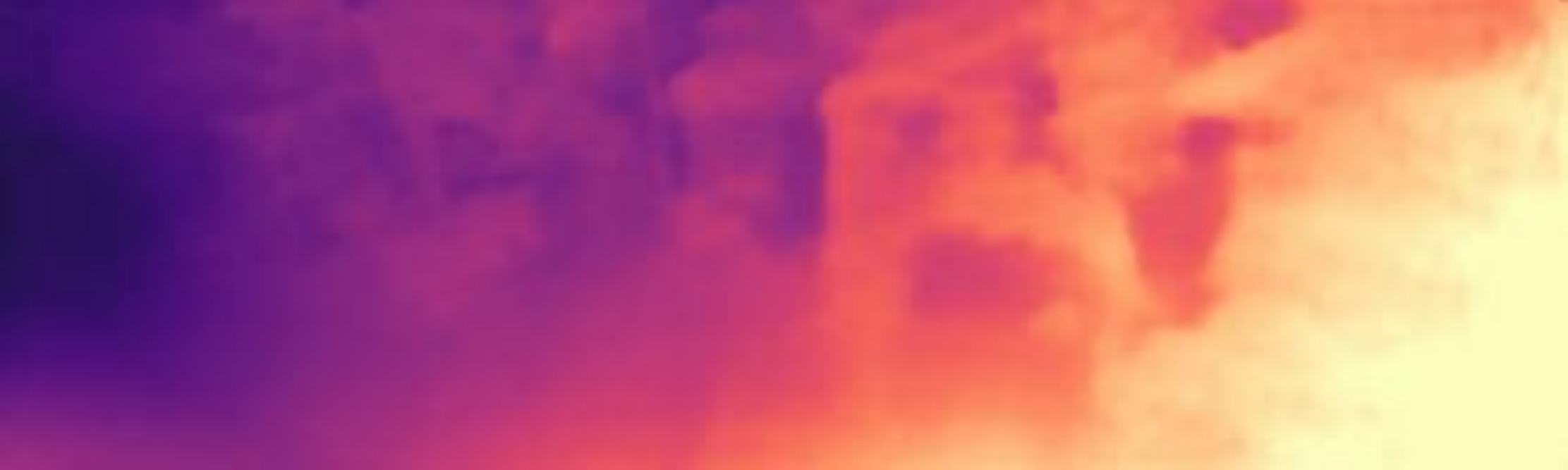}&
     \includegraphics[]{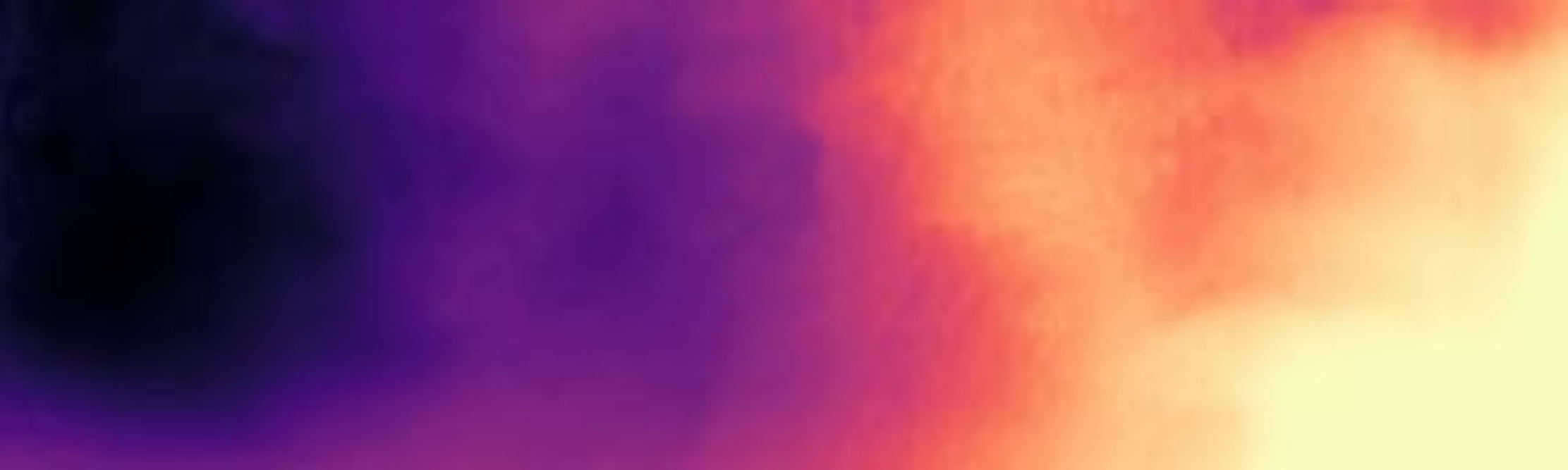}&
     \includegraphics[]{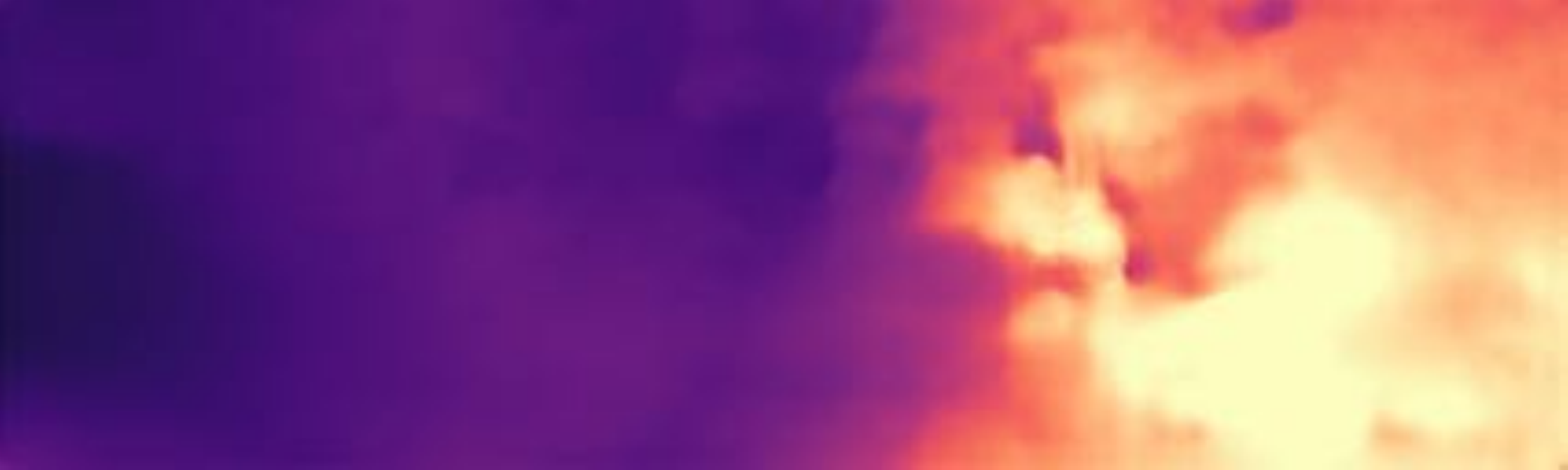}&
     \includegraphics[]{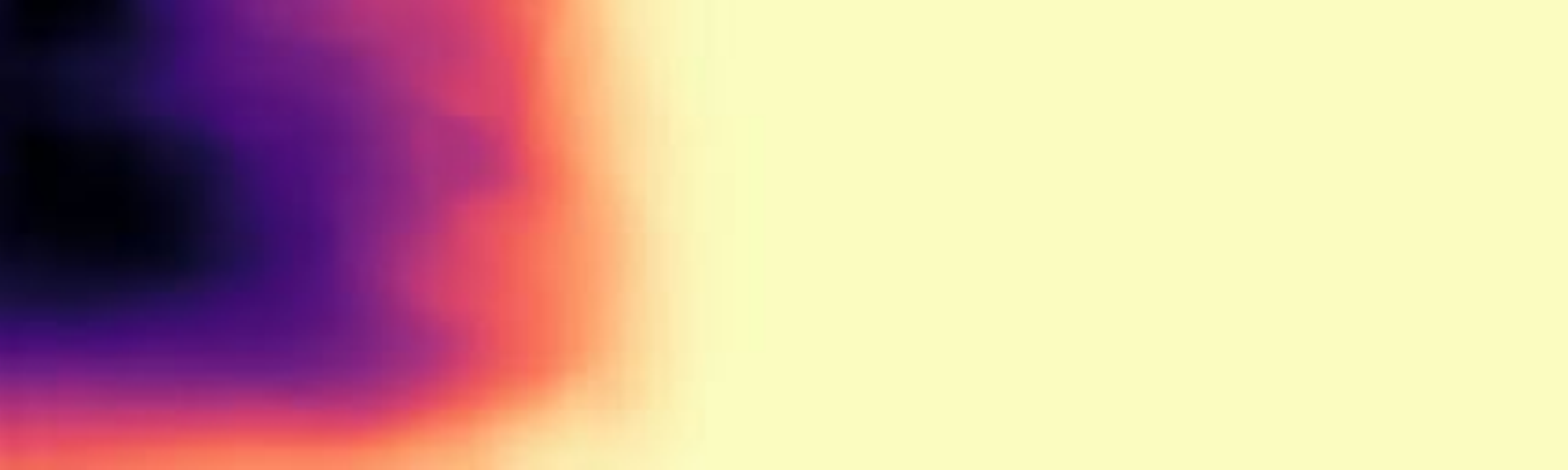}\\

          \includegraphics[]{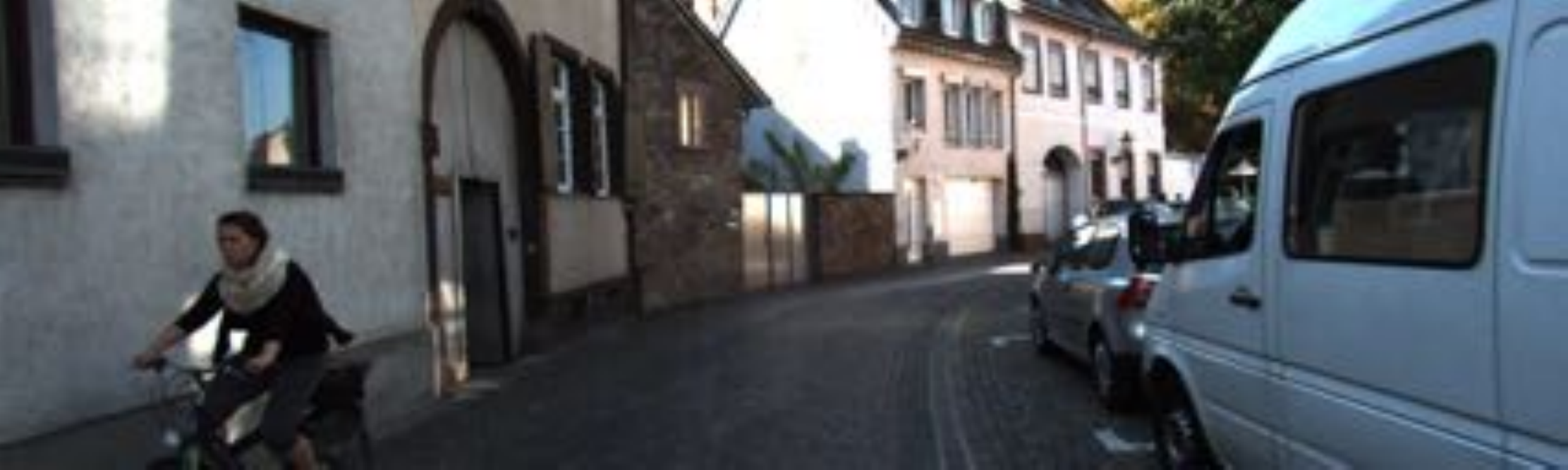}&
     \includegraphics[]{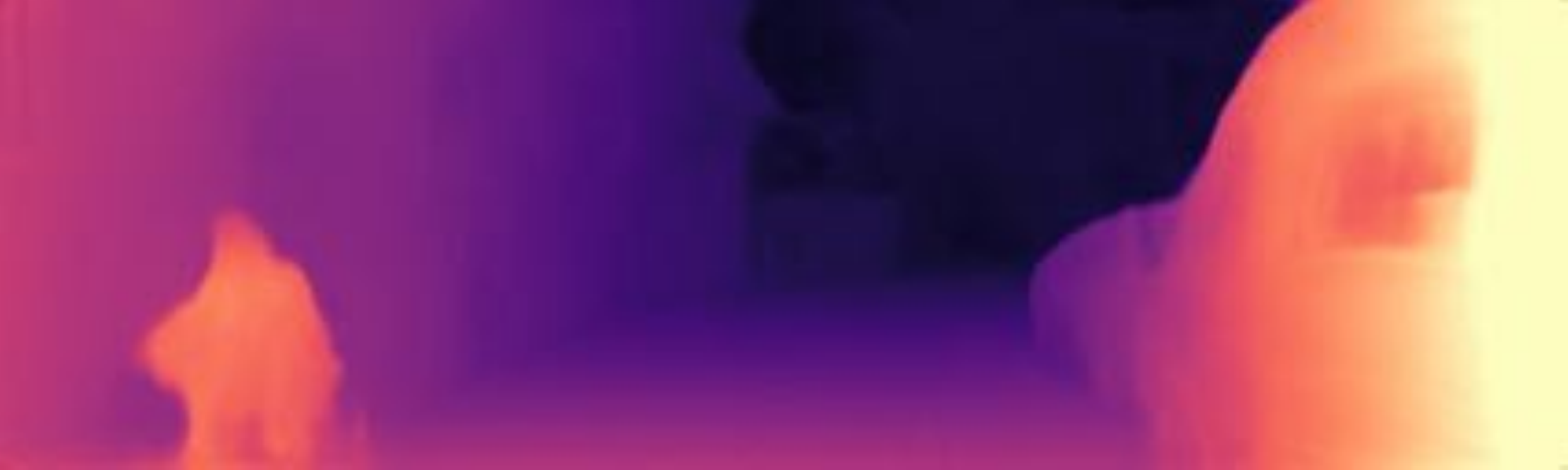}&
     \includegraphics[]{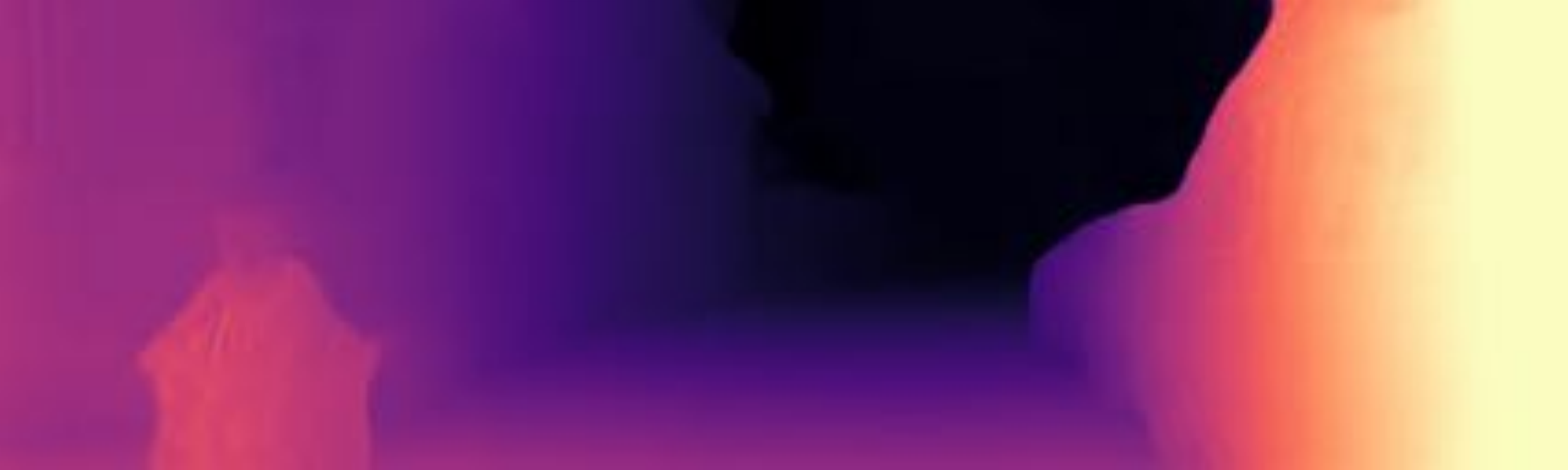}&
     \includegraphics[]{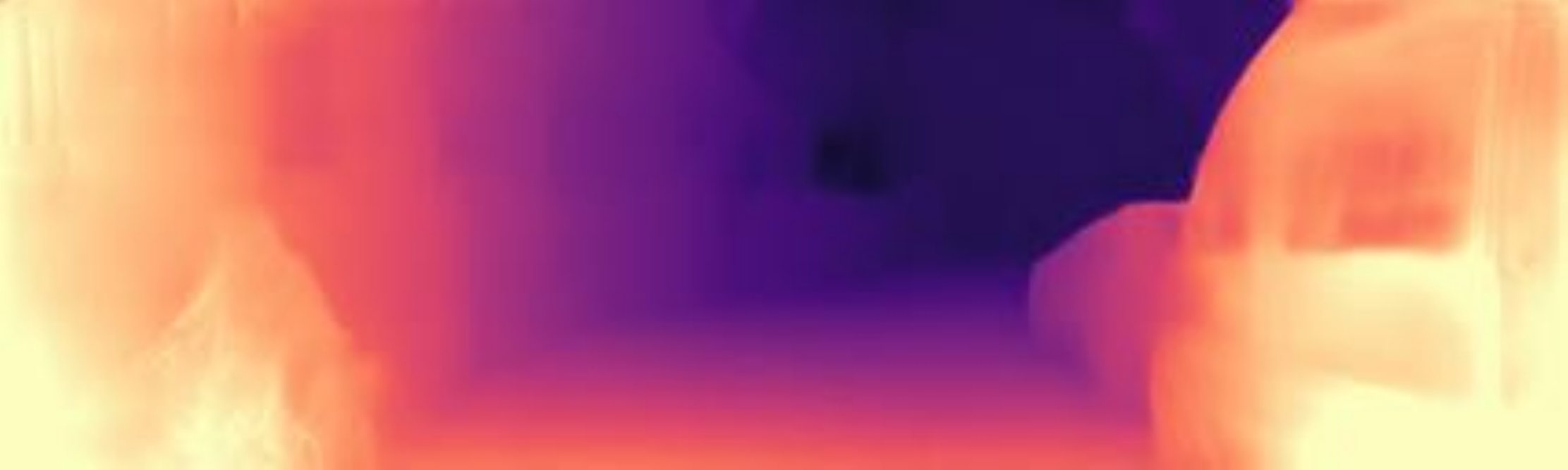}&
     \includegraphics[]{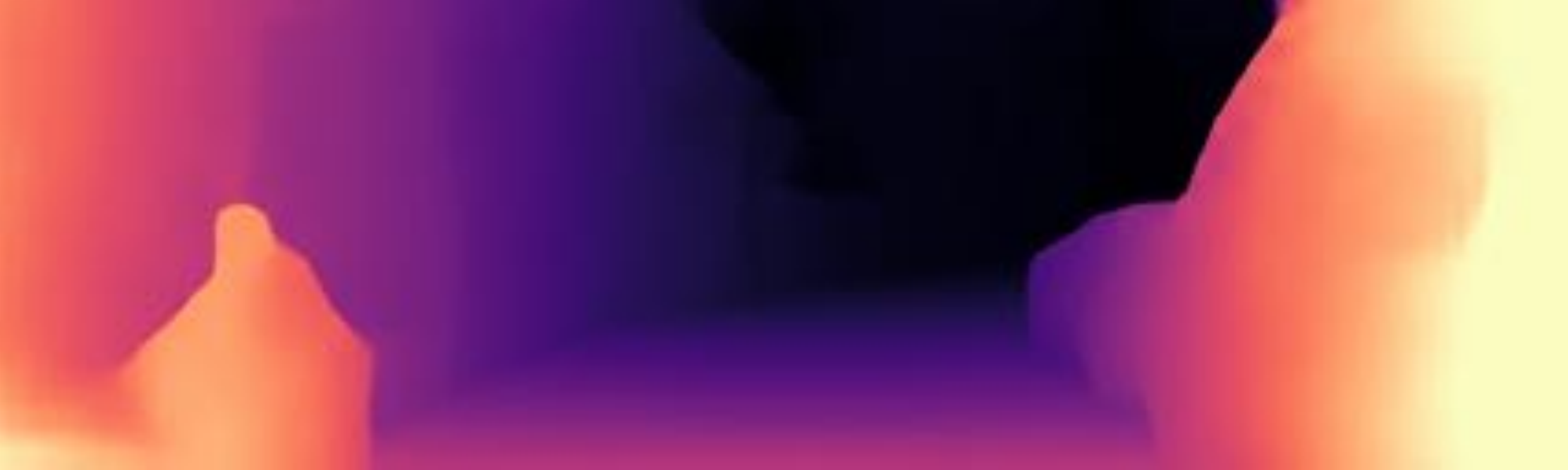}\\

          \includegraphics[]{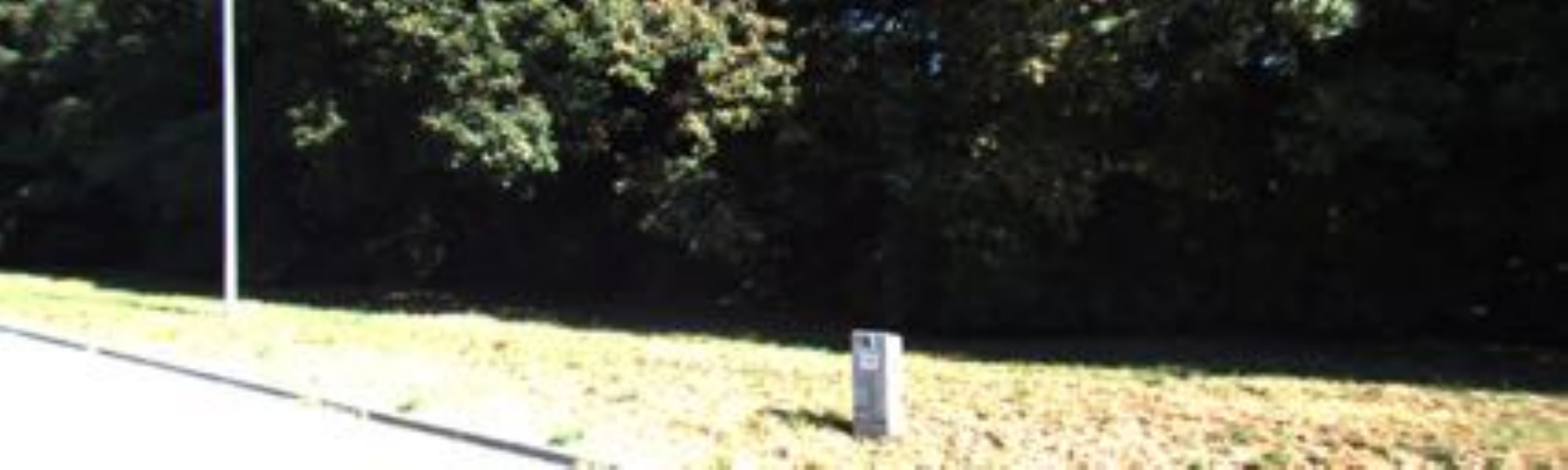}&
     \includegraphics[]{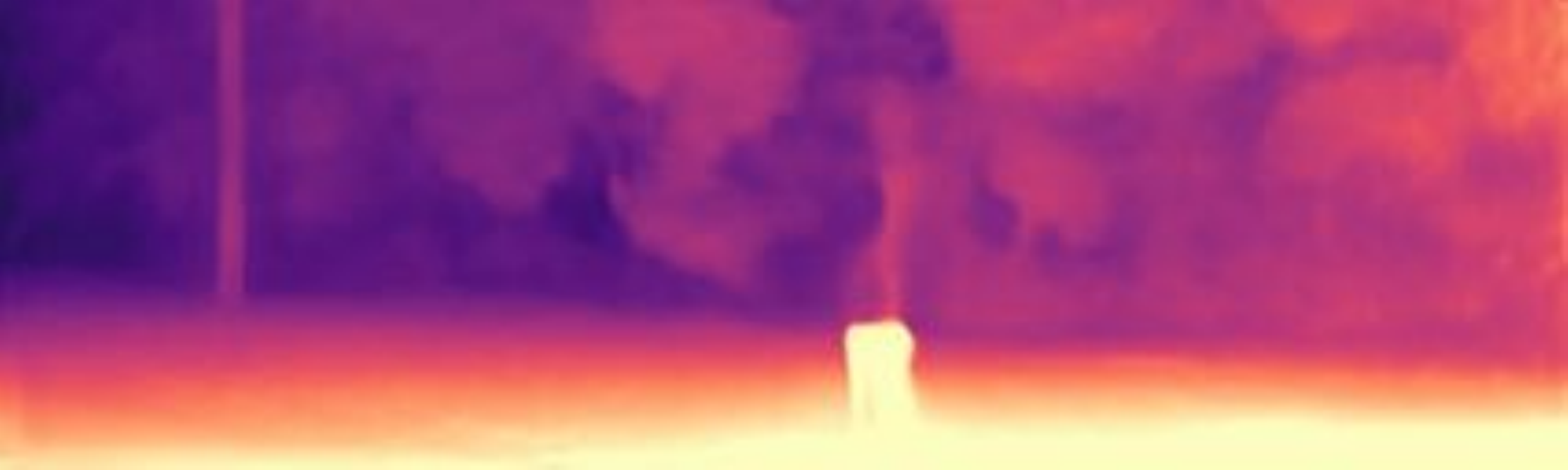}&
     \includegraphics[]{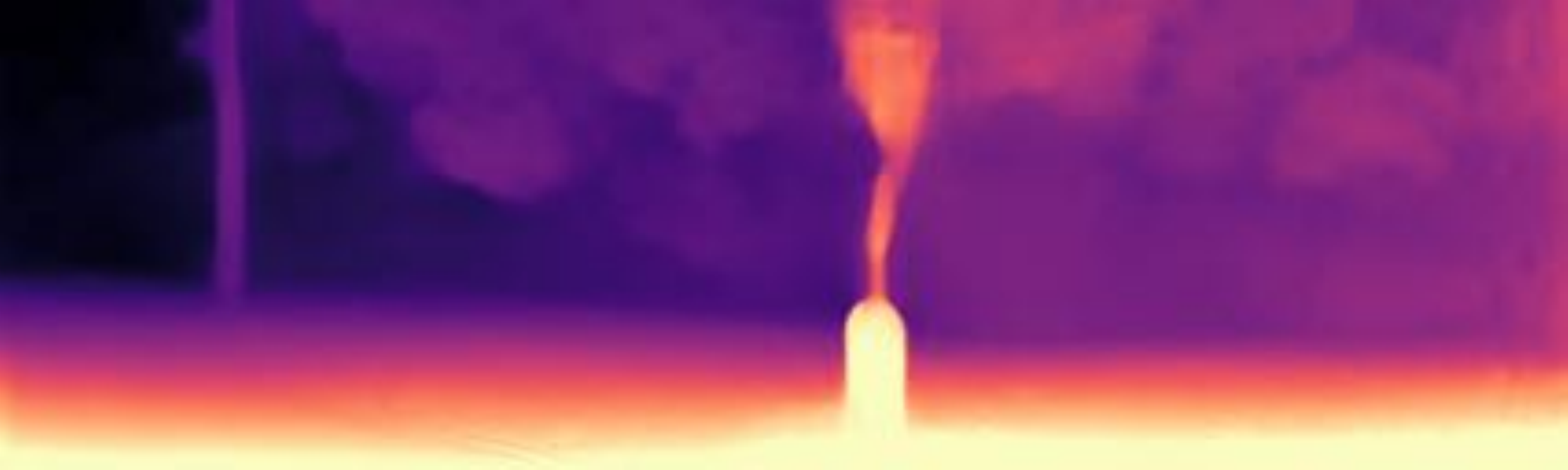}&
     \includegraphics[]{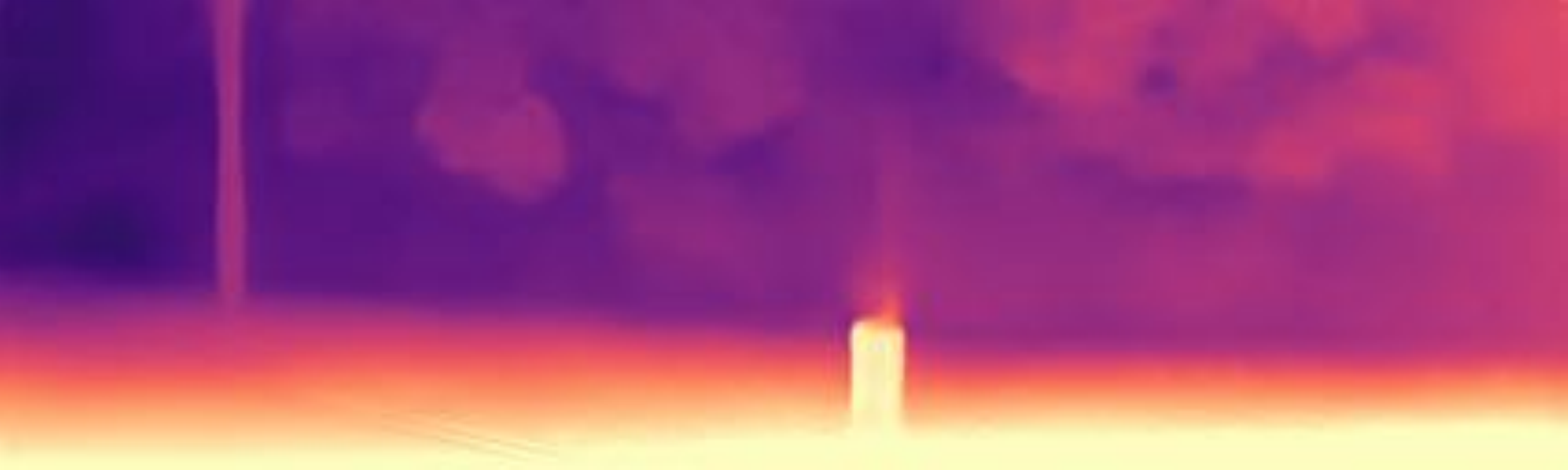}&
     \includegraphics[]{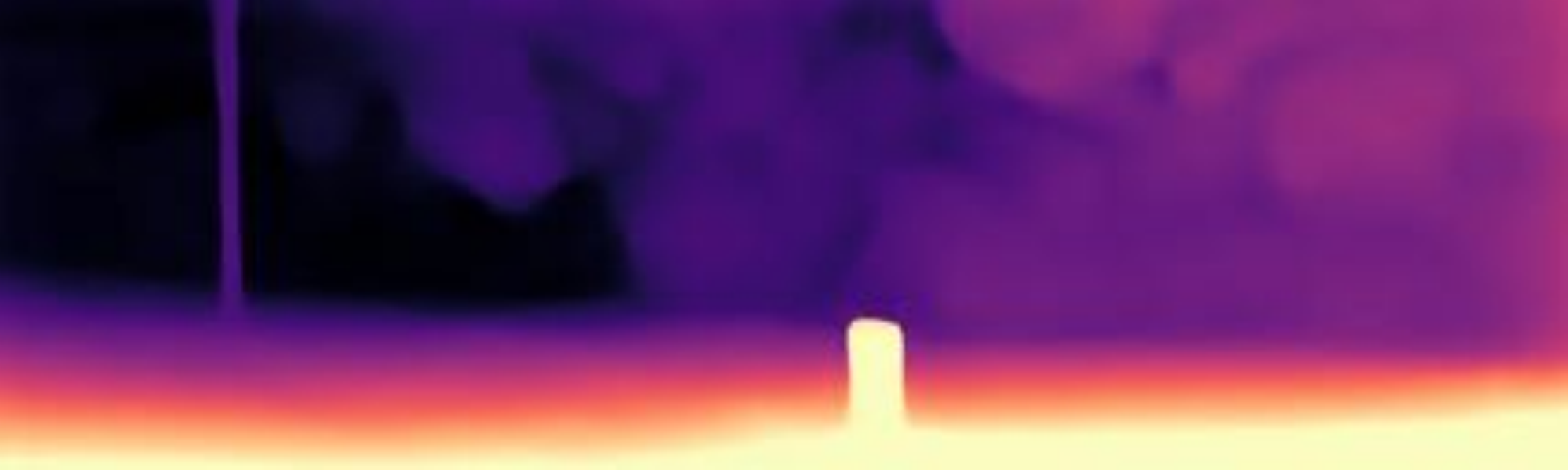}\\
     
     \fontsize{65}{15} \selectfont Input images & 
    \fontsize{65}{15} \selectfont Ours-Hybrid & 
    \fontsize{65}{15} \selectfont Monodepth2 &
    \fontsize{65}{15} \selectfont PackNet-SfM &
    \fontsize{65}{15} \selectfont R-MSFM6
    \end{tabular}}
    \vspace{-0.3cm}
    \caption{\textbf{Qualitative comparison to state-of-the-arts.} We use KITTI for training and testing.}
\label{figure_result_kitti_apdx}
    \end{subfigure}
     
     \begin{subfigure}
    \centering
    \resizebox{\linewidth}{!}{
    \begin{tabular}{ccccccc}
\includegraphics[width=\w,height=\h]{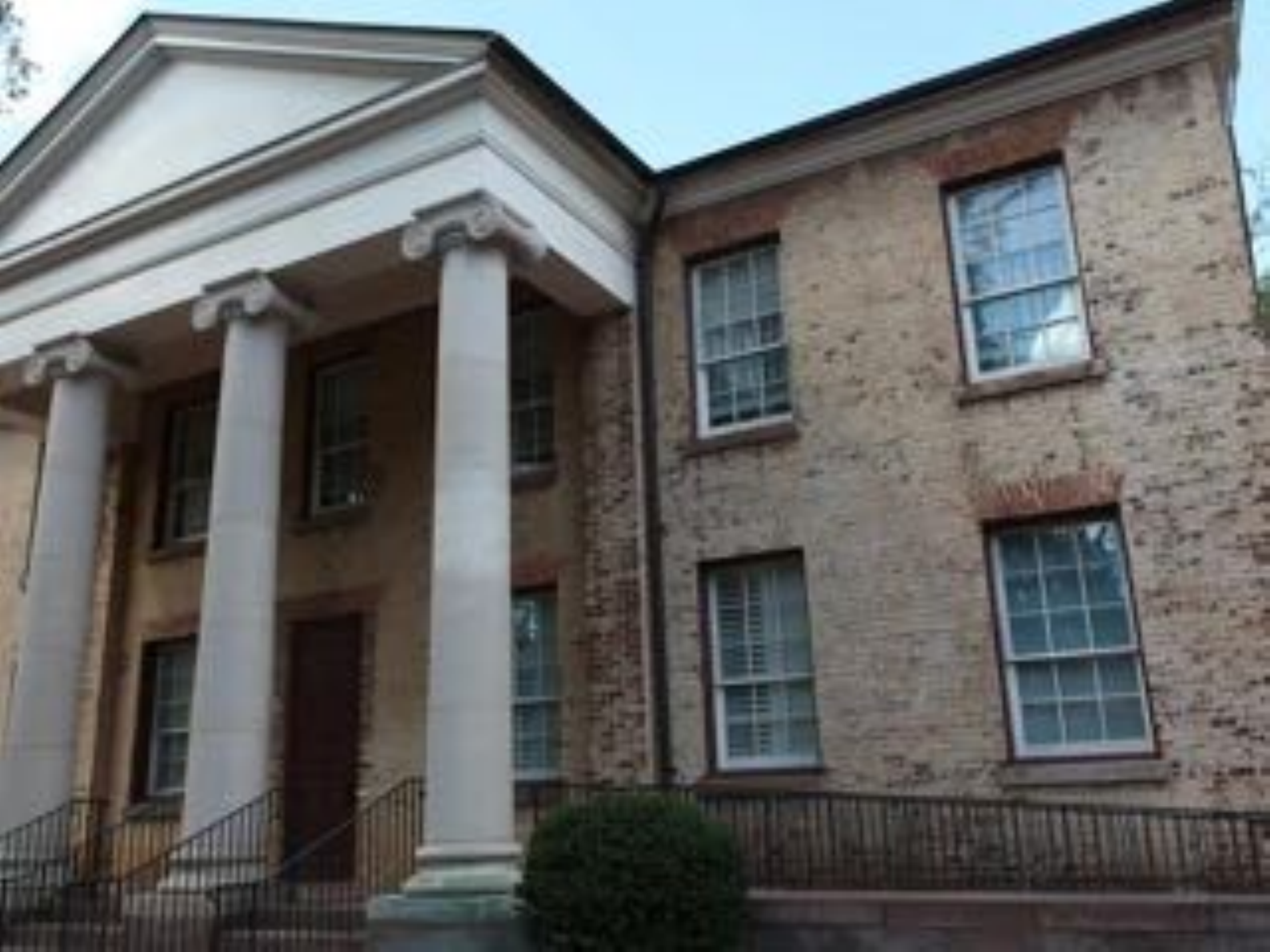}& 
\includegraphics[width=\w,height=\h]{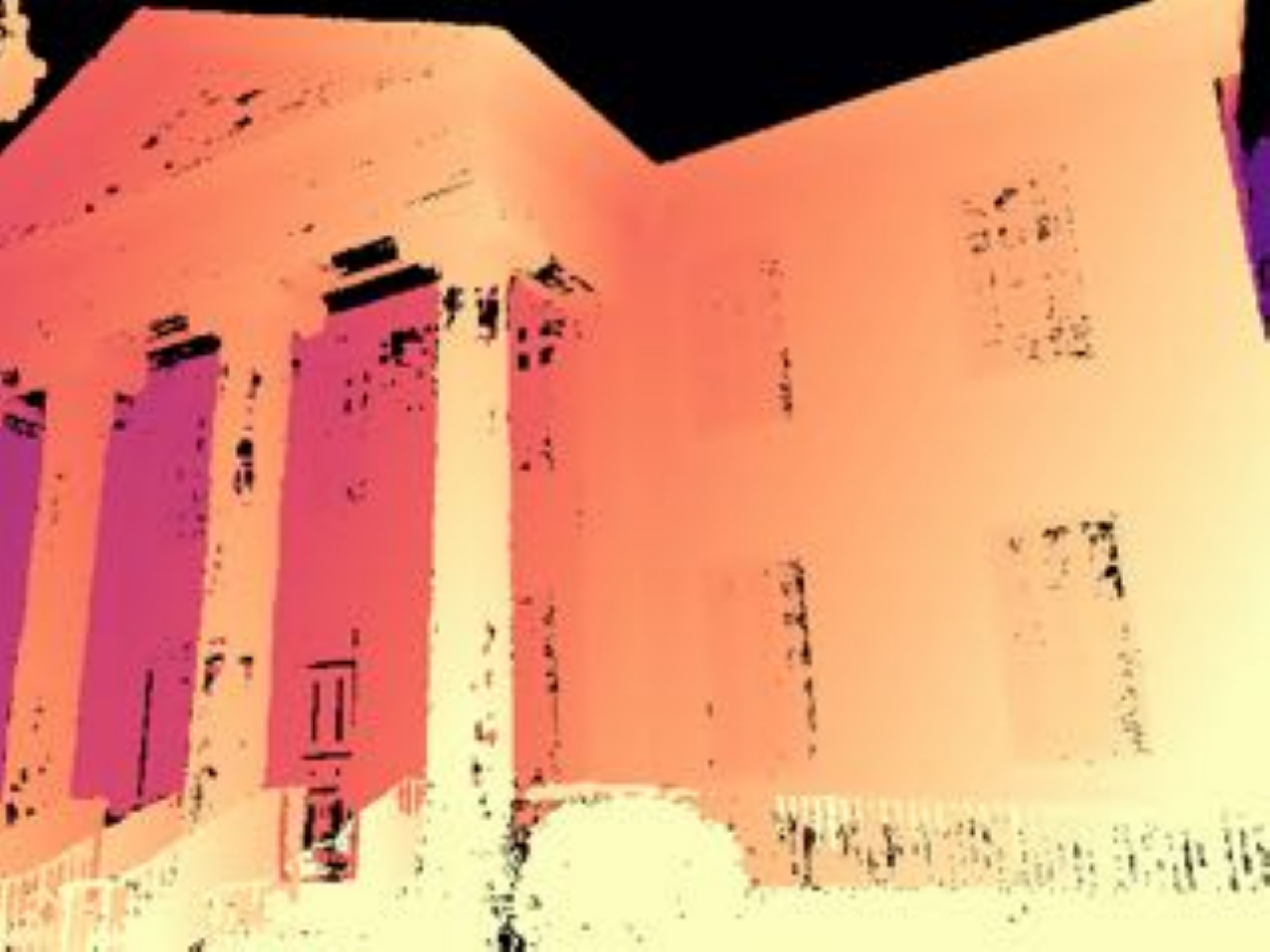}& 
\includegraphics[width=\w,height=\h]{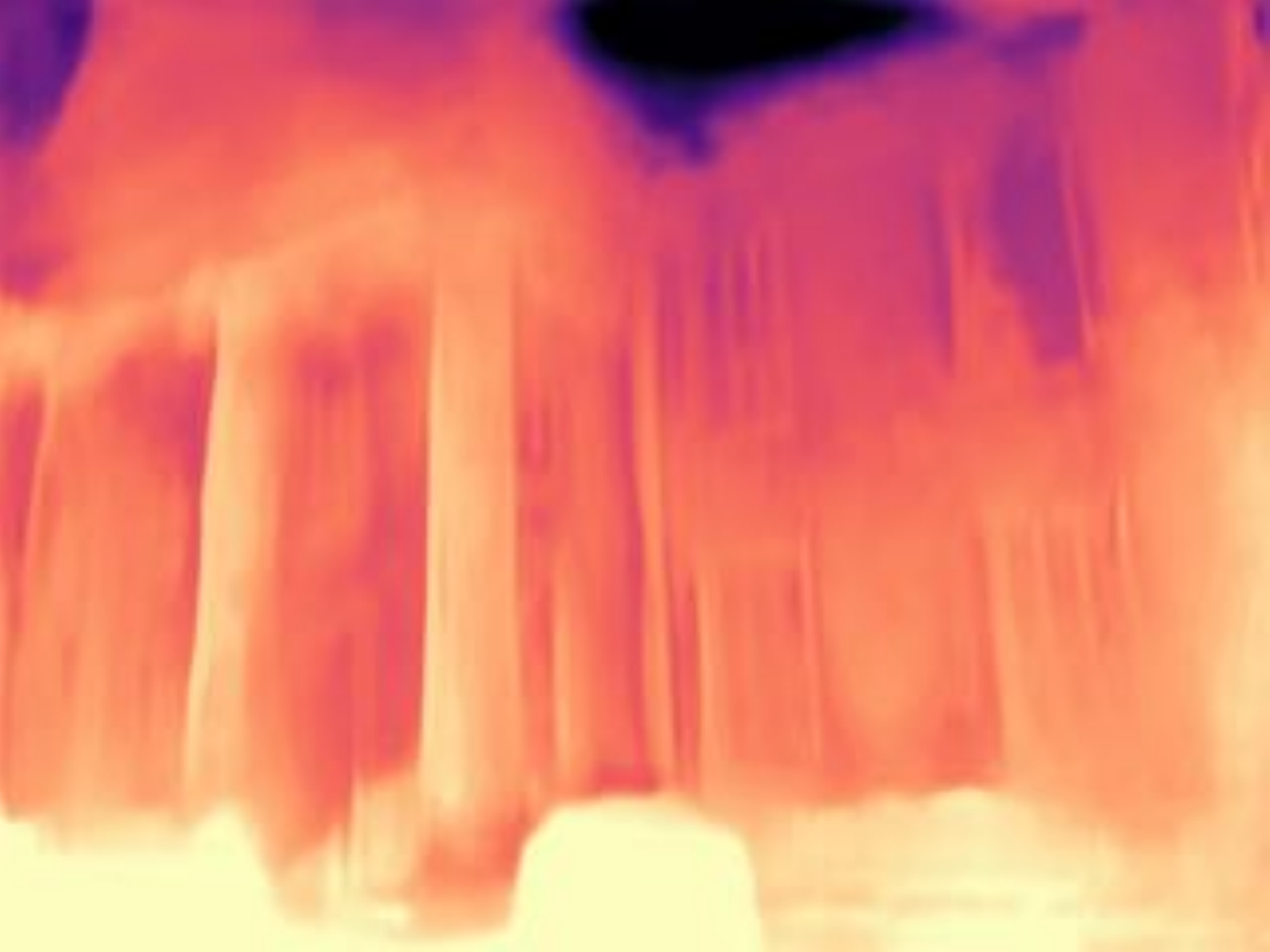}& 
\includegraphics[width=\w,height=\h]{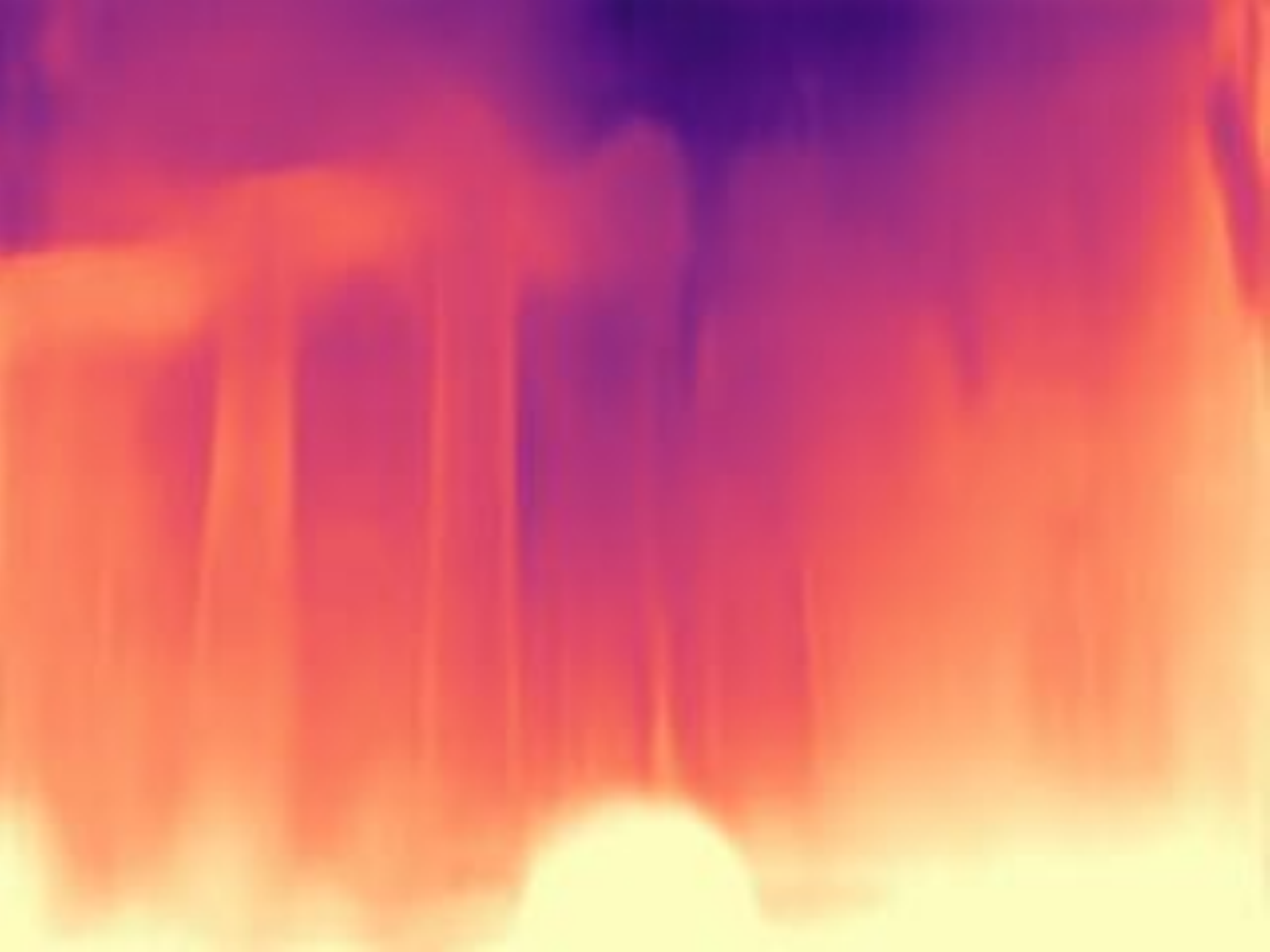}& 
\includegraphics[width=\w,height=\h]{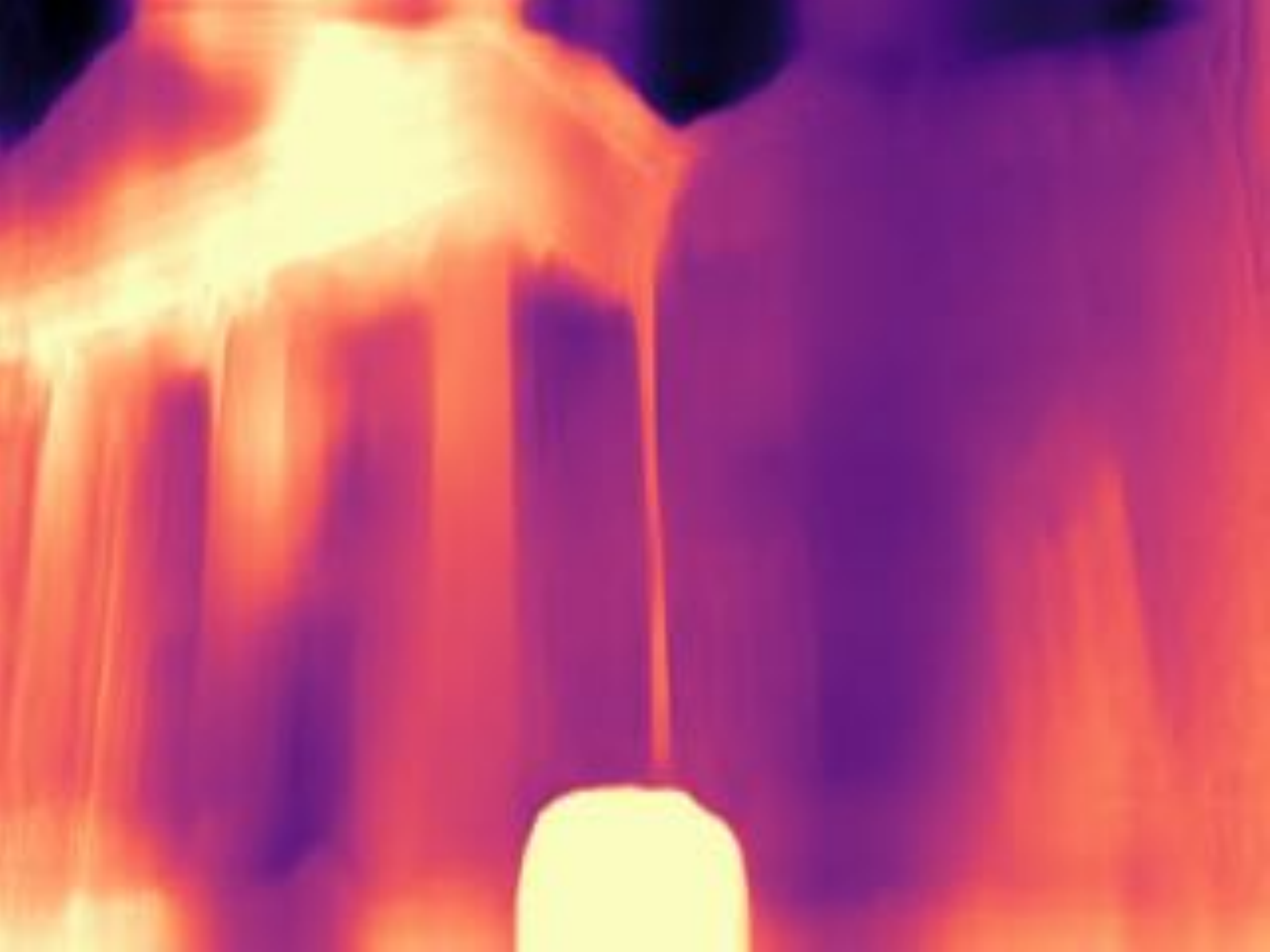}& 
\includegraphics[width=\w,height=\h]{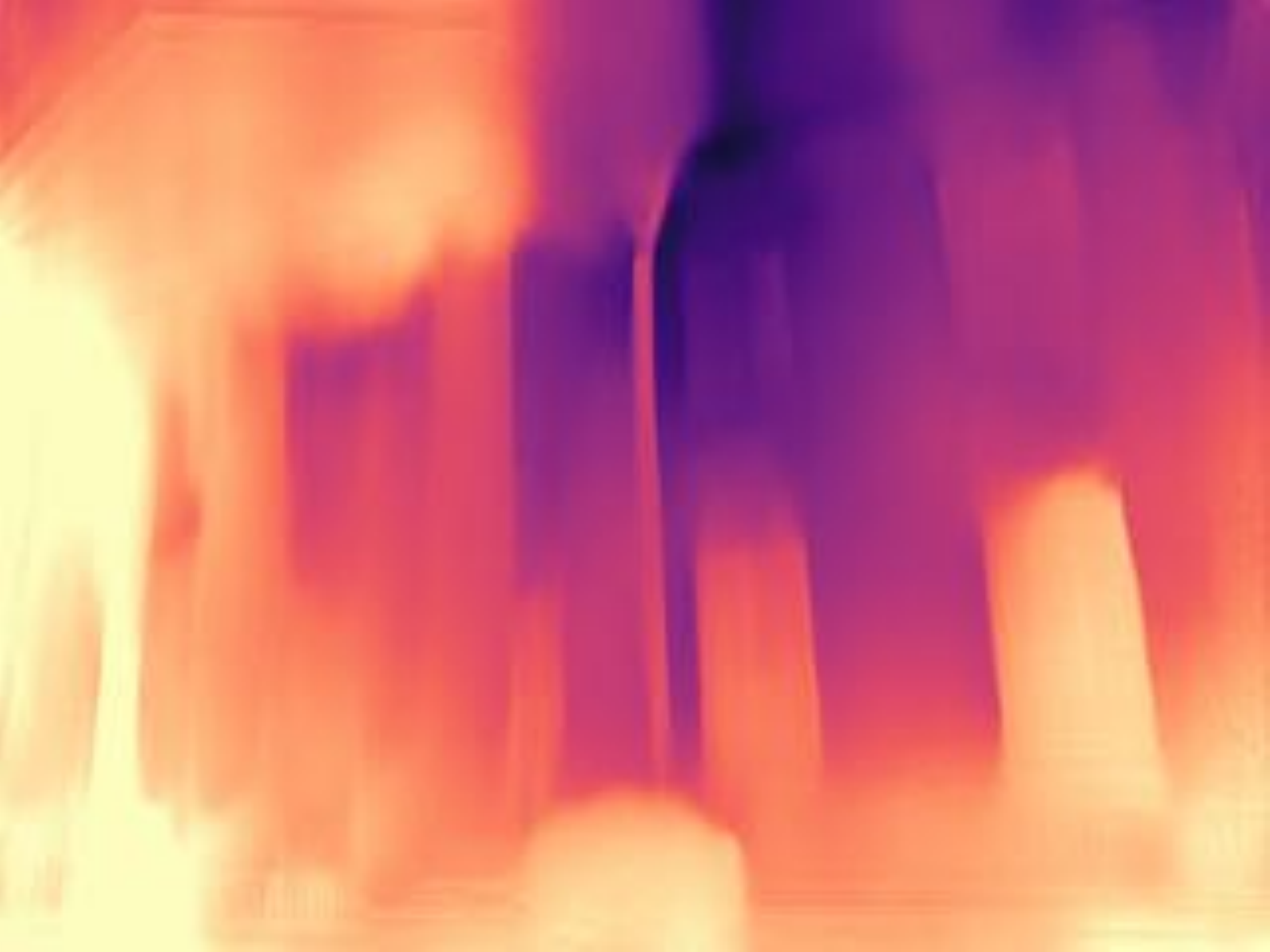}& 
\includegraphics[width=\w,height=\h]{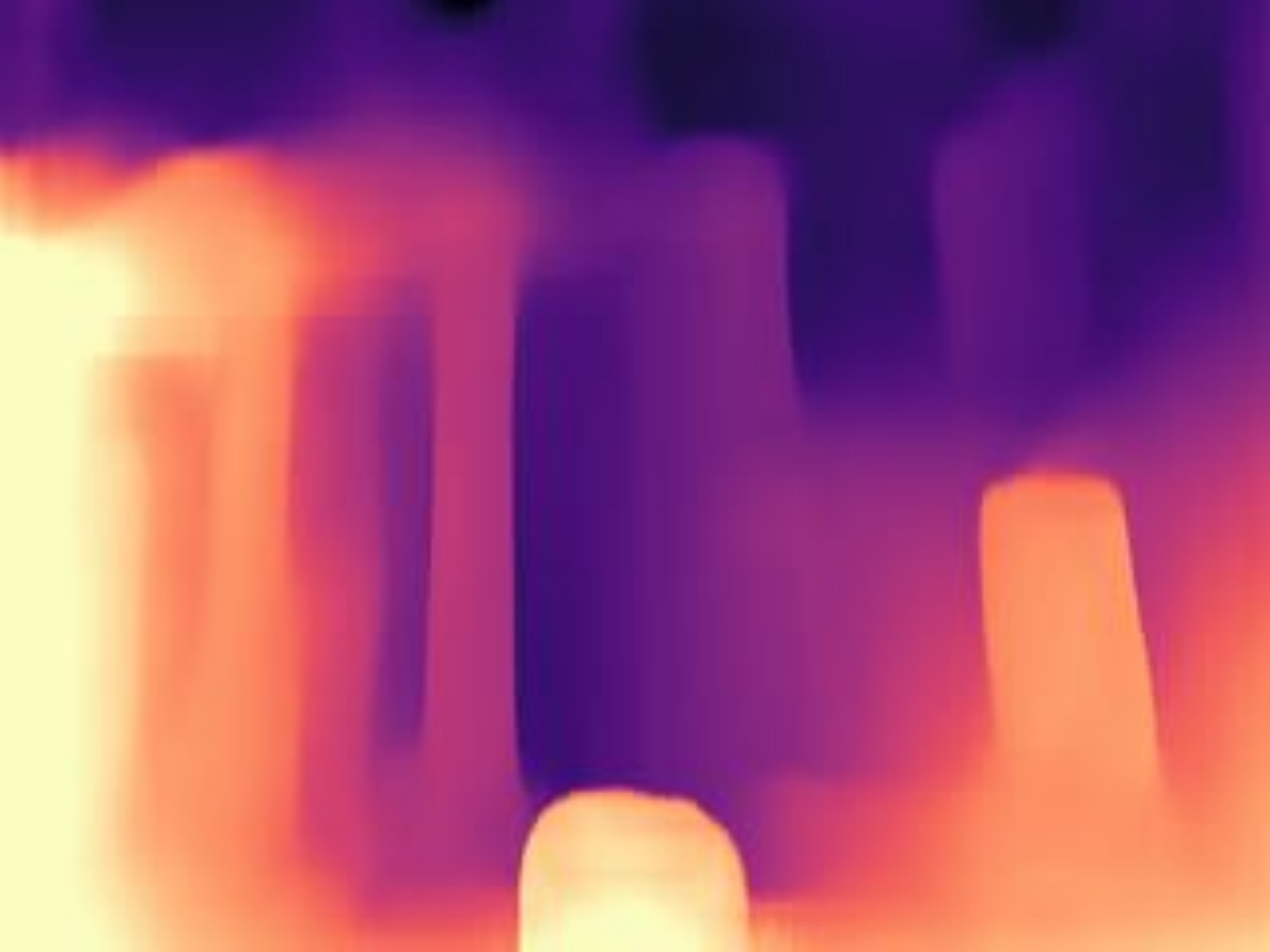}\\ 

\includegraphics[width=\w,height=\h]{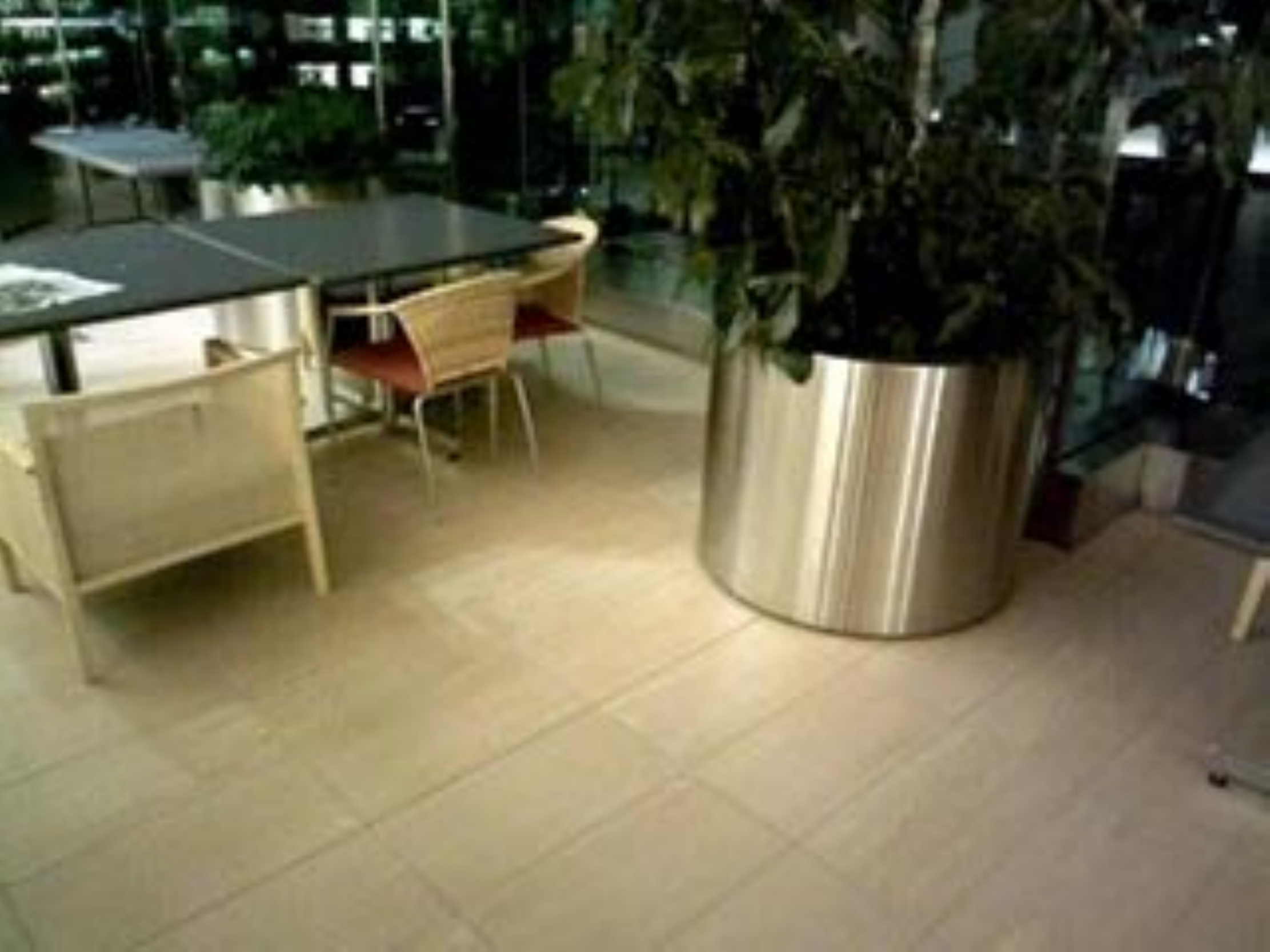}& 
\includegraphics[width=\w,height=\h]{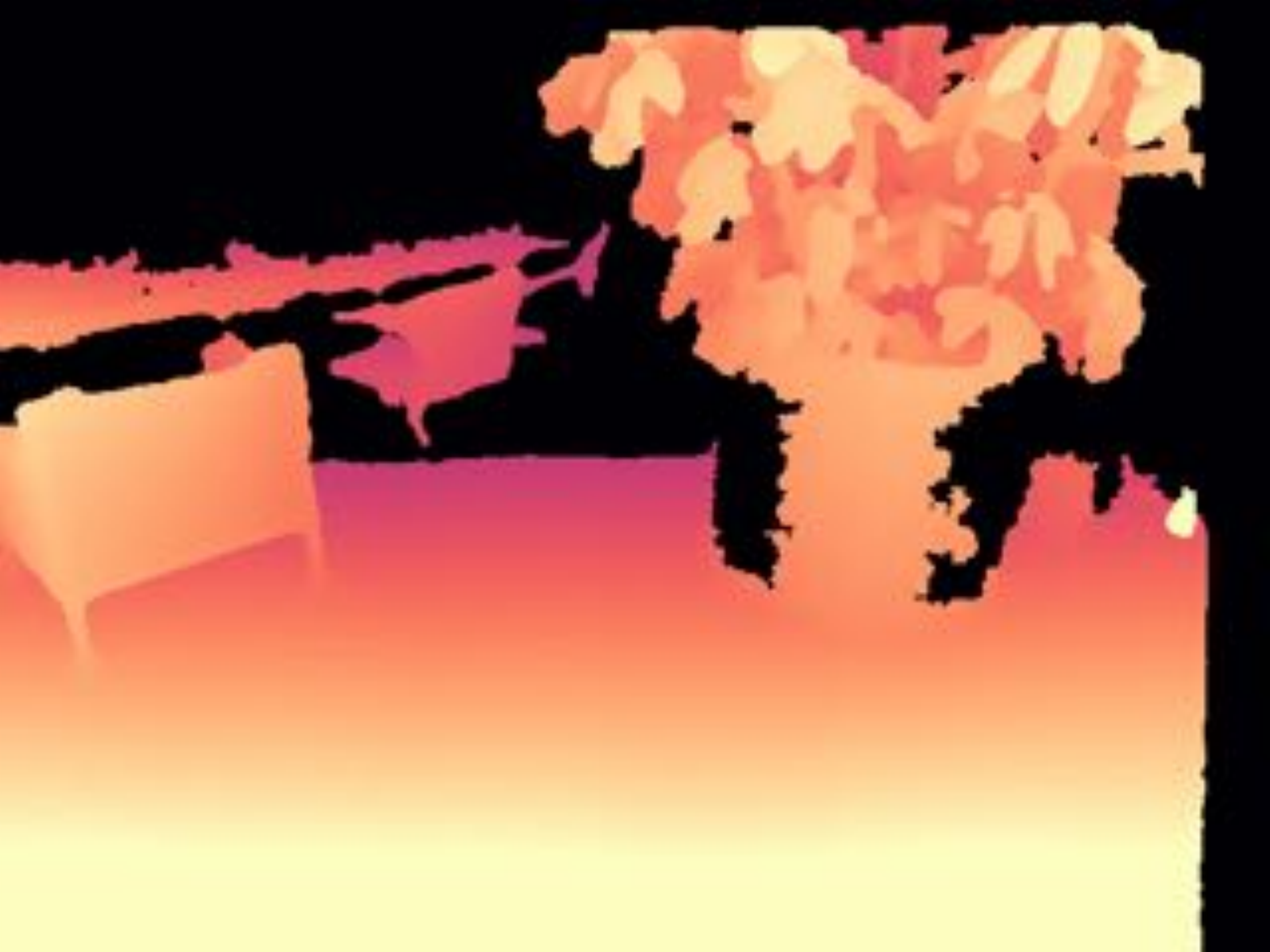}& 
\includegraphics[width=\w,height=\h]{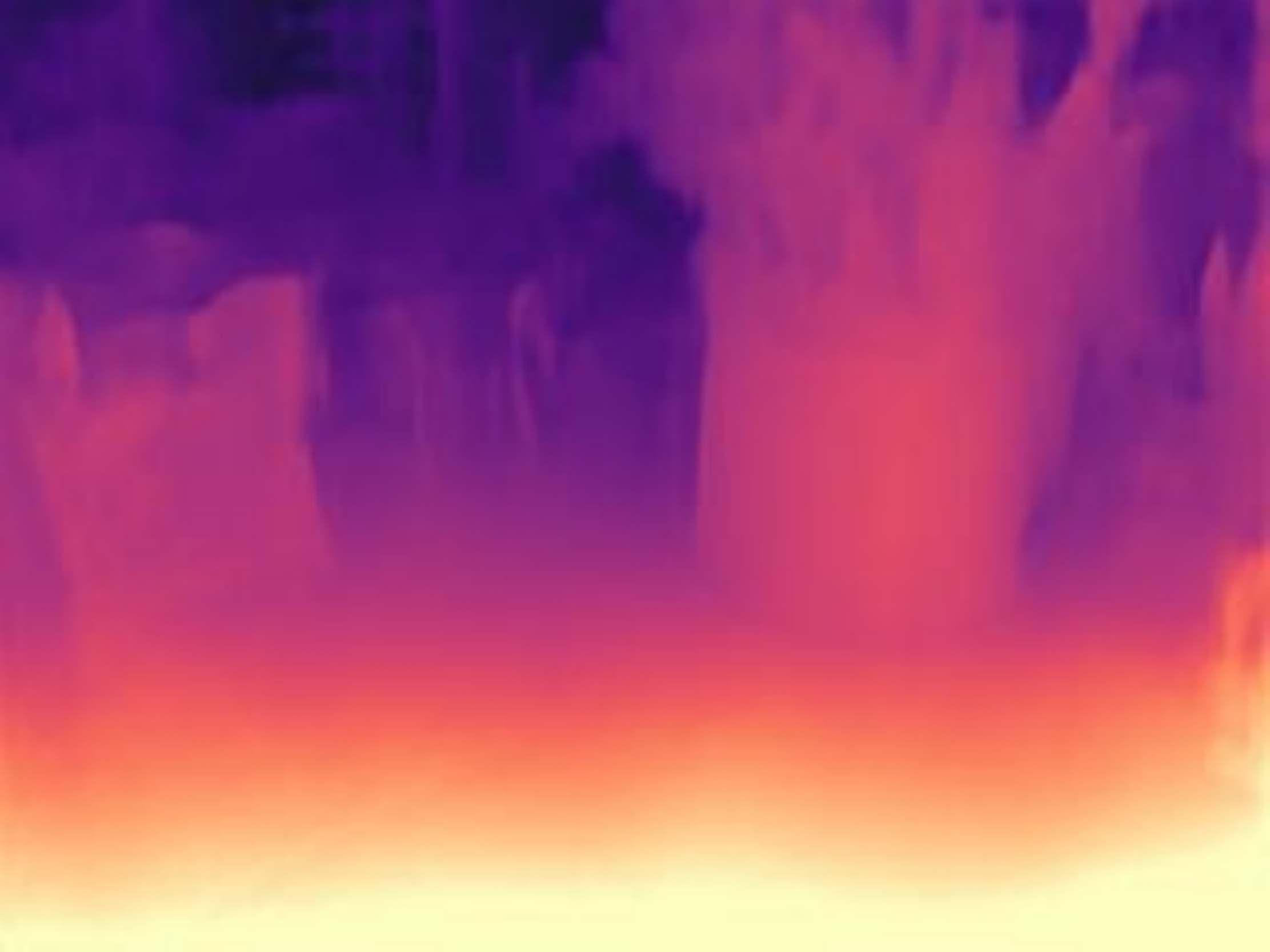}& 
\includegraphics[width=\w,height=\h]{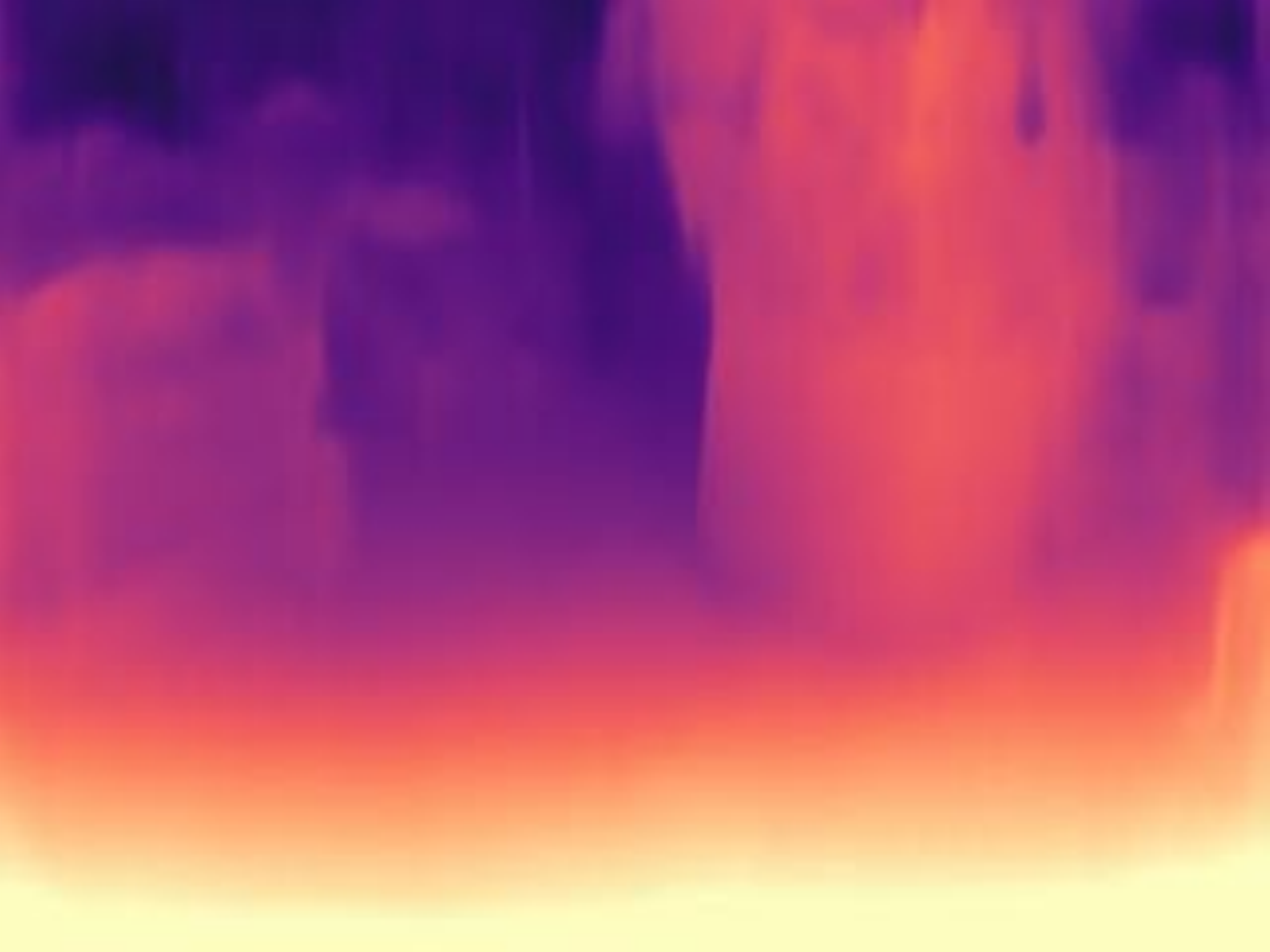}& 
\includegraphics[width=\w,height=\h]{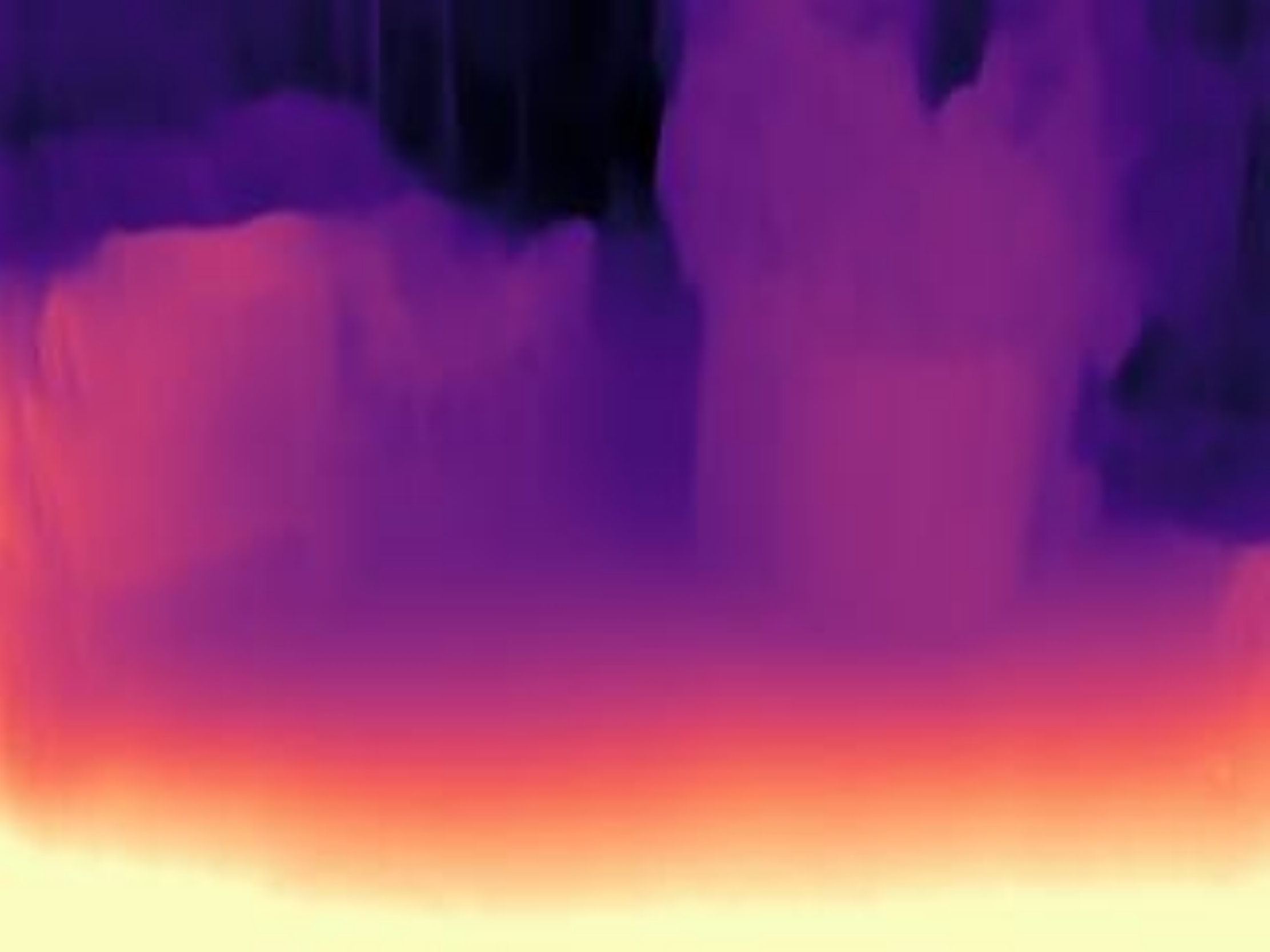}& 
\includegraphics[width=\w,height=\h]{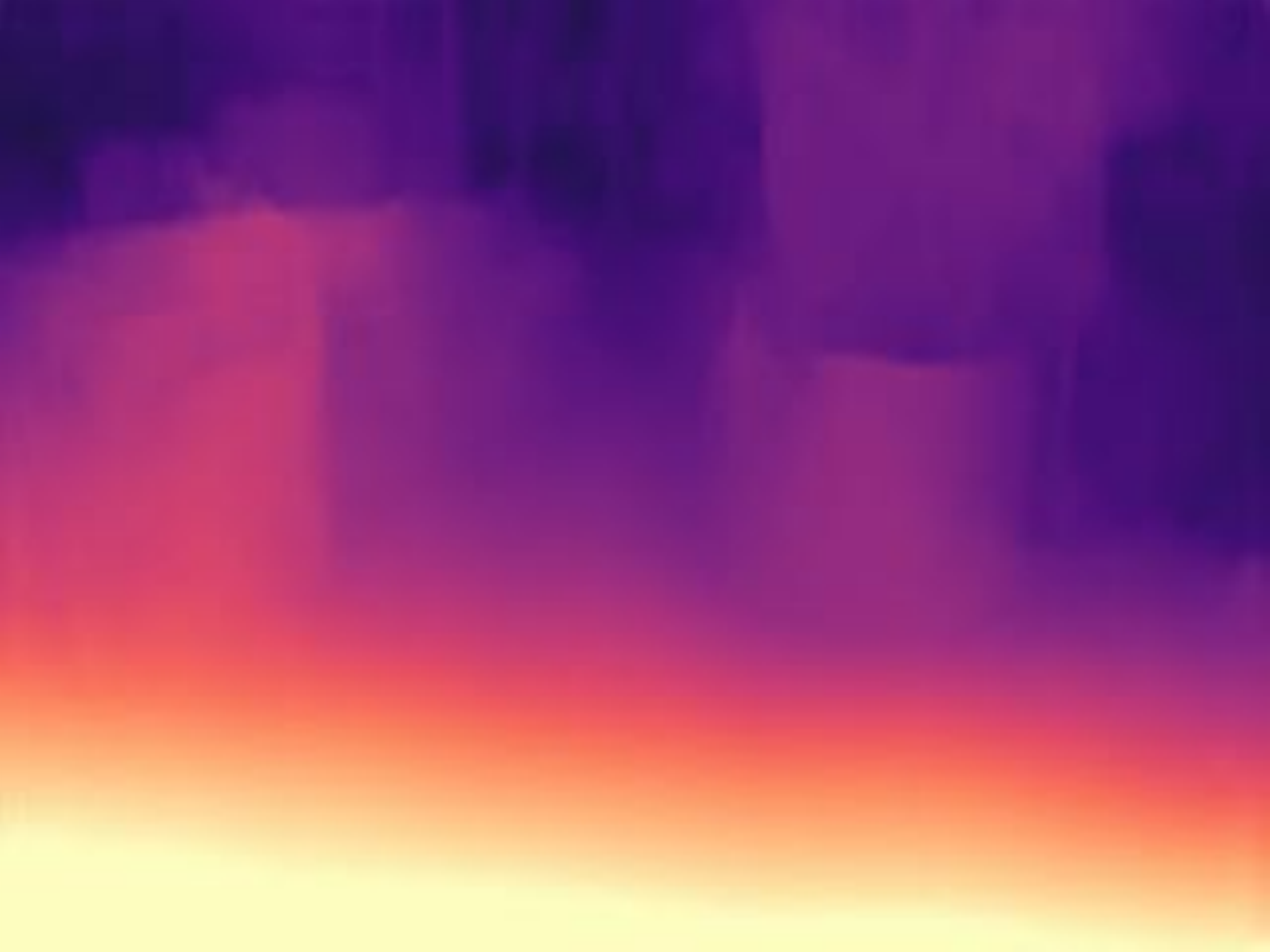}& 
\includegraphics[width=\w,height=\h]{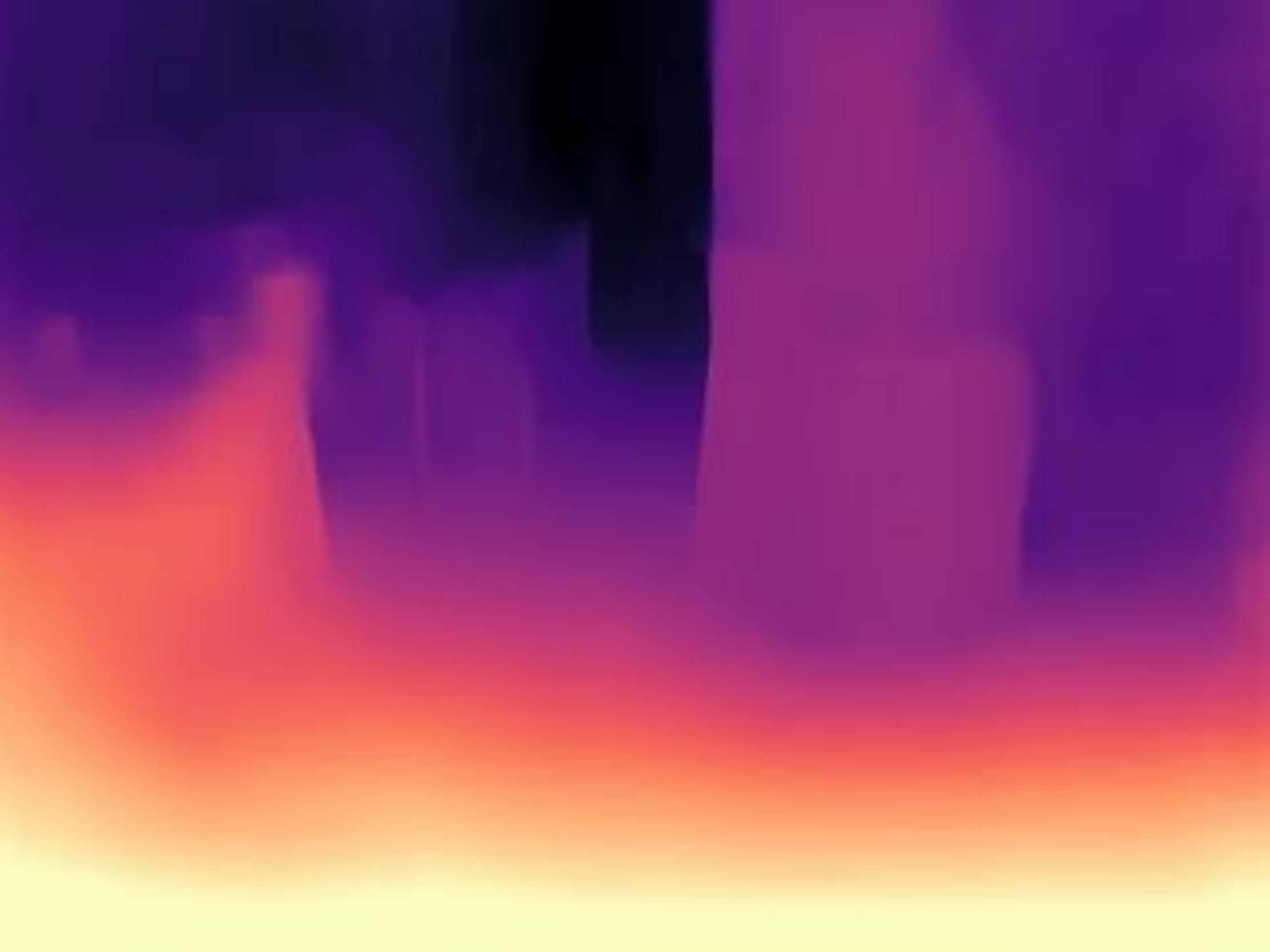}\\ 

\includegraphics[width=\w,height=\h]{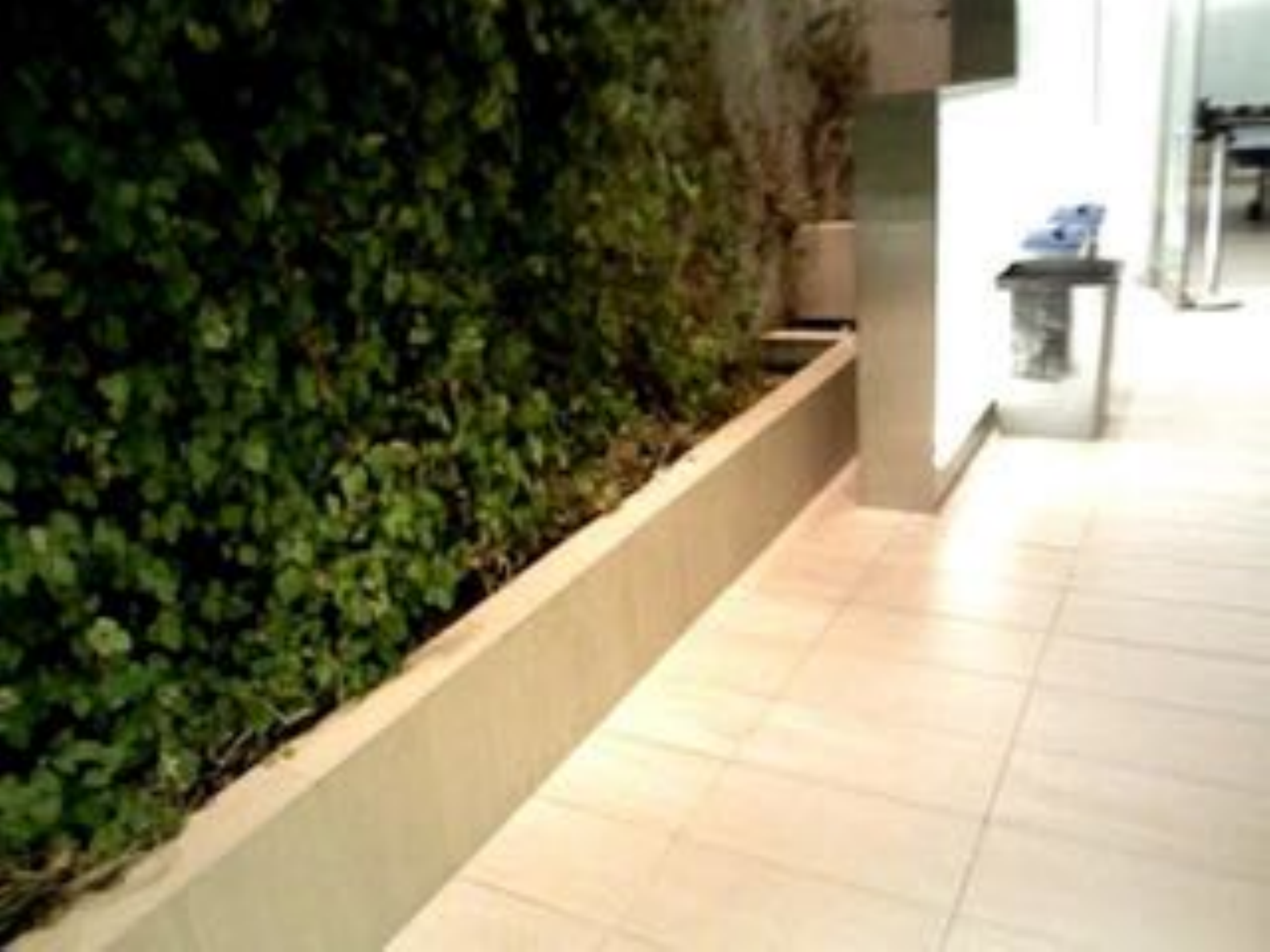}& 
\includegraphics[width=\w,height=\h]{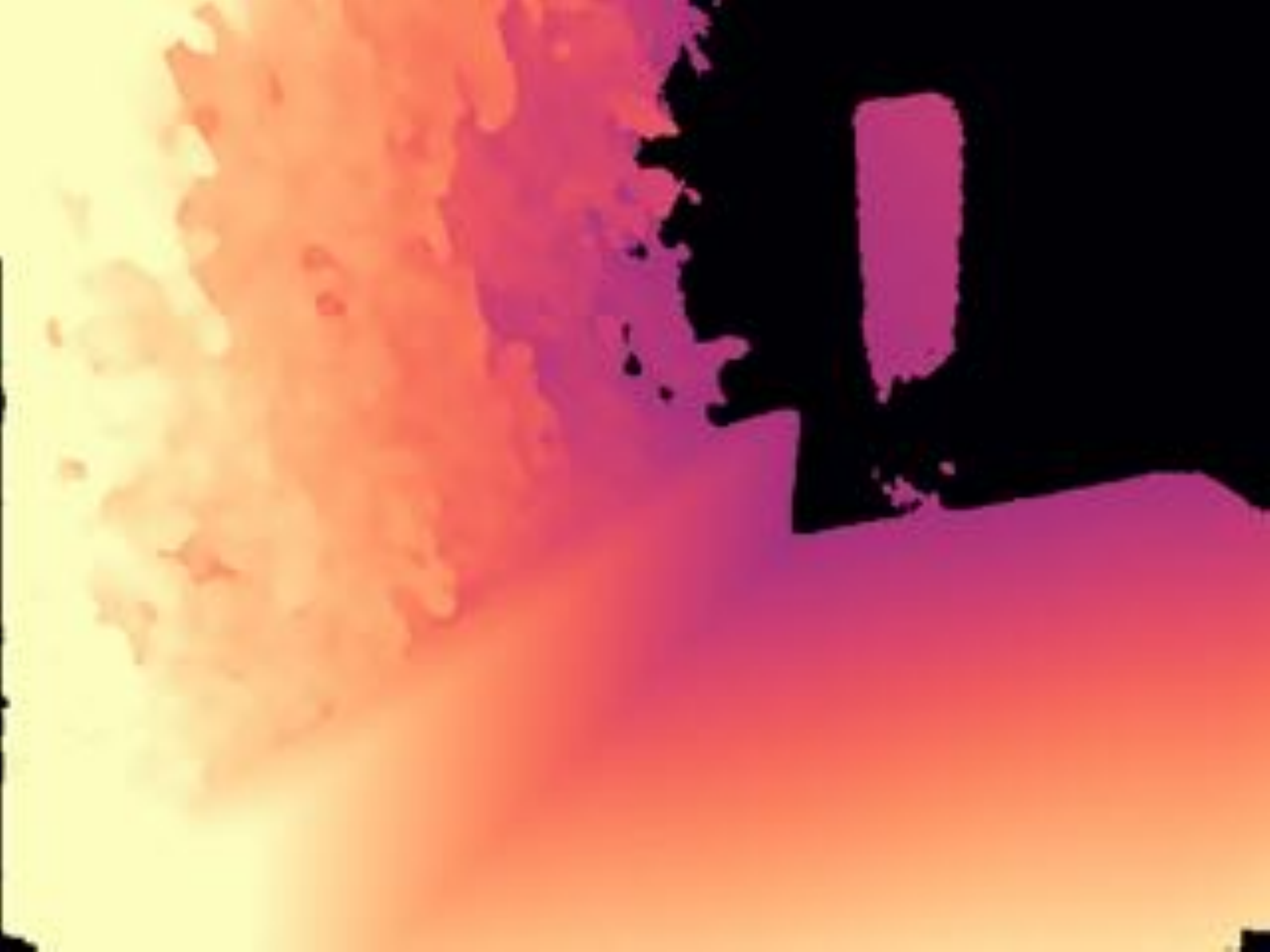}& 
\includegraphics[width=\w,height=\h]{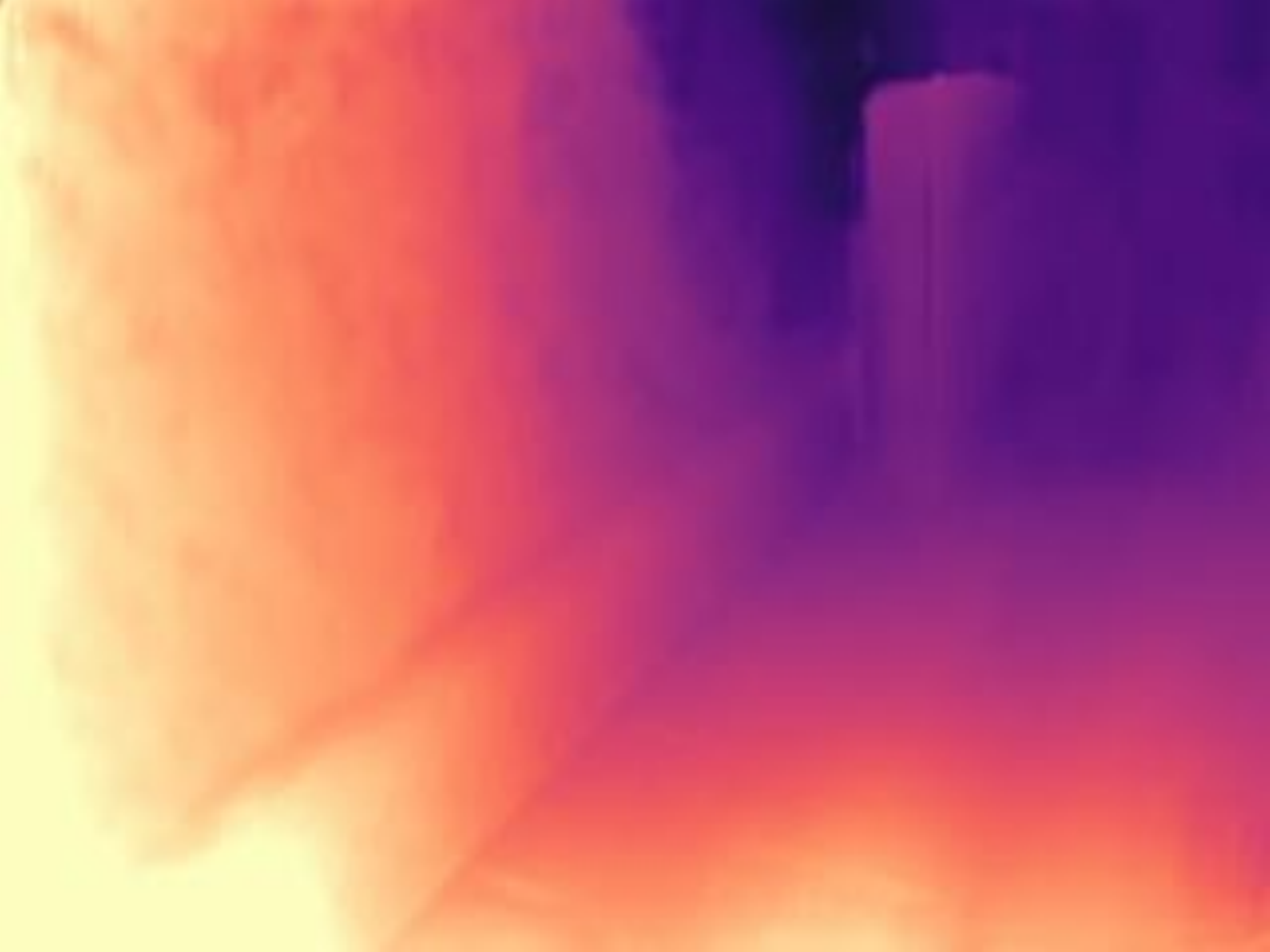}& 
\includegraphics[width=\w,height=\h]{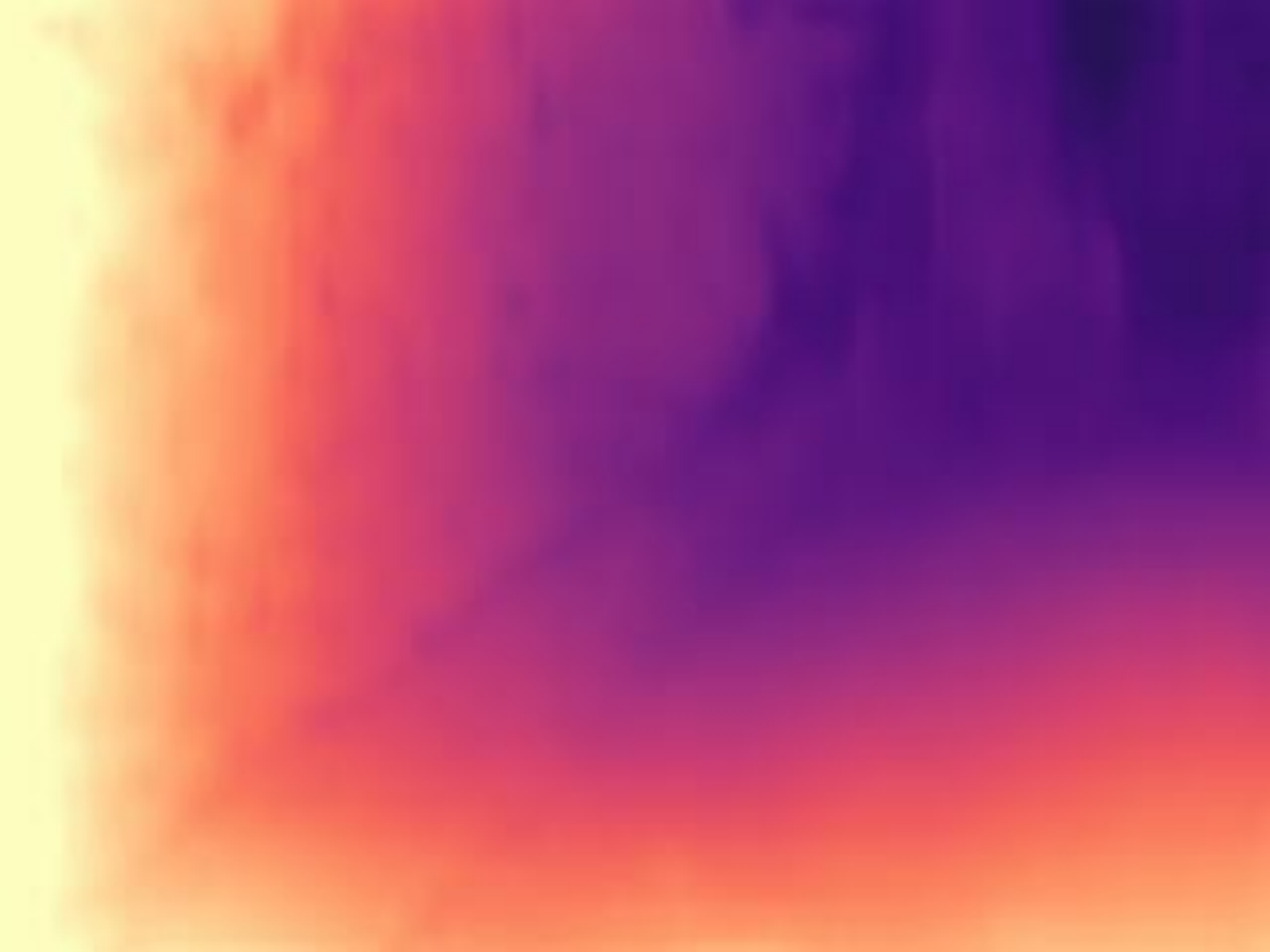}& 
\includegraphics[width=\w,height=\h]{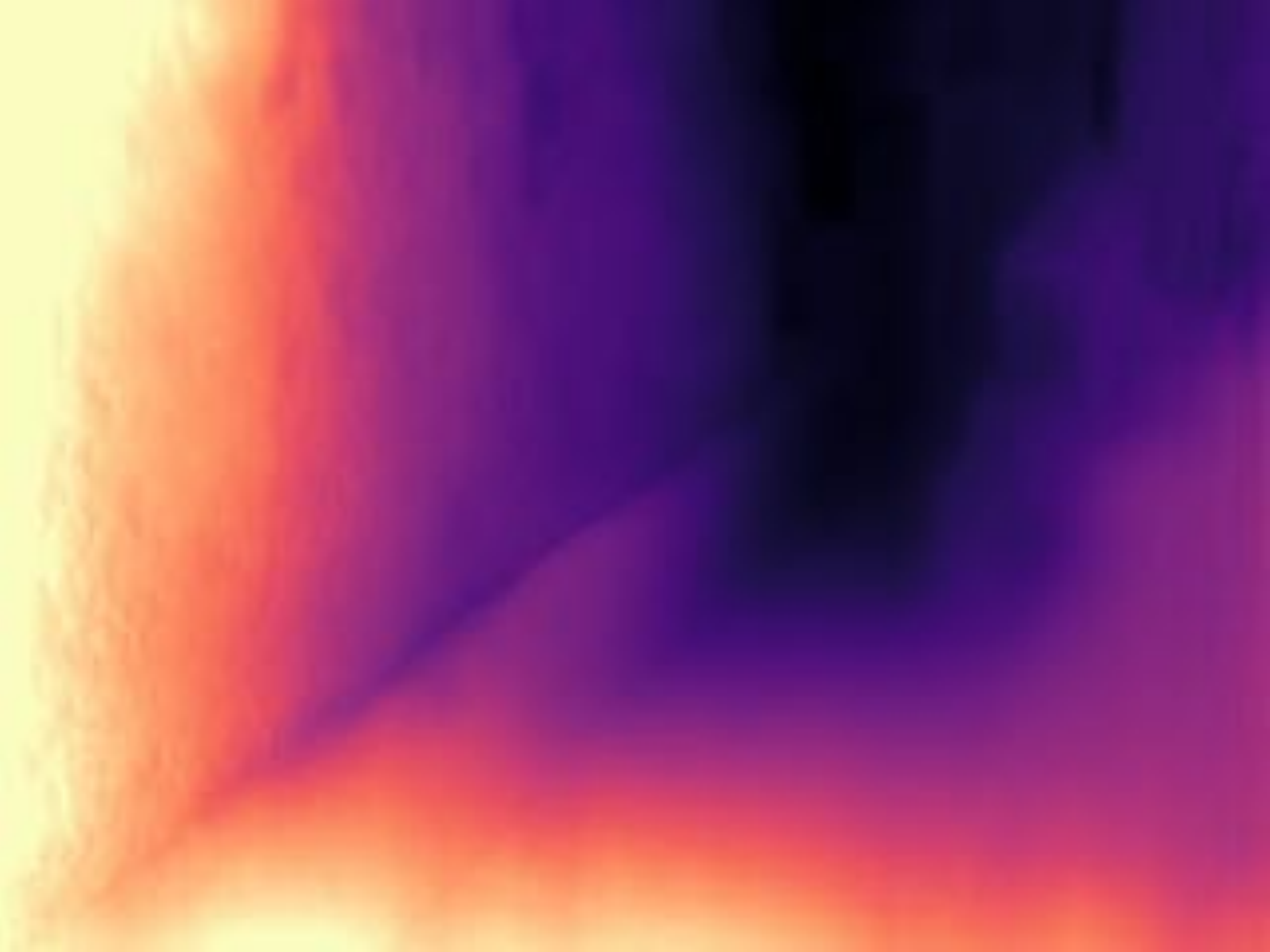}& 
\includegraphics[width=\w,height=\h]{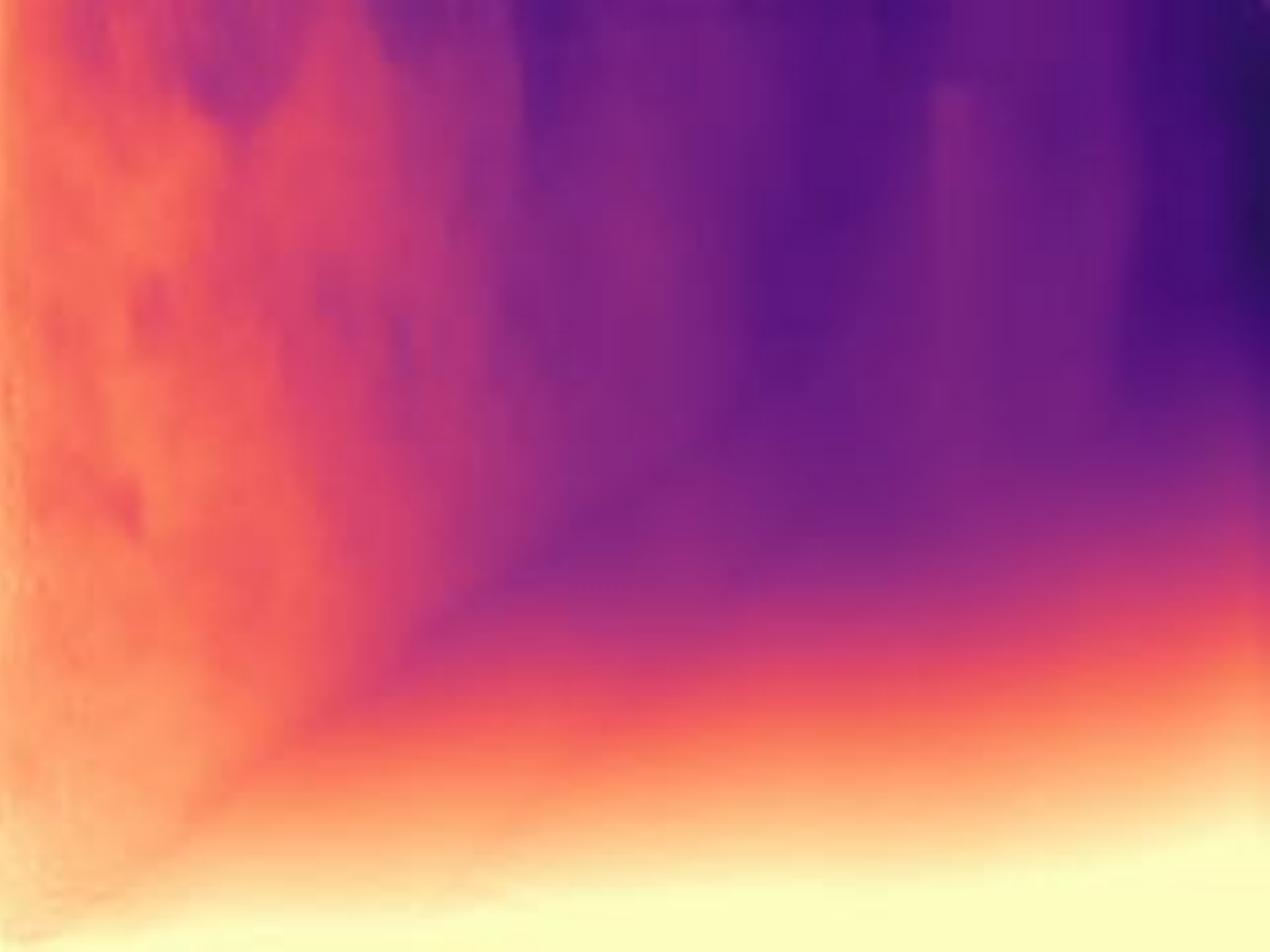}& 
\includegraphics[width=\w,height=\h]{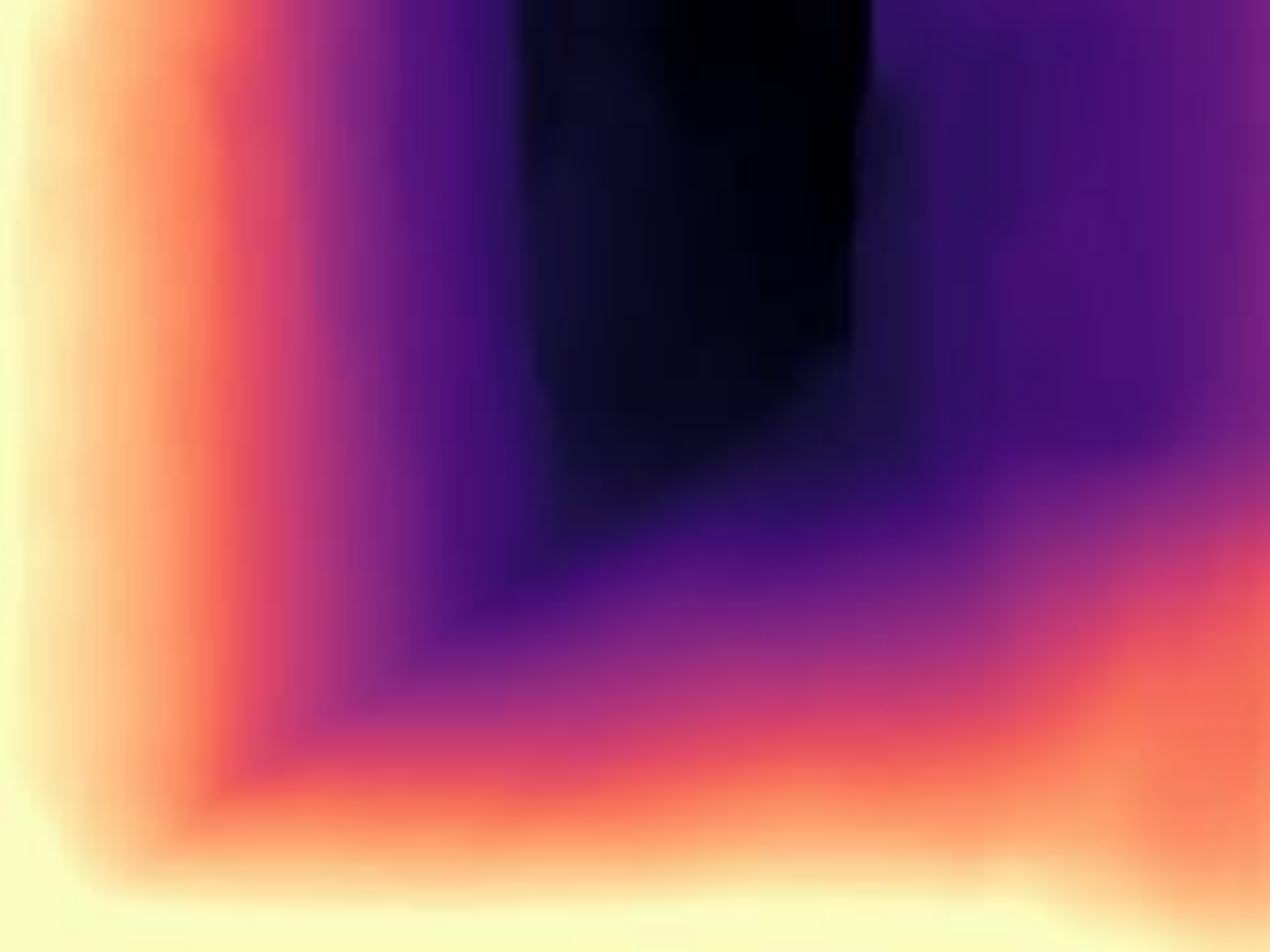}\\ 

\includegraphics[width=\w,height=\h]{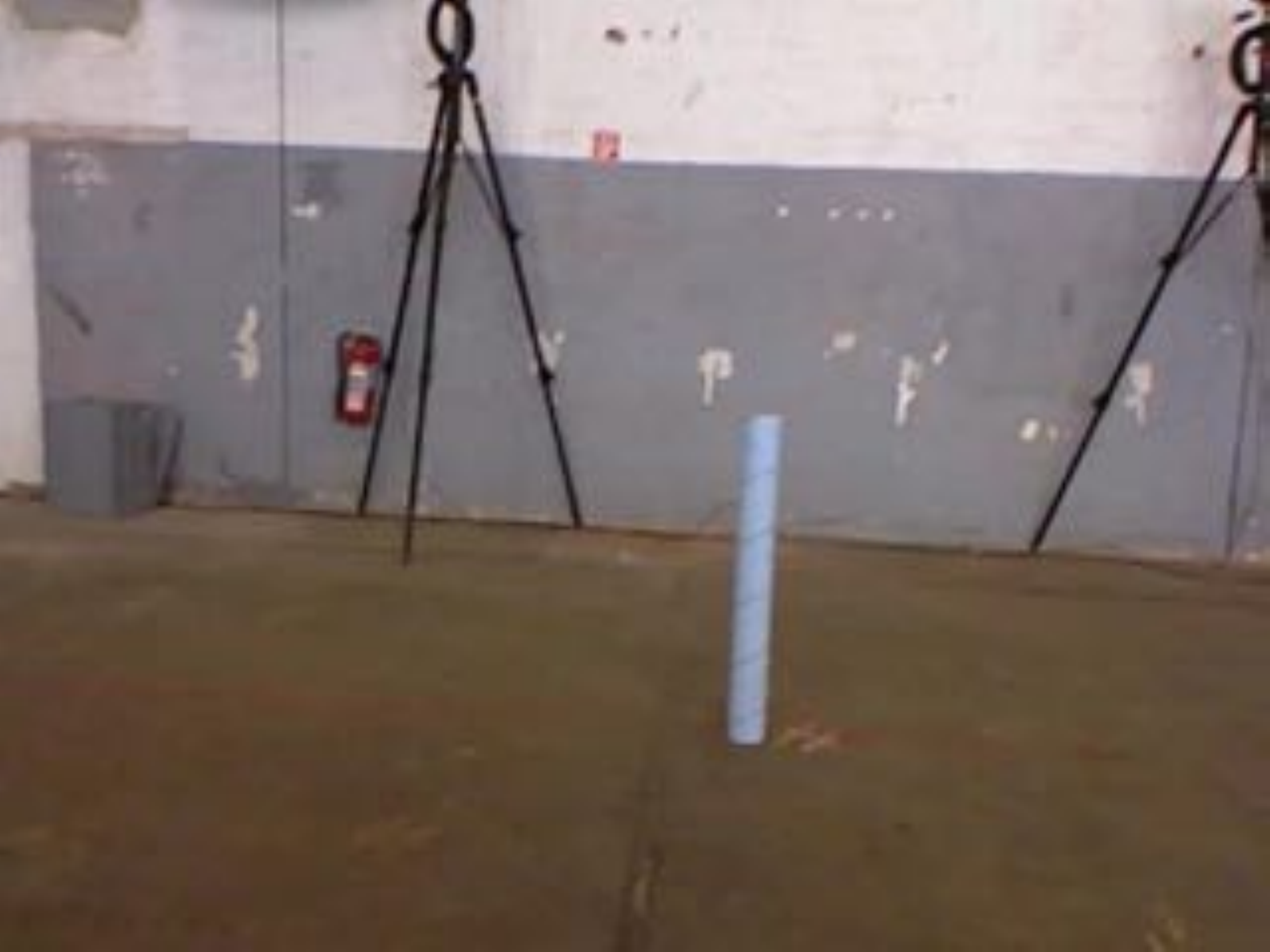}& 
\includegraphics[width=\w,height=\h]{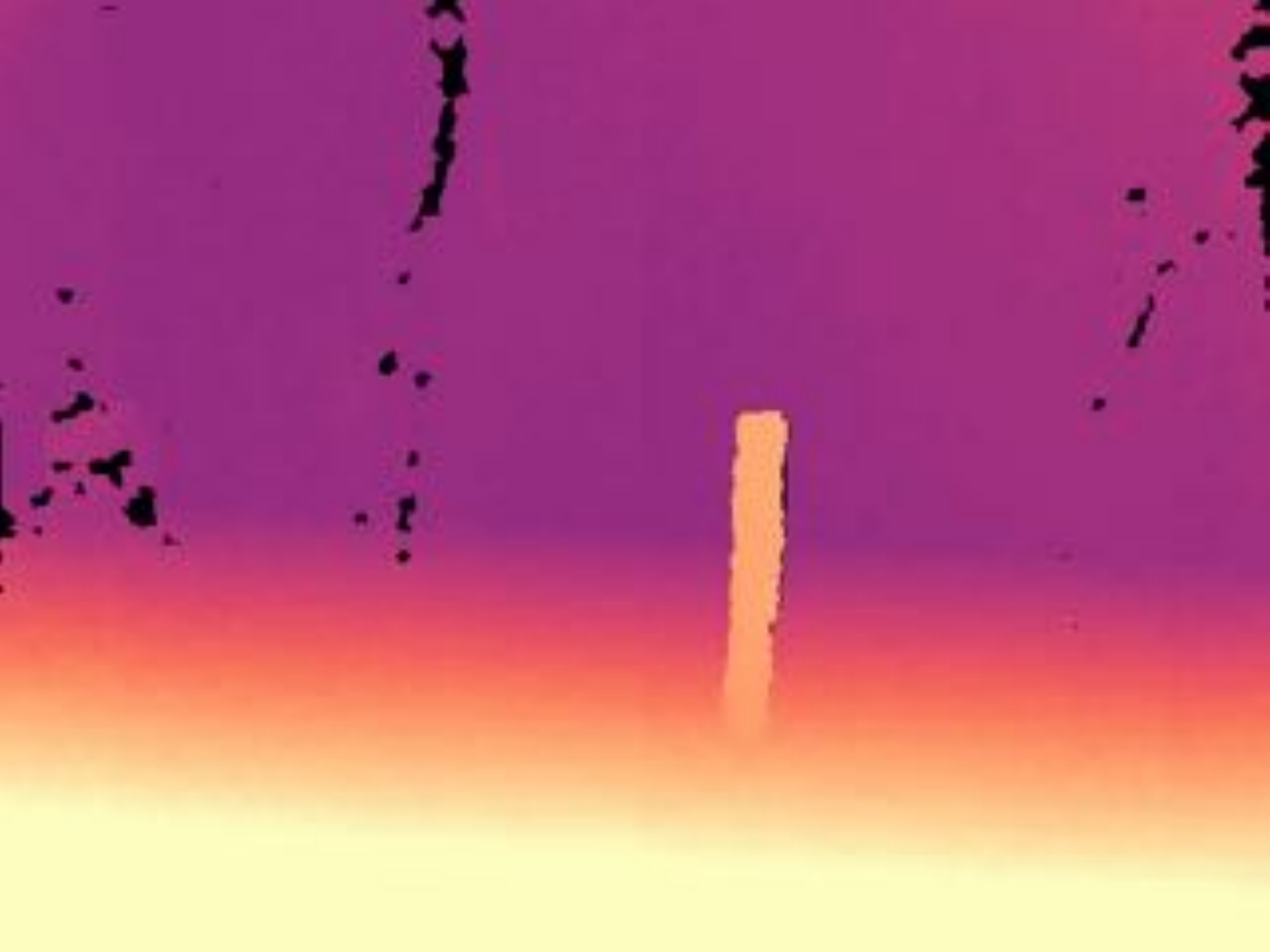}& 
\includegraphics[width=\w,height=\h]{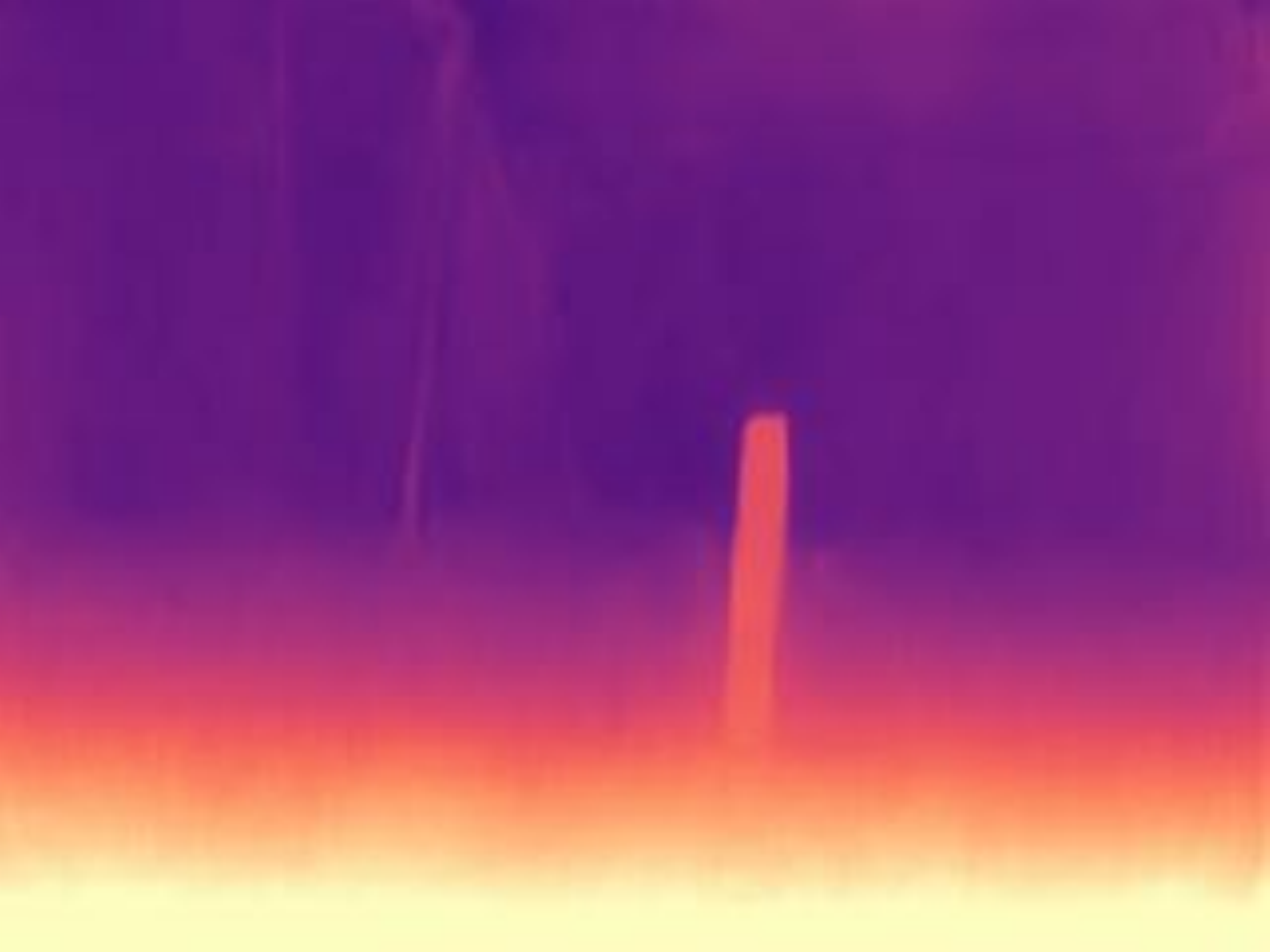}&
\includegraphics[width=\w,height=\h]{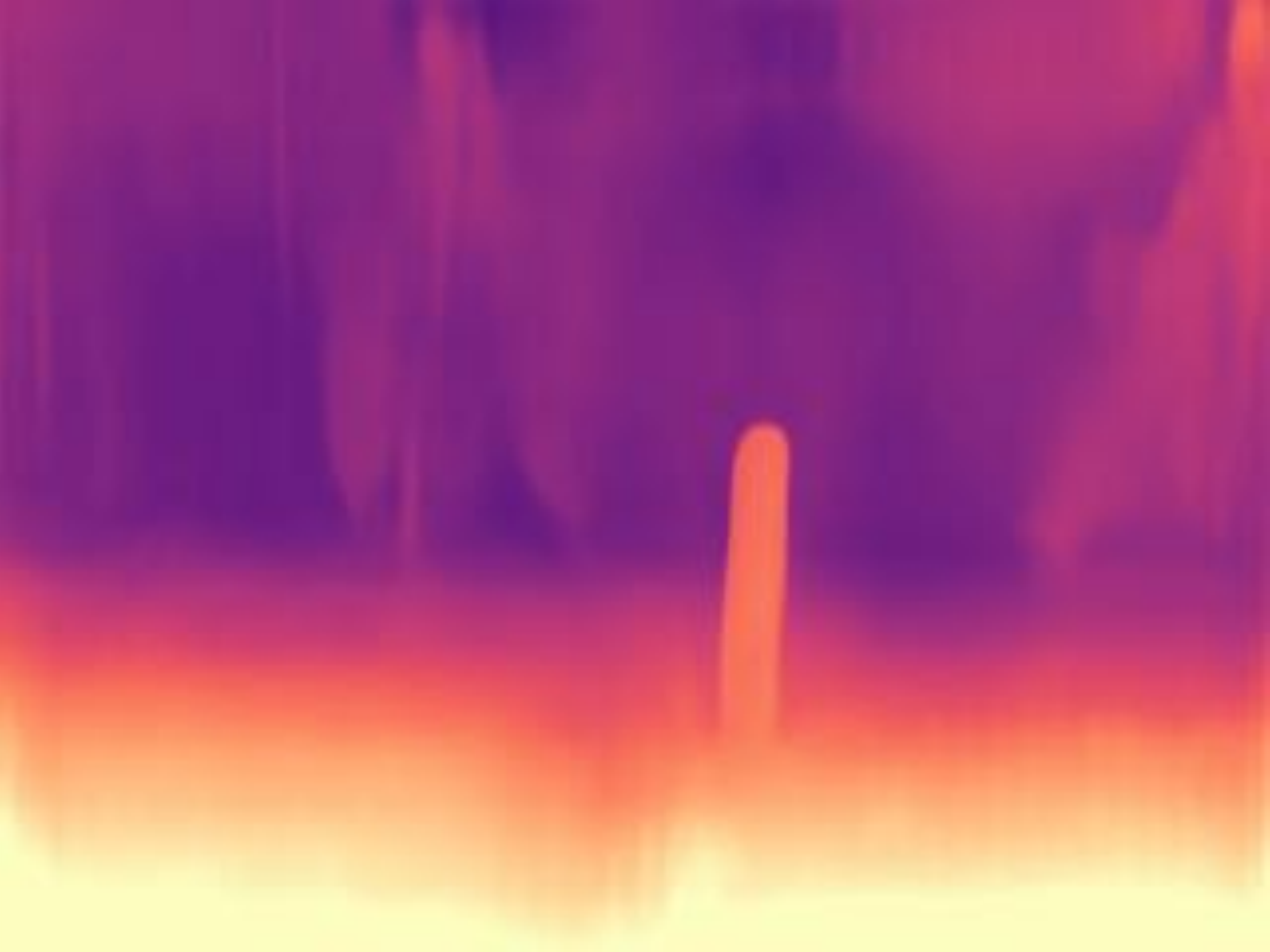}& 
\includegraphics[width=\w,height=\h]{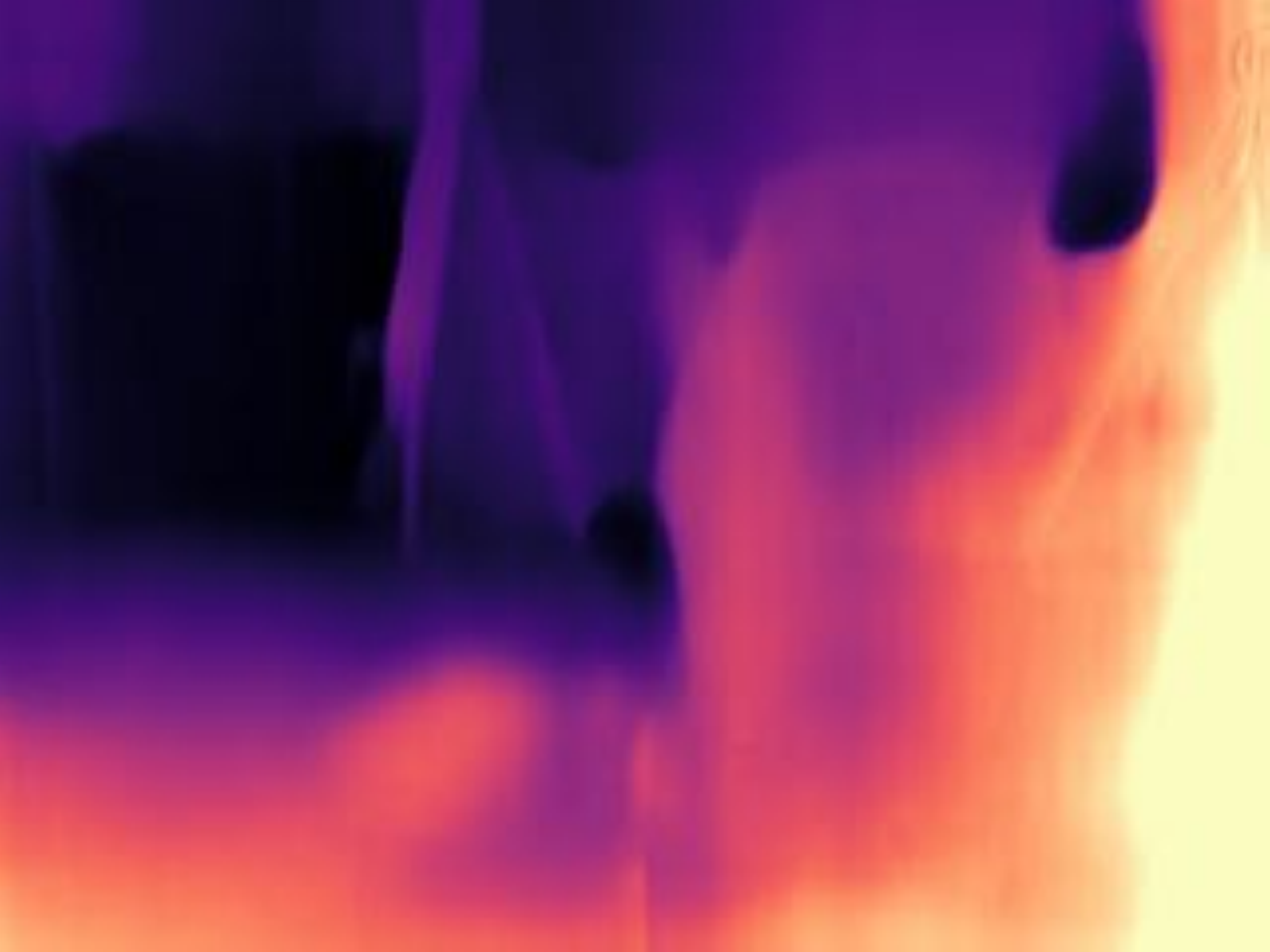}& 
\includegraphics[width=\w,height=\h]{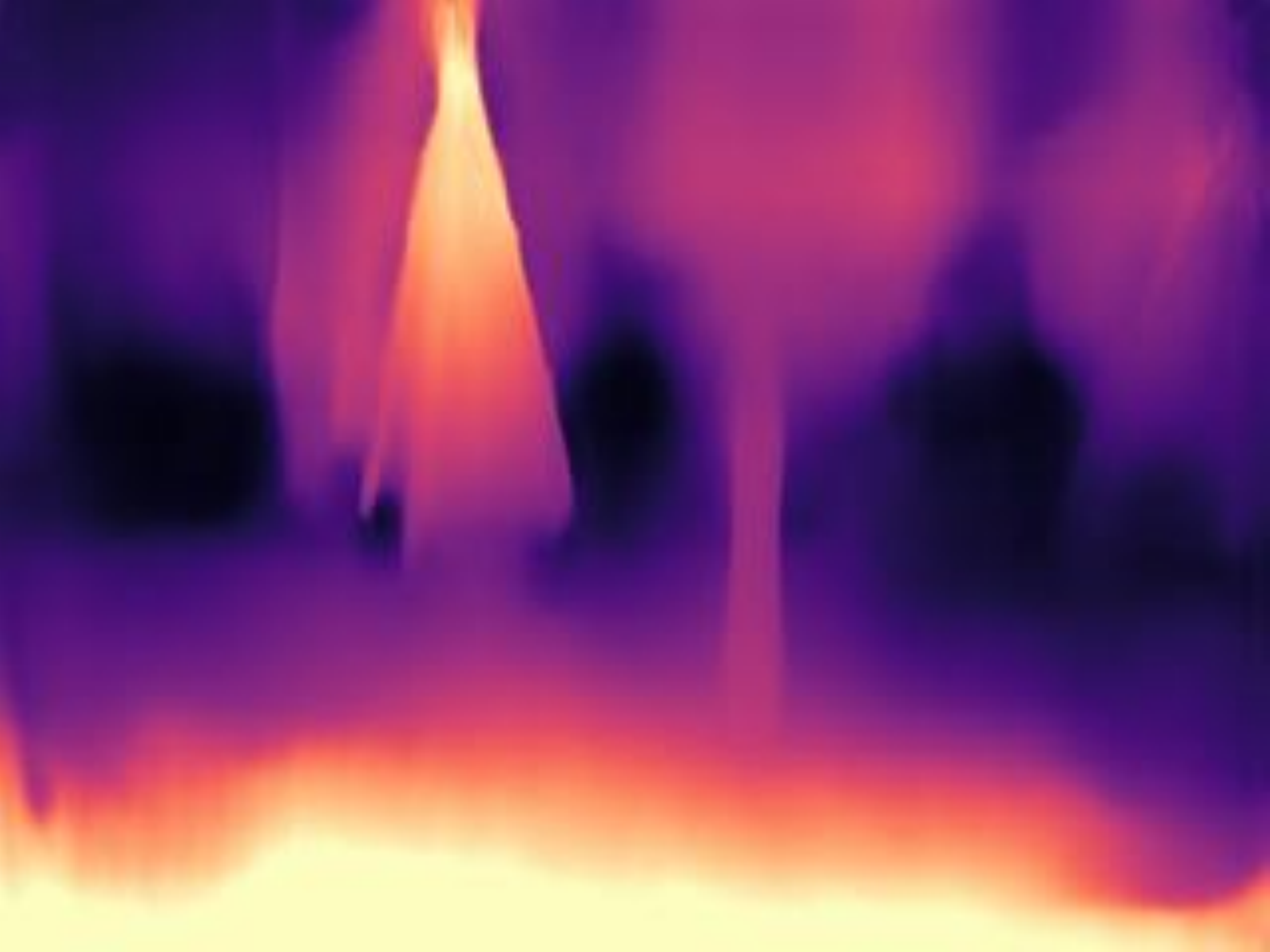}& 
\includegraphics[width=\w,height=\h]{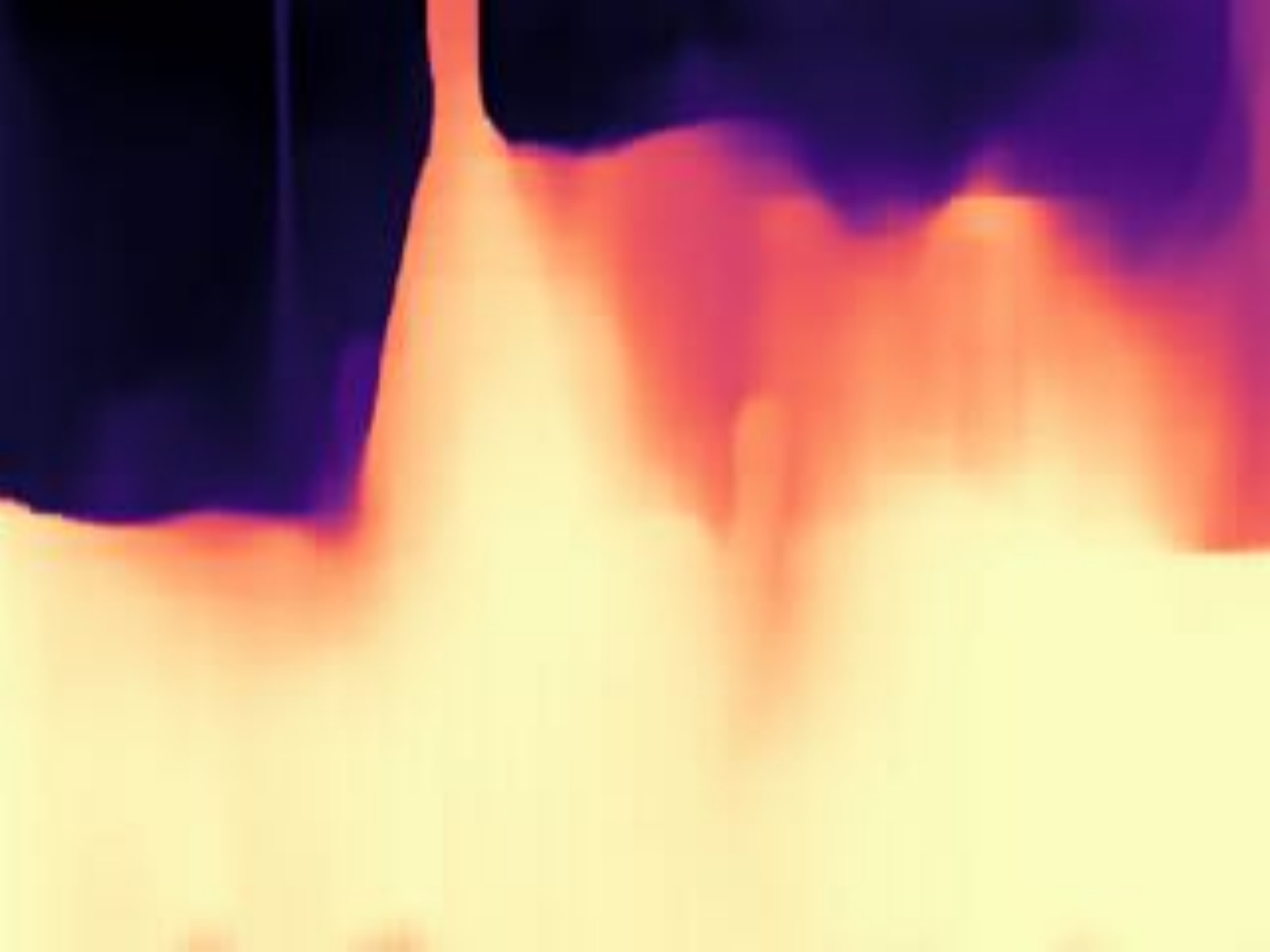}\\ 

\includegraphics[width=\w,height=\h]{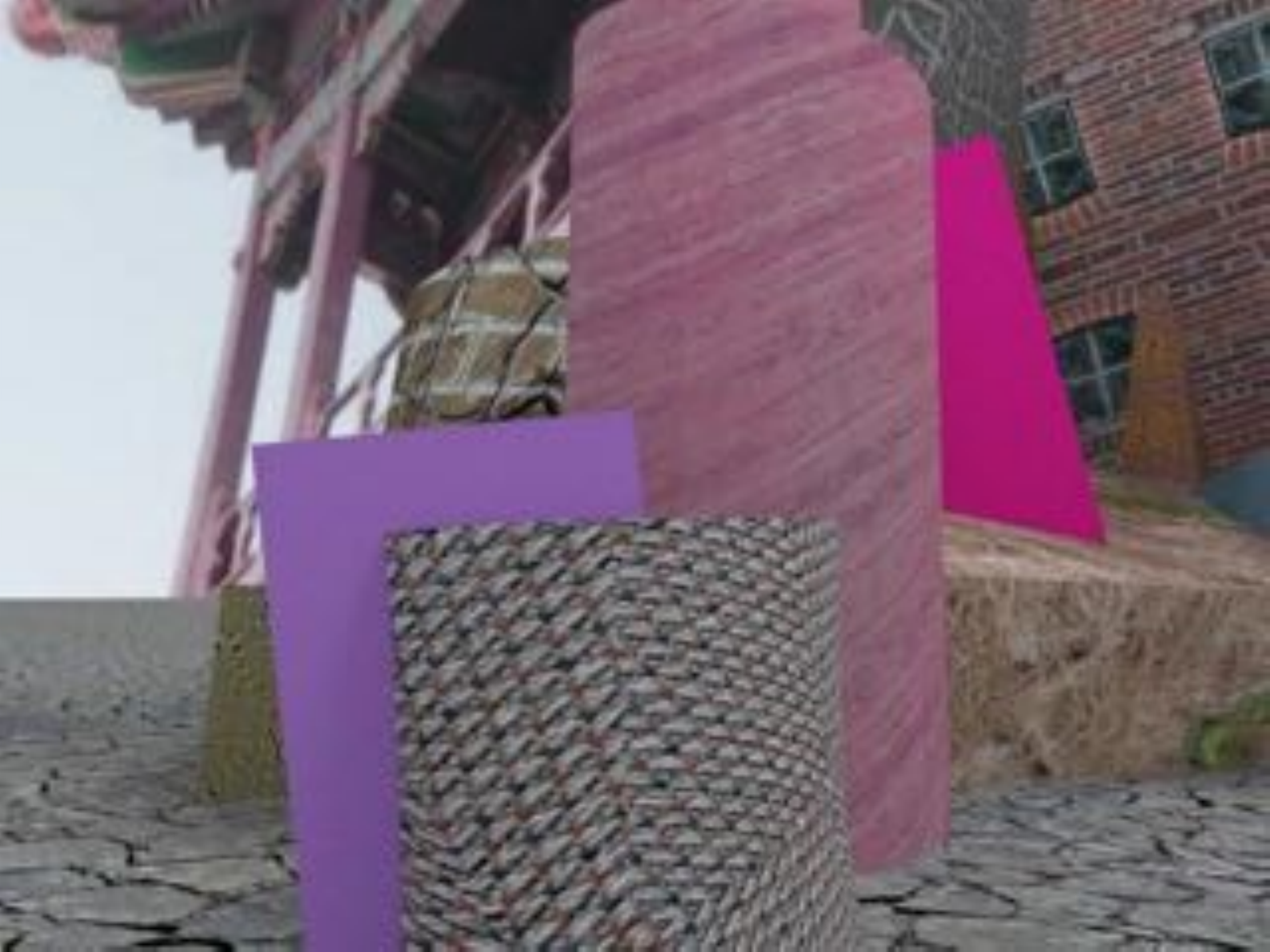}& 
\includegraphics[width=\w,height=\h]{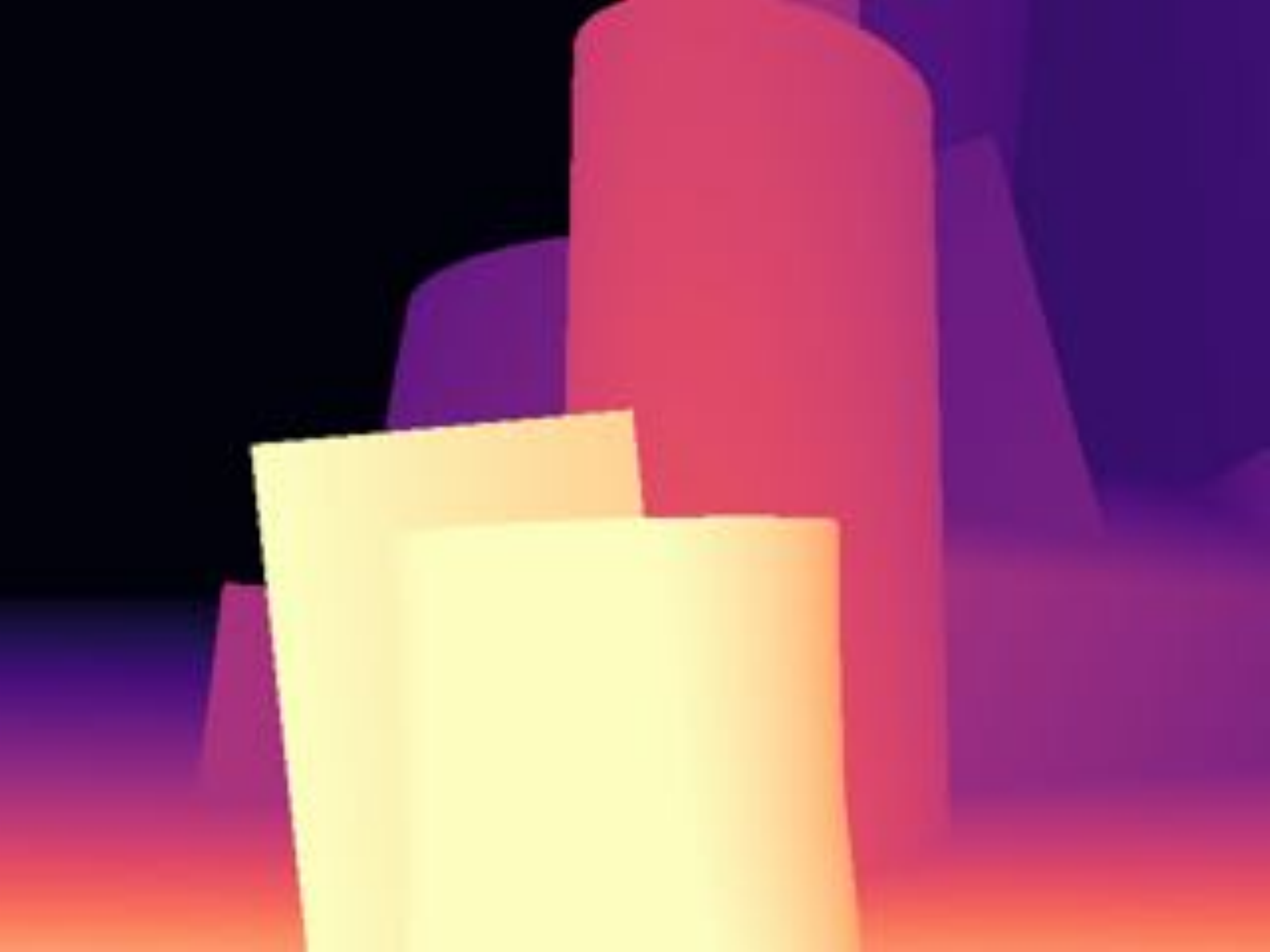}& 
\includegraphics[width=\w,height=\h]{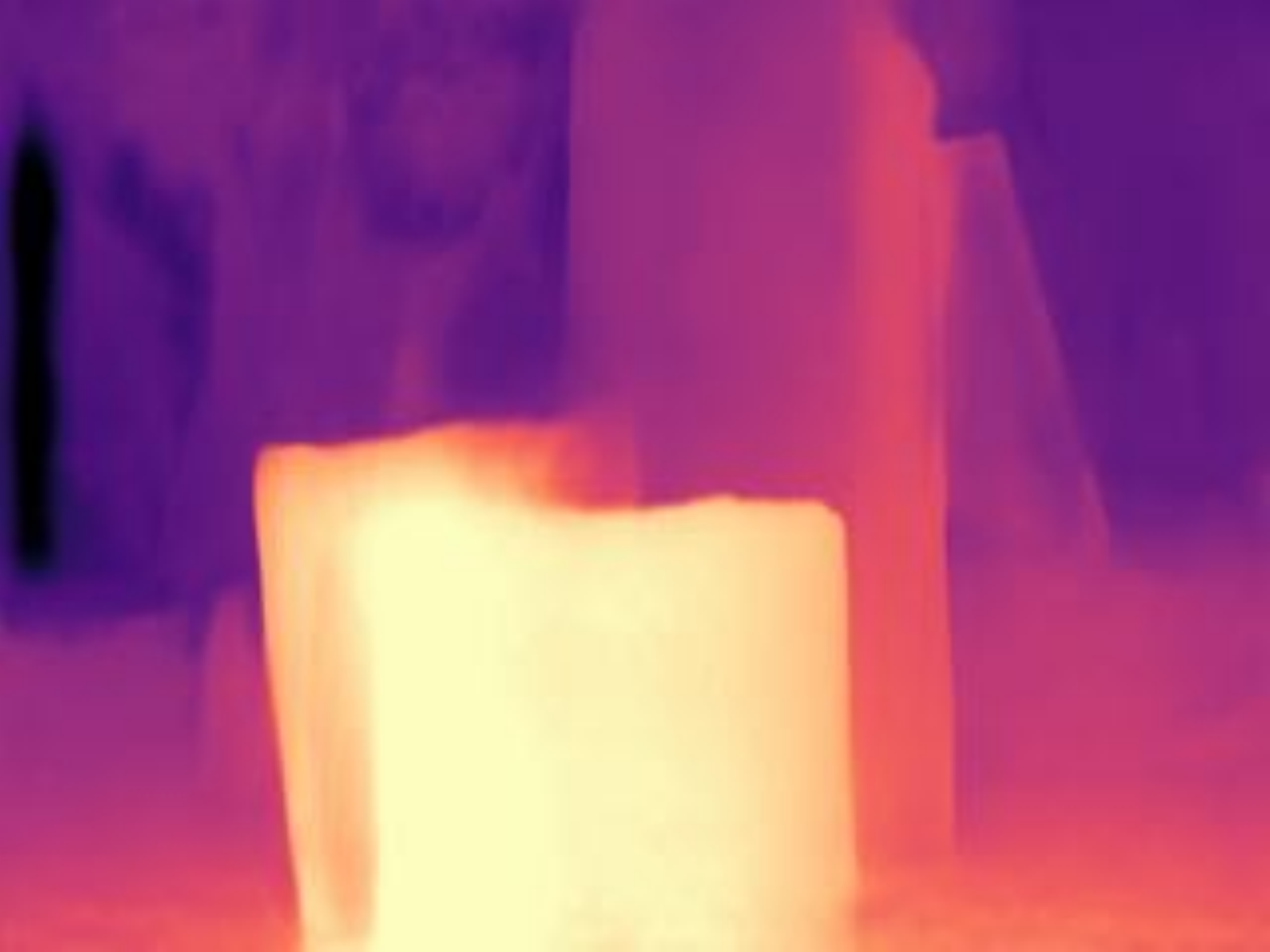}& 
\includegraphics[width=\w,height=\h]{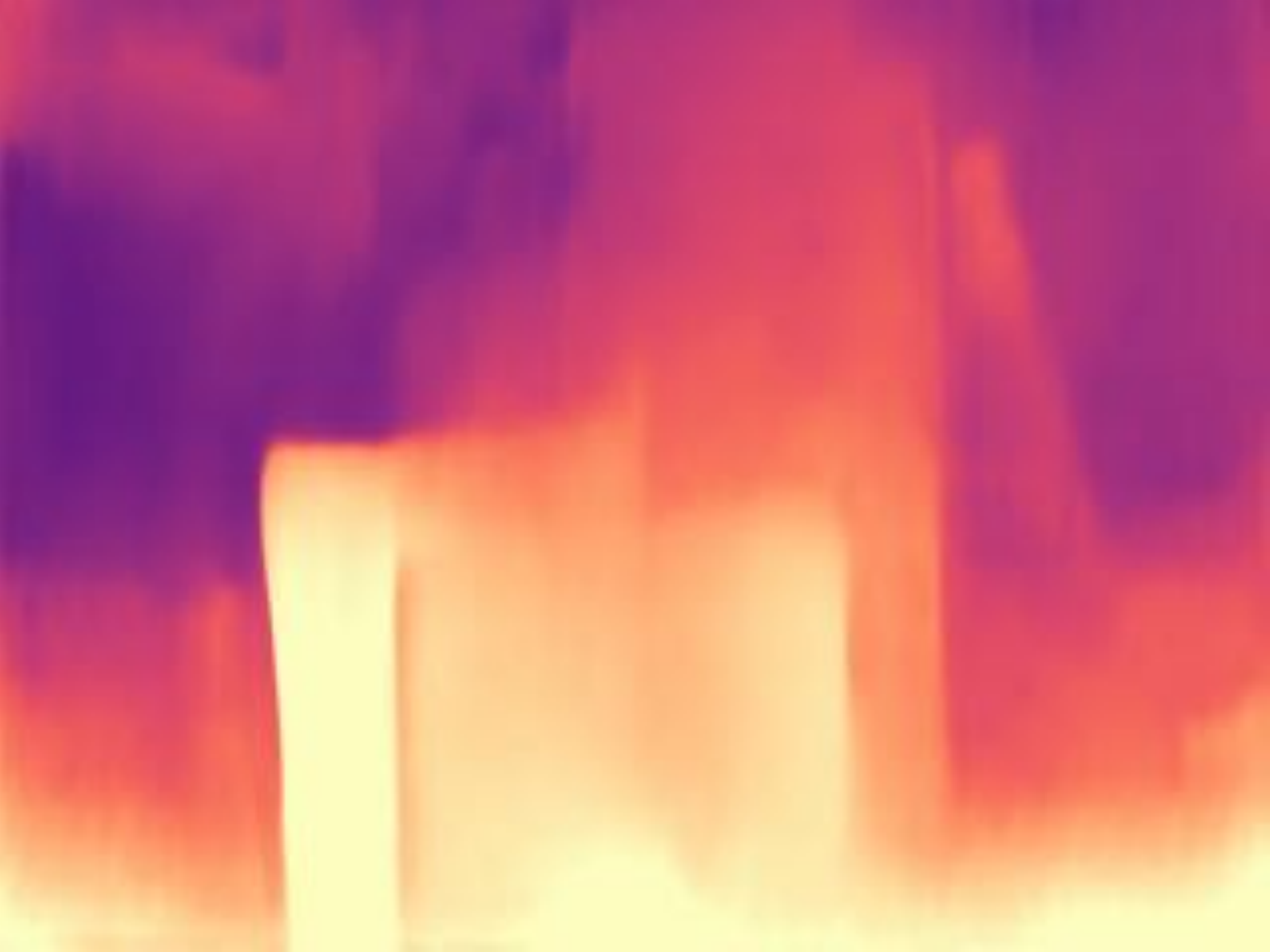}& 
\includegraphics[width=\w,height=\h]{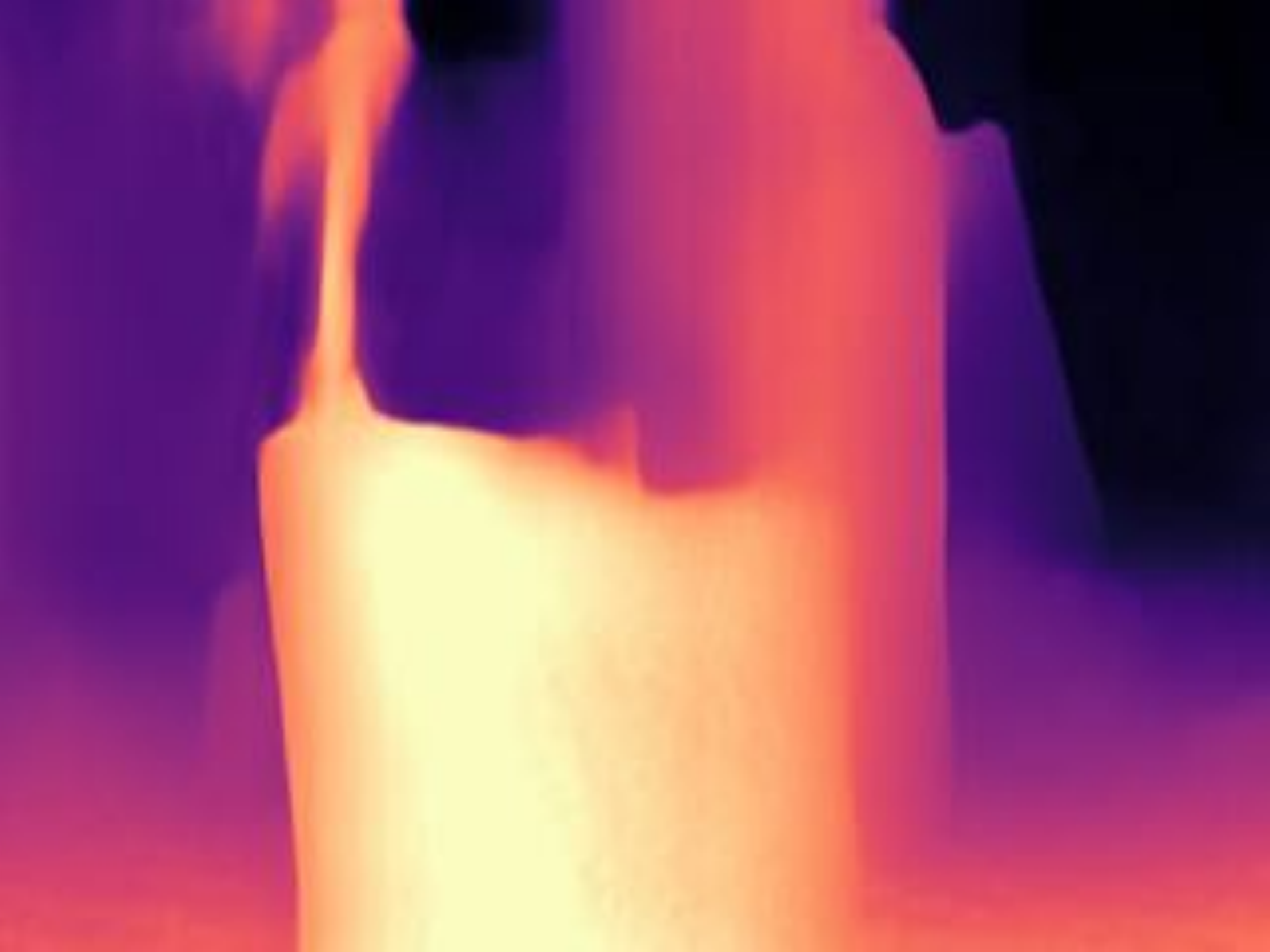}& 
\includegraphics[width=\w,height=\h]{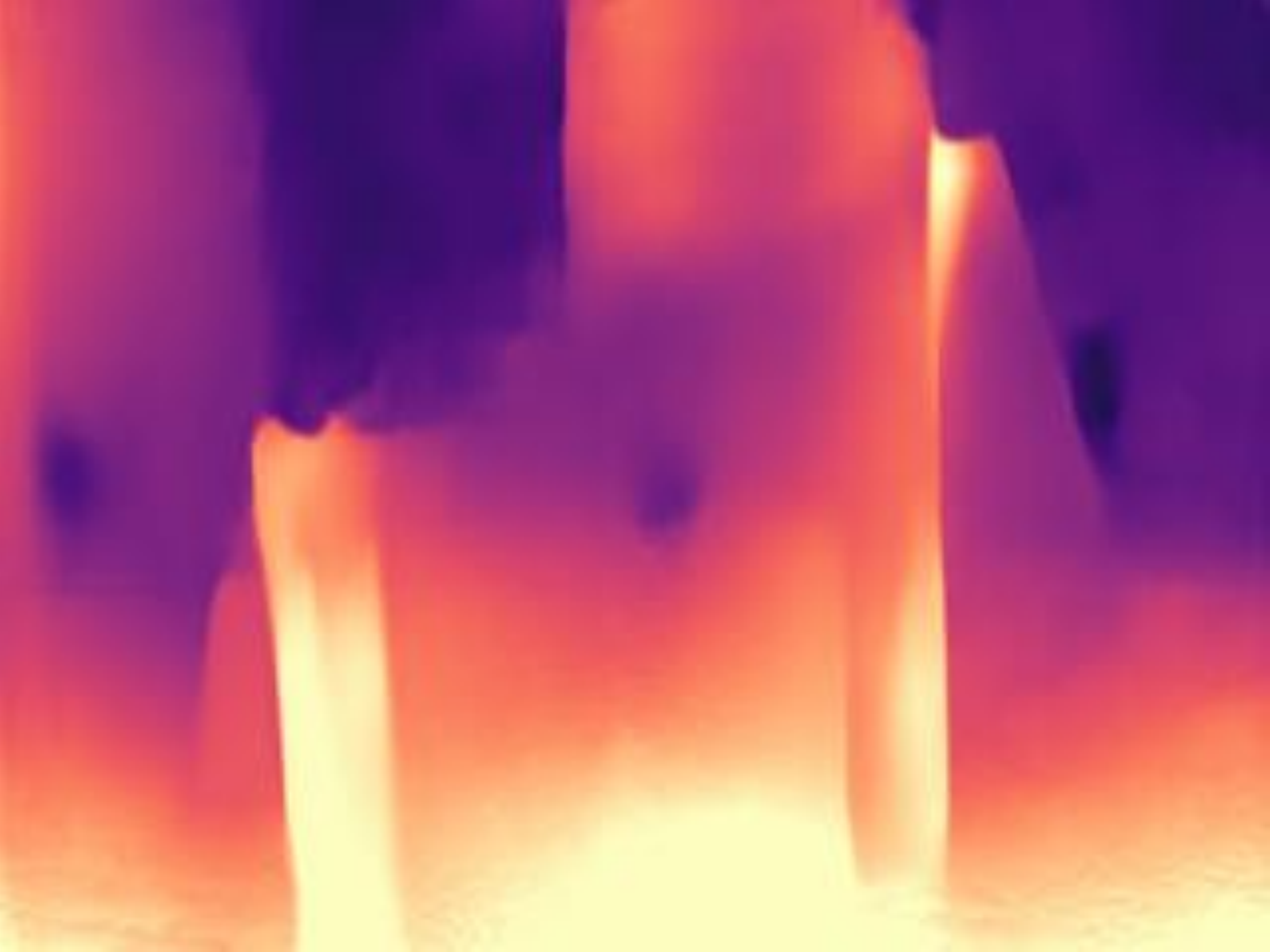}& 
\includegraphics[width=\w,height=\h]{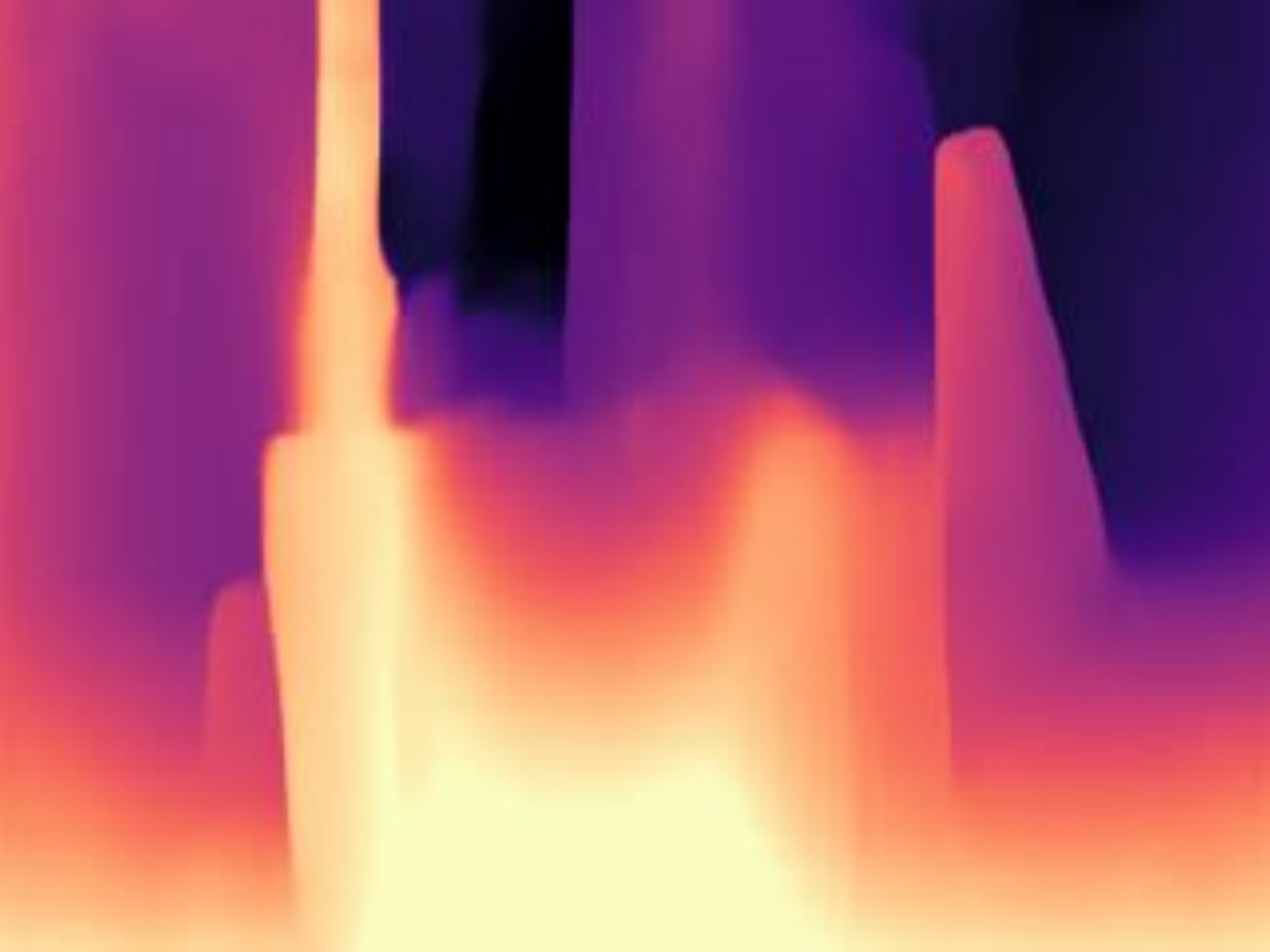}\\ 

\includegraphics[width=\w,height=\h]{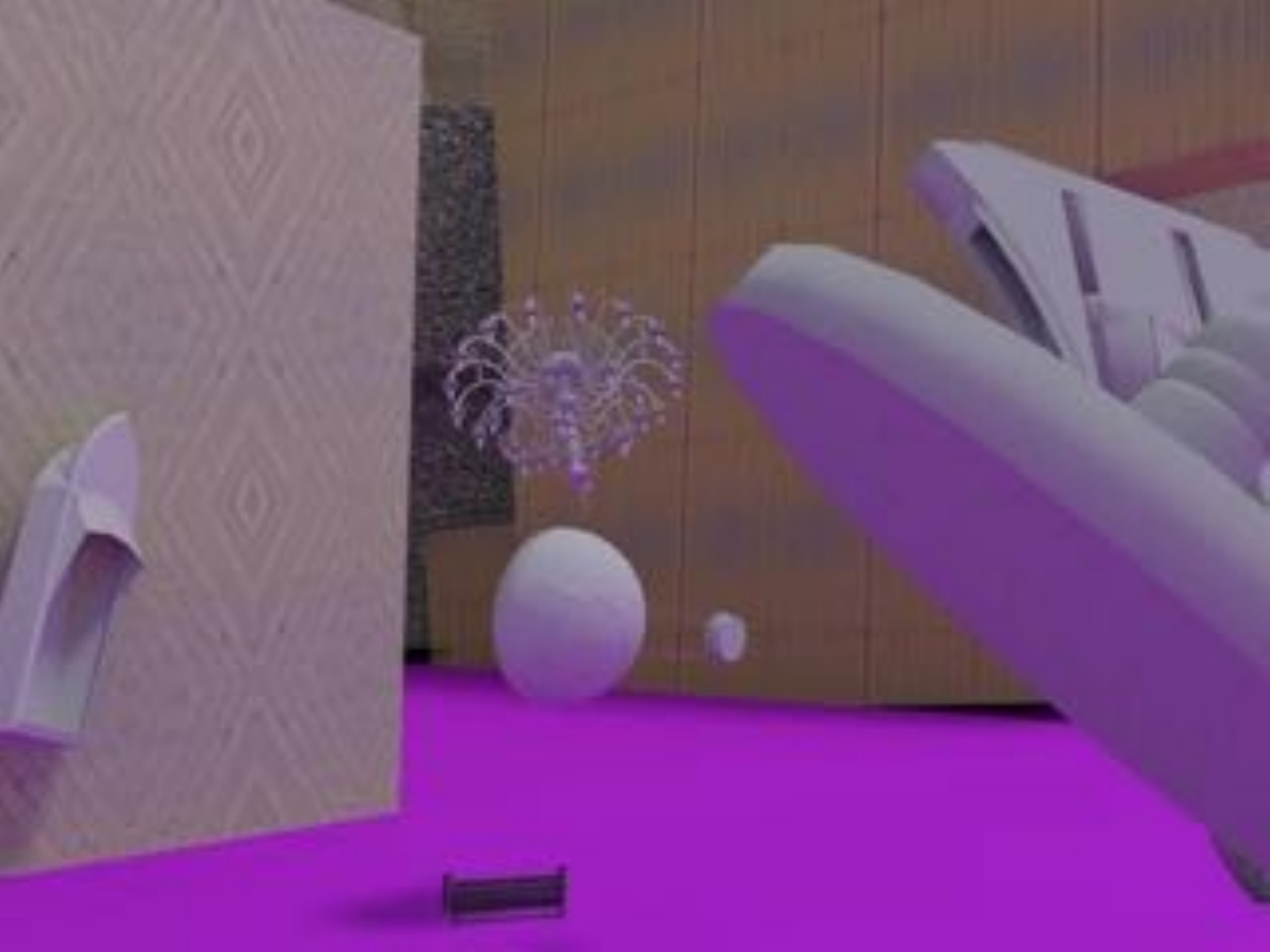}& 
\includegraphics[width=\w,height=\h]{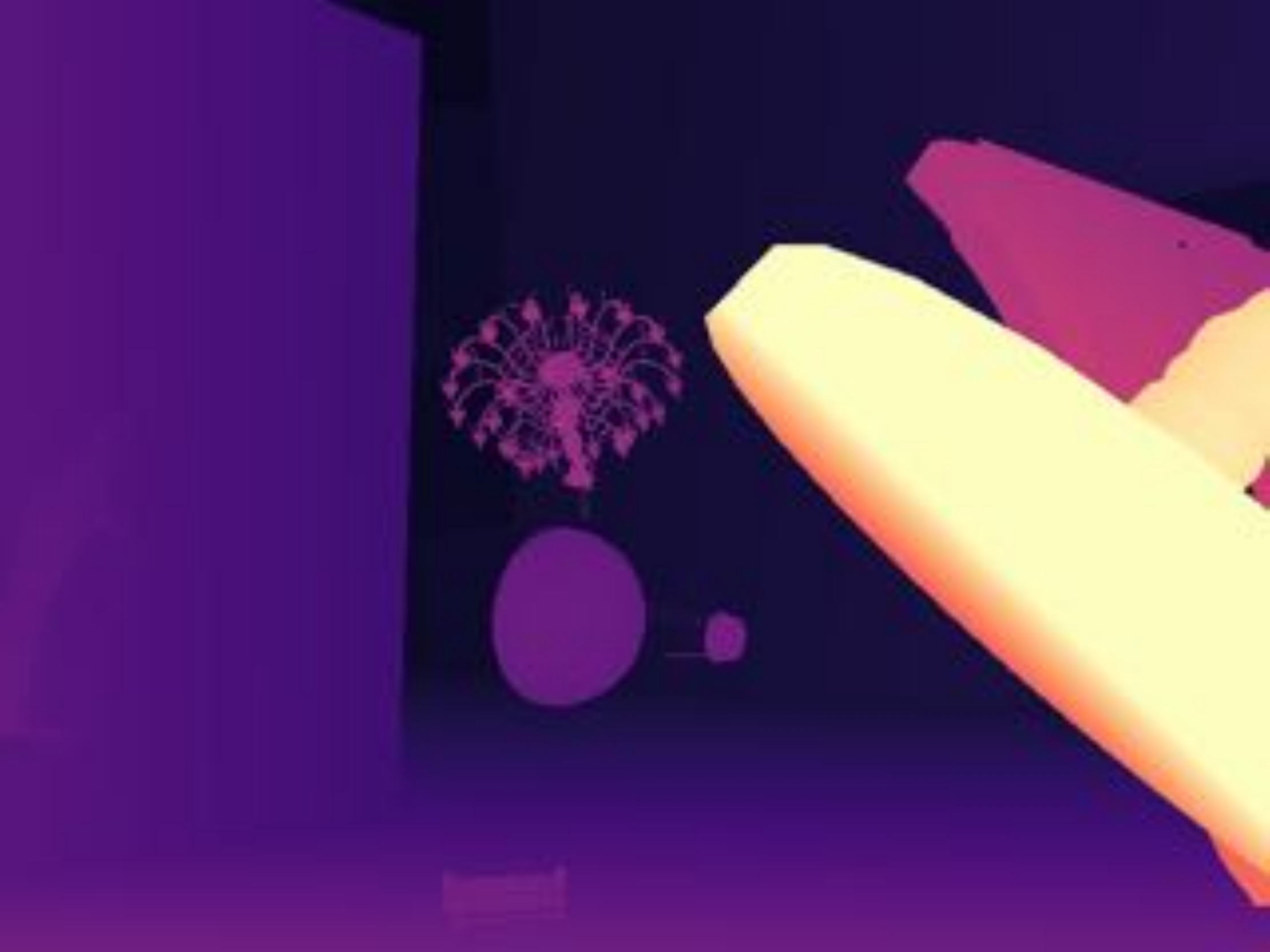}& 
\includegraphics[width=\w,height=\h]{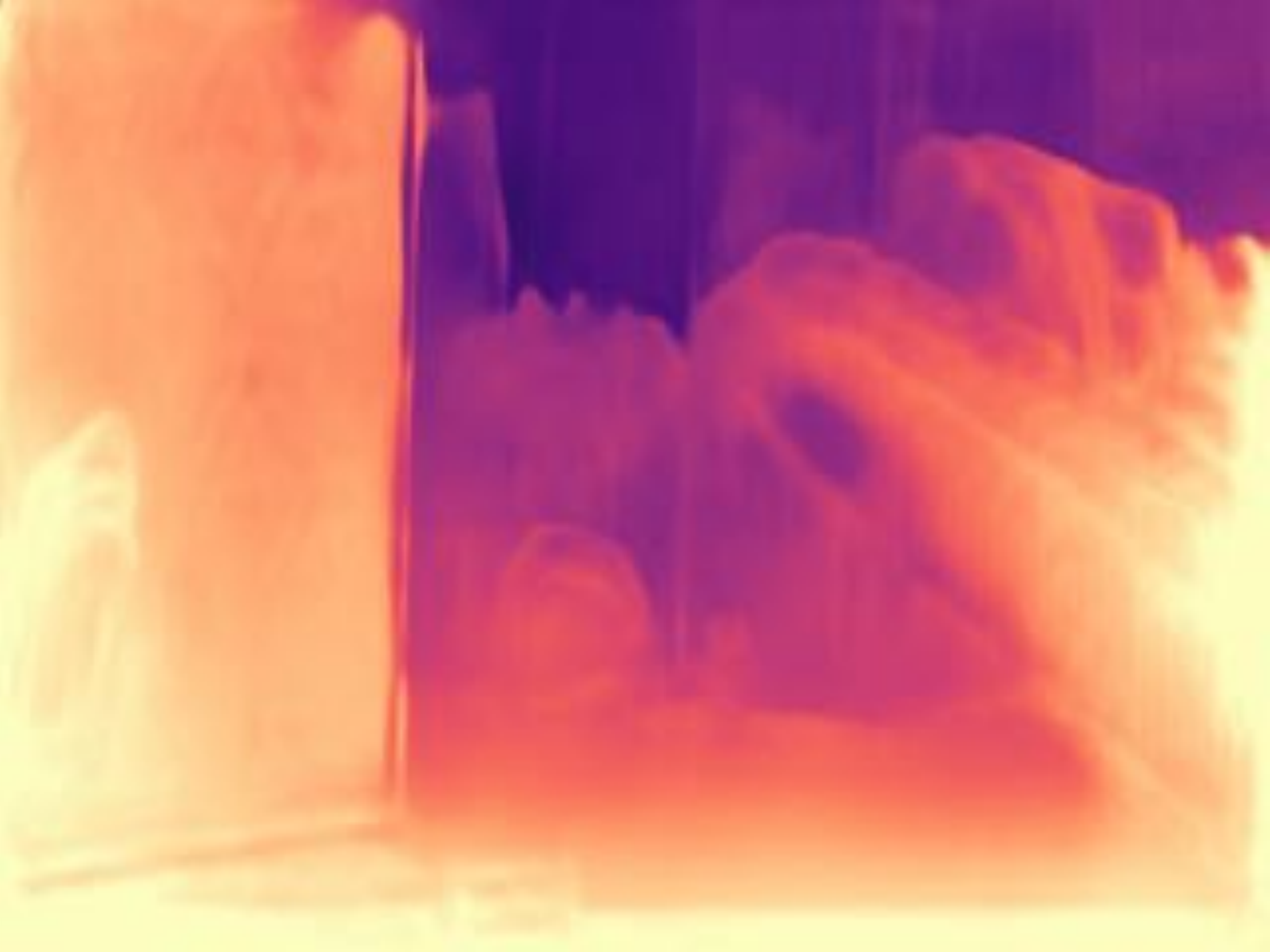}& 
\includegraphics[width=\w,height=\h]{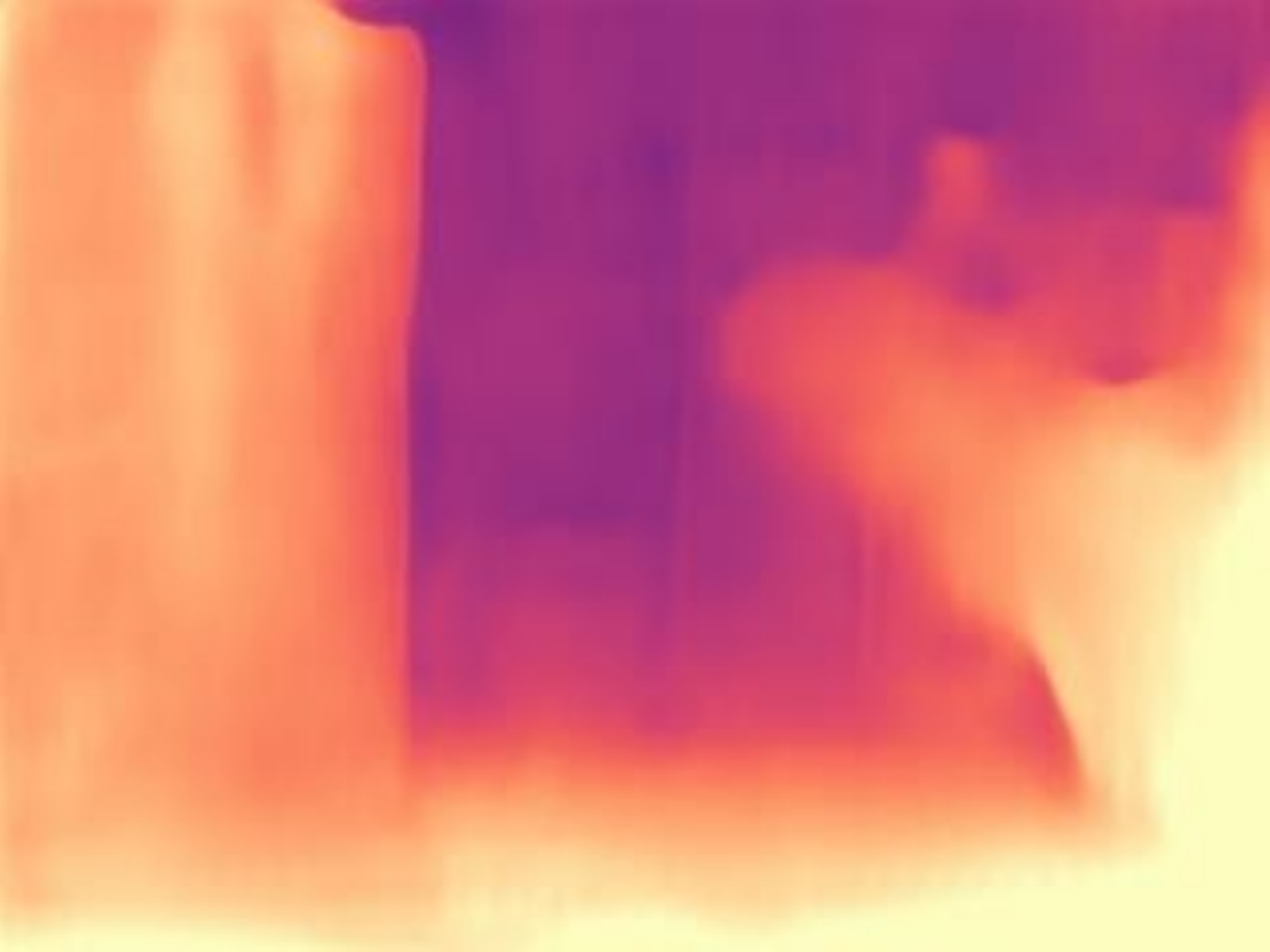}& 
\includegraphics[width=\w,height=\h]{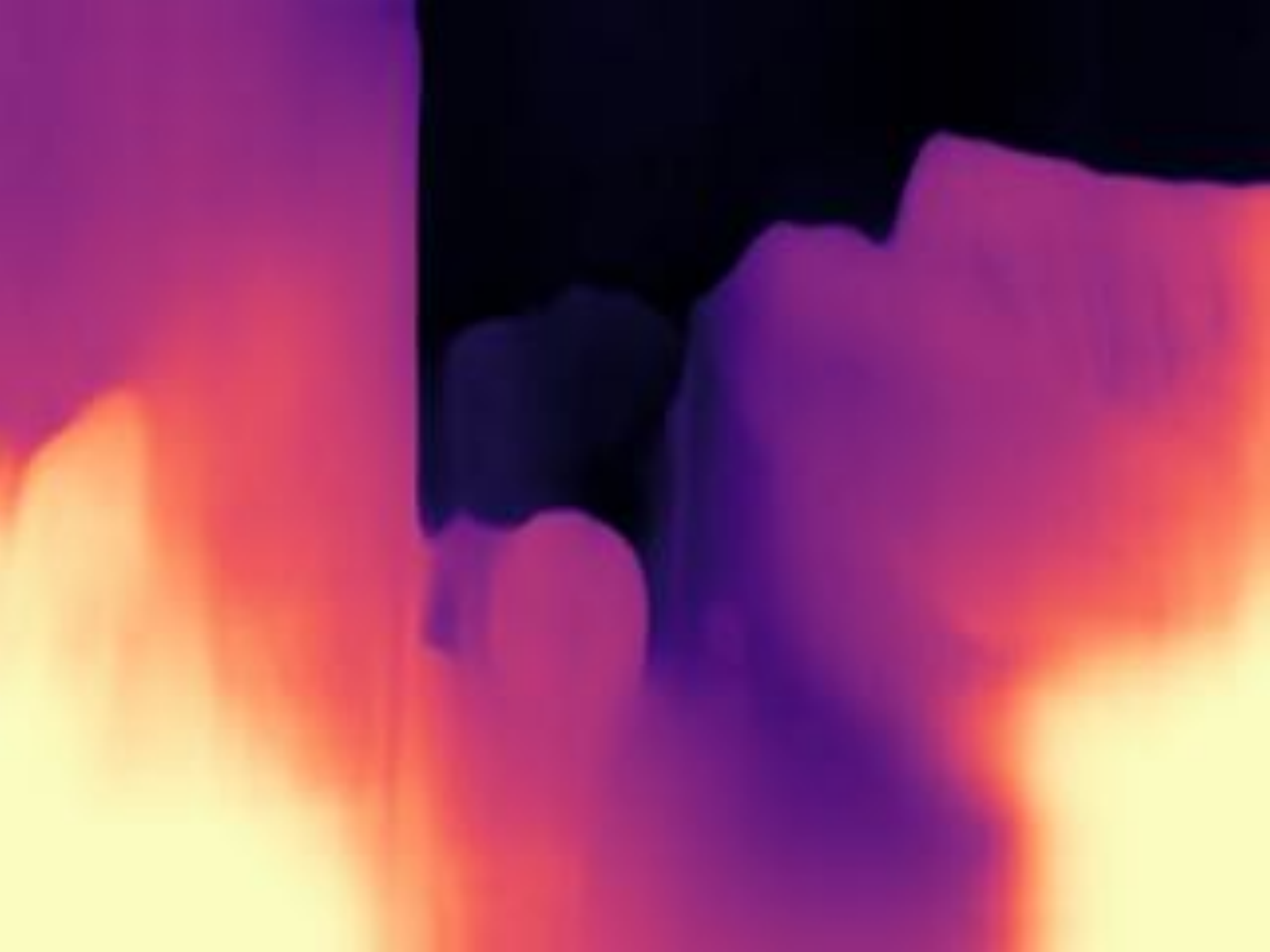}& 
\includegraphics[width=\w,height=\h]{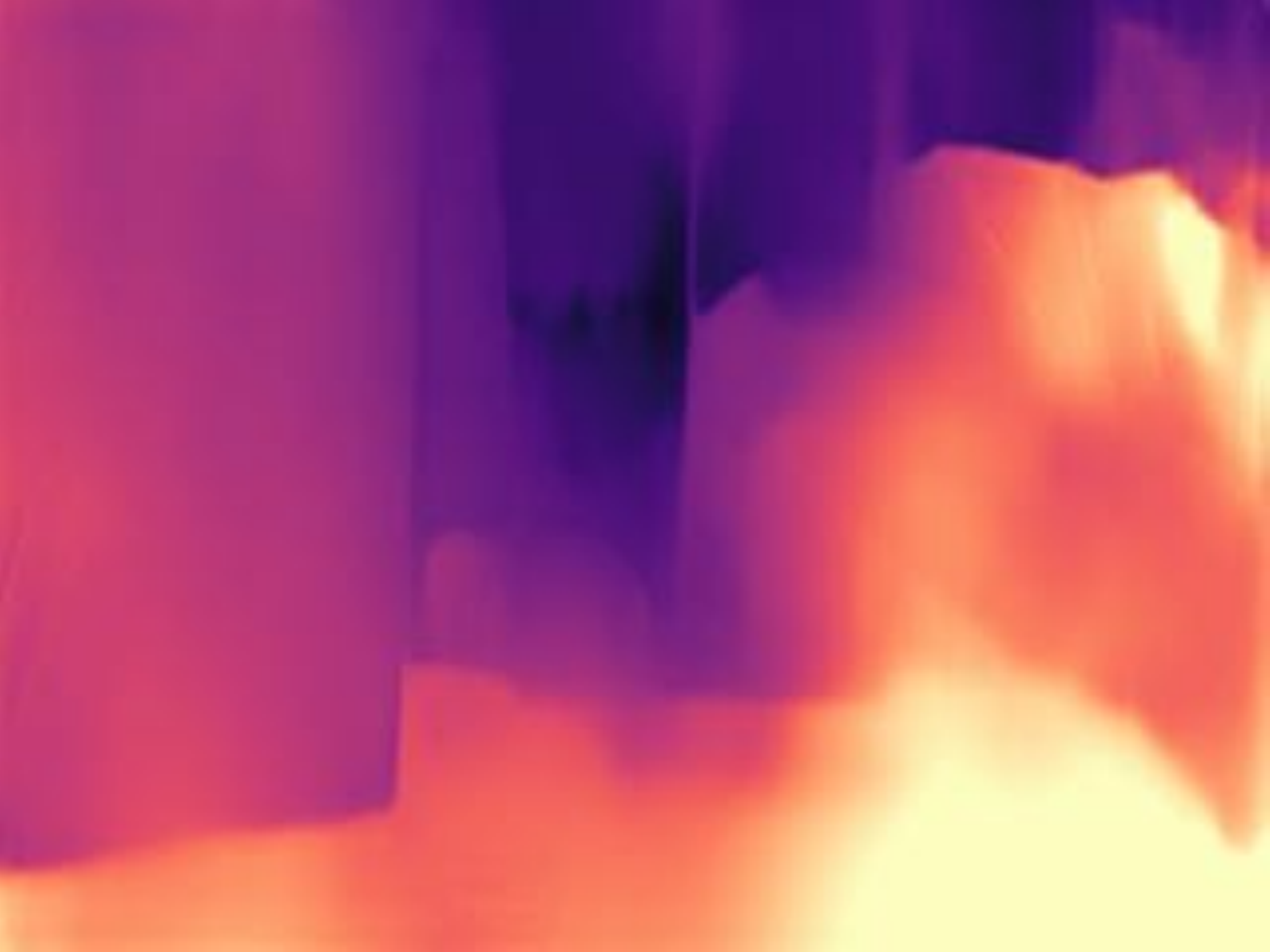}& 
\includegraphics[width=\w,height=\h]{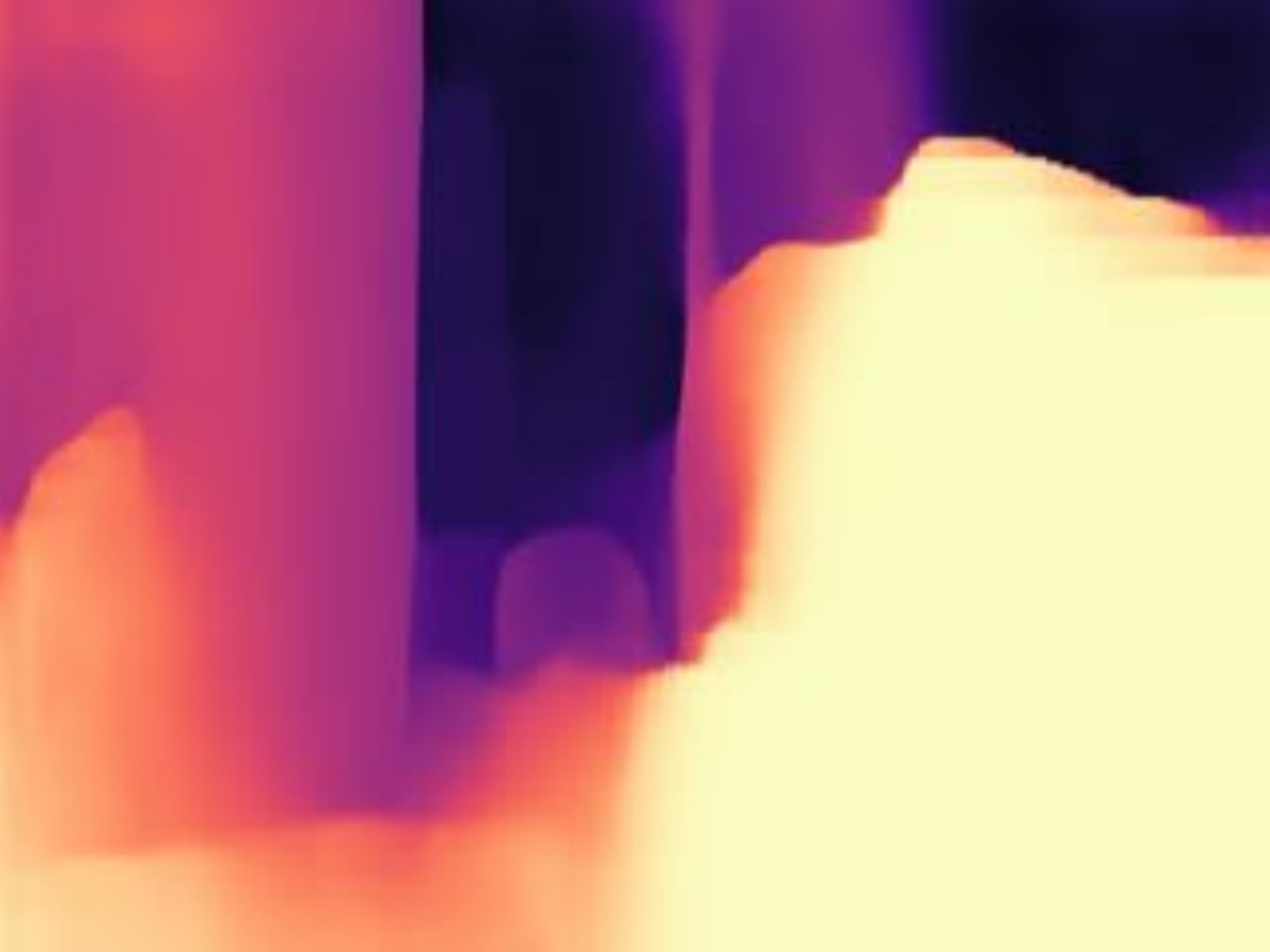}\\ 

     \fontsize{40}{30} \selectfont Input images & 
     \fontsize{40}{30} \selectfont GT depths &
     \fontsize{40}{30} \selectfont Ours-Hybrid &
     \fontsize{40}{30} \selectfont Ours-ViT &
     \fontsize{40}{30} \selectfont Monodepth2 & 
     \fontsize{40}{30} \selectfont PackNet-SfM & 
     \fontsize{40}{30} \selectfont R-MSFM6
    \end{tabular}}
    \vspace{-0.3cm}
    \caption{\textbf{Comparison of depth map results on various dataset.} We test our model and the competitive models trained on KITTI using MVS, SUN3D, RGBD and Scenes11 (Top to Bottom).}
    \label{figure_result_demons_apdx}
     \end{subfigure}
\begin{subfigure}
   \centering
    \resizebox{\linewidth}{!}{
\begin{tabular}{ccccccc}

\includegraphics[width=\w,height=\h]{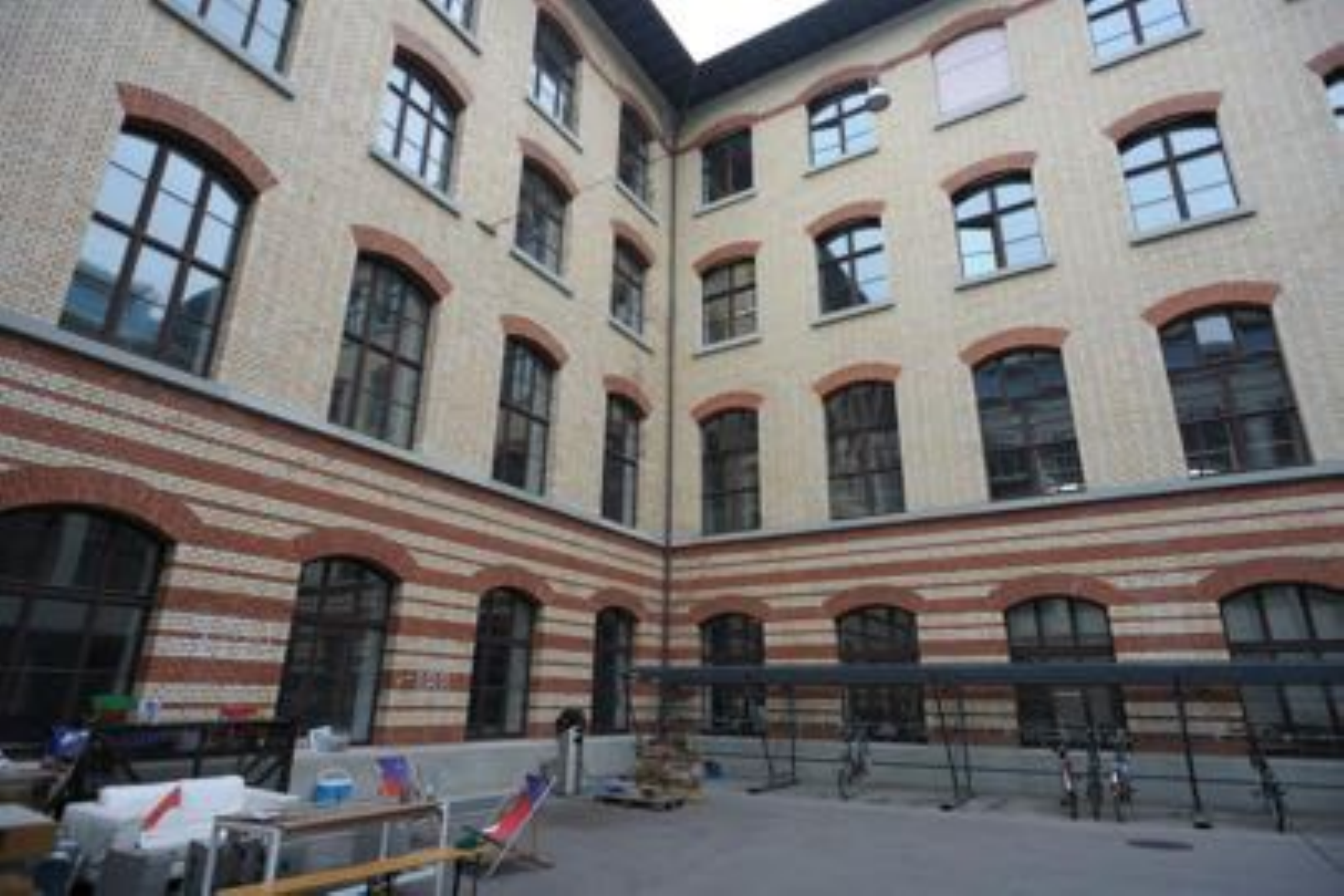}& 
\includegraphics[width=\w,height=\h]{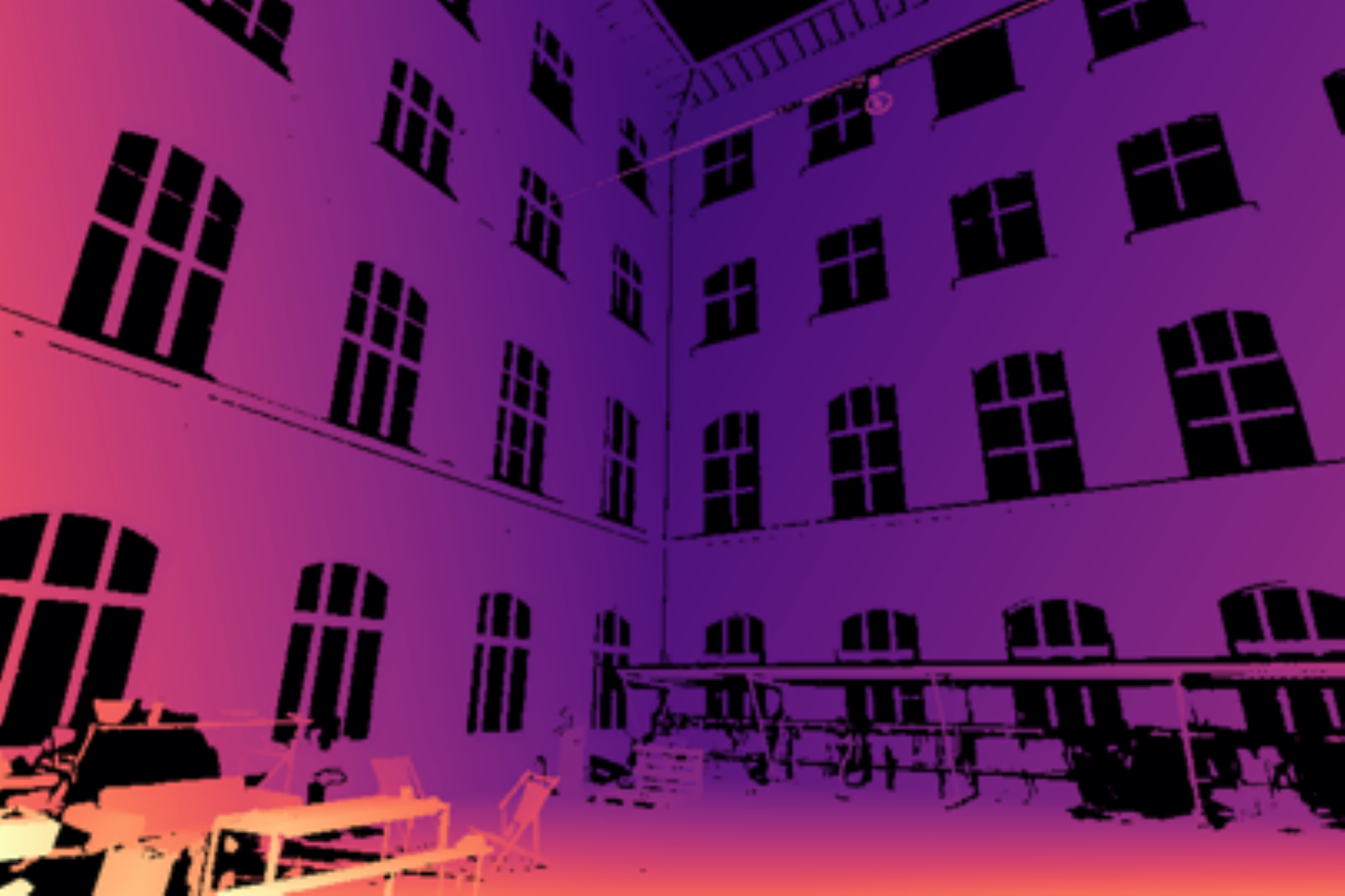}& 
\includegraphics[width=\w,height=\h]{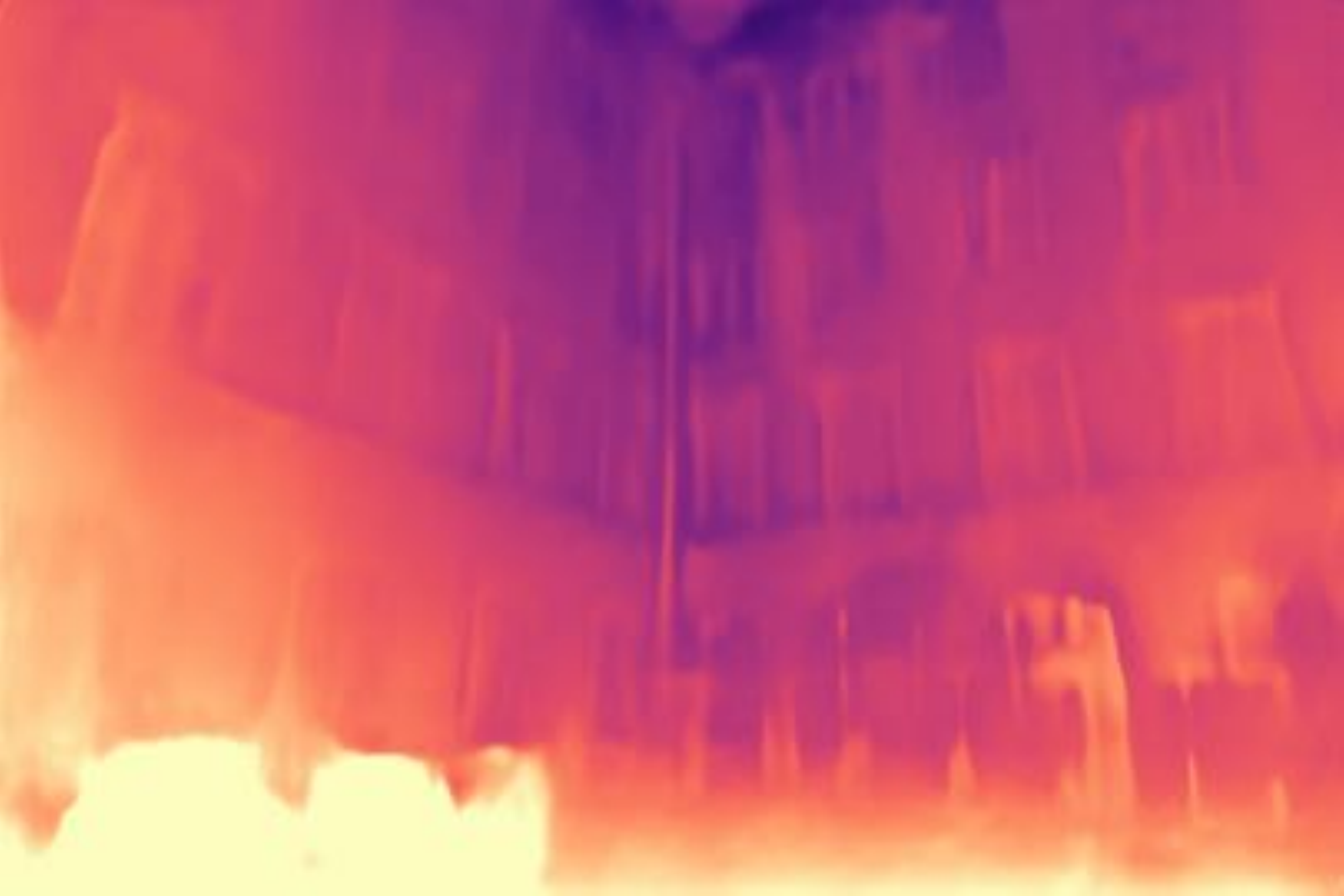}& 
\includegraphics[width=\w,height=\h]{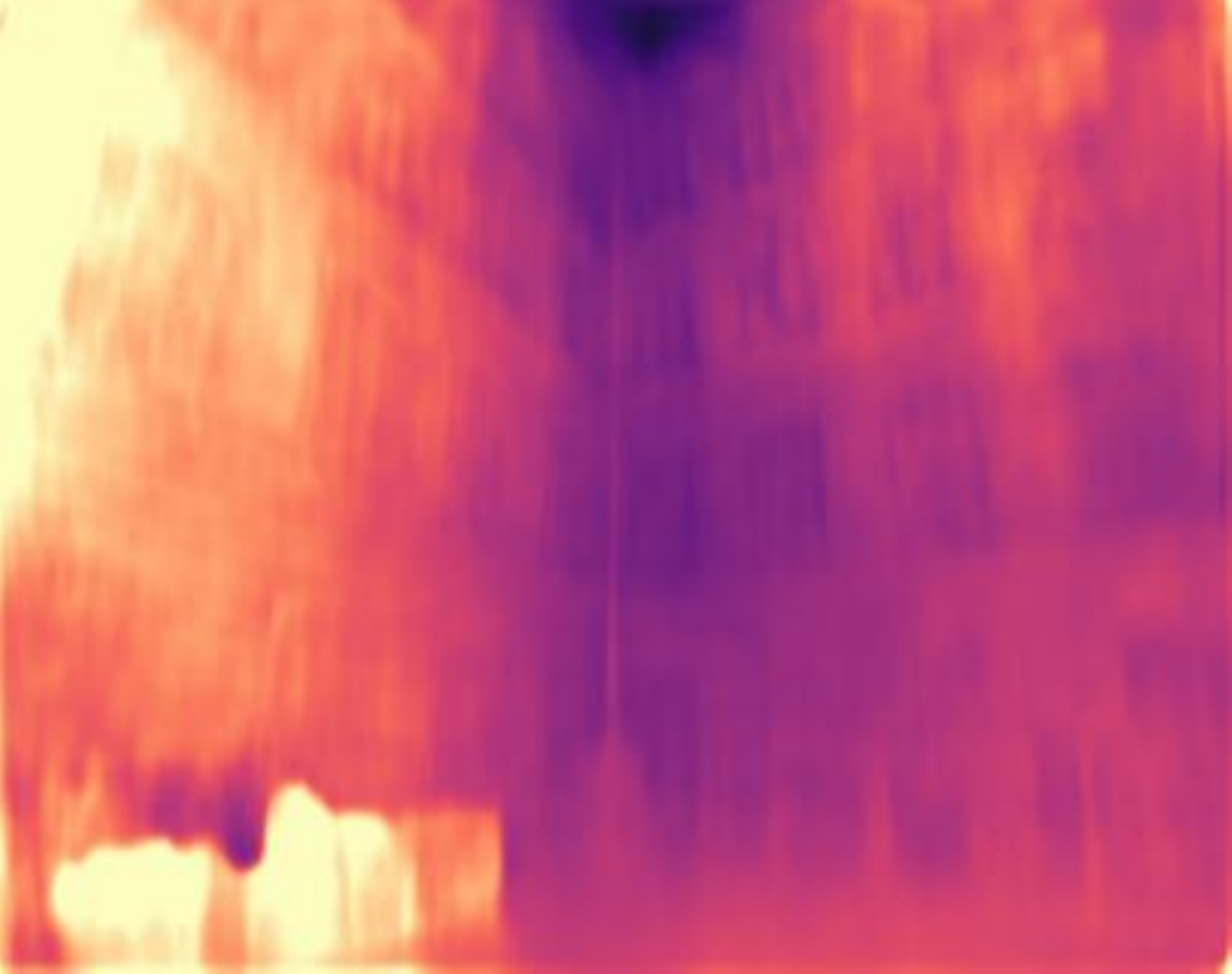}& 
\includegraphics[width=\w,height=\h]{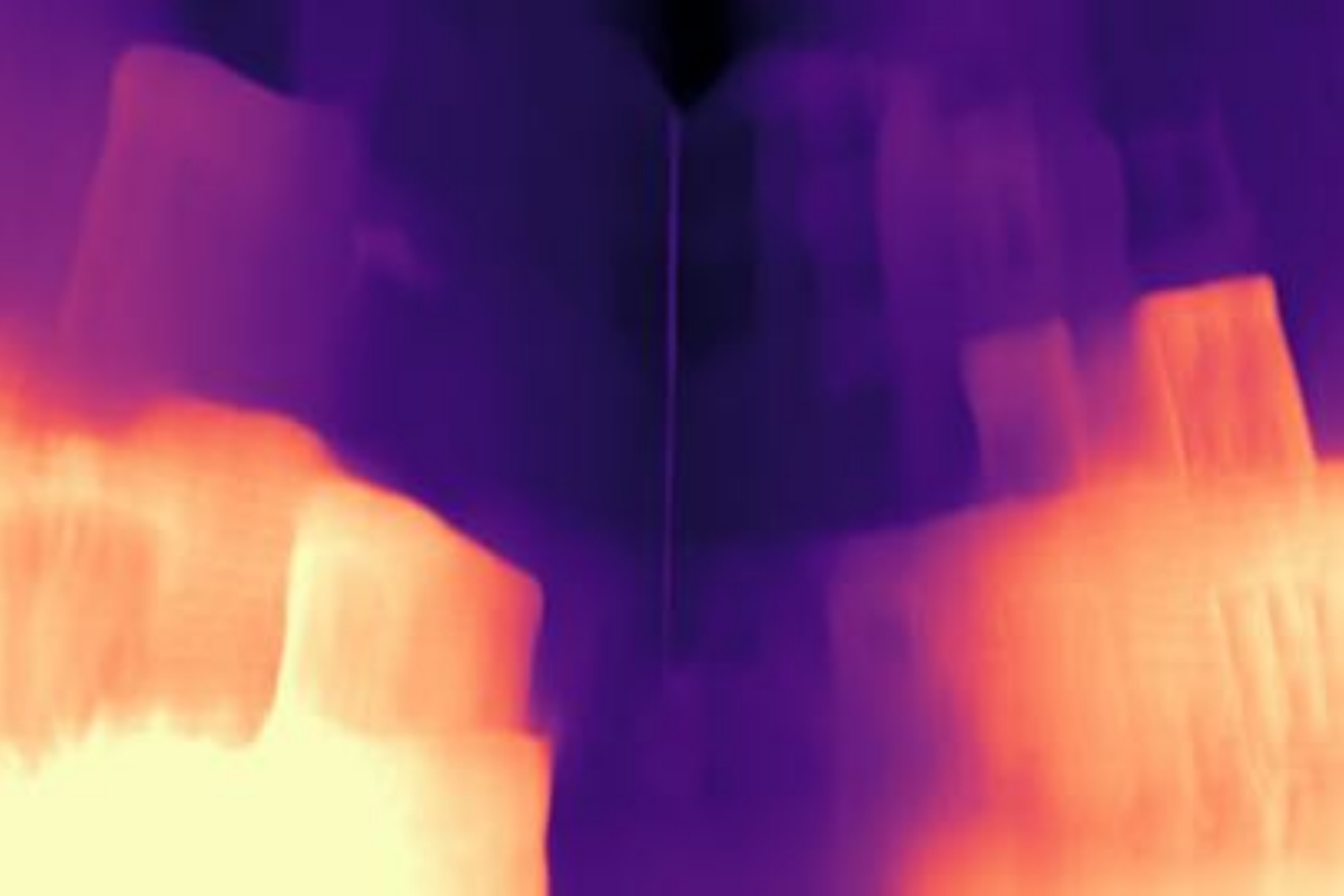}& 
\includegraphics[width=\w,height=\h]{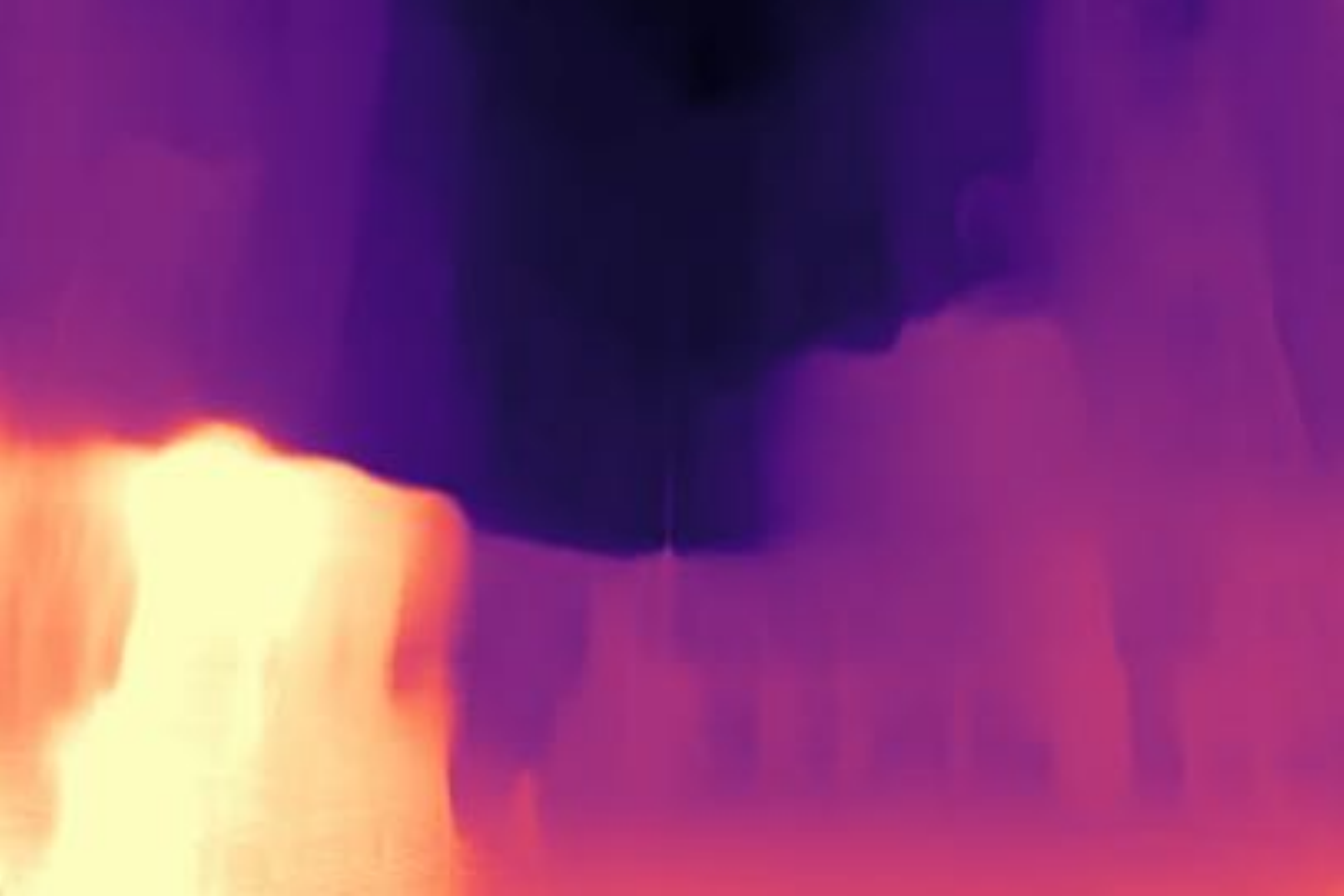}& 
\includegraphics[width=\w,height=\h]{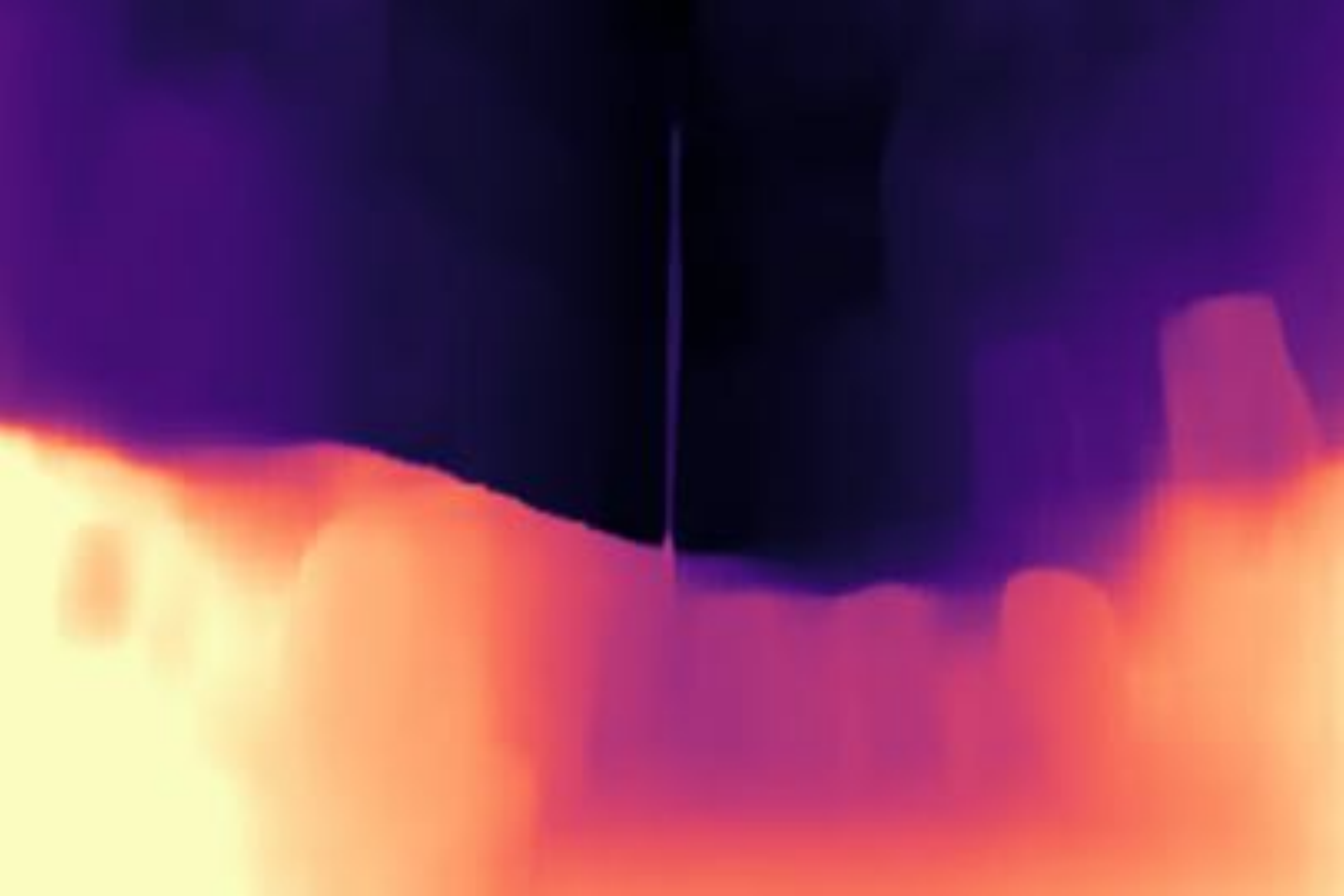}\\ 

\includegraphics[width=\w,height=\h]{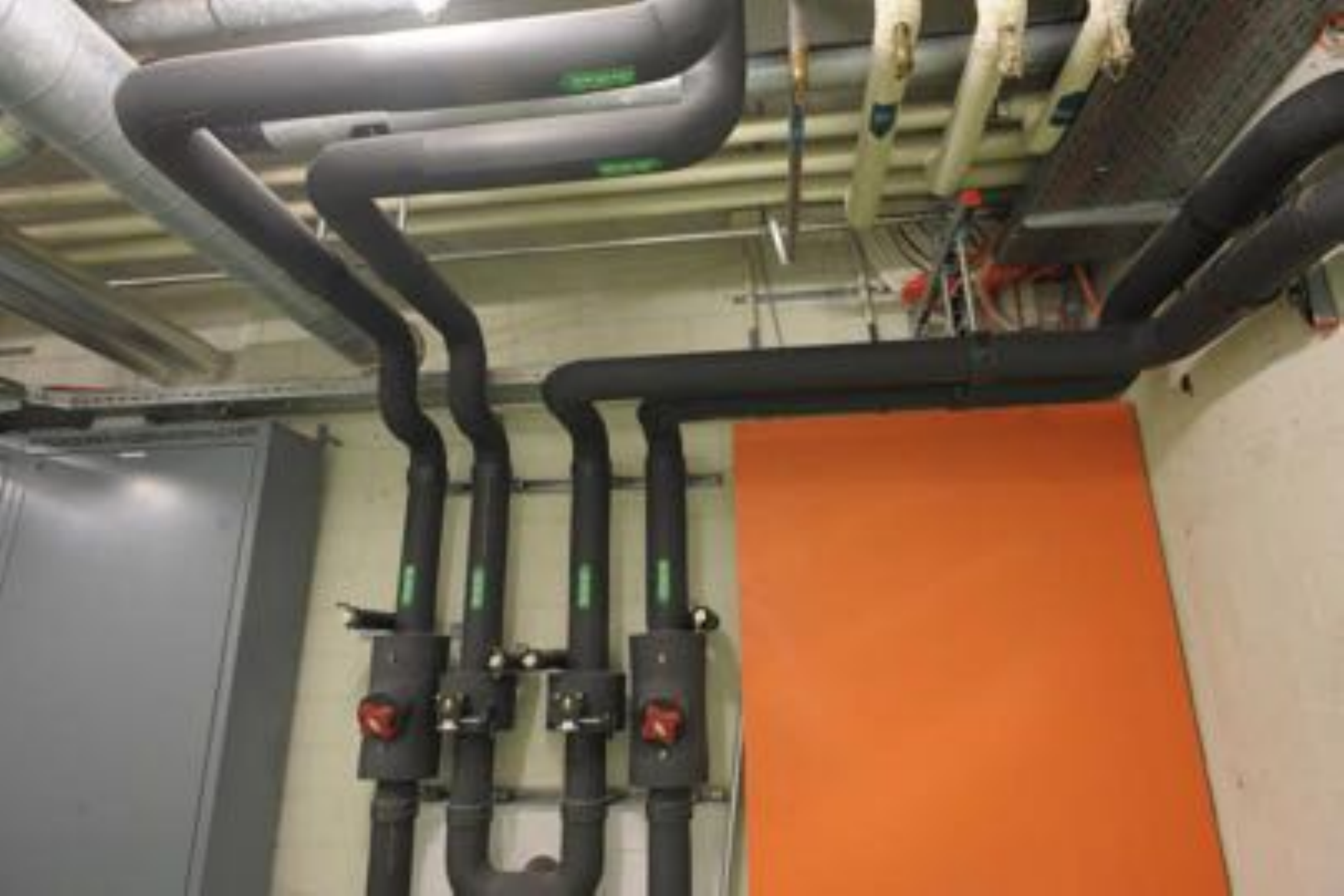}& 
\includegraphics[width=\w,height=\h]{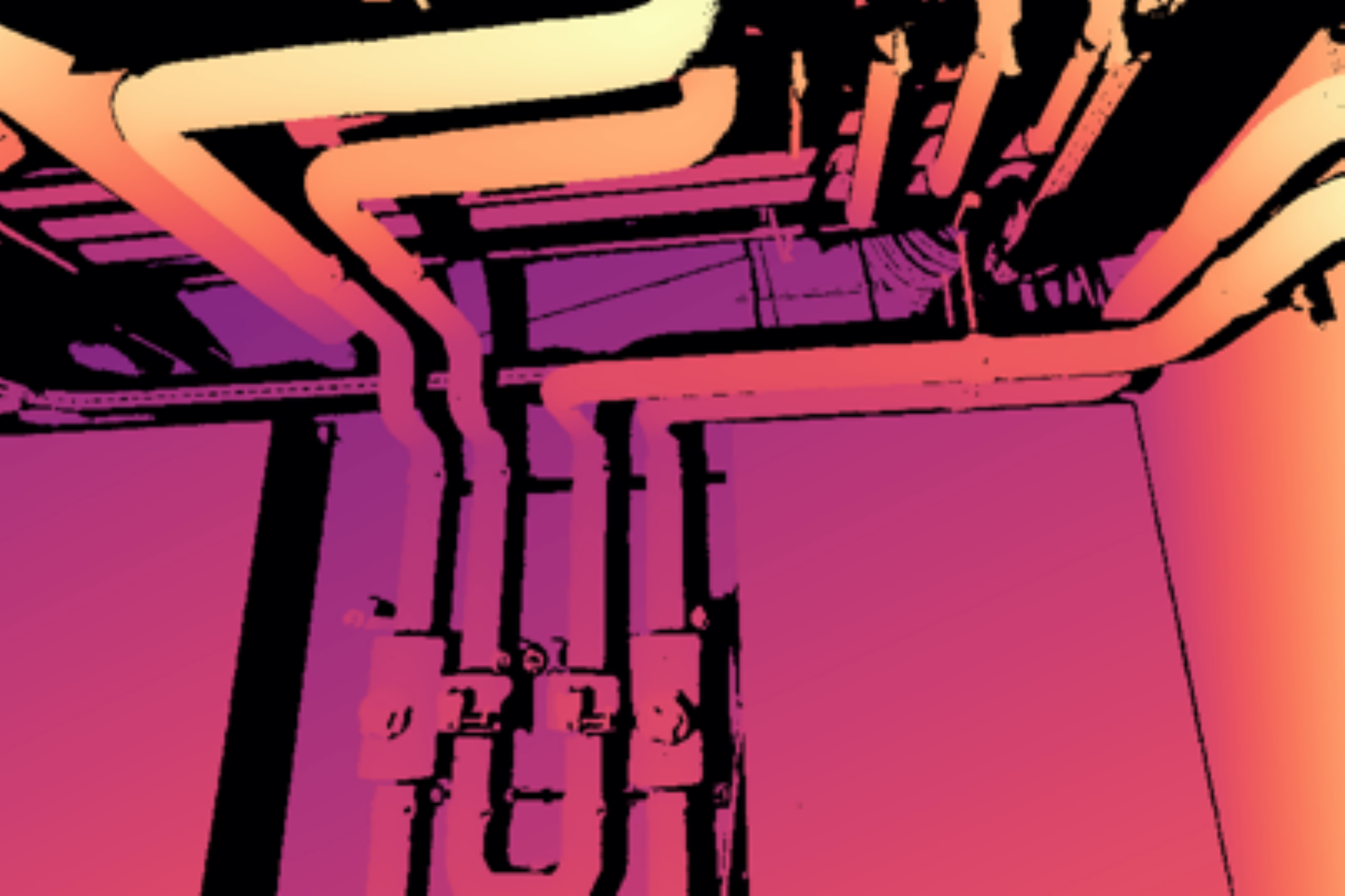}& 
\includegraphics[width=\w,height=\h]{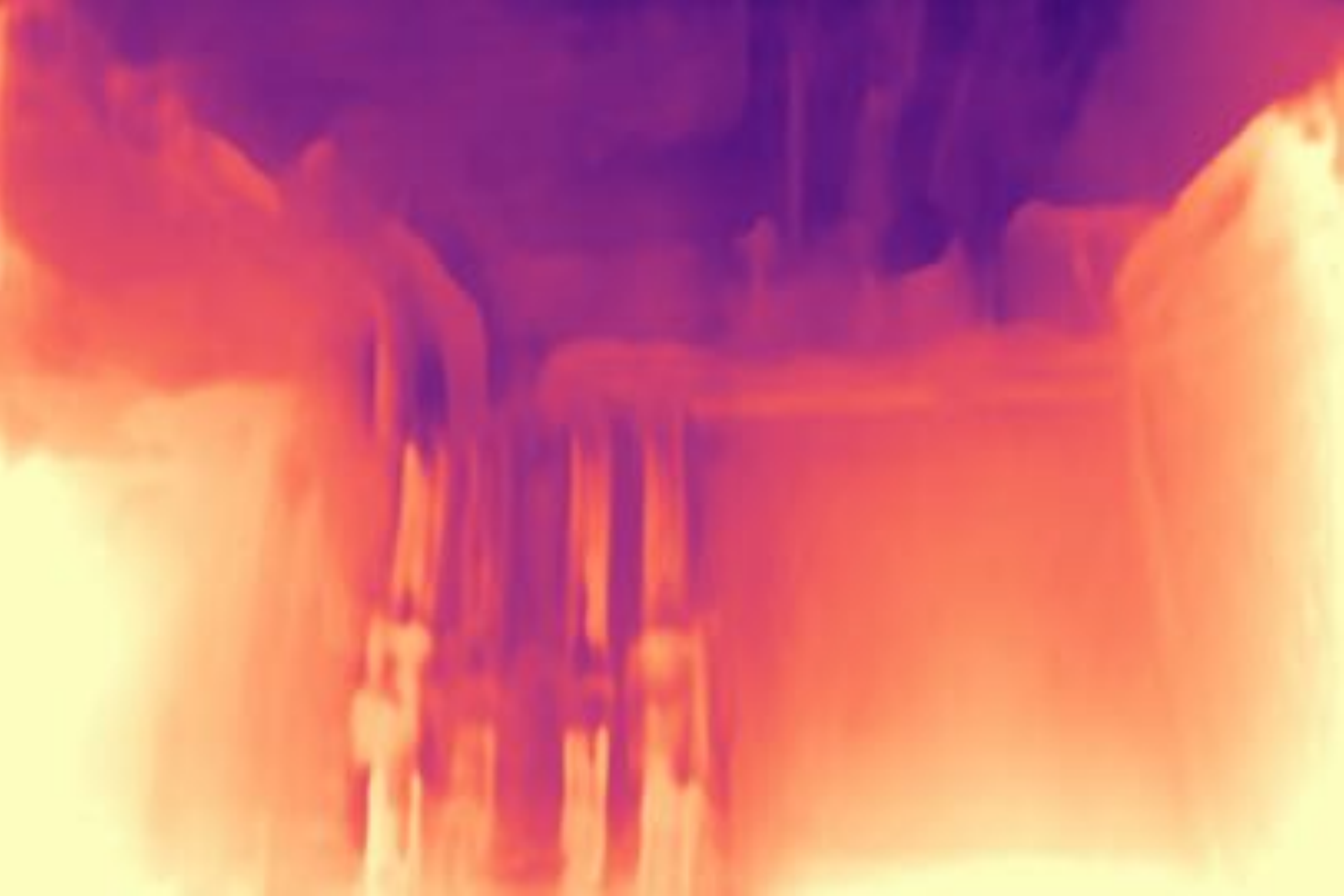}& 
\includegraphics[width=\w,height=\h]{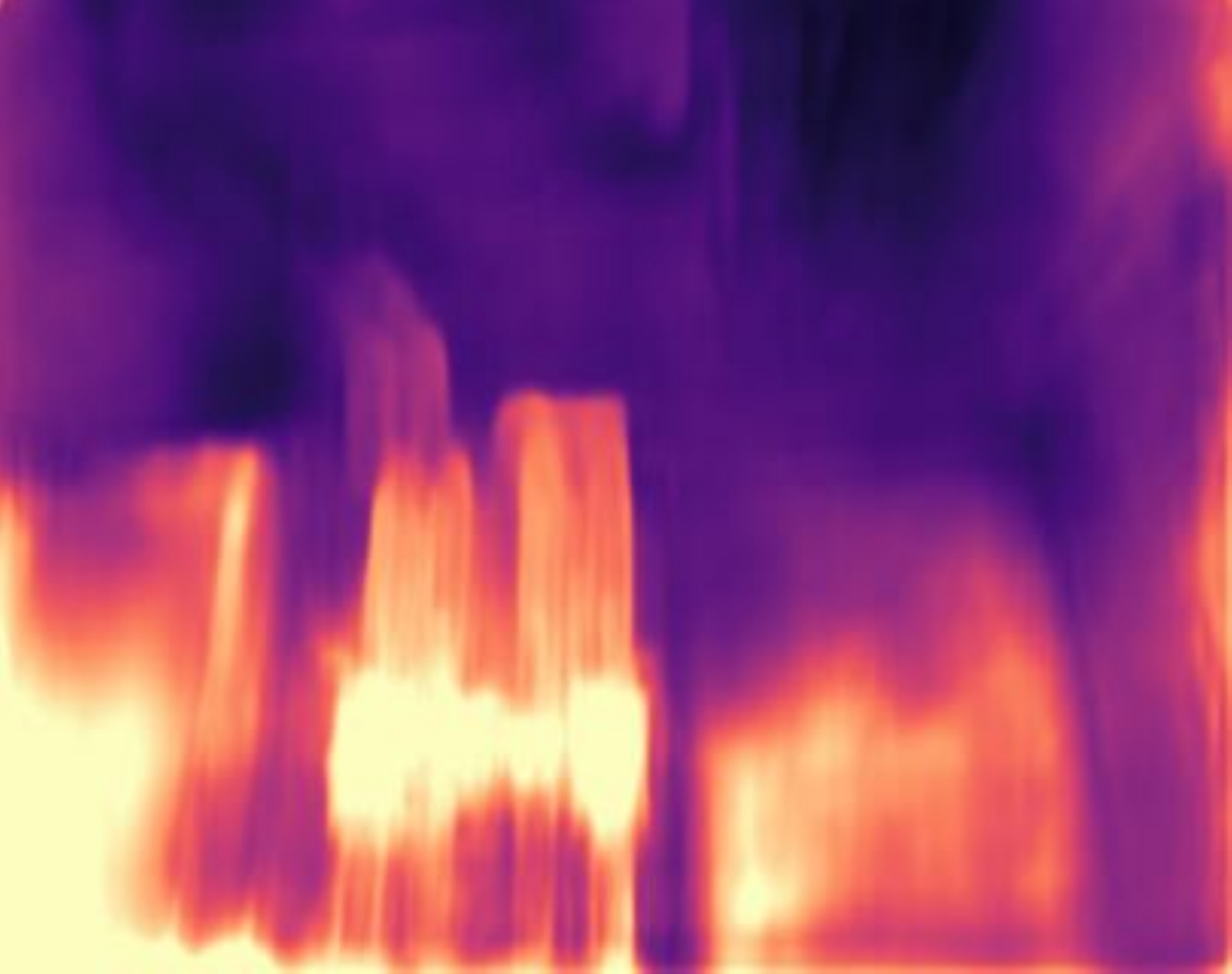}& 
\includegraphics[width=\w,height=\h]{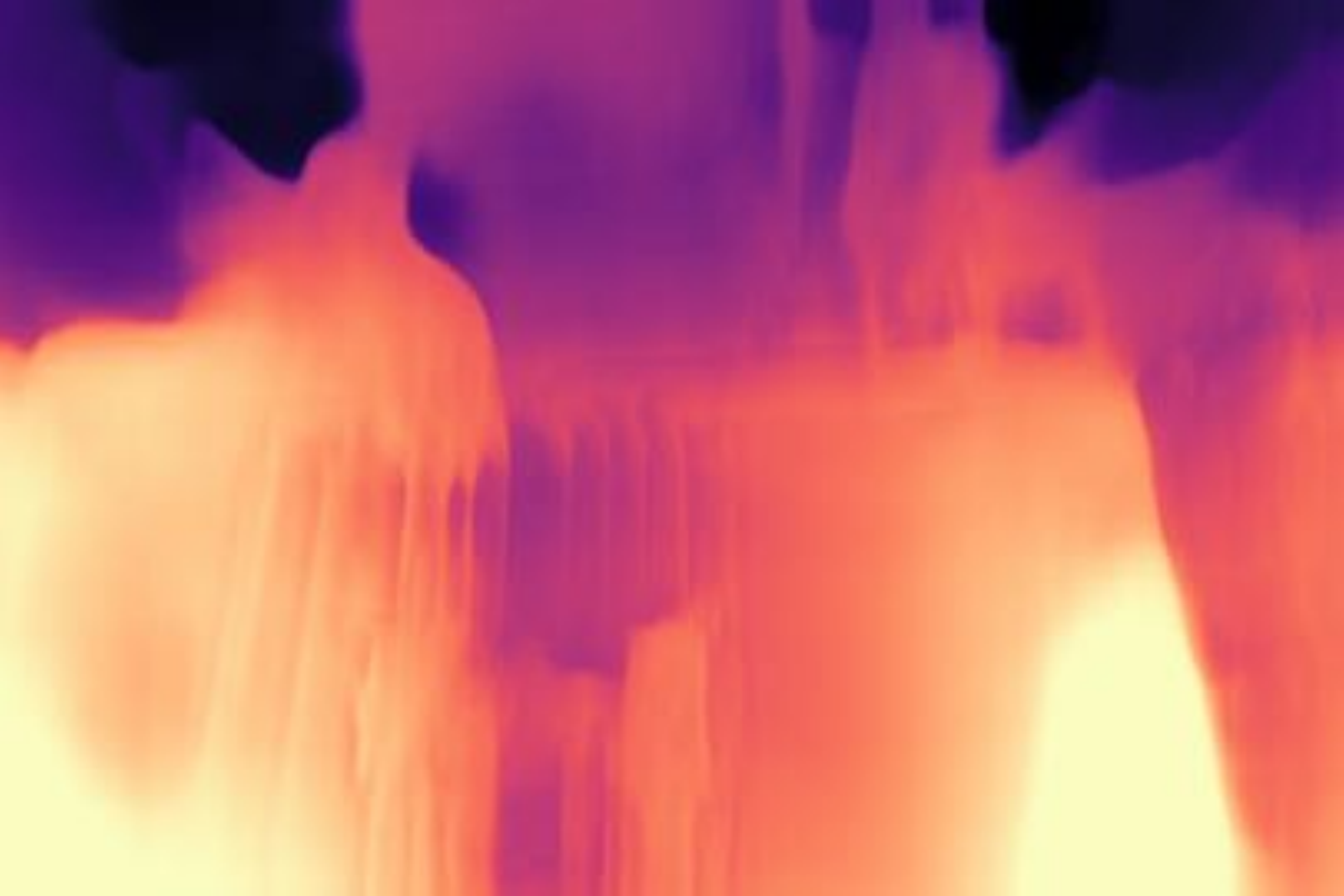}& 
\includegraphics[width=\w,height=\h]{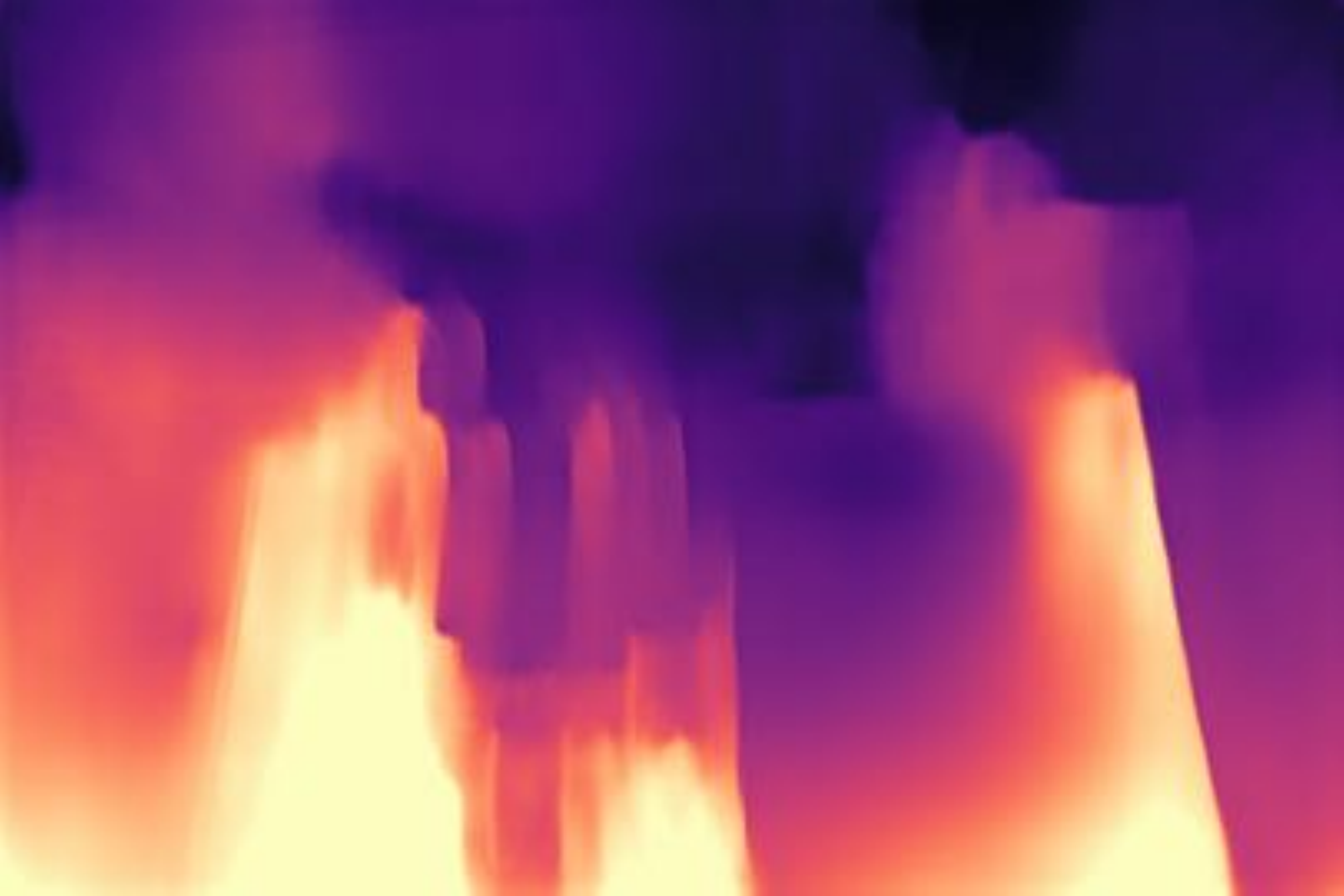}& 
\includegraphics[width=\w,height=\h]{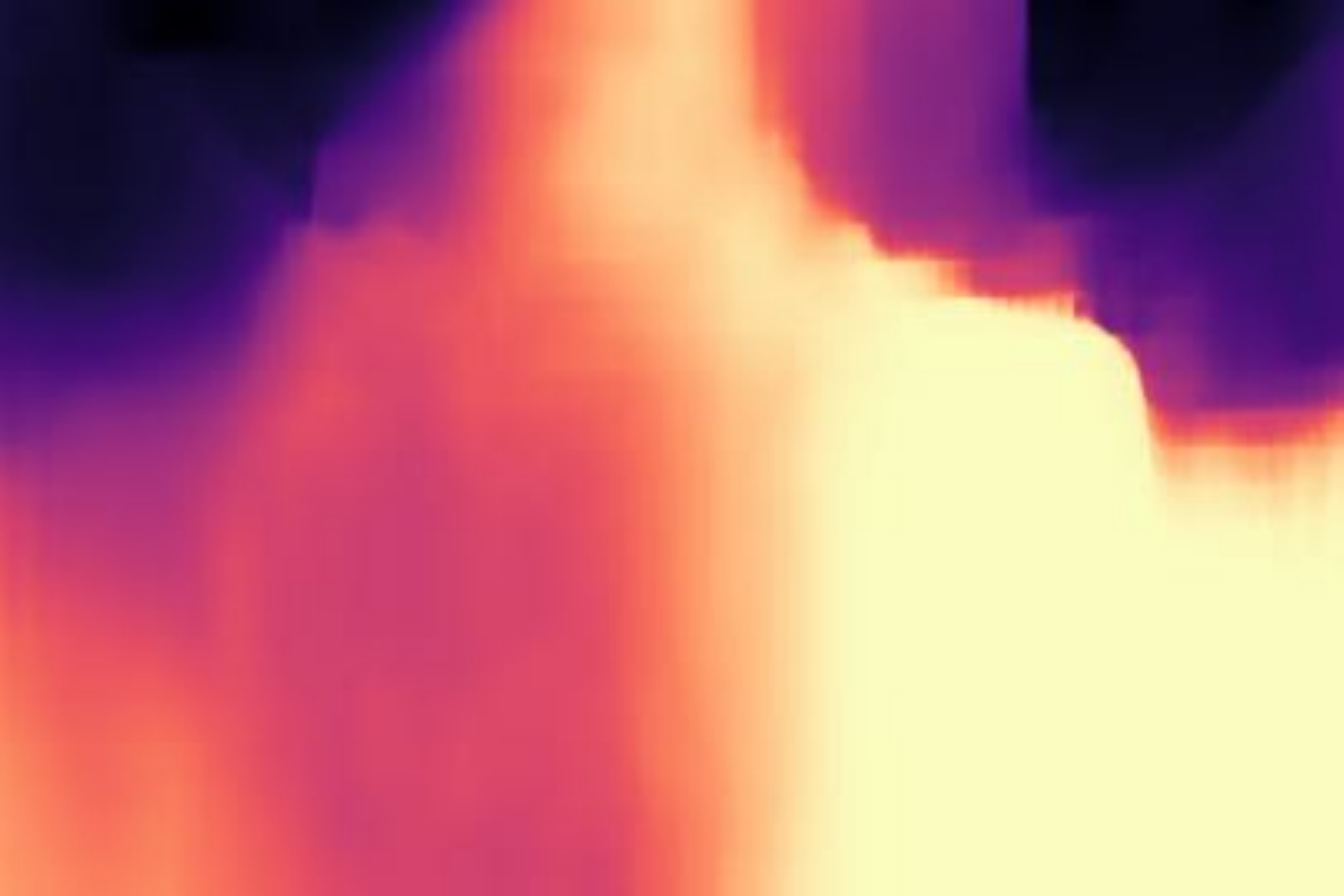}\\ 

\includegraphics[width=\w,height=\h]{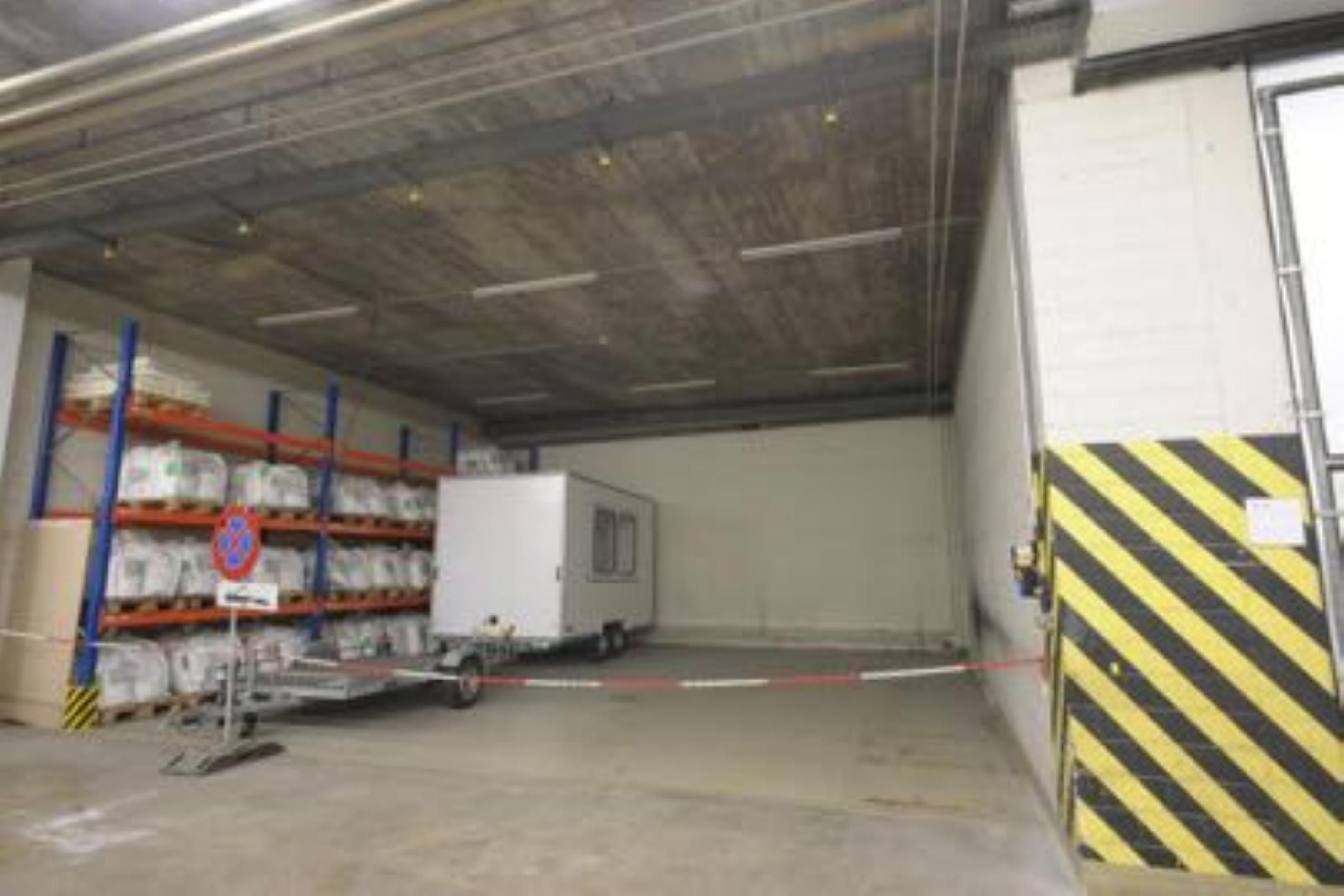}& 
\includegraphics[width=\w,height=\h]{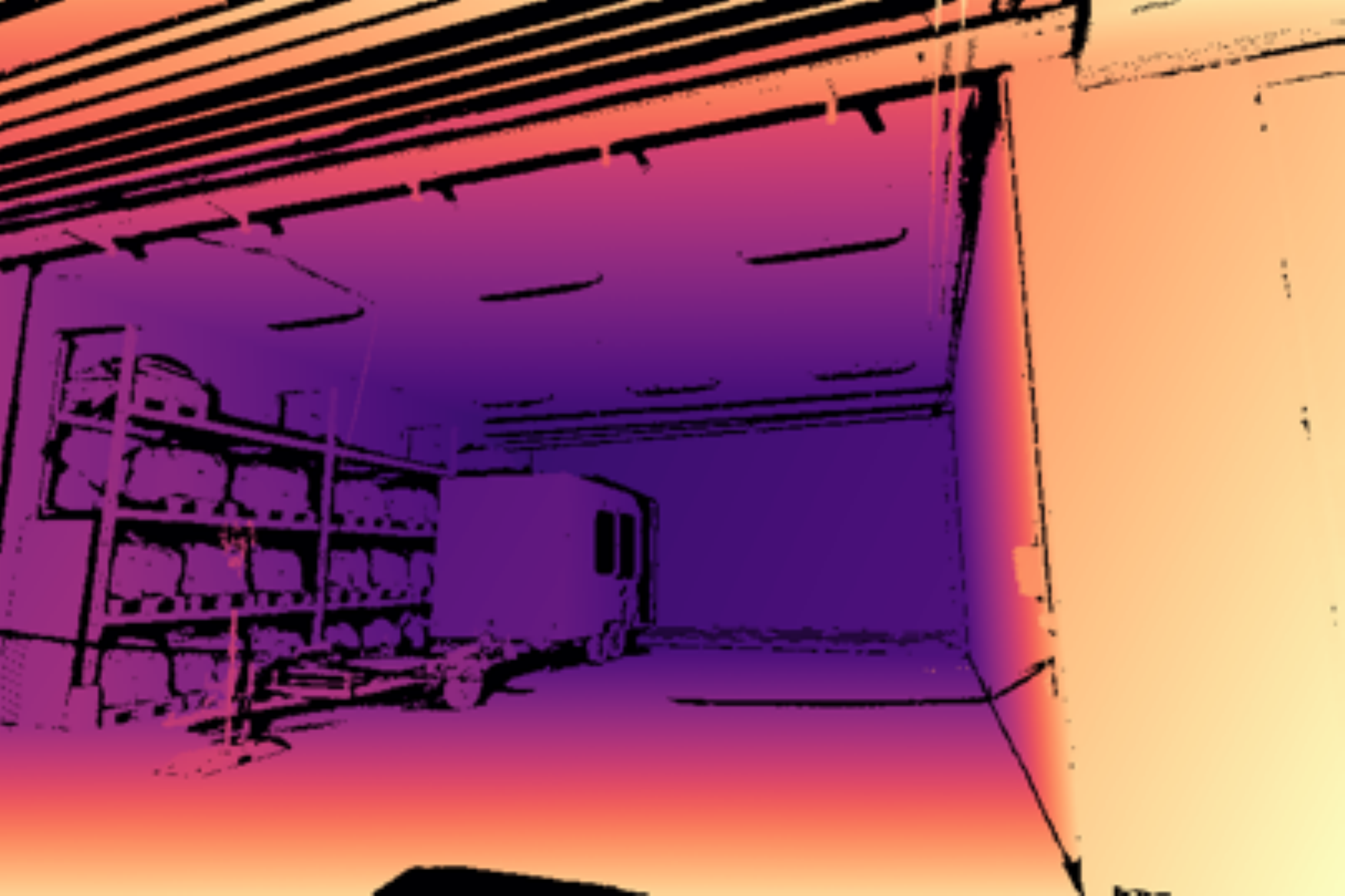}& 
\includegraphics[width=\w,height=\h]{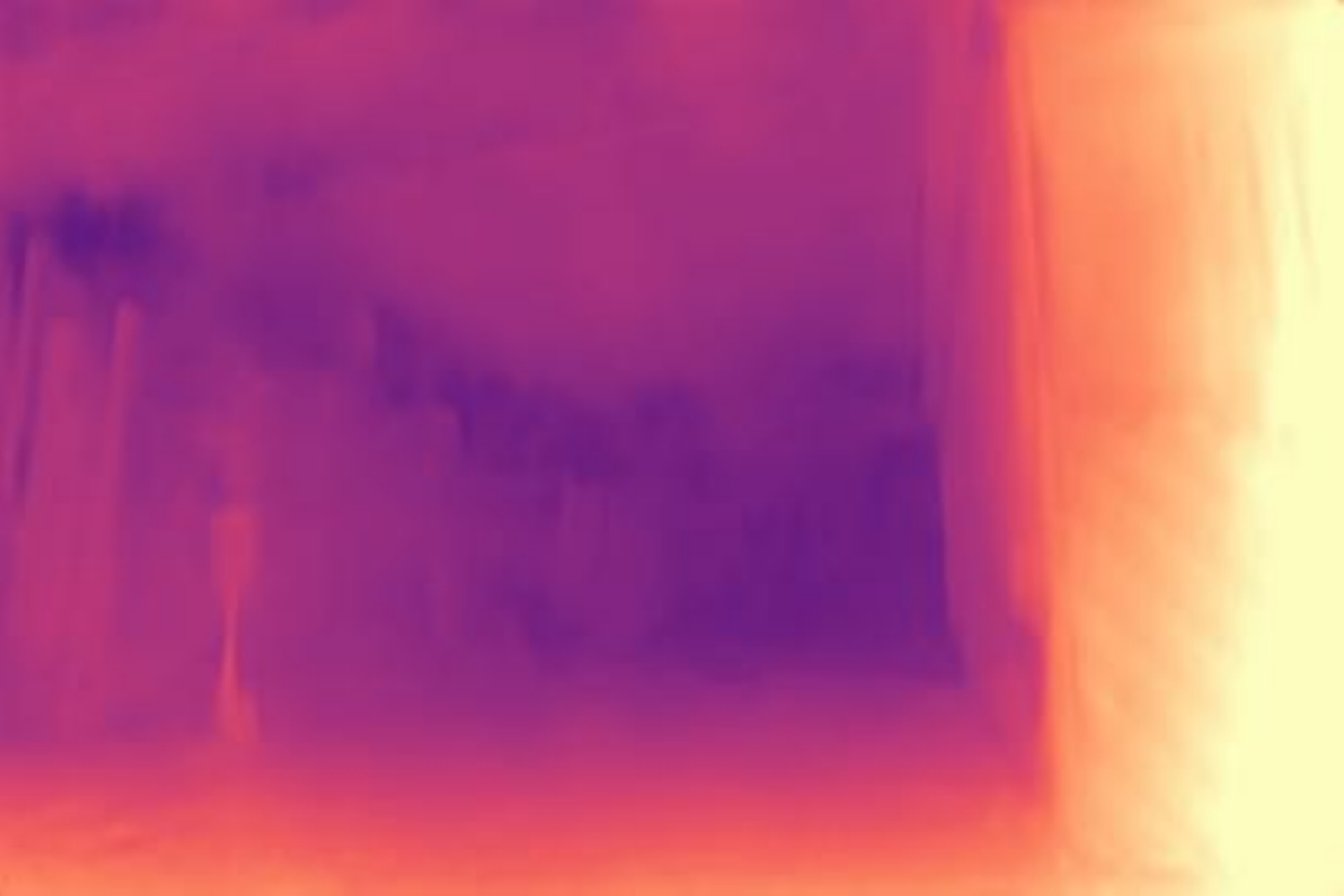}& 
\includegraphics[width=\w,height=\h]{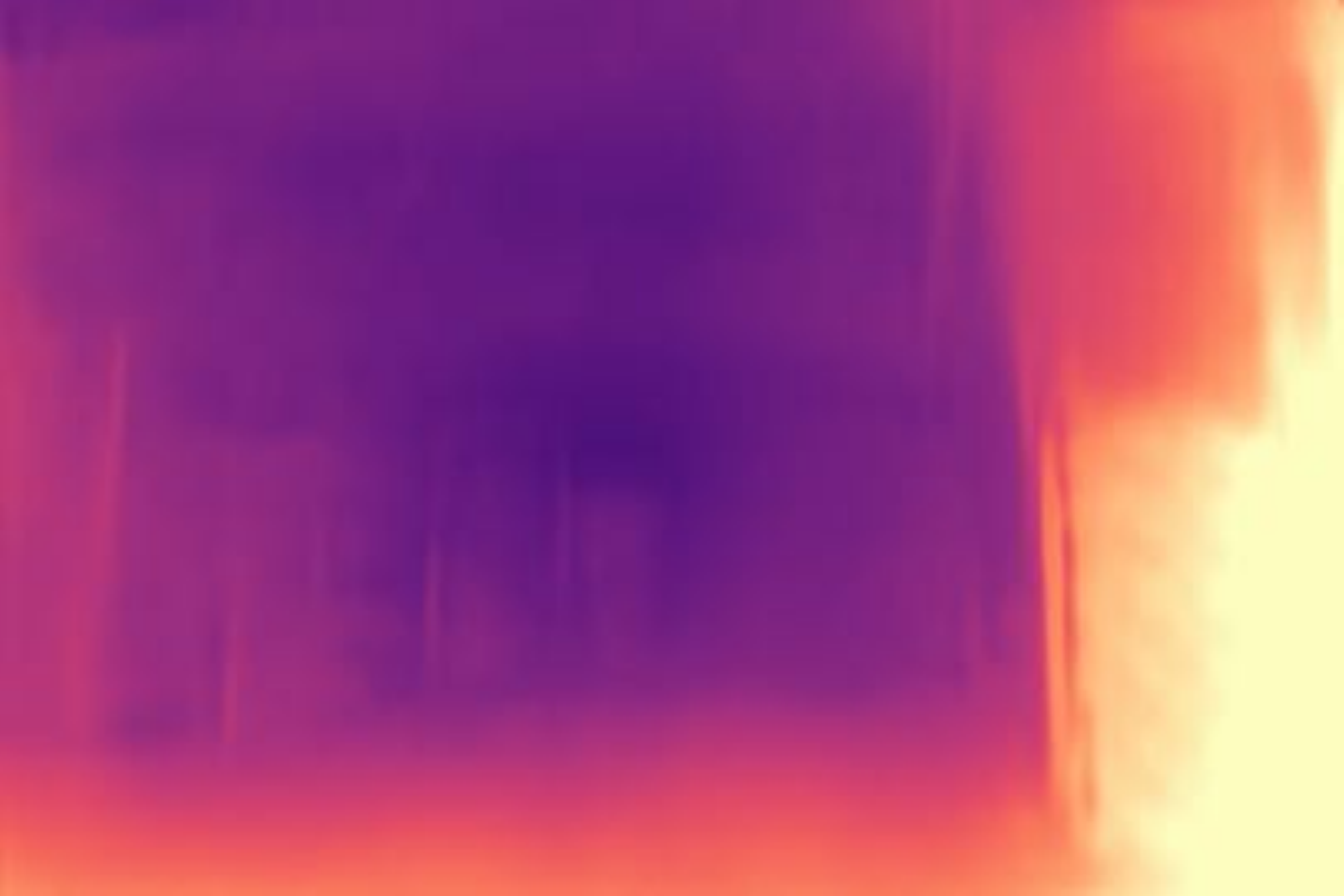}& 
\includegraphics[width=\w,height=\h]{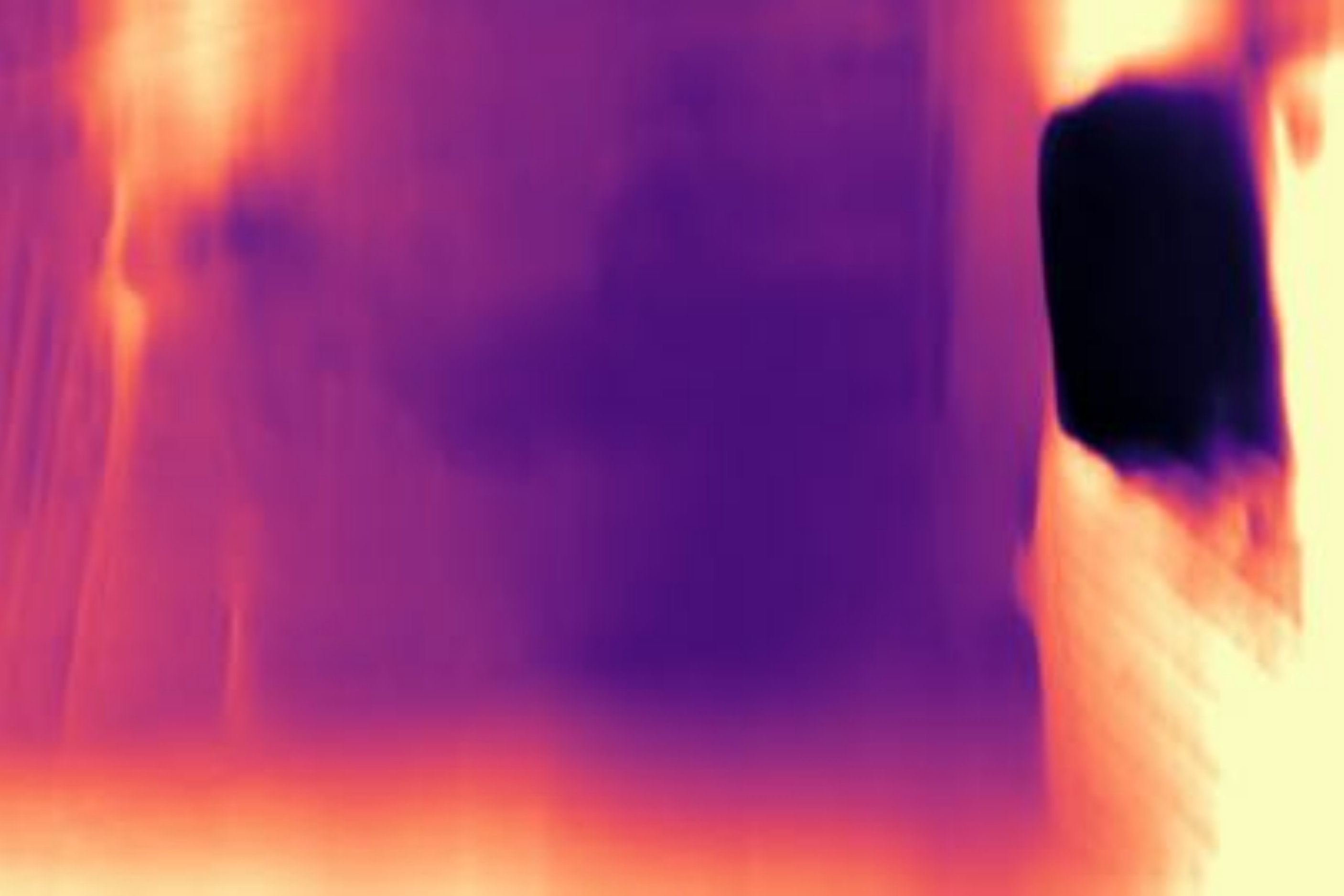}& 
\includegraphics[width=\w,height=\h]{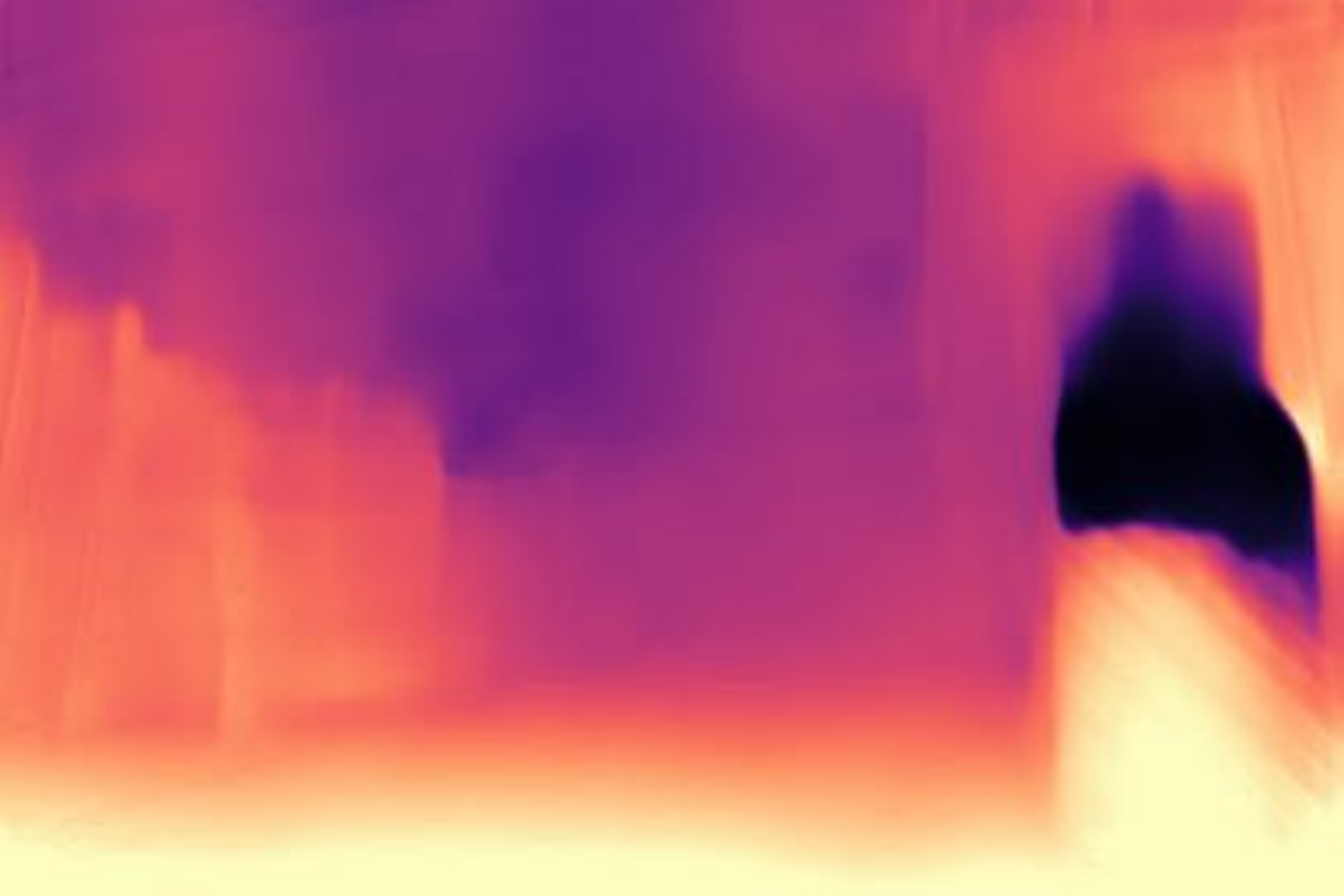}& 
\includegraphics[width=\w,height=\h]{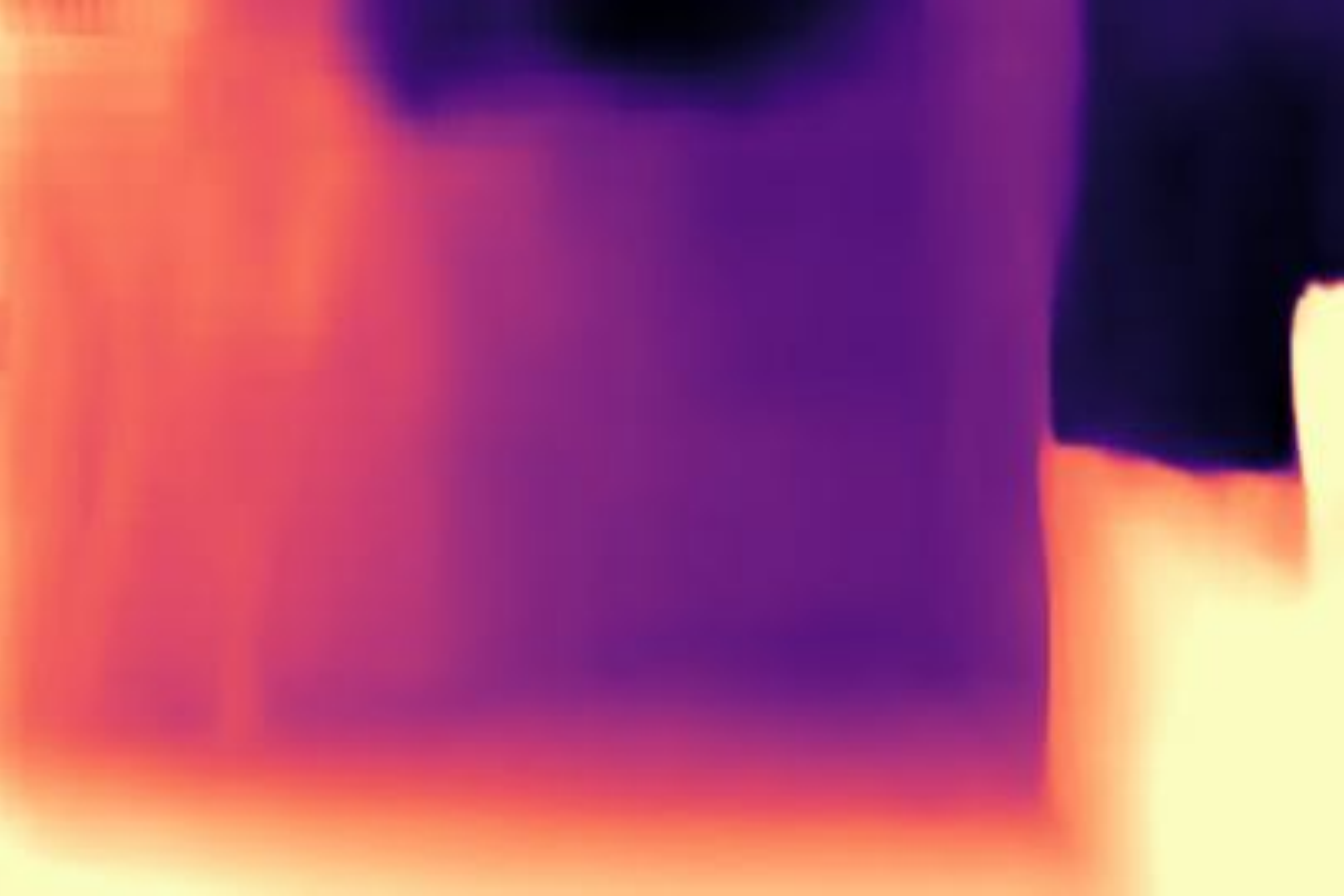}\\ 

\includegraphics[width=\w,height=\h]{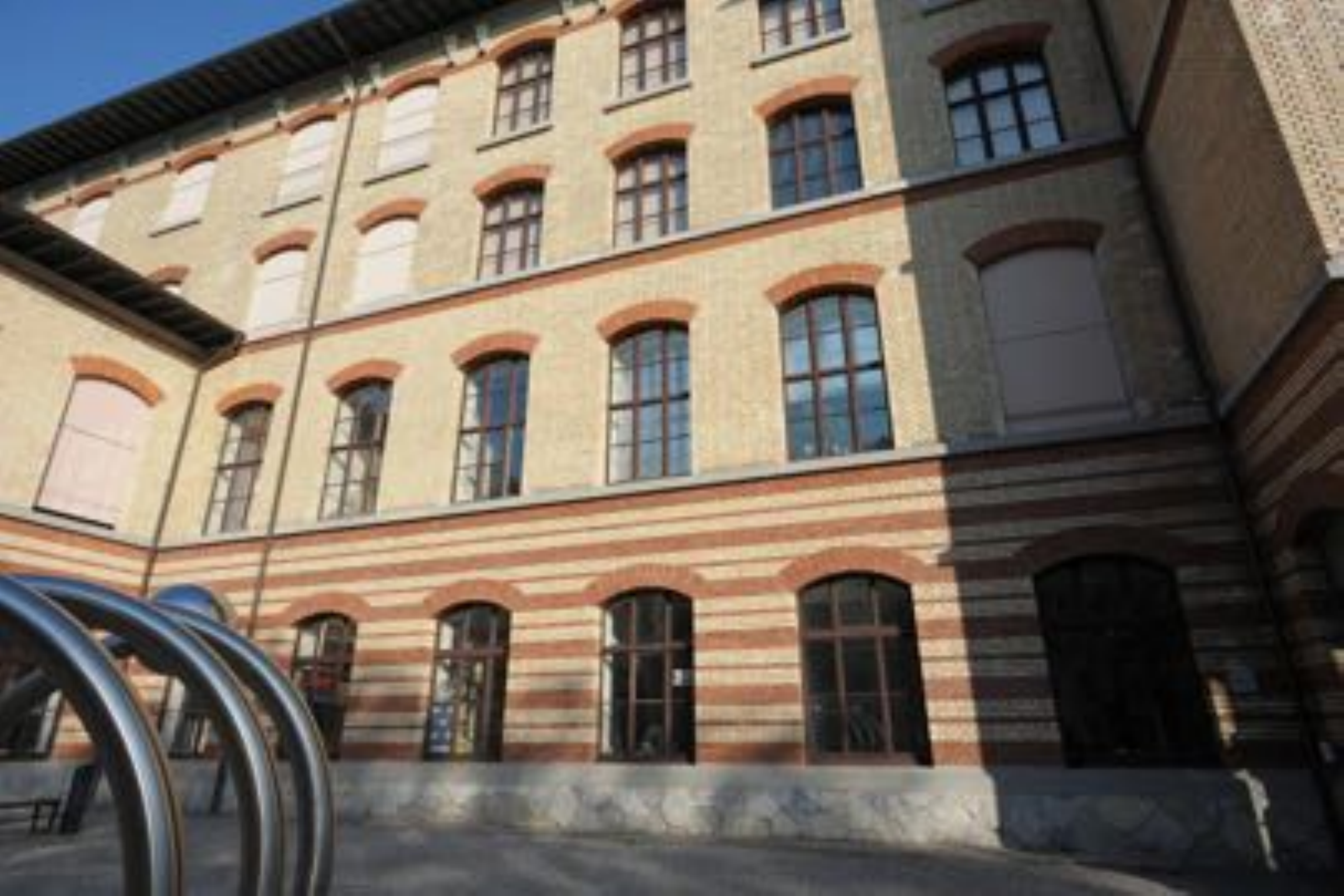}& 
\includegraphics[width=\w,height=\h]{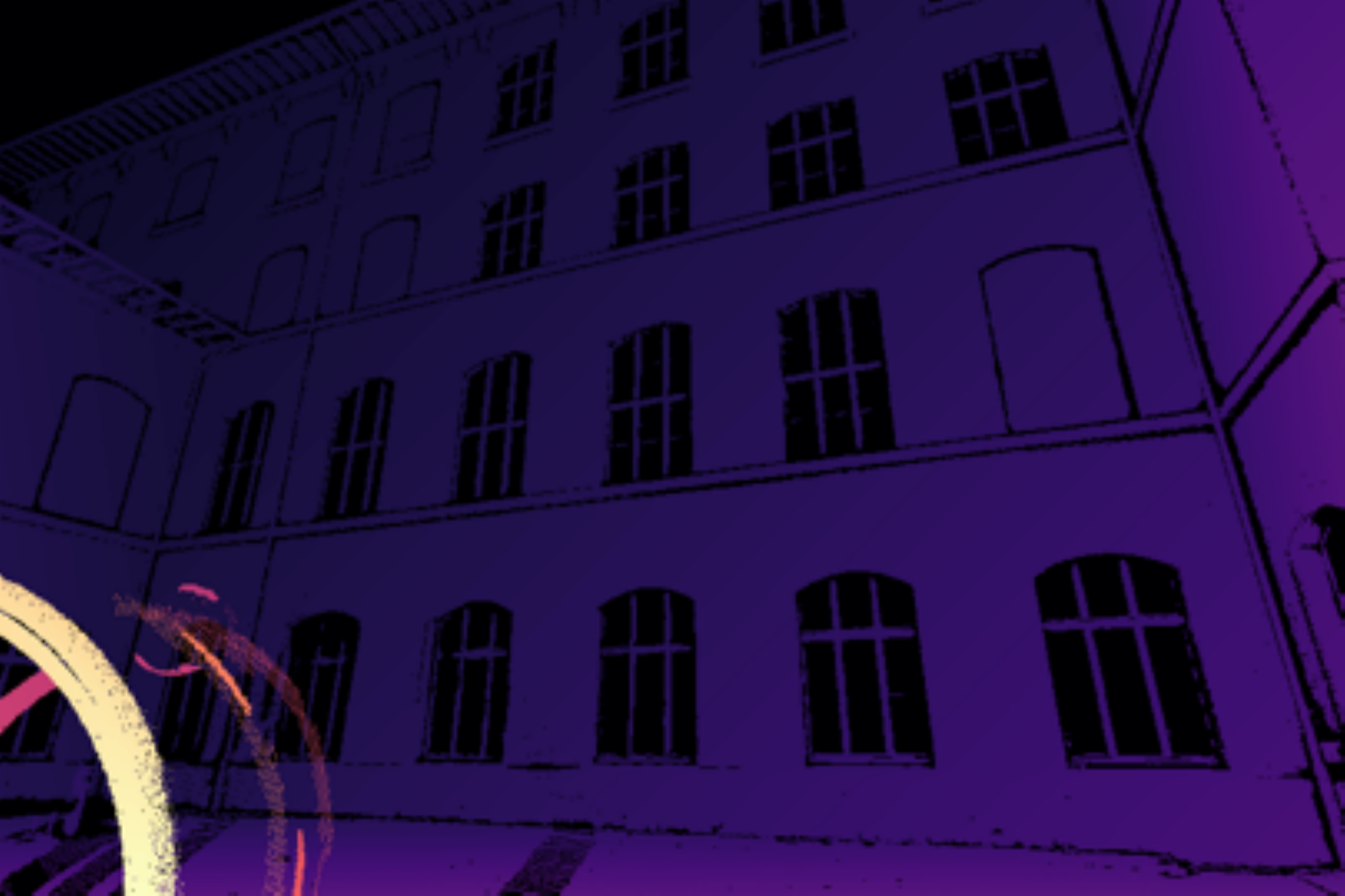}& 
\includegraphics[width=\w,height=\h]{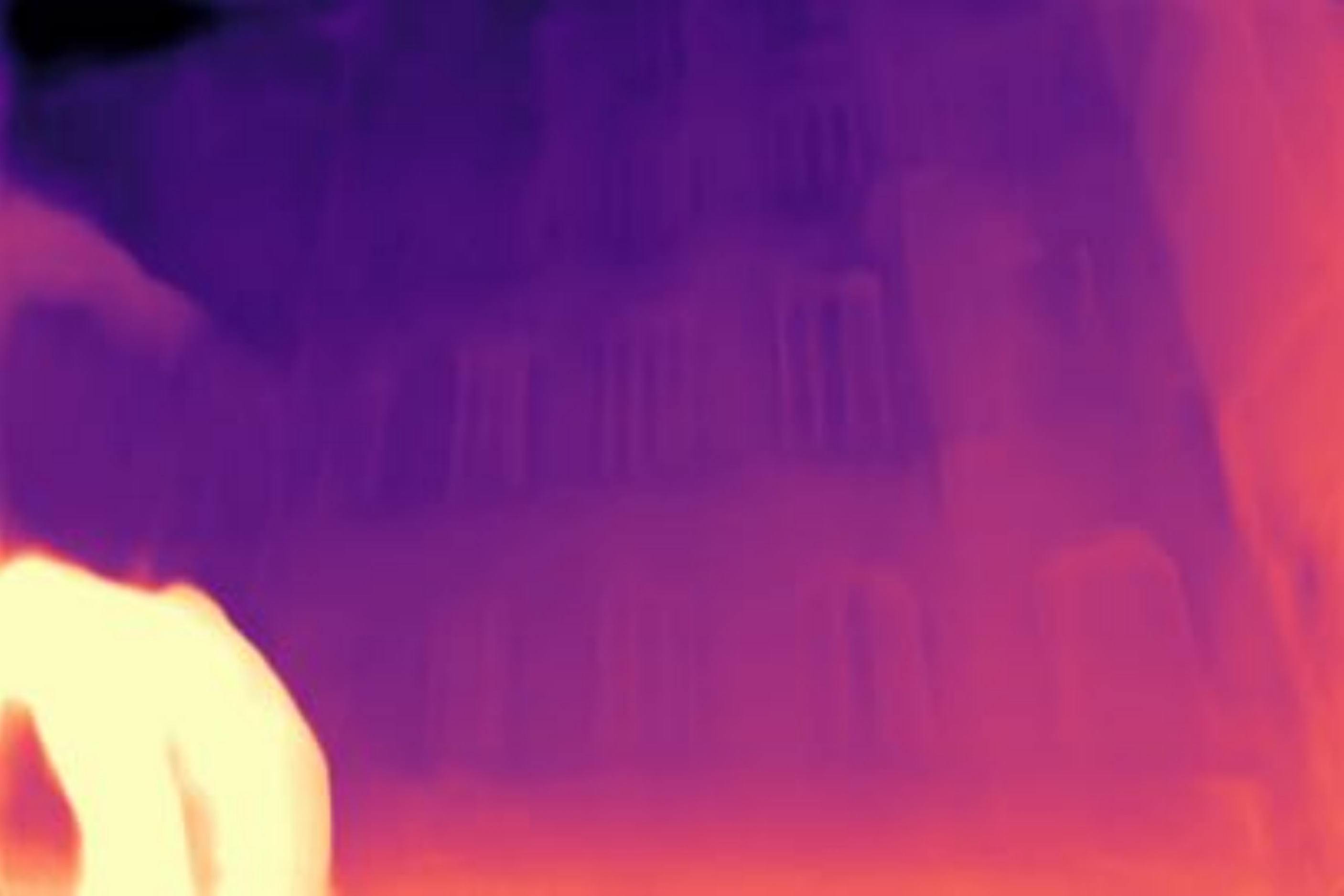}& 
\includegraphics[width=\w,height=\h]{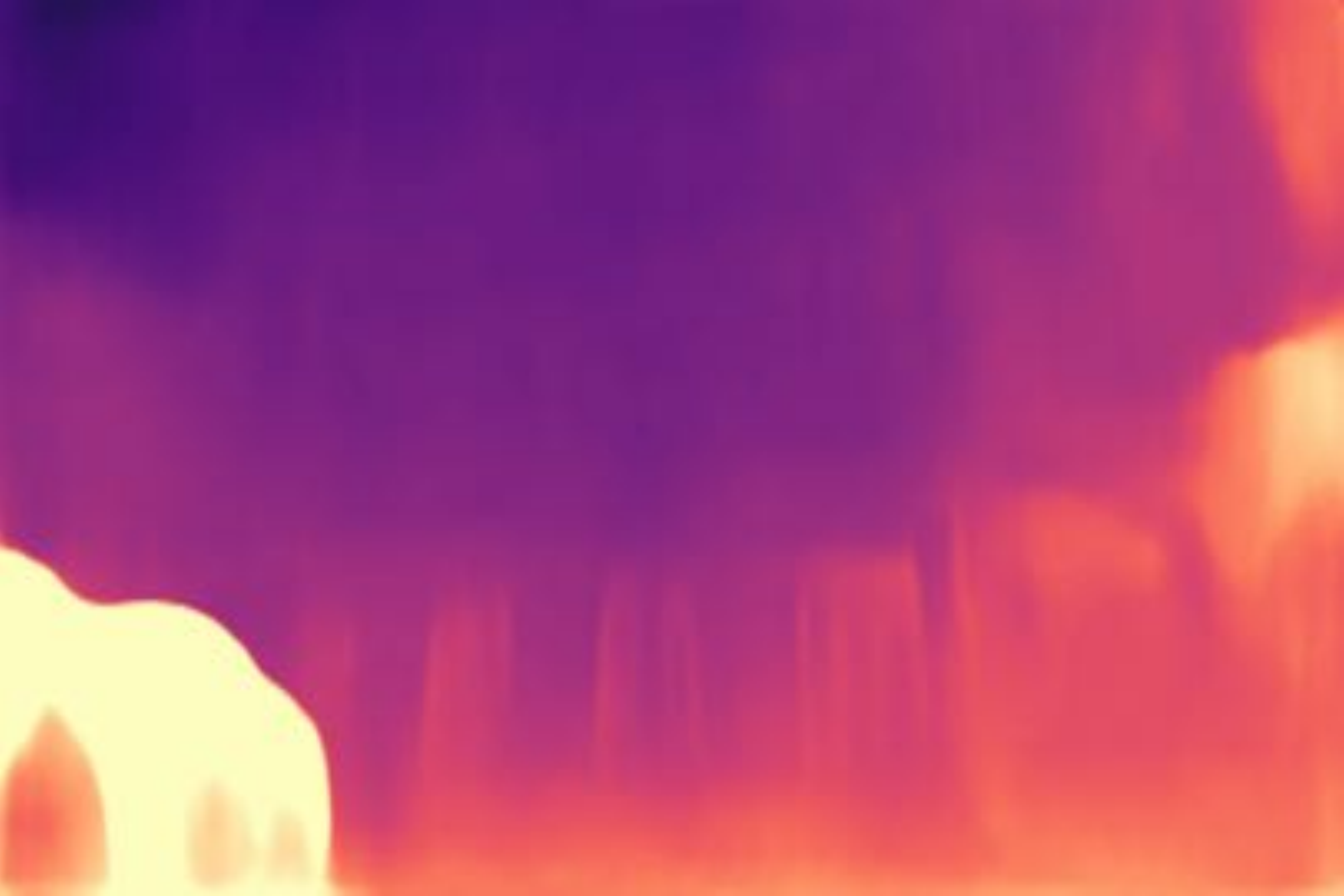}&
\includegraphics[width=\w,height=\h]{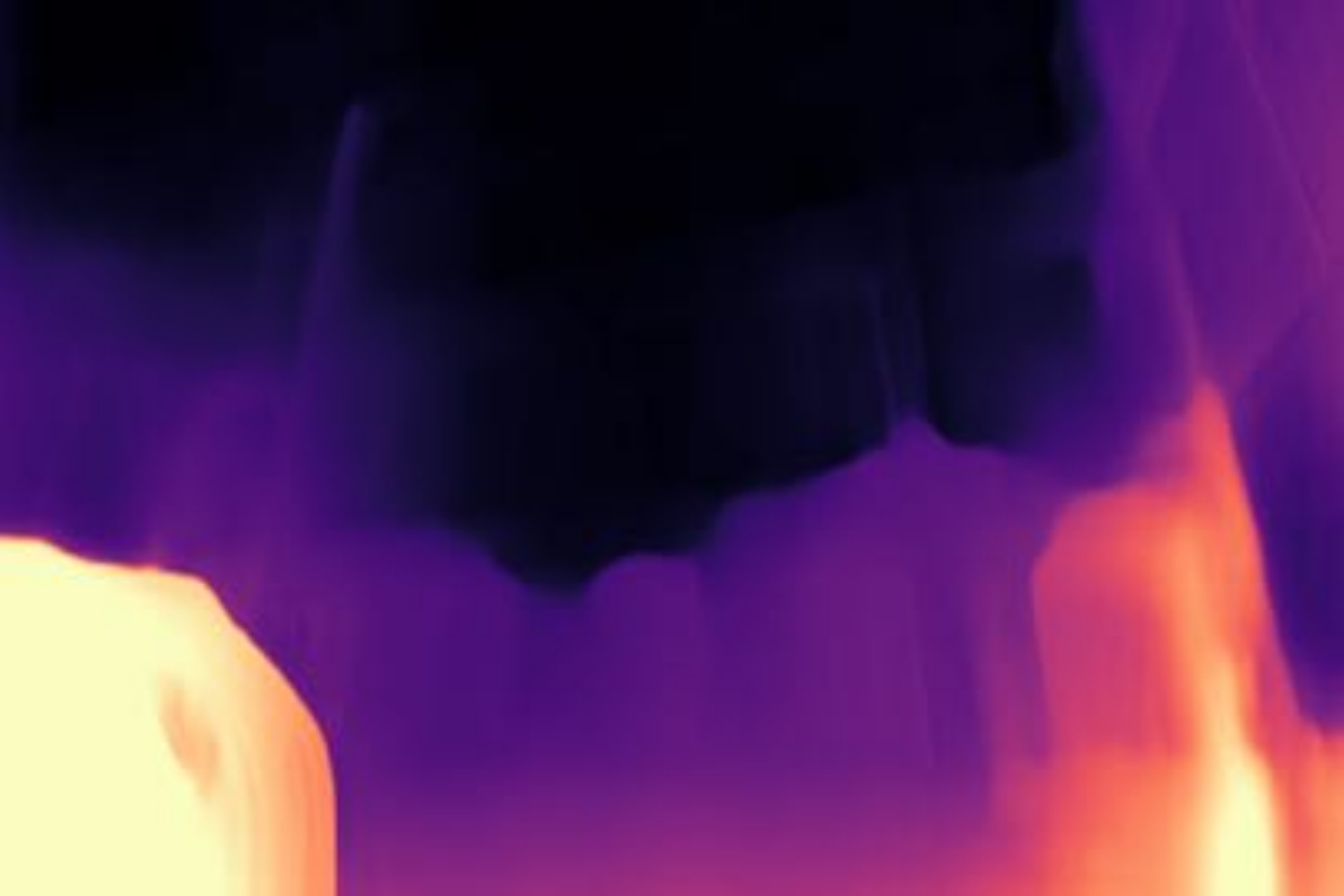}& 
\includegraphics[width=\w,height=\h]{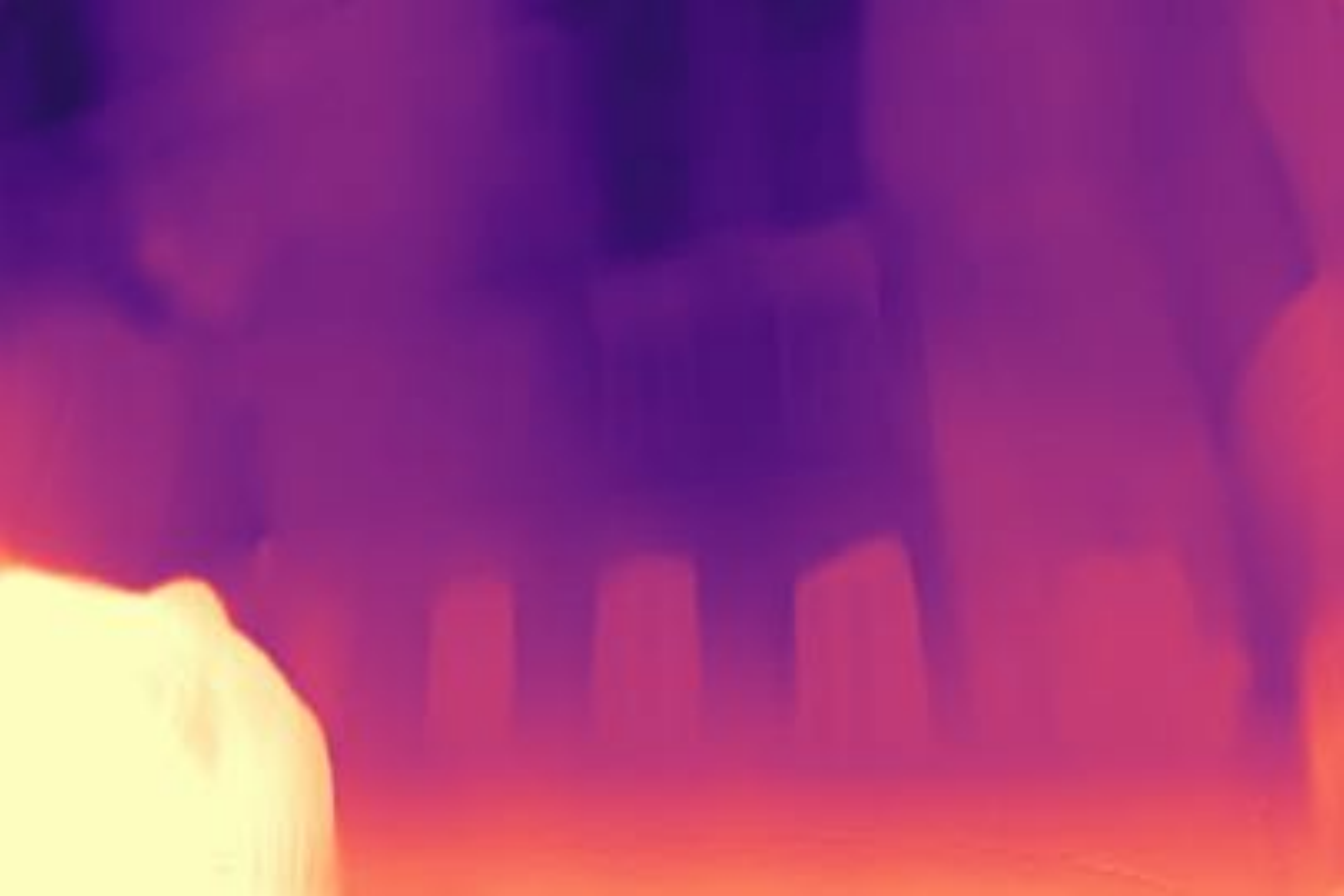}& 
\includegraphics[width=\w,height=\h]{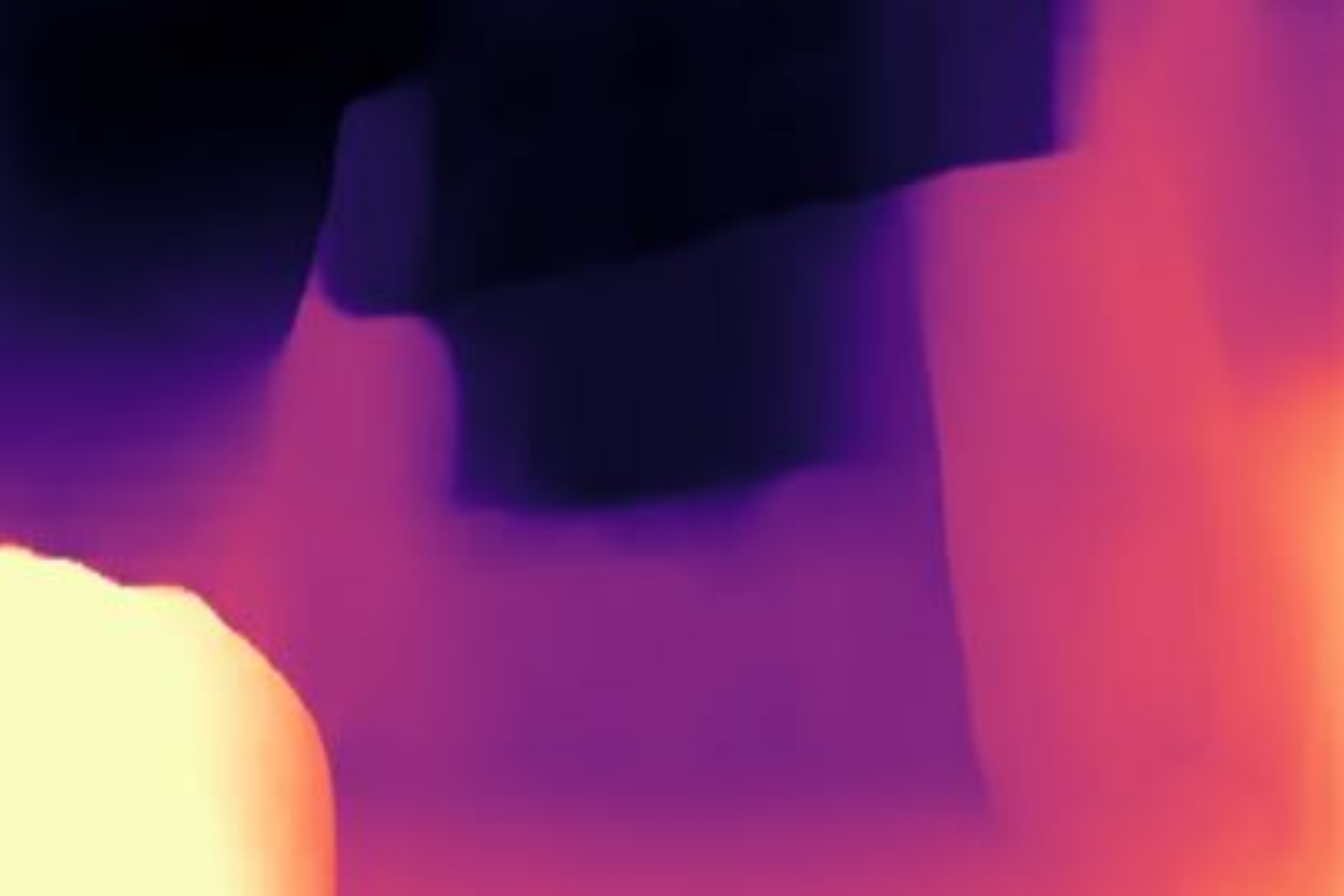}\\

\includegraphics[width=\w,height=\h]{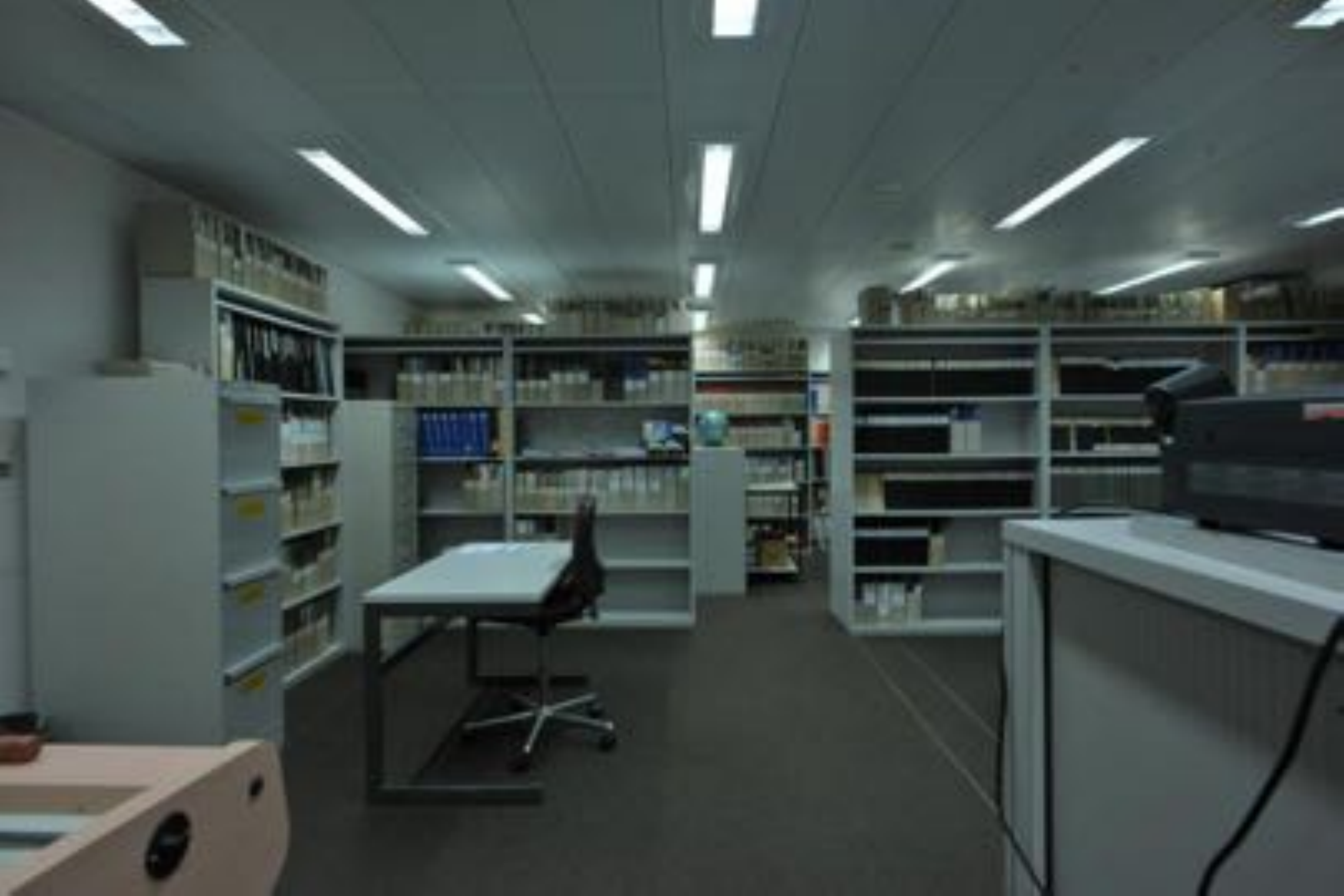}& 
\includegraphics[width=\w,height=\h]{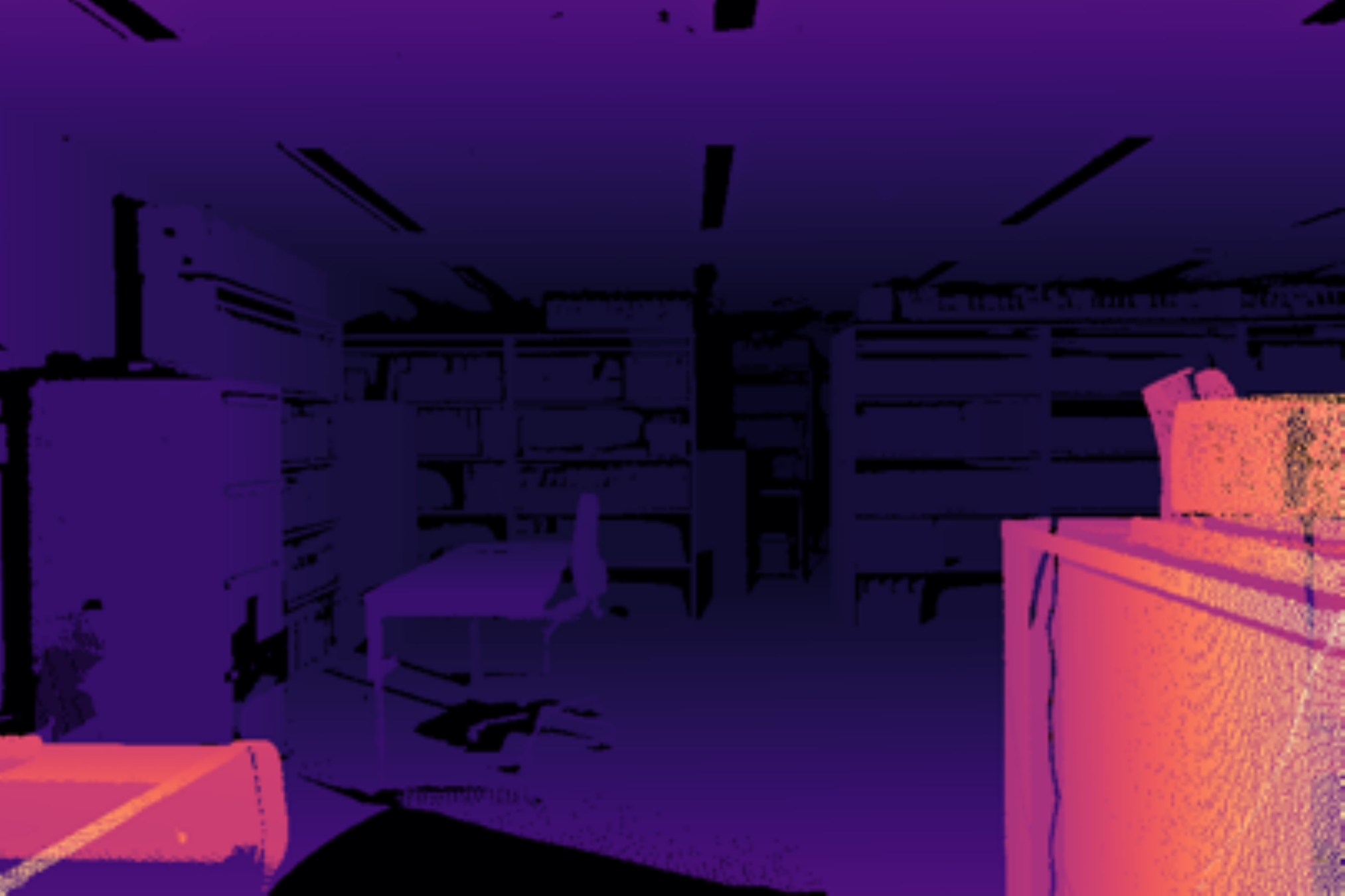}& 
\includegraphics[width=\w,height=\h]{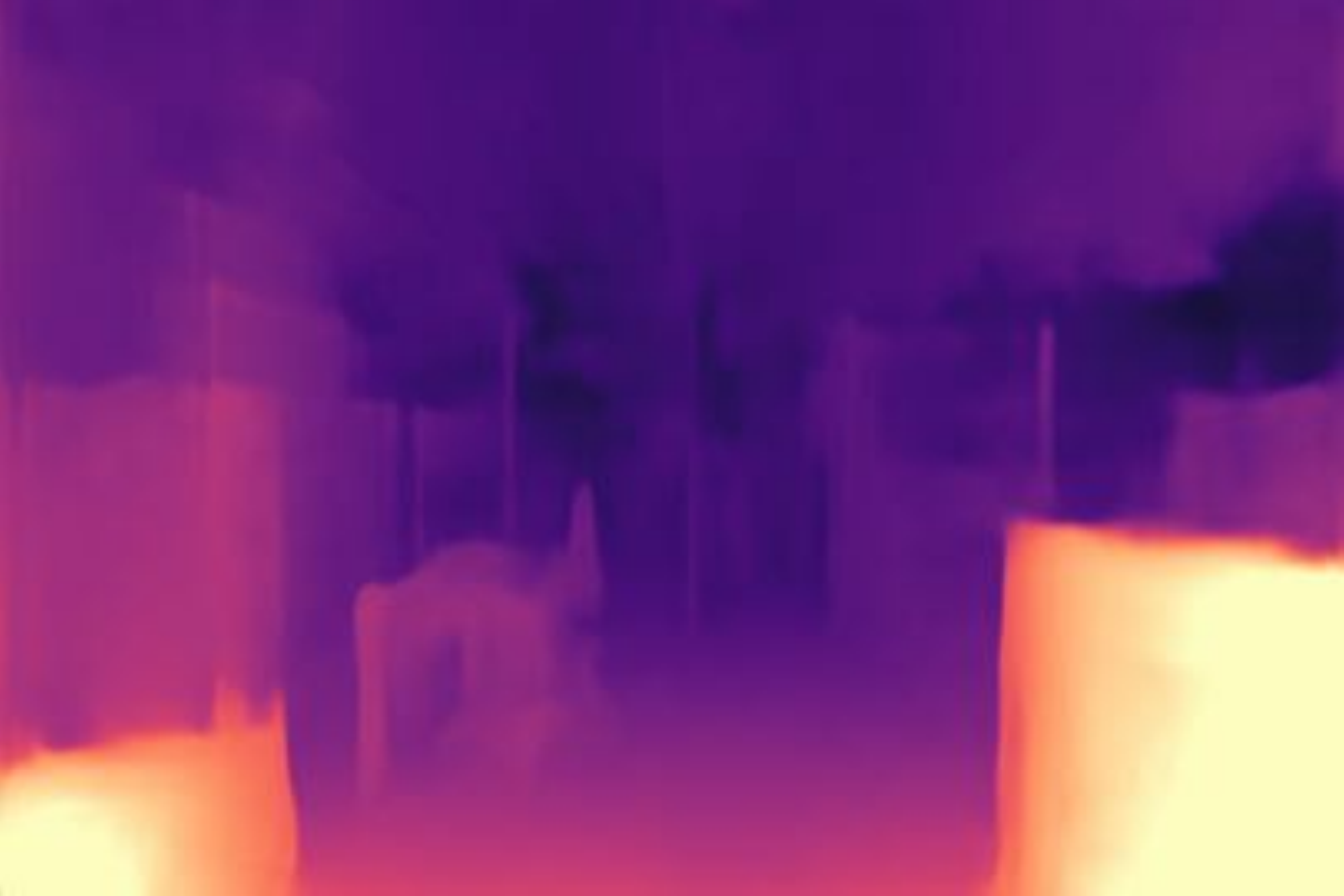}& 
\includegraphics[width=\w,height=\h]{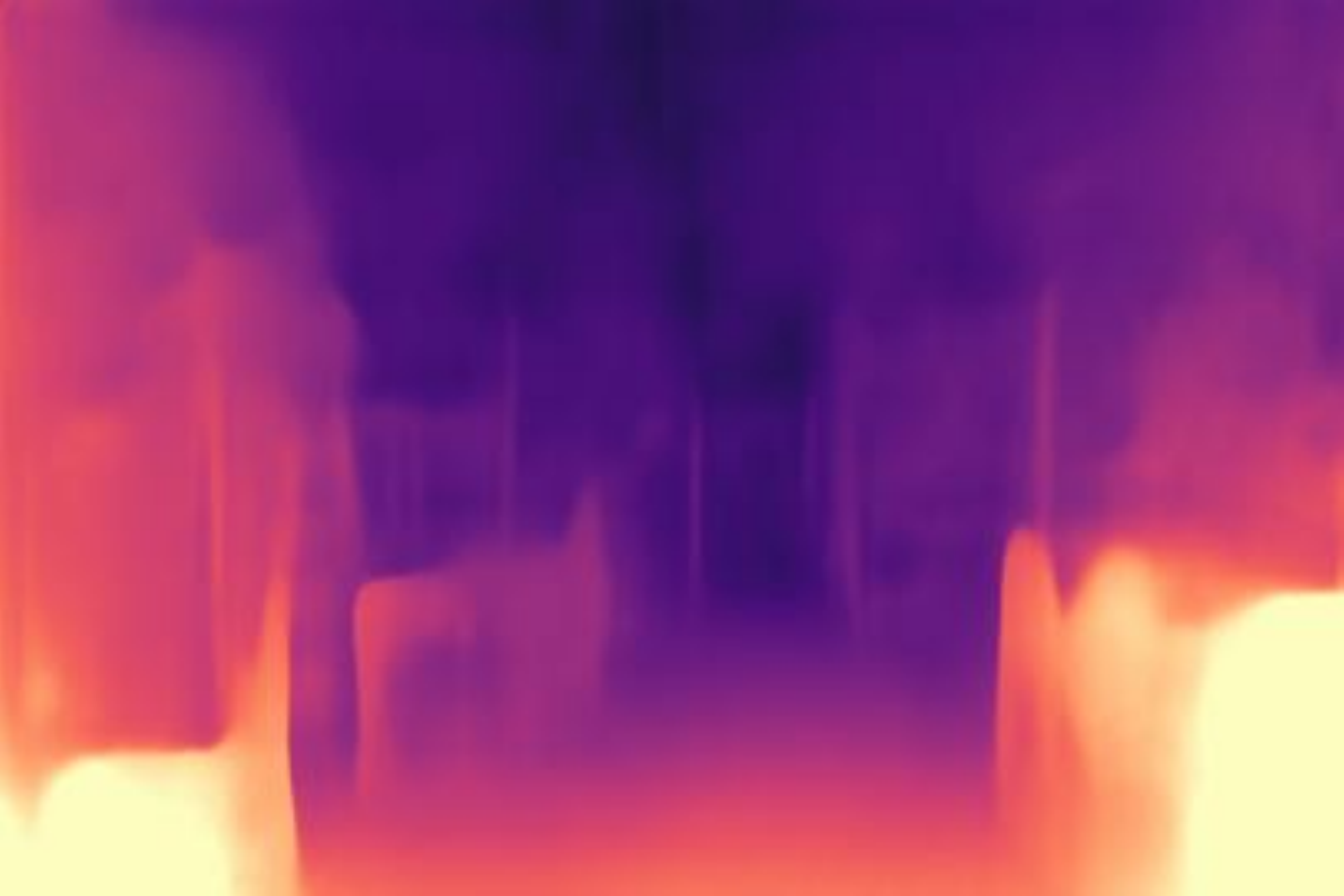}& 
\includegraphics[width=\w,height=\h]{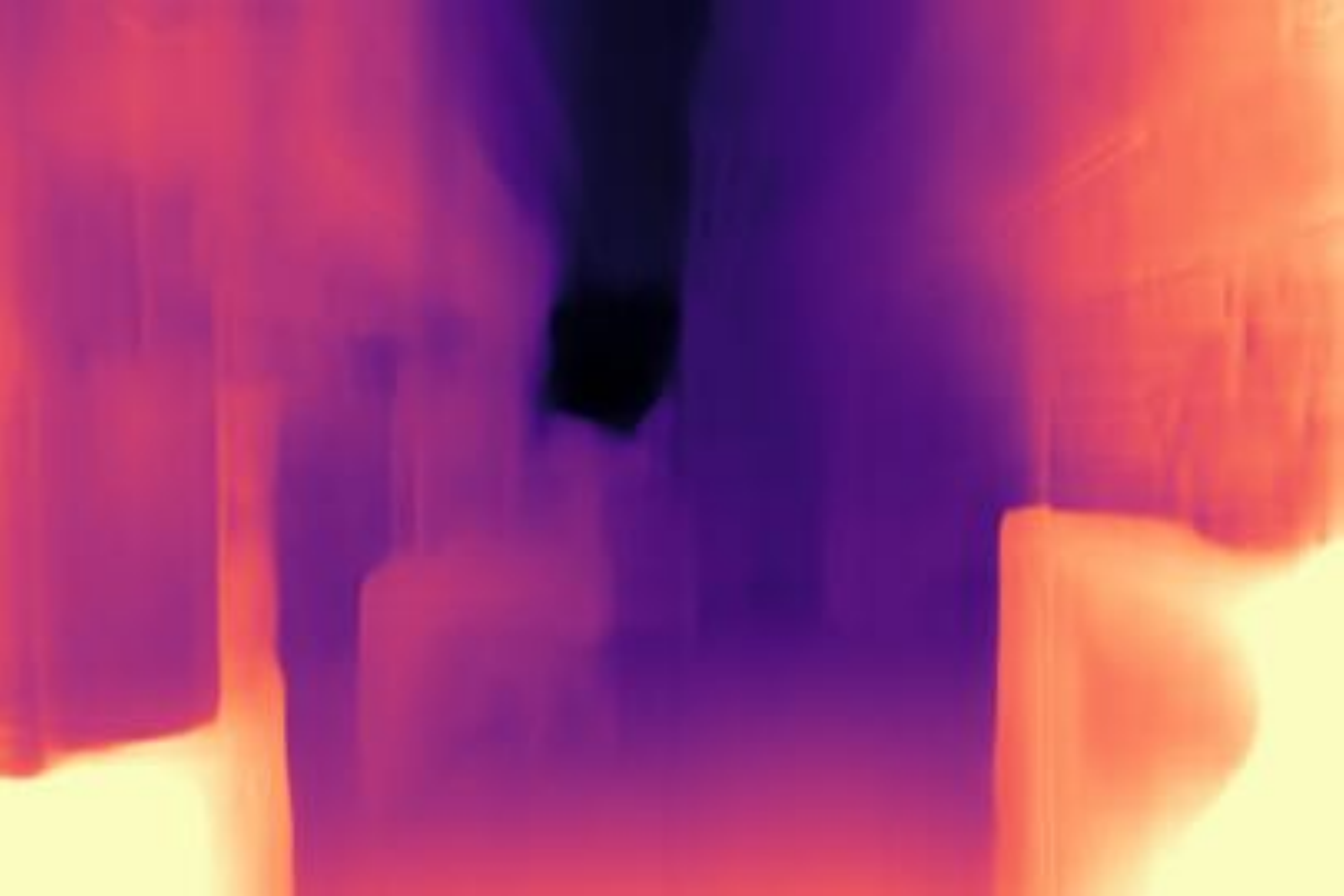}& 
\includegraphics[width=\w,height=\h]{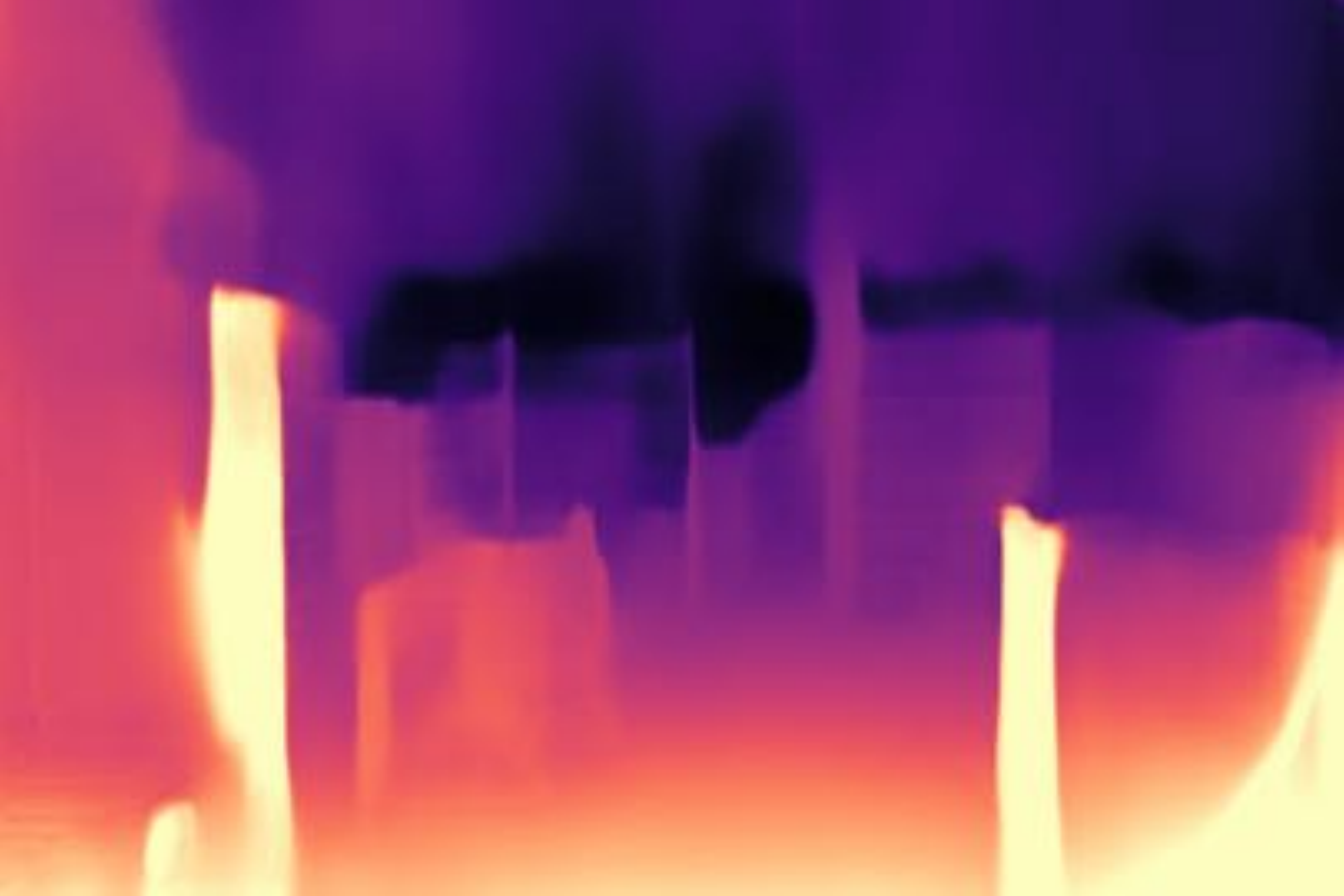}& 
\includegraphics[width=\w,height=\h]{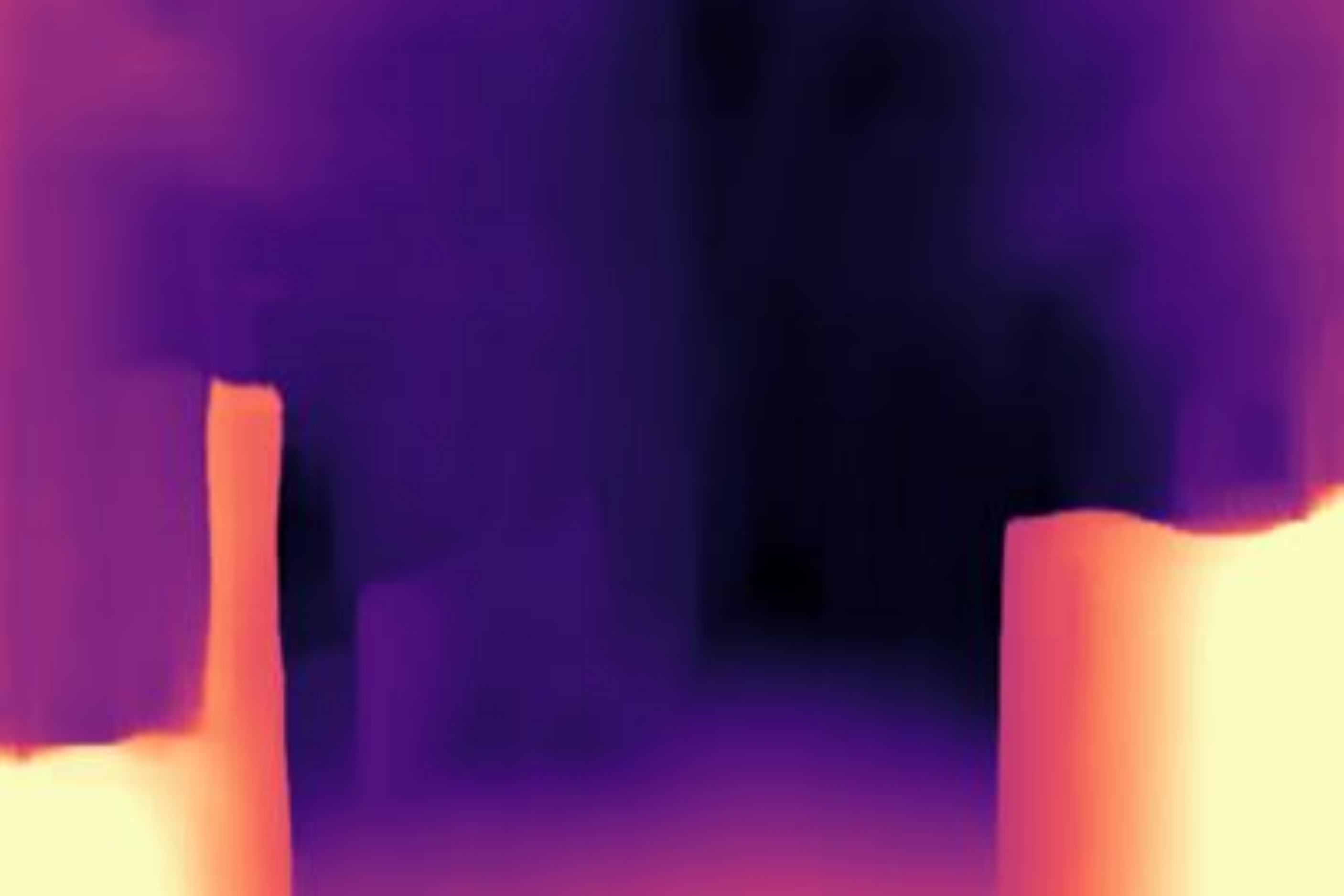}\\ 

\includegraphics[width=\w,height=\h]{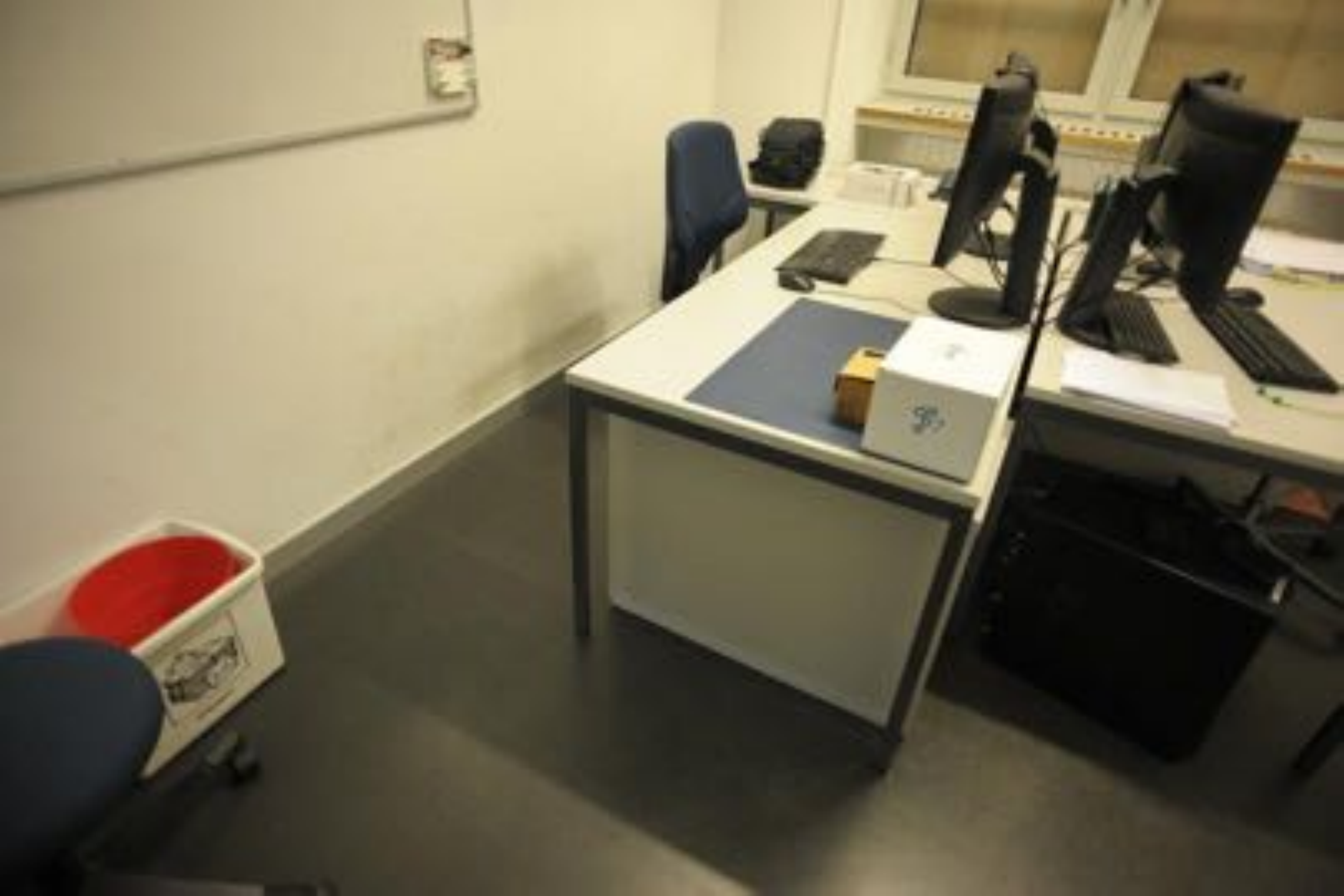}& 
\includegraphics[width=\w,height=\h]{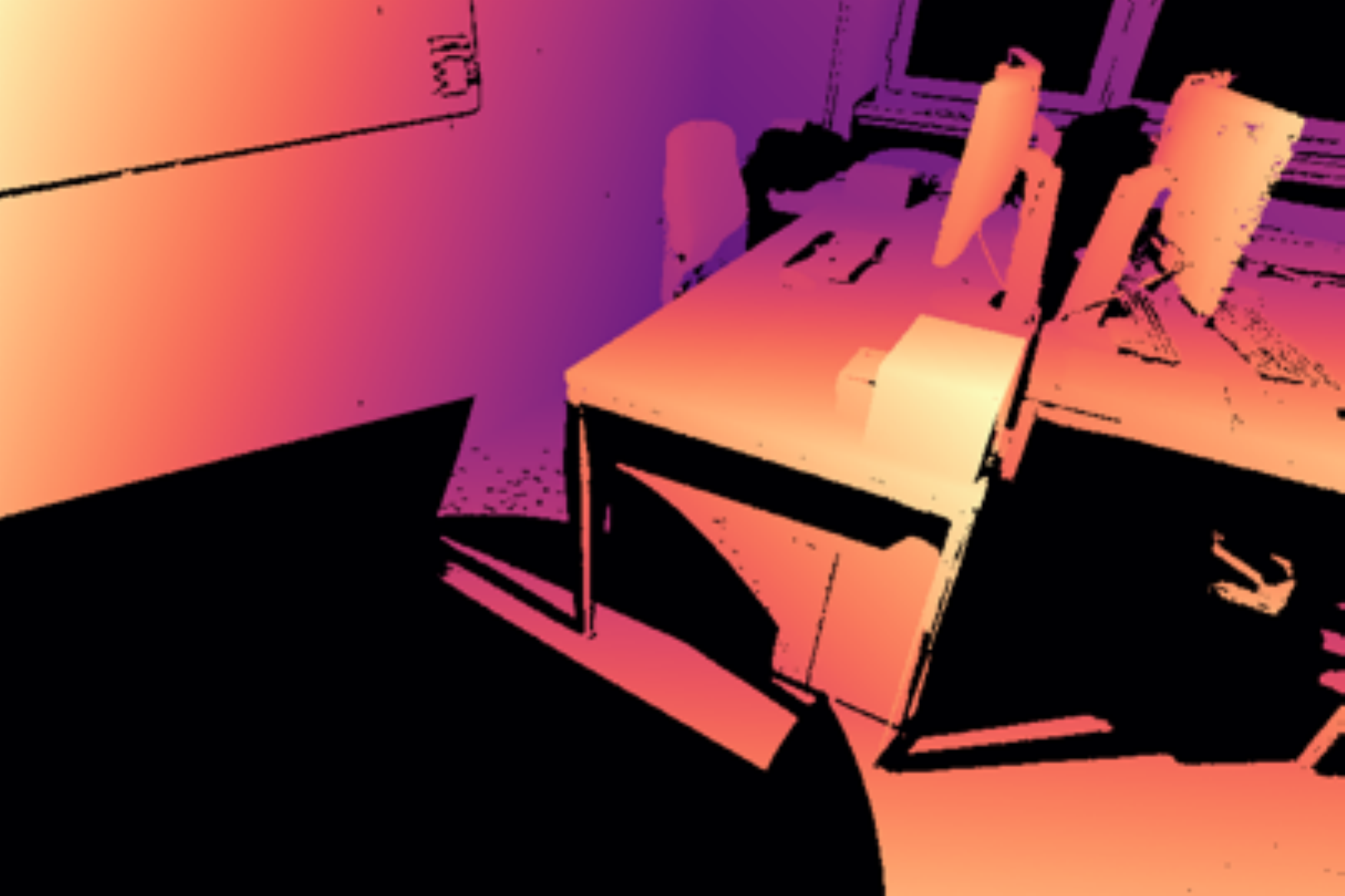}& 
\includegraphics[width=\w,height=\h]{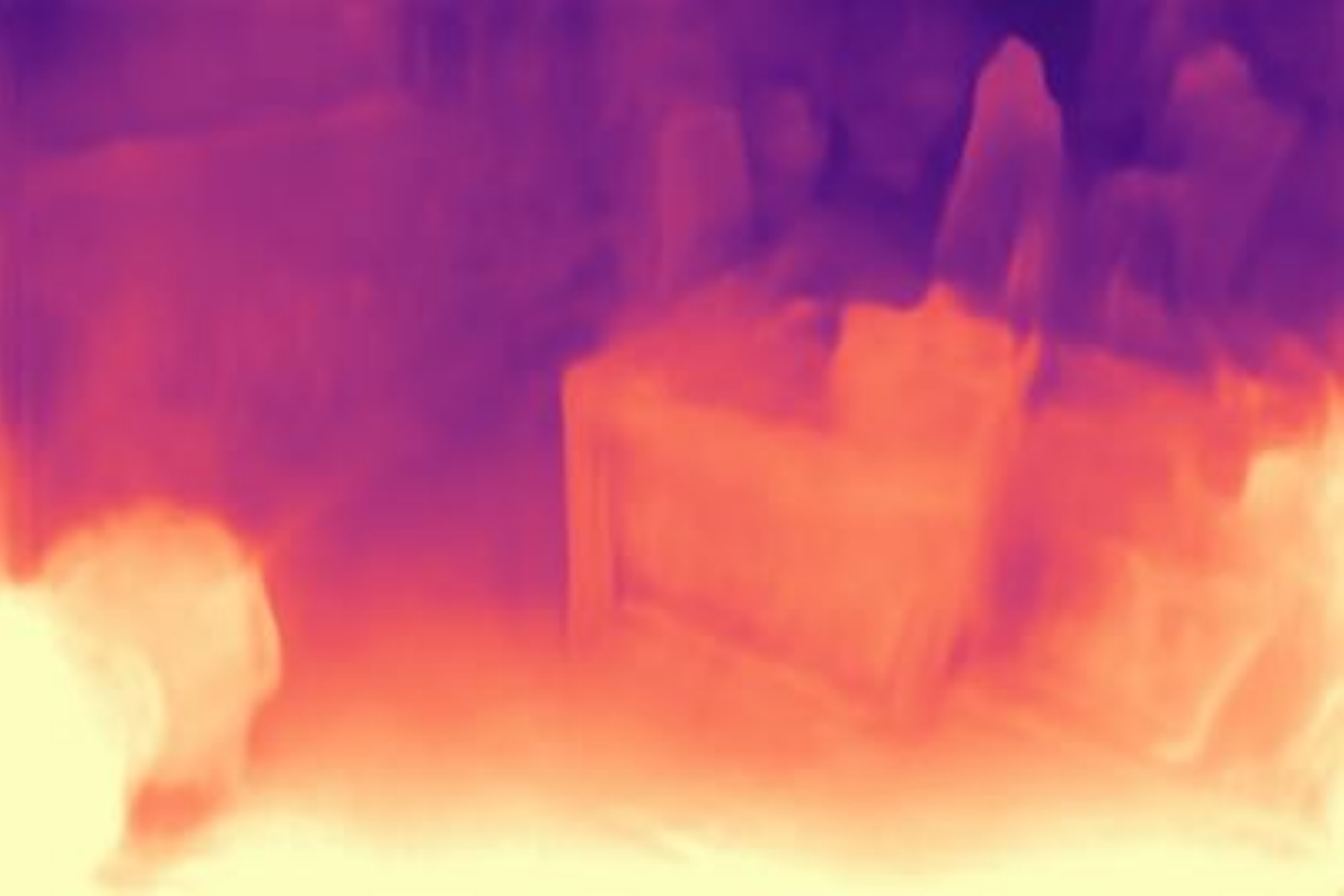}& 
\includegraphics[width=\w,height=\h]{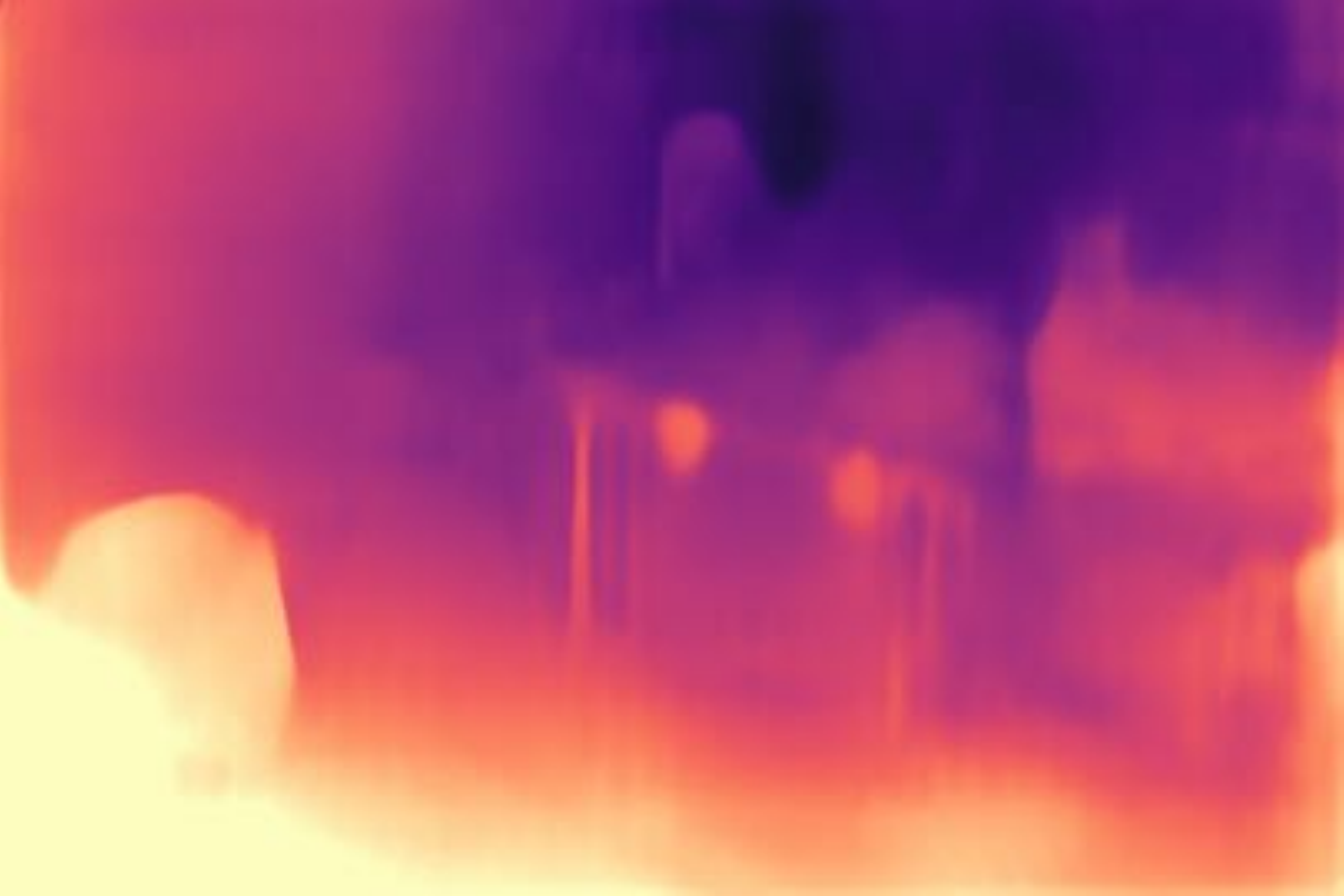}& 
\includegraphics[width=\w,height=\h]{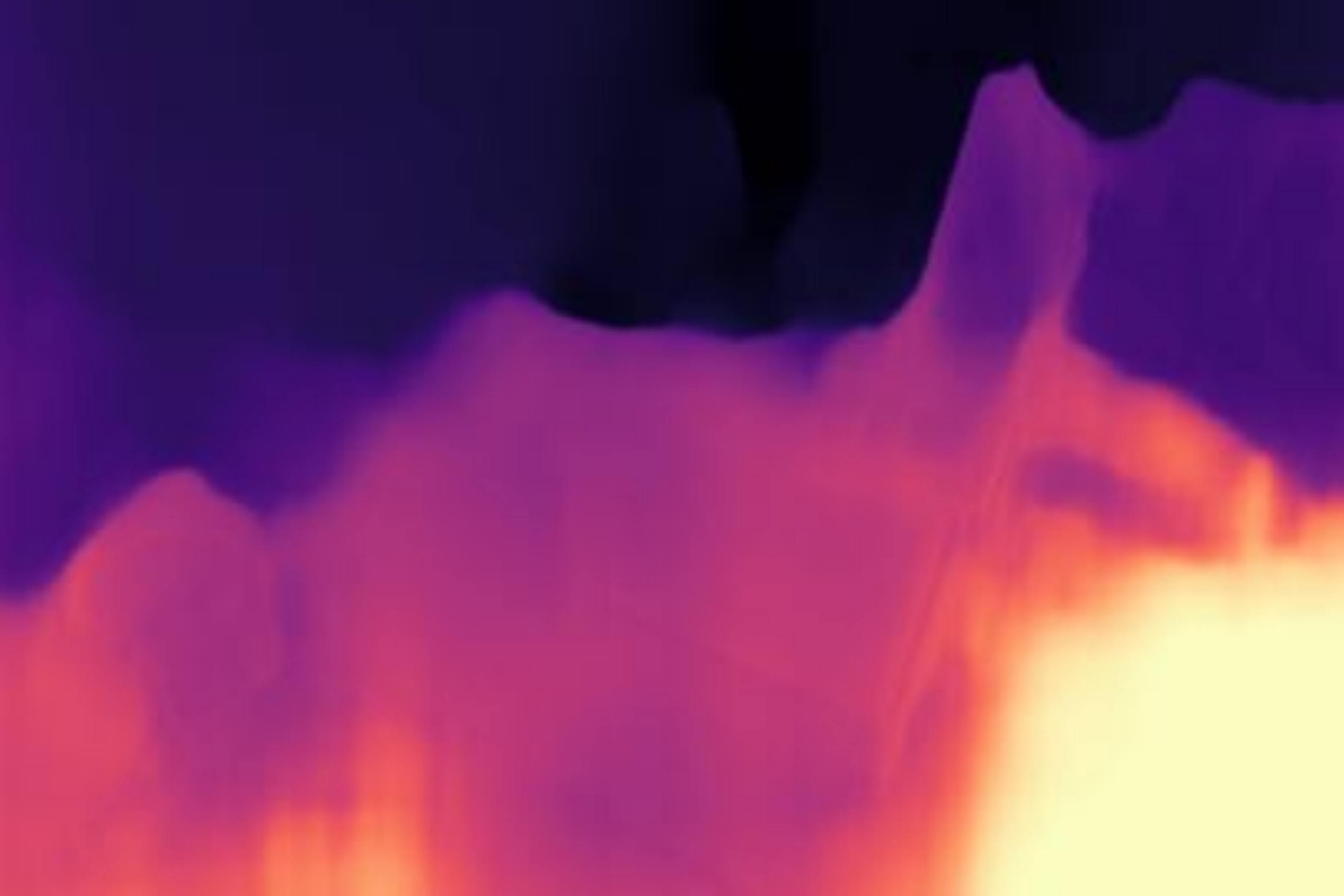}& 
\includegraphics[width=\w,height=\h]{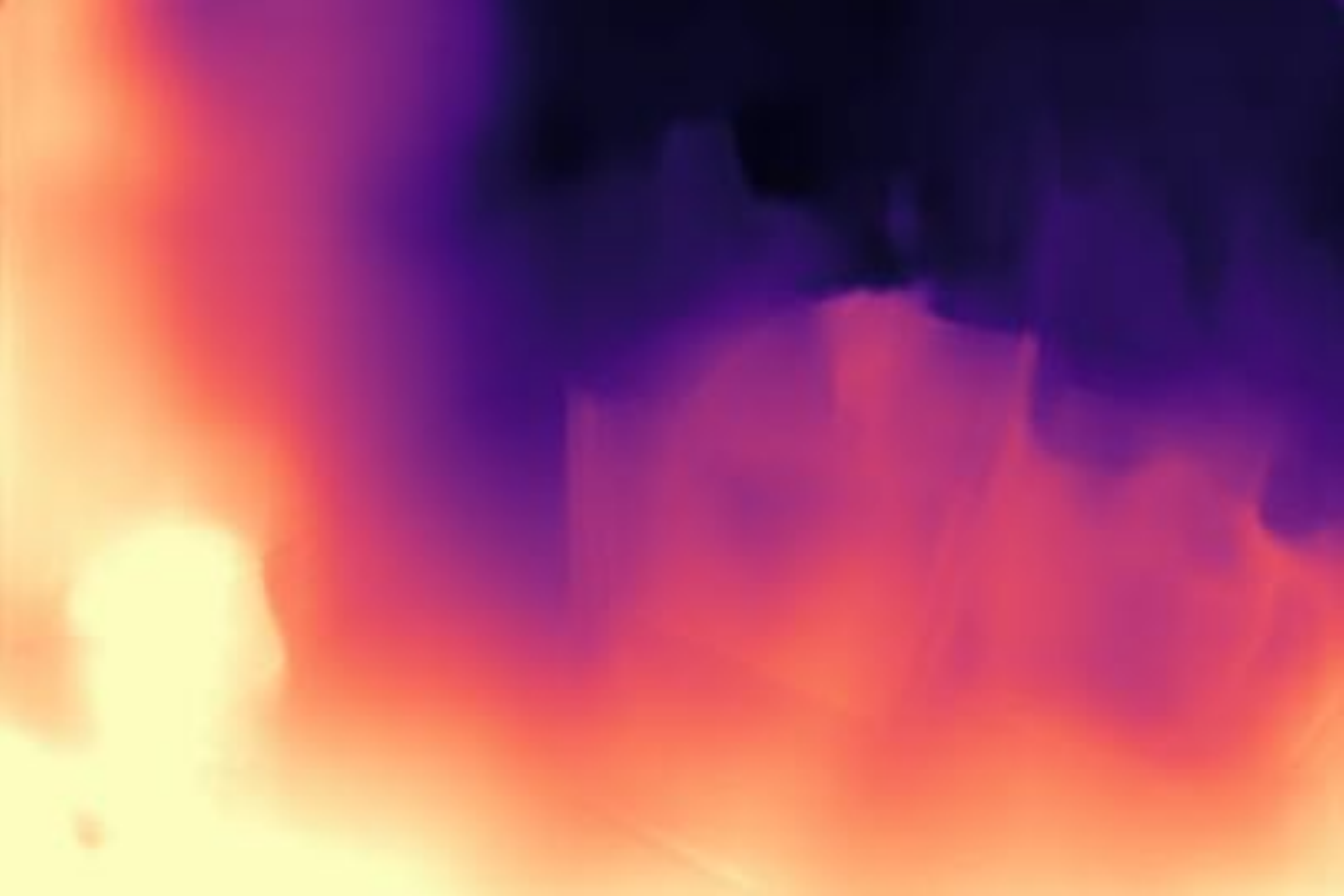}& 
\includegraphics[width=\w,height=\h]{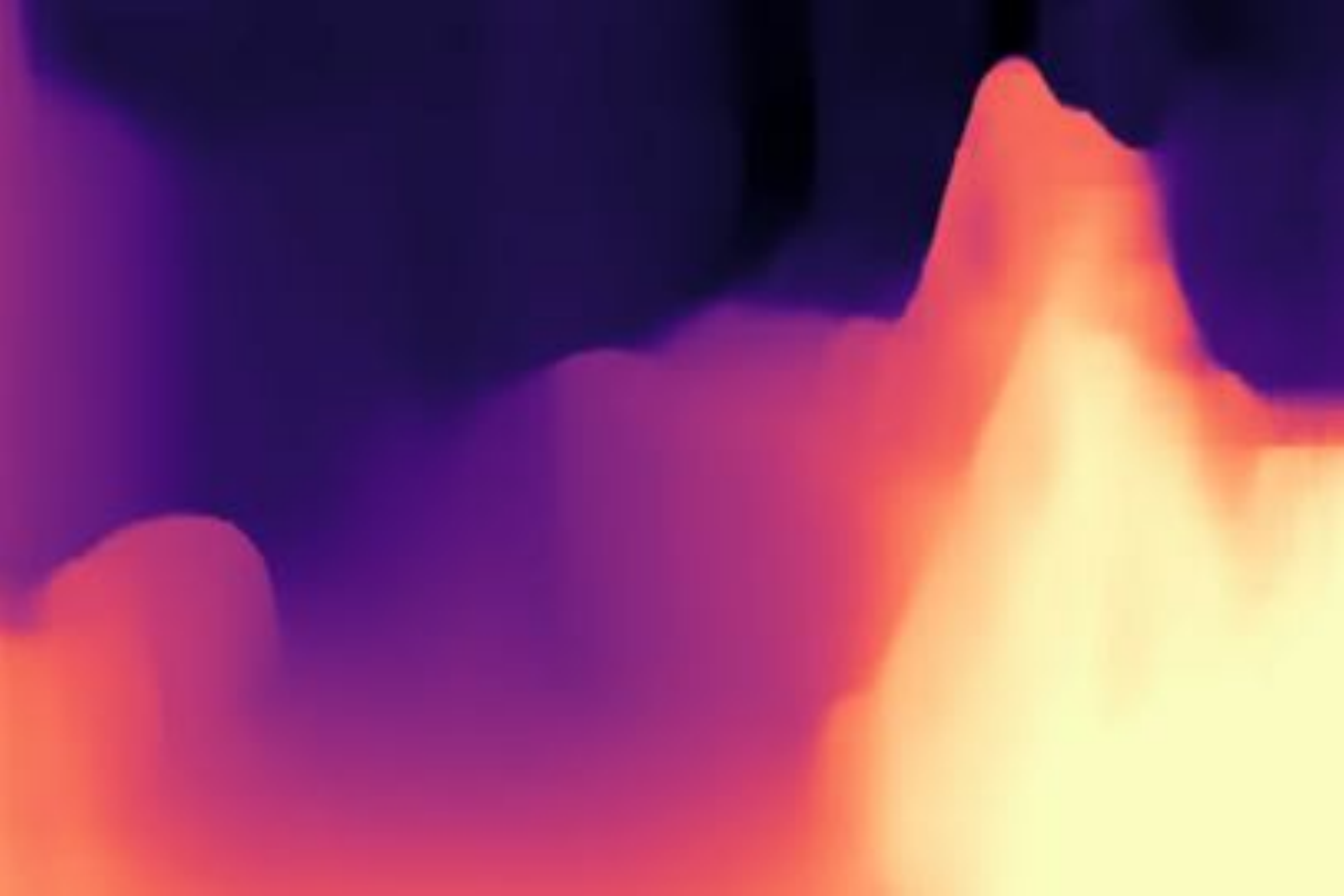}\\ 

\includegraphics[width=\w,height=\h]{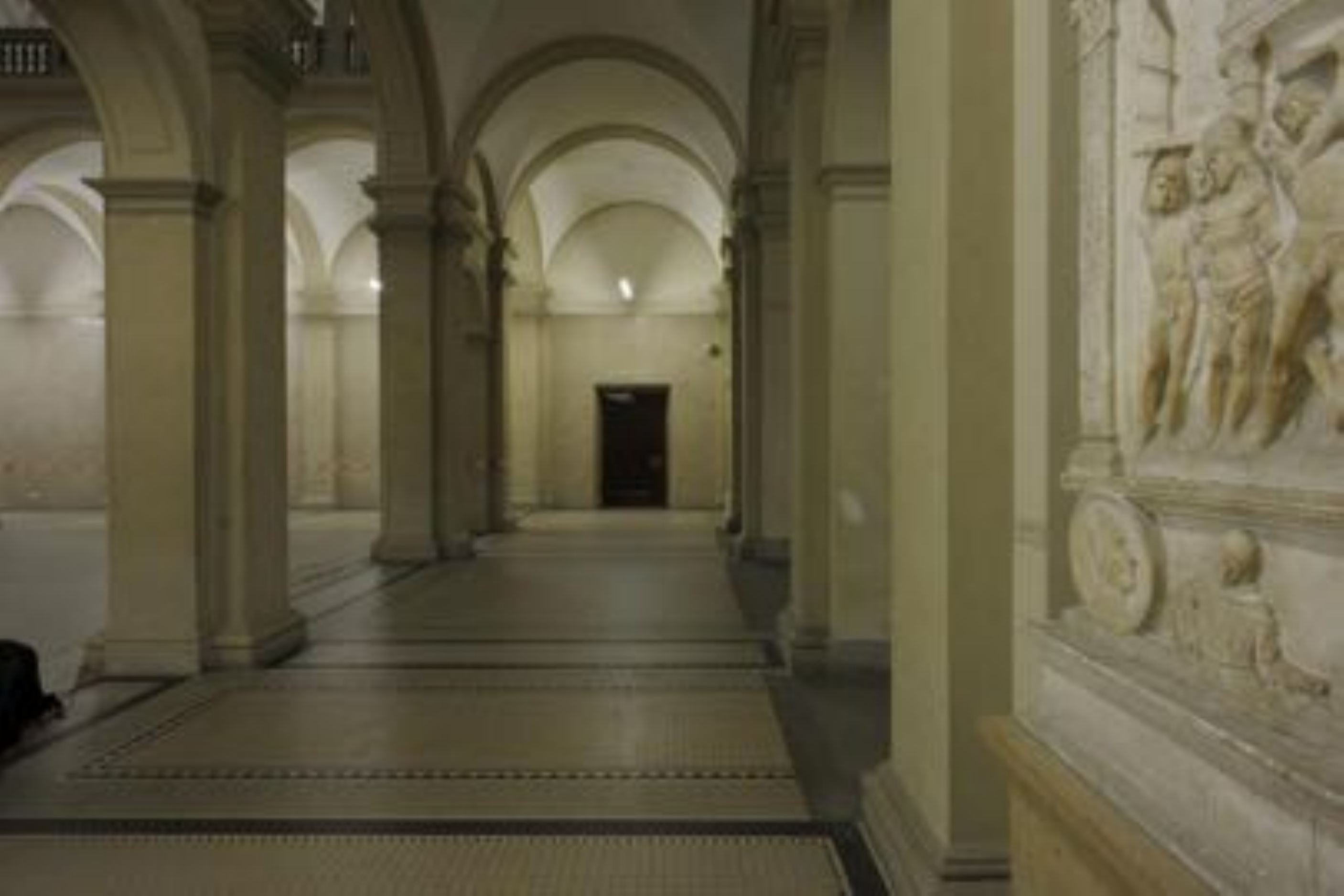}& 
\includegraphics[width=\w,height=\h]{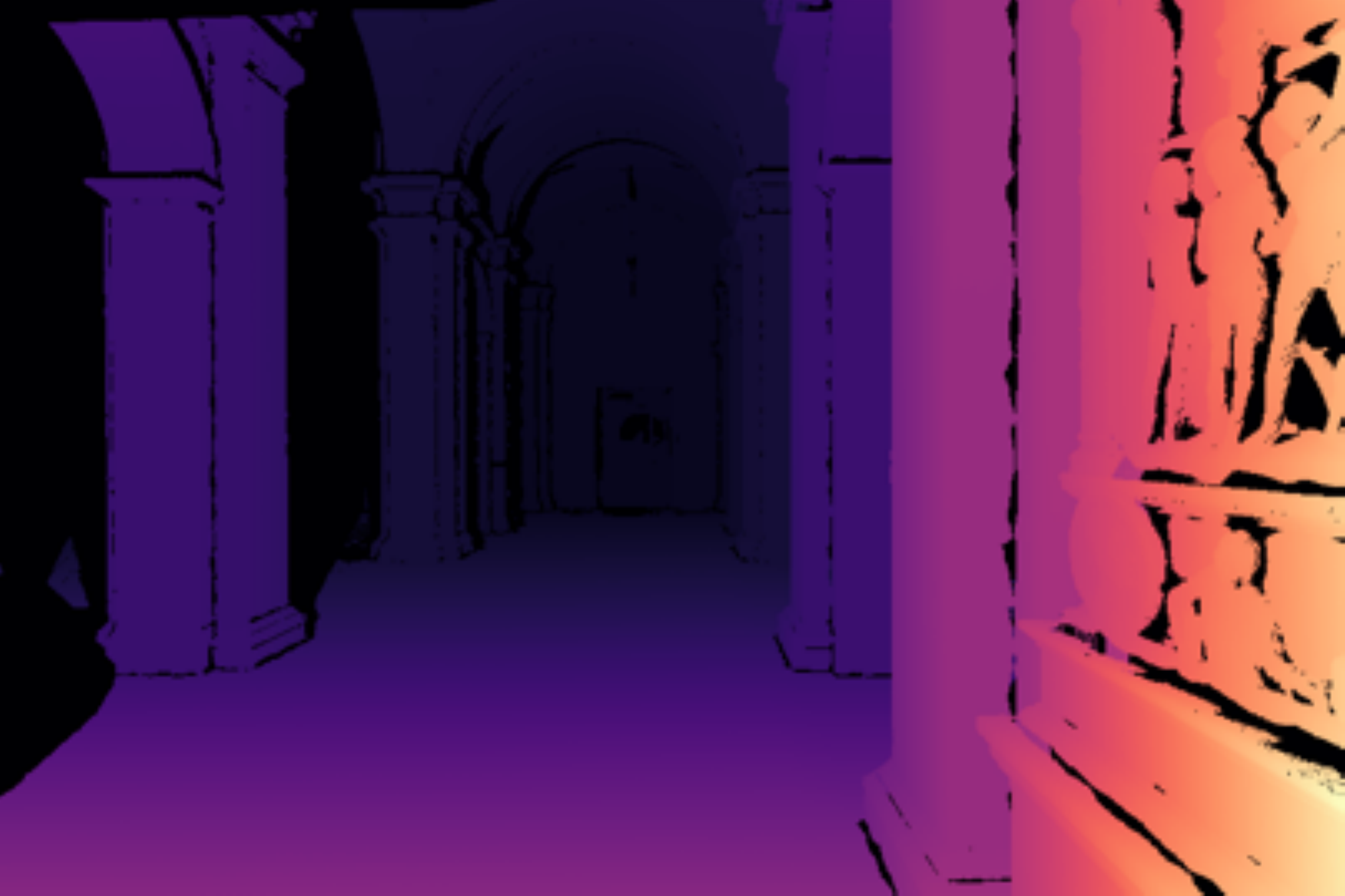}& 
\includegraphics[width=\w,height=\h]{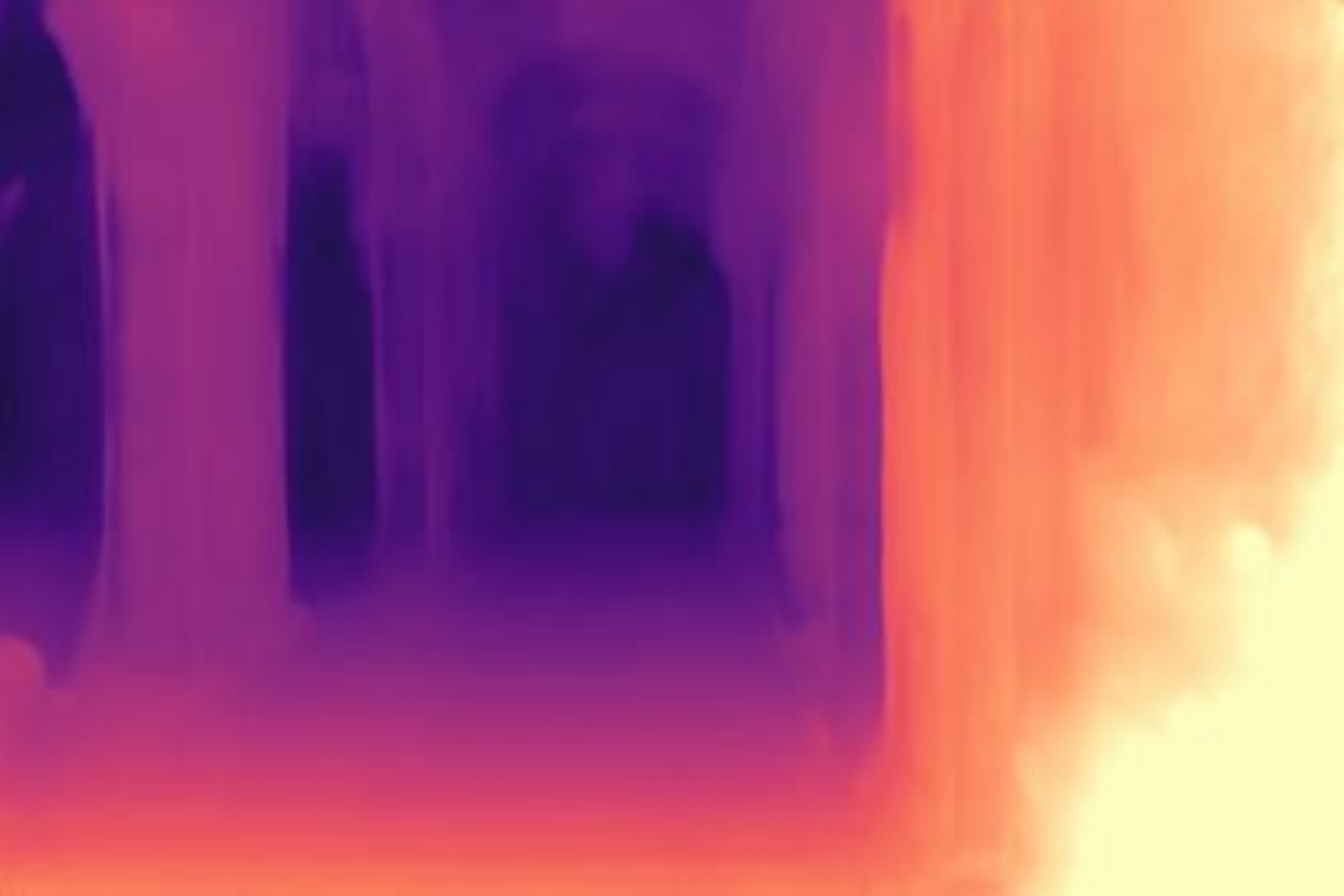}& 
\includegraphics[width=\w,height=\h]{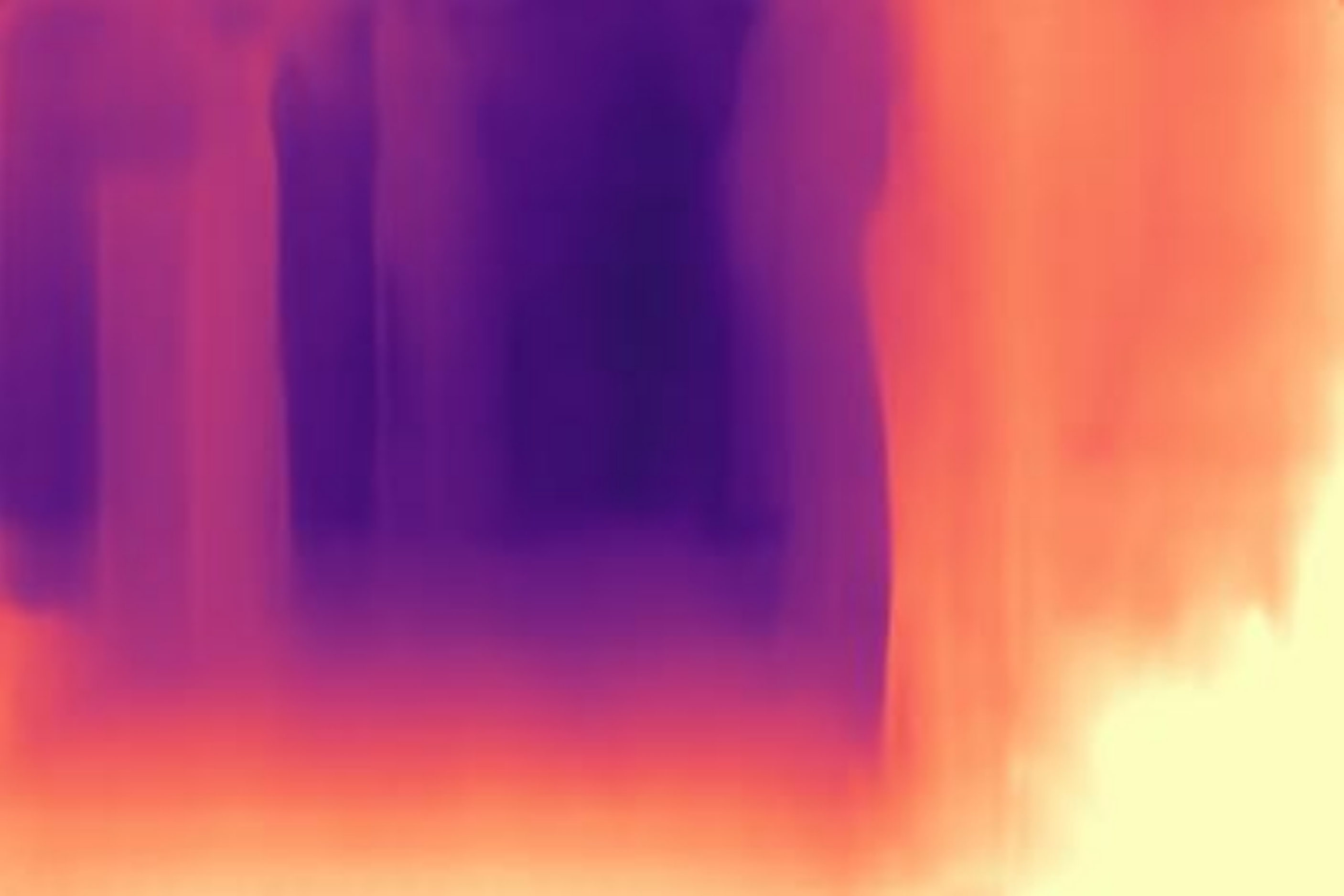}& 
\includegraphics[width=\w,height=\h]{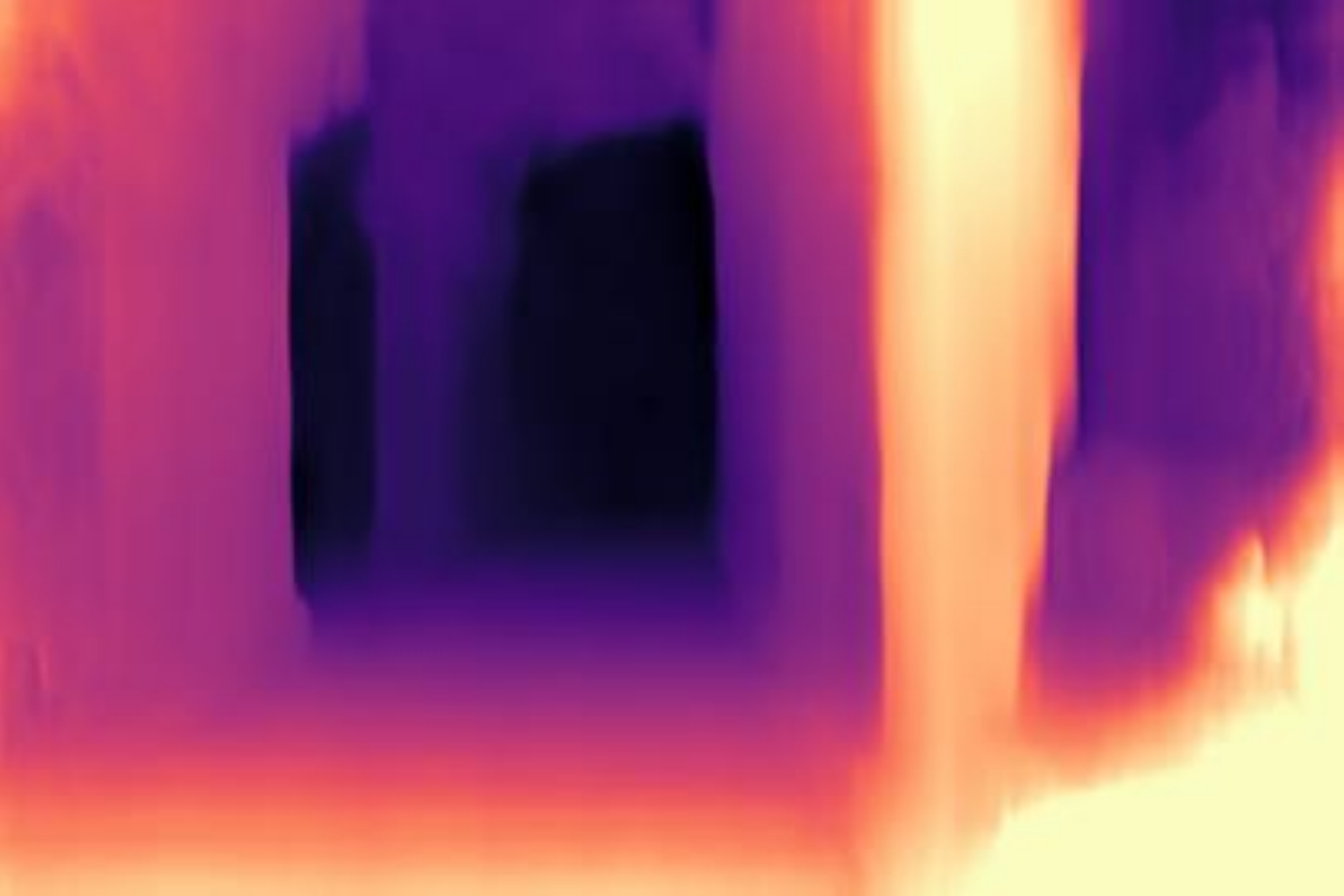}& 
\includegraphics[width=\w,height=\h]{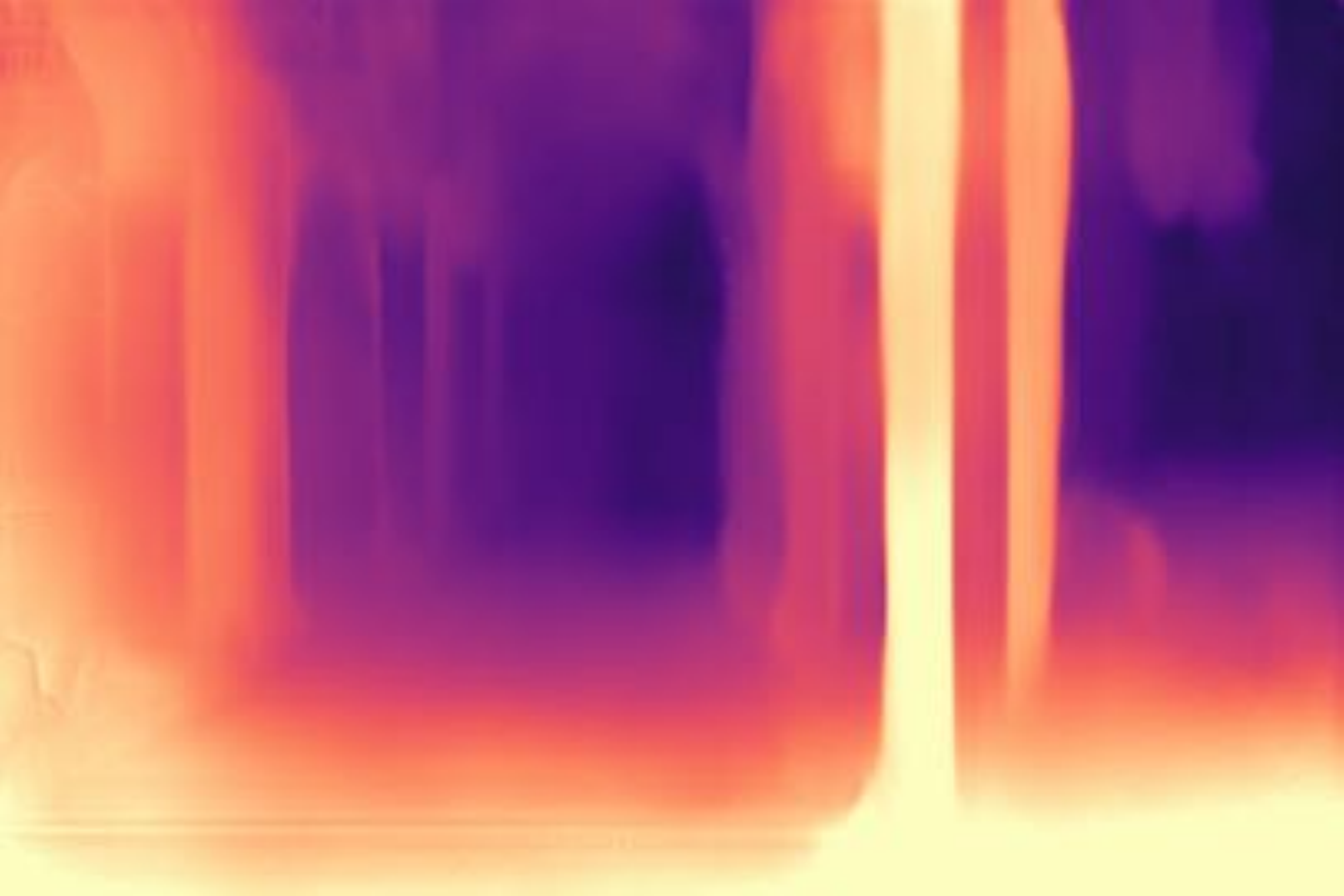}& 
\includegraphics[width=\w,height=\h]{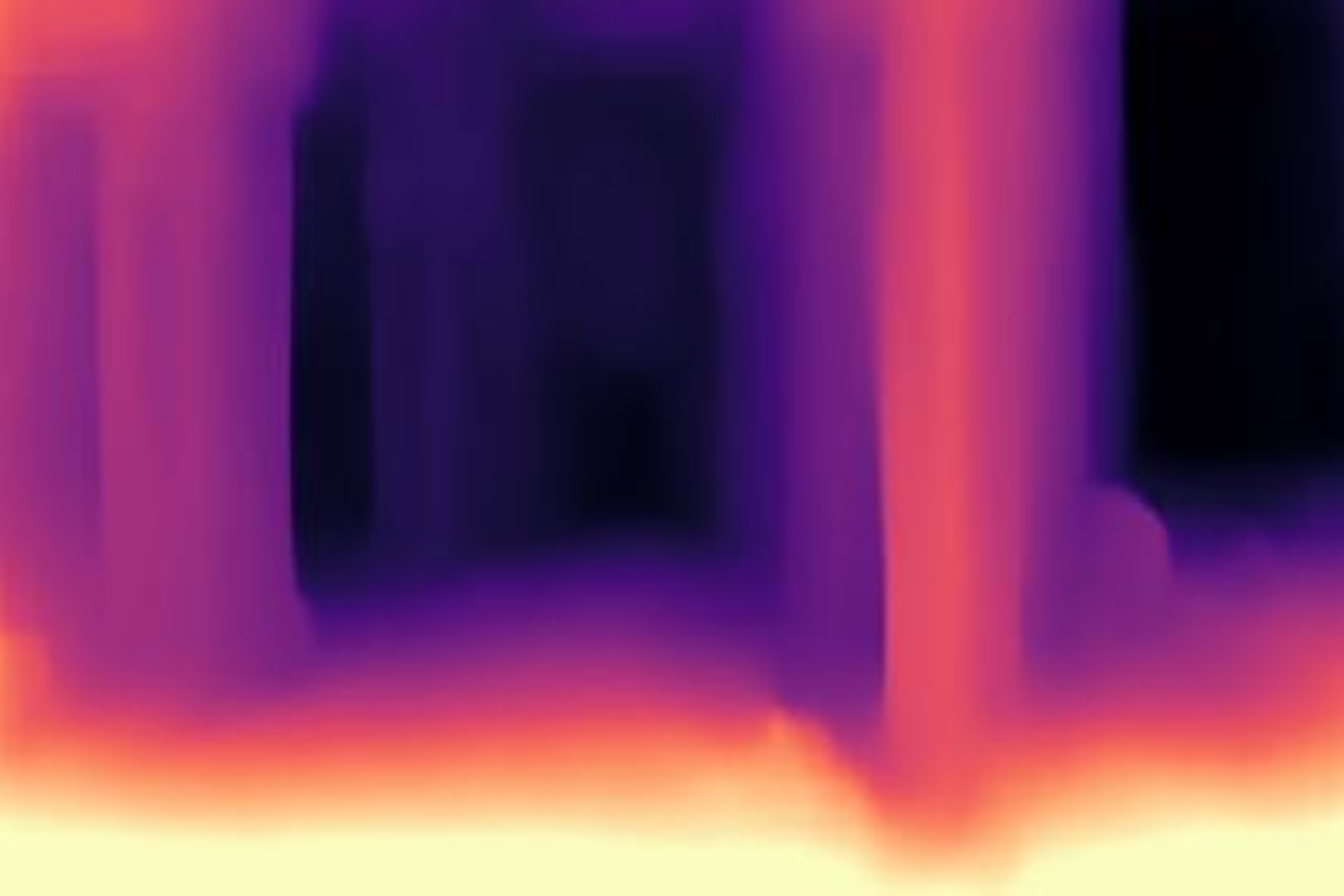}\\

     \fontsize{40}{30} \selectfont Input images &
     \fontsize{40}{30} \selectfont GT depths &
     \fontsize{40}{30} \selectfont Ours-Hybrid & 
      \fontsize{40}{30} \selectfont Ours-ViT & 
     \fontsize{40}{30} \selectfont Monodepth2 & 
     \fontsize{40}{30} \selectfont PackNet-SfM & 
     \fontsize{40}{30} \selectfont R-MSFM6
    \end{tabular}}
    \vspace{-0.3cm}
    \caption{\textbf{Comparison of depth map results on ETH3D dataset.}}
    \label{figure_result_ETH3D_apdx}
     \end{subfigure}
 \vspace{-0.3cm}

\end{figure*}

We provide additional qualitative results on KITTI datasets, various datasets(MVS, SUN3D, RGBD and Scenes11), and ETH3D in \figref{figure_result_kitti_apdx}
, \figref{figure_result_demons_apdx}, and \figref{figure_result_ETH3D_apdx}.

\subsection{Additional Details of Experiments}

\subsubsection{Training Loss and Implementation Detail}
We train both depth and motion networks using photometric consistency (L2 loss and SSIM loss) and edge-aware smoothness losses following the best practices of self-supervised monocular depth estimation \cite{zhou2017unsupervised,godard2019digging,guizilini20203d}.
We set the weight for SSIM, L2 photometric, and smoothness losses as $0.85$, $0.15$ and $0.001$, respectively. 
We use 7 convolution layers for 6DoF camera pose estimation following the work in \cite{zhou2017unsupervised}.
We implement our framework on PyTorch and train it on 4 Titan RTX GPUs. We use the Adam optimizer \cite{kingma2014adam} with $\beta_1 = 0.9$ and $\beta_2 = 0.999$. Our model is trained for 50 epochs with a batch size of 8. The learning rates for depth and pose network are $2 \times 10^{-5}$ and $5 \times 10^{-4}$, respectively. 
We will release the source code, the trained weights and the datasets once the paper is accepted.

\subsubsection{The Number of Encoder and Decoder Layers.}
We compare the performance of our model according to the number of encoder and decoder layers in \tabref{table_abliation_L_apdx}. 
We find out that the model with four transformer layers achieves the best performance.
Thus, we set $L$ as four for our MonoFormer.

\begin{table}[h]
    \centering
    \resizebox{\columnwidth}{!}{
    \begin{tabular}{cccccc}
    \hline
    $\#$ of layers & Abs Rel $\downarrow$ & RMSE $\downarrow$ &  & $\delta < 1.25$ $\uparrow$ & $\delta < 1.25^2 $ $\uparrow$ \\ \hline
    $L=2$ & 0.148 & 5.327  &   &0.810  & 0.939  \\
    $L=3$ & 0.112 & 4.745&  & 0.881  & \textbf{0.962} \\
    $L=4$ & \textbf{0.104} & \textbf{4.580} &  & 0.873 & \textbf{0.962} \\
    $L=5$ & 0.111 & 4.692 & &  \textbf{0.884} & \textbf{0.962} \\
    \hline
    \end{tabular}}
    \caption{\textbf{Ablation study on the number of encoder and decoder layers.}}
    \label{table_abliation_L_apdx}
\end{table}

\subsubsection{Implementation Detail of Image Generation}
The following is a summary of the image generation :
\begin{table}[!h]
\resizebox{\columnwidth}{!}{
\begin{tabular}{lll}  
\textit{Watercolor} & \multicolumn{2}{p{0.85\linewidth}}{\raggedright We smooth the texture details from original images while preserving the color cues using \texttt{cv2.stylization}. The image looks like a watercolor pictures.} \\ 
\textit{Pencil-sketch} & \multicolumn{2}{p{0.85\linewidth}}{\raggedright We remove both textures and color from original images using \texttt{cv2.pencilSketch}. The image seems like a sketch drawn with pencils.} \\
\textit{Style-transfer} & \multicolumn{2}{p{0.85\linewidth}}{\raggedright We apply a new texture to the original image (context) by utilizing other images (style) using a style transfer algorithm \cite{gatys2016image}. 
The textures of the original images are changed.} \\ 
\end{tabular}}
\end{table}

\begin{figure*}[t!]
    \begin{subfigure}
    \centering
    \resizebox{\linewidth}{!}{%
        \begin{tabular}{ccccccc}
                             &  
        \fontsize{120}{100}  {$\sigma_r = 10 , \sigma_s = 0.7$}&  
        \fontsize{120}{100} $\sigma_r = 190 , \sigma_s = 0.7$&  
        \fontsize{120}{100}  $\sigma_r = 60 , \sigma_s = 0.7$ \selectfont (default) &  
        \fontsize{120}{100}   $\sigma_r = 60 , \sigma_s = 0.1$&  
        \fontsize{120}{100}    $\sigma_r = 60 , \sigma_s = 0.9$\\
        
        \includegraphics[]{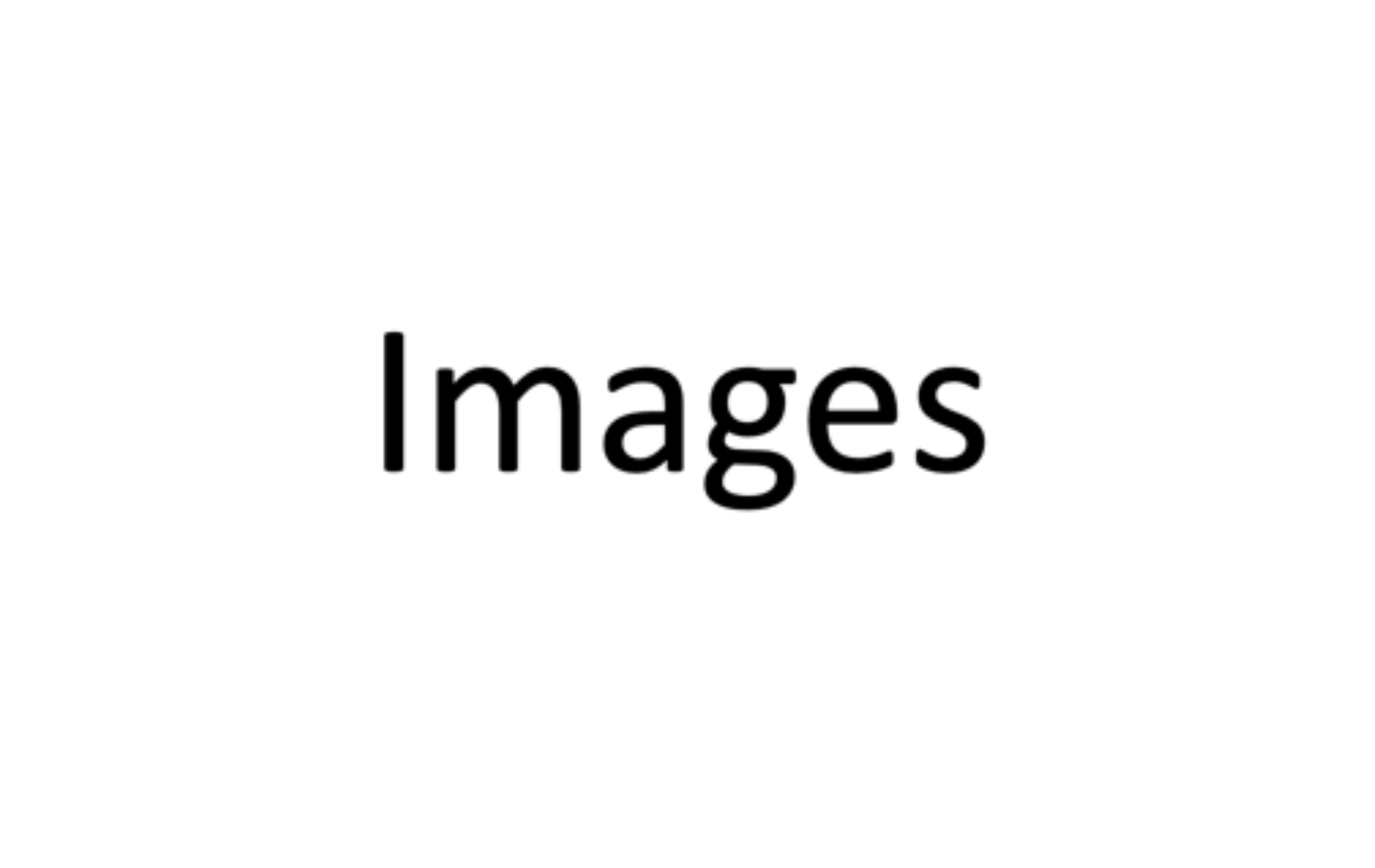} &  
        \includegraphics[]{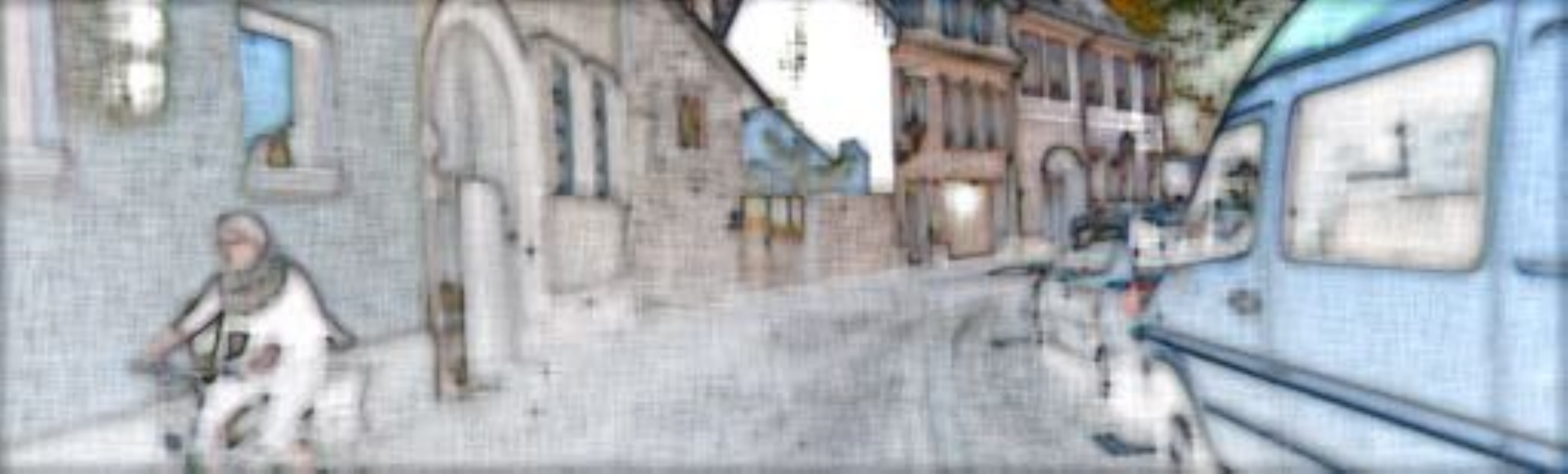}&  
        \includegraphics[]{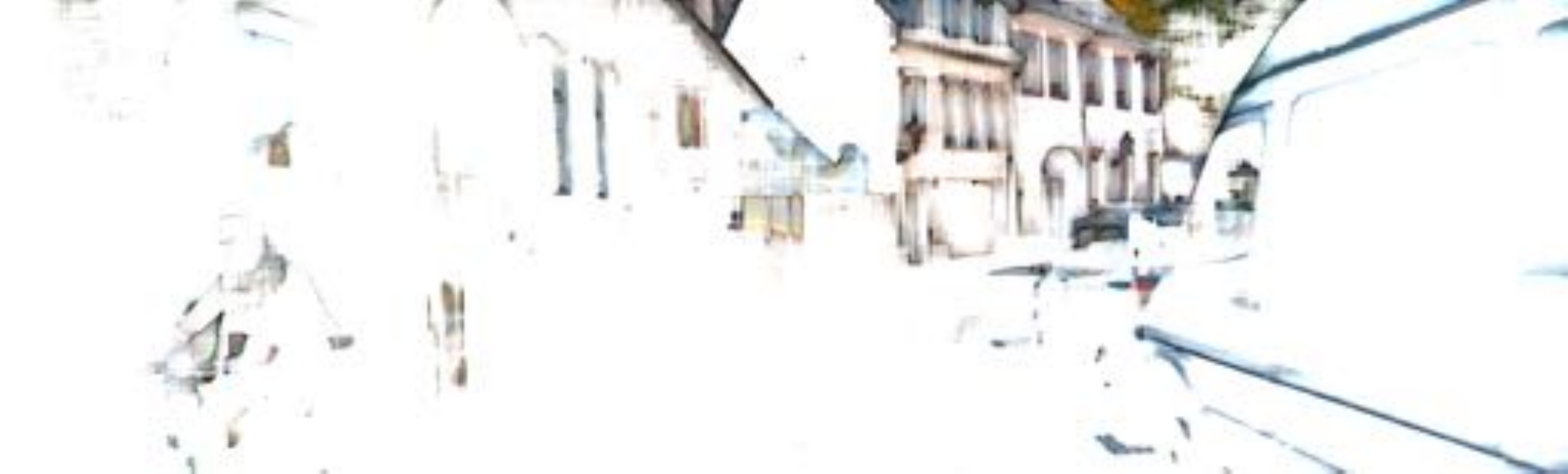}&  
        \includegraphics[]{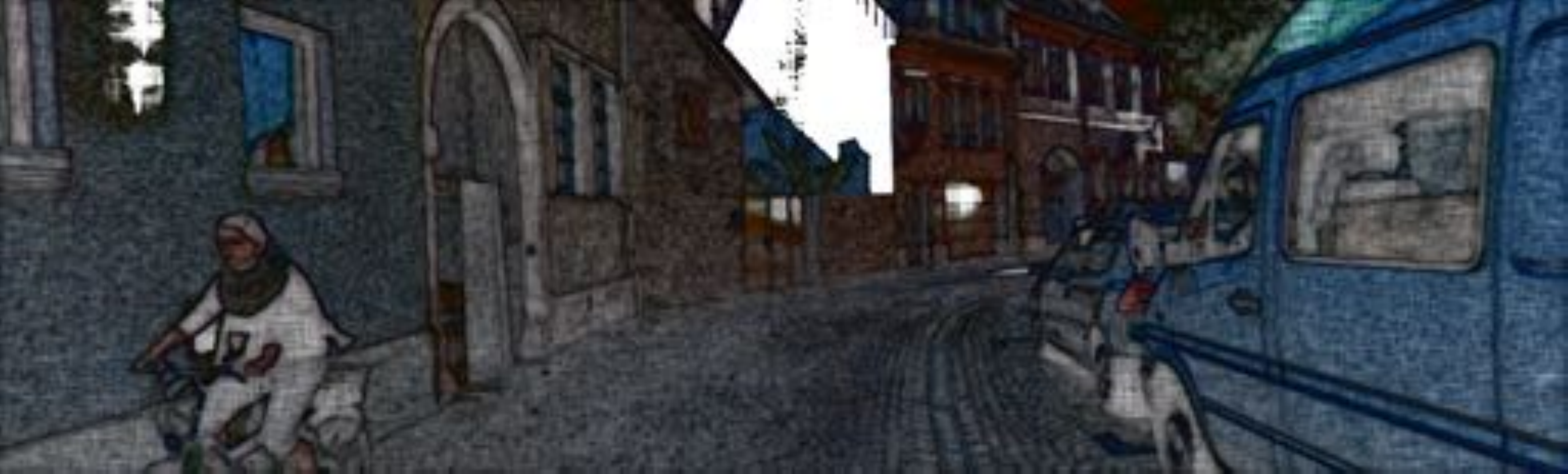}&  
        \includegraphics[]{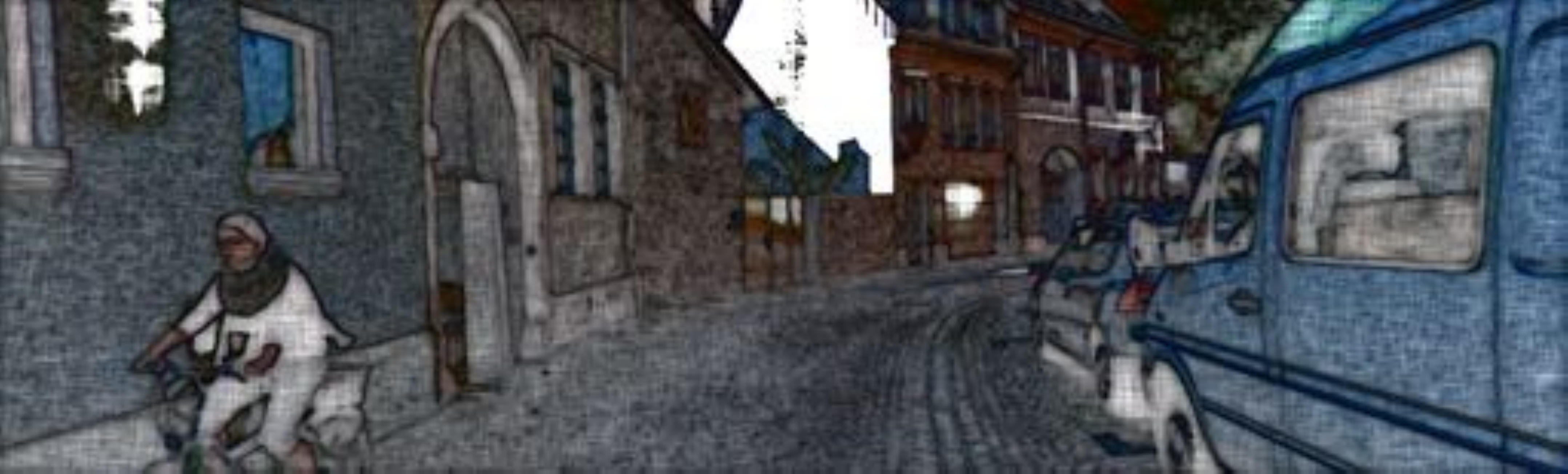}&  
        \includegraphics[]{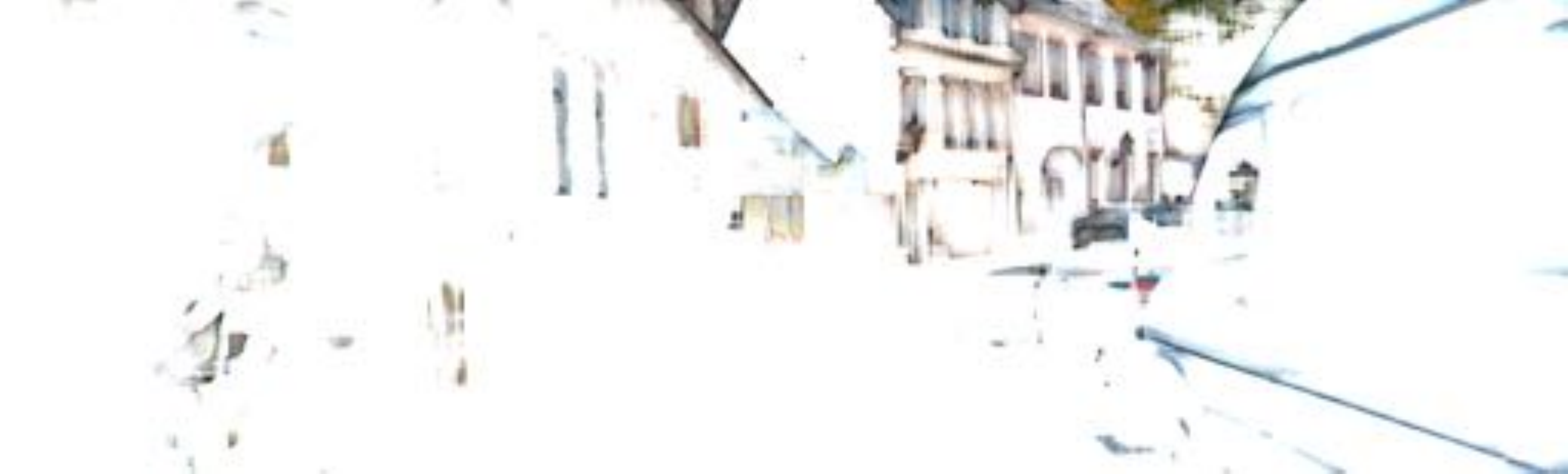}\\
        \includegraphics[]{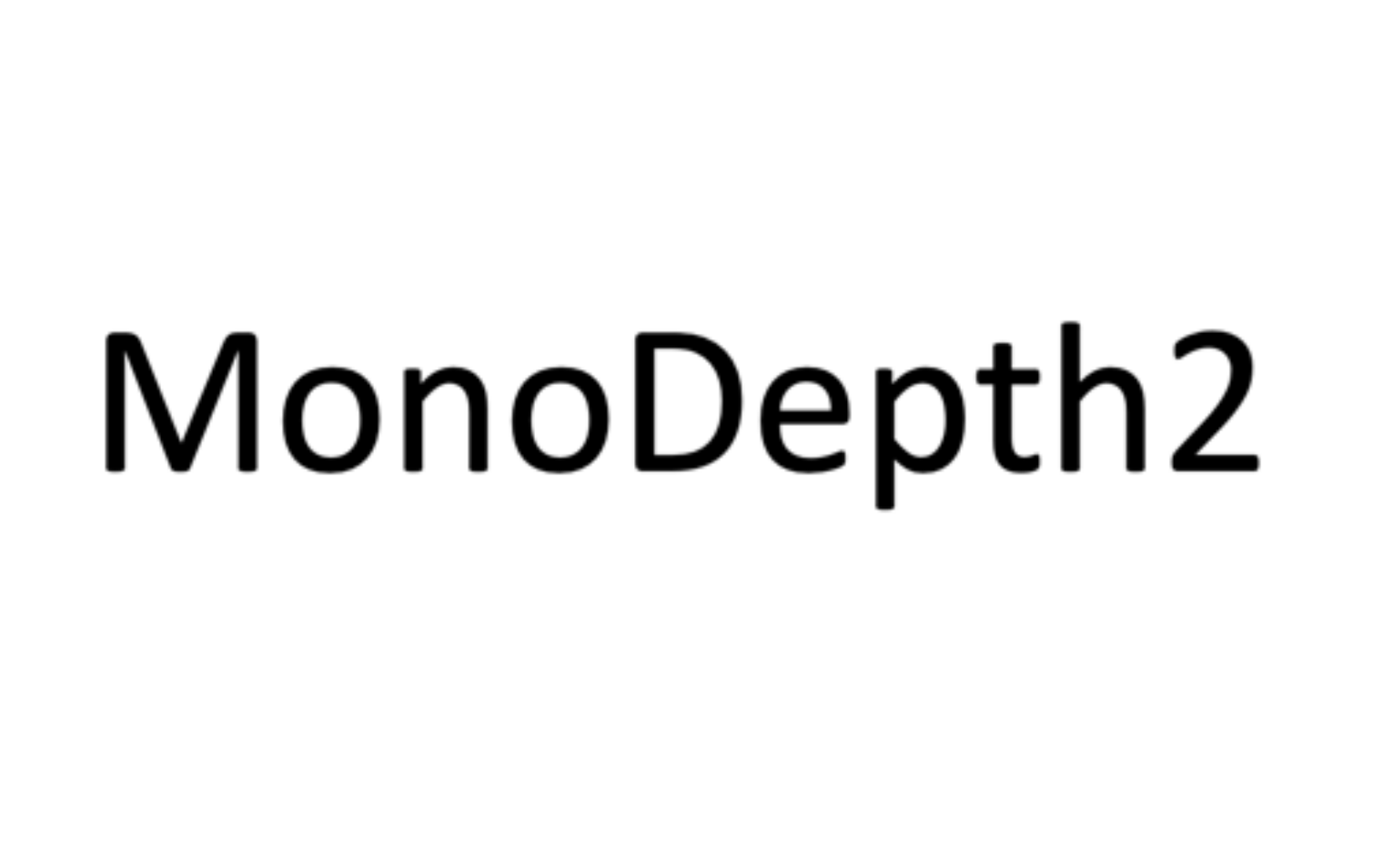}  &  
        \includegraphics[]{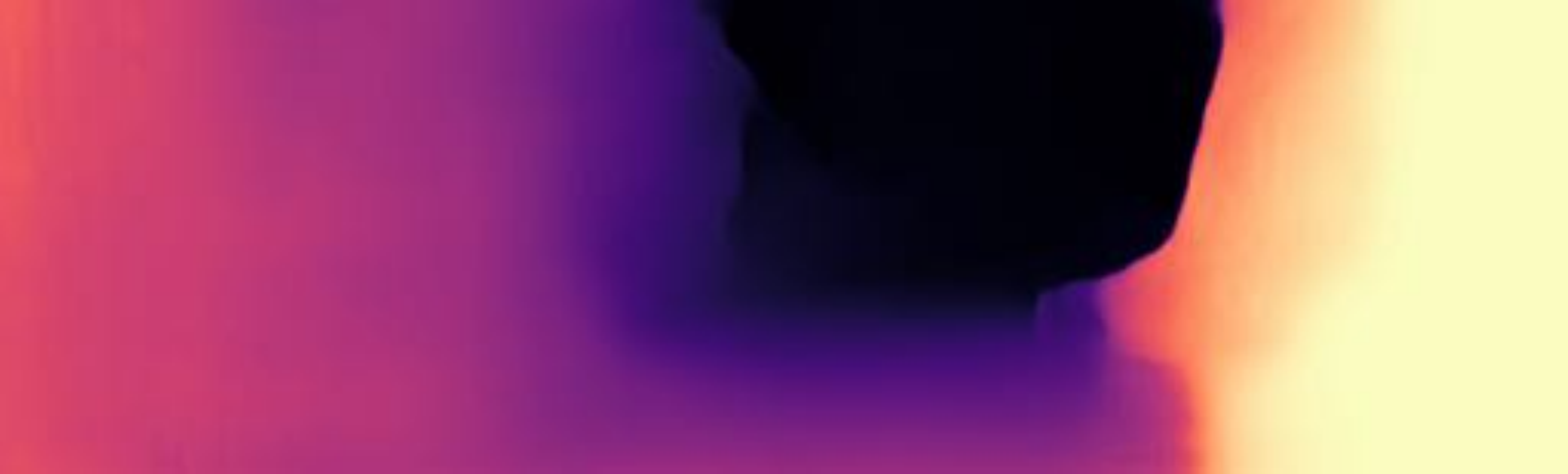}&  
        \includegraphics[]{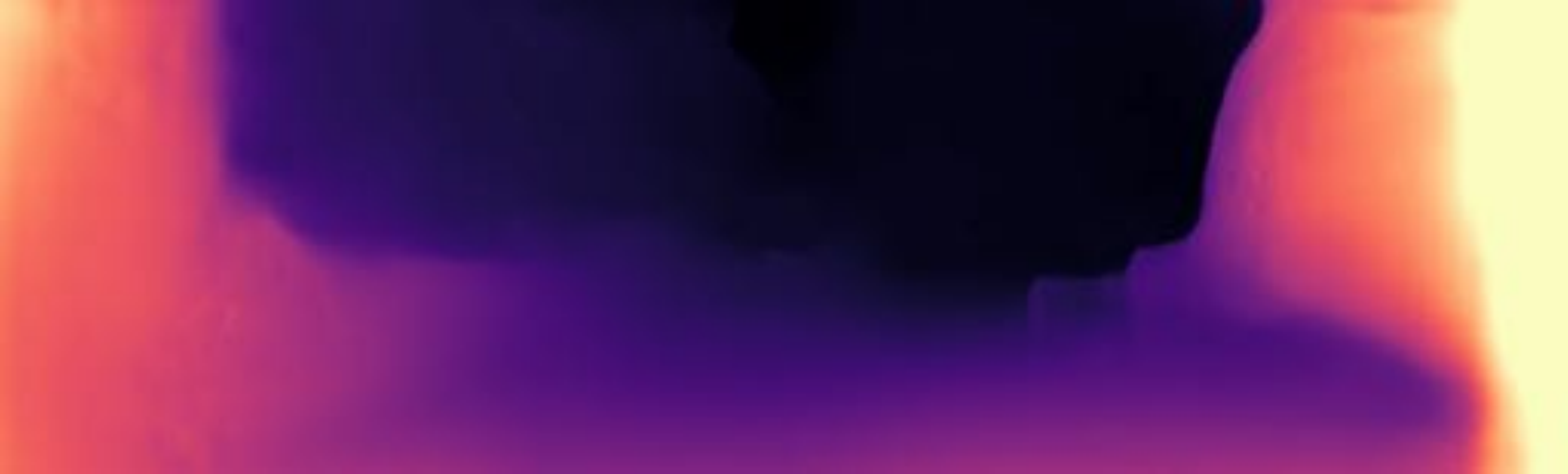}&  
        \includegraphics[]{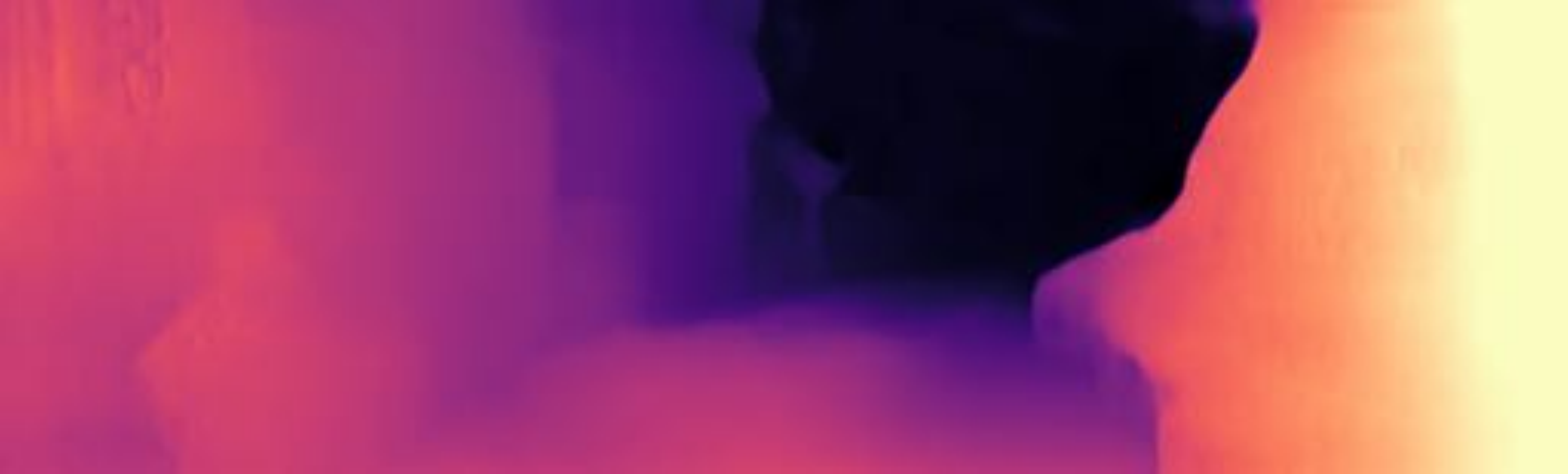}&  
        \includegraphics[]{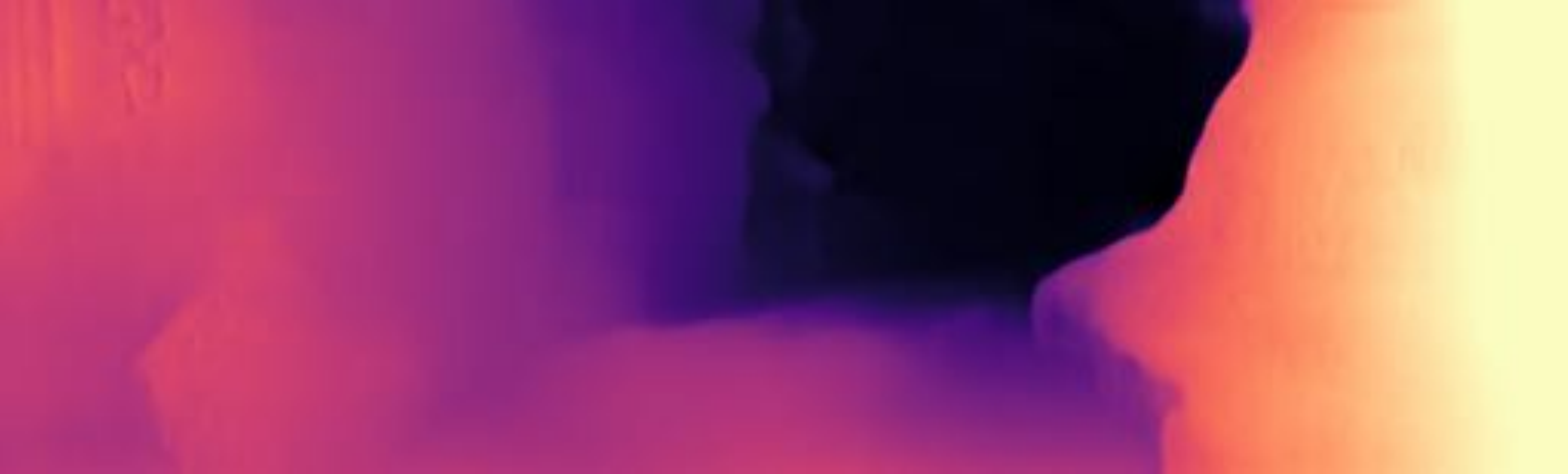}&  
        \includegraphics[]{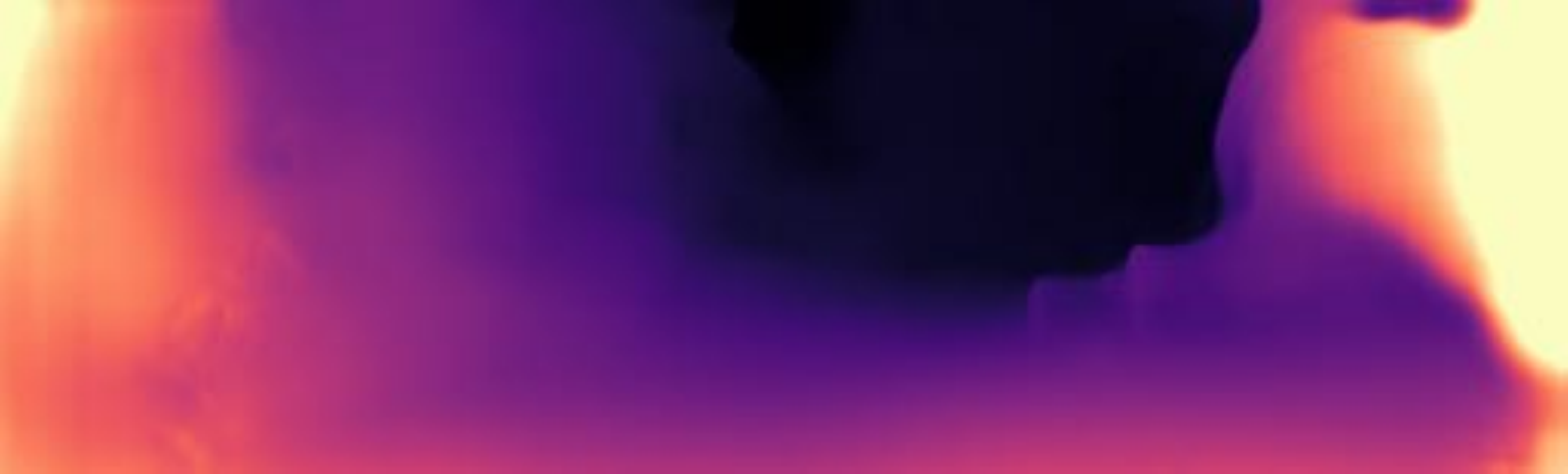}\\
        \includegraphics[]{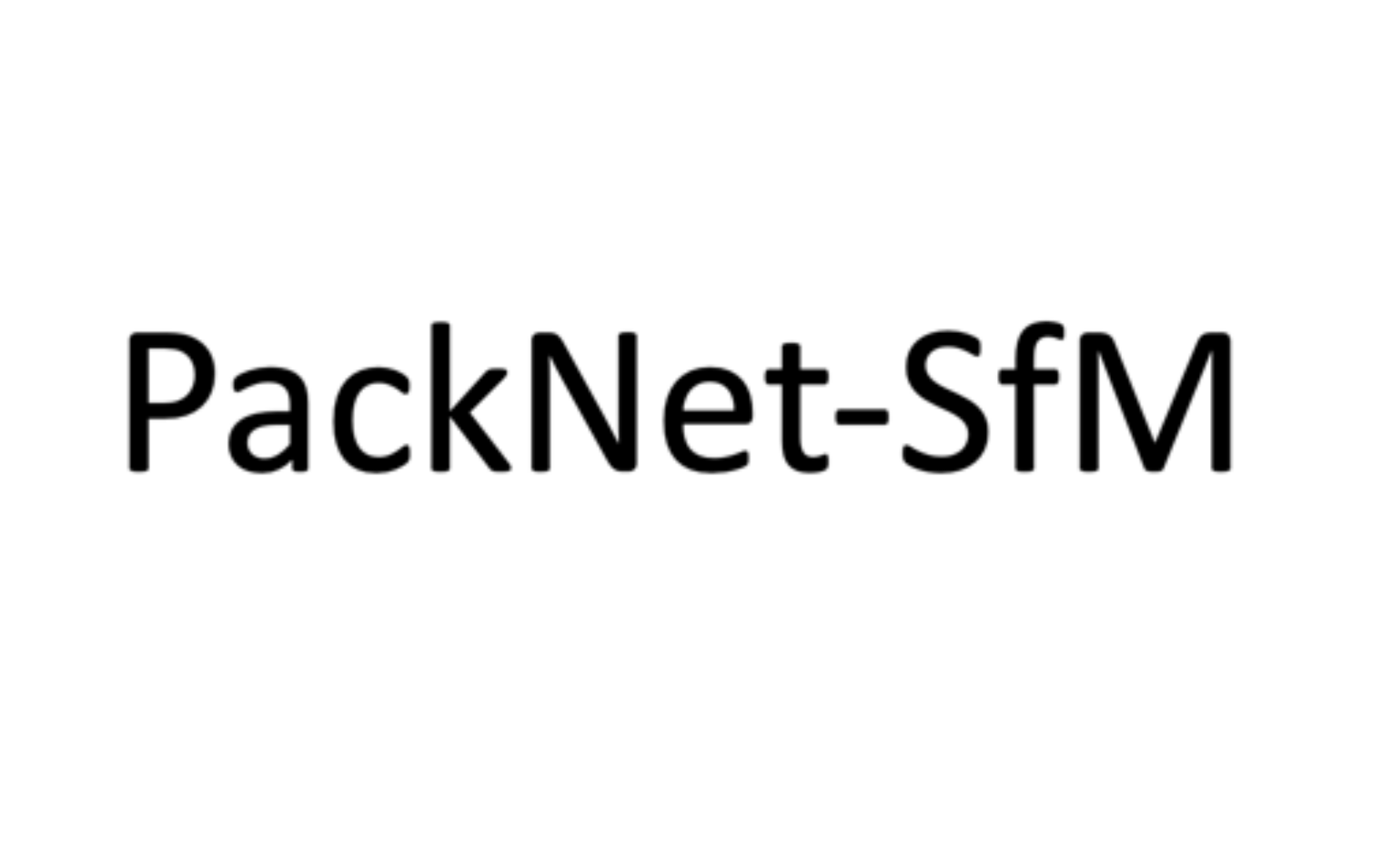} &  
        \includegraphics[]{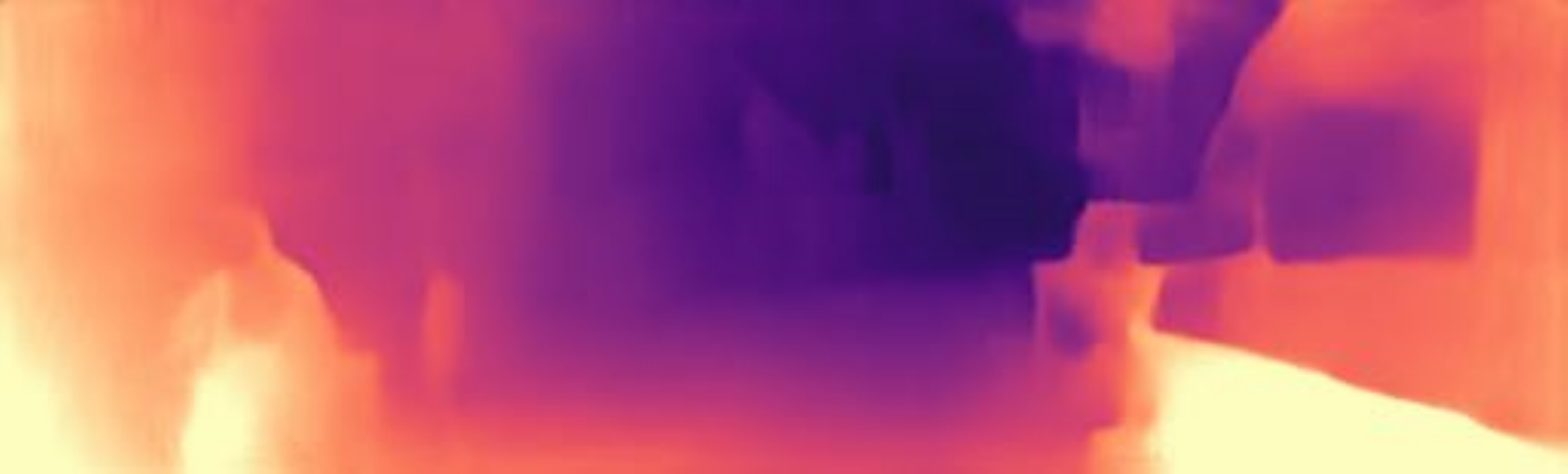}&  
        \includegraphics[]{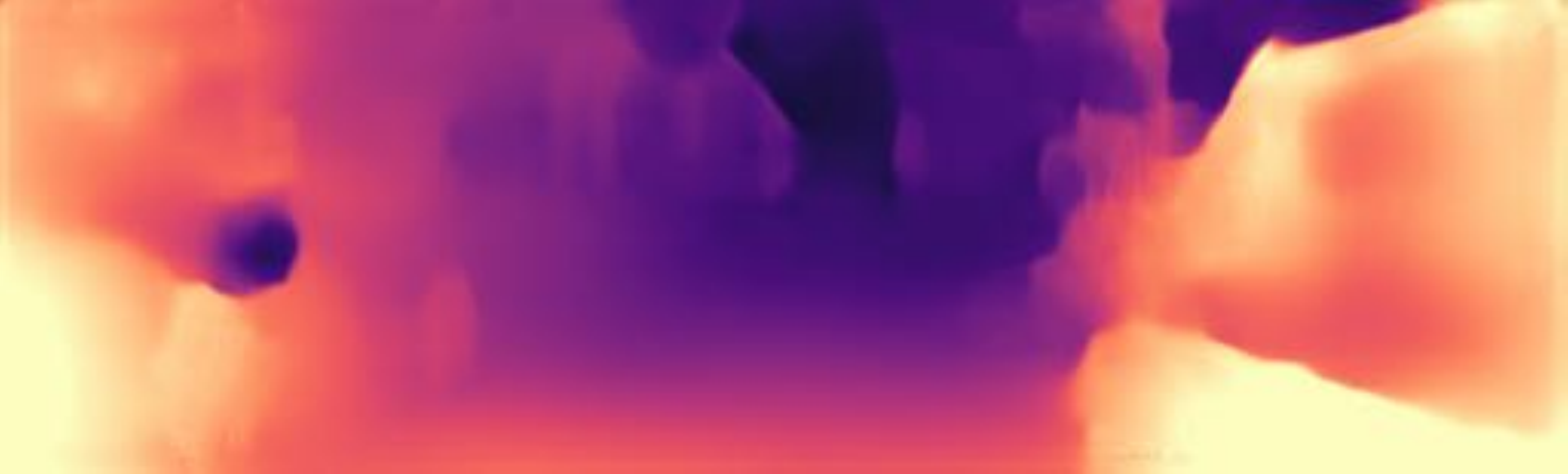}&  
        \includegraphics[]{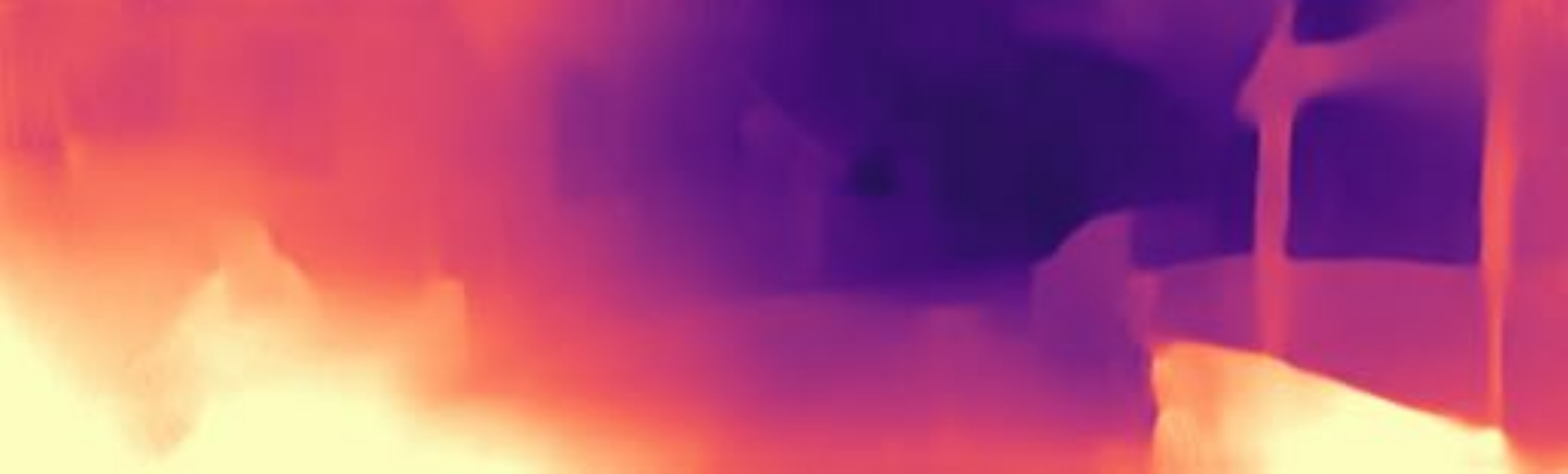}&  
        \includegraphics[]{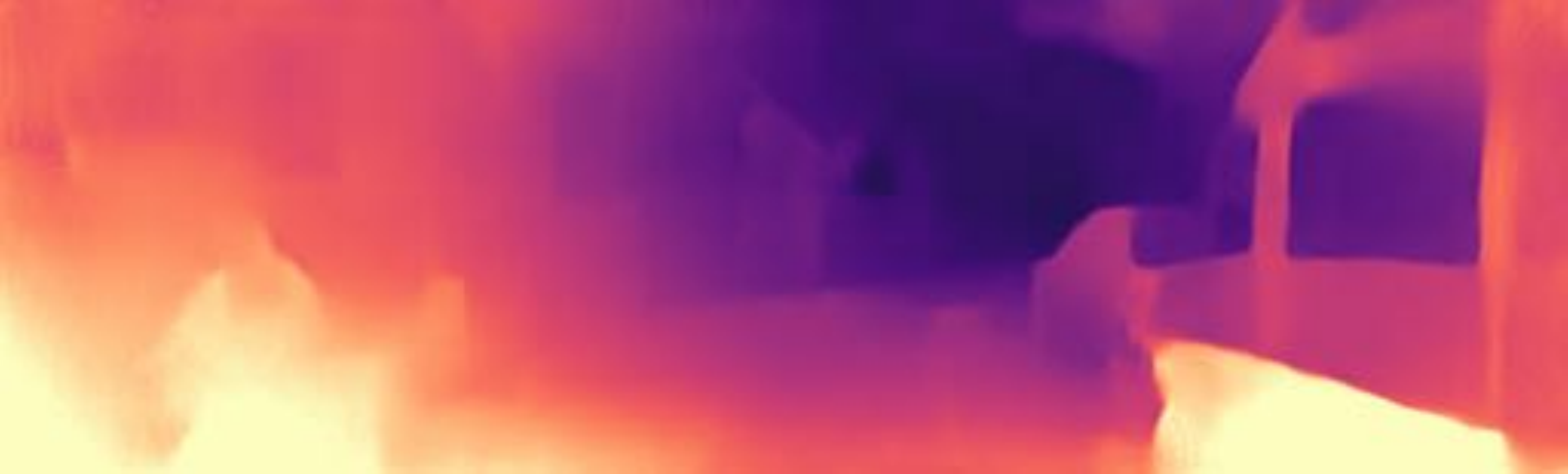}&  
        \includegraphics[]{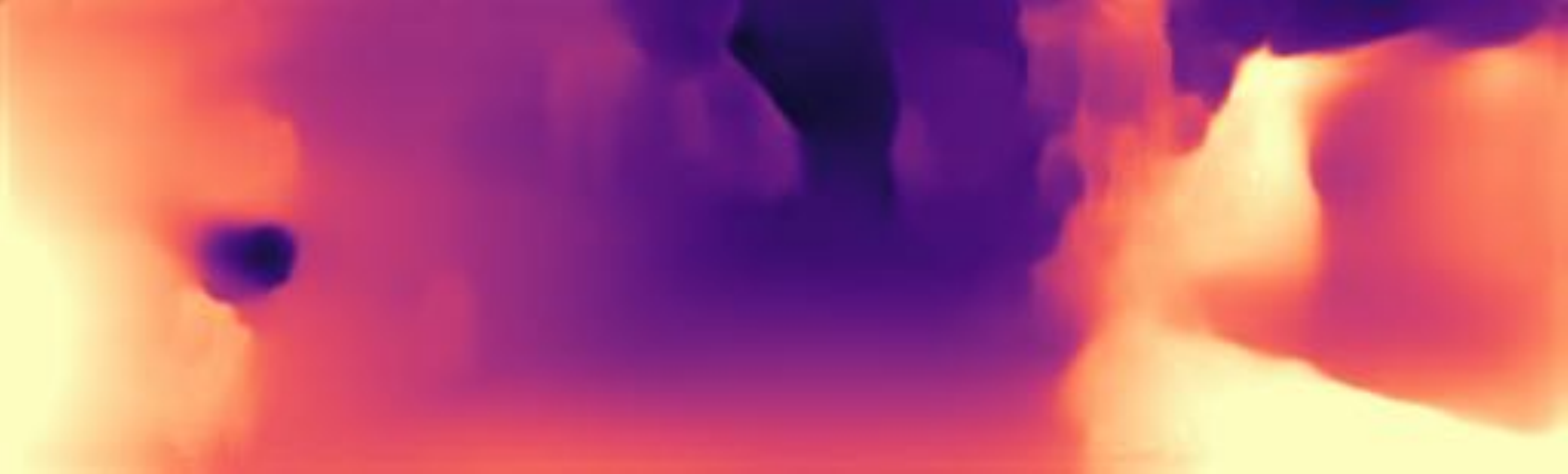}\\
        \includegraphics[]{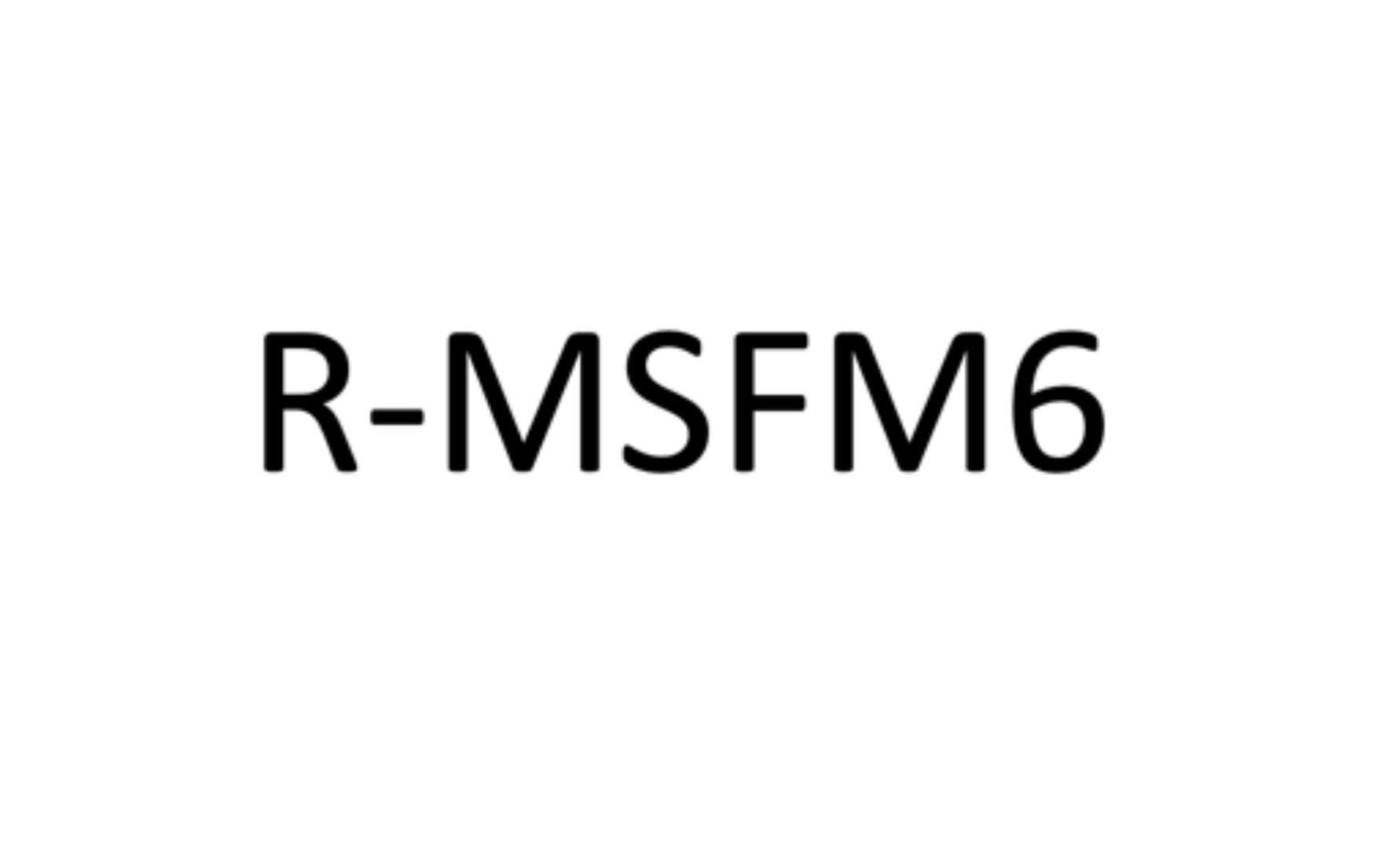}  &  
        \includegraphics[]{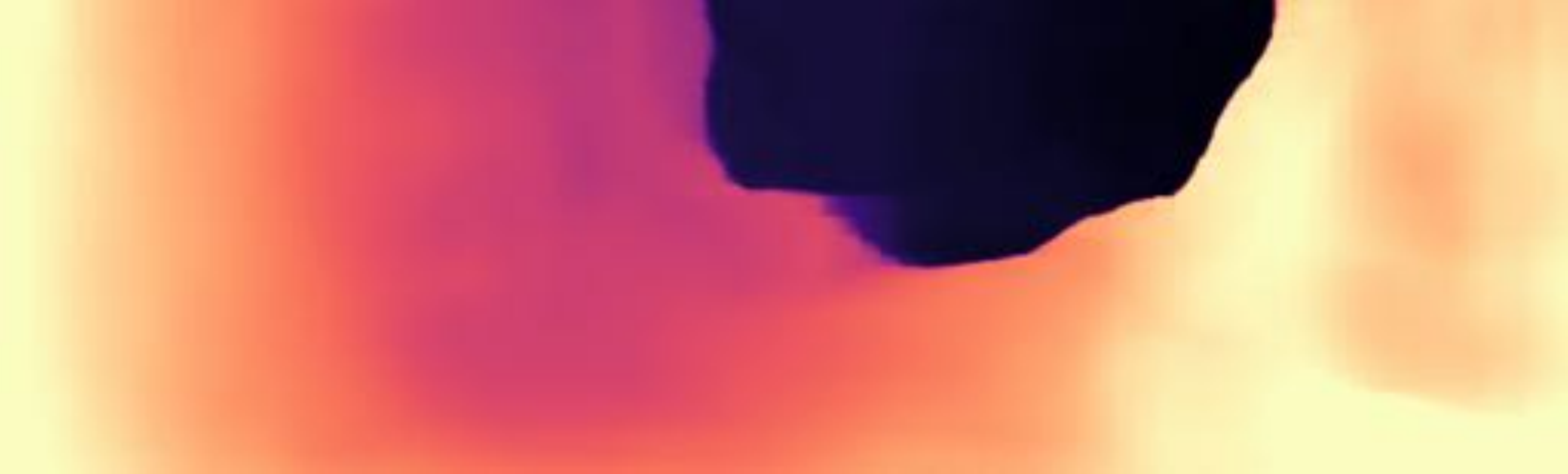}&  
        \includegraphics[]{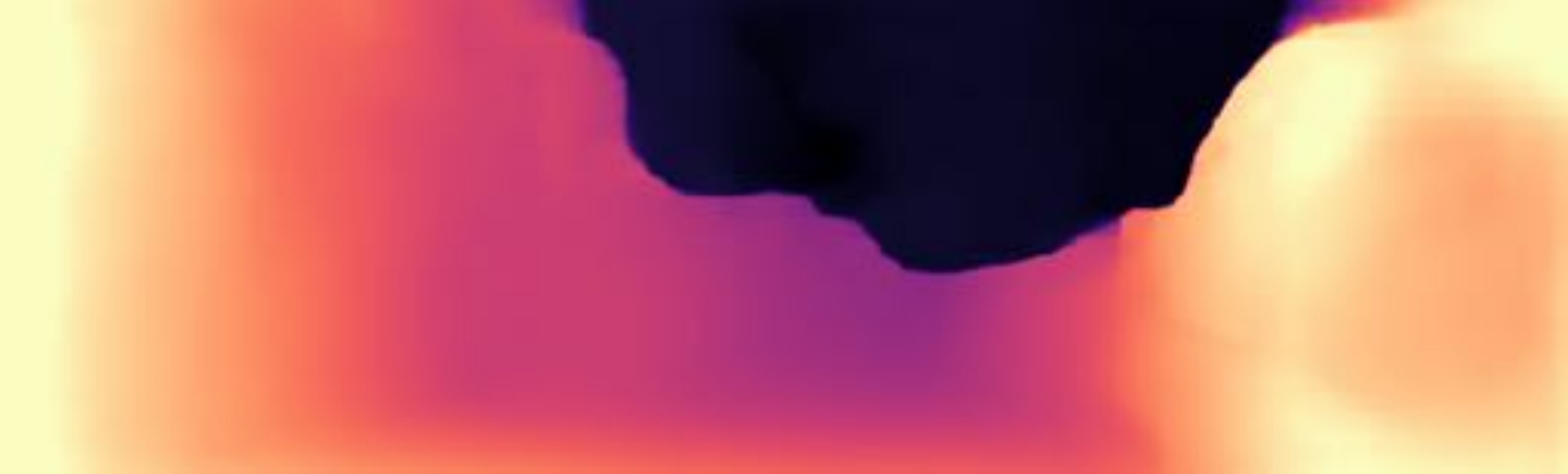}&  
        \includegraphics[]{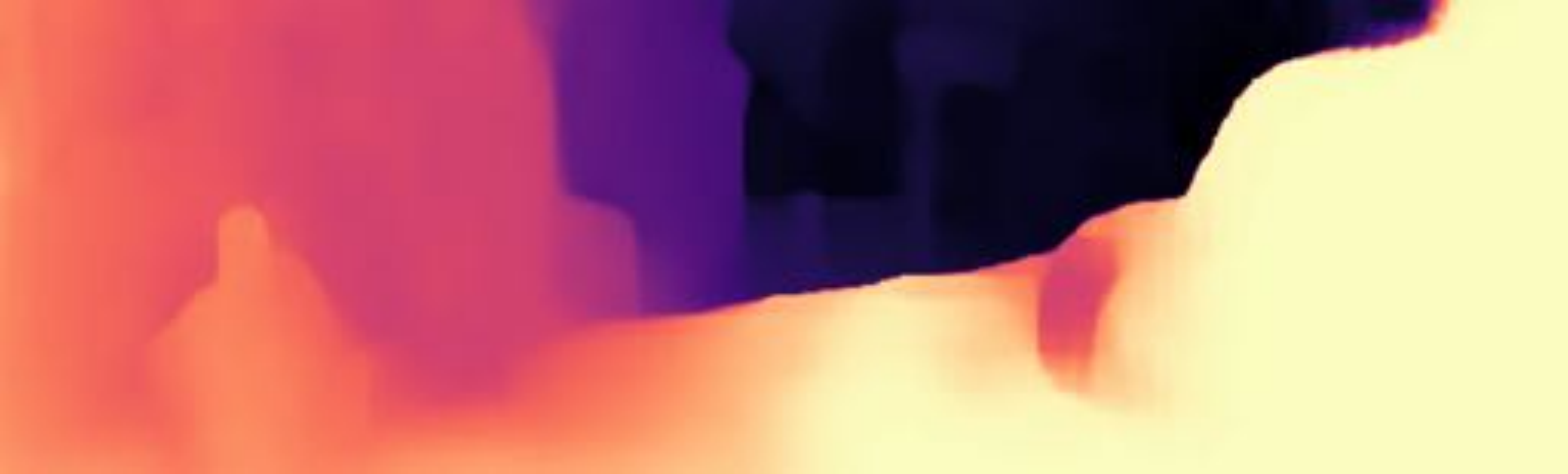}&  
        \includegraphics[]{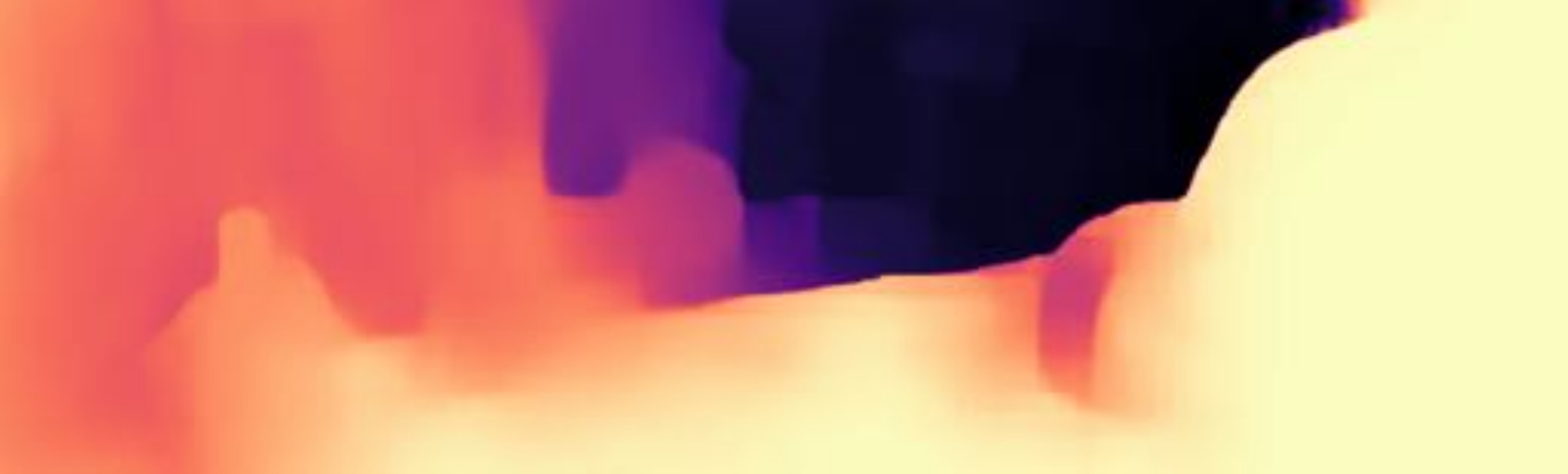}&  
        \includegraphics[]{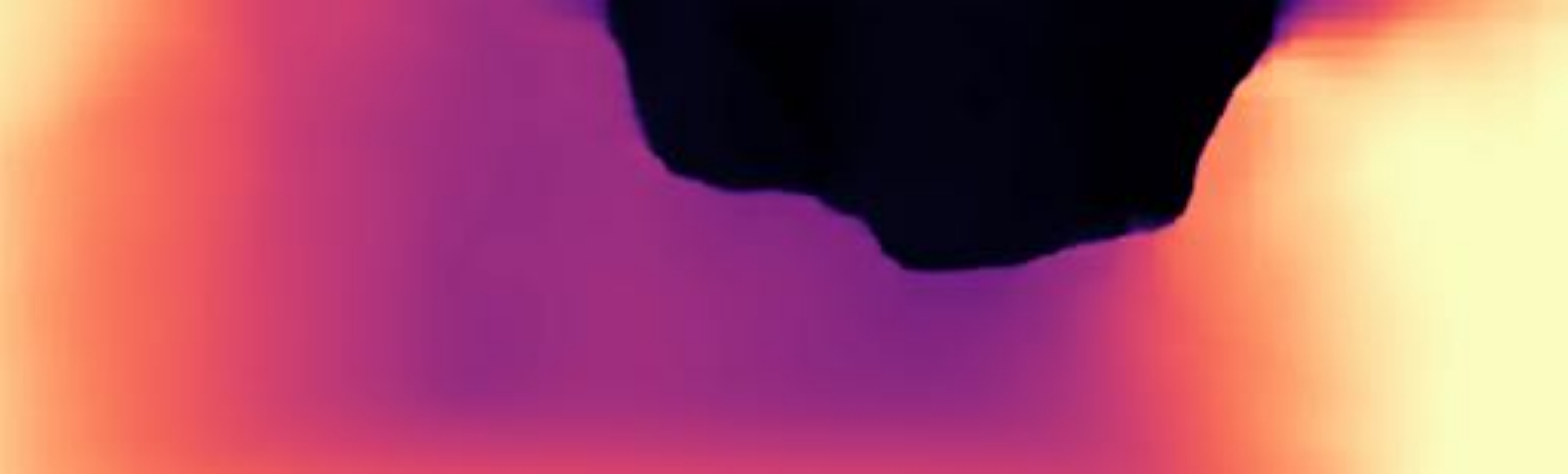}\\
        \includegraphics[]{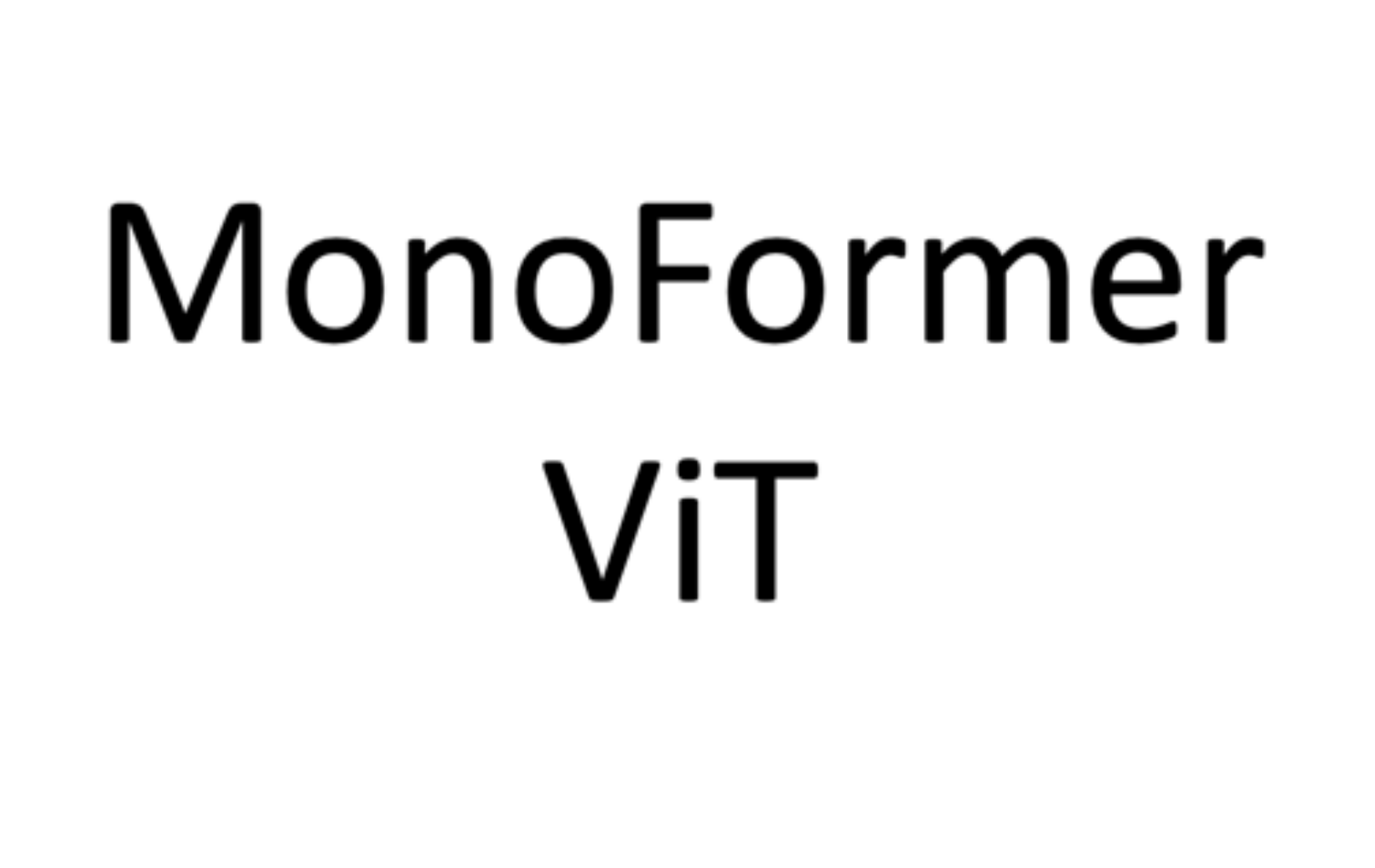} &  
        \includegraphics[]{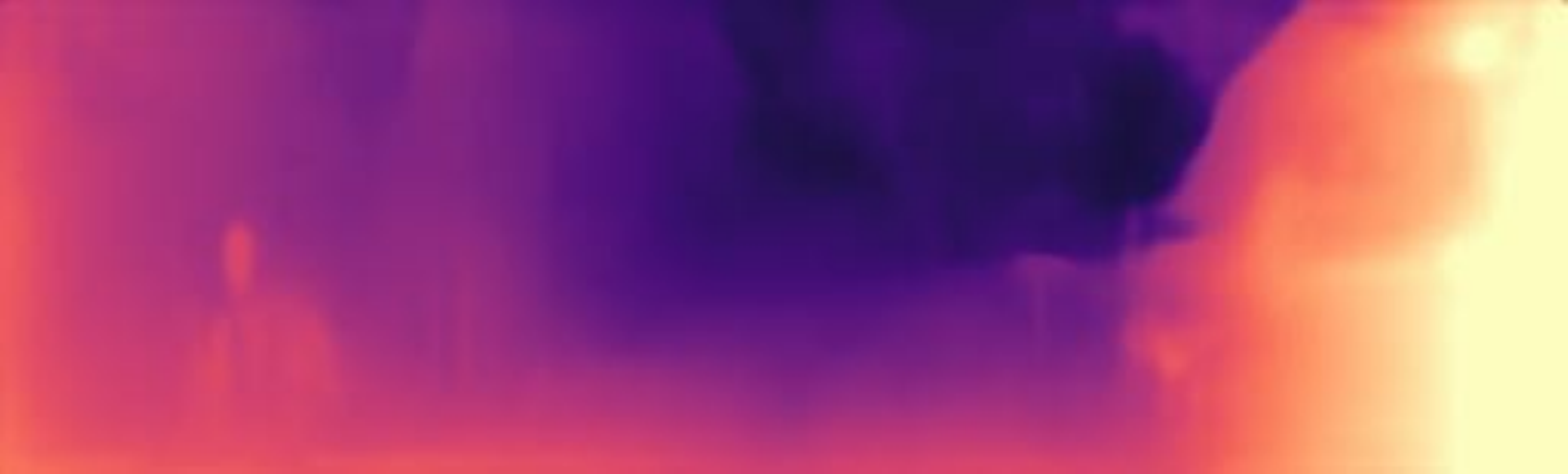}&  
        \includegraphics[]{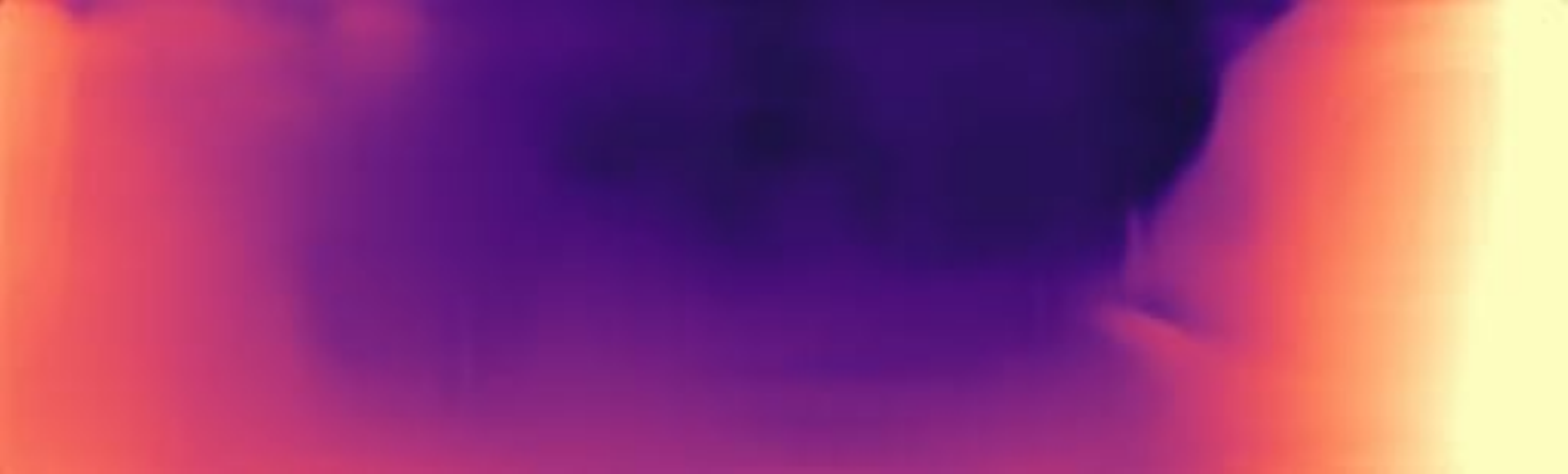}&  
        \includegraphics[]{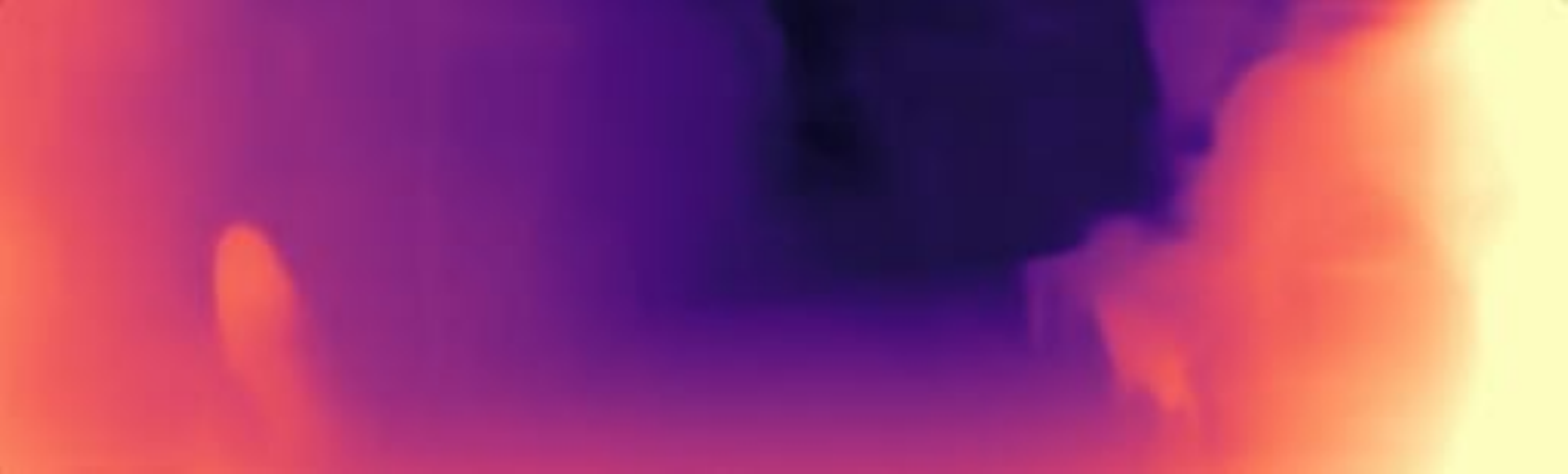}&  
        \includegraphics[]{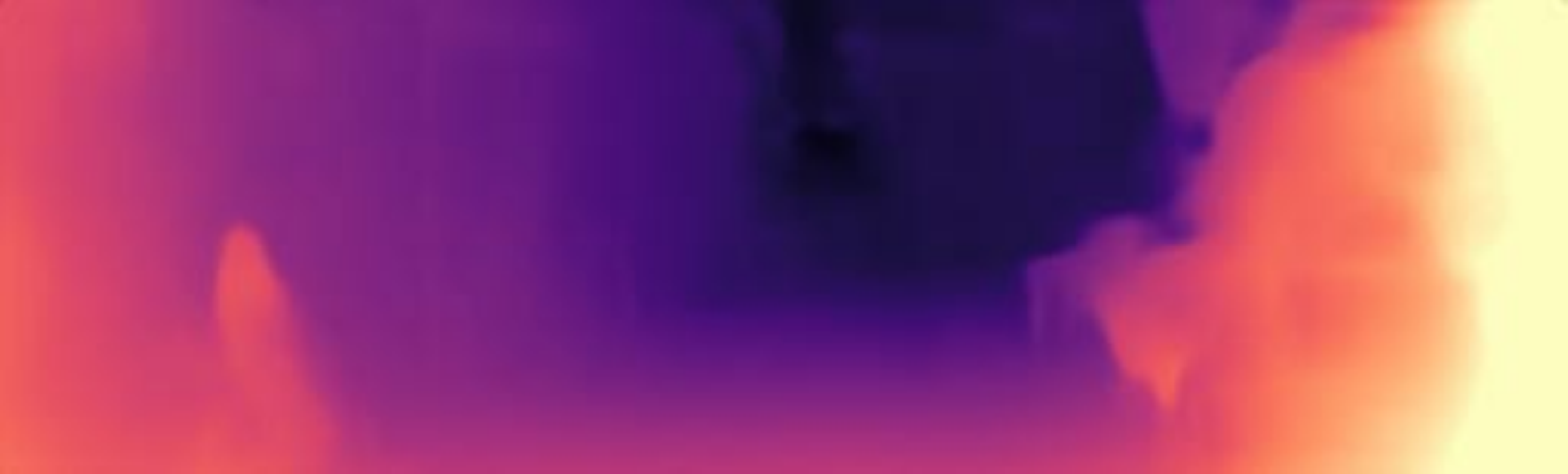}&  
        \includegraphics[]{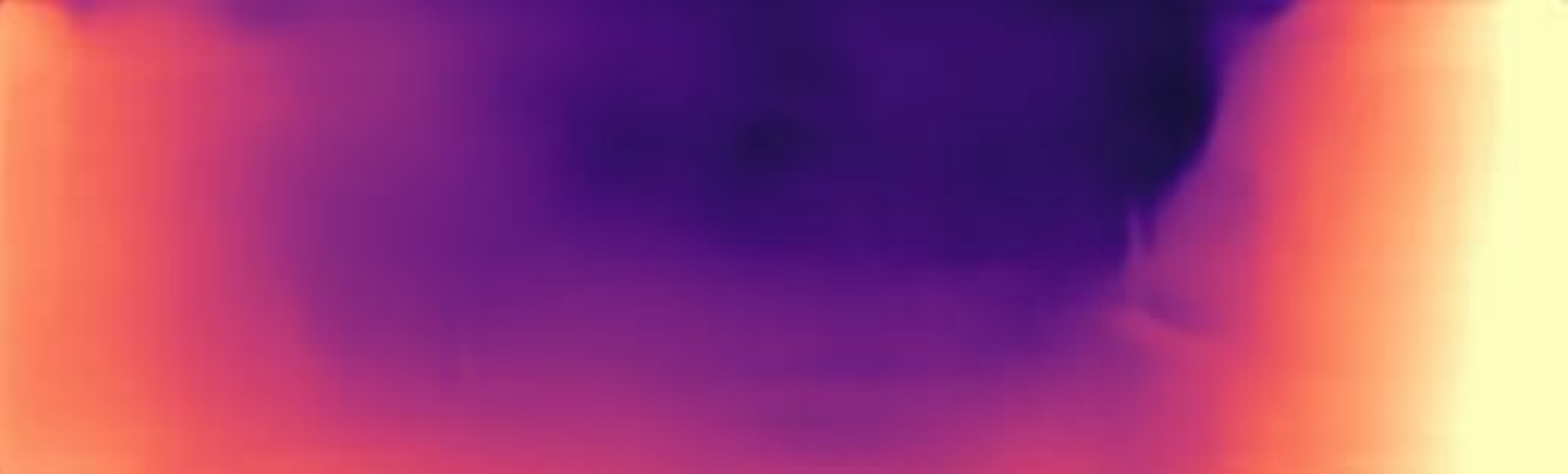}\\
        \includegraphics[]{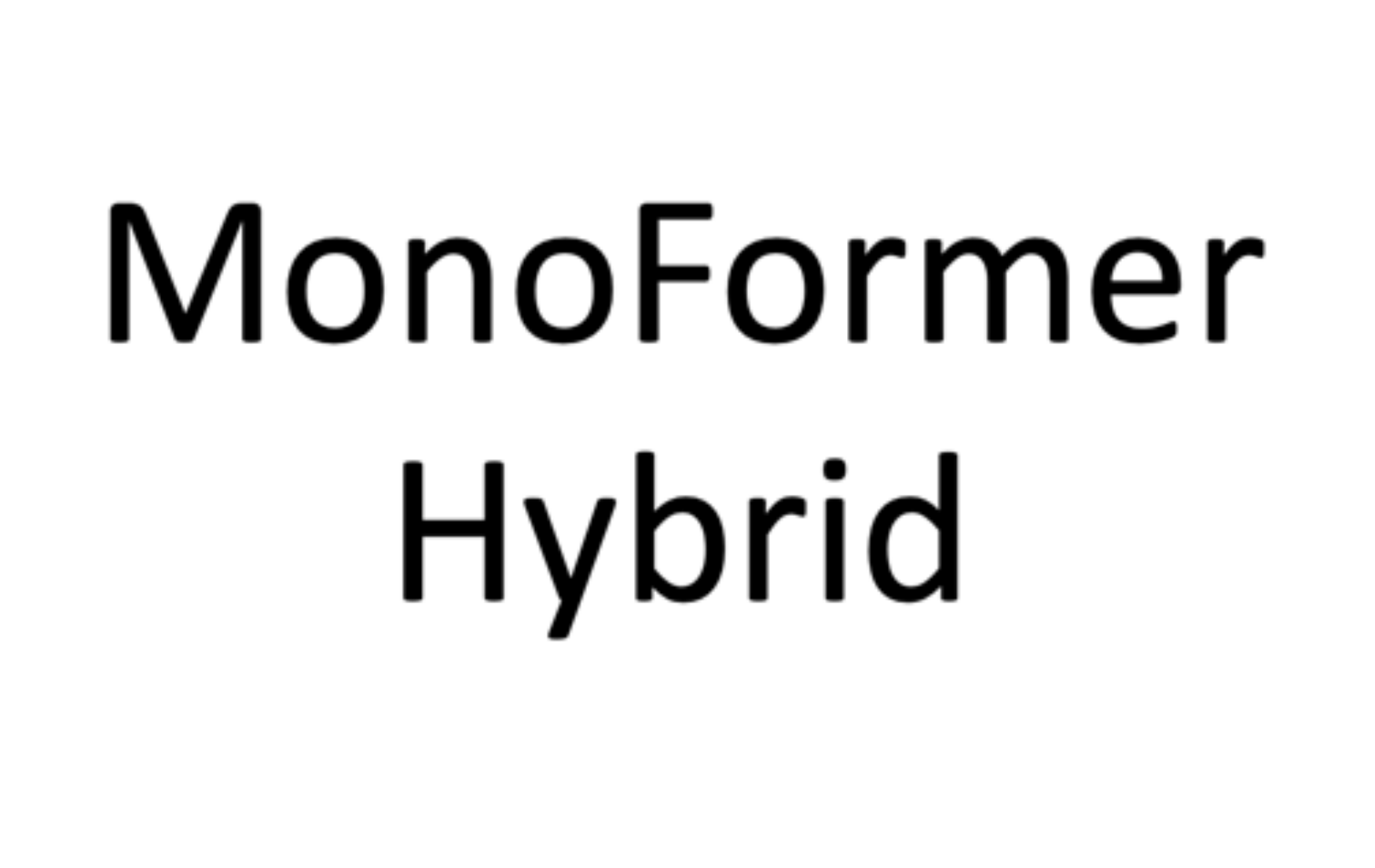} &  
        \includegraphics[]{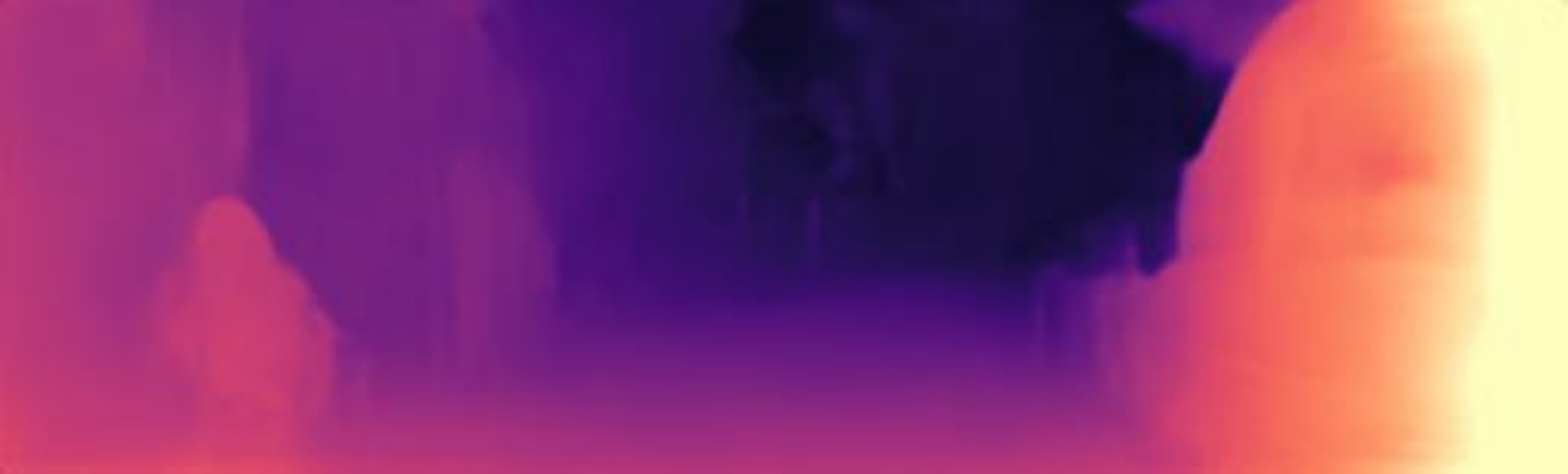}&  
        \includegraphics[]{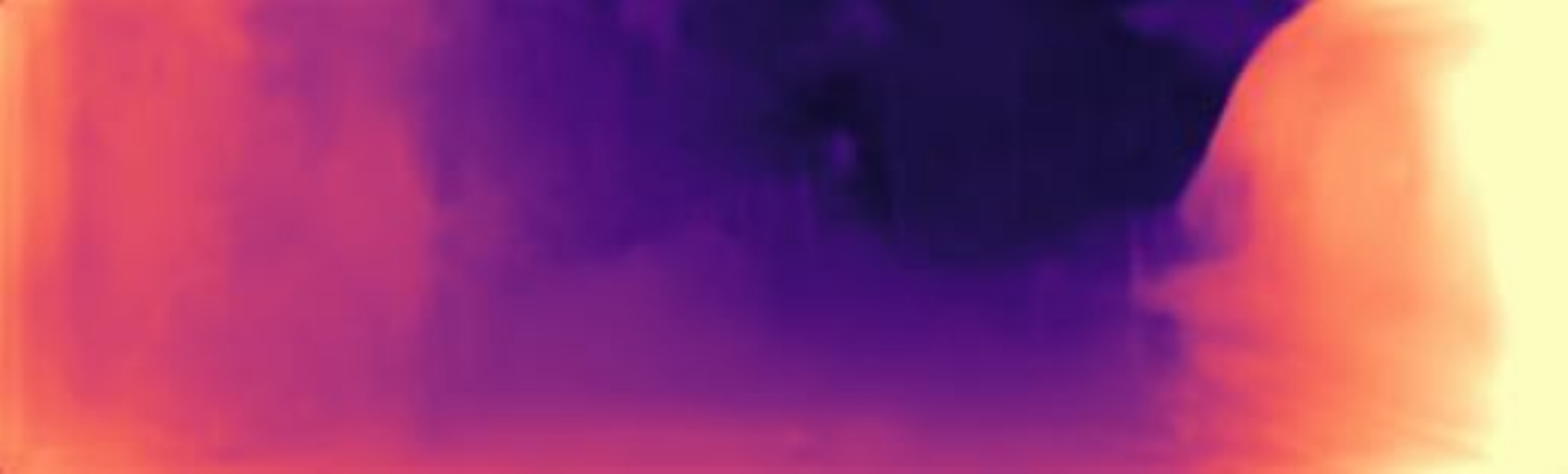}&  
        \includegraphics[]{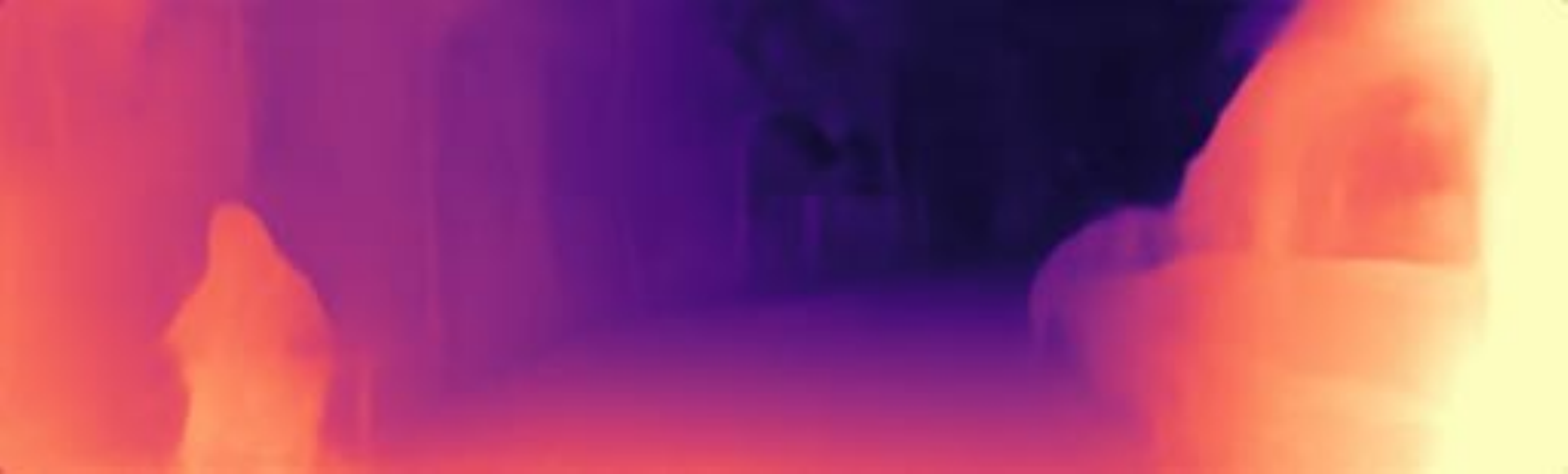}&  
        \includegraphics[]{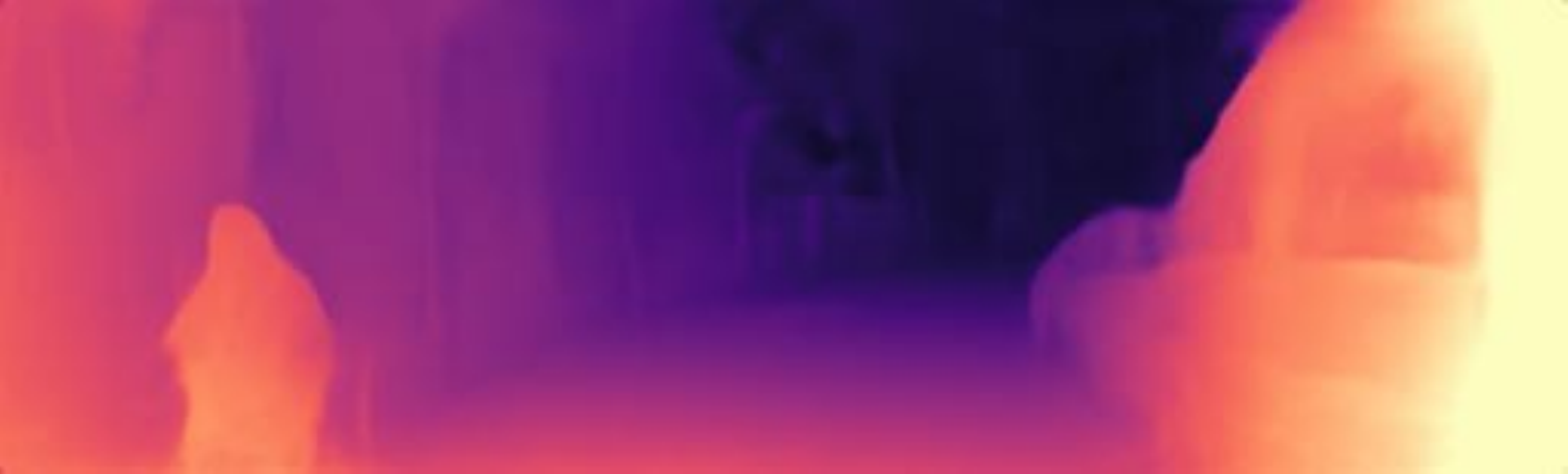}&  
        \includegraphics[]{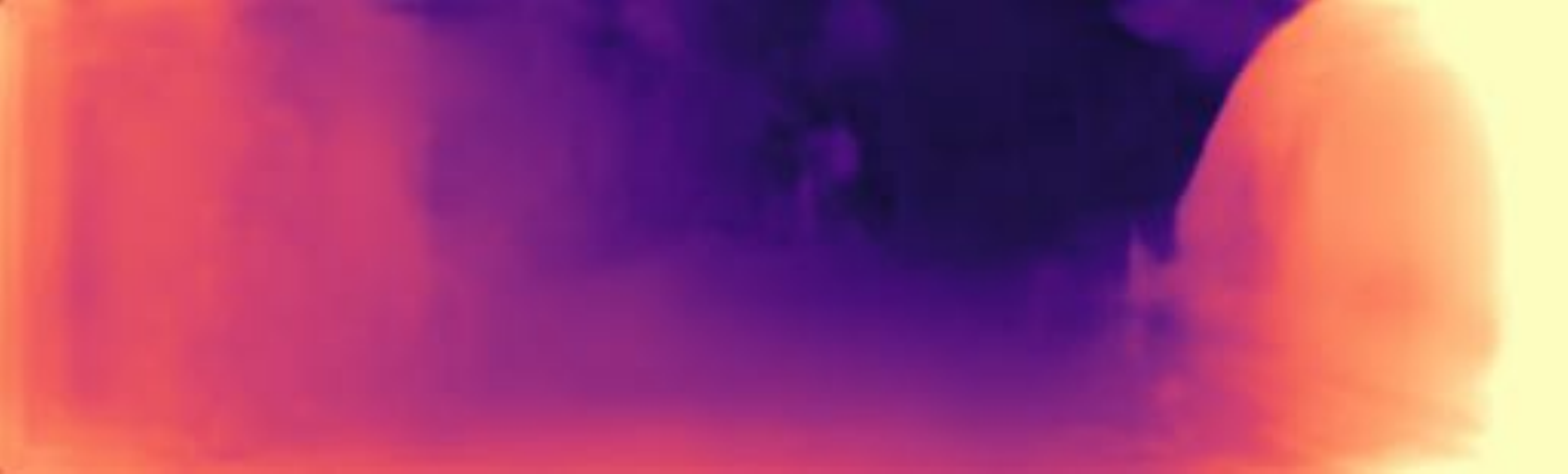}\\
        
        \end{tabular}
        }
    \caption{\textbf{Comparison of depth map results on various watercolor images.} The middle is the default of the OpenCV function. Based on the default, the left image changes $\sigma_r$, and the right image changes $\sigma_s$.}
    \label{watercolor_apdx}
    \end{subfigure}

    \begin{subfigure}
       \centering
        \resizebox{\linewidth}{!}{%
            \begin{tabular}{cccccc}
    &
                        \fontsize{120}{100} $\sigma_r = 10 , \sigma_s = 0.7$&  
                        \fontsize{120}{100} $\sigma_r = 190 , \sigma_s = 0.7$&  
                        \fontsize{120}{100}  $\sigma_r = 60 , \sigma_s = 0.7$ \selectfont (default) &  
                        \fontsize{120}{100}   $\sigma_r = 60 , \sigma_s = 0.1$&  
                        \fontsize{120}{100}    $\sigma_r = 60 , \sigma_s = 0.9$\\
    \includegraphics[]{figure/appendix/supp/images.pdf}             &  \includegraphics[]{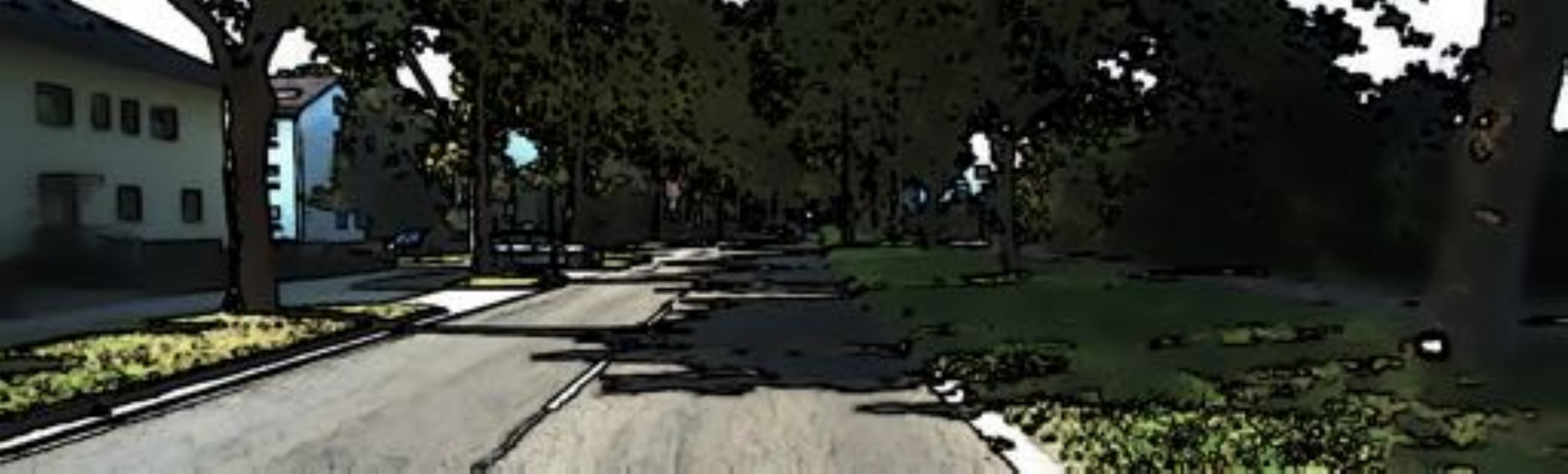}&  
    \includegraphics[]{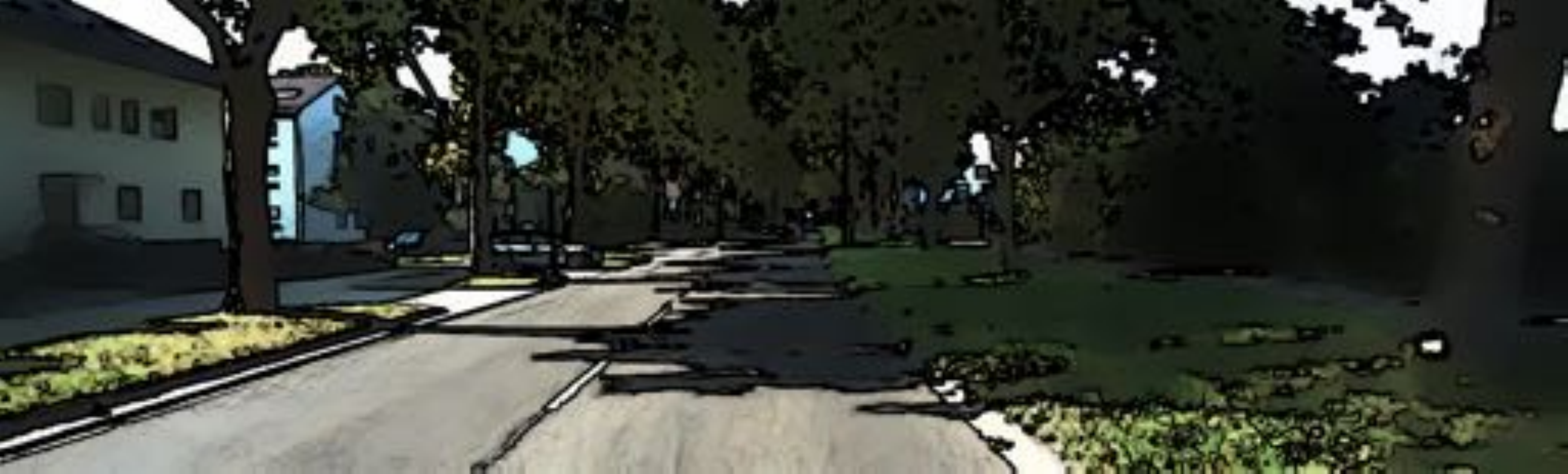}&  
    \includegraphics[]{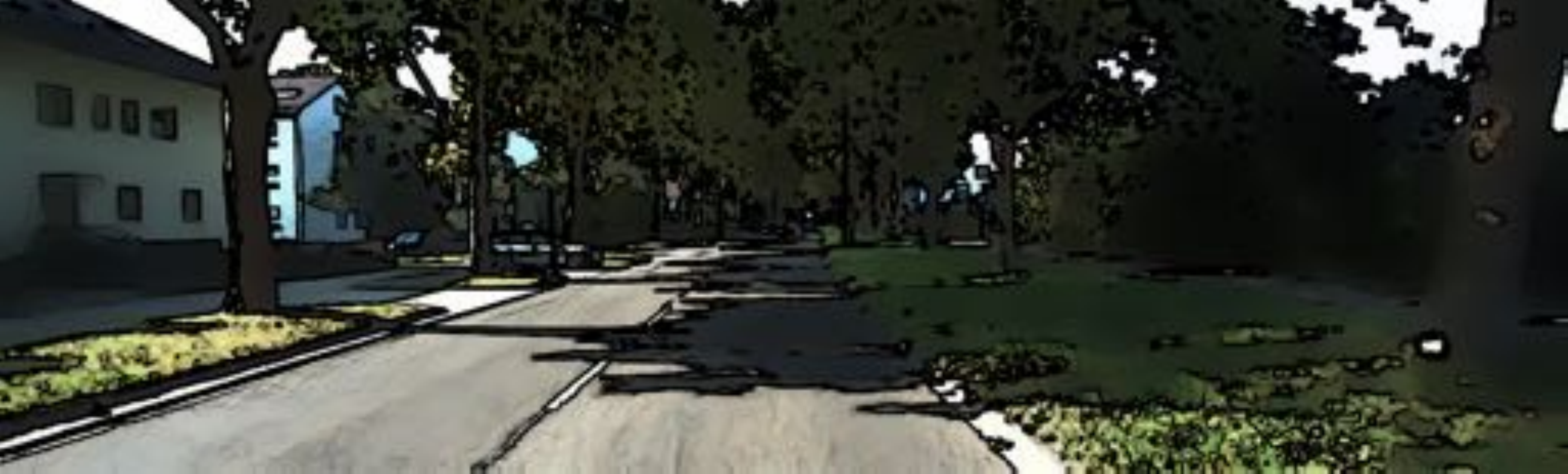}&  
    \includegraphics[]{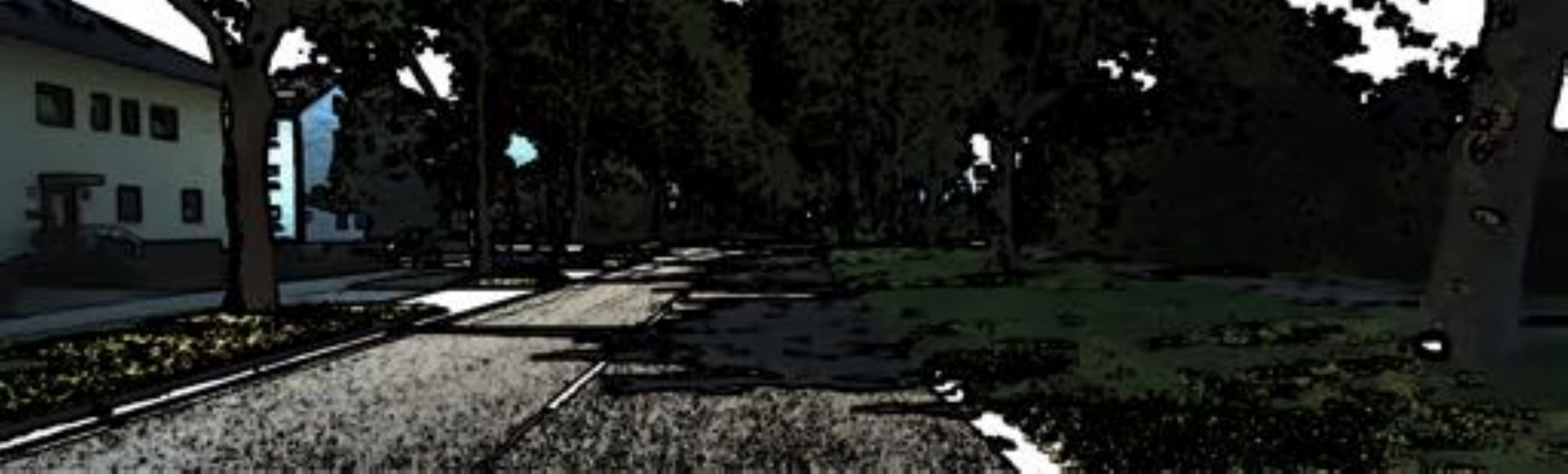}&  
    \includegraphics[]{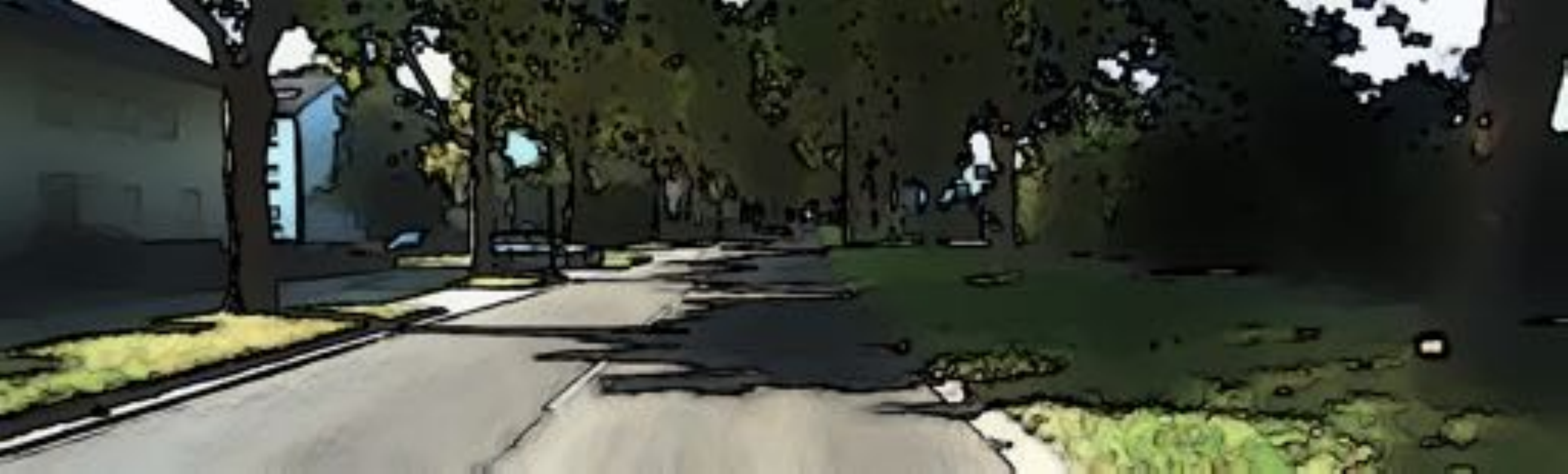}\\
    \includegraphics[]{figure/appendix/supp/monodpehth2.pdf}          &  \includegraphics[]{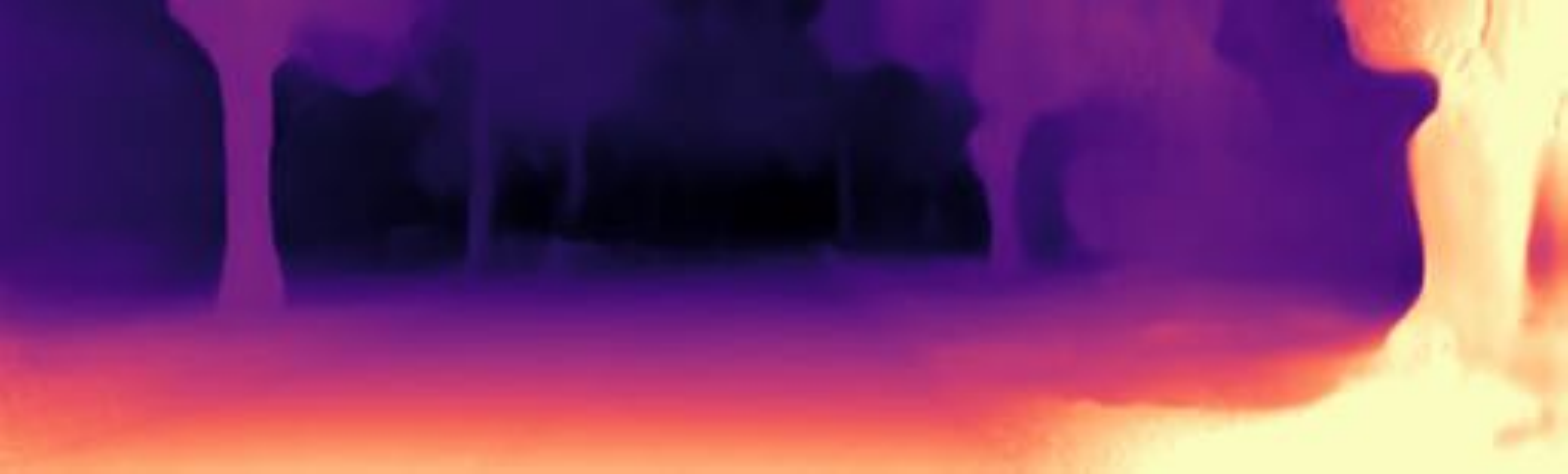}&  
    \includegraphics[]{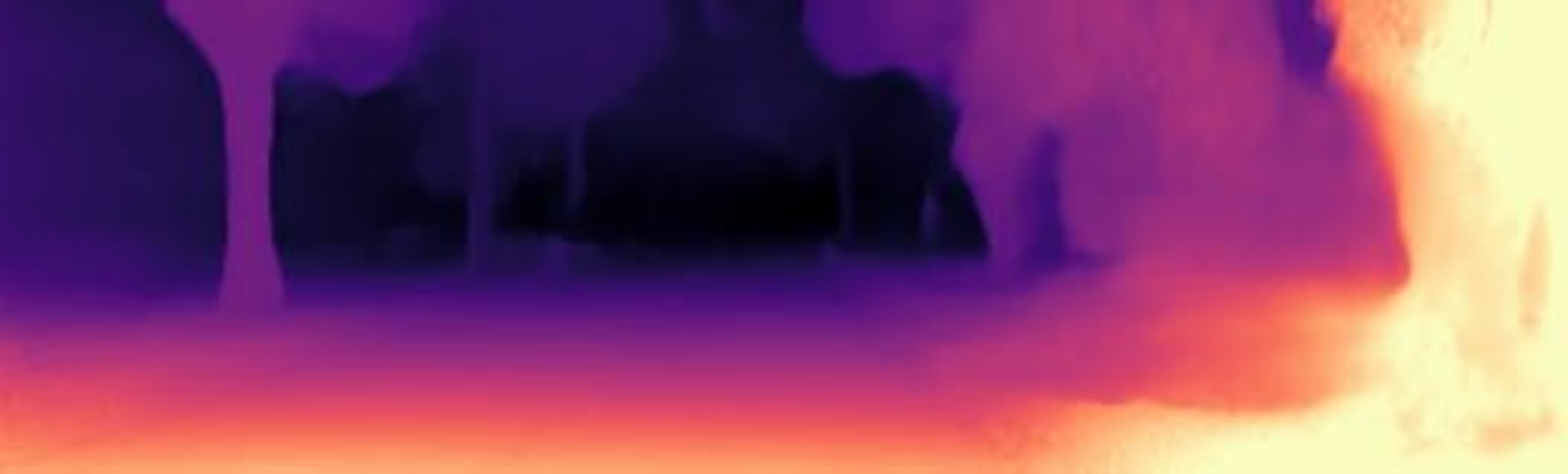}&  
    \includegraphics[]{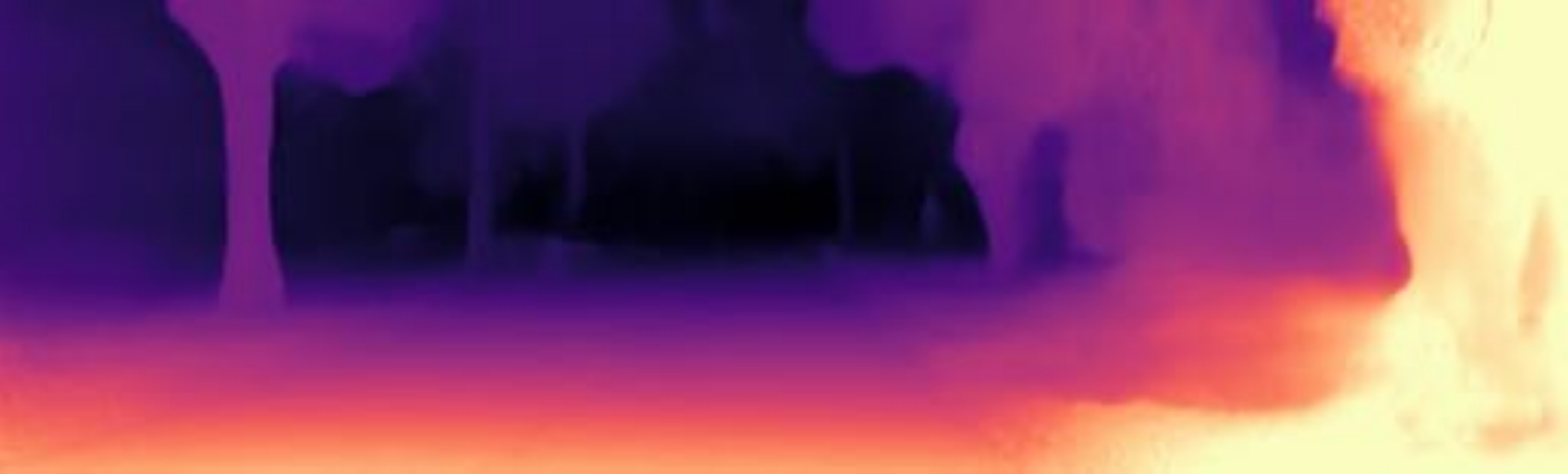}&  
    \includegraphics[]{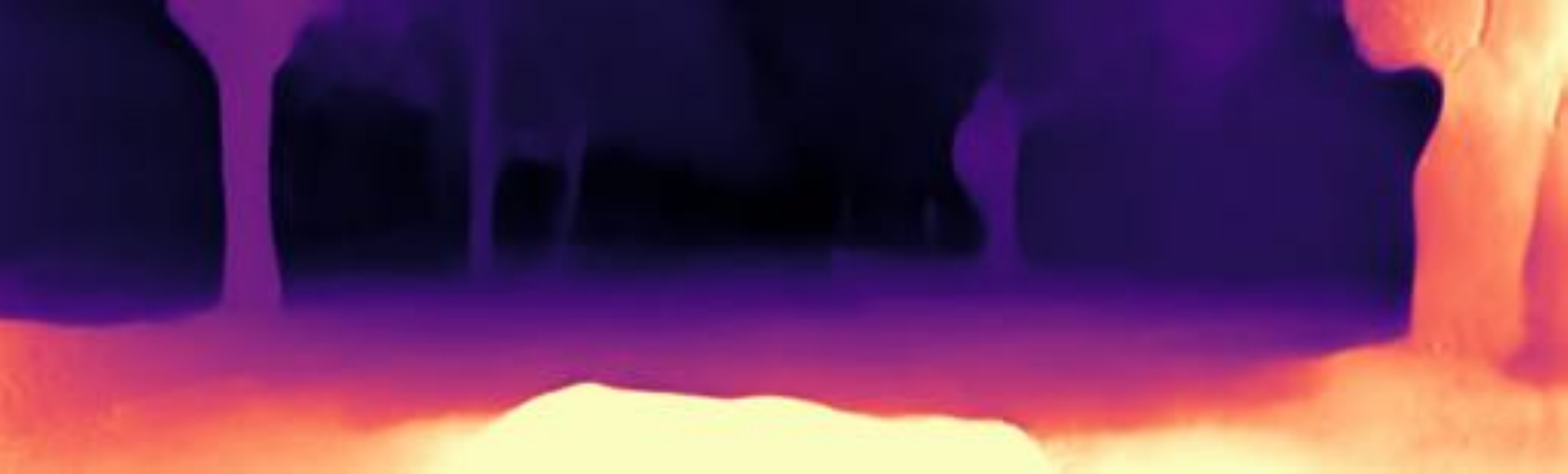}&  
    \includegraphics[]{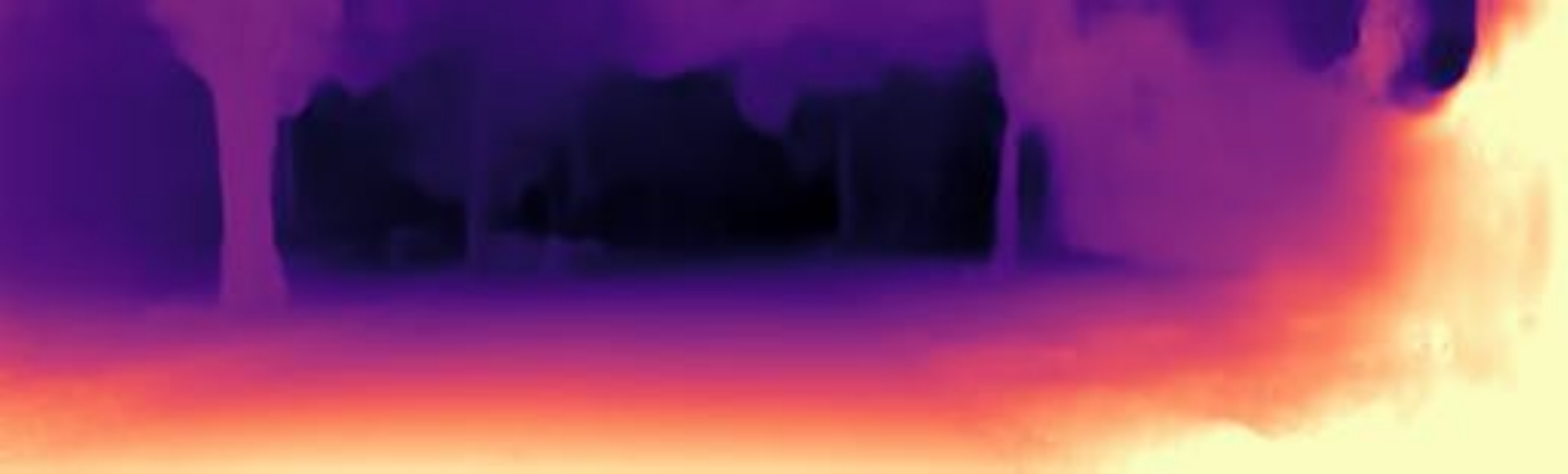}\\
    \includegraphics[]{figure/appendix/supp/packnet.pdf}         &  \includegraphics[]{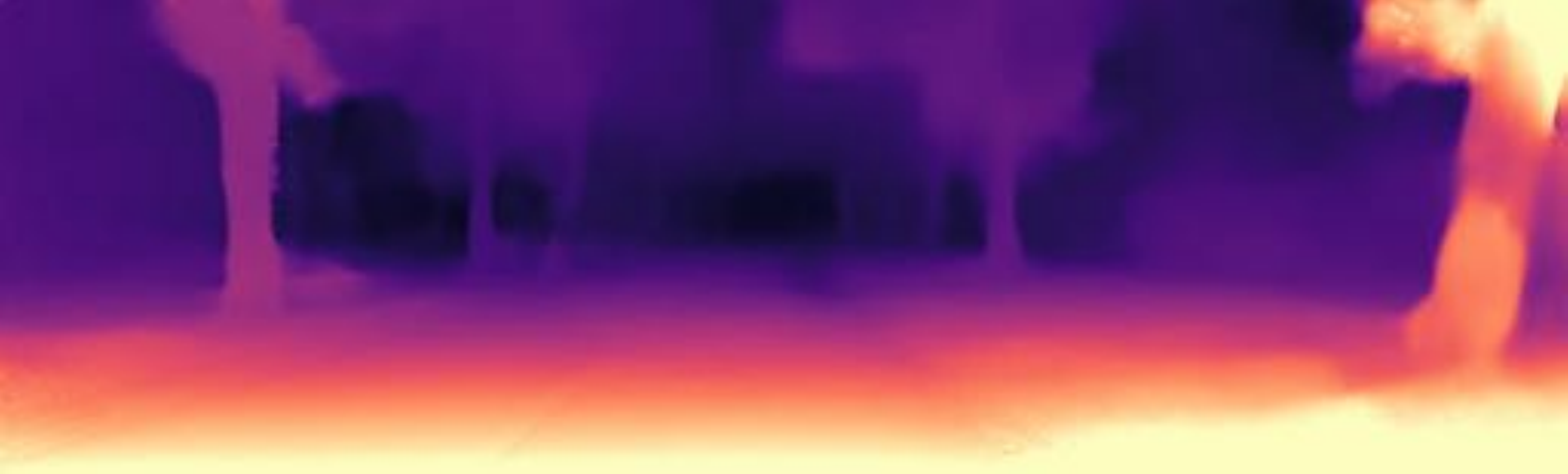}&  
    \includegraphics[]{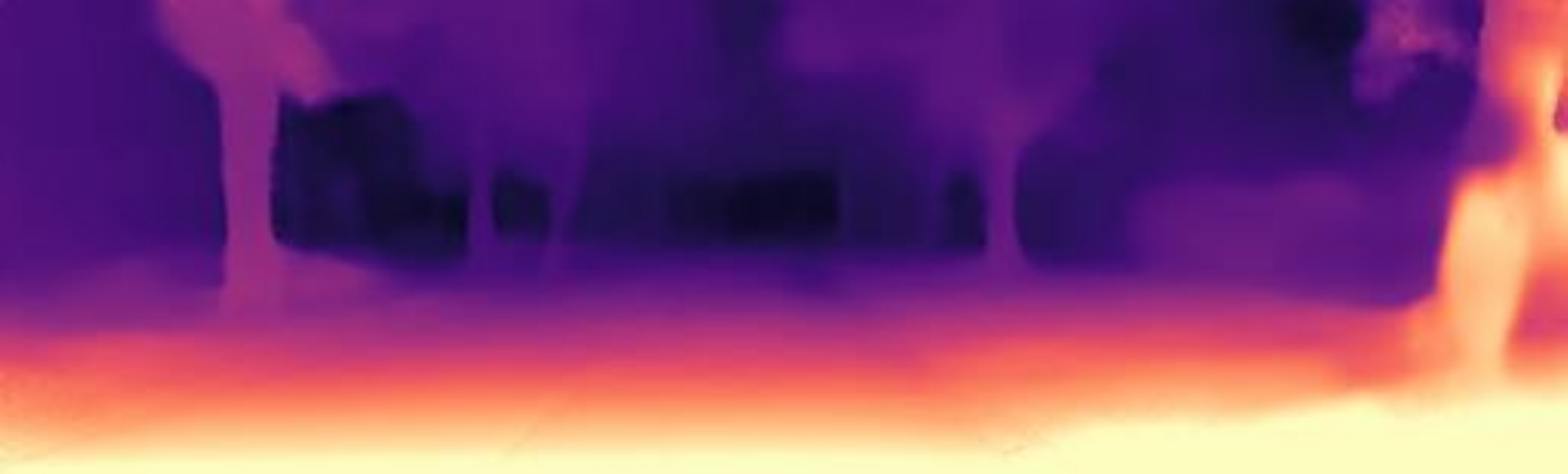}&  
    \includegraphics[]{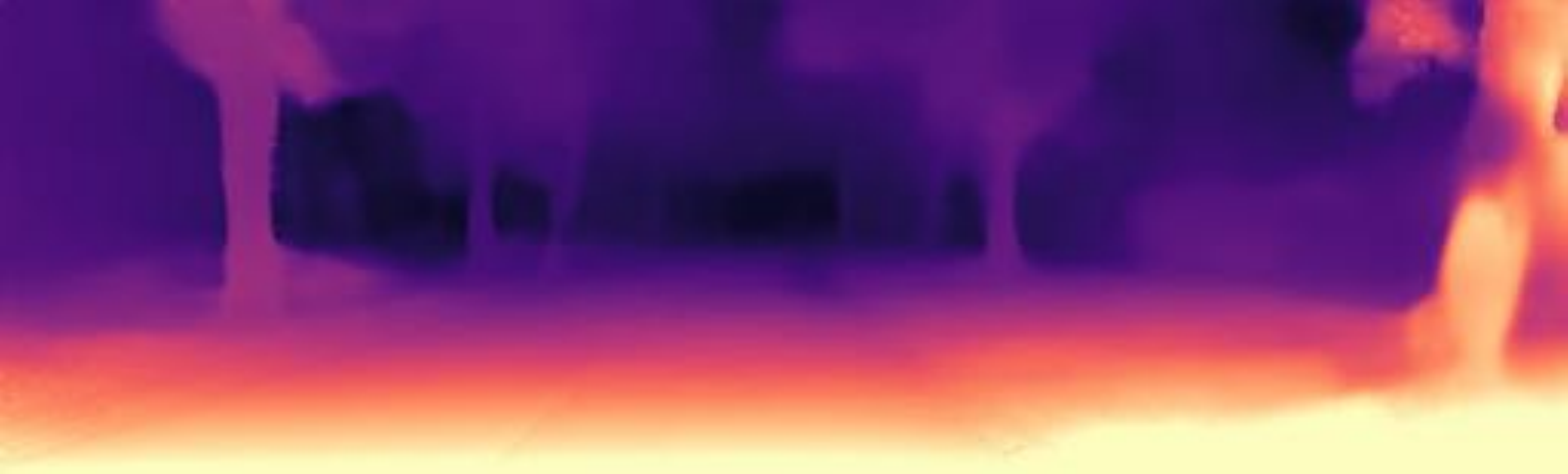}&  
    \includegraphics[]{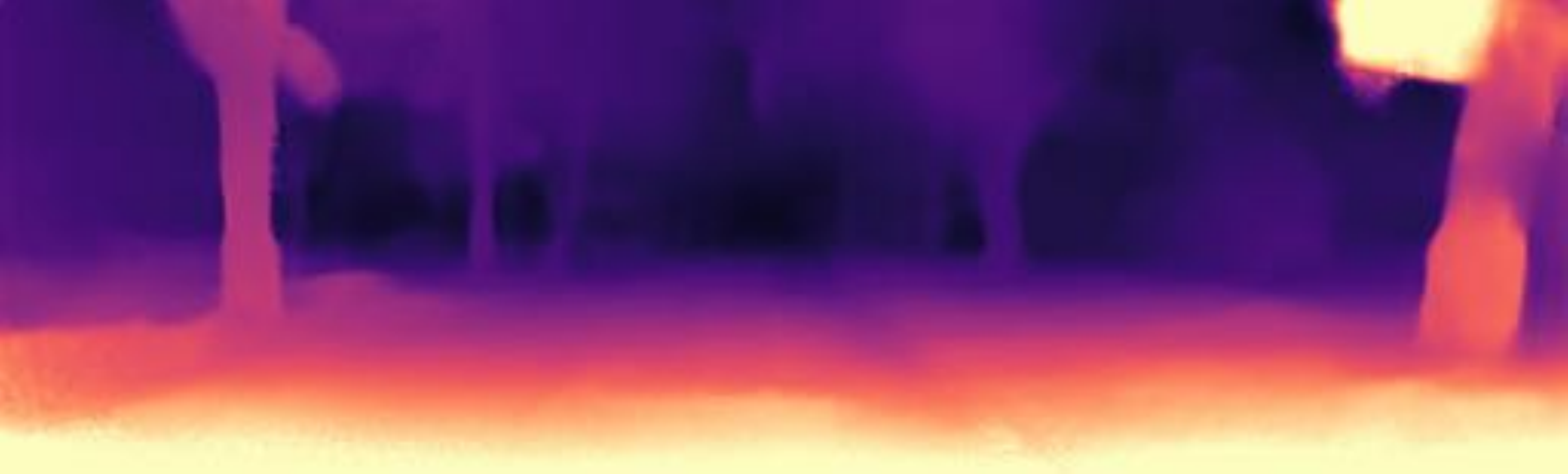}&  
    \includegraphics[]{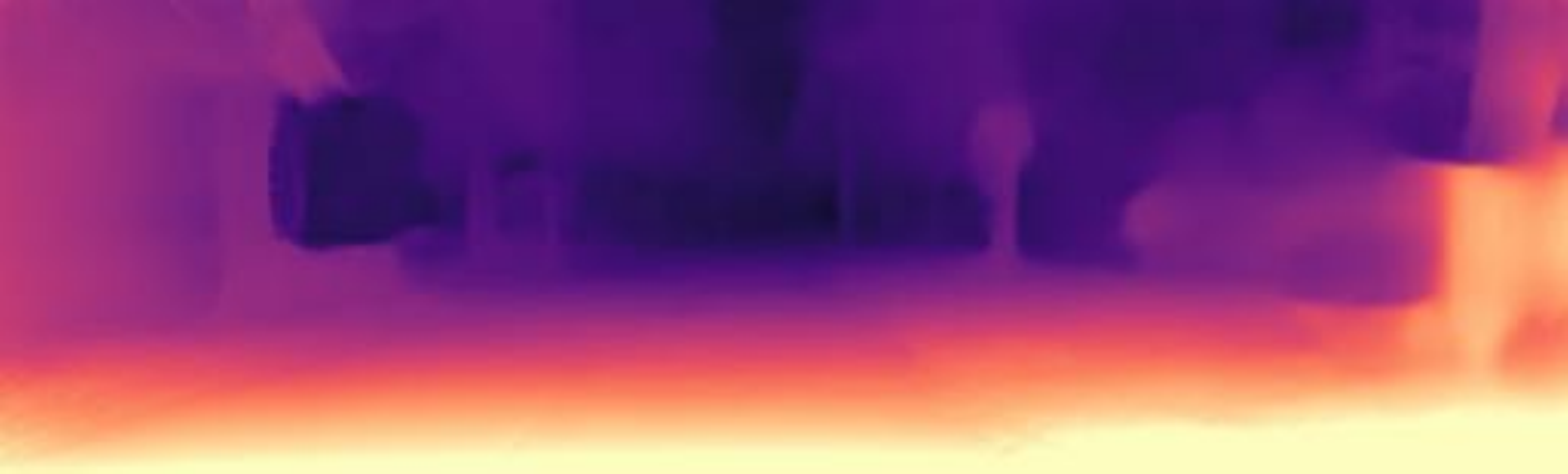}\\
    \includegraphics[]{figure/appendix/supp/rmsfm.pdf}             &  \includegraphics[]{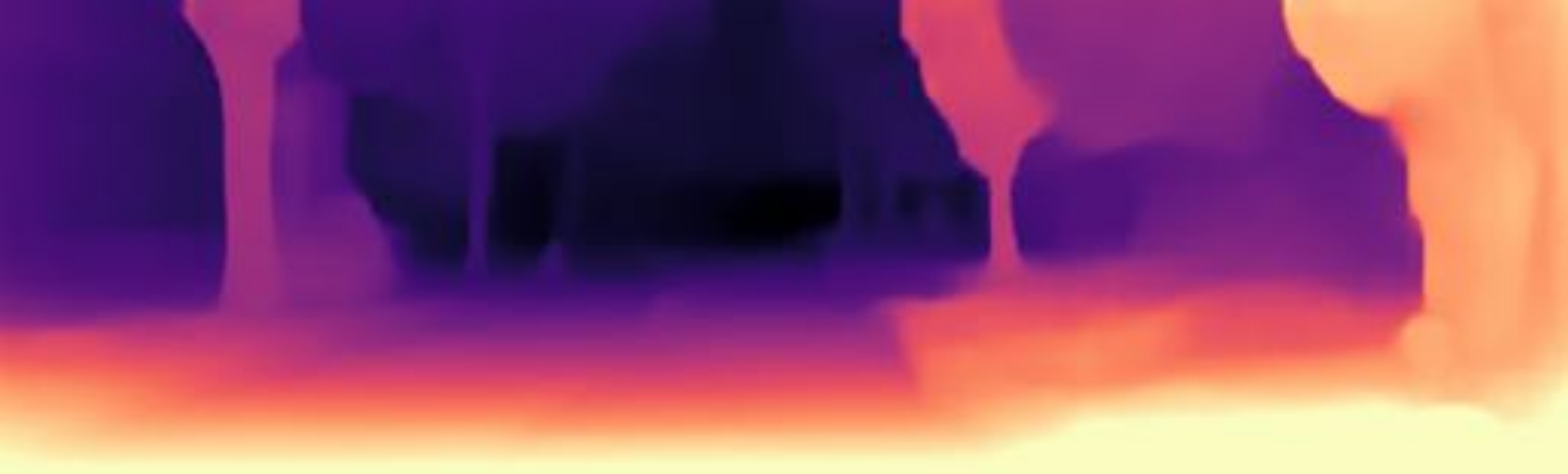}&  
    \includegraphics[]{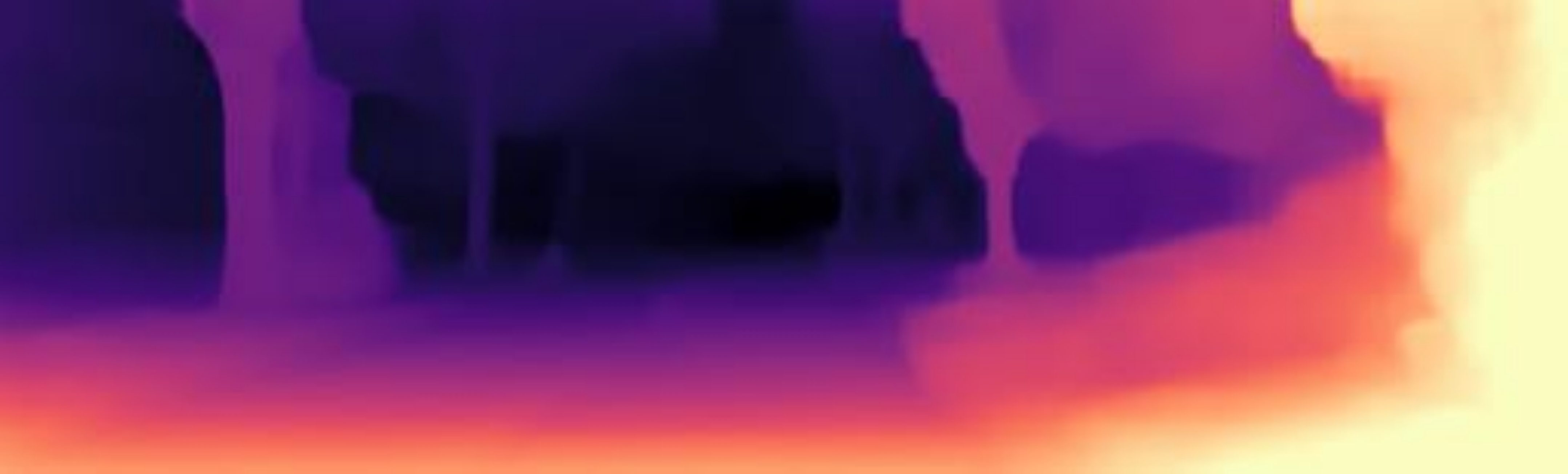}&  
    \includegraphics[]{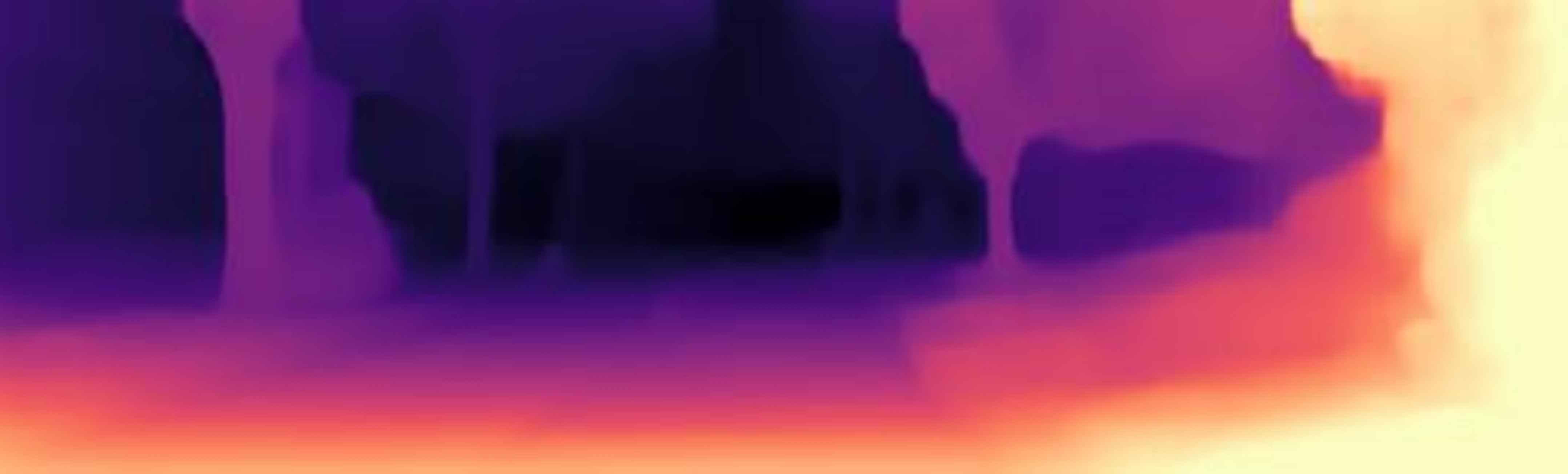}&  
    \includegraphics[]{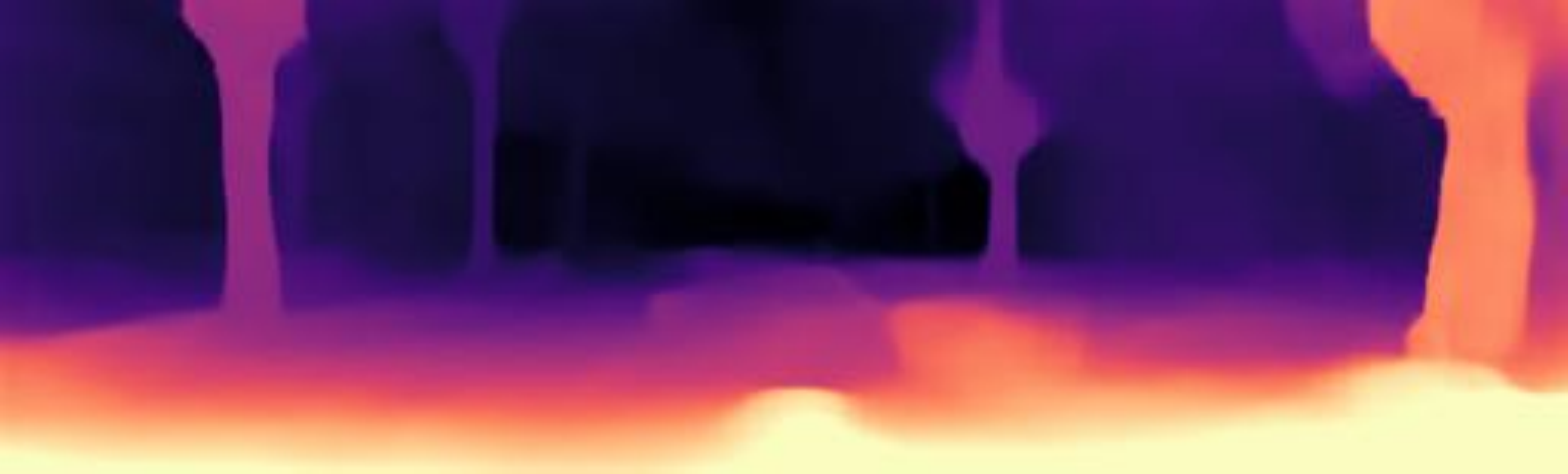}&  
    \includegraphics[]{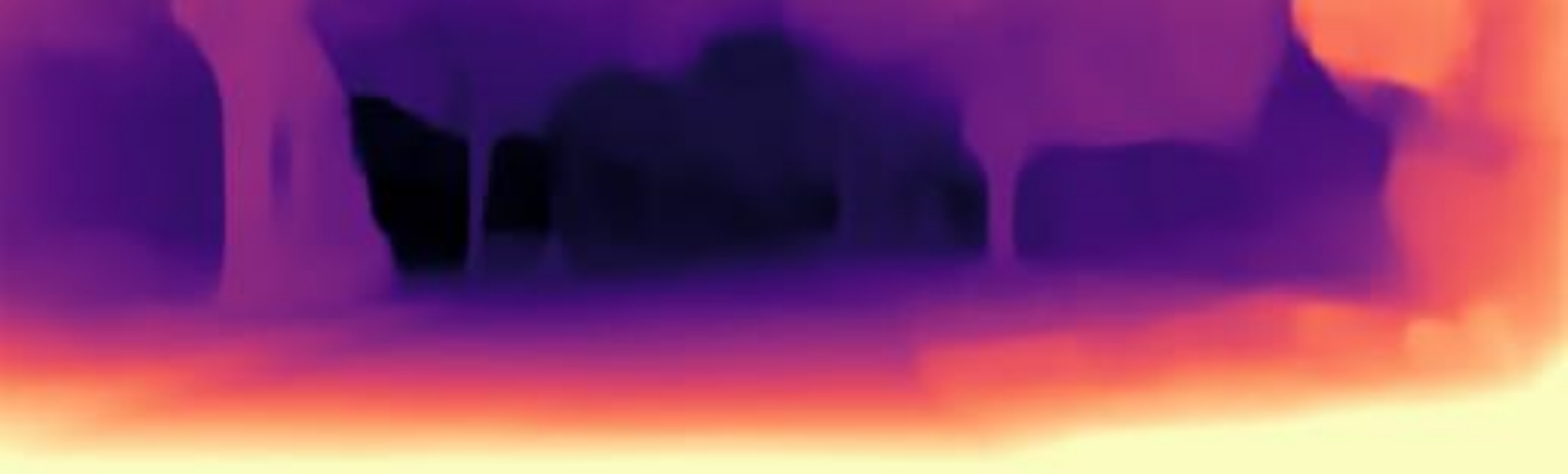}\\
    \includegraphics[]{figure/appendix/supp/monoformer-vi.pdf}    &  \includegraphics[]{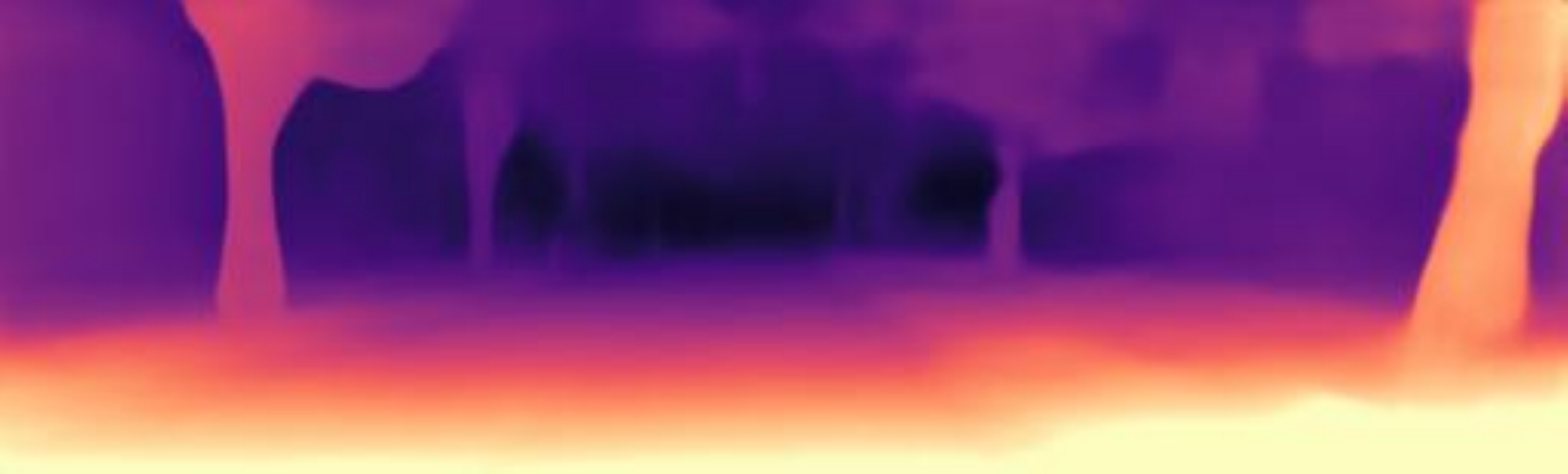}&  
    \includegraphics[]{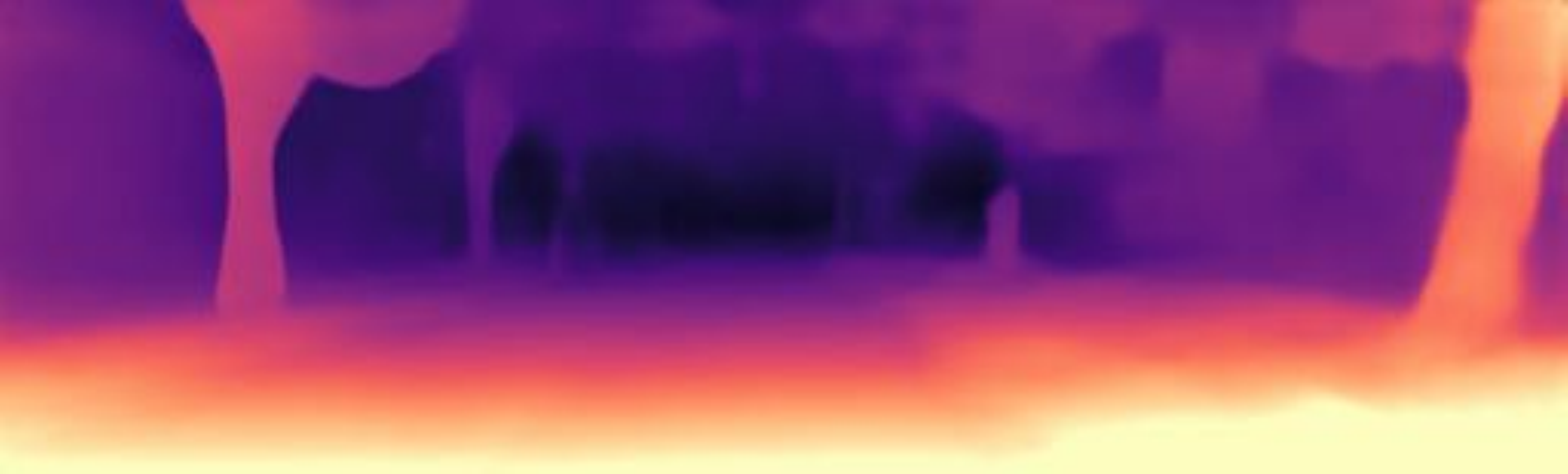}&  
    \includegraphics[]{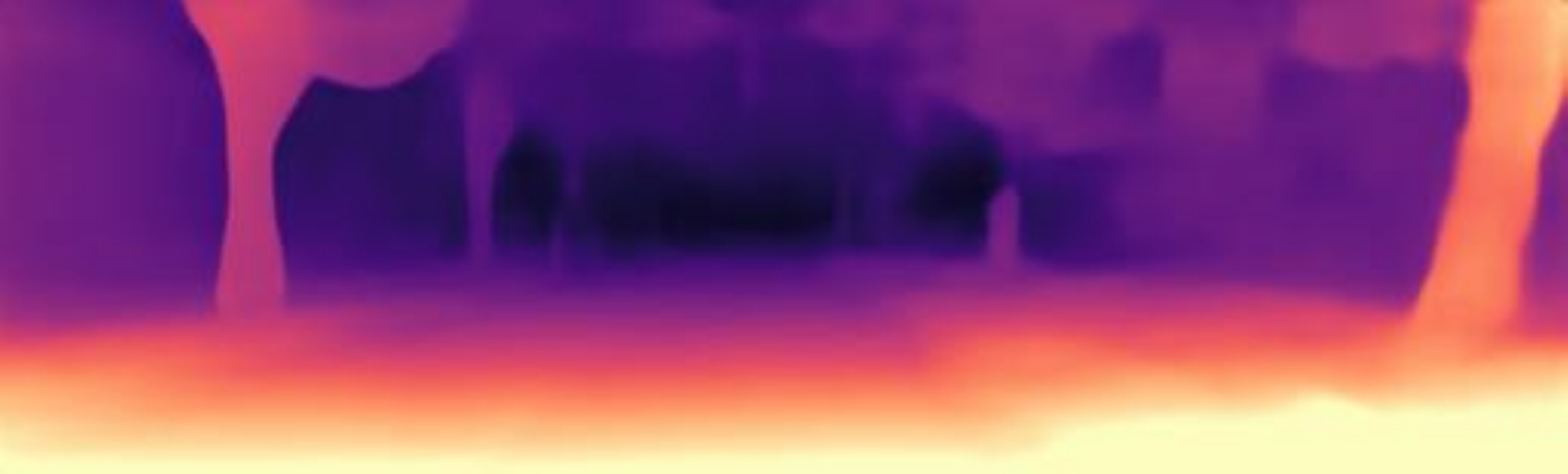}&  
    \includegraphics[]{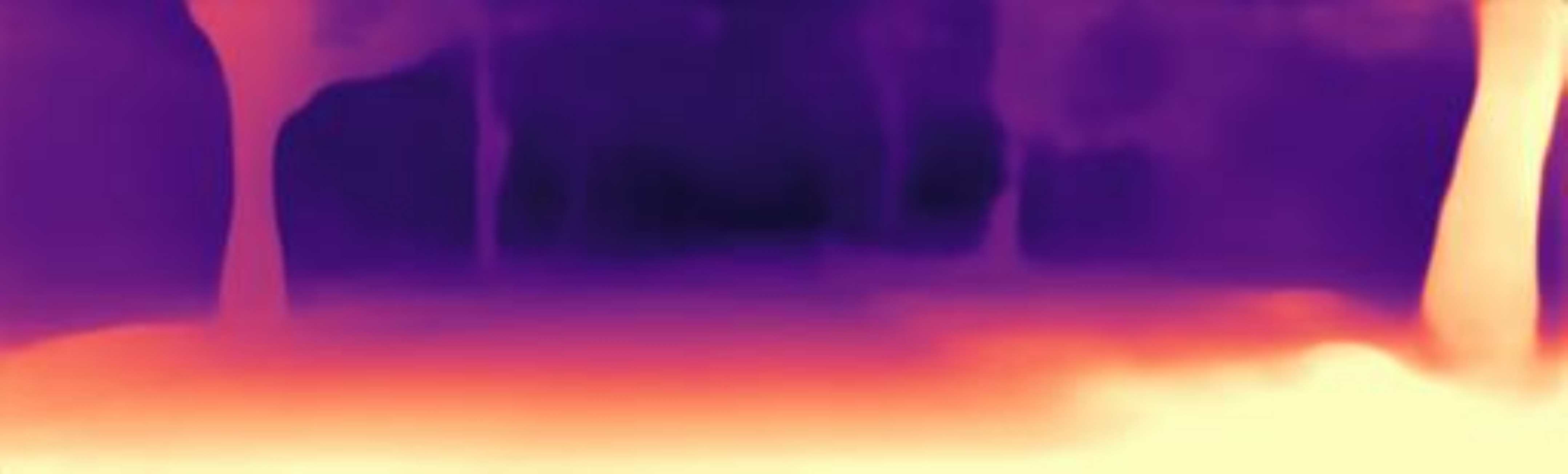}&  
    \includegraphics[]{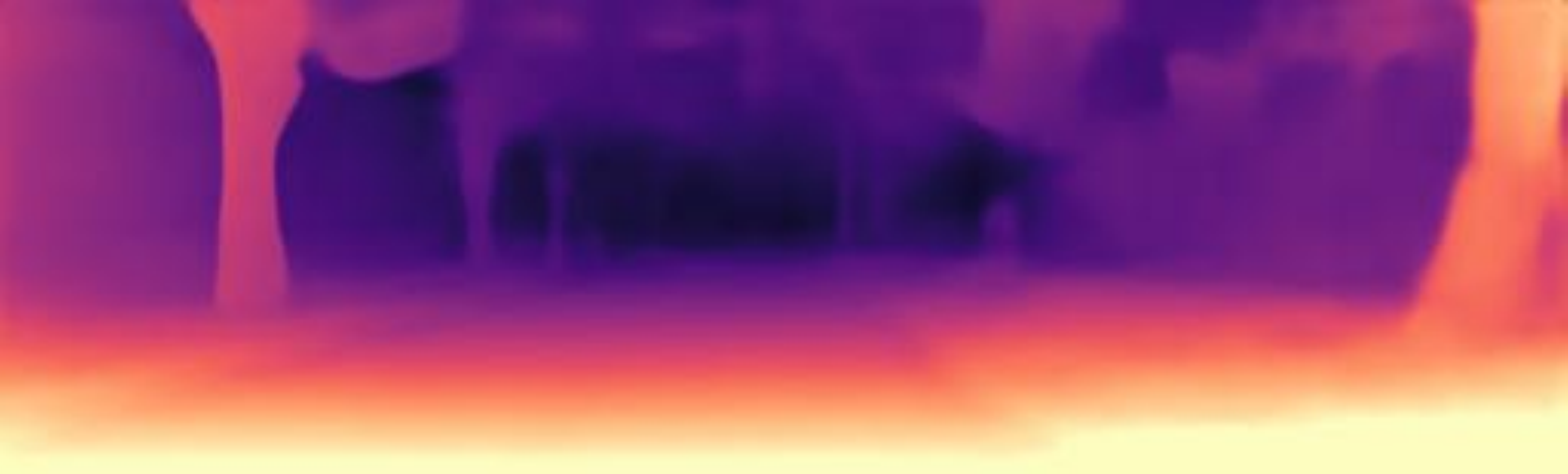}\\
    \includegraphics[]{figure/appendix/supp/monoformer-hy.pdf} &  \includegraphics[]{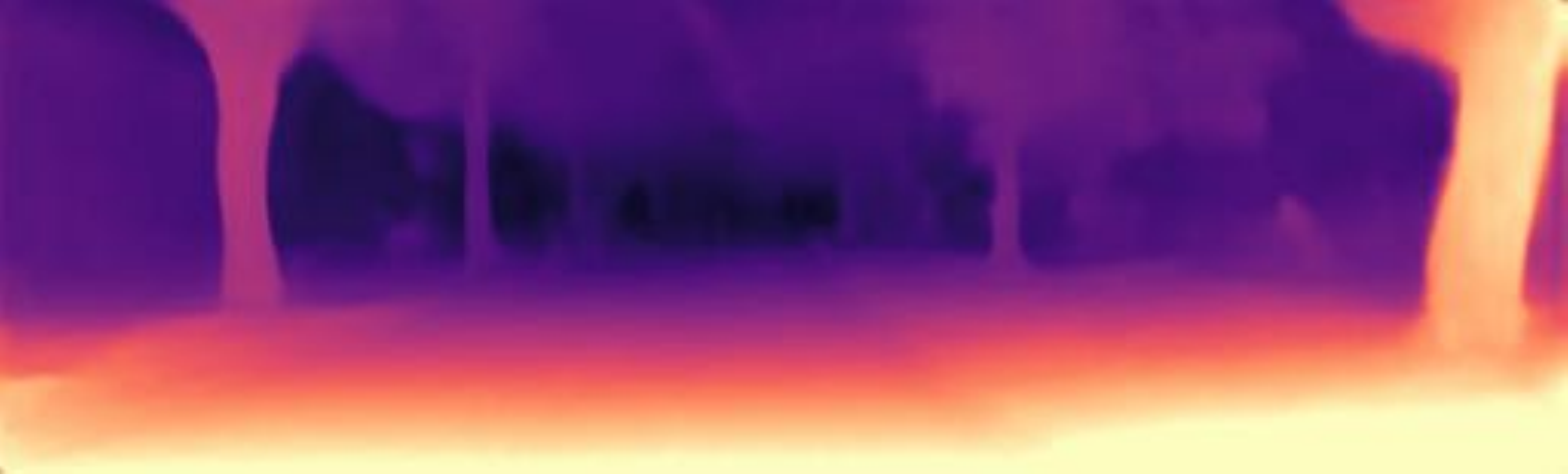}&  
    \includegraphics[]{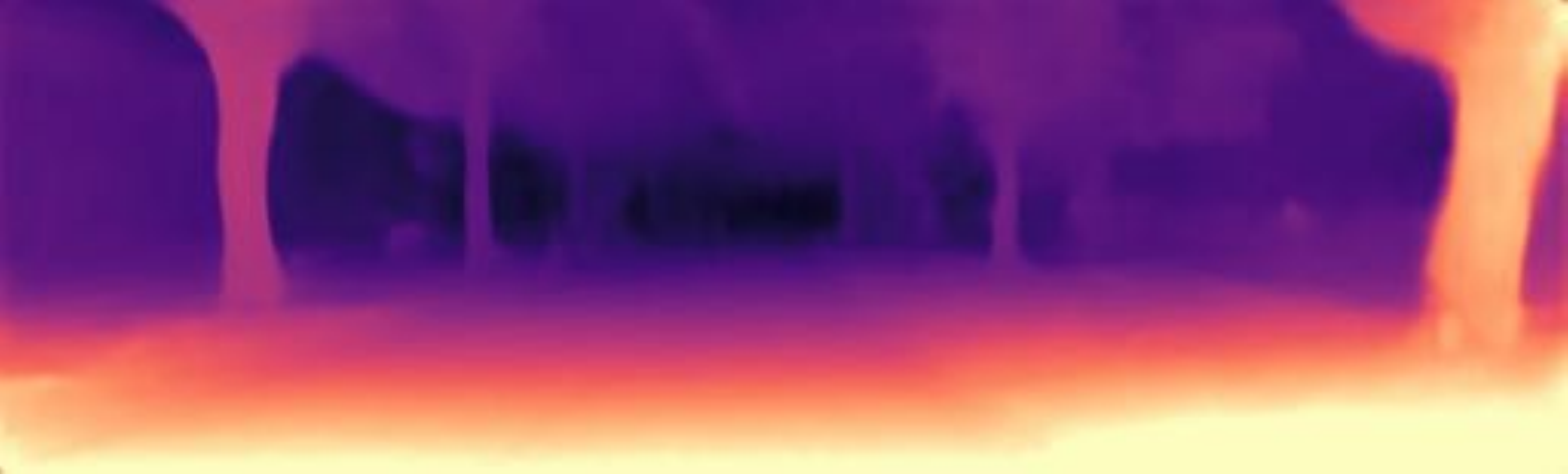}&  
    \includegraphics[]{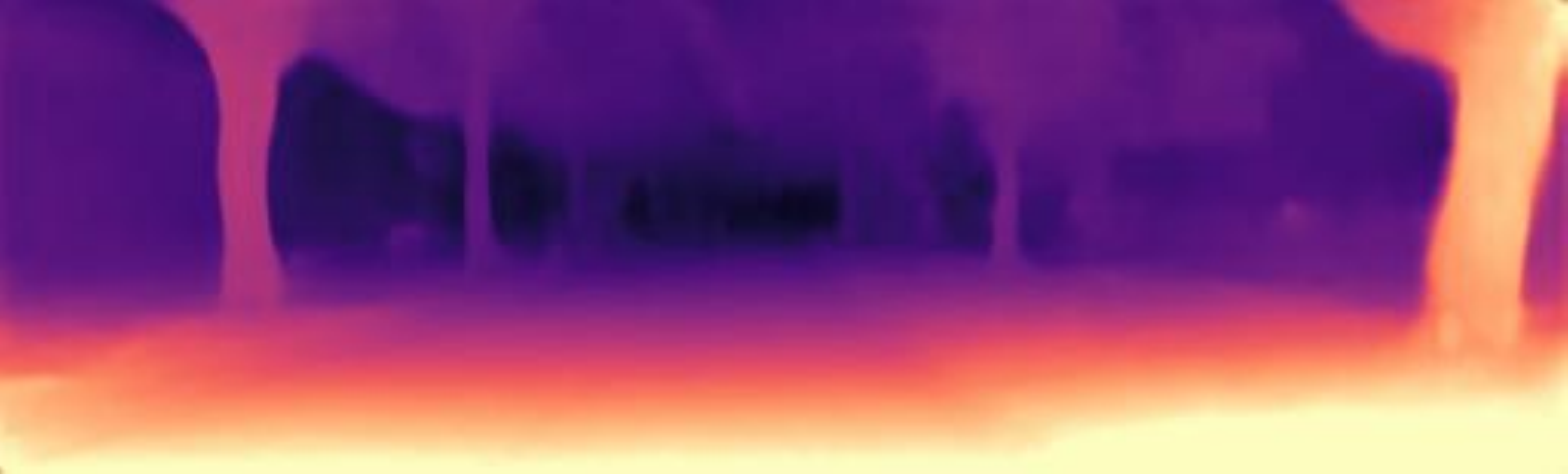}&  
    \includegraphics[]{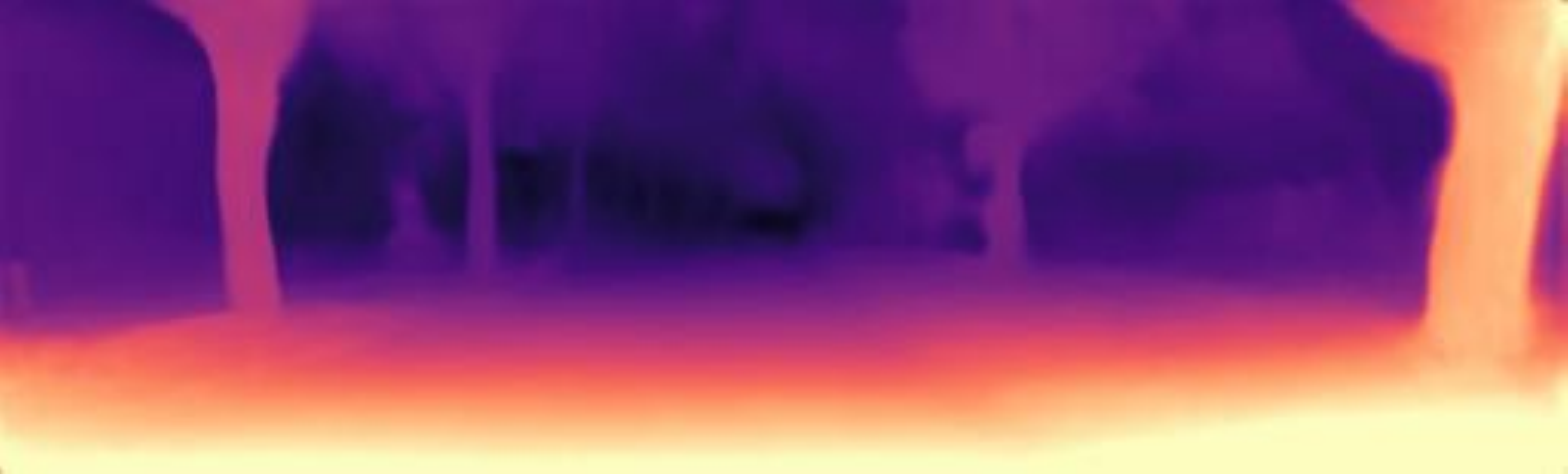}&  
    \includegraphics[]{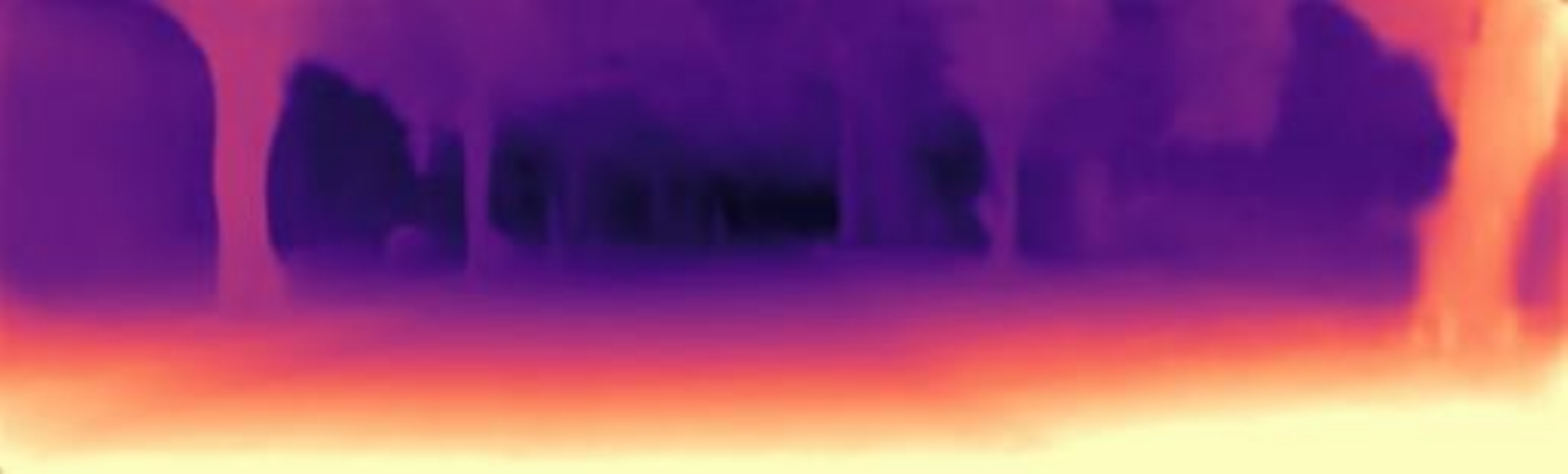}\\
    \end{tabular}
    }
        \caption{\textbf{Comparison of depth map results on various pencil-sketch images.} The middle is the default of the OpenCV function. Based on the default, the left image changes $\sigma_r$, and the right image changes $\sigma_s$.}
        \label{pencil-sketch_apdx}
    \end{subfigure}
\end{figure*}

We generate a style-transfer dataset using code and images provided by \cite{geirhos2018imagenet}\footnote{https://github.com/bethgelab/stylize-datasets}. Also, we generate synthetic datasets for analysis of the generalized performance using OpenCV. The watercolor and the pencil-sketch datasets are generated by using the function of \text{cv2.stylization}, and \text{cv2.pencilsketch}, respectively. Each function gets $\sigma_r$ and $\sigma_s$ as input parameters. The terms $\sigma_s$ and $\sigma_r$ determine the amount of smoothing effect applied to the image, and the number of edges preserved in the image, respectively. We conduct the same experiment as texture-shifted datasets in various parameters. \figref{watercolor_apdx} and \figref{pencil-sketch_apdx} show that CNN is texture-biased while Transformer is shape-biased regardless of the parameters for image generations.

\subsection{Computational Complexity of Models}
Transformer architecture and attention modules generally show somewhat heavier computational complexity than CNN methods. Our model also has more cost than other CNN-based methods in computation, as shown in \tabref{complexity_table_apdx}. However, the results show that the Transformer-based models are slower than CNN-based models, but Transformer models take about 15~25ms, which is a reasonable speed to be practically utilized. 

\begin{table}[h]
\centering
\resizebox{\columnwidth}{!}{%
\begin{tabular}{c|cccc}
\hline
Model             & Abs Error & \#Params & \#Inference time(ms) & \#FLOPs \\ \hline
Monodepth2        & 0.115     & 14.84M   & 4                    & 8       \\
PackNet-SfM       & 0.111     & 128M     & 24                   & 205     \\
R-MSFM6           & 0.112     & 3.8M     & 12                   & 31.2    \\
MonoFormer-ViT    & 0.118     & 138M     & 18                   & 96.86   \\
MonoFormer-Hybrid & 0.104     & 159M     & 25                   & 111.4   \\ \hline
\end{tabular}%
}
\caption{\textbf{Complexitytable of models.} All results are tested on a single A6000 GPU with an input image size of $640 \times 192$}
\label{complexity_table_apdx}
\vspace{-0.3cm}
\end{table}

\begin{figure*}[t] 
    \centering
    \resizebox{\linewidth}{!}{
    \begin{tabular}{cccccc}
    
    \includegraphics[]{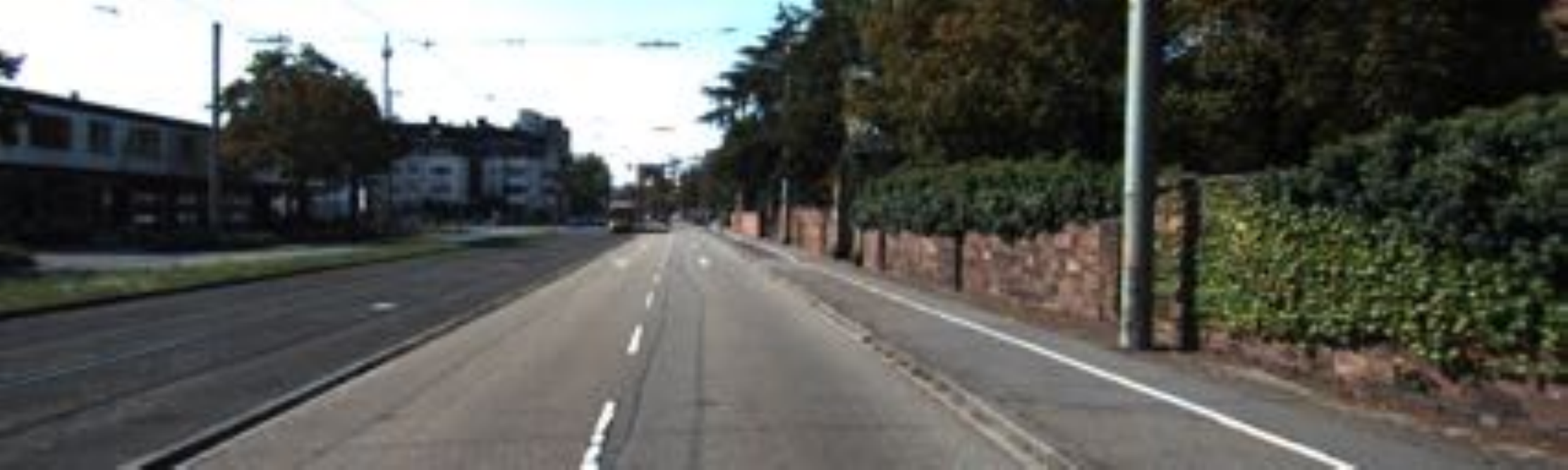}& 
    \includegraphics[]{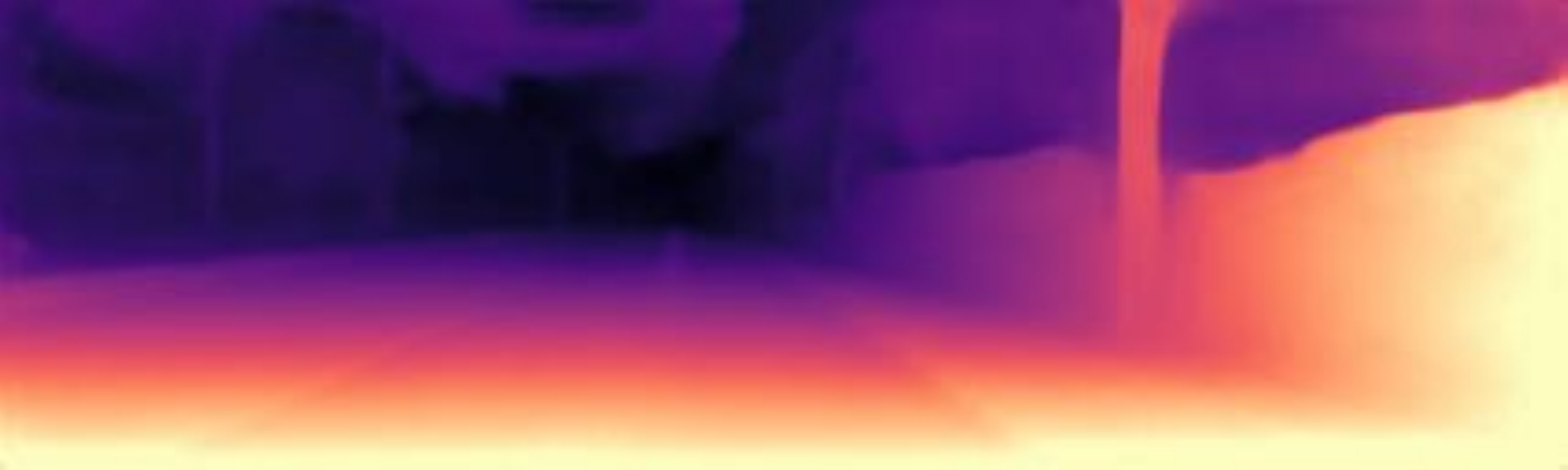}& 
    \includegraphics[]{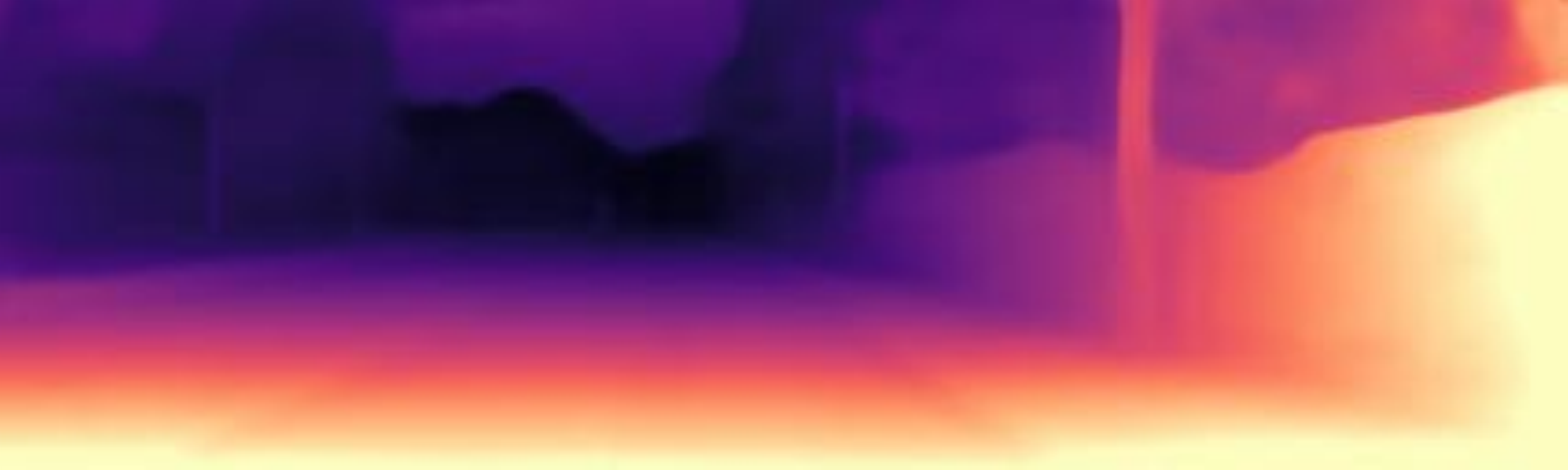}& 
    \includegraphics[]{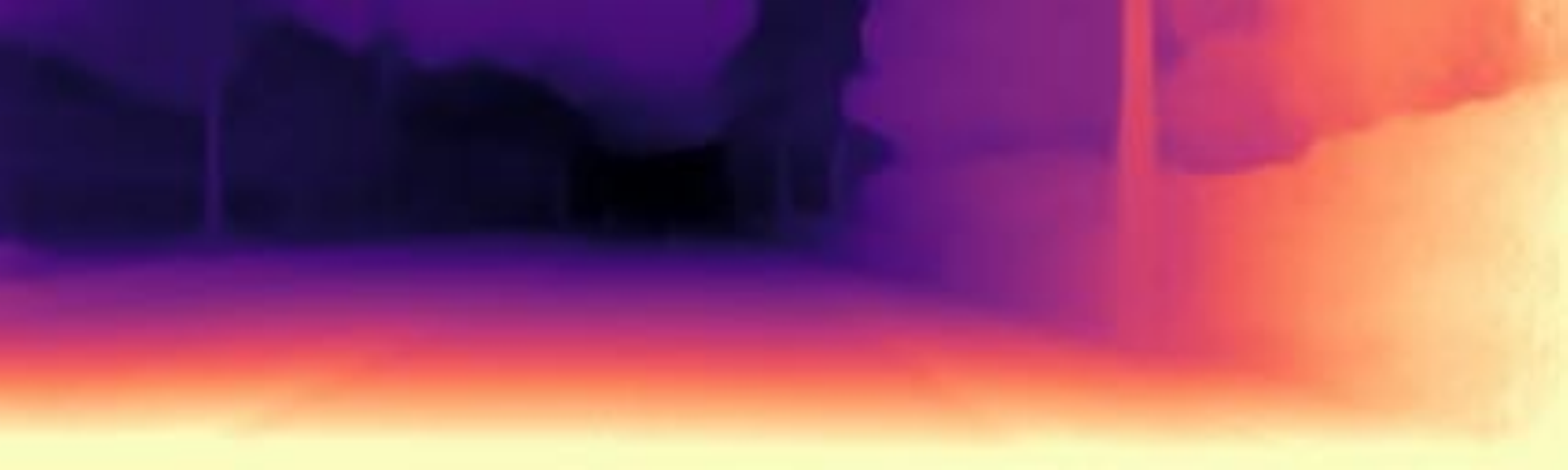}& 
    \includegraphics[]{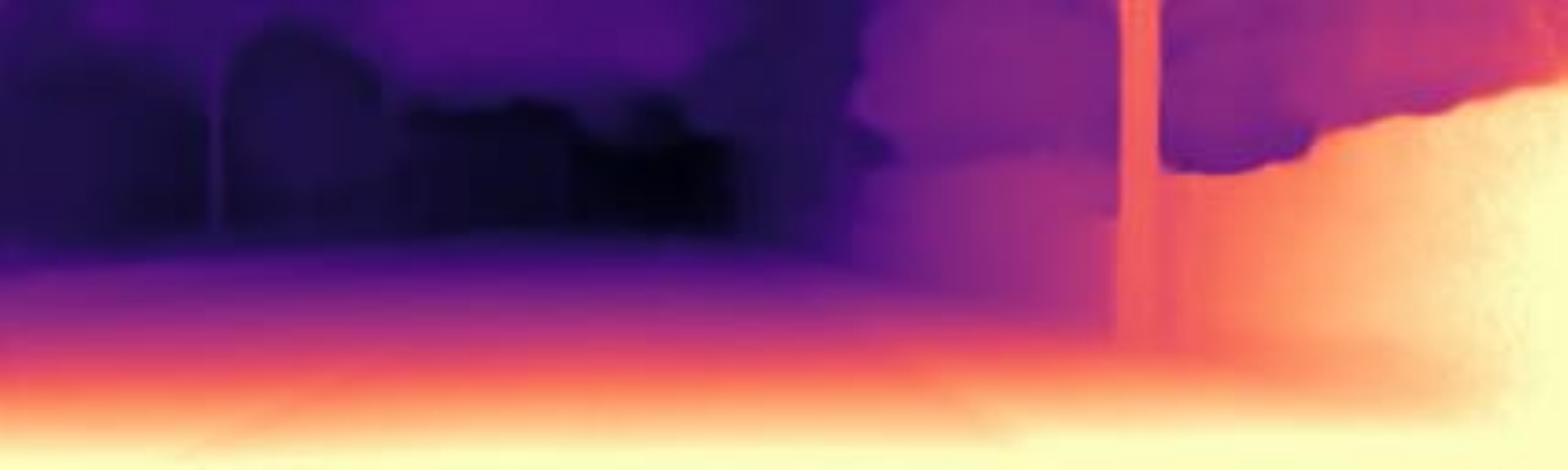}& 
    \includegraphics[]{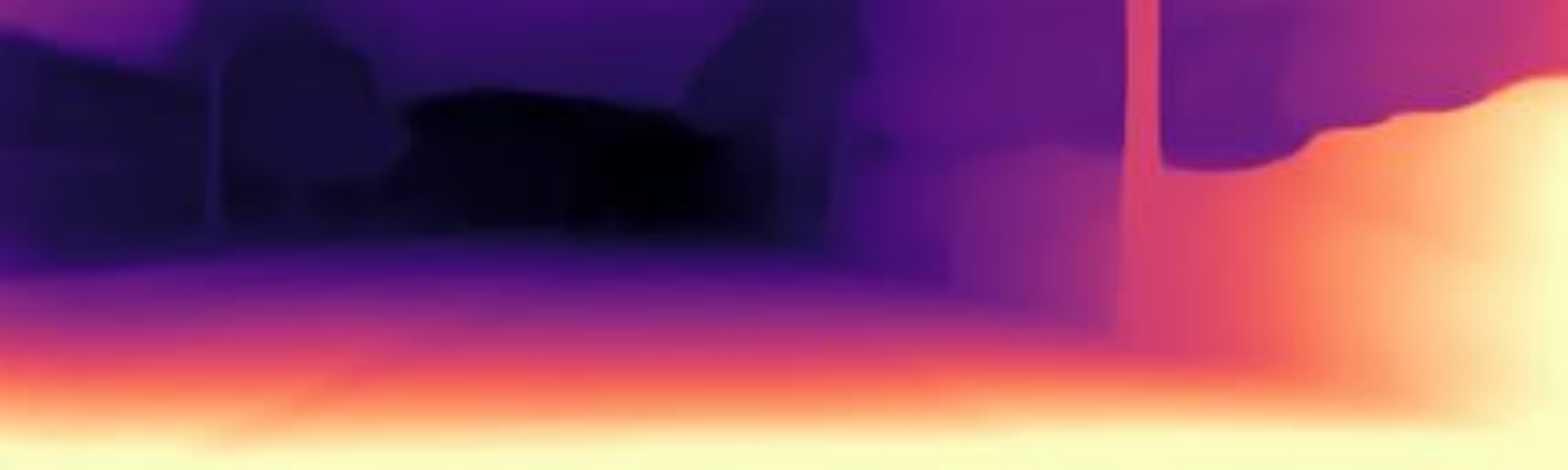}\\
    
    \includegraphics[]{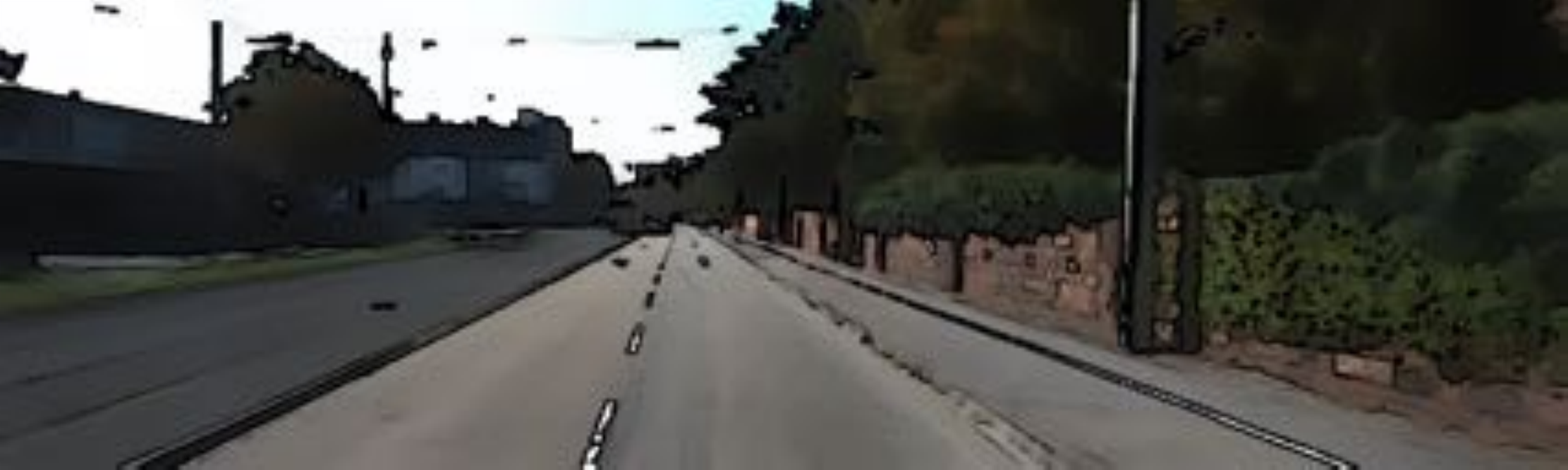}& 
    \includegraphics[]{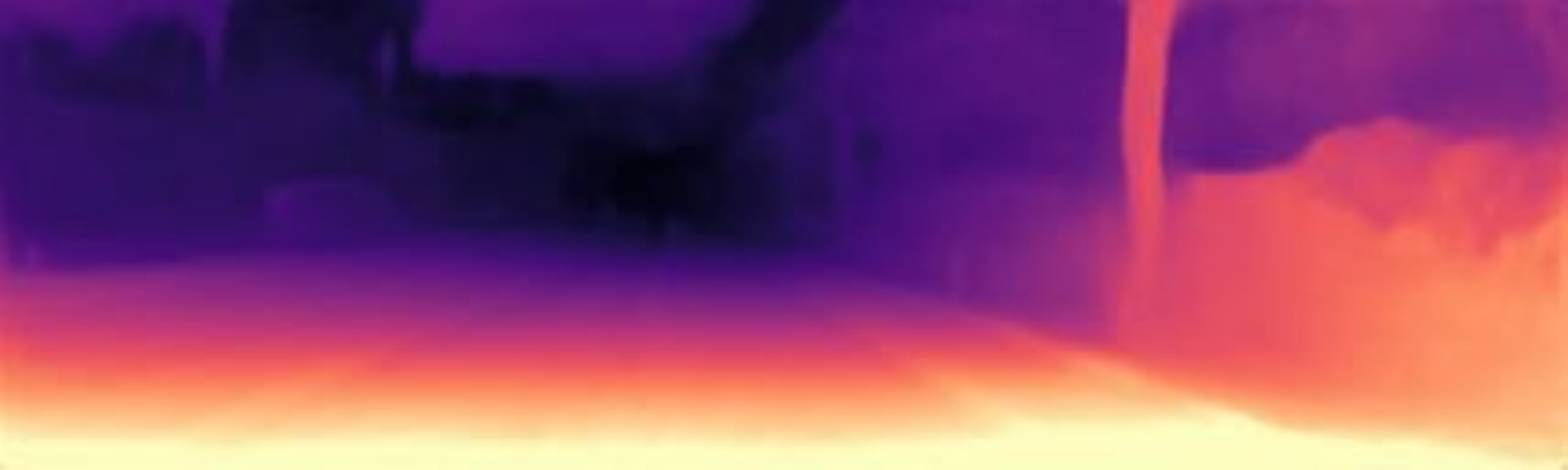}& 
    \includegraphics[]{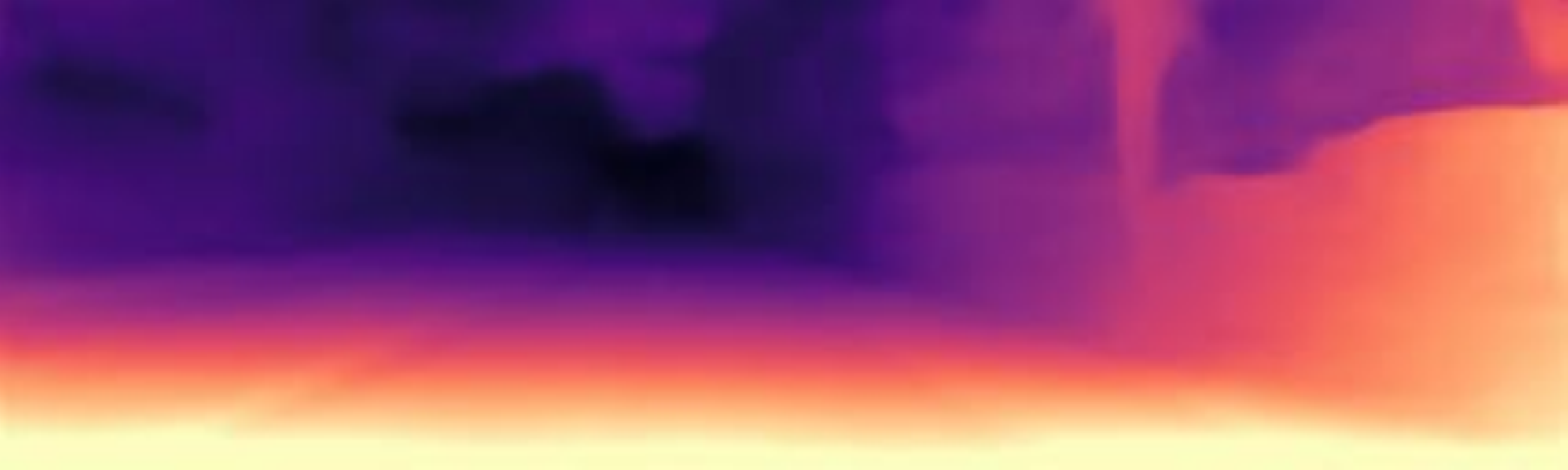}& 
    \includegraphics[]{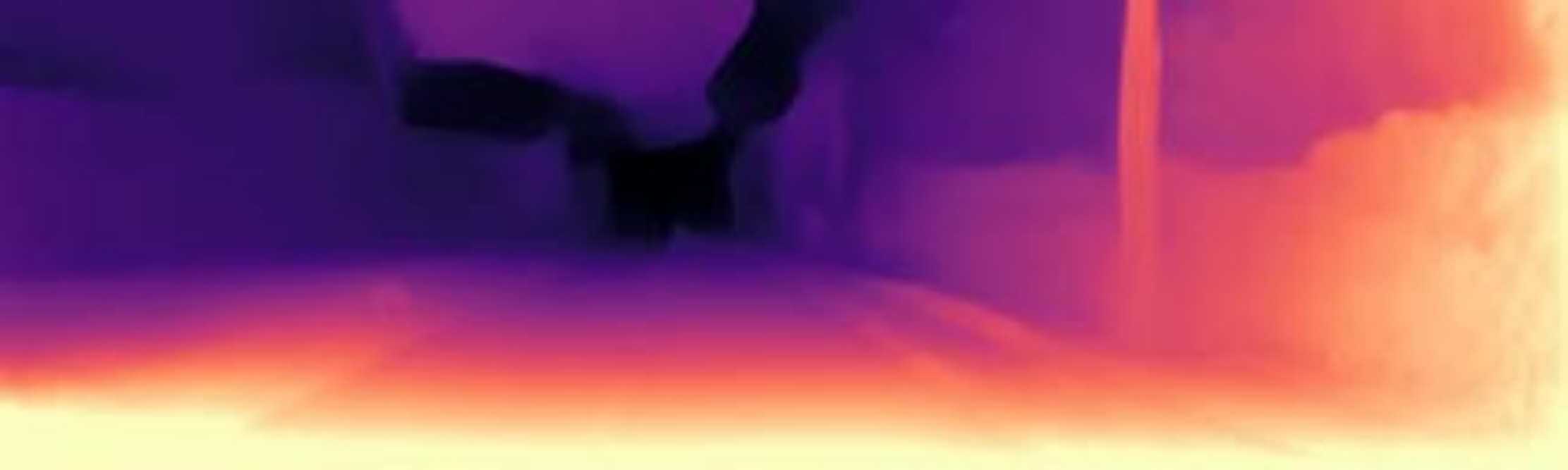}& 
    \includegraphics[]{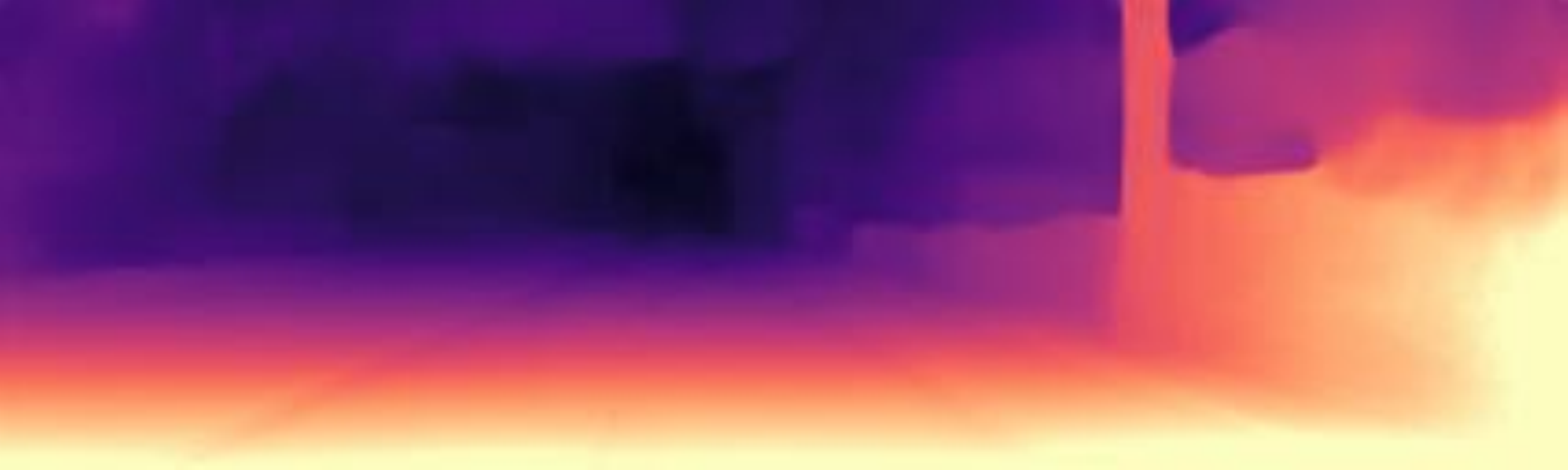}& 
    \includegraphics[]{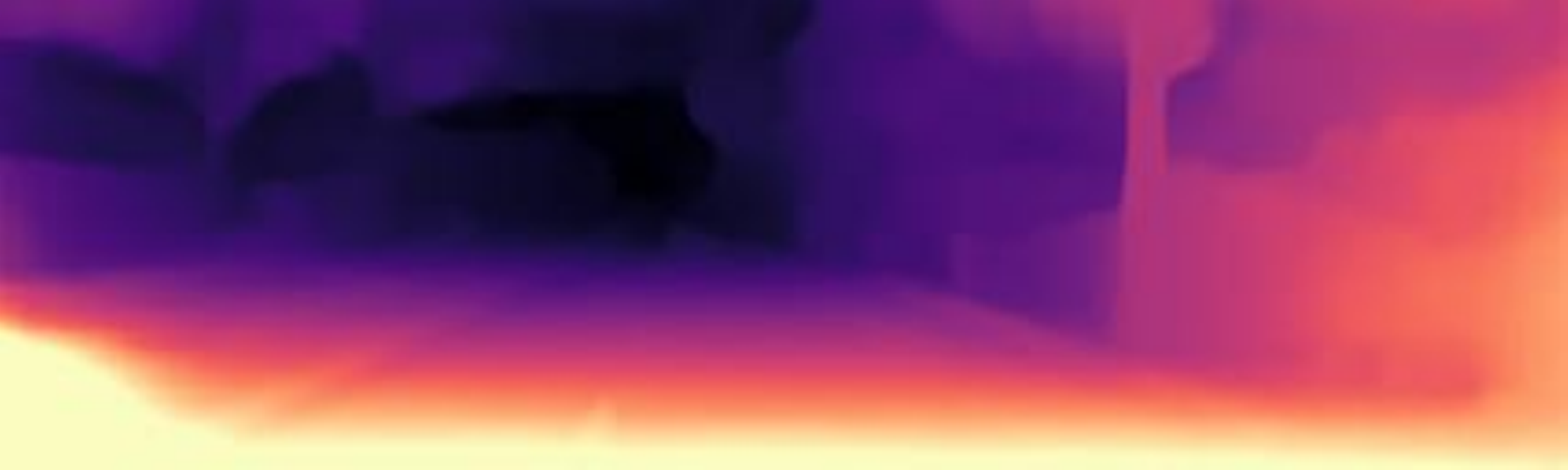}\\
    
    \includegraphics[]{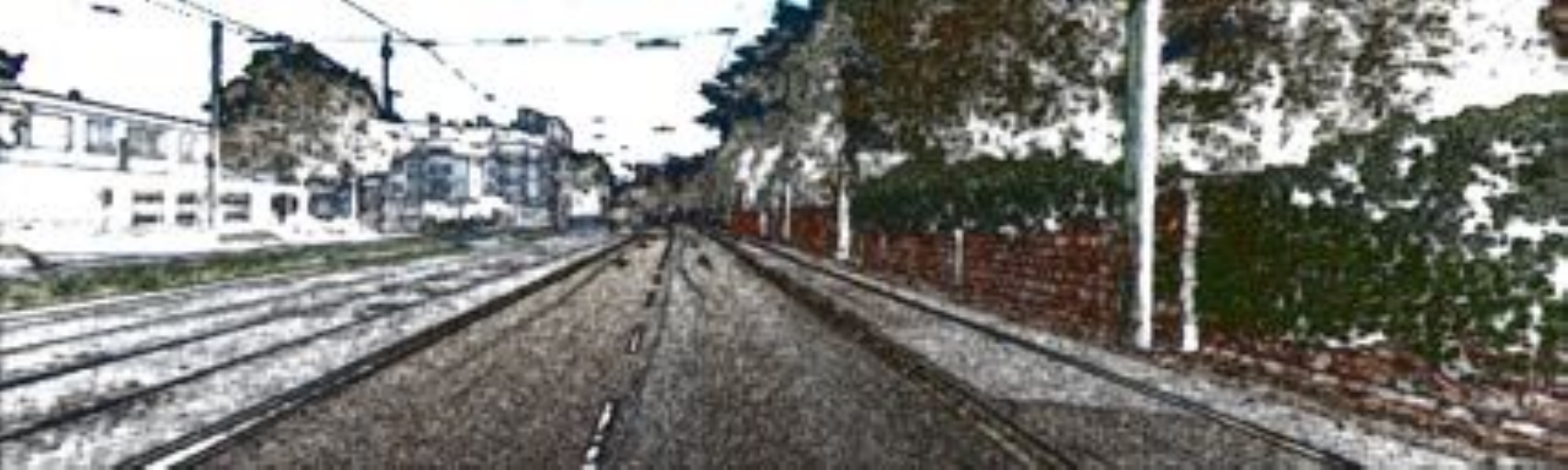}& 
    \includegraphics[]{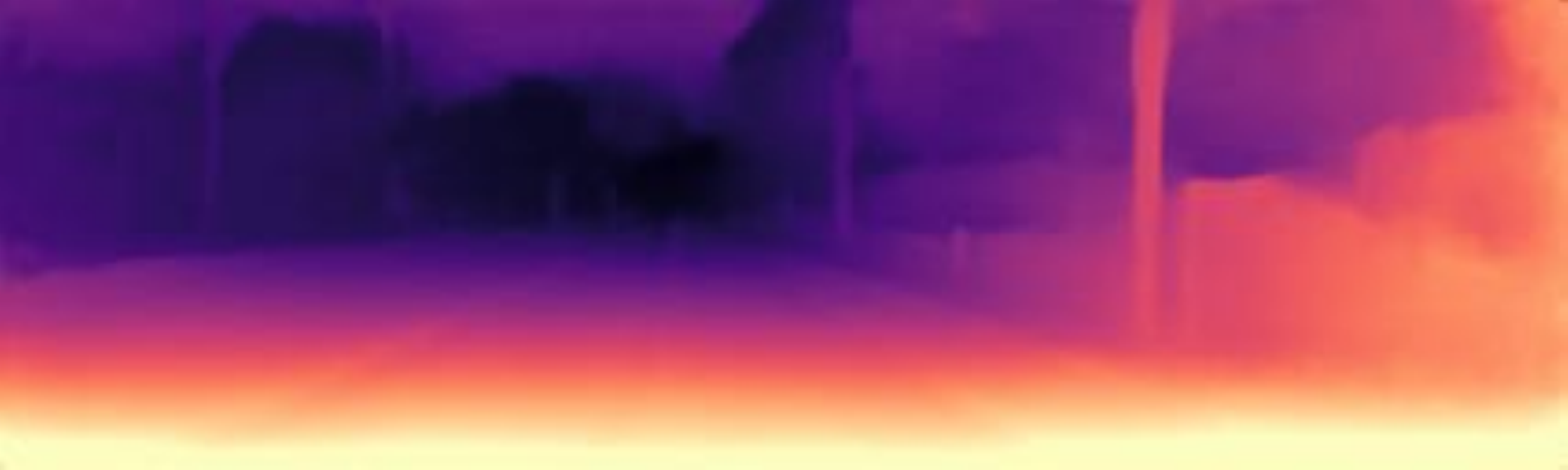}& 
    \includegraphics[]{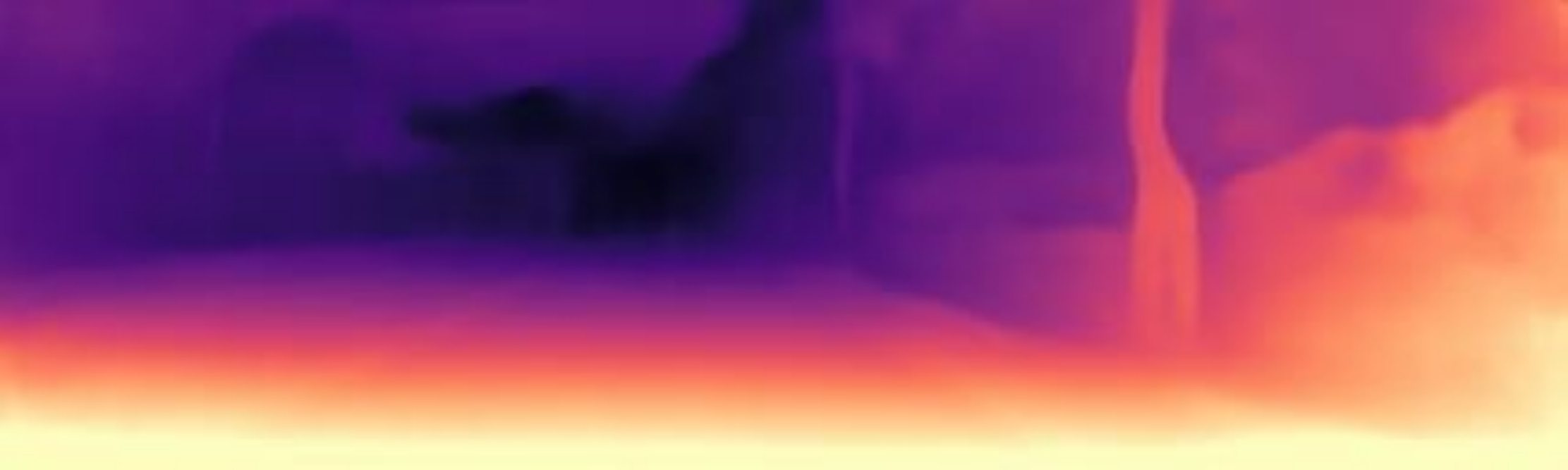}& 
    \includegraphics[]{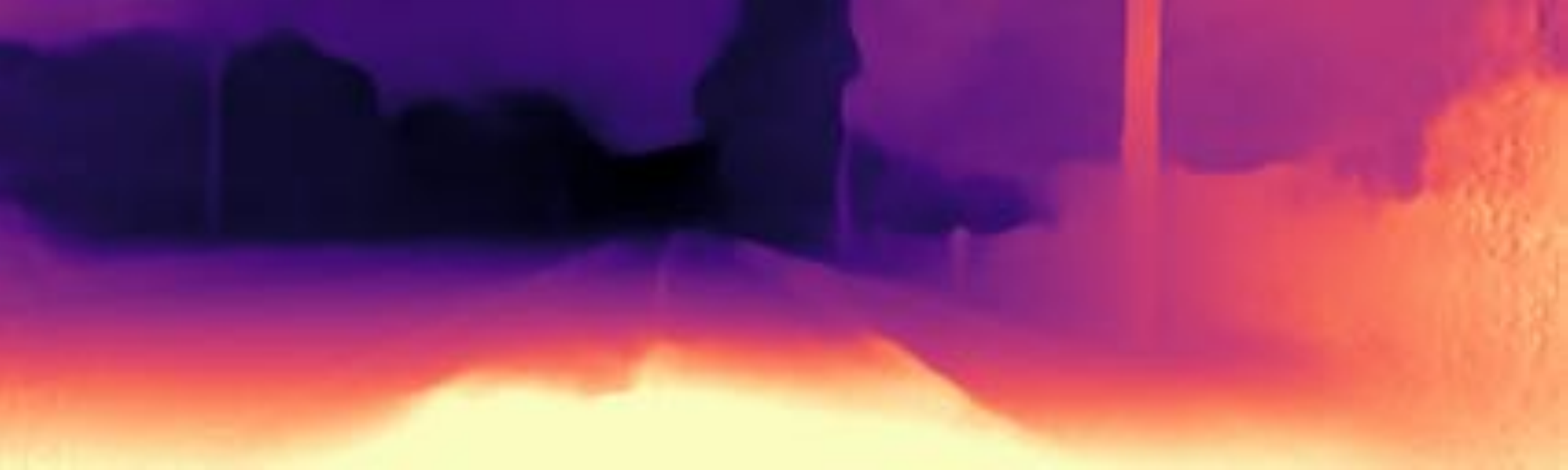}& 
    \includegraphics[]{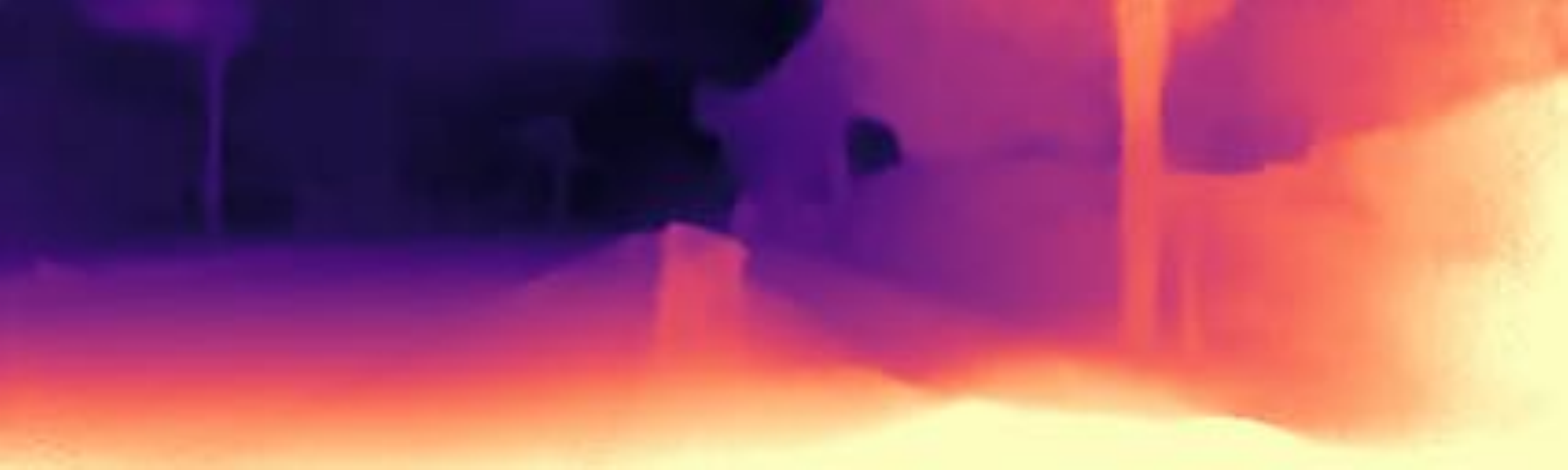}& 
    \includegraphics[]{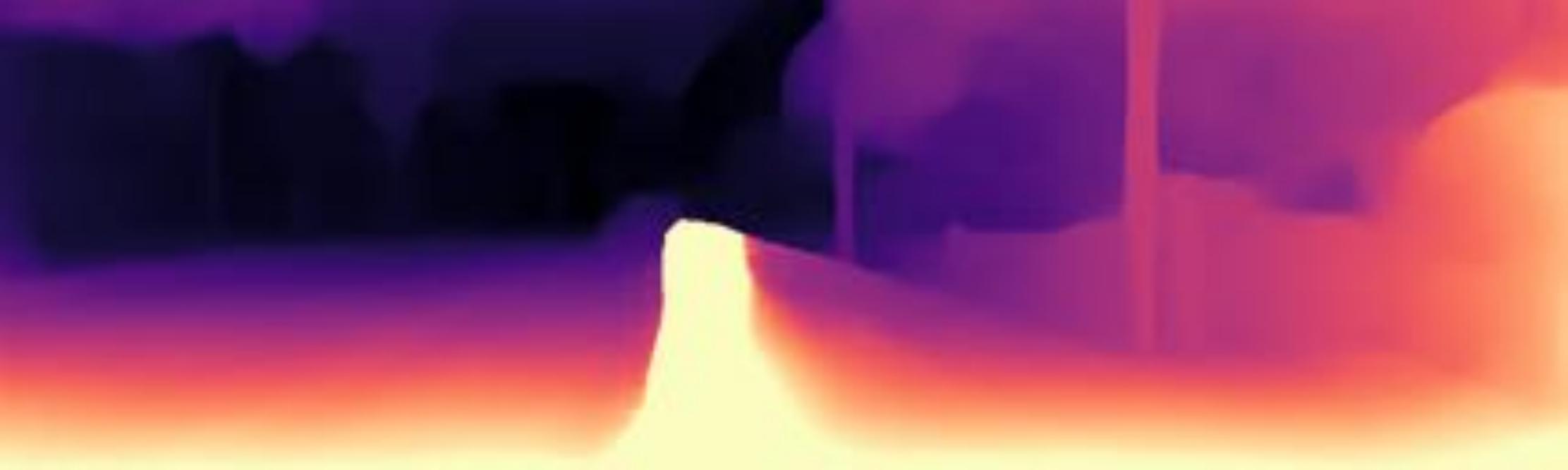} \\
    
    \includegraphics[]{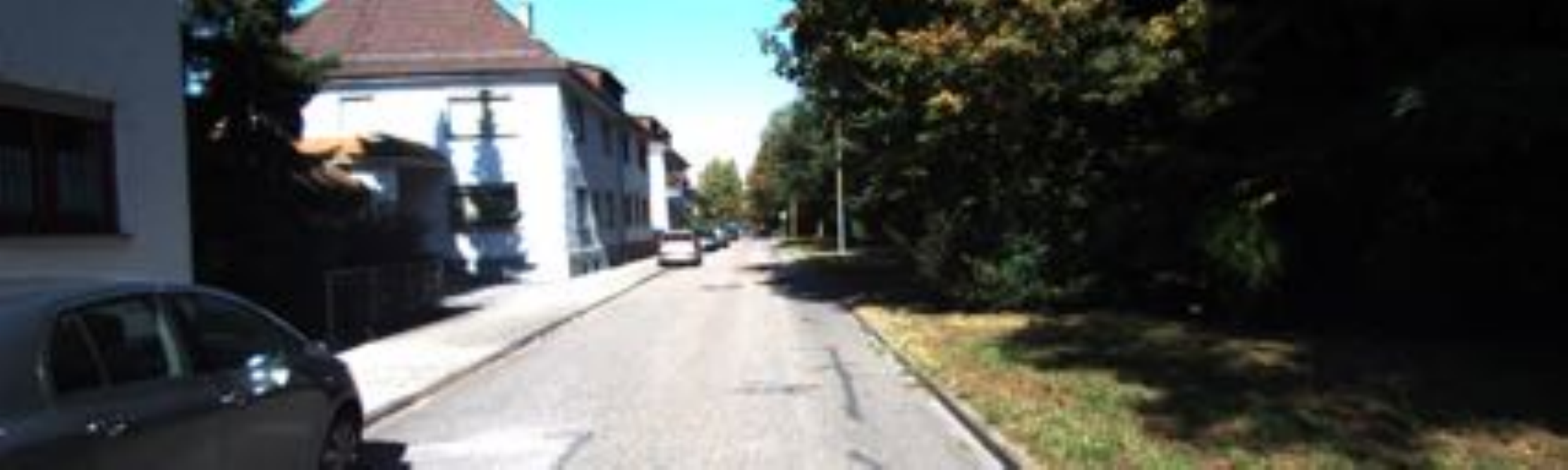}& 
    \includegraphics[]{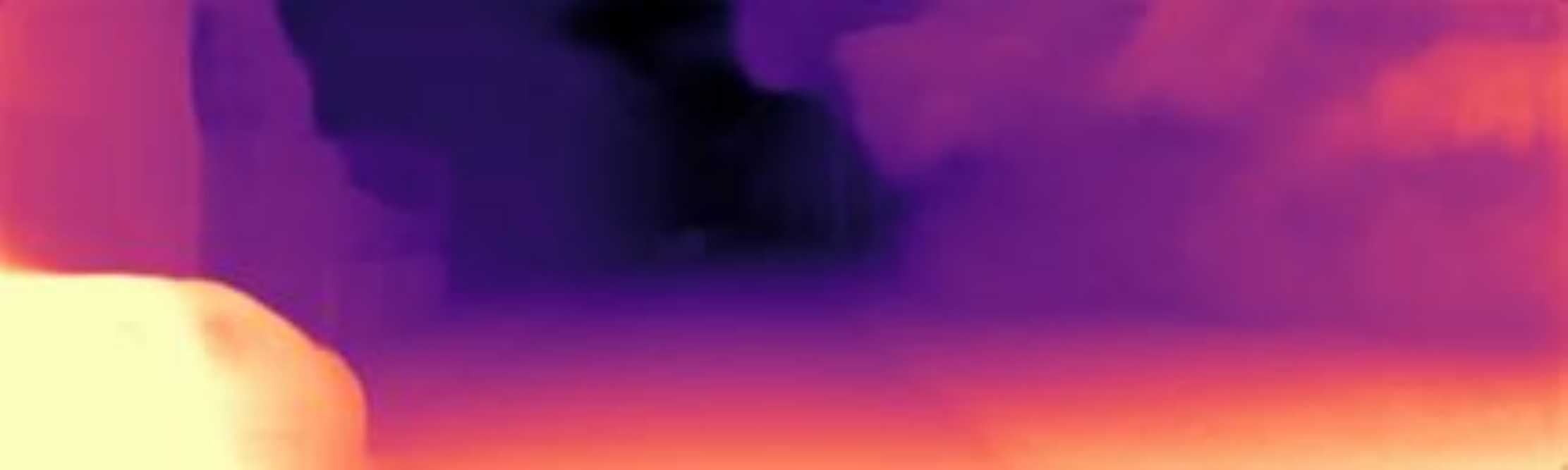}& 
    \includegraphics[]{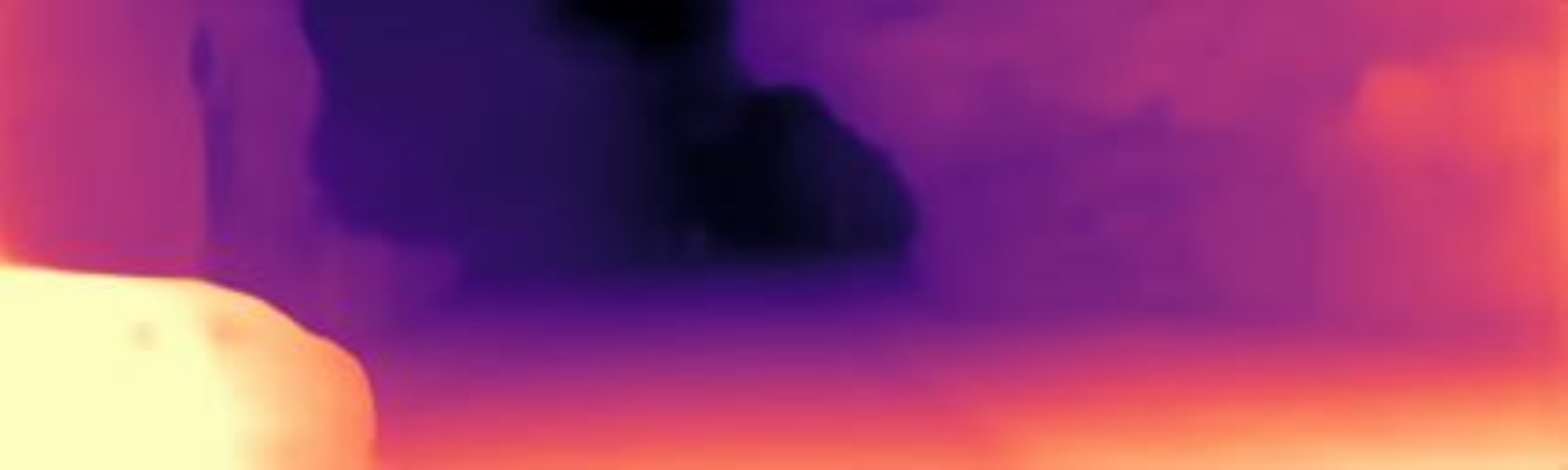}& 
    \includegraphics[]{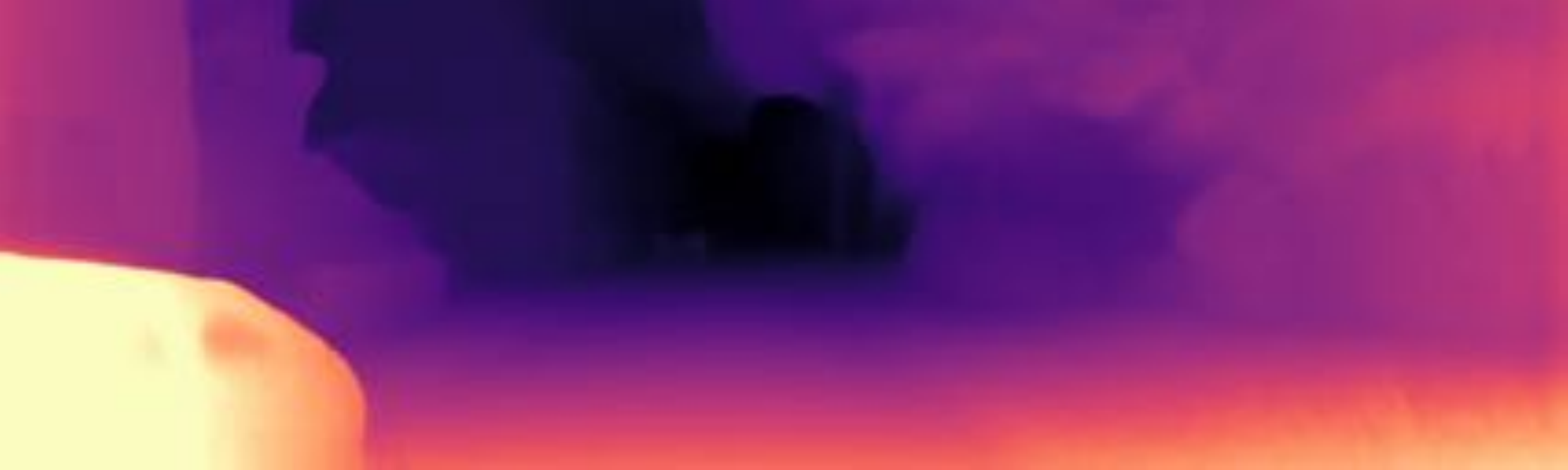}& 
    \includegraphics[]{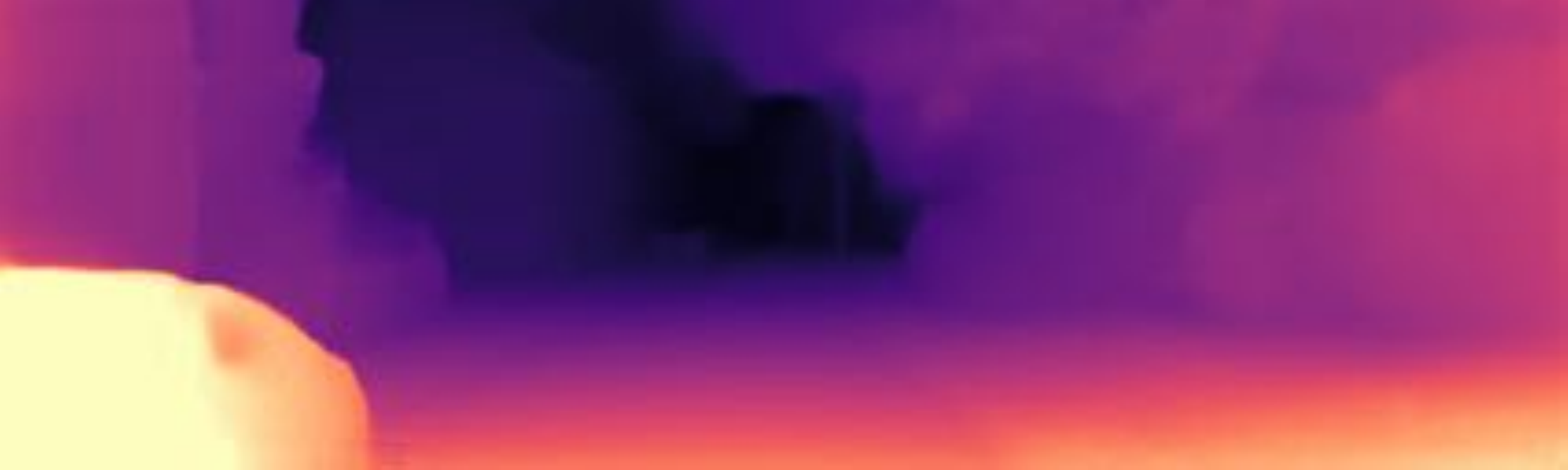}& 
    \includegraphics[]{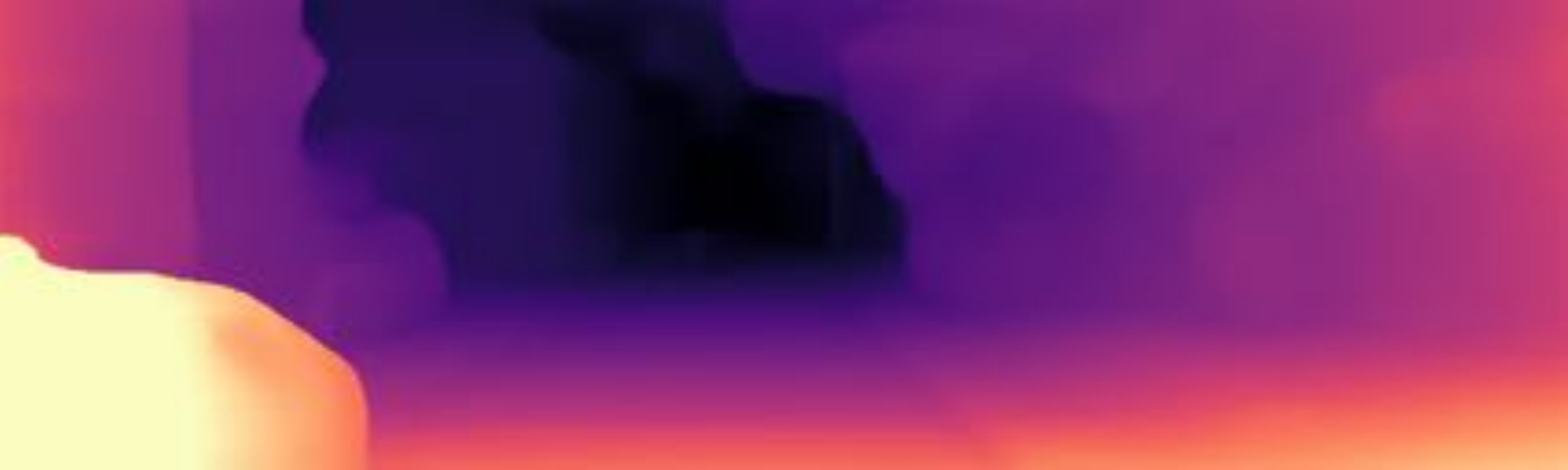}\\
    
    \includegraphics[]{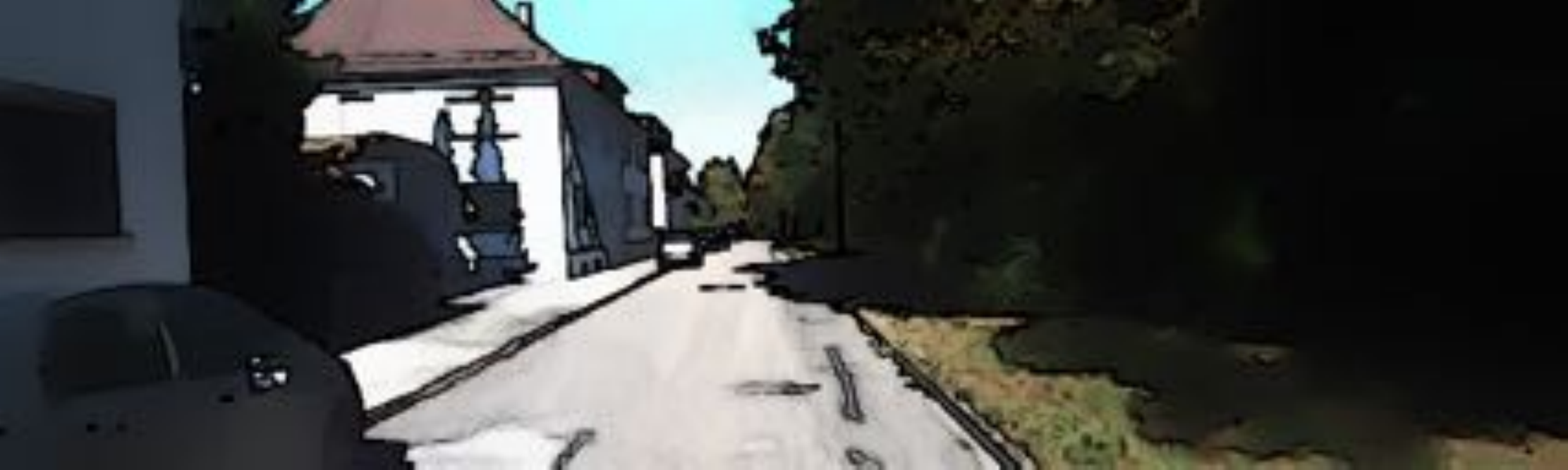}& 
    \includegraphics[]{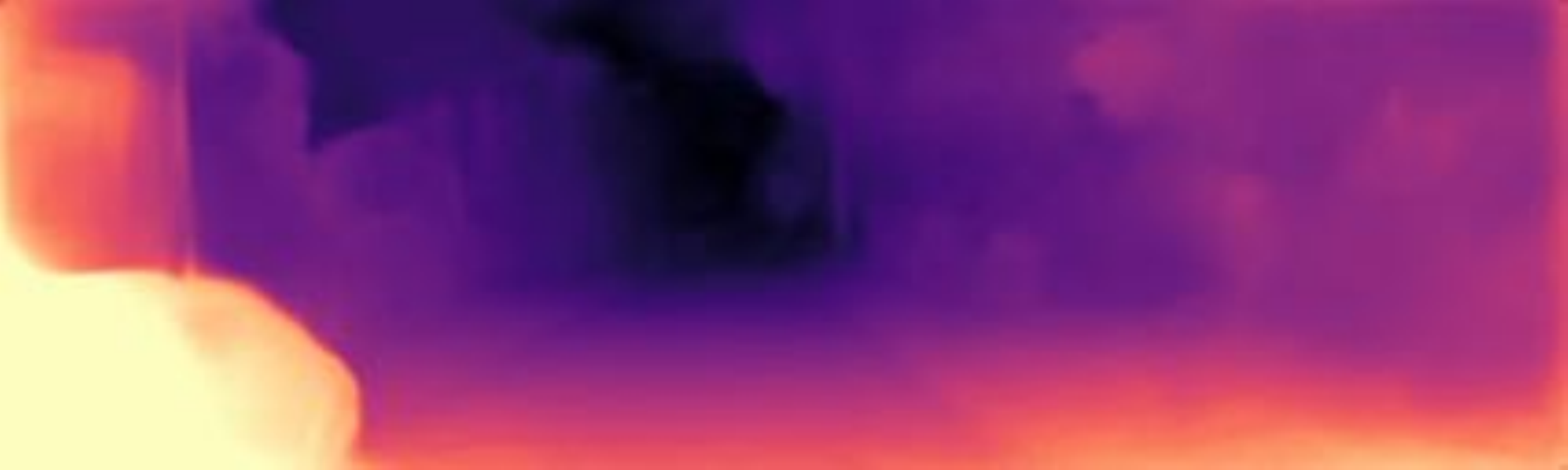}& 
    \includegraphics[]{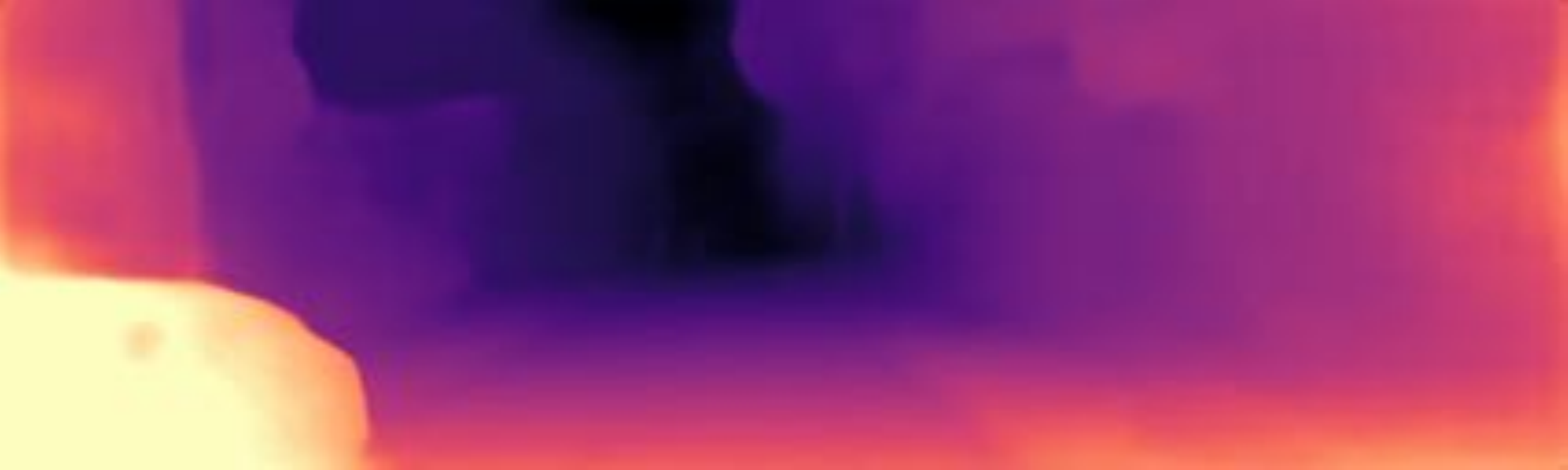}& 
        \includegraphics[]{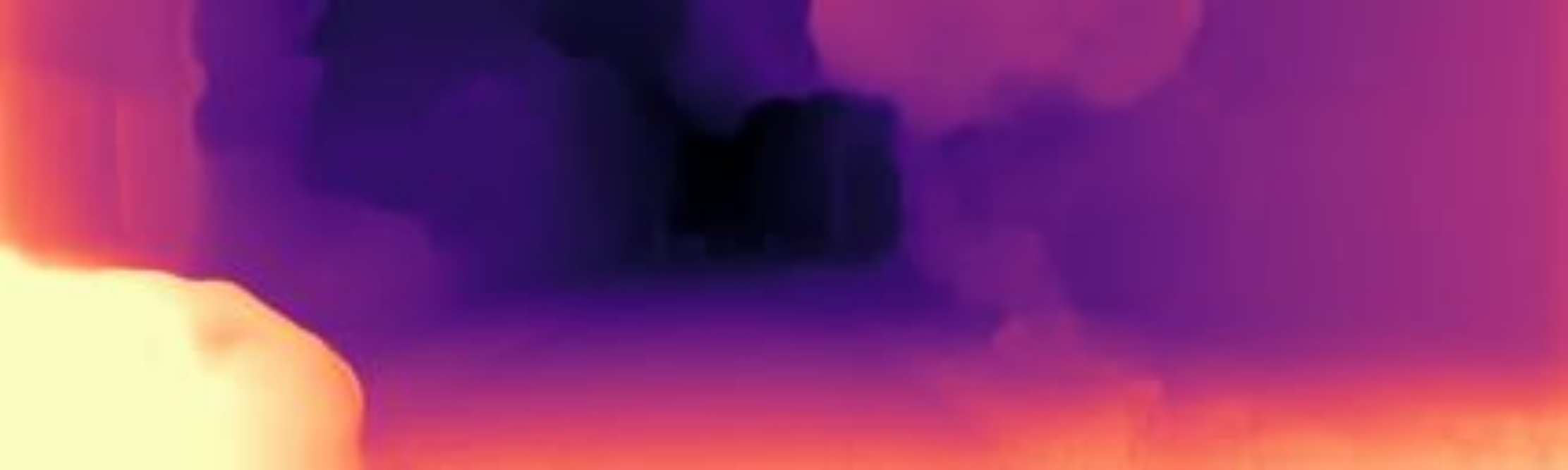}& 
    \includegraphics[]{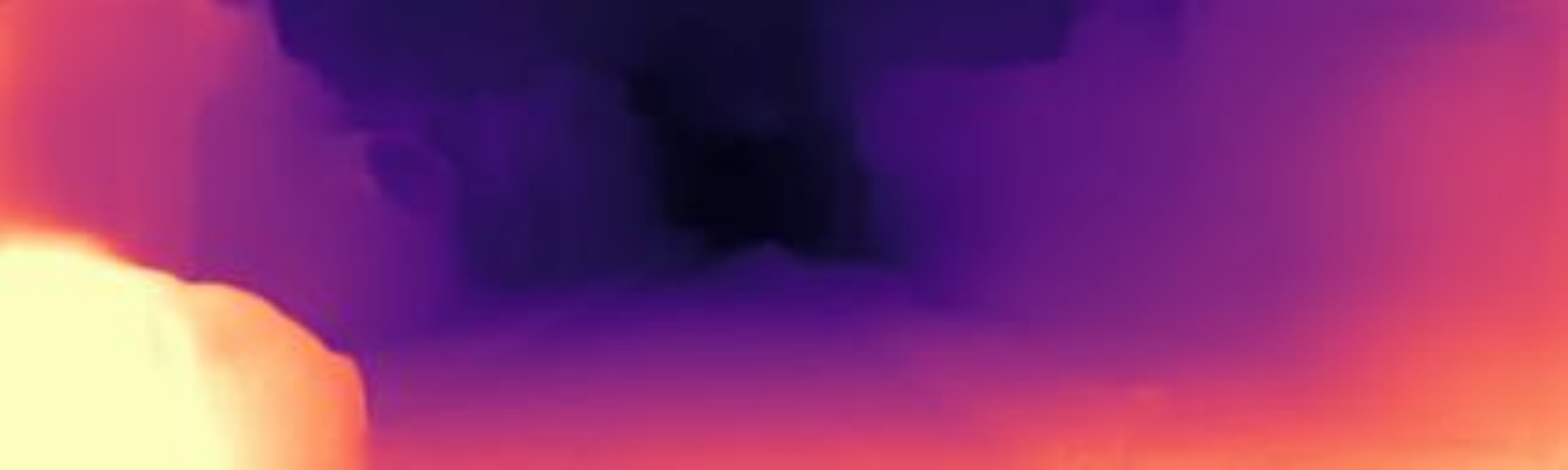}& 
    \includegraphics[]{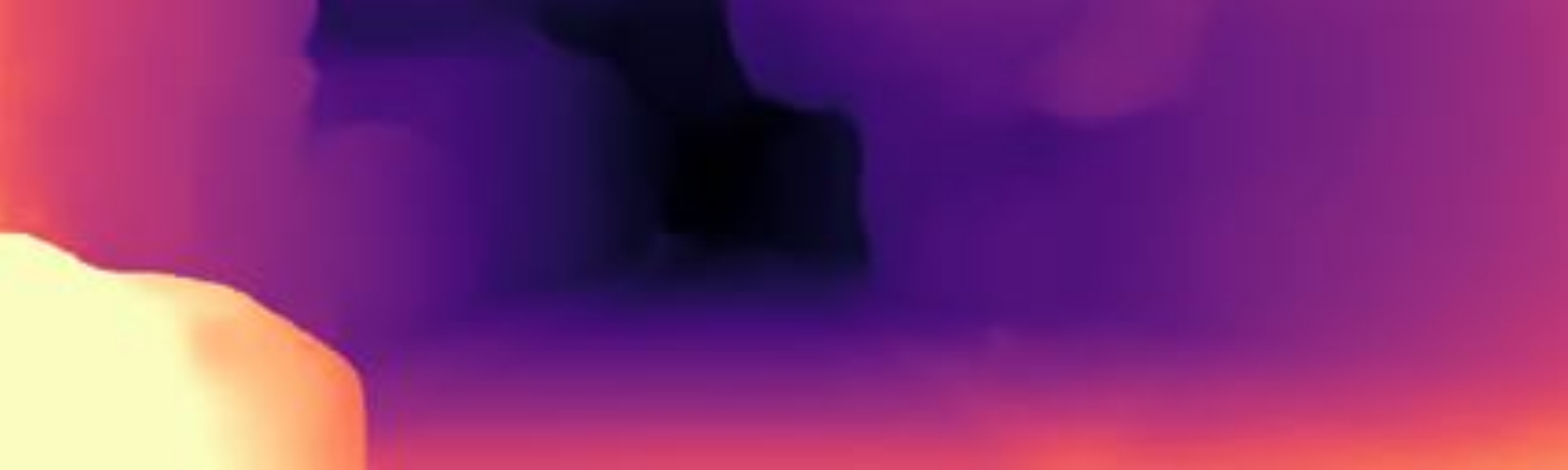}\\
    
    \includegraphics[]{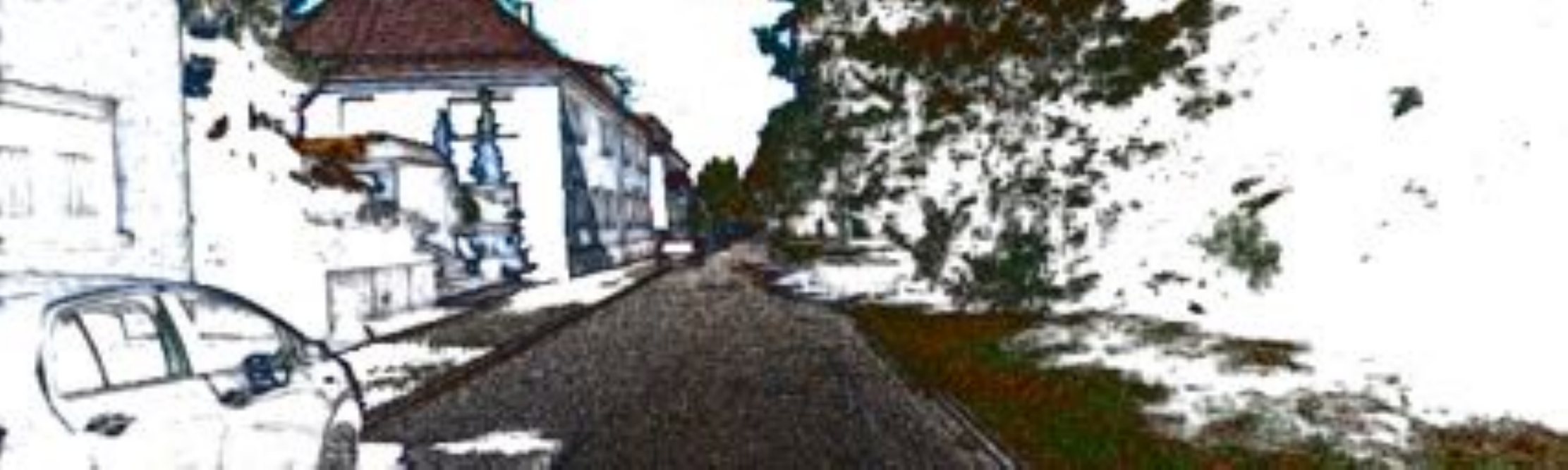}& 
    \includegraphics[]{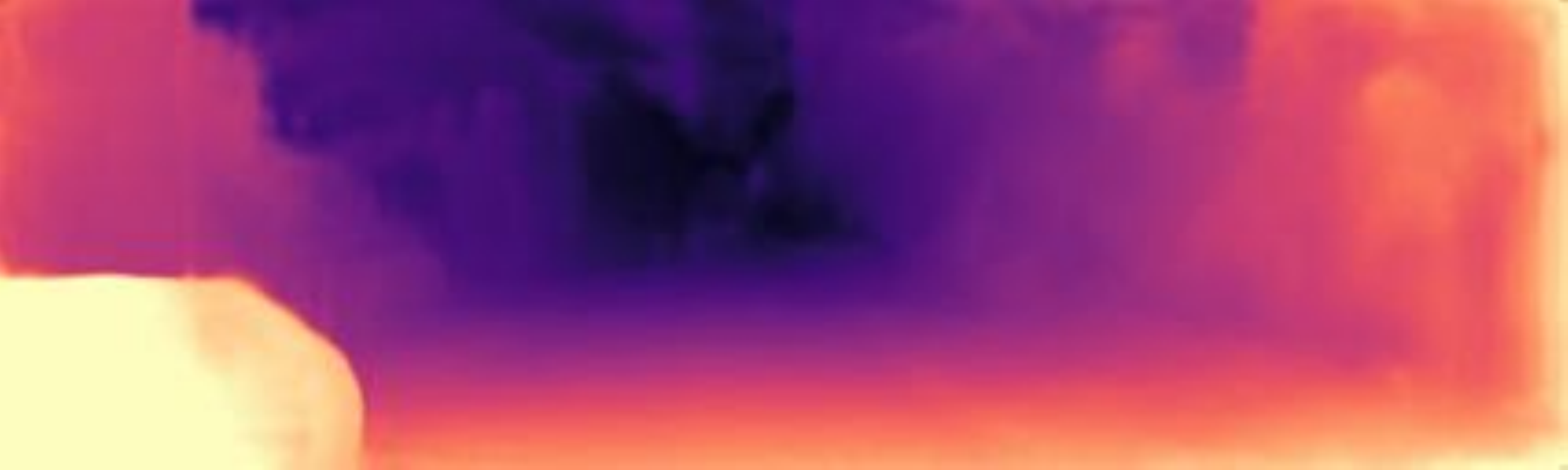}& 
    \includegraphics[]{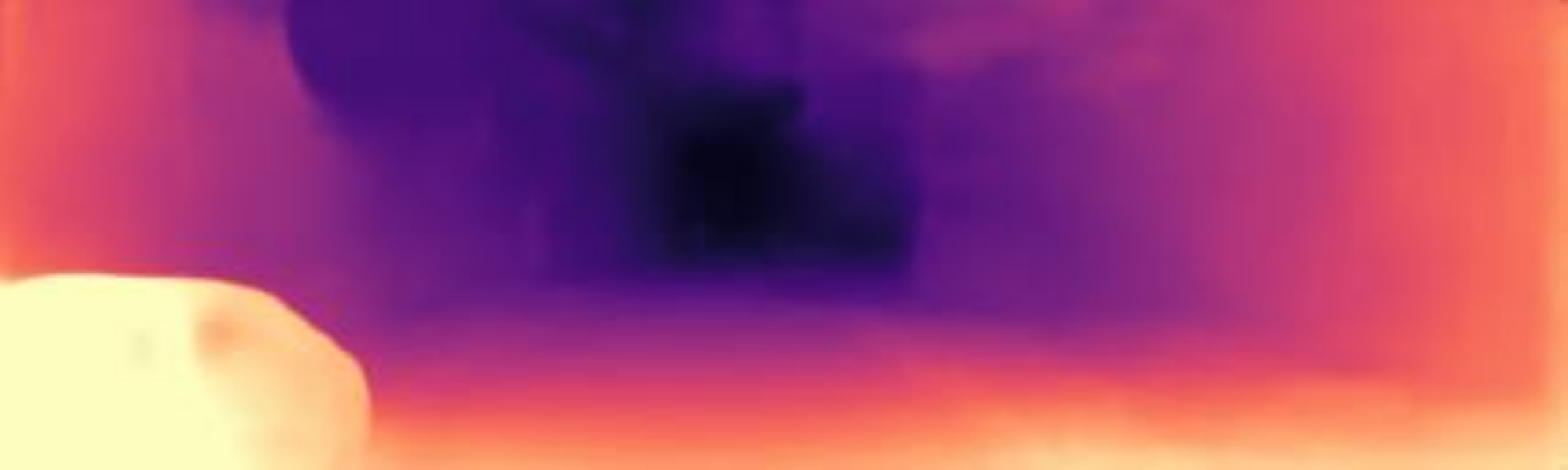}& 
    \includegraphics[]{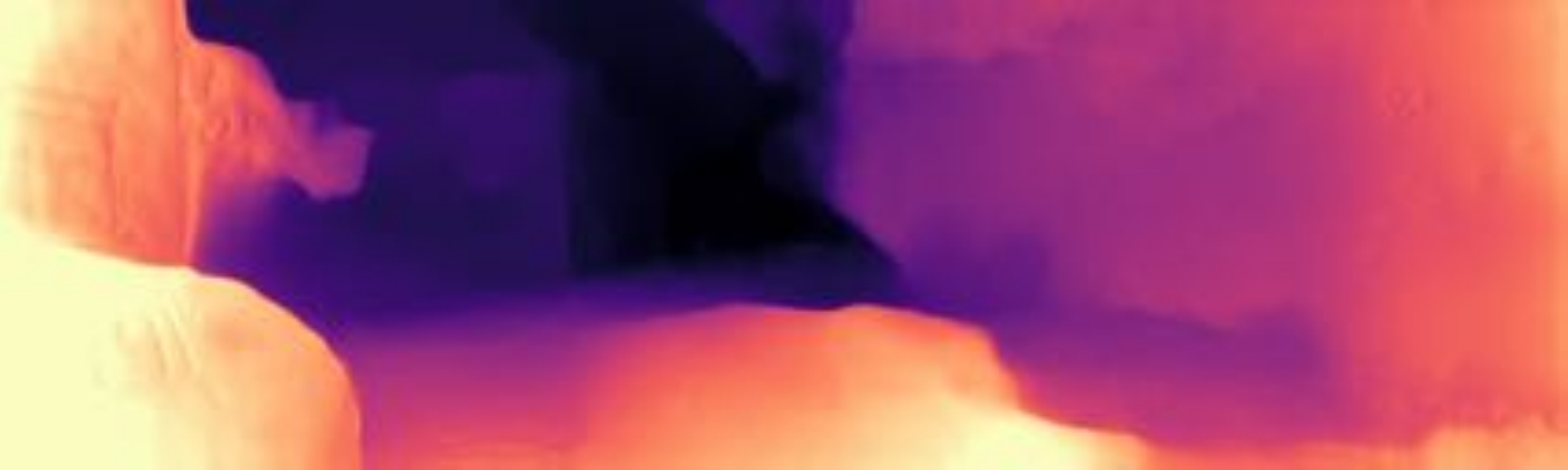}& 
    \includegraphics[]{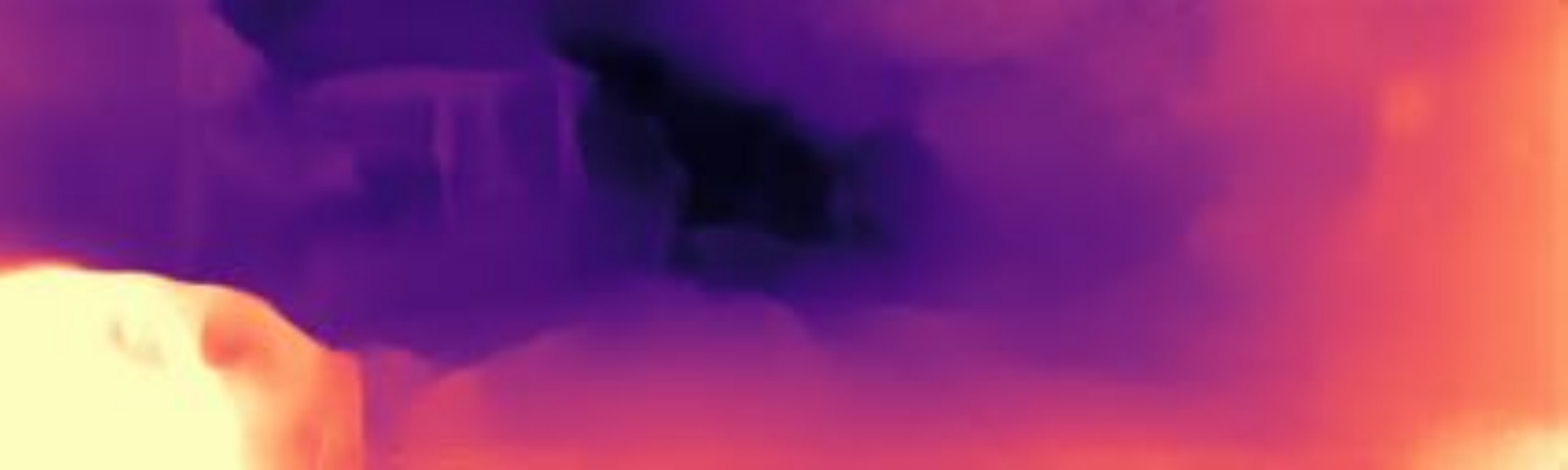}& 
    \includegraphics[]{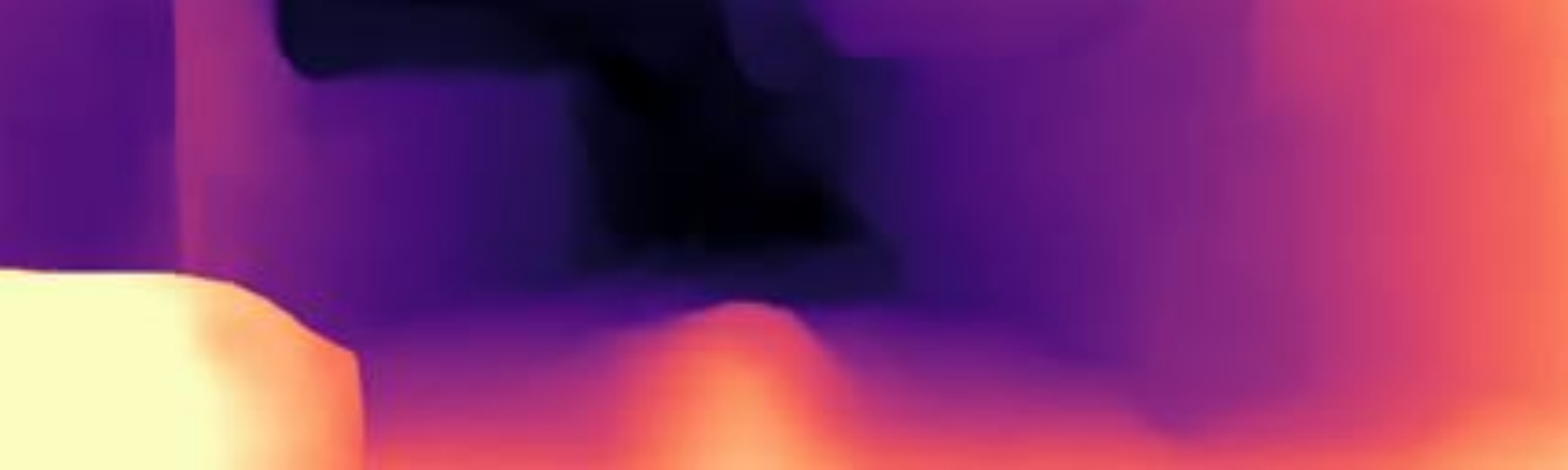}\\    
    
     \fontsize{80}{80} \selectfont Input images & 
     \fontsize{80}{80} \selectfont Ours-Hybrid &
     \fontsize{80}{80} \selectfont Ours-ViT &
     \fontsize{80}{80} \selectfont Monodepth2 &
     \fontsize{80}{80} \selectfont PackNet-SfM & 
     \fontsize{80}{80} \selectfont R-MSFM6
    \end{tabular}}
    \caption{\textbf{Depth map results on texture-shifted datasets and KITTI raw dataset.} We test our hybrid/ViT models and the competitive models trained on KITTI using original image, watercolor, and pencil-sketch (Top to Bottom). Note that the Ours-Hybrid is equivalent to MonoFormer.}
\label{figure_result_texture_apdx}
\end{figure*}

\bibliography{aaai23}

\end{document}